\PassOptionsToPackage{dvipsnames}{xcolor}
\documentclass[singlecolumn]{NTUAI4XPaperEN}
\usepackage[english]{babel}
\usepackage{tabularx}
\usepackage{placeins}
\usepackage{adjustbox}
\usepackage{mathtools}
\usepackage{listings}

\usepackage{longtable}
\usepackage{booktabs}
\usepackage{graphicx}
\usepackage{microtype}
\setthemecolor{19365b}

\settitlepreskip{-10ex}
\settitlelogogap{3em}

\titlelogocontent{%
	\begin{minipage}{\textwidth}
		\begin{minipage}[c]{0.35\textwidth}
			\raggedright
			\includegraphics[height=1cm]{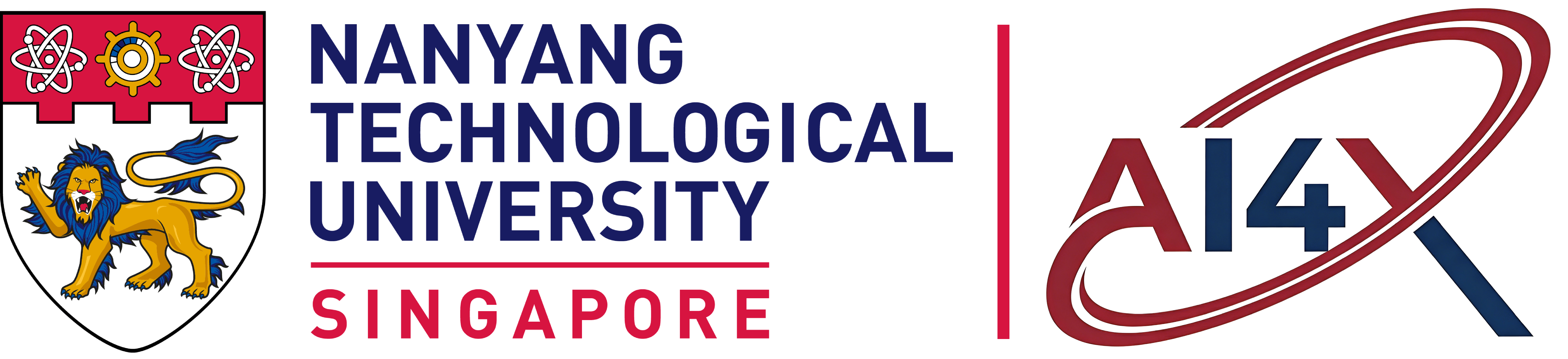}
		\end{minipage}%
		\\[0.4em]
		\textcolor{academicblue}{\rule{\textwidth}{0.6pt}}%
	\end{minipage}%
}

\showlogointitle

\hypersetup{
    pdftitle={Do You Miss the Nights We Spent Crafting Figures? Rethinking Teaser and Framework Figures in the Age of Generative AI},
    pdfauthor={Xiaoqi Zhao, Youwei Pang, Kailai Zhou},
}
\setthemecolor{19365b}
\newcolumntype{Y}{>{\raggedright\arraybackslash}X}
\newcommand{\model}{\texttt{gpt-6-astra}}

\newtcolorbox{scopebox}[1][]{enhanced,breakable,colback=academicblue!3,colframe=academicblue!65,boxrule=0.7pt,arc=2mm,left=10pt,right=10pt,top=8pt,bottom=8pt,title={Our scope: assistance without surrendering design},fonttitle=\bfseries,#1}
\title{Do You Miss the Nights We Spent Crafting Figures?
Rethinking Teaser and Framework Figures in the Age of Generative AI}
\shorttitle{Rethinking Teaser and Framework Figures in the Age of Generative AI}

\author{Xiaoqi Zhao, Youwei Pang, Kailai Zhou}
\affiliation{Nanyang Technological University, Singapore}
\abstract{%
Scientific figures give research ideas a visual form through choices of abstraction, composition, and emphasis. Generative AI offers new ways to realize these choices, raising questions about explanatory clarity and purposeful visual expression. We examine these questions through a visual retrospective of 371 motivation and framework figures from 200 papers. The corpus comprises a quota-based main sample of 160 papers spanning four temporal periods and a separate purposive style-contrast sample of 40 papers. A single multimodal judge evaluates six dimensions of communication and design, with three repeated judgments per figure. We attempt source-constrained revisions of all 371 figures, of which 369 pass the recorded fidelity checks. Original-figure ratings show no uniform decline across periods. Revisions increase mean clarity ratings by 0.428 points and decrease aesthetic coherence ratings by 0.111 points on a five-point scale. Gains in the six-dimension average are larger for main-sample motivation figures than for frameworks. Case analyses show that added explanation can clarify comparisons while leaving typographic hierarchy and visual economy unresolved. These model-based findings support researcher-led design, with AI assisting exploration and authors retaining responsibility for explanatory purpose, composition, and scientific relationships. The accompanying atlas and evaluation records are available at \url{https://ntu-ai4x.github.io/Rethinking_Figures/index_en.html}.
}
\begin{document}
\maketitle

\setcounter{tocdepth}{2}
\begingroup
\setlength{\parskip}{0pt}
\makeatletter
\renewcommand*{\l@section}[2]{%
  \@dottedtocline{1}{0em}{1.5em}{\textbf{#1}}{\textbf{#2}}%
}
\renewcommand*{\l@subsection}{%
  \@dottedtocline{2}{1.5em}{2.3em}%
}
\section*{\contentsname}
\vspace{-4pt}
\fontsize{9.5pt}{12pt}\selectfont
\@starttoc{toc}
\makeatother
\endgroup

\section{Introduction}
\label{sec:introduction}

A familiar scene unfolds before many paper deadlines. The manuscript is almost ready, yet Figure~1 remains open on the screen. An arrow is rerouted, a color is given a clearer meaning, and a generic box gives way to an example that reveals the central idea. Such revisions help shape the scientific argument. They influence what a reader notices, which relationships become intelligible, and how a contribution is remembered. The nights recalled in our title represent this effort to find a visual form that belongs to a particular idea.

\begin{figure*}[t]
\centering
\begin{minipage}[c]{0.62\linewidth}
\begin{subfigure}{\linewidth}
\centering
\includegraphics[width=\linewidth,height=7.0cm,keepaspectratio]{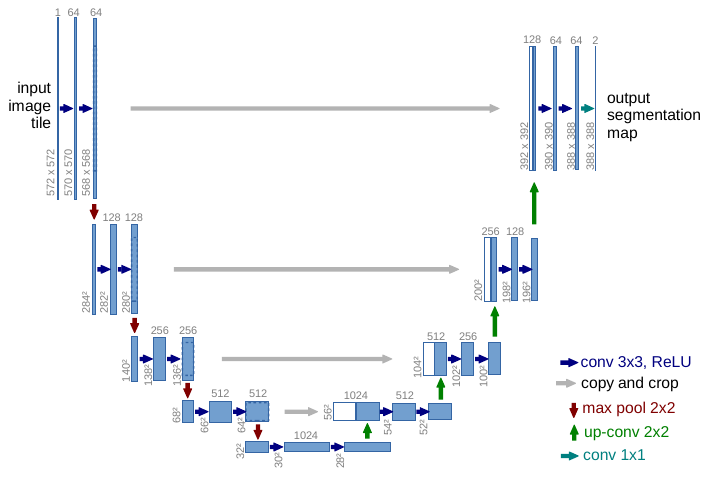}
\caption{U-Net (2015).}
\end{subfigure}
\par\vspace{12pt}
\begin{subfigure}{\linewidth}
\centering
\includegraphics[width=\linewidth,height=6.0cm,keepaspectratio]{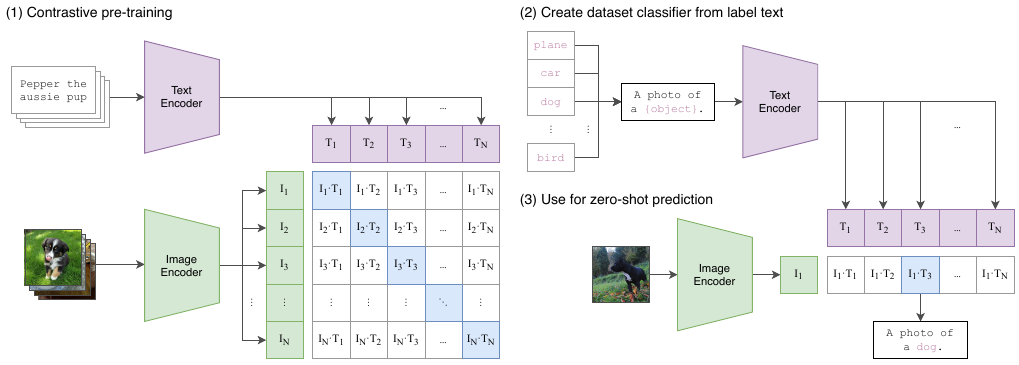}
\caption{CLIP (2021).}
\end{subfigure}
\end{minipage}\hfill
\begin{subfigure}[c]{0.34\linewidth}
\centering
\includegraphics[width=\linewidth,height=12.5cm,keepaspectratio]{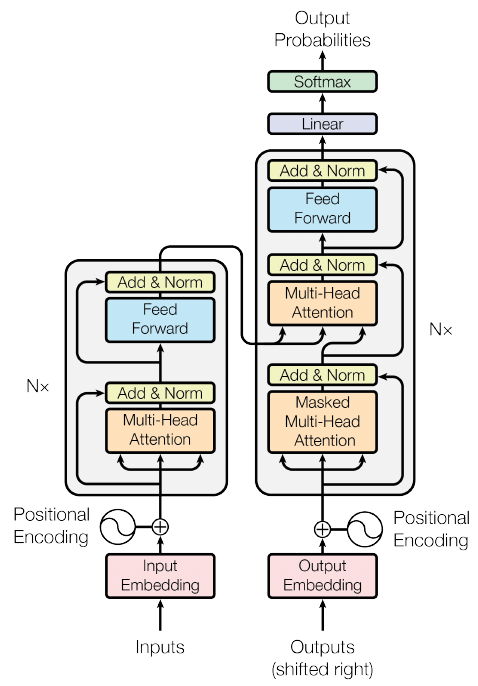}
\caption{Transformer (2017).}
\end{subfigure}
\caption{Distinctive visual designs in U-Net, CLIP, and the Transformer, reproduced from Figure~1 of the respective papers~\citep{ronneberger2015unet,radford2021clip,vaswani2017attention}. These illustrative examples are outside the analytical corpus.}
\label{fig:classics}
\end{figure*}

Teaser and framework figures serve complementary roles in this process. A teaser may introduce a problem, expose a limitation, or communicate an insight, while a framework figure presents the organization of a method. Our study focuses on motivation-oriented teasers and overall frameworks. Three familiar framework diagrams illustrate how composition can expose scientific structure. U-Net uses a U-shaped arrangement to express contraction, expansion, and cross-scale transfer. The Transformer organizes its principal operations around the relationship between an encoder and a decoder, while CLIP places contrastive pretraining alongside zero-shot classification~\citep{ronneberger2015unet,vaswani2017attention,radford2021clip}. Fig.~\ref{fig:classics} presents these deliberately selected landmarks as qualitative reference points outside the analytical corpus. Their value here lies in the correspondence between visual arrangement and scientific structure.

Generative models broaden the ways in which researchers develop such compositions. They can propose layouts, create illustrative elements, and support iterative refinement. Recent systems also target scientific diagrams explicitly~\citep{zhu2026paperbanana}. This assistance makes alternative designs easier to explore, but authors must still decide which arrangement best explains the intended claim. An attractive draft may obscure a dependency, repeat information, or give a secondary detail undue prominence. Additional labels may resolve ambiguity while increasing reading effort. Assessing AI assistance therefore requires examining how revisions affect communication, visual economy, and scientific fidelity.

Visual appearance alone cannot establish how a figure was produced or how effectively it communicates. Likewise, an overall quality score can conceal improvements in clarity that coincide with weaker organization or aesthetic coherence. A useful assessment must distinguish production provenance from visible design properties and examine communication through multiple dimensions. This motivates an analysis that places existing figures in their historical context and follows changes within individual figures.

Within this multidimensional assessment, we give particular attention to \emph{purposeful visual inventiveness} and \emph{problem-specific design}. A revealing comparison, a meaningful spatial correspondence, or a carefully organized example can make an unfamiliar concept accessible. These qualities can appear in a restrained block diagram or a detailed illustration, depending on the message and audience~\citep{rougier2014figures}. They give a concrete interpretation to the ``human touch'' that motivates this study without treating appearance as evidence of authorship or effort. A polished figure can still leave its central claim implicit, and a detailed diagram can identify its components without making their interactions traceable.

Evidence from other creative activities further highlights the distinction between individual evaluations and collective diversity. Doshi and Hauser report improved individual evaluations of AI-assisted stories alongside greater similarity among outputs~\citep{doshi2024creativity}. Ashkinaze et al. find that extensive exposure to AI-generated ideas increases collective diversity without improving individual creativity~\citep{ashkinaze2025ideas}. These task-dependent findings motivate distinguishing communicative quality, expressive design, and stylistic diversity. Our evaluation addresses the first two, while claims about community-wide stylistic convergence require separate evidence.

We combine a retrospective analysis of existing figures with a source-constrained revision study. The corpus contains 371 figures from 200 papers, including a quota-based main sample of 160 papers spanning four periods and a separate purposive style-contrast sample of 40 papers. A fixed protocol uses a single multimodal judge to assess six dimensions of communication and design (Section~\ref{sec:scope}). Under this protocol, original-figure ratings show no uniform deterioration across periods. Revisions receive higher mean clarity ratings but lower aesthetic-coherence ratings, with larger increases in the six-dimension average for main-sample motivation figures than for frameworks. Case analyses connect these outcomes to specific design decisions, particularly the balance between making explanations explicit and preserving a coherent composition.

This study makes three contributions:
\begin{enumerate}
\item We construct a traceable corpus and illustrated atlas of 200 papers and 371 figures, retaining source and visual-version evidence and distinguishing the main sample from the purposive contrast sample.
\item We evaluate original figures and 369 study-generated revisions that pass the recorded fidelity checks across six dimensions, preserving repeated judgments and reporting both dimension-level tradeoffs and fidelity failures.
\item We develop practical guidance for researcher-led figure design, connecting observed tradeoffs to decisions about visual narrative, typographic hierarchy, information selection, and scientific fidelity.
\end{enumerate}

\FloatBarrier
\section{Study Overview and Research Questions}
\label{sec:scope}

\begin{scopebox}
\textbf{Design principle.} Researchers should determine what a figure communicates and how its composition, colors, symbols, and connections express the scientific idea.

\textbf{Study approach.} We combine a retrospective analysis of existing figures with paired assessment of source-constrained revisions, documenting both design tradeoffs and fidelity failures.

\textbf{Evaluation focus.} Six dimensions characterize visual communication and design under a fixed multimodal judging protocol.
\end{scopebox}

The study distinguishes visual appearance, production provenance, and communication quality. Appearance can be inspected directly, while provenance requires evidence about how a figure was produced. We assess communication and design through six dimensions: Communicative Clarity, Organization and Readability, Graphical Consistency, Aesthetic Coherence, Purposeful Visual Inventiveness, and Design Specificity. The latter two respectively assess whether a visual device contributes to explanation and whether the depicted elements and relationships are specific to the research problem. These dimensions are examined separately to preserve tradeoffs that an overall score may conceal.

Two research questions guide the analysis:
\begin{enumerate}[label=RQ\arabic*.,leftmargin=3.8em]
\item \textbf{How do communication and design ratings vary across periods and figure roles?} The retrospective analysis compares motivation and framework figures across the four periods within the main sample. The purposive contrast sample supports closer examination of selected visual conventions and is analyzed separately.
\item \textbf{How do these ratings change following source-constrained AI-assisted revision?} Paired comparisons identify gains, losses, and unchanged ratings across dimensions. Revisions that fail fidelity checks are retained as workflow outcomes without receiving quality scores.
\end{enumerate}

Both questions are descriptive and concern the documented corpus and configured evaluator. The fixed protocol provides a standardized assessment procedure, while the correspondence between its ratings and human understanding or preference remains to be established~\citep{chen2024mllmjudge}. These analyses inform our argument for researcher-led design and identify questions for subsequent reader studies.

\section{Dataset Construction}
\label{sec:dataset}

We constructed a traceable corpus of 200 papers and 371 motivation and framework figures through two complementary collection paths. The quota-based main sample contains 160 papers and 320 figures for comparisons across temporal stages and research topics. A separate purposive style-contrast sample contains 40 papers and 51 figures for examining selected visual conventions. Papers are the units of inclusion and deduplication, while figures are the units of assessment. Eligible material was publicly available by 20 October 2026, and collection was finalized on 29 October 2026. The sampling procedure was refined through pilot collection and checks of candidate availability and was not preregistered.

\subsection{Sampling Design: Stages, Topics, and Sources}
\label{sec:strata}
\label{sec:eligibility}

\begin{figure*}[t]
\centering
\includegraphics[width=\textwidth]{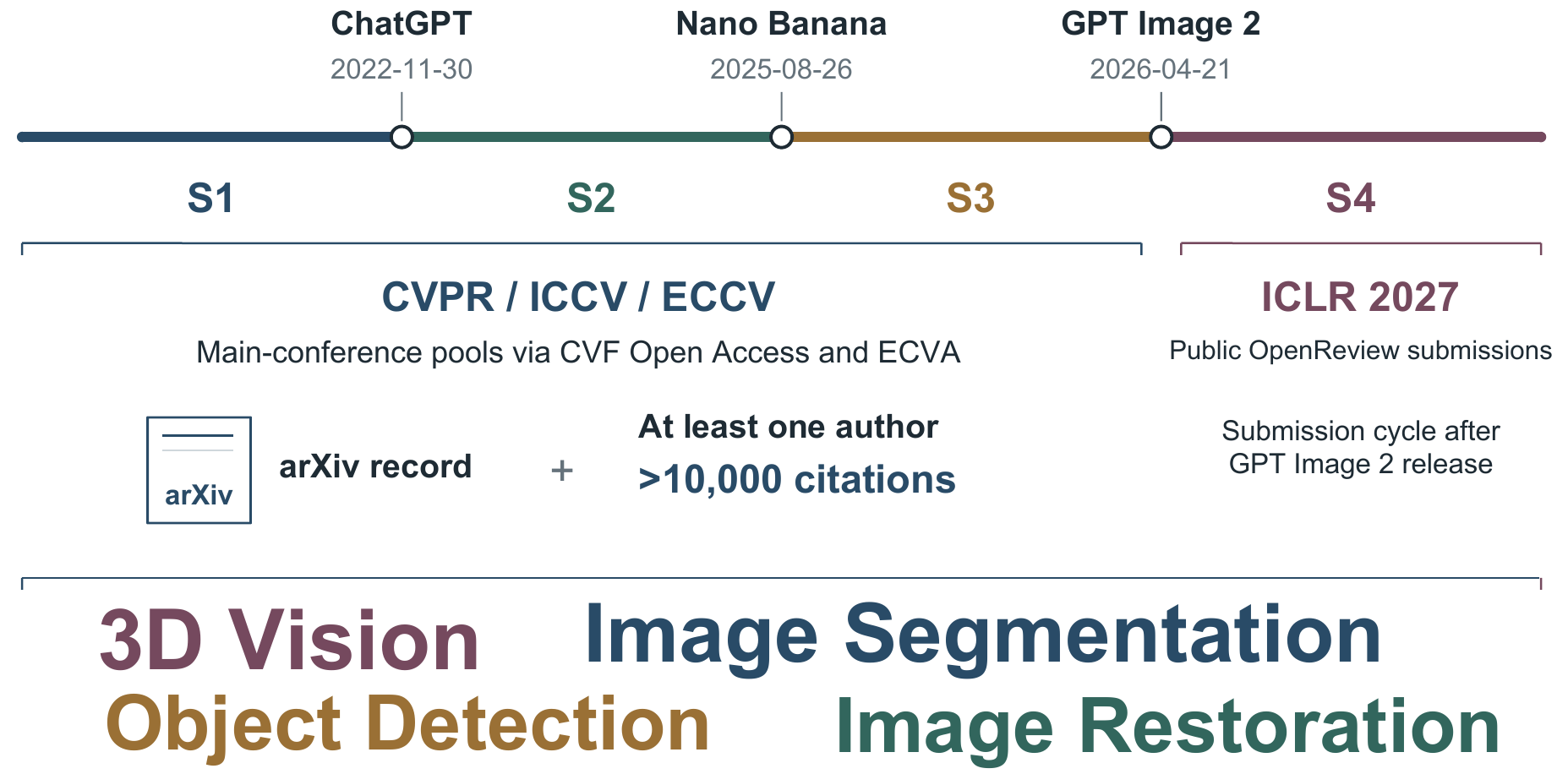}
\caption{Main-sample design across temporal stages, candidate source pools, and recurring research topics. Source labels identify the search scope, and release milestones indicate tool availability. Timeline lengths are schematic.}
\label{fig:stages}
\end{figure*}

The main sample crosses four temporal stages with four research topics that recur throughout the study period (Fig.~\ref{fig:stages}). Object detection primarily concerns 2D objects, while image segmentation includes semantic and instance segmentation. Image restoration covers denoising, deblurring, and super-resolution, and 3D vision primarily concerns reconstruction. Each of the 16 combinations of stage and topic contains 10 papers, with each paper contributing one motivation (M) figure and one distinct framework (F) figure. These quotas maintain a common topical scope while accommodating changes in methods and subtopics within each research area.

The eligibility window extends from 1 January 2016 to 20 October 2026. The public releases of ChatGPT on 30 November 2022, Gemini 2.5 Flash Image (Nano Banana) on 26 August 2025, and ChatGPT Images 2.0 (GPT Image 2) on 21 April 2026 mark the beginnings of S2, S3, and S4, respectively~\citep{openai2022chatgpt,google2025flashimage,openai2026images}. Each preceding stage ends on the previous day. Stage assignment follows public evidence for the sampled visual version. These milestones indicate tool availability and do not establish use by the original authors.

For S1--S3, discovery began with CVPR, ICCV, and ECCV main-conference directories on CVF Open Access and ECVA, supplemented by arXiv version records. Eligibility required an arXiv record and at least one author with more than 10,000 cumulative citations on a verifiable public scholarly profile. Citation snapshots were collected on 21--28 October 2026, after the material cutoff. This criterion reflects author visibility at retrieval and does not validate paper quality.

For S4, discovery used public ICLR 2027 submissions on OpenReview. The September 2026 submission cycle followed the release of GPT Image 2~\citep{iclr2027dates}, and the submissions were under review when collected. We initially applied the same arXiv and author-citation requirements. When reviewed candidates satisfying both requirements did not fill a topic quota, either or both requirements were waived. Papers admitted under the original and relaxed criteria remain separately identified.

\subsection{Collection and Figure Verification}
\label{sec:collection}
\label{sec:selection}
\label{sec:roles-dates}

\begin{figure*}[t]
\centering
\includegraphics[width=\textwidth]{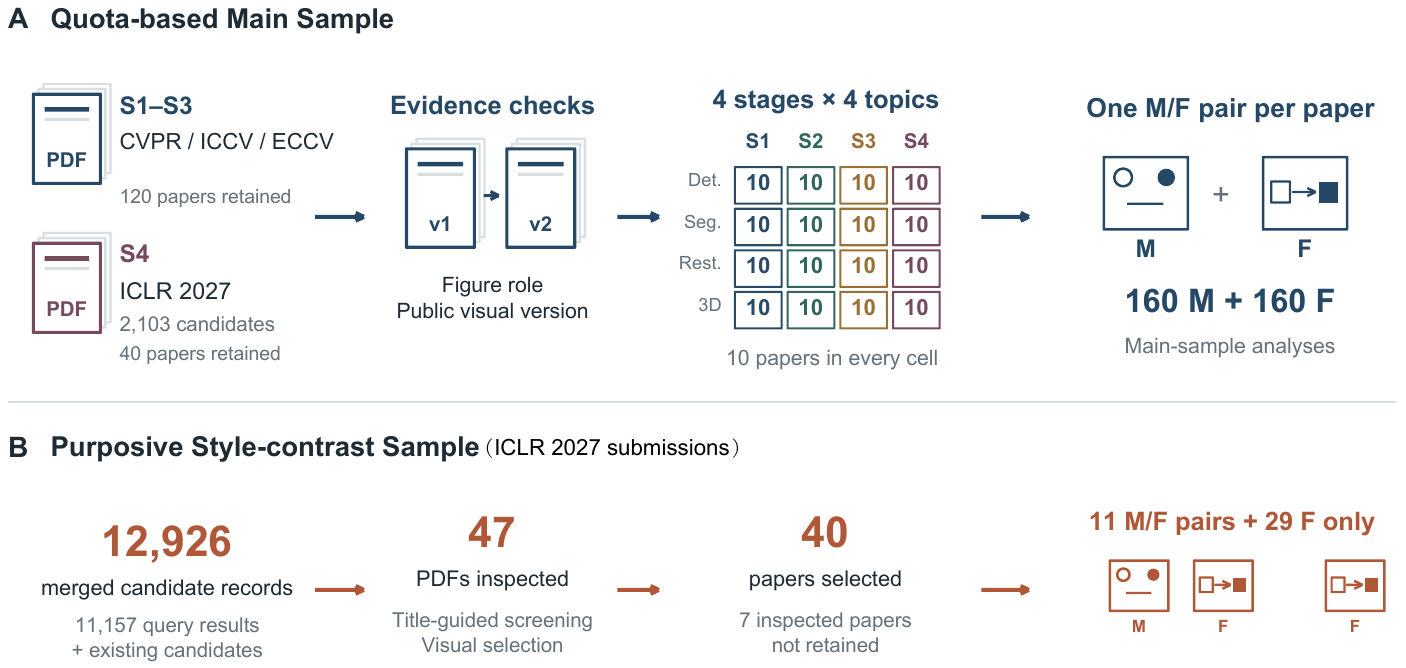}
\caption{Two collection paths produce 200 papers and 371 figures. \textbf{A:} Role and version checks support quota-based selection of paired motivation (M) and framework (F) figures. \textbf{B:} Purposive selection retains eligible M and/or F figures. Candidate pools overlap. In the contrast pool, 47 PDFs underwent visual inspection.}
\label{fig:construction}
\end{figure*}

Fig.~\ref{fig:construction} summarizes the two collection paths. Main-sample selection used figure-role and version checks to fill the temporal and topical quotas without considering aesthetic appeal or perceived AI-like style. The contrast sample used title-guided inspection and purposive visual selection without requiring paired M and F figures. The retained paper sets are disjoint, although their candidate pools overlap and their retrieval counts cannot be summed.

Main-sample searches used topic-specific terms. Detection queries included \texttt{object detection} and \texttt{object detector}, while segmentation queries included \texttt{semantic segmentation} and \texttt{instance segmentation}. Restoration queries covered \texttt{image restoration}, \texttt{denoising}, \texttt{deblurring}, and \texttt{super-resolution}. Queries for 3D vision included \texttt{3D reconstruction}, \texttt{multi-view reconstruction}, and \texttt{surface reconstruction}. Retrieval snapshots preserve the query scope and pagination. The 11 S4 queries yielded 2,103 distinct candidate records within this search scope and do not constitute a census of the conference. The cumulative S1--S3 retrieval history has not been reconstructed into a deduplicated screening funnel, so aggregate retrieval and exclusion counts are unavailable for those stages.

Figure roles were assessed from the complete image, its caption, and relevant body text. An M figure introduces a problem through limitations, failure cases, conceptual contrasts, or application needs. An F figure depicts an overall architecture, major module relationships, or a principal training or inference workflow. Performance-only charts, ordinary result montages, and isolated module diagrams do not qualify solely because they contain relevant visual elements. When several figures qualified, we prioritized the first problem-introducing figure for M and the explicitly designated overall framework for F. Remaining ties were resolved by figure number.

For each main-sample figure, we sought the earliest verifiable public occurrence of the sampled visual version. Composition, elements, colors, and connections were compared across available versions, with posting and submission records establishing dates or date ranges. Changes to surrounding text or experiments do not alter the visual-version date when the sampled figure remains unchanged. A later redraw is dated from its own verifiable public appearance. Both M and F figures must qualify for the paper's assigned stage, and an earlier arXiv figure may therefore be retained instead of a later conference version. A figure was not assigned to a stage when a possible earlier-stage occurrence remained unresolved. These records establish public availability rather than creation dates. Historical tracing is more limited for S4 papers without matched arXiv records, and the manifest preserves the available date precision.

The contrast path searched additional ICLR 2027 submissions without topic quotas, arXiv matching, or author-citation requirements. The four queries were \texttt{multi-agent}, \texttt{agentic}, \texttt{multimodal}, and \texttt{benchmark}. They returned 11,157 distinct records, and merging these results with existing candidates yielded a pool of 12,926 records. Title relevance guided inspection of 47 PDFs, of which 40 were retained and seven were not. Selected figures prominently combined conventions such as nested boxes, labeled panels, explanatory text, icons, and pastel colors. Selection required neither a numerical threshold nor the presence of every feature. Candidates that were not inspected are not counted as exclusions following full-text assessment.

Eleven contrast papers contribute both M and F figures, and 29 contribute F figures only. The legacy label \texttt{S4-AI-Style} identifies this collection path. It does not establish AI authorship or an earliest visual occurrence within S4. Fixed submission PDFs and public-date evidence are archived, but earliest matching visual versions were not systematically traced across arXiv and project histories. The contrast sample is therefore analyzed separately from temporal stage means and cannot estimate the prevalence of its selected conventions.

\label{sec:archive}
Complete figure regions were cropped from fixed PDFs, retaining all panels, labels, legends, arrows, and symbols. The cropped PDF assets preserve the original PDF objects, including embedded raster content. PNG versions were rendered at 300 dpi with the original aspect ratio, without redrawing, recoloring, text replacement, or generative enhancement. Each export was visually inspected and linked to its caption, source page, crop coordinates, URL, version evidence, and file hash. Anonymous evaluation copies omit external source labels while retaining content within the figure. Drawing-tool disclosures are recorded separately, and production provenance remains unknown in the absence of direct evidence.

The archive preserves paper and figure manifests, exclusions, retrieval logs, fixed sources, and blind-ID mappings. The accompanying methods project provides aggregate plotting data, source-file hashes, editable vector figures, and figure-generation code. Corpus counts derive from the main-sample manifests of 160 papers and 320 figures and the contrast-sample manifests of 40 papers and 51 figures. Collection, role assessment, version comparison, and cropping used OpenAI's \model\ under researcher-specified rules, in sessions separate from quality assessment. Independent expert verification and inter-rater reliability were not established.

\subsection{Corpus Characteristics and Comparison Scope}
\label{sec:composition}

\begin{figure*}[t]
\centering
\includegraphics[width=\textwidth]{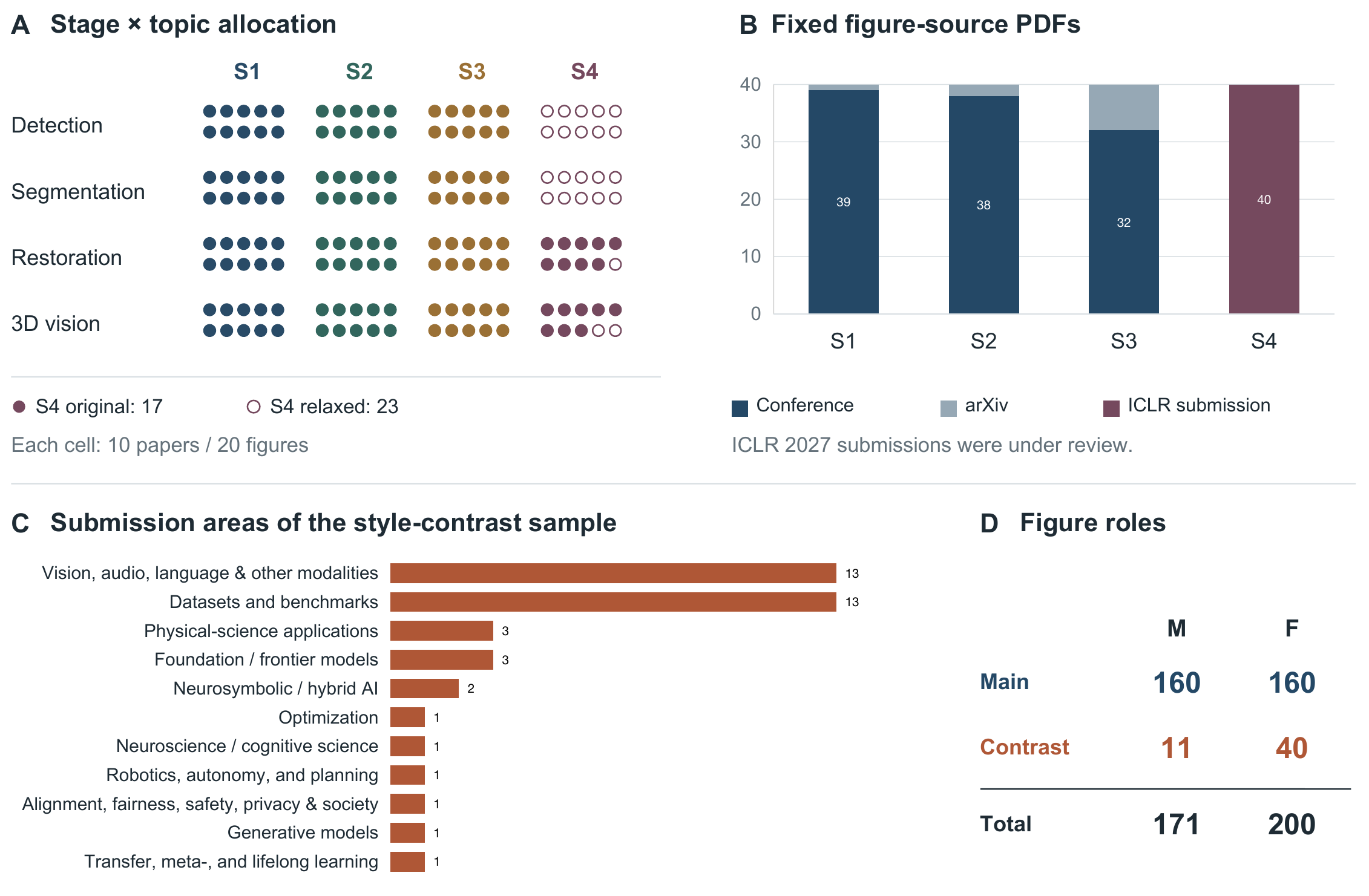}
\caption{Corpus composition. \textbf{A:} Main-sample allocation and S4 eligibility tiers. \textbf{B:} Fixed PDF sources, distinct from publication venues. \textbf{C:} OpenReview submission areas of the contrast sample, with shortened metadata labels. \textbf{D:} Motivation (M) and framework (F) figure counts. The contrast sample is excluded from temporal stage averages.}
\label{fig:distribution}
\end{figure*}

Fig.~\ref{fig:distribution} summarizes the realized corpus and the differences that remain despite balanced quotas. Each main-sample stage contains 40 papers and 80 figures, but S4 includes two eligibility tiers (panel~A). Its 17 original-criteria papers comprise nine restoration papers and eight 3D papers. The 23 relaxed-criteria papers comprise all 10 detection papers, all 10 segmentation papers, one restoration paper, and two 3D papers.

Source composition also varies across stages (panel~B). At the paper level, 39, 38, and 32 papers in S1, S2, and S3, respectively, use conference PDFs as their fixed figure sources. The corresponding arXiv-source counts are one, two, and eight, yielding 109 conference-source papers and 11 arXiv-source papers overall. The retained conference labels are CVPR and ECCV. No ICCV paper was retained despite its inclusion in the search scope. Three papers with arXiv figure sources also have a CVPR venue label, demonstrating that the fixed source version does not determine publication status. All S4 figure sources are fixed ICLR 2027 submission PDFs.

The contrast sample is concentrated in two OpenReview \texttt{primary\_area} categories (panel~C). Modality applications and datasets and benchmarks each contribute 13 papers, with the remaining 14 papers distributed across nine other areas. These metadata categories differ from the four task families used in the main sample. Panel~D reports 160 M and 160 F figures in the main sample and 11 M and 40 F figures in the contrast sample, yielding 171 M and 200 F figures overall.

Temporal coverage is also uneven (Table~\ref{tab:years}). Stage windows differ in duration, and sampling was not balanced by year. In S1, 52 of the 80 figures date to 2022, and none date to 2016 or 2017. The start of the eligibility window therefore precedes the earliest retained figure versions.

\begin{table*}[t]
\centering
\small
\setlength{\tabcolsep}{4pt}
\renewcommand{\arraystretch}{1.12}
\caption{Temporal distribution of verified figure versions in the main sample. Entries count figures, and dashes denote zero.}
\label{tab:years}
\begin{tabular*}{\textwidth}{@{\extracolsep{\fill}}lrrrrrrrrrr@{}}
\toprule
& \multicolumn{9}{c}{Year of verified figure version} & \\
\cmidrule(lr){2-10}
Stage & 2018 & 2019 & 2020 & 2021 & 2022 & 2023 & 2024 & 2025 & 2026 & Total \\
\midrule
S1 & 3 & 3 & 1 & 21 & 52 & -- & -- & -- & -- & 80 \\
S2 & -- & -- & -- & -- & -- & 3 & 35 & 42 & -- & 80 \\
S3 & -- & -- & -- & -- & -- & -- & -- & 20 & 60 & 80 \\
S4 & -- & -- & -- & -- & -- & -- & -- & -- & 80 & 80 \\
\midrule
Total & 3 & 3 & 1 & 21 & 52 & 3 & 35 & 62 & 140 & 320 \\
\bottomrule
\end{tabular*}
\end{table*}

These characteristics delimit the comparisons supported by the corpus. Neither collection path produces a probability sample. Equal quotas do not equalize source populations, eligibility conditions, review status, or the distribution of years within each stage. The contrast sample additionally reflects purposive selection by appearance and a different topical scope. Temporal differences therefore cannot isolate the effects of AI assistance, and figures drawn from the same paper must be treated as clustered observations.
\FloatBarrier
\section{Multimodal Evaluation and Source-Constrained Revision}
\label{sec:evaluation}

Scientific figure revision seeks to improve visual communication while preserving scientific meaning. Our protocol evaluates these objectives separately through anonymous image-based quality assessment and source-informed fidelity review. Revisions that pass fidelity review are compared with their originals using the same quality criteria.

\subsection{Assessment Criteria}

The six dimensions in Table~\ref{tab:dimensions} assess the accessibility of an explanation, the coherence of its visual organization, and the specificity and inventiveness of its expression. Scores range from 1 to 5, with explicit anchors at 1, 3, and 5 and intermediate scores of 2 and 4. The definitions and anchors remain fixed across original and revised figures. These study-defined criteria operationalize the design concerns introduced in Section~\ref{sec:scope} and do not constitute a validated psychometric instrument. Because each dimension uses distinct criteria, differences between their means describe a rubric profile rather than calibrated differences on a common quality scale.

\begin{table*}[t]
\centering
\small
\setlength{\tabcolsep}{5pt}
\renewcommand{\arraystretch}{1.15}
\caption{Operational dimensions and compact scoring anchors. Scores of 2 and 4 indicate performance between adjacent anchors. The same frozen rubric is used for original and revised figures.}
\label{tab:dimensions}
\begin{tabularx}{\textwidth}{@{}p{0.17\textwidth}XXX@{}}
\toprule
Dimension & Score 1 & Score 3 & Score 5 \\
\midrule
Communicative Clarity &
The central message is unclear, and key objects or relationships require substantial inference. &
The main problem or process is identifiable, but important relationships or emphasis remain unclear. &
The intended problem or contrast, or the main inputs, operations, relationships, and outputs, are readily understood. \\
\addlinespace
Organization \& Readability &
Confusing hierarchy, crowding, or illegible key text obstructs the reading sequence. &
Most information can be followed, but grouping, density, hierarchy, or text requires noticeable effort. &
Reading order and hierarchy are clear, grouping manages complexity, and key labels are legible. \\
\addlinespace
Graphical Consistency &
Conflicting or unexplained visual encodings make important relationships ambiguous. &
Encodings are mostly consistent, with some ambiguous labels, directions, connections, or repeated elements. &
Visual encodings are consistent and traceable, with necessary labels or legends clarifying relationships. \\
\addlinespace
Aesthetic Coherence &
Clashing typography, colors, spacing, or styles compete with the message. &
The composition is broadly coherent but has uneven alignment, spacing, emphasis, or balance. &
Color, typography, spacing, alignment, and balance jointly support the information. \\
\addlinespace
Purposeful Visual Inventiveness &
Generic arrangement or ornament contributes little problem-specific understanding. &
A deliberate comparison, arrangement, example, or explanatory device helps, but much remains generic. &
A well-chosen explanatory device provides a particularly effective account of the concept. \\
\addlinespace
Design Specificity &
Replacing the labels would allow the design to represent many unrelated problems. &
Some elements are tailored to the problem, while much of the design remains interchangeable. &
The elements and their arrangement express distinctive problem-specific objects, relationships, or operations. \\
\bottomrule
\end{tabularx}
\end{table*}

The rubric evaluates visual devices by their explanatory function. Conventional boxes may receive high scores, while illustrations, icons, gradients, and hand-drawn marks receive no automatic preference. Framework overviews are assessed at their intended level of abstraction. The judge is instructed to avoid inferences about authorship, effort, historical period, software, or paper reputation. Because external captions and manuscript text are withheld, clarity and specificity concern the information conveyed by the supplied figure. Correspondence with source evidence is assessed separately during fidelity review.

Responses follow a structured JSON format. For each dimension, the judge provides an integer score or a null value, a concrete visual observation, and an assessment limitation. A null value requires an explicit reason. Image-level fields record readability and general limitations. Retaining this evidence alongside the scores makes the basis of each judgment available for inspection.

\subsection{Anonymous Repeated Judgments}

Each request contains the frozen rubric, the figure's motivation (M) or framework (F) role, and three image views. These comprise the complete anonymized figure and two deterministic crops along its longer axis. The crops jointly cover the figure with a 10\% overlap, with boundaries rounded outward to integer pixels. Local preparation preserves the pixels of the prepared rating image without resizing or enhancement. External captions, paper titles, author identities, venues, years, stage and sample labels, and previous ratings are withheld. Text and imagery within the figure remain intact, so familiar figures may still be recognizable. Any additional image processing within the model service is unobserved.

Each original figure receives three repeated judgments, yielding 1,113 validated responses for 371 figures. Failed requests are archived and retried within the same slot defined by figure and repetition. Only one validated response per slot enters the analysis. Visual observations are preserved in the judge's original English wording. A procedural pilot used eight figures spanning all combinations of stage and role and collectively covering the four main topics. The pilot assessed instruction compliance and the grounding of explanations in visible evidence. Its responses are excluded from the primary analysis.

All reported quality judgments were obtained using OpenAI's \model\ through Codex (version \texttt{0.163.0-alpha.10.2}). Original figures were evaluated on 31 October 2026, and revision evaluation continued through 1 November 2026. Each repetition used a fresh ephemeral request with external retrieval, persistent memory, shell and multi-agent tools, apps, plugins, project-document loading, and host-skill discovery disabled. These controls standardize the supplied context but do not remove implicit service instructions. Because a hosted model can change behind a fixed identifier, timestamps, requested settings, request metadata, and raw responses are archived. All reported quality ratings come from this configuration. The same model was also used, in separate sessions, for corpus construction, revision briefs, and fidelity review (Sections~\ref{sec:collection} and~\ref{sec:revision}). Across the entire project, the workflow consumed an estimated one billion tokens. Cross-model agreement, counterbalanced pairwise preferences, and agreement with human assessments were not evaluated.

\subsection{Source-Constrained Revision and Fidelity Review}
\label{sec:revision}

We attempted a revision of every original figure. Each brief combined the source image, its caption and relevant manuscript explanation, and limitations identified in the original ratings. Briefs were prepared and revisions directed by \model\ in sessions separate from quality assessment. The revision process therefore had access to source information withheld during quality assessment. It also incorporated feedback from the same model used for subsequent scoring, although revised figures were scored in fresh sessions without access to the briefs or the original ratings. Paired changes characterize this feedback-guided workflow under the configured evaluator.

The workflow combined OpenAI's \texttt{gpt-image-2.5} for image generation, with programmatic restoration of source content where necessary. Photographs, masks, heatmaps, curves, numerical values, and other protected data-bearing regions were retained from the source. The archive records source regions, their placements, and applicable pixel-preservation checks. Revisions range from reorganized diagrams to retained scientific panels supplemented with reading guides or explanatory keys. Observed changes can therefore reflect both altered presentation and additional information made explicit within the image. The intervention does not isolate the contributions of layout, color, or generative rendering.

Fidelity review used \model\ in sessions separate from quality assessment to compare each derivative with the source figure and available caption and body text. The review examined objects, relationships, labels, and scope alongside applicable pixel-restoration checks. Passing indicates that the recorded checks found no disqualifying discrepancy and does not constitute independent expert validation. Of 371 attempted revisions, 369 passed and received three fresh anonymous quality judgments each, yielding 1,107 final-version responses. Two figures remained unsuccessful after an initial draft and two semantic repairs. The EAGLE motivation figure incorrectly connected image inputs to authenticity instructions. The \emph{Make It Up} framework omitted a pooling-to-local-prototype update and introduced an unsupported reverse connection. Both remain in the revision-attempt denominator without revision quality scores. Paired quality analysis is therefore conditional on passing fidelity review.

Semantic repairs do not constitute additional intervention samples, and accepted revisions were not regenerated in response to low quality scores. An earlier scored PhoenixSR revision was superseded after further verification. Its three responses are retained as historical records and excluded from the final analysis. Checksums link images, anonymous views, prompts, responses, and accepted derivatives to ensure that scores correspond to the final image version. Original rating records remain unchanged.

\subsection{Paired Analysis and Uncertainty}
\label{sec:analysis}

Let $y^{t}_{idr}$ denote the score for figure $i$, dimension $d$, repeat $r\in\{1,2,3\}$, and version $t\in\{\mathrm{original},\mathrm{revision}\}$. We summarize valid numerical responses by
\begin{equation}
m^{t}_{id}
=
\operatorname{median}_{r\in\mathcal{R}^{t}_{id}}
y^{t}_{idr},
\end{equation}
where $\mathcal{R}^{t}_{id}$ contains repeats with numerical scores. The number and range of valid numerical scores are retained. When this set is empty, the median is undefined, and missing values are not imputed. For a revision that passes fidelity review and has numerical medians in both versions for dimension $d$, the paired change is
\begin{equation}
\Delta_{id}
=
m^{\mathrm{revision}}_{id}
-
m^{\mathrm{original}}_{id}.
\label{eq:change}
\end{equation}

The six dimensions remain the primary outcomes. Group profiles report arithmetic means of figure-level medians. Paired summaries report mean changes and counts of increases, ties, and decreases with their valid denominators. Main-sample analyses distinguish temporal stages and figure roles while retaining topic, source, and eligibility-tier information. The contrast sample is analyzed separately and excluded from temporal stage means. Explicitly pooled paired results describe the accepted revision set.

The gallery additionally reports a post hoc equal-weight mean across the six dimensions,
\begin{equation}
\bar m^{t}_{i}
=
\frac{1}{6}\sum_{d=1}^{6}m^{t}_{id},
\qquad
\Delta\bar m_i
=
\bar m^{\mathrm{revision}}_i
-
\bar m^{\mathrm{original}}_i.
\end{equation}
This mean is computed only when all six dimension medians are available for the corresponding version. It is a descriptive average of ordinal ratings and does not constitute a validated overall-quality scale. Differences are calculated before display rounding.

Uncertainty in reported mean changes is summarized using 95\% paper-cluster bootstrap percentile intervals with 5,000 resamples and random seed 20261020. Paper identifiers are sampled uniformly with replacement within the relevant analysis set. Each sampled paper contributes all its included figures and their matched versions, and each replicate recomputes the figure-weighted mean. The 2.5th and 97.5th percentiles define the interval. This procedure preserves within-paper dependence and the original figure weighting, using frozen score files linked to paper identities through the manifest. Neither the 2,220 validated responses nor the 2,214 paired figure--dimension outcomes are treated as independent sampling units. The intervals quantify resampling variability within the corpus. They do not account for non-probability selection, source and eligibility differences, service drift, or systematic evaluator bias.

\section{Results and Summary}
\label{sec:results}

We begin with 40 paper-level cases, comprising eight papers from each of S1--S4, with two per topic, and eight from the separate contrast sample. The complete atlas of 200 papers, including original figures, revision outcomes, and evaluation scores, is available at \url{https://ntu-ai4x.github.io/Rethinking_Figures/index_en.html}.

The gallery provides visual context for analyses based on the complete eligible samples. Motivation (M) and framework (F) figures are distinguished throughout. C, O, G, A, I, and S denote clarity, organization and readability, graphical consistency, aesthetic coherence, purposeful visual inventiveness, and design specificity. Displayed scores are figure-level medians of three judgments. The six-dimension mean is a post hoc descriptive summary, and changes are calculated before rounding.

\phantomsection
\label{atlas:corpus001}
\pdfbookmark[1]{1. A Dual Weighting Label Assignment Scheme for Object Detection}{corpus001}
{\Large\bfseries\color{academicblue} 001 \quad \href{https://openaccess.thecvf.com/content/CVPR2022/html/Li\_A\_Dual\_Weighting\_Label\_Assignment\_Scheme\_for\_Object\_Detection\_CVPR\_2022\_paper.html}{A Dual Weighting Label Assignment Scheme for Object Detection} \citep{corpus001}\par}
\vspace{3pt}
{\small S1 \enspace $\vert$ \enspace Object detection \enspace $\vert$ \enspace CVPR 2022\par}
\vspace{4pt}
\vspace{5pt}
{\bfseries Motivation (source Fig.\ 1)} \hfill {\footnotesize \href{https://openaccess.thecvf.com/content/CVPR2022/papers/Li\_A\_Dual\_Weighting\_Label\_Assignment\_Scheme\_for\_Object\_Detection\_CVPR\_2022\_paper.pdf}{Source PDF, p.\ 1}; ID: P65}\par
\vspace{2pt}
\noindent\begin{minipage}[t]{0.486\linewidth}
\centering\textbf{Original}\par\vspace{2pt}
\begin{minipage}[c][0.255\textheight][c]{\linewidth}\centering\includegraphics[width=\linewidth,height=0.255\textheight,keepaspectratio]{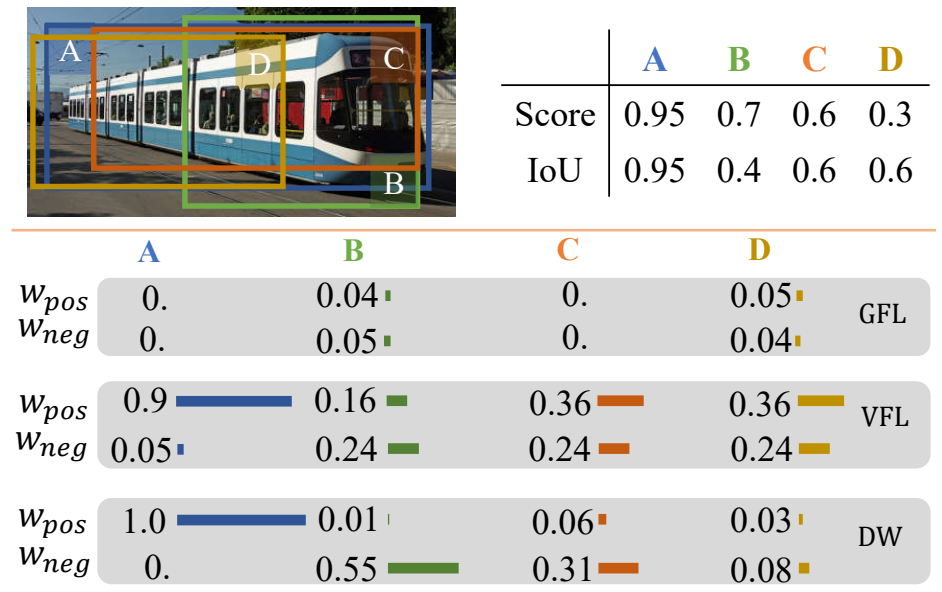}\end{minipage}\par
\vspace{3pt}{\scriptsize \setlength{\tabcolsep}{1.5pt}\begin{tabular*}{\linewidth}{@{\extracolsep{\fill}}rrrrrrr@{}}
\toprule
C & O & G & A & I & S & Mean \\
3 & 4 & 5 & 4 & 4 & 5 & 4.17\\
\bottomrule
\end{tabular*}
}
\end{minipage}\hfill
\begin{minipage}[t]{0.486\linewidth}
\centering\textbf{Study-generated derivative}\par\vspace{2pt}
\begin{minipage}[c][0.255\textheight][c]{\linewidth}\centering\includegraphics[width=\linewidth,height=0.255\textheight,keepaspectratio]{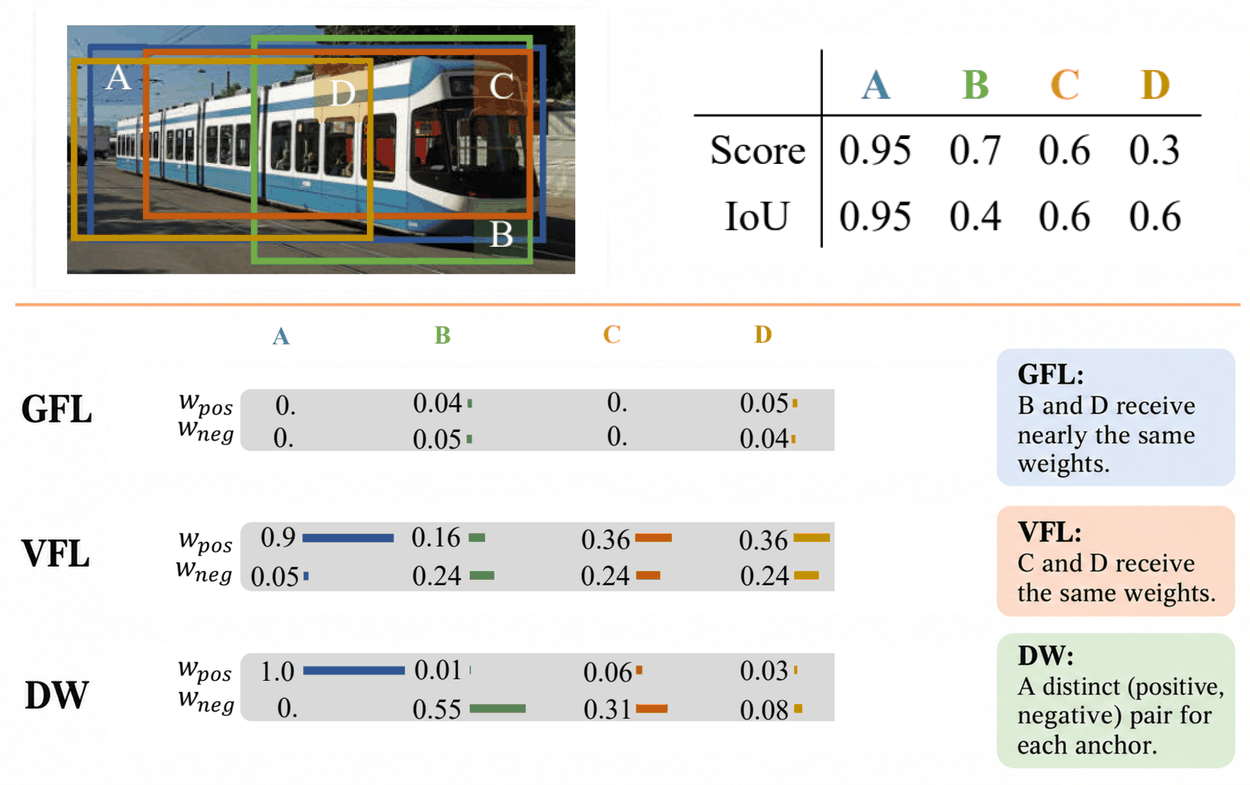}\end{minipage}\par
\vspace{3pt}
{\scriptsize \setlength{\tabcolsep}{1.5pt}\begin{tabular*}{\linewidth}{@{\extracolsep{\fill}}rrrrrrr@{}}
\toprule
C & O & G & A & I & S & Mean \\
4 (+1) & 5 (+1) & 5 (+0) & 4 (+0) & 4 (+0) & 5 (+0) & 4.50 (+0.33)\\
\bottomrule
\end{tabular*}
}
\end{minipage}\par
\vspace{3pt}
{\scriptsize Source public-date proxy: 2022-03-18. All displayed scores are medians of three runs; parentheses show revised minus original.\par}
\vspace{5pt}
{\bfseries Framework (source Fig.\ 2)} \hfill {\footnotesize \href{https://openaccess.thecvf.com/content/CVPR2022/papers/Li\_A\_Dual\_Weighting\_Label\_Assignment\_Scheme\_for\_Object\_Detection\_CVPR\_2022\_paper.pdf}{Source PDF, p.\ 4}; ID: P66}\par
\vspace{2pt}
\noindent\begin{minipage}[t]{0.486\linewidth}
\centering\textbf{Original}\par\vspace{2pt}
\begin{minipage}[c][0.255\textheight][c]{\linewidth}\centering\includegraphics[width=\linewidth,height=0.255\textheight,keepaspectratio]{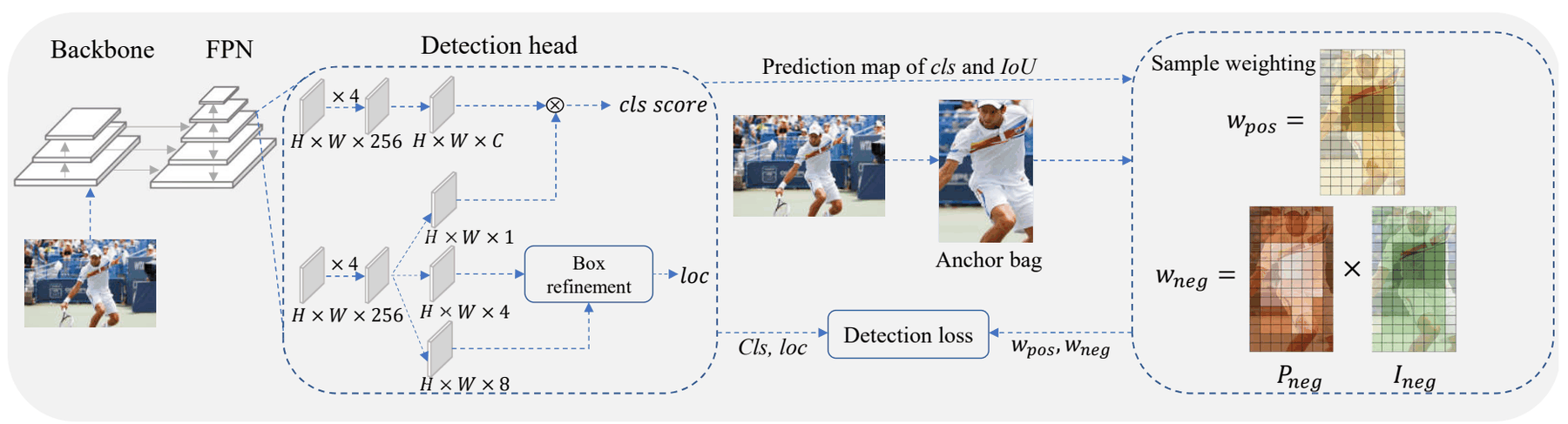}\end{minipage}\par
\vspace{3pt}{\scriptsize \setlength{\tabcolsep}{1.5pt}\begin{tabular*}{\linewidth}{@{\extracolsep{\fill}}rrrrrrr@{}}
\toprule
C & O & G & A & I & S & Mean \\
3 & 4 & 4 & 4 & 3 & 5 & 3.83\\
\bottomrule
\end{tabular*}
}
\end{minipage}\hfill
\begin{minipage}[t]{0.486\linewidth}
\centering\textbf{Study-generated derivative}\par\vspace{2pt}
\begin{minipage}[c][0.255\textheight][c]{\linewidth}\centering\includegraphics[width=\linewidth,height=0.255\textheight,keepaspectratio]{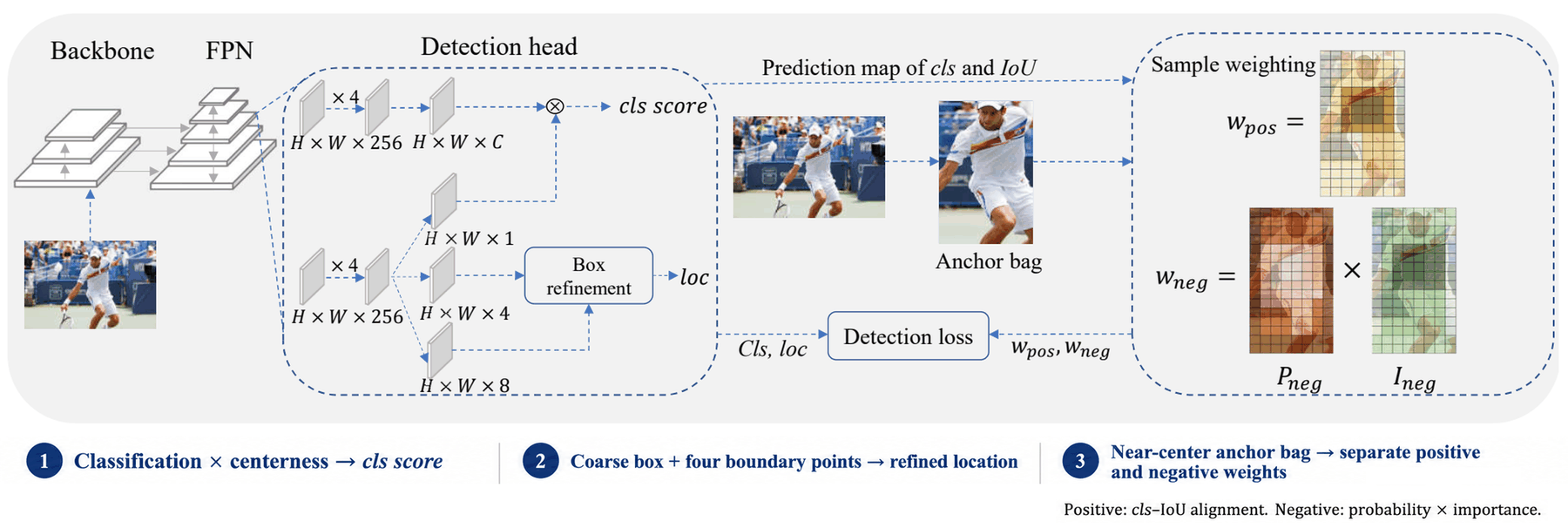}\end{minipage}\par
\vspace{3pt}
{\scriptsize \setlength{\tabcolsep}{1.5pt}\begin{tabular*}{\linewidth}{@{\extracolsep{\fill}}rrrrrrr@{}}
\toprule
C & O & G & A & I & S & Mean \\
4 (+1) & 4 (+0) & 4 (+0) & 4 (+0) & 3 (+0) & 4 (-1) & 3.83 (+0.00)\\
\bottomrule
\end{tabular*}
}
\end{minipage}\par
\vspace{3pt}
{\scriptsize Source public-date proxy: 2022-03-18. All displayed scores are medians of three runs; parentheses show revised minus original.\par}
\vfill
{\scriptsize C: clarity; O: organization/readability; G: graphical consistency; A: aesthetic coherence; I: visual inventiveness; S: design specificity. Mean is the arithmetic mean of six dimension medians, a descriptive display value, not a validated quality index.\par}

\clearpage
\phantomsection
\label{atlas:corpus004}
\pdfbookmark[1]{4. Localization Distillation for Dense Object Detection}{corpus004}
{\Large\bfseries\color{academicblue} 004 \quad \href{https://openaccess.thecvf.com/content/CVPR2022/html/Zheng\_Localization\_Distillation\_for\_Dense\_Object\_Detection\_CVPR\_2022\_paper.html}{Localization Distillation for Dense Object Detection} \citep{corpus004}\par}
\vspace{3pt}
{\small S1 \enspace $\vert$ \enspace Object detection \enspace $\vert$ \enspace CVPR 2022\par}
\vspace{4pt}
\vspace{5pt}
{\bfseries Motivation (source Fig.\ 1)} \hfill {\footnotesize \href{https://openaccess.thecvf.com/content/CVPR2022/papers/Zheng\_Localization\_Distillation\_for\_Dense\_Object\_Detection\_CVPR\_2022\_paper.pdf}{Source PDF, p.\ 1}; ID: P213}\par
\vspace{2pt}
\noindent\begin{minipage}[t]{0.486\linewidth}
\centering\textbf{Original}\par\vspace{2pt}
\begin{minipage}[c][0.255\textheight][c]{\linewidth}\centering\includegraphics[width=\linewidth,height=0.255\textheight,keepaspectratio]{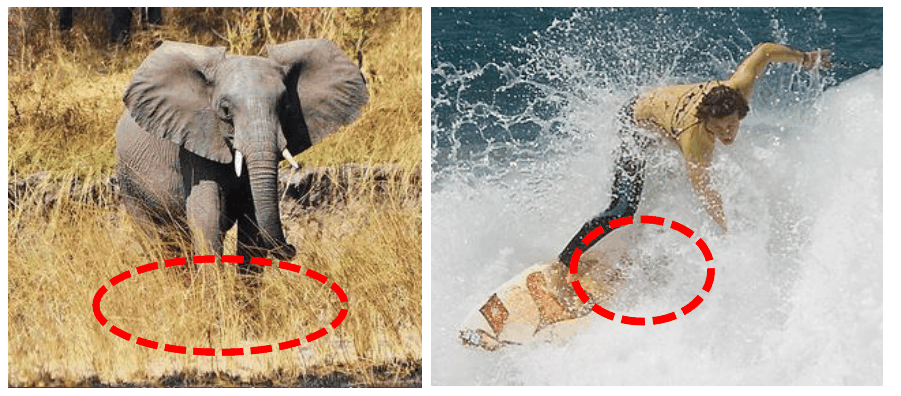}\end{minipage}\par
\vspace{3pt}{\scriptsize \setlength{\tabcolsep}{1.5pt}\begin{tabular*}{\linewidth}{@{\extracolsep{\fill}}rrrrrrr@{}}
\toprule
C & O & G & A & I & S & Mean \\
2 & 4 & 4 & 4 & 2 & 3 & 3.17\\
\bottomrule
\end{tabular*}
}
\end{minipage}\hfill
\begin{minipage}[t]{0.486\linewidth}
\centering\textbf{Study-generated derivative}\par\vspace{2pt}
\begin{minipage}[c][0.255\textheight][c]{\linewidth}\centering\includegraphics[width=\linewidth,height=0.255\textheight,keepaspectratio]{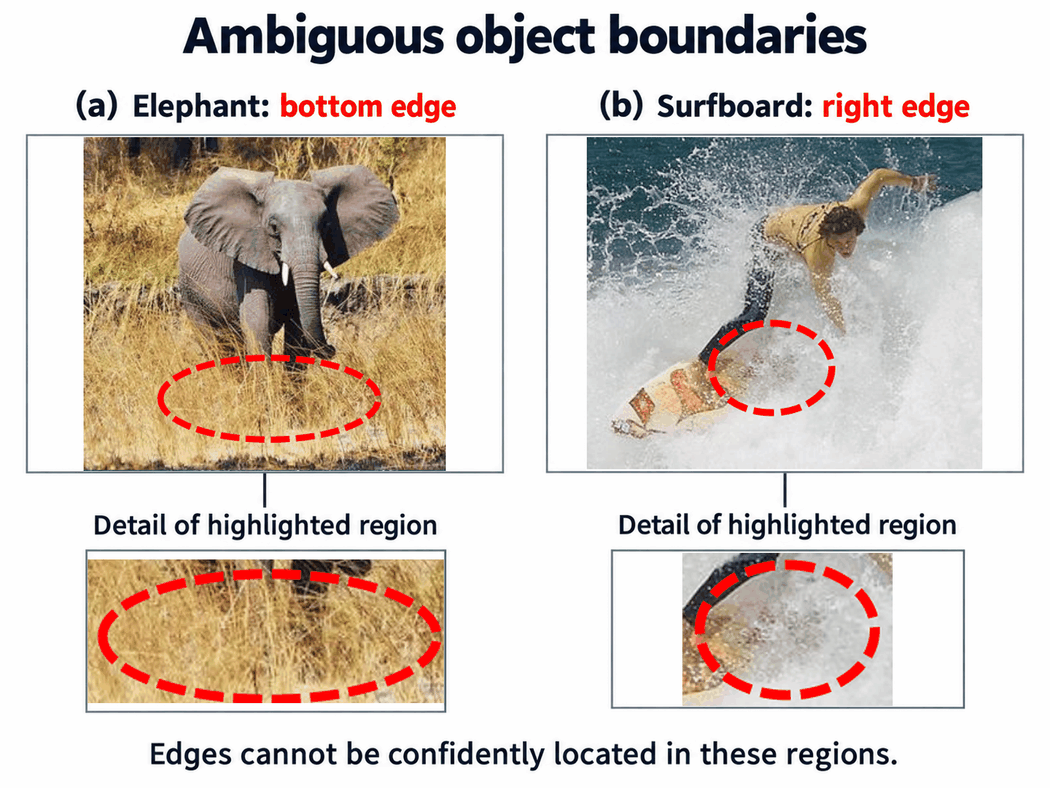}\end{minipage}\par
\vspace{3pt}
{\scriptsize \setlength{\tabcolsep}{1.5pt}\begin{tabular*}{\linewidth}{@{\extracolsep{\fill}}rrrrrrr@{}}
\toprule
C & O & G & A & I & S & Mean \\
5 (+3) & 5 (+1) & 5 (+1) & 4 (+0) & 3 (+1) & 4 (+1) & 4.33 (+1.17)\\
\bottomrule
\end{tabular*}
}
\end{minipage}\par
\vspace{3pt}
{\scriptsize Source public-date proxy: 2021-02-24. All displayed scores are medians of three runs; parentheses show revised minus original.\par}
\vspace{5pt}
{\bfseries Framework (source Fig.\ 3)} \hfill {\footnotesize \href{https://openaccess.thecvf.com/content/CVPR2022/papers/Zheng\_Localization\_Distillation\_for\_Dense\_Object\_Detection\_CVPR\_2022\_paper.pdf}{Source PDF, p.\ 3}; ID: P214}\par
\vspace{2pt}
\noindent\begin{minipage}[t]{0.486\linewidth}
\centering\textbf{Original}\par\vspace{2pt}
\begin{minipage}[c][0.255\textheight][c]{\linewidth}\centering\includegraphics[width=\linewidth,height=0.255\textheight,keepaspectratio]{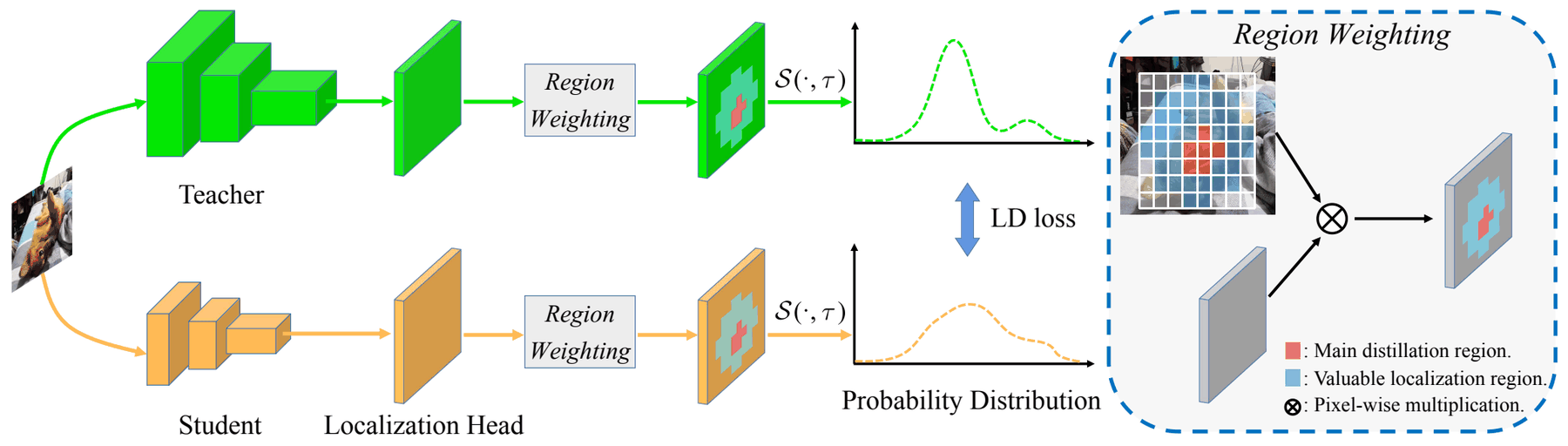}\end{minipage}\par
\vspace{3pt}{\scriptsize \setlength{\tabcolsep}{1.5pt}\begin{tabular*}{\linewidth}{@{\extracolsep{\fill}}rrrrrrr@{}}
\toprule
C & O & G & A & I & S & Mean \\
4 & 4 & 4 & 3 & 3 & 4 & 3.67\\
\bottomrule
\end{tabular*}
}
\end{minipage}\hfill
\begin{minipage}[t]{0.486\linewidth}
\centering\textbf{Study-generated derivative}\par\vspace{2pt}
\begin{minipage}[c][0.255\textheight][c]{\linewidth}\centering\includegraphics[width=\linewidth,height=0.255\textheight,keepaspectratio]{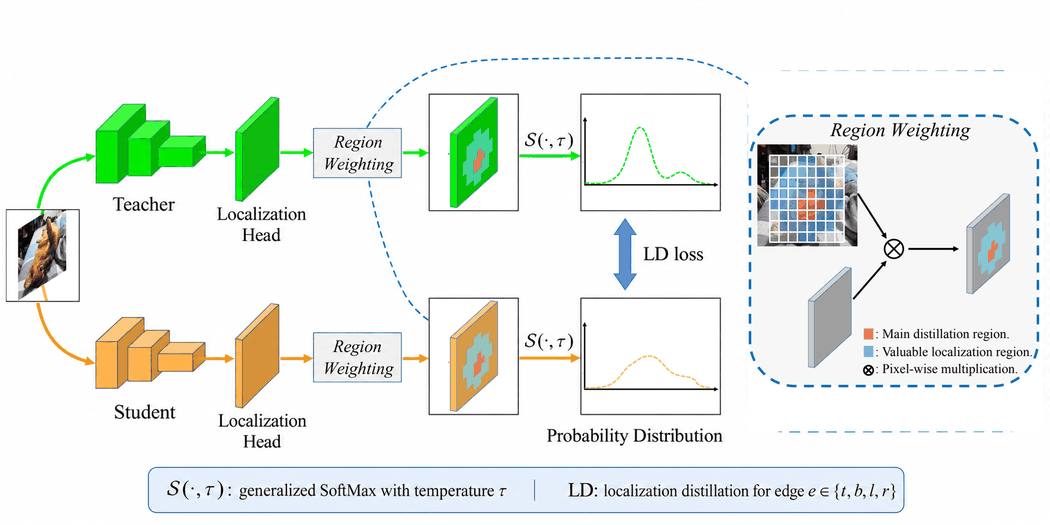}\end{minipage}\par
\vspace{3pt}
{\scriptsize \setlength{\tabcolsep}{1.5pt}\begin{tabular*}{\linewidth}{@{\extracolsep{\fill}}rrrrrrr@{}}
\toprule
C & O & G & A & I & S & Mean \\
4 (+0) & 4 (+0) & 4 (+0) & 3 (+0) & 3 (+0) & 4 (+0) & 3.67 (+0.00)\\
\bottomrule
\end{tabular*}
}
\end{minipage}\par
\vspace{3pt}
{\scriptsize Source public-date proxy: 2021-11-25. All displayed scores are medians of three runs; parentheses show revised minus original.\par}
\vfill
{\scriptsize C: clarity; O: organization/readability; G: graphical consistency; A: aesthetic coherence; I: visual inventiveness; S: design specificity. Mean is the arithmetic mean of six dimension medians, a descriptive display value, not a validated quality index.\par}

\clearpage
\phantomsection
\label{atlas:corpus012}
\pdfbookmark[1]{12. Decoupling Zero-Shot Semantic Segmentation}{corpus012}
{\Large\bfseries\color{academicblue} 012 \quad \href{https://openaccess.thecvf.com/content/CVPR2022/html/Ding\_Decoupling\_Zero-Shot\_Semantic\_Segmentation\_CVPR\_2022\_paper.html}{Decoupling Zero-Shot Semantic Segmentation} \citep{corpus012}\par}
\vspace{3pt}
{\small S1 \enspace $\vert$ \enspace Image segmentation \enspace $\vert$ \enspace CVPR 2022\par}
\vspace{4pt}
\vspace{5pt}
{\bfseries Motivation (source Fig.\ 1)} \hfill {\footnotesize \href{https://openaccess.thecvf.com/content/CVPR2022/papers/Ding\_Decoupling\_Zero-Shot\_Semantic\_Segmentation\_CVPR\_2022\_paper.pdf}{Source PDF, p.\ 1}; ID: P125}\par
\vspace{2pt}
\noindent\begin{minipage}[t]{0.486\linewidth}
\centering\textbf{Original}\par\vspace{2pt}
\begin{minipage}[c][0.255\textheight][c]{\linewidth}\centering\includegraphics[width=\linewidth,height=0.255\textheight,keepaspectratio]{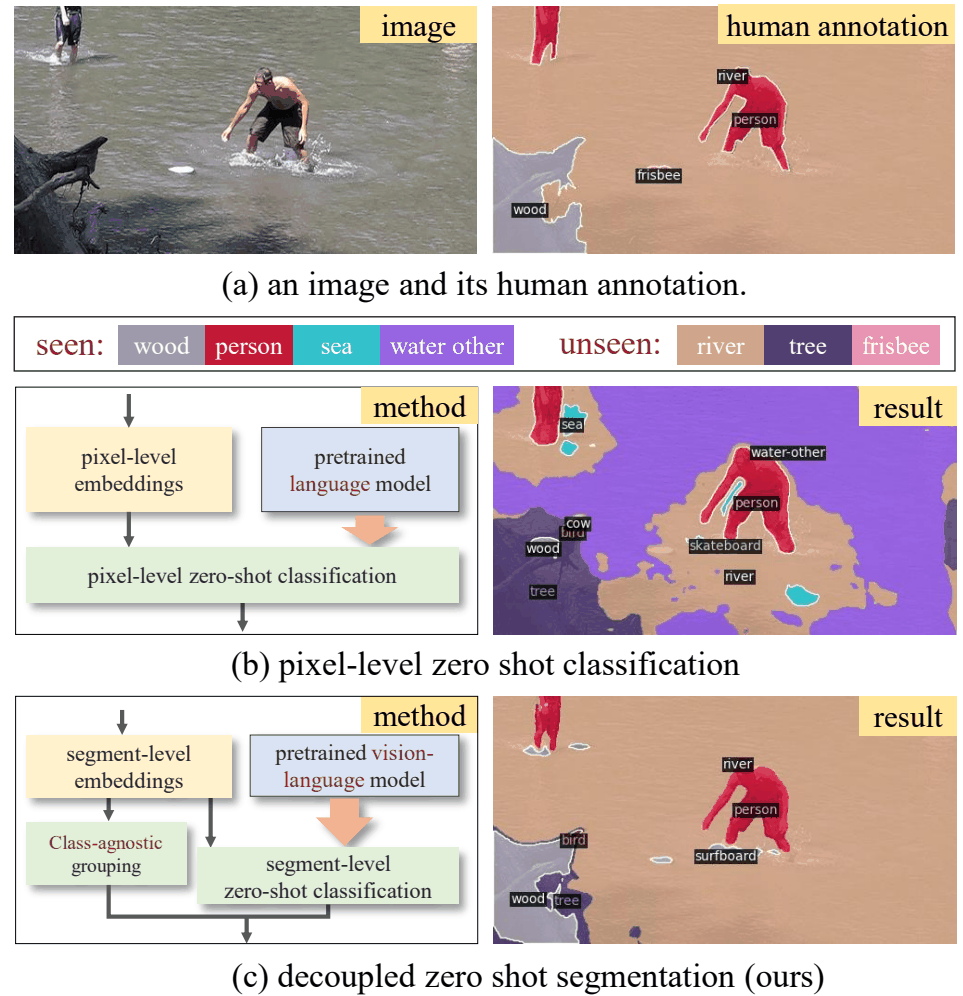}\end{minipage}\par
\vspace{3pt}{\scriptsize \setlength{\tabcolsep}{1.5pt}\begin{tabular*}{\linewidth}{@{\extracolsep{\fill}}rrrrrrr@{}}
\toprule
C & O & G & A & I & S & Mean \\
4 & 4 & 4 & 3 & 4 & 5 & 4.00\\
\bottomrule
\end{tabular*}
}
\end{minipage}\hfill
\begin{minipage}[t]{0.486\linewidth}
\centering\textbf{Study-generated derivative}\par\vspace{2pt}
\begin{minipage}[c][0.255\textheight][c]{\linewidth}\centering\includegraphics[width=\linewidth,height=0.255\textheight,keepaspectratio]{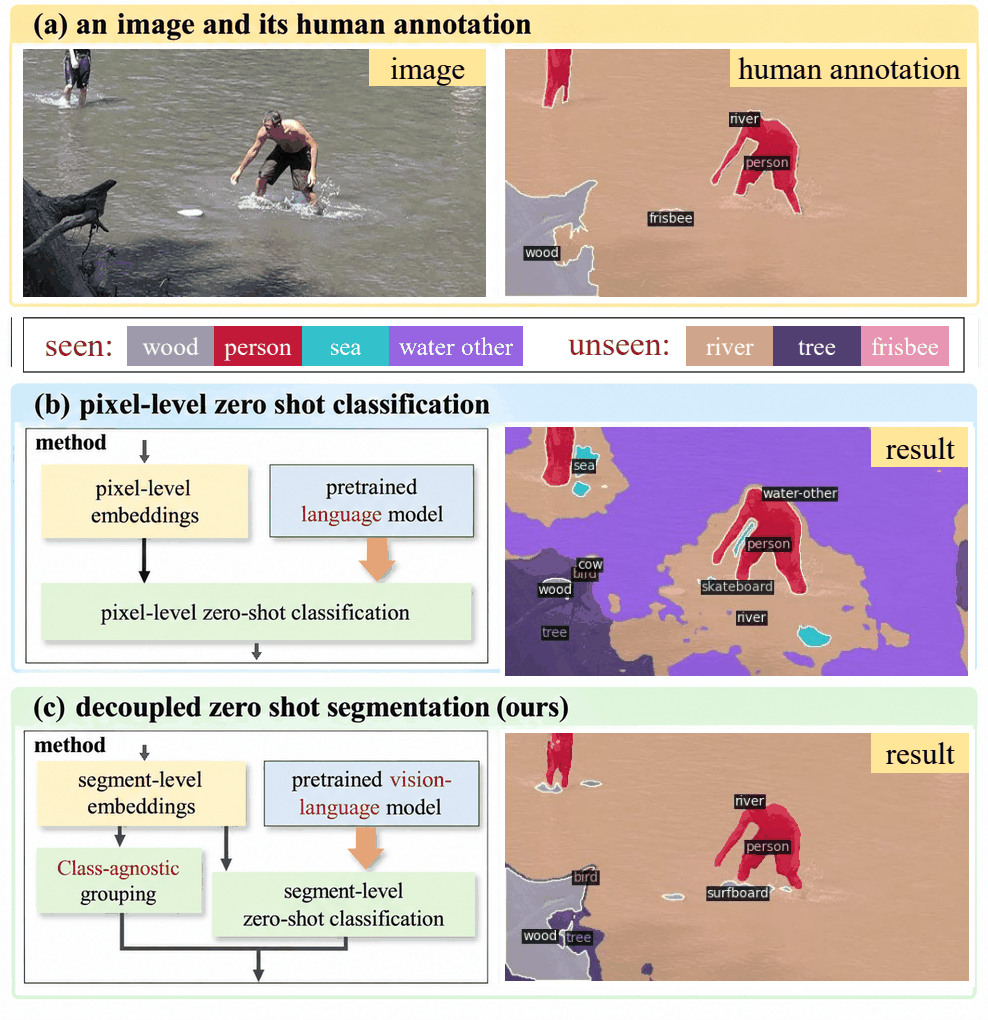}\end{minipage}\par
\vspace{3pt}
{\scriptsize \setlength{\tabcolsep}{1.5pt}\begin{tabular*}{\linewidth}{@{\extracolsep{\fill}}rrrrrrr@{}}
\toprule
C & O & G & A & I & S & Mean \\
4 (+0) & 4 (+0) & 4 (+0) & 4 (+1) & 4 (+0) & 5 (+0) & 4.17 (+0.17)\\
\bottomrule
\end{tabular*}
}
\end{minipage}\par
\vspace{3pt}
{\scriptsize Source public-date proxy: 2022-04-15. All displayed scores are medians of three runs; parentheses show revised minus original.\par}
\vspace{5pt}
{\bfseries Framework (source Fig.\ 2)} \hfill {\footnotesize \href{https://openaccess.thecvf.com/content/CVPR2022/papers/Ding\_Decoupling\_Zero-Shot\_Semantic\_Segmentation\_CVPR\_2022\_paper.pdf}{Source PDF, p.\ 4}; ID: P126}\par
\vspace{2pt}
\noindent\begin{minipage}[t]{0.486\linewidth}
\centering\textbf{Original}\par\vspace{2pt}
\begin{minipage}[c][0.255\textheight][c]{\linewidth}\centering\includegraphics[width=\linewidth,height=0.255\textheight,keepaspectratio]{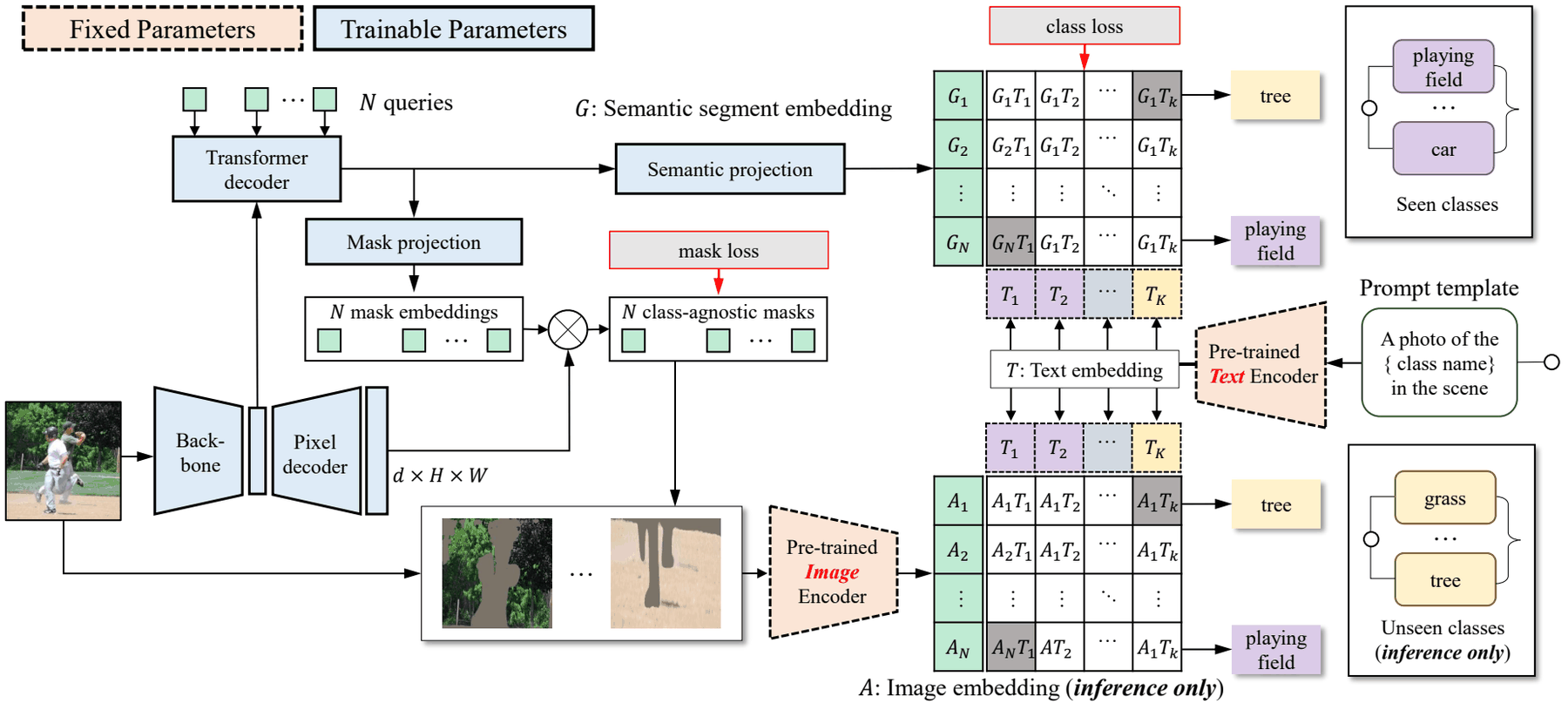}\end{minipage}\par
\vspace{3pt}{\scriptsize \setlength{\tabcolsep}{1.5pt}\begin{tabular*}{\linewidth}{@{\extracolsep{\fill}}rrrrrrr@{}}
\toprule
C & O & G & A & I & S & Mean \\
4 & 4 & 3 & 4 & 4 & 5 & 4.00\\
\bottomrule
\end{tabular*}
}
\end{minipage}\hfill
\begin{minipage}[t]{0.486\linewidth}
\centering\textbf{Study-generated derivative}\par\vspace{2pt}
\begin{minipage}[c][0.255\textheight][c]{\linewidth}\centering\includegraphics[width=\linewidth,height=0.255\textheight,keepaspectratio]{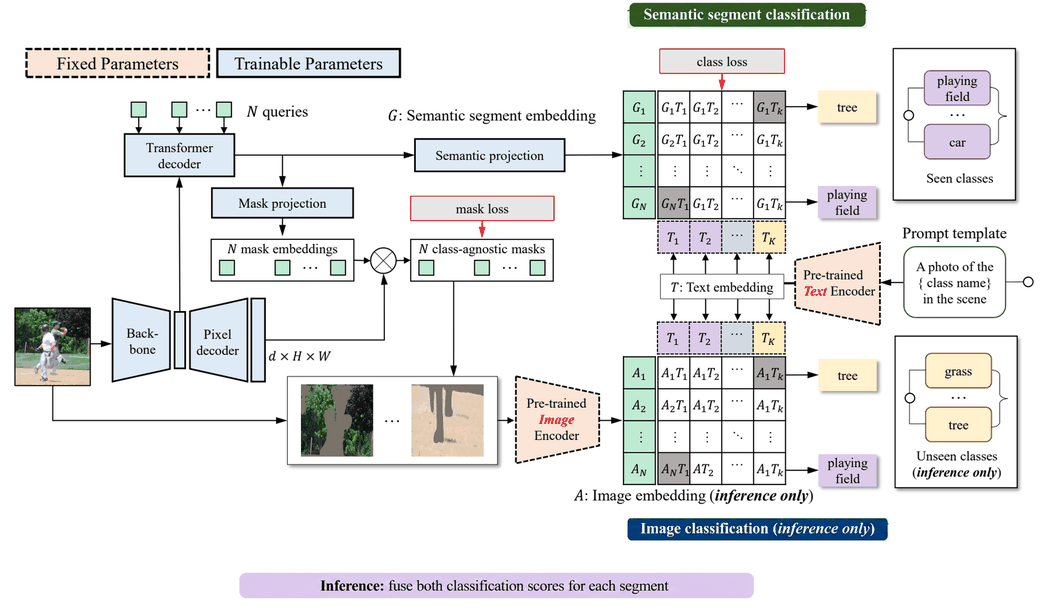}\end{minipage}\par
\vspace{3pt}
{\scriptsize \setlength{\tabcolsep}{1.5pt}\begin{tabular*}{\linewidth}{@{\extracolsep{\fill}}rrrrrrr@{}}
\toprule
C & O & G & A & I & S & Mean \\
4 (+0) & 4 (+0) & 3 (+0) & 3 (-1) & 4 (+0) & 5 (+0) & 3.83 (-0.17)\\
\bottomrule
\end{tabular*}
}
\end{minipage}\par
\vspace{3pt}
{\scriptsize Source public-date proxy: 2022-04-15. All displayed scores are medians of three runs; parentheses show revised minus original.\par}
\vfill
{\scriptsize C: clarity; O: organization/readability; G: graphical consistency; A: aesthetic coherence; I: visual inventiveness; S: design specificity. Mean is the arithmetic mean of six dimension medians, a descriptive display value, not a validated quality index.\par}

\clearpage
\phantomsection
\label{atlas:corpus016}
\pdfbookmark[1]{16. Open-World Instance Segmentation: Exploiting Pseudo Ground Truth From Learned Pairwise Affinity}{corpus016}
{\Large\bfseries\color{academicblue} 016 \quad \href{https://openaccess.thecvf.com/content/CVPR2022/html/Wang_Open-World_Instance_Segmentation_Exploiting_Pseudo_Ground_Truth_From_Learned_Pairwise_CVPR_2022_paper.html}{Open-World Instance Segmentation: Exploiting Pseudo Ground Truth From Learned Pairwise Affinity} \citep{corpus016}\par}
\vspace{3pt}
{\small S1 \enspace $\vert$ \enspace Image segmentation \enspace $\vert$ \enspace CVPR 2022\par}
\vspace{4pt}
\vspace{5pt}
{\bfseries Motivation (source Fig.\ 1)} \hfill {\footnotesize \href{https://openaccess.thecvf.com/content/CVPR2022/papers/Wang\_Open-World\_Instance\_Segmentation\_Exploiting\_Pseudo\_Ground\_Truth\_From\_Learned\_Pairwise\_CVPR\_2022\_paper.pdf}{Source PDF, p.\ 1}; ID: P49}\par
\vspace{2pt}
\noindent\begin{minipage}[t]{0.486\linewidth}
\centering\textbf{Original}\par\vspace{2pt}
\begin{minipage}[c][0.255\textheight][c]{\linewidth}\centering\includegraphics[width=\linewidth,height=0.255\textheight,keepaspectratio]{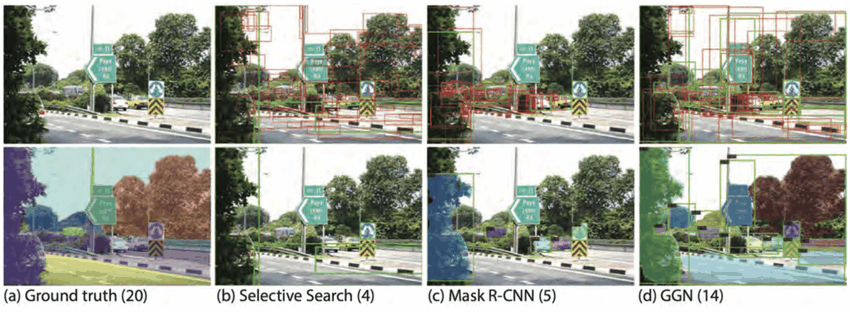}\end{minipage}\par
\vspace{3pt}{\scriptsize \setlength{\tabcolsep}{1.5pt}\begin{tabular*}{\linewidth}{@{\extracolsep{\fill}}rrrrrrr@{}}
\toprule
C & O & G & A & I & S & Mean \\
3 & 4 & 3 & 3 & 3 & 4 & 3.33\\
\bottomrule
\end{tabular*}
}
\end{minipage}\hfill
\begin{minipage}[t]{0.486\linewidth}
\centering\textbf{Study-generated derivative}\par\vspace{2pt}
\begin{minipage}[c][0.255\textheight][c]{\linewidth}\centering\includegraphics[width=\linewidth,height=0.255\textheight,keepaspectratio]{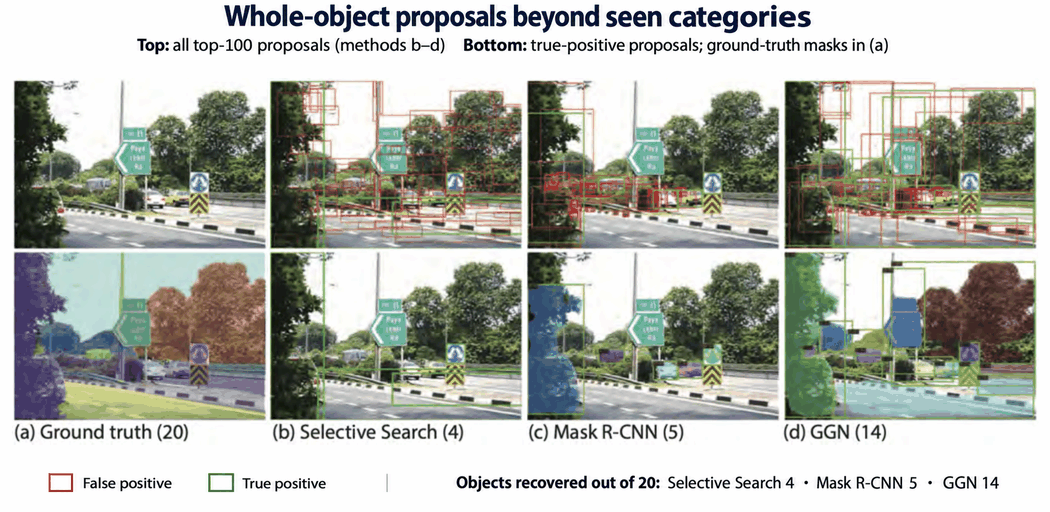}\end{minipage}\par
\vspace{3pt}
{\scriptsize \setlength{\tabcolsep}{1.5pt}\begin{tabular*}{\linewidth}{@{\extracolsep{\fill}}rrrrrrr@{}}
\toprule
C & O & G & A & I & S & Mean \\
4 (+1) & 4 (+0) & 4 (+1) & 4 (+1) & 3 (+0) & 5 (+1) & 4.00 (+0.67)\\
\bottomrule
\end{tabular*}
}
\end{minipage}\par
\vspace{3pt}
{\scriptsize Source public-date proxy: 2022-04-12. All displayed scores are medians of three runs; parentheses show revised minus original.\par}
\vspace{5pt}
{\bfseries Framework (source Fig.\ 2)} \hfill {\footnotesize \href{https://openaccess.thecvf.com/content/CVPR2022/papers/Wang\_Open-World\_Instance\_Segmentation\_Exploiting\_Pseudo\_Ground\_Truth\_From\_Learned\_Pairwise\_CVPR\_2022\_paper.pdf}{Source PDF, p.\ 3}; ID: P50}\par
\vspace{2pt}
\noindent\begin{minipage}[t]{0.486\linewidth}
\centering\textbf{Original}\par\vspace{2pt}
\begin{minipage}[c][0.255\textheight][c]{\linewidth}\centering\includegraphics[width=\linewidth,height=0.255\textheight,keepaspectratio]{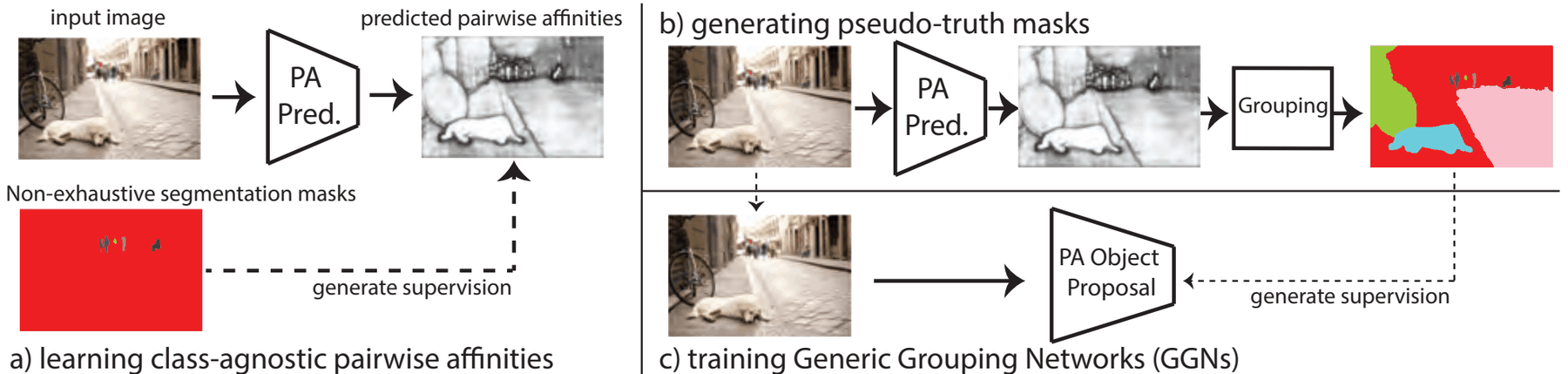}\end{minipage}\par
\vspace{3pt}{\scriptsize \setlength{\tabcolsep}{1.5pt}\begin{tabular*}{\linewidth}{@{\extracolsep{\fill}}rrrrrrr@{}}
\toprule
C & O & G & A & I & S & Mean \\
4 & 4 & 4 & 3 & 3 & 4 & 3.67\\
\bottomrule
\end{tabular*}
}
\end{minipage}\hfill
\begin{minipage}[t]{0.486\linewidth}
\centering\textbf{Study-generated derivative}\par\vspace{2pt}
\begin{minipage}[c][0.255\textheight][c]{\linewidth}\centering\includegraphics[width=\linewidth,height=0.255\textheight,keepaspectratio]{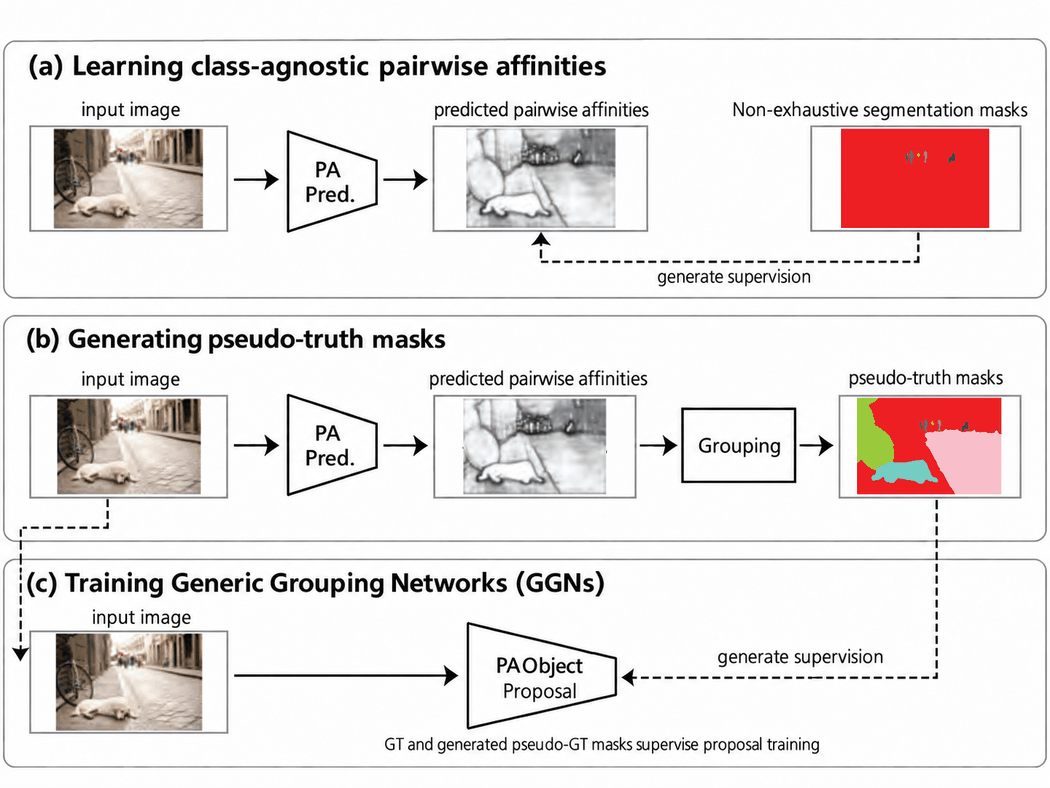}\end{minipage}\par
\vspace{3pt}
{\scriptsize \setlength{\tabcolsep}{1.5pt}\begin{tabular*}{\linewidth}{@{\extracolsep{\fill}}rrrrrrr@{}}
\toprule
C & O & G & A & I & S & Mean \\
4 (+0) & 5 (+1) & 4 (+0) & 4 (+1) & 3 (+0) & 4 (+0) & 4.00 (+0.33)\\
\bottomrule
\end{tabular*}
}
\end{minipage}\par
\vspace{3pt}
{\scriptsize Source public-date proxy: 2022-04-12. All displayed scores are medians of three runs; parentheses show revised minus original.\par}
\vfill
{\scriptsize C: clarity; O: organization/readability; G: graphical consistency; A: aesthetic coherence; I: visual inventiveness; S: design specificity. Mean is the arithmetic mean of six dimension medians, a descriptive display value, not a validated quality index.\par}

\clearpage
\phantomsection
\label{atlas:corpus024}
\pdfbookmark[1]{24. Deep Constrained Least Squares for Blind Image Super-Resolution}{corpus024}
{\Large\bfseries\color{academicblue} 024 \quad \href{https://openaccess.thecvf.com/content/CVPR2022/html/Luo\_Deep\_Constrained\_Least\_Squares\_for\_Blind\_Image\_Super-Resolution\_CVPR\_2022\_paper.html}{Deep Constrained Least Squares for Blind Image Super-Resolution} \citep{corpus024}\par}
\vspace{3pt}
{\small S1 \enspace $\vert$ \enspace Image restoration \enspace $\vert$ \enspace CVPR 2022\par}
\vspace{4pt}
\vspace{5pt}
{\bfseries Motivation (source Fig.\ 2)} \hfill {\footnotesize \href{https://openaccess.thecvf.com/content/CVPR2022/papers/Luo\_Deep\_Constrained\_Least\_Squares\_for\_Blind\_Image\_Super-Resolution\_CVPR\_2022\_paper.pdf}{Source PDF, p.\ 2}; ID: P215}\par
\vspace{2pt}
\noindent\begin{minipage}[t]{0.486\linewidth}
\centering\textbf{Original}\par\vspace{2pt}
\begin{minipage}[c][0.255\textheight][c]{\linewidth}\centering\includegraphics[width=\linewidth,height=0.255\textheight,keepaspectratio]{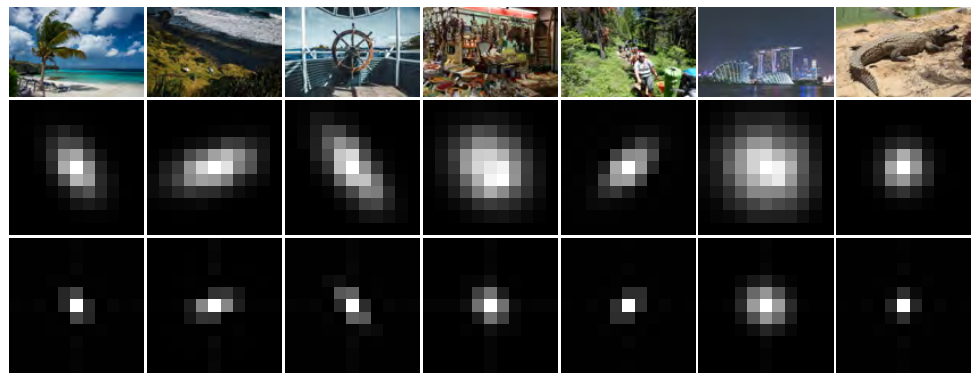}\end{minipage}\par
\vspace{3pt}{\scriptsize \setlength{\tabcolsep}{1.5pt}\begin{tabular*}{\linewidth}{@{\extracolsep{\fill}}rrrrrrr@{}}
\toprule
C & O & G & A & I & S & Mean \\
1 & 4 & 3 & 4 & 3 & 3 & 3.00\\
\bottomrule
\end{tabular*}
}
\end{minipage}\hfill
\begin{minipage}[t]{0.486\linewidth}
\centering\textbf{Study-generated derivative}\par\vspace{2pt}
\begin{minipage}[c][0.255\textheight][c]{\linewidth}\centering\includegraphics[width=\linewidth,height=0.255\textheight,keepaspectratio]{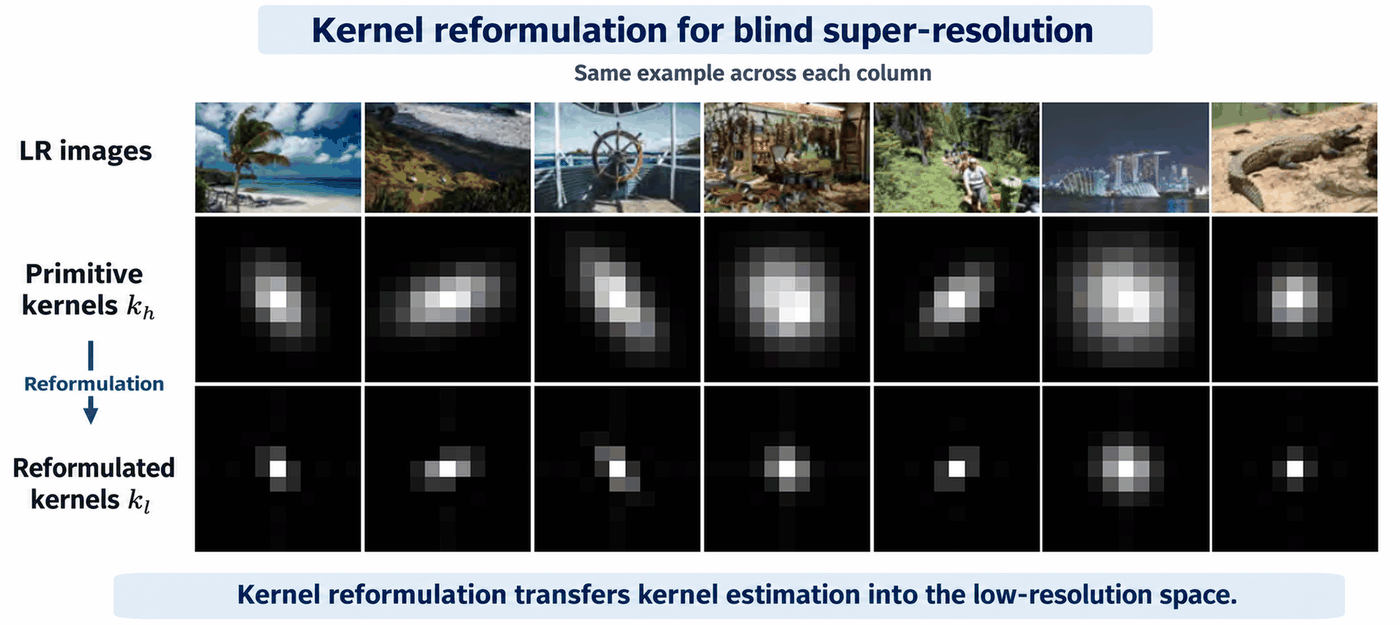}\end{minipage}\par
\vspace{3pt}
{\scriptsize \setlength{\tabcolsep}{1.5pt}\begin{tabular*}{\linewidth}{@{\extracolsep{\fill}}rrrrrrr@{}}
\toprule
C & O & G & A & I & S & Mean \\
4 (+3) & 5 (+1) & 5 (+2) & 4 (+0) & 3 (+0) & 4 (+1) & 4.17 (+1.17)\\
\bottomrule
\end{tabular*}
}
\end{minipage}\par
\vspace{3pt}
{\scriptsize Source public-date proxy: 2022-02-15. All displayed scores are medians of three runs; parentheses show revised minus original.\par}
\vspace{5pt}
{\bfseries Framework (source Fig.\ 4)} \hfill {\footnotesize \href{https://openaccess.thecvf.com/content/CVPR2022/papers/Luo\_Deep\_Constrained\_Least\_Squares\_for\_Blind\_Image\_Super-Resolution\_CVPR\_2022\_paper.pdf}{Source PDF, p.\ 4}; ID: P216}\par
\vspace{2pt}
\noindent\begin{minipage}[t]{0.486\linewidth}
\centering\textbf{Original}\par\vspace{2pt}
\begin{minipage}[c][0.255\textheight][c]{\linewidth}\centering\includegraphics[width=\linewidth,height=0.255\textheight,keepaspectratio]{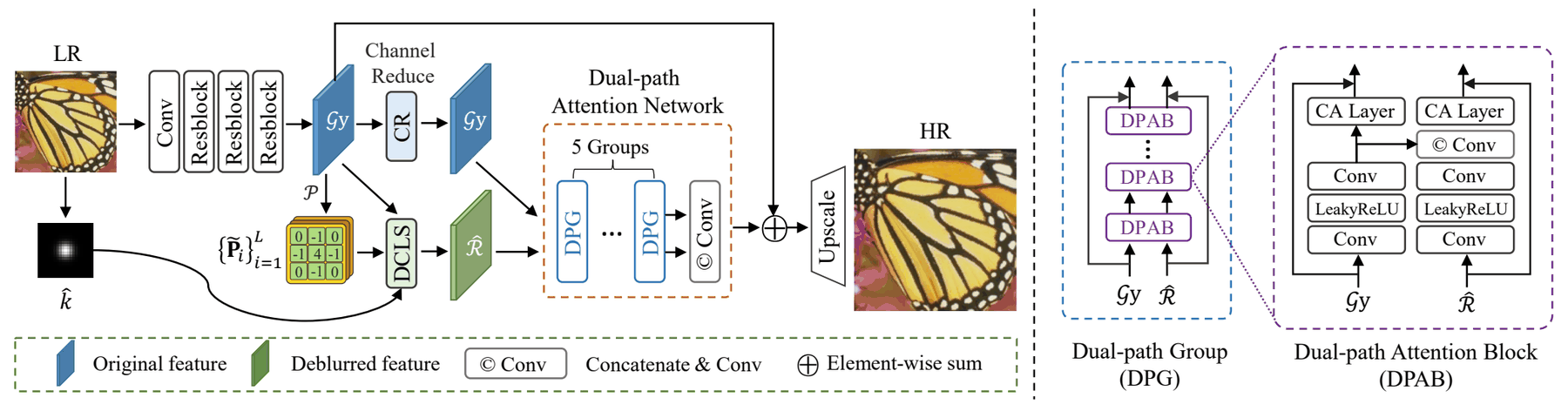}\end{minipage}\par
\vspace{3pt}{\scriptsize \setlength{\tabcolsep}{1.5pt}\begin{tabular*}{\linewidth}{@{\extracolsep{\fill}}rrrrrrr@{}}
\toprule
C & O & G & A & I & S & Mean \\
4 & 4 & 4 & 4 & 3 & 5 & 4.00\\
\bottomrule
\end{tabular*}
}
\end{minipage}\hfill
\begin{minipage}[t]{0.486\linewidth}
\centering\textbf{Study-generated derivative}\par\vspace{2pt}
\begin{minipage}[c][0.255\textheight][c]{\linewidth}\centering\includegraphics[width=\linewidth,height=0.255\textheight,keepaspectratio]{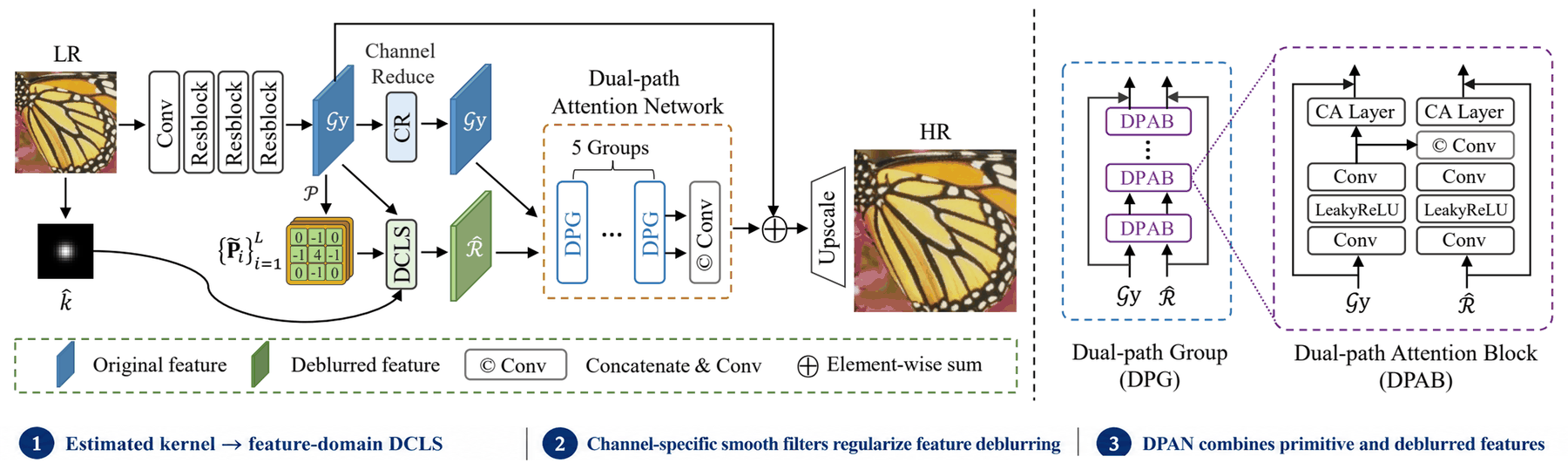}\end{minipage}\par
\vspace{3pt}
{\scriptsize \setlength{\tabcolsep}{1.5pt}\begin{tabular*}{\linewidth}{@{\extracolsep{\fill}}rrrrrrr@{}}
\toprule
C & O & G & A & I & S & Mean \\
4 (+0) & 4 (+0) & 4 (+0) & 4 (+0) & 3 (+0) & 5 (+0) & 4.00 (+0.00)\\
\bottomrule
\end{tabular*}
}
\end{minipage}\par
\vspace{3pt}
{\scriptsize Source public-date proxy: 2022-02-15. All displayed scores are medians of three runs; parentheses show revised minus original.\par}
\vfill
{\scriptsize C: clarity; O: organization/readability; G: graphical consistency; A: aesthetic coherence; I: visual inventiveness; S: design specificity. Mean is the arithmetic mean of six dimension medians, a descriptive display value, not a validated quality index.\par}

\clearpage
\phantomsection
\label{atlas:corpus030}
\pdfbookmark[1]{30. Unsupervised Real-World Image Super Resolution via Domain-Distance Aware Training}{corpus030}
{\Large\bfseries\color{academicblue} 030 \quad \href{https://openaccess.thecvf.com/content/CVPR2021/html/Wei_Unsupervised_Real-World_Image_Super_Resolution_via_Domain-Distance_Aware_Training_CVPR_2021_paper.html}{Unsupervised Real-World Image Super Resolution via Domain-Distance Aware Training} \citep{corpus030}\par}
\vspace{3pt}
{\small S1 \enspace $\vert$ \enspace Image restoration \enspace $\vert$ \enspace CVPR 2021\par}
\vspace{4pt}
\vspace{5pt}
{\bfseries Motivation (source Fig.\ 1)} \hfill {\footnotesize \href{https://openaccess.thecvf.com/content/CVPR2021/papers/Wei\_Unsupervised\_Real-World\_Image\_Super\_Resolution\_via\_Domain-Distance\_Aware\_Training\_CVPR\_2021\_paper.pdf}{Source PDF, p.\ 1}; ID: P325}\par
\vspace{2pt}
\noindent\begin{minipage}[t]{0.486\linewidth}
\centering\textbf{Original}\par\vspace{2pt}
\begin{minipage}[c][0.255\textheight][c]{\linewidth}\centering\includegraphics[width=\linewidth,height=0.255\textheight,keepaspectratio]{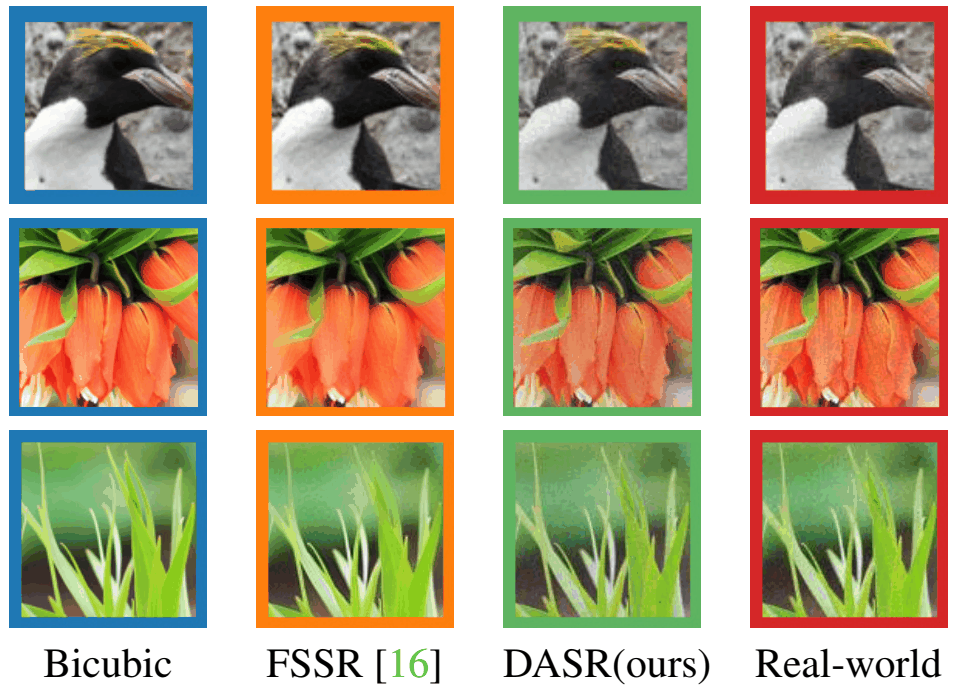}\end{minipage}\par
\vspace{3pt}{\scriptsize \setlength{\tabcolsep}{1.5pt}\begin{tabular*}{\linewidth}{@{\extracolsep{\fill}}rrrrrrr@{}}
\toprule
C & O & G & A & I & S & Mean \\
3 & 5 & 5 & 4 & 3 & 3 & 3.83\\
\bottomrule
\end{tabular*}
}
\end{minipage}\hfill
\begin{minipage}[t]{0.486\linewidth}
\centering\textbf{Study-generated derivative}\par\vspace{2pt}
\begin{minipage}[c][0.255\textheight][c]{\linewidth}\centering\includegraphics[width=\linewidth,height=0.255\textheight,keepaspectratio]{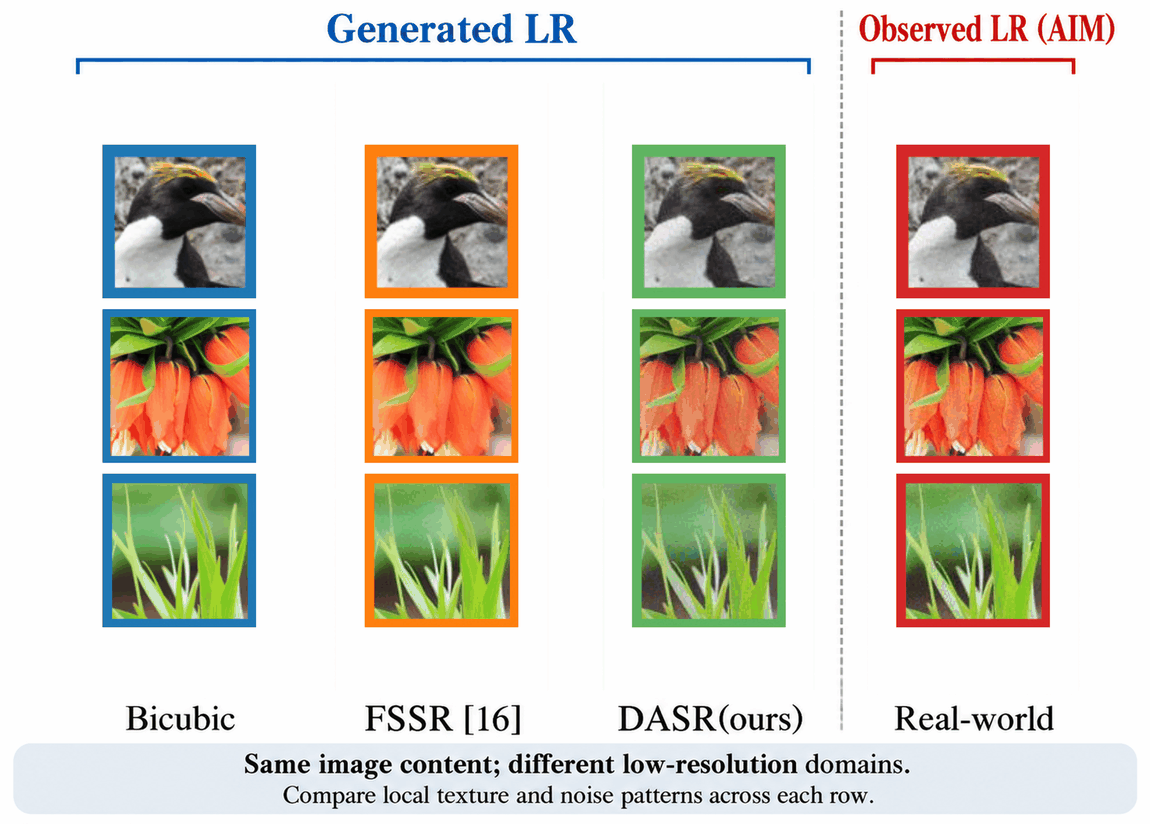}\end{minipage}\par
\vspace{3pt}
{\scriptsize \setlength{\tabcolsep}{1.5pt}\begin{tabular*}{\linewidth}{@{\extracolsep{\fill}}rrrrrrr@{}}
\toprule
C & O & G & A & I & S & Mean \\
4 (+1) & 5 (+0) & 5 (+0) & 4 (+0) & 3 (+0) & 4 (+1) & 4.17 (+0.33)\\
\bottomrule
\end{tabular*}
}
\end{minipage}\par
\vspace{3pt}
{\scriptsize Source public-date proxy: 2021-06. All displayed scores are medians of three runs; parentheses show revised minus original.\par}
\vspace{5pt}
{\bfseries Framework (source Fig.\ 2)} \hfill {\footnotesize \href{https://openaccess.thecvf.com/content/CVPR2021/papers/Wei\_Unsupervised\_Real-World\_Image\_Super\_Resolution\_via\_Domain-Distance\_Aware\_Training\_CVPR\_2021\_paper.pdf}{Source PDF, p.\ 3}; ID: P326}\par
\vspace{2pt}
\noindent\begin{minipage}[t]{0.486\linewidth}
\centering\textbf{Original}\par\vspace{2pt}
\begin{minipage}[c][0.255\textheight][c]{\linewidth}\centering\includegraphics[width=\linewidth,height=0.255\textheight,keepaspectratio]{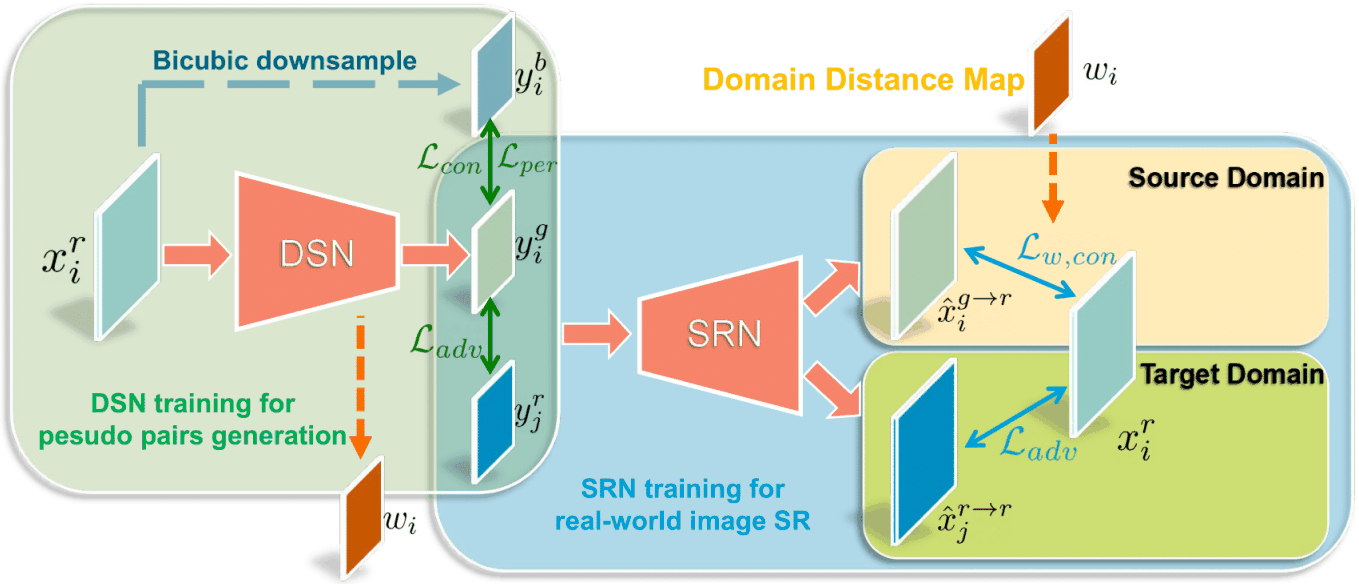}\end{minipage}\par
\vspace{3pt}{\scriptsize \setlength{\tabcolsep}{1.5pt}\begin{tabular*}{\linewidth}{@{\extracolsep{\fill}}rrrrrrr@{}}
\toprule
C & O & G & A & I & S & Mean \\
3 & 3 & 4 & 3 & 3 & 4 & 3.33\\
\bottomrule
\end{tabular*}
}
\end{minipage}\hfill
\begin{minipage}[t]{0.486\linewidth}
\centering\textbf{Study-generated derivative}\par\vspace{2pt}
\begin{minipage}[c][0.255\textheight][c]{\linewidth}\centering\includegraphics[width=\linewidth,height=0.255\textheight,keepaspectratio]{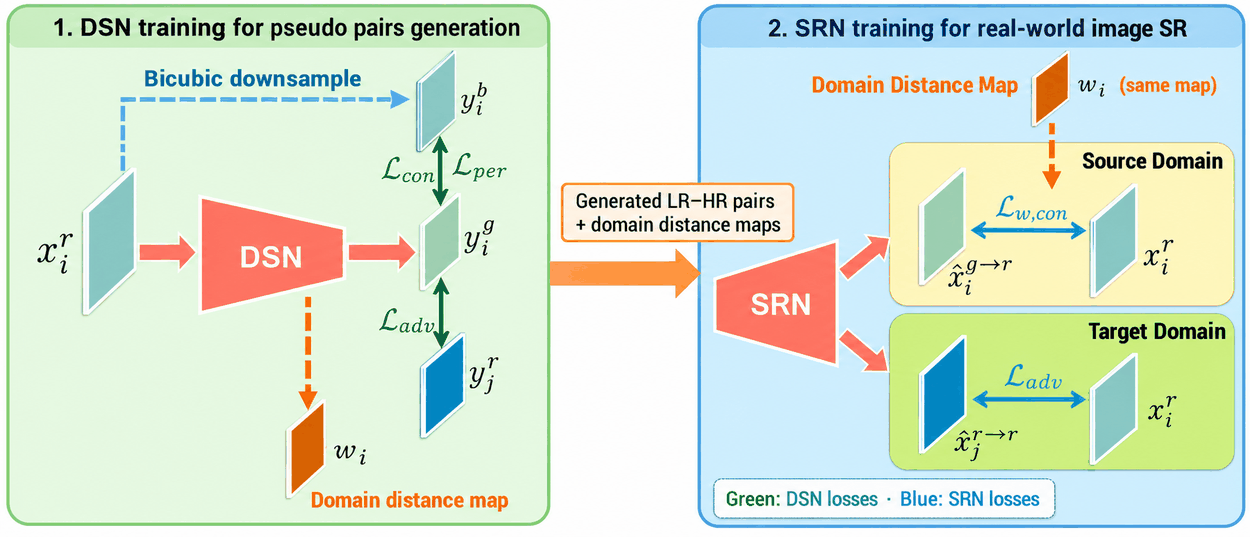}\end{minipage}\par
\vspace{3pt}
{\scriptsize \setlength{\tabcolsep}{1.5pt}\begin{tabular*}{\linewidth}{@{\extracolsep{\fill}}rrrrrrr@{}}
\toprule
C & O & G & A & I & S & Mean \\
4 (+1) & 5 (+2) & 4 (+0) & 3 (+0) & 3 (+0) & 4 (+0) & 3.83 (+0.50)\\
\bottomrule
\end{tabular*}
}
\end{minipage}\par
\vspace{3pt}
{\scriptsize Source public-date proxy: 2021-06. All displayed scores are medians of three runs; parentheses show revised minus original.\par}
\vfill
{\scriptsize C: clarity; O: organization/readability; G: graphical consistency; A: aesthetic coherence; I: visual inventiveness; S: design specificity. Mean is the arithmetic mean of six dimension medians, a descriptive display value, not a validated quality index.\par}

\clearpage
\phantomsection
\label{atlas:corpus034}
\pdfbookmark[1]{34. NeuralRecon: Real-Time Coherent 3D Reconstruction From Monocular Video}{corpus034}
{\Large\bfseries\color{academicblue} 034 \quad \href{https://openaccess.thecvf.com/content/CVPR2021/html/Sun_NeuralRecon_Real-Time_Coherent_3D_Reconstruction_From_Monocular_Video_CVPR_2021_paper.html}{NeuralRecon: Real-Time Coherent 3D Reconstruction From Monocular Video} \citep{corpus034}\par}
\vspace{3pt}
{\small S1 \enspace $\vert$ \enspace 3D vision \enspace $\vert$ \enspace CVPR 2021\par}
\vspace{4pt}
\vspace{5pt}
{\bfseries Motivation (source Fig.\ 1)} \hfill {\footnotesize \href{https://openaccess.thecvf.com/content/CVPR2021/papers/Sun\_NeuralRecon\_Real-Time\_Coherent\_3D\_Reconstruction\_From\_Monocular\_Video\_CVPR\_2021\_paper.pdf}{Source PDF, p.\ 1}; ID: P165}\par
\vspace{2pt}
\noindent\begin{minipage}[t]{0.486\linewidth}
\centering\textbf{Original}\par\vspace{2pt}
\begin{minipage}[c][0.255\textheight][c]{\linewidth}\centering\includegraphics[width=\linewidth,height=0.255\textheight,keepaspectratio]{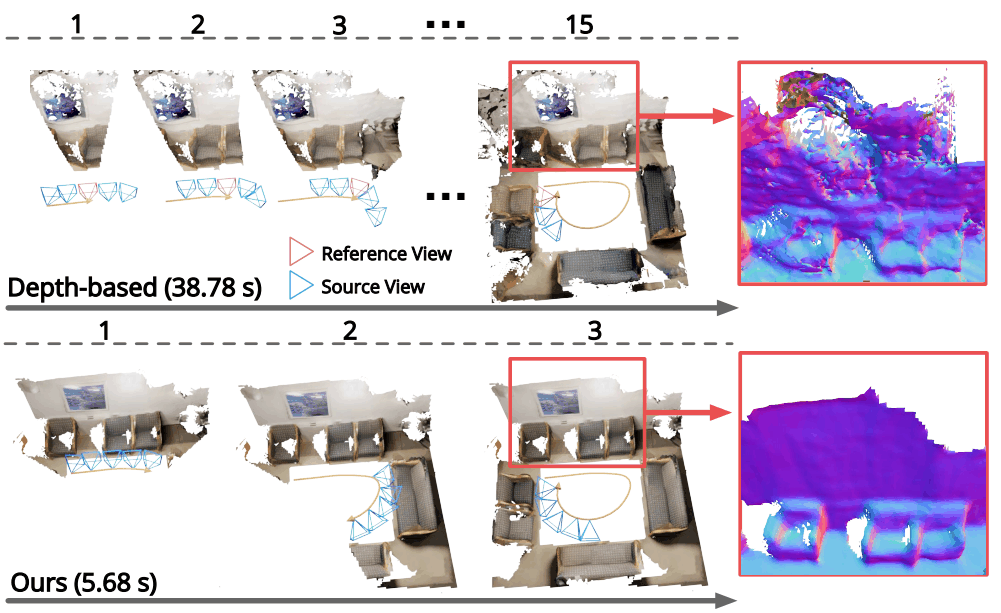}\end{minipage}\par
\vspace{3pt}{\scriptsize \setlength{\tabcolsep}{1.5pt}\begin{tabular*}{\linewidth}{@{\extracolsep{\fill}}rrrrrrr@{}}
\toprule
C & O & G & A & I & S & Mean \\
4 & 5 & 4 & 4 & 4 & 5 & 4.33\\
\bottomrule
\end{tabular*}
}
\end{minipage}\hfill
\begin{minipage}[t]{0.486\linewidth}
\centering\textbf{Study-generated derivative}\par\vspace{2pt}
\begin{minipage}[c][0.255\textheight][c]{\linewidth}\centering\includegraphics[width=\linewidth,height=0.255\textheight,keepaspectratio]{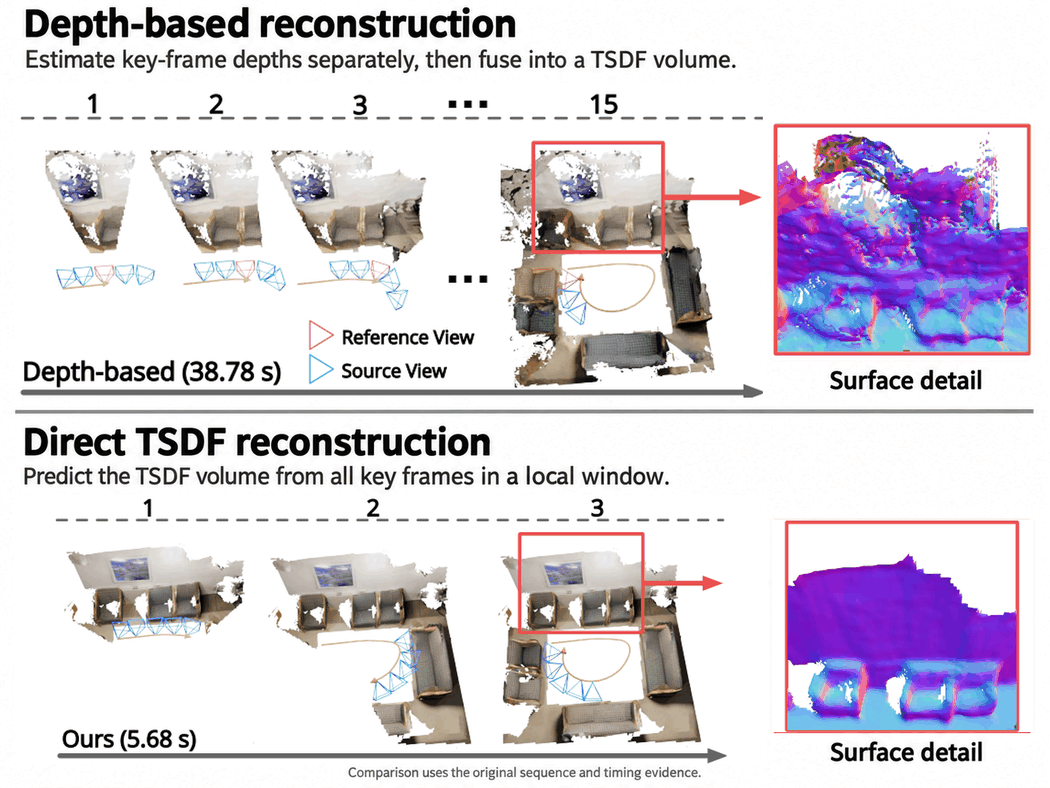}\end{minipage}\par
\vspace{3pt}
{\scriptsize \setlength{\tabcolsep}{1.5pt}\begin{tabular*}{\linewidth}{@{\extracolsep{\fill}}rrrrrrr@{}}
\toprule
C & O & G & A & I & S & Mean \\
5 (+1) & 4 (-1) & 4 (+0) & 4 (+0) & 4 (+0) & 5 (+0) & 4.33 (+0.00)\\
\bottomrule
\end{tabular*}
}
\end{minipage}\par
\vspace{3pt}
{\scriptsize Source public-date proxy: 2021-04-01. All displayed scores are medians of three runs; parentheses show revised minus original.\par}
\vspace{5pt}
{\bfseries Framework (source Fig.\ 2)} \hfill {\footnotesize \href{https://openaccess.thecvf.com/content/CVPR2021/papers/Sun\_NeuralRecon\_Real-Time\_Coherent\_3D\_Reconstruction\_From\_Monocular\_Video\_CVPR\_2021\_paper.pdf}{Source PDF, p.\ 3}; ID: P166}\par
\vspace{2pt}
\noindent\begin{minipage}[t]{0.486\linewidth}
\centering\textbf{Original}\par\vspace{2pt}
\begin{minipage}[c][0.255\textheight][c]{\linewidth}\centering\includegraphics[width=\linewidth,height=0.255\textheight,keepaspectratio]{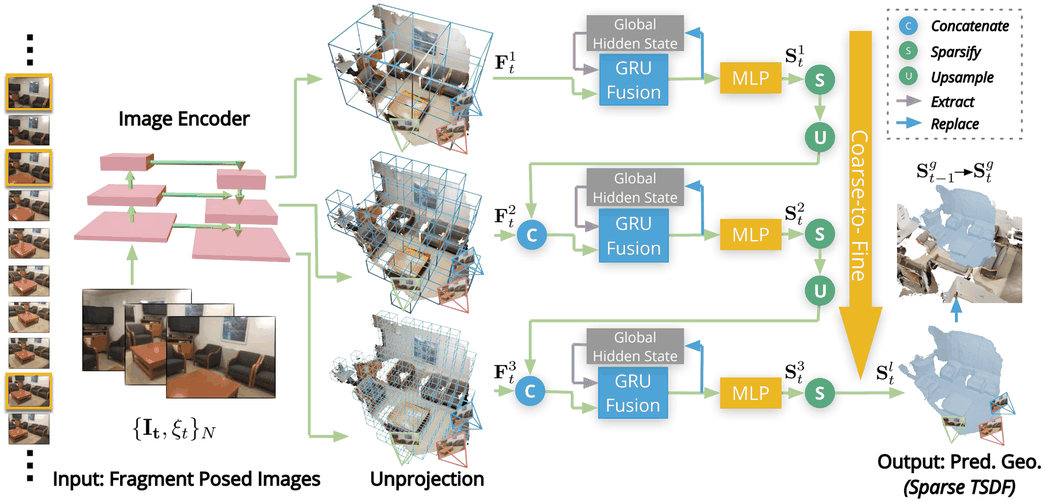}\end{minipage}\par
\vspace{3pt}{\scriptsize \setlength{\tabcolsep}{1.5pt}\begin{tabular*}{\linewidth}{@{\extracolsep{\fill}}rrrrrrr@{}}
\toprule
C & O & G & A & I & S & Mean \\
4 & 4 & 5 & 4 & 4 & 5 & 4.33\\
\bottomrule
\end{tabular*}
}
\end{minipage}\hfill
\begin{minipage}[t]{0.486\linewidth}
\centering\textbf{Study-generated derivative}\par\vspace{2pt}
\begin{minipage}[c][0.255\textheight][c]{\linewidth}\centering\includegraphics[width=\linewidth,height=0.255\textheight,keepaspectratio]{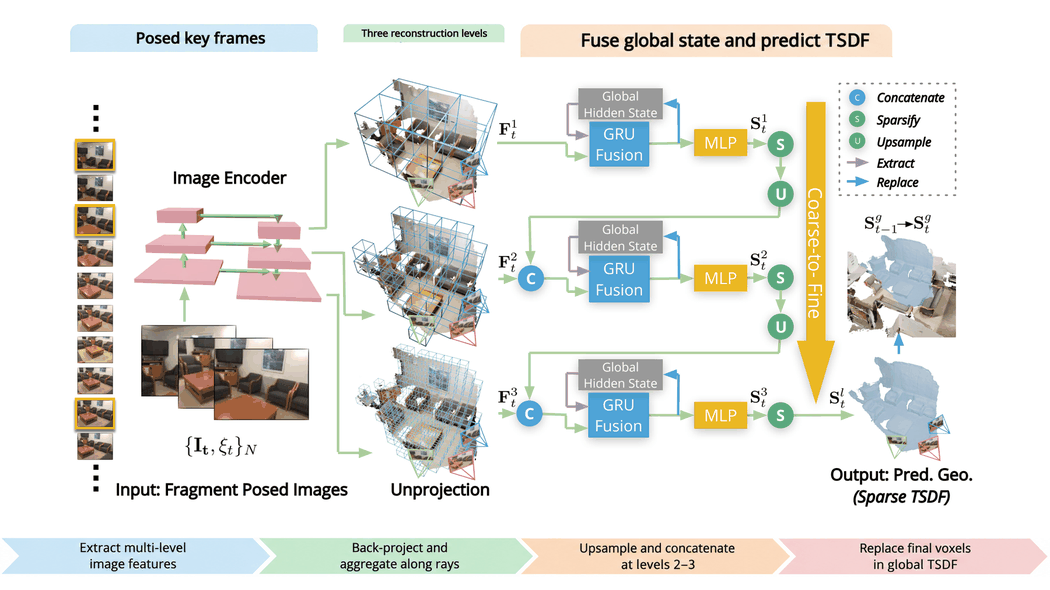}\end{minipage}\par
\vspace{3pt}
{\scriptsize \setlength{\tabcolsep}{1.5pt}\begin{tabular*}{\linewidth}{@{\extracolsep{\fill}}rrrrrrr@{}}
\toprule
C & O & G & A & I & S & Mean \\
4 (+0) & 4 (+0) & 5 (+0) & 4 (+0) & 4 (+0) & 5 (+0) & 4.33 (+0.00)\\
\bottomrule
\end{tabular*}
}
\end{minipage}\par
\vspace{3pt}
{\scriptsize Source public-date proxy: 2021-04-01. All displayed scores are medians of three runs; parentheses show revised minus original.\par}
\vfill
{\scriptsize C: clarity; O: organization/readability; G: graphical consistency; A: aesthetic coherence; I: visual inventiveness; S: design specificity. Mean is the arithmetic mean of six dimension medians, a descriptive display value, not a validated quality index.\par}

\clearpage
\phantomsection
\label{atlas:corpus035}
\pdfbookmark[1]{35. Occupancy Networks: Learning 3D Reconstruction in Function Space}{corpus035}
{\Large\bfseries\color{academicblue} 035 \quad \href{https://openaccess.thecvf.com/content\_CVPR\_2019/html/Mescheder\_Occupancy\_Networks\_Learning\_3D\_Reconstruction\_in\_Function\_Space\_CVPR\_2019\_paper.html}{Occupancy Networks: Learning 3D Reconstruction in Function Space} \citep{corpus035}\par}
\vspace{3pt}
{\small S1 \enspace $\vert$ \enspace 3D vision \enspace $\vert$ \enspace CVPR 2019\par}
\vspace{4pt}
\vspace{5pt}
{\bfseries Motivation (source Fig.\ 1)} \hfill {\footnotesize \href{https://openaccess.thecvf.com/content\_CVPR\_2019/papers/Mescheder\_Occupancy\_Networks\_Learning\_3D\_Reconstruction\_in\_Function\_Space\_CVPR\_2019\_paper.pdf}{Source PDF, p.\ 1}; ID: P167}\par
\vspace{2pt}
\noindent\begin{minipage}[t]{0.486\linewidth}
\centering\textbf{Original}\par\vspace{2pt}
\begin{minipage}[c][0.255\textheight][c]{\linewidth}\centering\includegraphics[width=\linewidth,height=0.255\textheight,keepaspectratio]{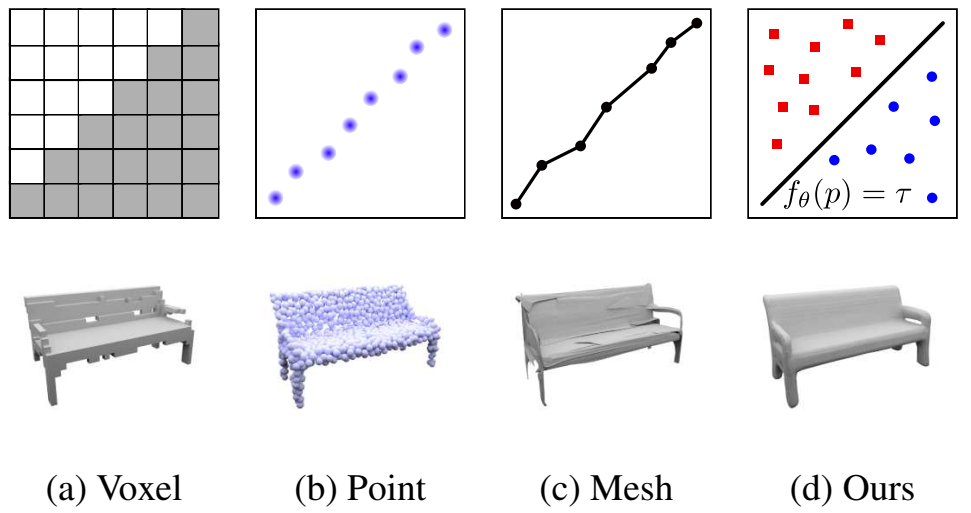}\end{minipage}\par
\vspace{3pt}{\scriptsize \setlength{\tabcolsep}{1.5pt}\begin{tabular*}{\linewidth}{@{\extracolsep{\fill}}rrrrrrr@{}}
\toprule
C & O & G & A & I & S & Mean \\
4 & 5 & 4 & 4 & 4 & 4 & 4.17\\
\bottomrule
\end{tabular*}
}
\end{minipage}\hfill
\begin{minipage}[t]{0.486\linewidth}
\centering\textbf{Study-generated derivative}\par\vspace{2pt}
\begin{minipage}[c][0.255\textheight][c]{\linewidth}\centering\includegraphics[width=\linewidth,height=0.255\textheight,keepaspectratio]{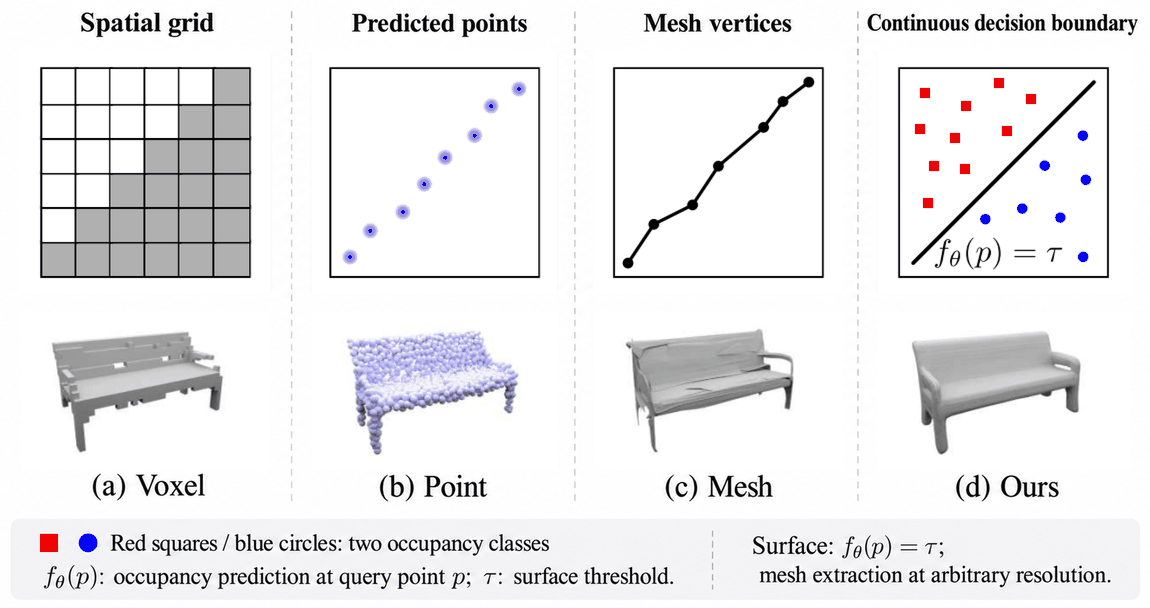}\end{minipage}\par
\vspace{3pt}
{\scriptsize \setlength{\tabcolsep}{1.5pt}\begin{tabular*}{\linewidth}{@{\extracolsep{\fill}}rrrrrrr@{}}
\toprule
C & O & G & A & I & S & Mean \\
4 (+0) & 5 (+0) & 5 (+1) & 4 (+0) & 4 (+0) & 5 (+1) & 4.50 (+0.33)\\
\bottomrule
\end{tabular*}
}
\end{minipage}\par
\vspace{3pt}
{\scriptsize Source public-date proxy: 2019-04-30. All displayed scores are medians of three runs; parentheses show revised minus original.\par}
\vspace{5pt}
{\bfseries Framework (source Fig.\ 2)} \hfill {\footnotesize \href{https://openaccess.thecvf.com/content\_CVPR\_2019/papers/Mescheder\_Occupancy\_Networks\_Learning\_3D\_Reconstruction\_in\_Function\_Space\_CVPR\_2019\_paper.pdf}{Source PDF, p.\ 4}; ID: P168}\par
\vspace{2pt}
\noindent\begin{minipage}[t]{0.486\linewidth}
\centering\textbf{Original}\par\vspace{2pt}
\begin{minipage}[c][0.255\textheight][c]{\linewidth}\centering\includegraphics[width=\linewidth,height=0.255\textheight,keepaspectratio]{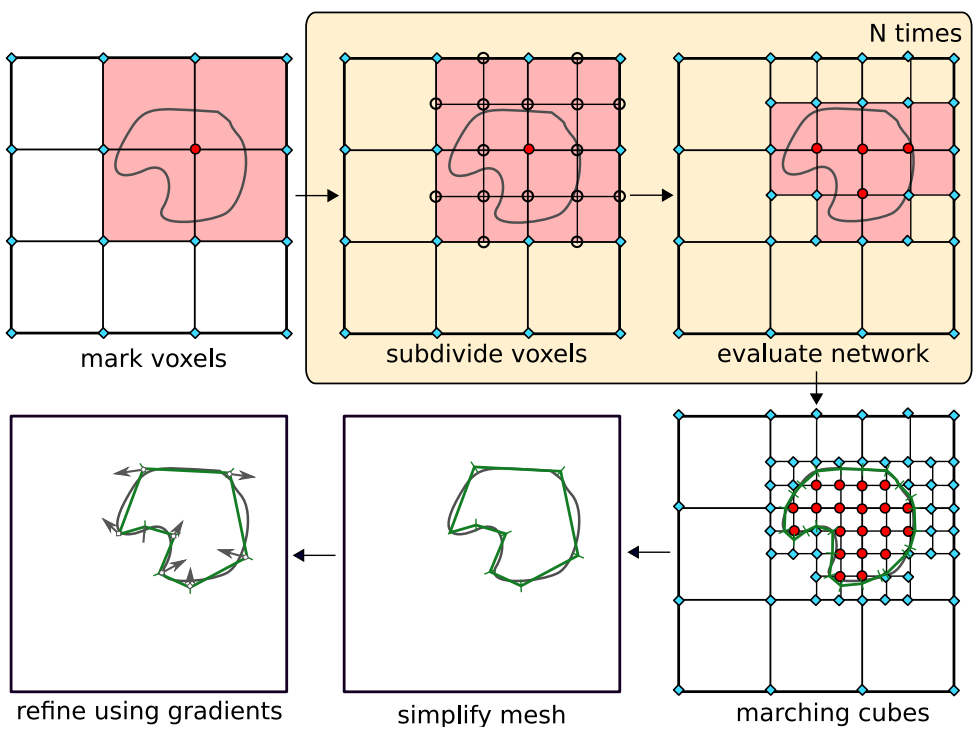}\end{minipage}\par
\vspace{3pt}{\scriptsize \setlength{\tabcolsep}{1.5pt}\begin{tabular*}{\linewidth}{@{\extracolsep{\fill}}rrrrrrr@{}}
\toprule
C & O & G & A & I & S & Mean \\
4 & 5 & 4 & 4 & 4 & 5 & 4.33\\
\bottomrule
\end{tabular*}
}
\end{minipage}\hfill
\begin{minipage}[t]{0.486\linewidth}
\centering\textbf{Study-generated derivative}\par\vspace{2pt}
\begin{minipage}[c][0.255\textheight][c]{\linewidth}\centering\includegraphics[width=\linewidth,height=0.255\textheight,keepaspectratio]{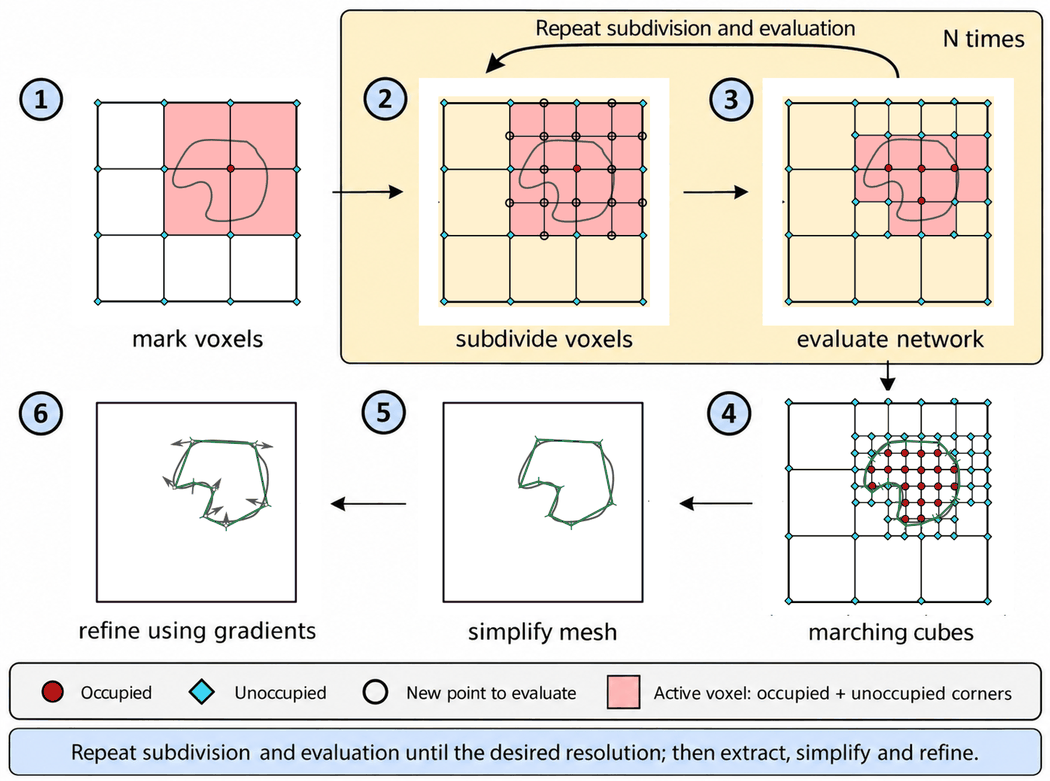}\end{minipage}\par
\vspace{3pt}
{\scriptsize \setlength{\tabcolsep}{1.5pt}\begin{tabular*}{\linewidth}{@{\extracolsep{\fill}}rrrrrrr@{}}
\toprule
C & O & G & A & I & S & Mean \\
4 (+0) & 5 (+0) & 4 (+0) & 4 (+0) & 4 (+0) & 5 (+0) & 4.33 (+0.00)\\
\bottomrule
\end{tabular*}
}
\end{minipage}\par
\vspace{3pt}
{\scriptsize Source public-date proxy: 2018-12-10. All displayed scores are medians of three runs; parentheses show revised minus original.\par}
\vfill
{\scriptsize C: clarity; O: organization/readability; G: graphical consistency; A: aesthetic coherence; I: visual inventiveness; S: design specificity. Mean is the arithmetic mean of six dimension medians, a descriptive display value, not a validated quality index.\par}

\clearpage
\phantomsection
\label{atlas:corpus042}
\pdfbookmark[1]{42. Efficient Test-time Adaptive Object Detection via Sensitivity-Guided Pruning}{corpus042}
{\Large\bfseries\color{academicblue} 042 \quad \href{https://openaccess.thecvf.com/content/CVPR2025/html/Wang\_Efficient\_Test-time\_Adaptive\_Object\_Detection\_via\_Sensitivity-Guided\_Pruning\_CVPR\_2025\_paper.html}{Efficient Test-time Adaptive Object Detection via Sensitivity-Guided Pruning} \citep{corpus042}\par}
\vspace{3pt}
{\small S2 \enspace $\vert$ \enspace Object detection \enspace $\vert$ \enspace CVPR 2025\par}
\vspace{4pt}
\vspace{5pt}
{\bfseries Motivation (source Fig.\ 1)} \hfill {\footnotesize \href{https://openaccess.thecvf.com/content/CVPR2025/papers/Wang\_Efficient\_Test-time\_Adaptive\_Object\_Detection\_via\_Sensitivity-Guided\_Pruning\_CVPR\_2025\_paper.pdf}{Source PDF, p.\ 1}; ID: P83}\par
\vspace{2pt}
\noindent\begin{minipage}[t]{0.486\linewidth}
\centering\textbf{Original}\par\vspace{2pt}
\begin{minipage}[c][0.255\textheight][c]{\linewidth}\centering\includegraphics[width=\linewidth,height=0.255\textheight,keepaspectratio]{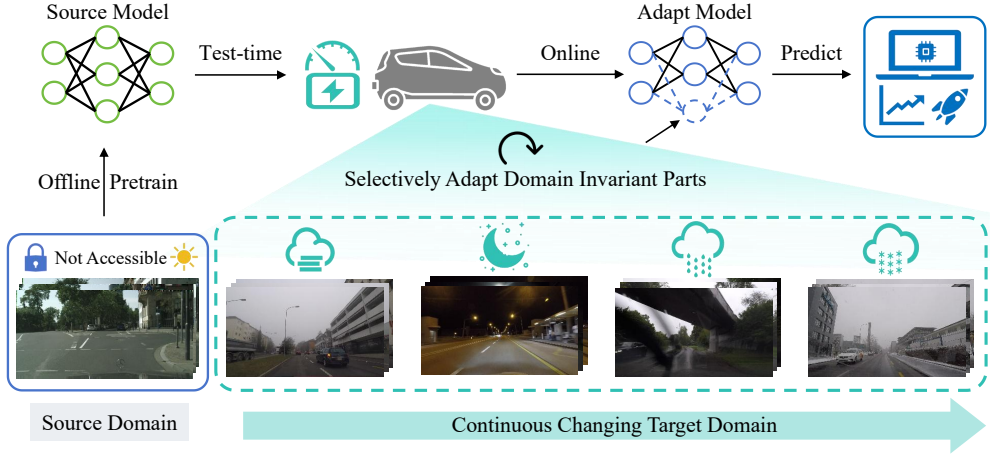}\end{minipage}\par
\vspace{3pt}{\scriptsize \setlength{\tabcolsep}{1.5pt}\begin{tabular*}{\linewidth}{@{\extracolsep{\fill}}rrrrrrr@{}}
\toprule
C & O & G & A & I & S & Mean \\
4 & 5 & 4 & 4 & 4 & 4 & 4.17\\
\bottomrule
\end{tabular*}
}
\end{minipage}\hfill
\begin{minipage}[t]{0.486\linewidth}
\centering\textbf{Study-generated derivative}\par\vspace{2pt}
\begin{minipage}[c][0.255\textheight][c]{\linewidth}\centering\includegraphics[width=\linewidth,height=0.255\textheight,keepaspectratio]{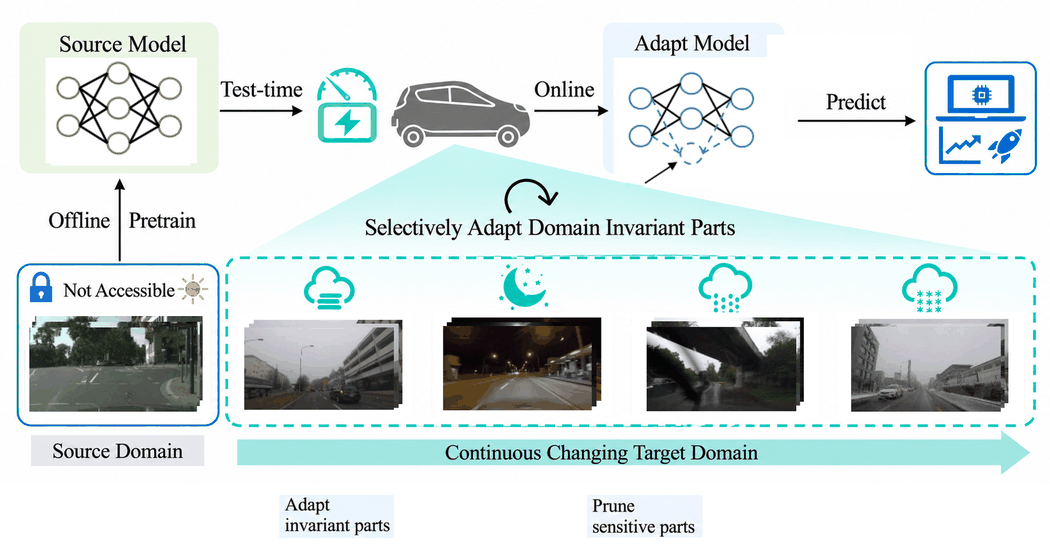}\end{minipage}\par
\vspace{3pt}
{\scriptsize \setlength{\tabcolsep}{1.5pt}\begin{tabular*}{\linewidth}{@{\extracolsep{\fill}}rrrrrrr@{}}
\toprule
C & O & G & A & I & S & Mean \\
3 (-1) & 3 (-2) & 3 (-1) & 3 (-1) & 3 (-1) & 4 (+0) & 3.17 (-1.00)\\
\bottomrule
\end{tabular*}
}
\end{minipage}\par
\vspace{3pt}
{\scriptsize Source public-date proxy: 2025-06-03. All displayed scores are medians of three runs; parentheses show revised minus original.\par}
\vspace{5pt}
{\bfseries Framework (source Fig.\ 4)} \hfill {\footnotesize \href{https://openaccess.thecvf.com/content/CVPR2025/papers/Wang\_Efficient\_Test-time\_Adaptive\_Object\_Detection\_via\_Sensitivity-Guided\_Pruning\_CVPR\_2025\_paper.pdf}{Source PDF, p.\ 3}; ID: P84}\par
\vspace{2pt}
\noindent\begin{minipage}[t]{0.486\linewidth}
\centering\textbf{Original}\par\vspace{2pt}
\begin{minipage}[c][0.255\textheight][c]{\linewidth}\centering\includegraphics[width=\linewidth,height=0.255\textheight,keepaspectratio]{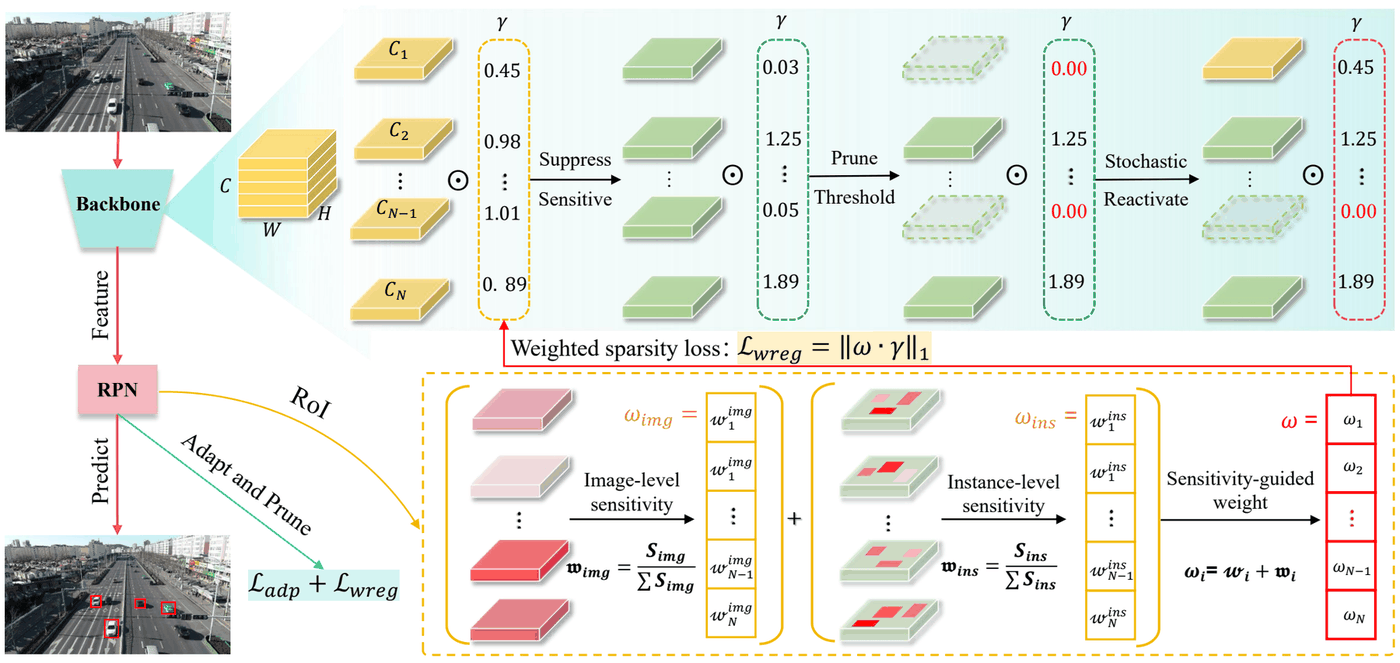}\end{minipage}\par
\vspace{3pt}{\scriptsize \setlength{\tabcolsep}{1.5pt}\begin{tabular*}{\linewidth}{@{\extracolsep{\fill}}rrrrrrr@{}}
\toprule
C & O & G & A & I & S & Mean \\
4 & 3 & 3 & 3 & 4 & 5 & 3.67\\
\bottomrule
\end{tabular*}
}
\end{minipage}\hfill
\begin{minipage}[t]{0.486\linewidth}
\centering\textbf{Study-generated derivative}\par\vspace{2pt}
\begin{minipage}[c][0.255\textheight][c]{\linewidth}\centering\includegraphics[width=\linewidth,height=0.255\textheight,keepaspectratio]{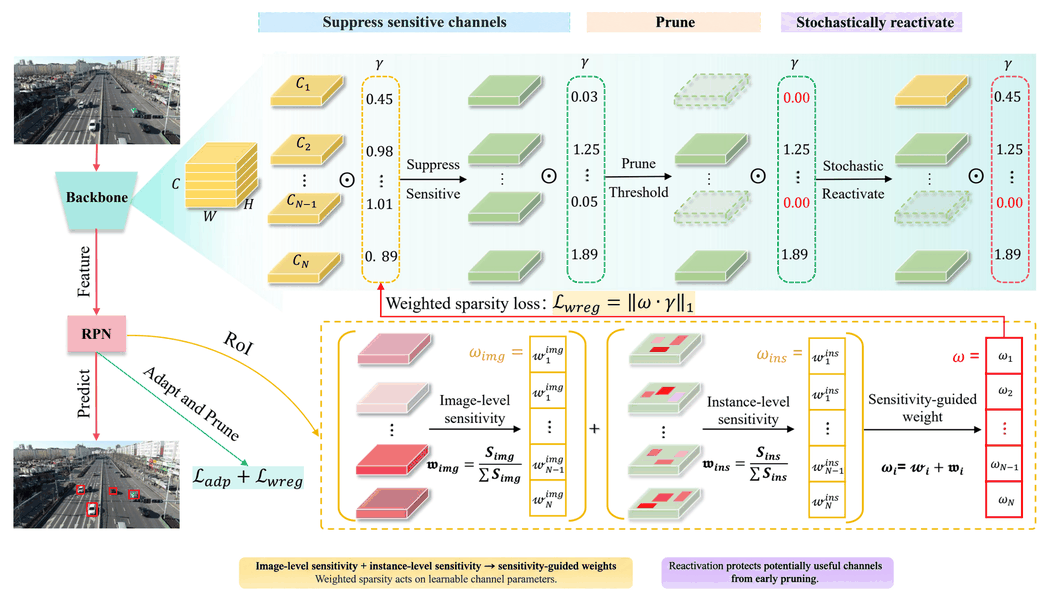}\end{minipage}\par
\vspace{3pt}
{\scriptsize \setlength{\tabcolsep}{1.5pt}\begin{tabular*}{\linewidth}{@{\extracolsep{\fill}}rrrrrrr@{}}
\toprule
C & O & G & A & I & S & Mean \\
4 (+0) & 3 (+0) & 3 (+0) & 3 (+0) & 4 (+0) & 5 (+0) & 3.67 (+0.00)\\
\bottomrule
\end{tabular*}
}
\end{minipage}\par
\vspace{3pt}
{\scriptsize Source public-date proxy: 2025-06-03. All displayed scores are medians of three runs; parentheses show revised minus original.\par}
\vfill
{\scriptsize C: clarity; O: organization/readability; G: graphical consistency; A: aesthetic coherence; I: visual inventiveness; S: design specificity. Mean is the arithmetic mean of six dimension medians, a descriptive display value, not a validated quality index.\par}

\clearpage
\phantomsection
\label{atlas:corpus043}
\pdfbookmark[1]{43. Fractal Calibration for Long-tailed Object Detection}{corpus043}
{\Large\bfseries\color{academicblue} 043 \quad \href{https://openaccess.thecvf.com/content/CVPR2025/html/Alexandridis_Fractal_Calibration_for_Long-tailed_Object_Detection_CVPR_2025_paper.html}{Fractal Calibration for Long-tailed Object Detection} \citep{corpus043}\par}
\vspace{3pt}
{\small S2 \enspace $\vert$ \enspace Object detection \enspace $\vert$ \enspace CVPR 2025\par}
\vspace{4pt}
\vspace{5pt}
{\bfseries Motivation (source Fig.\ 1)} \hfill {\footnotesize \href{https://openaccess.thecvf.com/content/CVPR2025/papers/Alexandridis\_Fractal\_Calibration\_for\_Long-tailed\_Object\_Detection\_CVPR\_2025\_paper.pdf}{Source PDF, p.\ 1}; ID: P233}\par
\vspace{2pt}
\noindent\begin{minipage}[t]{0.486\linewidth}
\centering\textbf{Original}\par\vspace{2pt}
\begin{minipage}[c][0.255\textheight][c]{\linewidth}\centering\includegraphics[width=\linewidth,height=0.255\textheight,keepaspectratio]{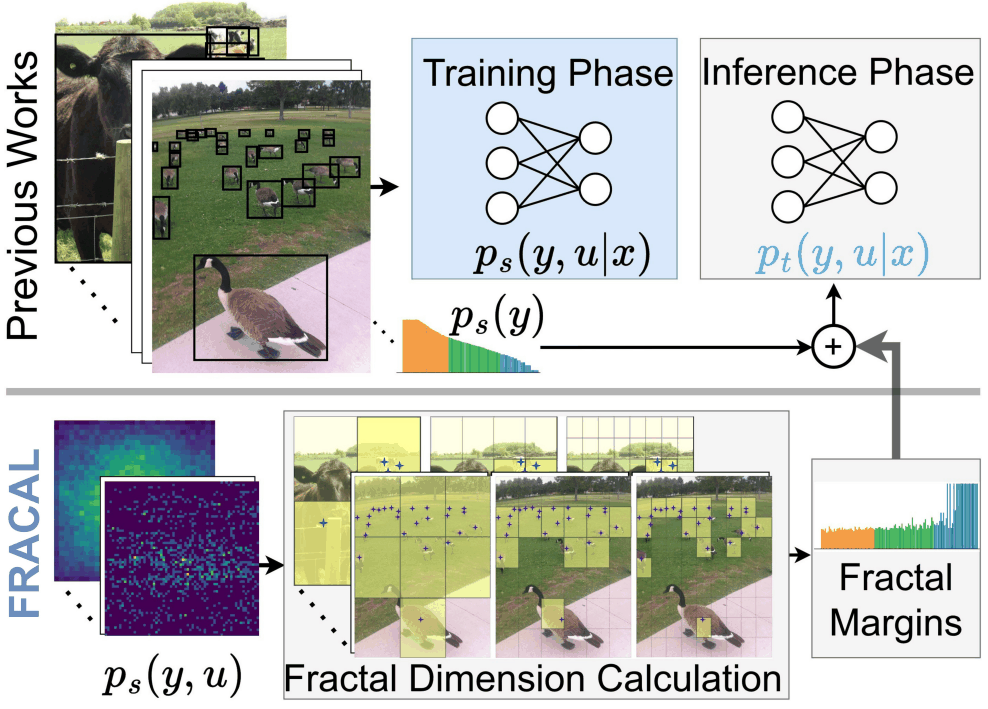}\end{minipage}\par
\vspace{3pt}{\scriptsize \setlength{\tabcolsep}{1.5pt}\begin{tabular*}{\linewidth}{@{\extracolsep{\fill}}rrrrrrr@{}}
\toprule
C & O & G & A & I & S & Mean \\
3 & 4 & 3 & 3 & 4 & 4 & 3.50\\
\bottomrule
\end{tabular*}
}
\end{minipage}\hfill
\begin{minipage}[t]{0.486\linewidth}
\centering\textbf{Study-generated derivative}\par\vspace{2pt}
\begin{minipage}[c][0.255\textheight][c]{\linewidth}\centering\includegraphics[width=\linewidth,height=0.255\textheight,keepaspectratio]{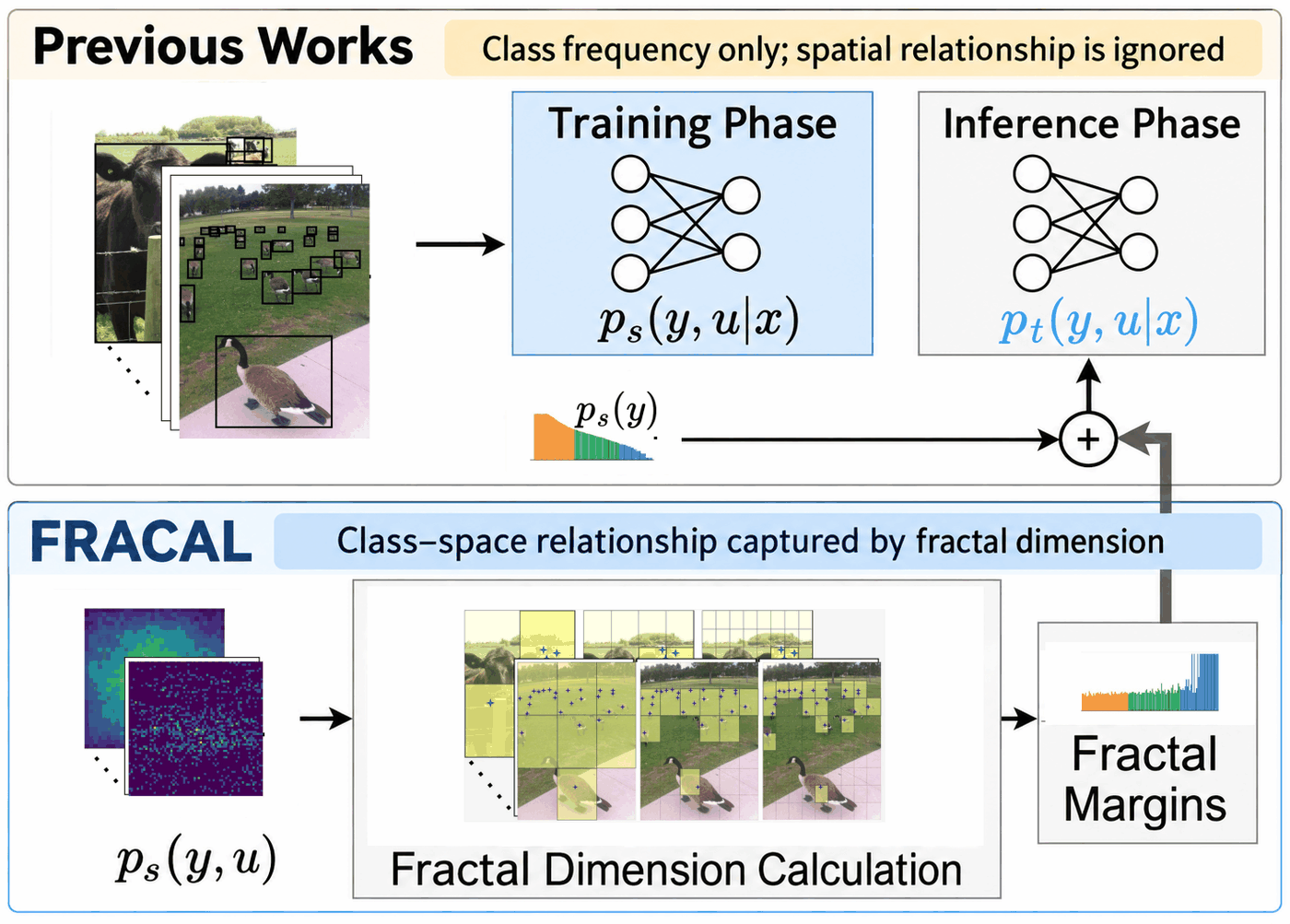}\end{minipage}\par
\vspace{3pt}
{\scriptsize \setlength{\tabcolsep}{1.5pt}\begin{tabular*}{\linewidth}{@{\extracolsep{\fill}}rrrrrrr@{}}
\toprule
C & O & G & A & I & S & Mean \\
4 (+1) & 4 (+0) & 3 (+0) & 3 (+0) & 4 (+0) & 5 (+1) & 3.83 (+0.33)\\
\bottomrule
\end{tabular*}
}
\end{minipage}\par
\vspace{3pt}
{\scriptsize Source public-date proxy: 2025-03-05. All displayed scores are medians of three runs; parentheses show revised minus original.\par}
\vspace{5pt}
{\bfseries Framework (source Fig.\ 2)} \hfill {\footnotesize \href{https://openaccess.thecvf.com/content/CVPR2025/papers/Alexandridis\_Fractal\_Calibration\_for\_Long-tailed\_Object\_Detection\_CVPR\_2025\_paper.pdf}{Source PDF, p.\ 3}; ID: P234}\par
\vspace{2pt}
\noindent\begin{minipage}[t]{0.486\linewidth}
\centering\textbf{Original}\par\vspace{2pt}
\begin{minipage}[c][0.255\textheight][c]{\linewidth}\centering\includegraphics[width=\linewidth,height=0.255\textheight,keepaspectratio]{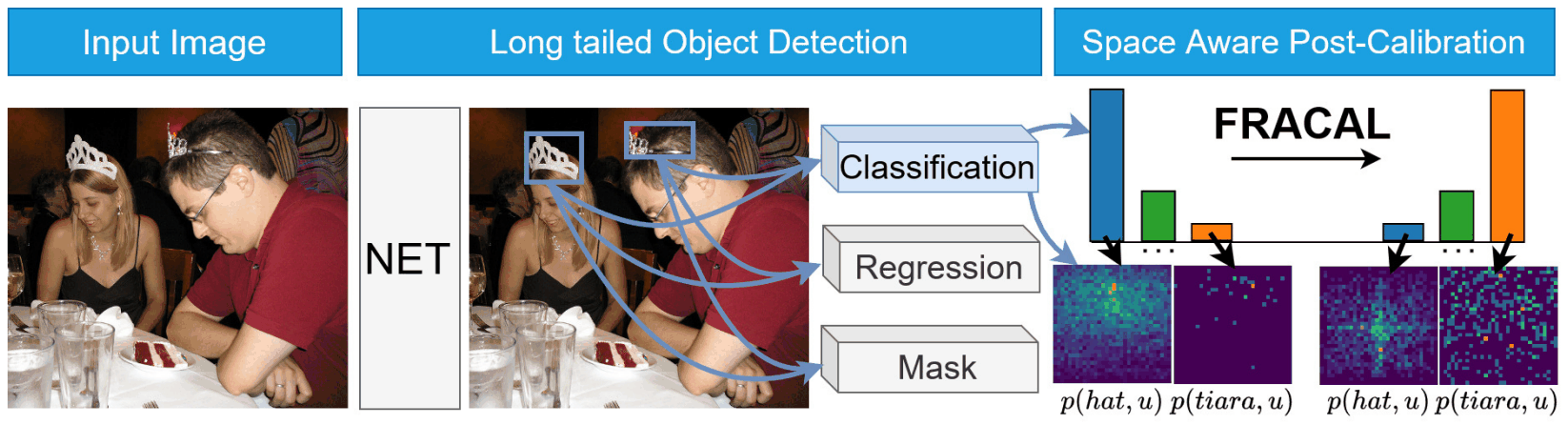}\end{minipage}\par
\vspace{3pt}{\scriptsize \setlength{\tabcolsep}{1.5pt}\begin{tabular*}{\linewidth}{@{\extracolsep{\fill}}rrrrrrr@{}}
\toprule
C & O & G & A & I & S & Mean \\
3 & 4 & 3 & 3 & 4 & 4 & 3.50\\
\bottomrule
\end{tabular*}
}
\end{minipage}\hfill
\begin{minipage}[t]{0.486\linewidth}
\centering\textbf{Study-generated derivative}\par\vspace{2pt}
\begin{minipage}[c][0.255\textheight][c]{\linewidth}\centering\includegraphics[width=\linewidth,height=0.255\textheight,keepaspectratio]{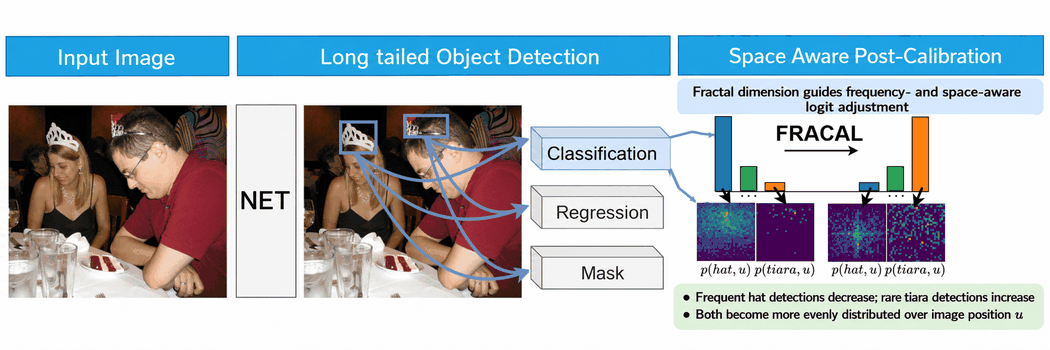}\end{minipage}\par
\vspace{3pt}
{\scriptsize \setlength{\tabcolsep}{1.5pt}\begin{tabular*}{\linewidth}{@{\extracolsep{\fill}}rrrrrrr@{}}
\toprule
C & O & G & A & I & S & Mean \\
4 (+1) & 4 (+0) & 3 (+0) & 3 (+0) & 4 (+0) & 4 (+0) & 3.67 (+0.17)\\
\bottomrule
\end{tabular*}
}
\end{minipage}\par
\vspace{3pt}
{\scriptsize Source public-date proxy: 2024-10-15. All displayed scores are medians of three runs; parentheses show revised minus original.\par}
\vfill
{\scriptsize C: clarity; O: organization/readability; G: graphical consistency; A: aesthetic coherence; I: visual inventiveness; S: design specificity. Mean is the arithmetic mean of six dimension medians, a descriptive display value, not a validated quality index.\par}

\clearpage
\phantomsection
\label{atlas:corpus052}
\pdfbookmark[1]{52. DFormerv2: Geometry Self-Attention for RGBD Semantic Segmentation}{corpus052}
{\Large\bfseries\color{academicblue} 052 \quad \href{https://openaccess.thecvf.com/content/CVPR2025/html/Yin\_DFormerv2\_Geometry\_Self-Attention\_for\_RGBD\_Semantic\_Segmentation\_CVPR\_2025\_paper.html}{DFormerv2: Geometry Self-Attention for RGBD Semantic Segmentation} \citep{corpus052}\par}
\vspace{3pt}
{\small S2 \enspace $\vert$ \enspace Image segmentation \enspace $\vert$ \enspace CVPR 2025\par}
\vspace{4pt}
\vspace{5pt}
{\bfseries Motivation (source Fig.\ 1)} \hfill {\footnotesize \href{https://openaccess.thecvf.com/content/CVPR2025/papers/Yin\_DFormerv2\_Geometry\_Self-Attention\_for\_RGBD\_Semantic\_Segmentation\_CVPR\_2025\_paper.pdf}{Source PDF, p.\ 1}; ID: P179}\par
\vspace{2pt}
\noindent\begin{minipage}[t]{0.486\linewidth}
\centering\textbf{Original}\par\vspace{2pt}
\begin{minipage}[c][0.255\textheight][c]{\linewidth}\centering\includegraphics[width=\linewidth,height=0.255\textheight,keepaspectratio]{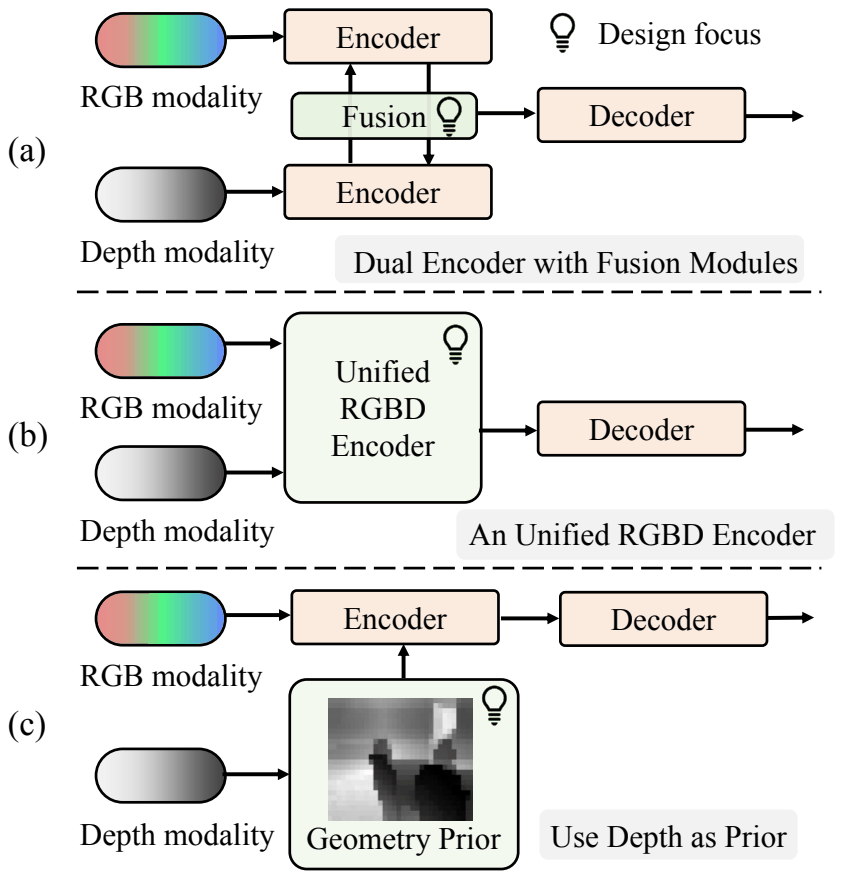}\end{minipage}\par
\vspace{3pt}{\scriptsize \setlength{\tabcolsep}{1.5pt}\begin{tabular*}{\linewidth}{@{\extracolsep{\fill}}rrrrrrr@{}}
\toprule
C & O & G & A & I & S & Mean \\
4 & 5 & 4 & 4 & 3 & 4 & 4.00\\
\bottomrule
\end{tabular*}
}
\end{minipage}\hfill
\begin{minipage}[t]{0.486\linewidth}
\centering\textbf{Study-generated derivative}\par\vspace{2pt}
\begin{minipage}[c][0.255\textheight][c]{\linewidth}\centering\includegraphics[width=\linewidth,height=0.255\textheight,keepaspectratio]{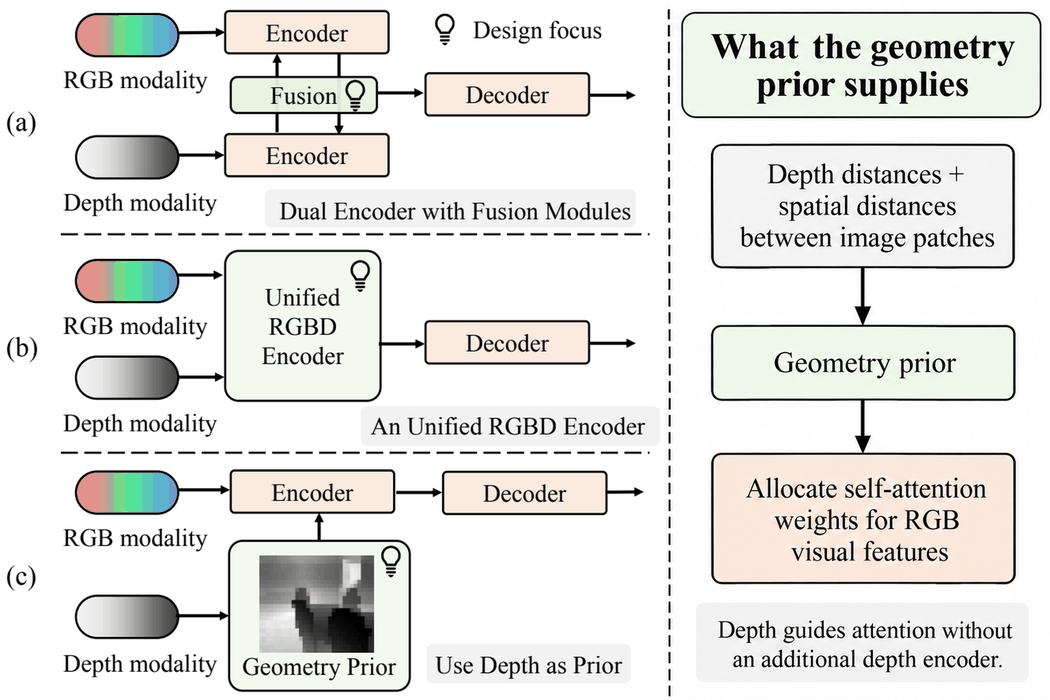}\end{minipage}\par
\vspace{3pt}
{\scriptsize \setlength{\tabcolsep}{1.5pt}\begin{tabular*}{\linewidth}{@{\extracolsep{\fill}}rrrrrrr@{}}
\toprule
C & O & G & A & I & S & Mean \\
4 (+0) & 5 (+0) & 4 (+0) & 4 (+0) & 3 (+0) & 4 (+0) & 4.00 (+0.00)\\
\bottomrule
\end{tabular*}
}
\end{minipage}\par
\vspace{3pt}
{\scriptsize Source public-date proxy: 2025-04-07. All displayed scores are medians of three runs; parentheses show revised minus original.\par}
\vspace{5pt}
{\bfseries Framework (source Fig.\ 4)} \hfill {\footnotesize \href{https://openaccess.thecvf.com/content/CVPR2025/papers/Yin\_DFormerv2\_Geometry\_Self-Attention\_for\_RGBD\_Semantic\_Segmentation\_CVPR\_2025\_paper.pdf}{Source PDF, p.\ 4}; ID: P180}\par
\vspace{2pt}
\noindent\begin{minipage}[t]{0.486\linewidth}
\centering\textbf{Original}\par\vspace{2pt}
\begin{minipage}[c][0.255\textheight][c]{\linewidth}\centering\includegraphics[width=\linewidth,height=0.255\textheight,keepaspectratio]{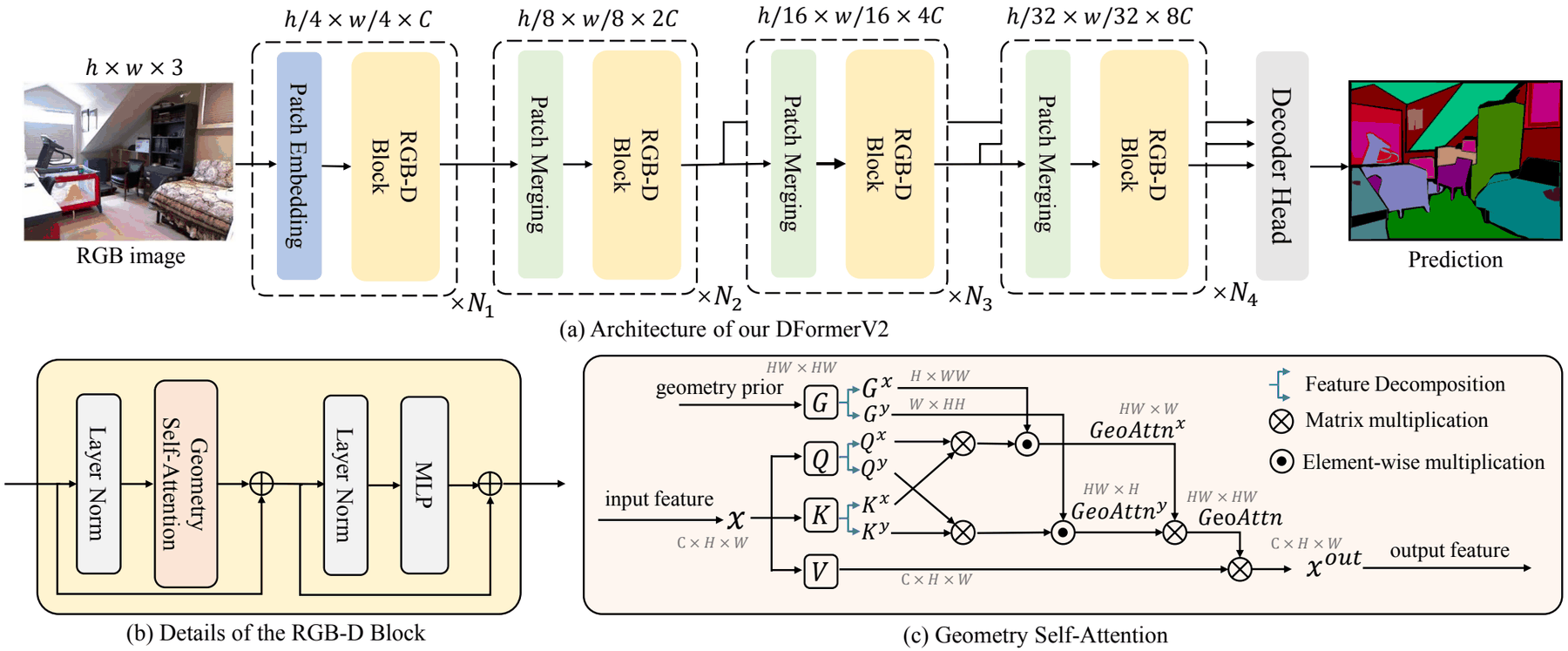}\end{minipage}\par
\vspace{3pt}{\scriptsize \setlength{\tabcolsep}{1.5pt}\begin{tabular*}{\linewidth}{@{\extracolsep{\fill}}rrrrrrr@{}}
\toprule
C & O & G & A & I & S & Mean \\
4 & 4 & 4 & 4 & 3 & 5 & 4.00\\
\bottomrule
\end{tabular*}
}
\end{minipage}\hfill
\begin{minipage}[t]{0.486\linewidth}
\centering\textbf{Study-generated derivative}\par\vspace{2pt}
\begin{minipage}[c][0.255\textheight][c]{\linewidth}\centering\includegraphics[width=\linewidth,height=0.255\textheight,keepaspectratio]{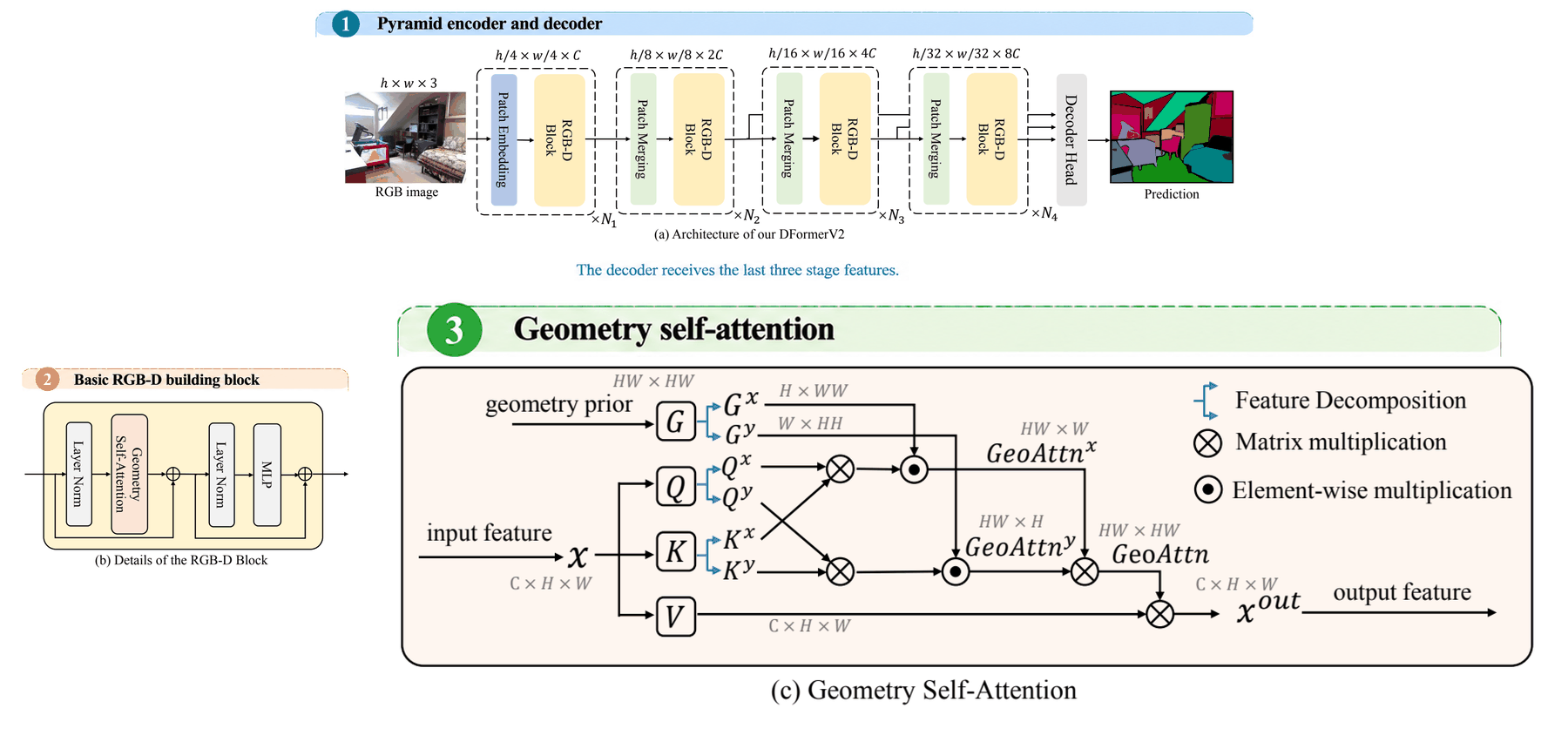}\end{minipage}\par
\vspace{3pt}
{\scriptsize \setlength{\tabcolsep}{1.5pt}\begin{tabular*}{\linewidth}{@{\extracolsep{\fill}}rrrrrrr@{}}
\toprule
C & O & G & A & I & S & Mean \\
4 (+0) & 3 (-1) & 4 (+0) & 3 (-1) & 3 (+0) & 5 (+0) & 3.67 (-0.33)\\
\bottomrule
\end{tabular*}
}
\end{minipage}\par
\vspace{3pt}
{\scriptsize Source public-date proxy: 2025-04-07. All displayed scores are medians of three runs; parentheses show revised minus original.\par}
\vfill
{\scriptsize C: clarity; O: organization/readability; G: graphical consistency; A: aesthetic coherence; I: visual inventiveness; S: design specificity. Mean is the arithmetic mean of six dimension medians, a descriptive display value, not a validated quality index.\par}

\clearpage
\phantomsection
\label{atlas:corpus053}
\pdfbookmark[1]{53. Distilling Spectral Graph for Object-Context Aware Open-Vocabulary Semantic Segmentation}{corpus053}
{\Large\bfseries\color{academicblue} 053 \quad \href{https://openaccess.thecvf.com/content/CVPR2025/html/Kim\_Distilling\_Spectral\_Graph\_for\_Object-Context\_Aware\_Open-Vocabulary\_Semantic\_Segmentation\_CVPR\_2025\_paper.html}{Distilling Spectral Graph for Object-Context Aware Open-Vocabulary Semantic Segmentation} \citep{corpus053}\par}
\vspace{3pt}
{\small S2 \enspace $\vert$ \enspace Image segmentation \enspace $\vert$ \enspace CVPR 2025\par}
\vspace{4pt}
\vspace{5pt}
{\bfseries Motivation (source Fig.\ 3)} \hfill {\footnotesize \href{https://openaccess.thecvf.com/content/CVPR2025/papers/Kim\_Distilling\_Spectral\_Graph\_for\_Object-Context\_Aware\_Open-Vocabulary\_Semantic\_Segmentation\_CVPR\_2025\_paper.pdf}{Source PDF, p.\ 4}; ID: P127}\par
\vspace{2pt}
\noindent\begin{minipage}[t]{0.486\linewidth}
\centering\textbf{Original}\par\vspace{2pt}
\begin{minipage}[c][0.255\textheight][c]{\linewidth}\centering\includegraphics[width=\linewidth,height=0.255\textheight,keepaspectratio]{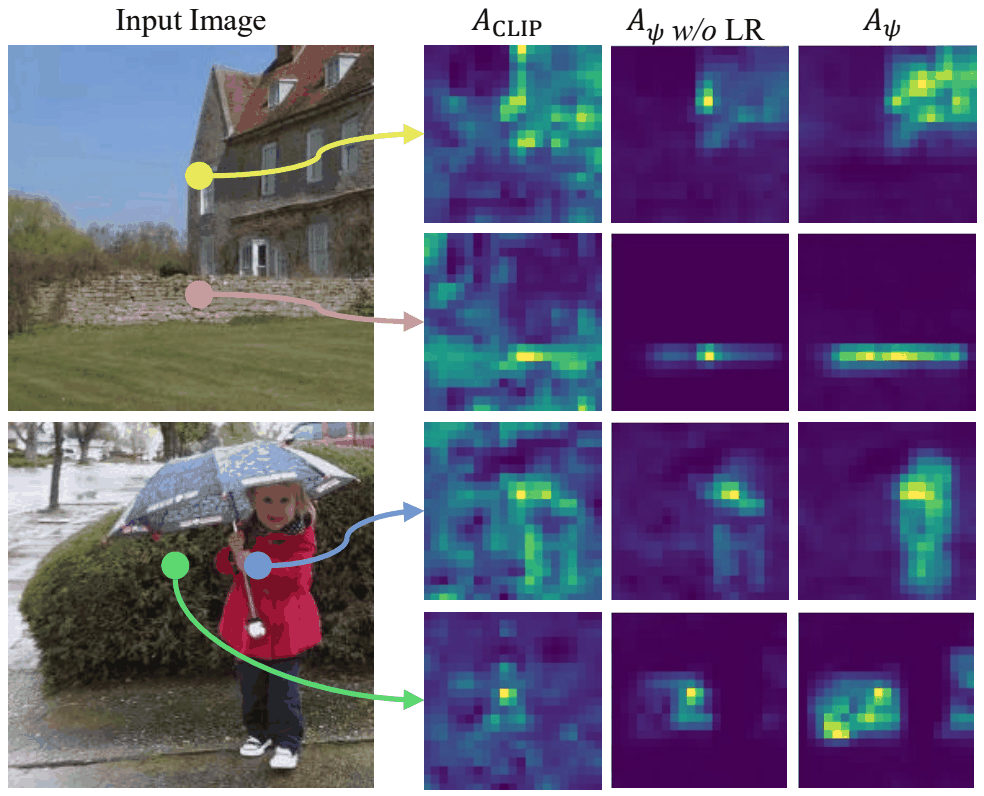}\end{minipage}\par
\vspace{3pt}{\scriptsize \setlength{\tabcolsep}{1.5pt}\begin{tabular*}{\linewidth}{@{\extracolsep{\fill}}rrrrrrr@{}}
\toprule
C & O & G & A & I & S & Mean \\
3 & 5 & 4 & 5 & 3 & 4 & 4.00\\
\bottomrule
\end{tabular*}
}
\end{minipage}\hfill
\begin{minipage}[t]{0.486\linewidth}
\centering\textbf{Study-generated derivative}\par\vspace{2pt}
\begin{minipage}[c][0.255\textheight][c]{\linewidth}\centering\includegraphics[width=\linewidth,height=0.255\textheight,keepaspectratio]{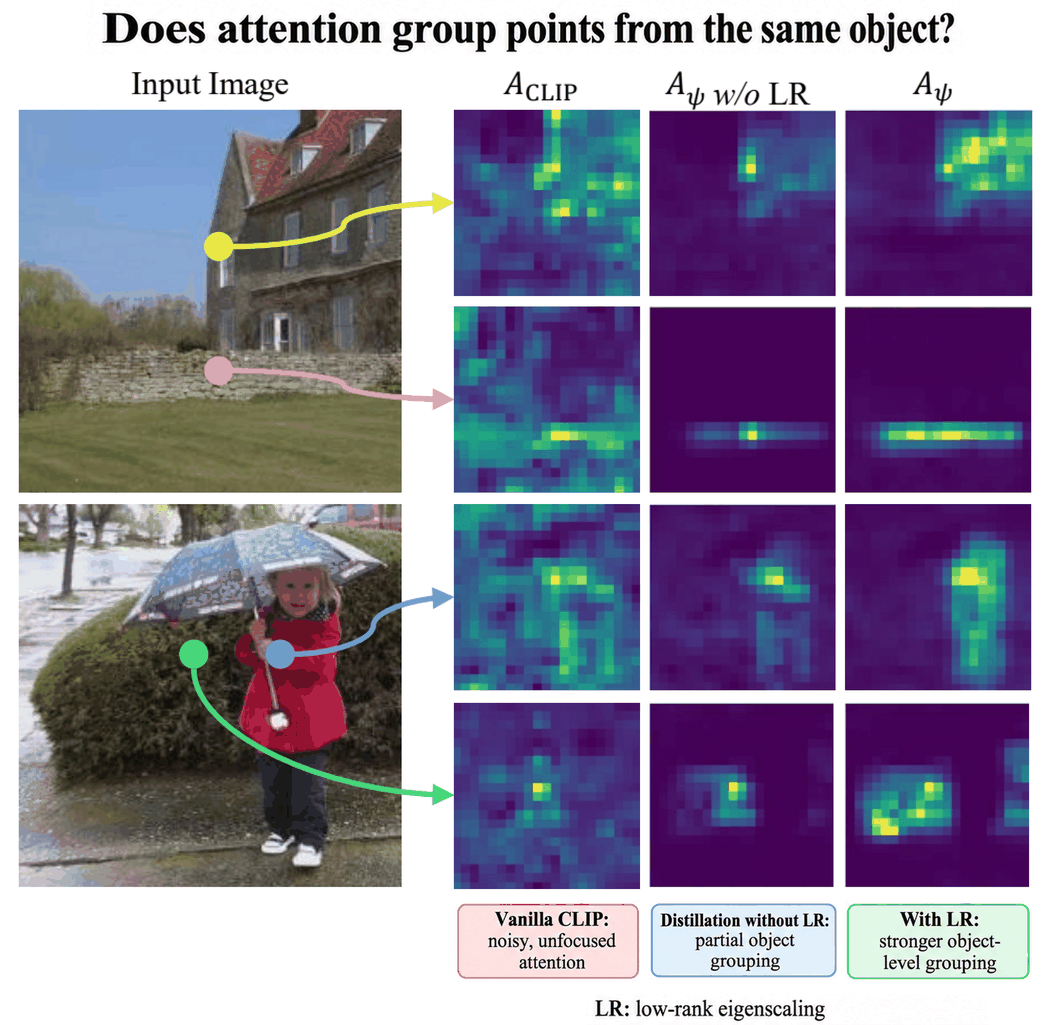}\end{minipage}\par
\vspace{3pt}
{\scriptsize \setlength{\tabcolsep}{1.5pt}\begin{tabular*}{\linewidth}{@{\extracolsep{\fill}}rrrrrrr@{}}
\toprule
C & O & G & A & I & S & Mean \\
5 (+2) & 5 (+0) & 4 (+0) & 4 (-1) & 4 (+1) & 5 (+1) & 4.50 (+0.50)\\
\bottomrule
\end{tabular*}
}
\end{minipage}\par
\vspace{3pt}
{\scriptsize Source public-date proxy: 2024-11-26. All displayed scores are medians of three runs; parentheses show revised minus original.\par}
\vspace{5pt}
{\bfseries Framework (source Fig.\ 1)} \hfill {\footnotesize \href{https://openaccess.thecvf.com/content/CVPR2025/papers/Kim\_Distilling\_Spectral\_Graph\_for\_Object-Context\_Aware\_Open-Vocabulary\_Semantic\_Segmentation\_CVPR\_2025\_paper.pdf}{Source PDF, p.\ 1}; ID: P128}\par
\vspace{2pt}
\noindent\begin{minipage}[t]{0.486\linewidth}
\centering\textbf{Original}\par\vspace{2pt}
\begin{minipage}[c][0.255\textheight][c]{\linewidth}\centering\includegraphics[width=\linewidth,height=0.255\textheight,keepaspectratio]{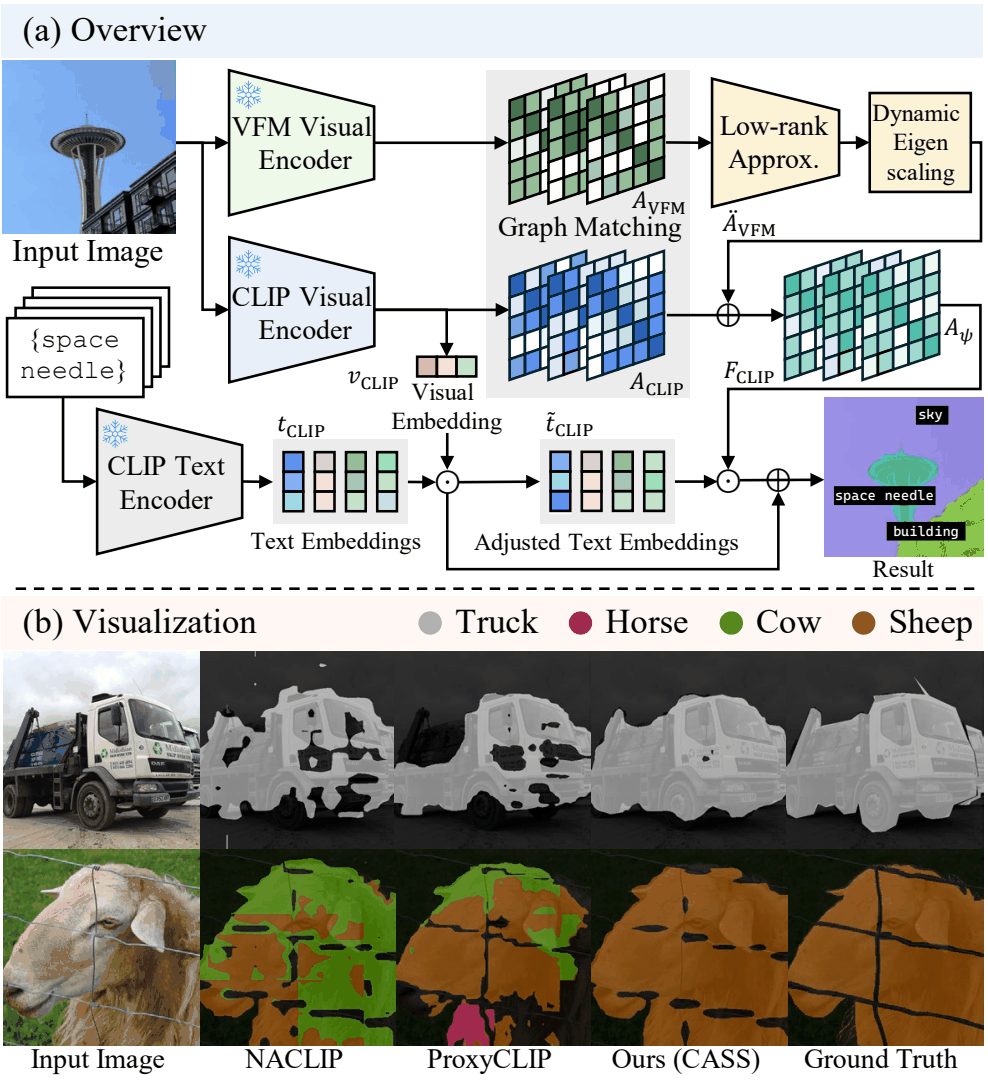}\end{minipage}\par
\vspace{3pt}{\scriptsize \setlength{\tabcolsep}{1.5pt}\begin{tabular*}{\linewidth}{@{\extracolsep{\fill}}rrrrrrr@{}}
\toprule
C & O & G & A & I & S & Mean \\
4 & 4 & 4 & 4 & 3 & 5 & 4.00\\
\bottomrule
\end{tabular*}
}
\end{minipage}\hfill
\begin{minipage}[t]{0.486\linewidth}
\centering\textbf{Study-generated derivative}\par\vspace{2pt}
\begin{minipage}[c][0.255\textheight][c]{\linewidth}\centering\includegraphics[width=\linewidth,height=0.255\textheight,keepaspectratio]{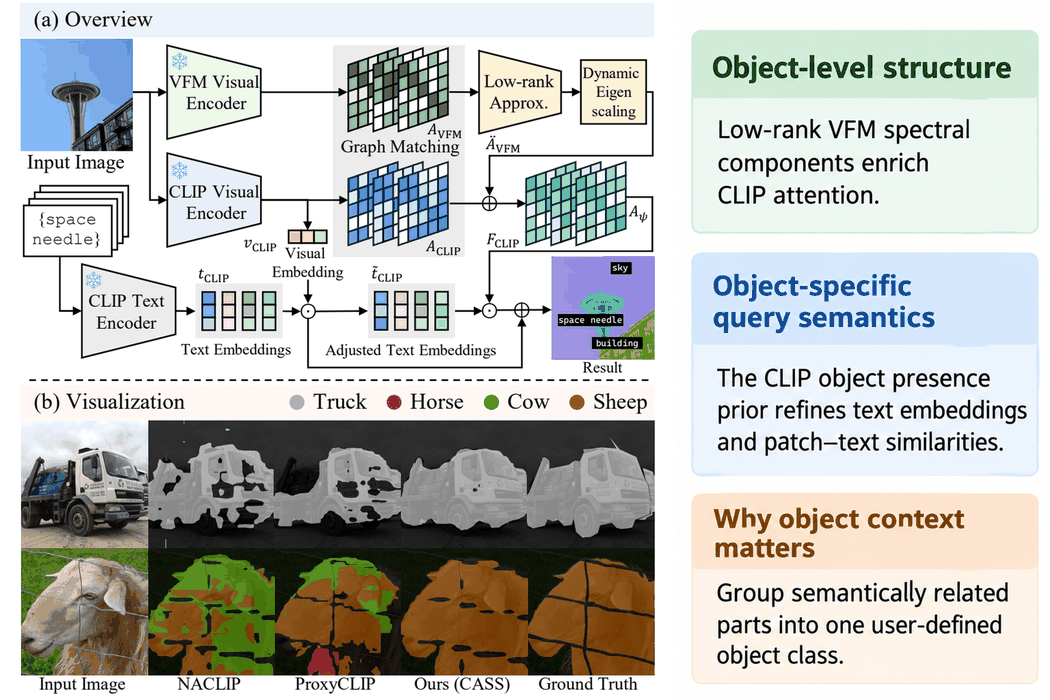}\end{minipage}\par
\vspace{3pt}
{\scriptsize \setlength{\tabcolsep}{1.5pt}\begin{tabular*}{\linewidth}{@{\extracolsep{\fill}}rrrrrrr@{}}
\toprule
C & O & G & A & I & S & Mean \\
4 (+0) & 4 (+0) & 4 (+0) & 3 (-1) & 3 (+0) & 5 (+0) & 3.83 (-0.17)\\
\bottomrule
\end{tabular*}
}
\end{minipage}\par
\vspace{3pt}
{\scriptsize Source public-date proxy: 2025-03-16. All displayed scores are medians of three runs; parentheses show revised minus original.\par}
\vfill
{\scriptsize C: clarity; O: organization/readability; G: graphical consistency; A: aesthetic coherence; I: visual inventiveness; S: design specificity. Mean is the arithmetic mean of six dimension medians, a descriptive display value, not a validated quality index.\par}

\clearpage
\phantomsection
\label{atlas:corpus062}
\pdfbookmark[1]{62. Complexity Experts are Task-Discriminative Learners for Any Image Restoration}{corpus062}
{\Large\bfseries\color{academicblue} 062 \quad \href{https://openaccess.thecvf.com/content/CVPR2025/html/Zamfir\_Complexity\_Experts\_are\_Task-Discriminative\_Learners\_for\_Any\_Image\_Restoration\_CVPR\_2025\_paper.html}{Complexity Experts are Task-Discriminative Learners for Any Image Restoration} \citep{corpus062}\par}
\vspace{3pt}
{\small S2 \enspace $\vert$ \enspace Image restoration \enspace $\vert$ \enspace CVPR 2025\par}
\vspace{4pt}
\vspace{5pt}
{\bfseries Motivation (source Fig.\ 1)} \hfill {\footnotesize \href{https://openaccess.thecvf.com/content/CVPR2025/papers/Zamfir\_Complexity\_Experts\_are\_Task-Discriminative\_Learners\_for\_Any\_Image\_Restoration\_CVPR\_2025\_paper.pdf}{Source PDF, p.\ 1}; ID: P67}\par
\vspace{2pt}
\noindent\begin{minipage}[t]{0.486\linewidth}
\centering\textbf{Original}\par\vspace{2pt}
\begin{minipage}[c][0.255\textheight][c]{\linewidth}\centering\includegraphics[width=\linewidth,height=0.255\textheight,keepaspectratio]{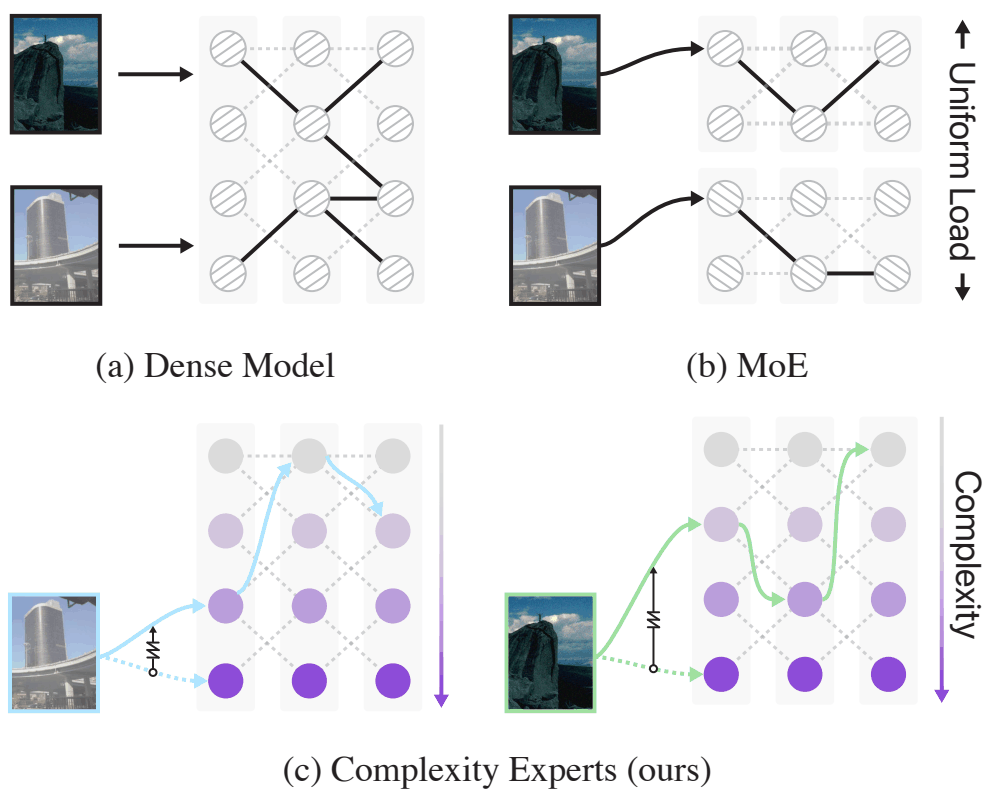}\end{minipage}\par
\vspace{3pt}{\scriptsize \setlength{\tabcolsep}{1.5pt}\begin{tabular*}{\linewidth}{@{\extracolsep{\fill}}rrrrrrr@{}}
\toprule
C & O & G & A & I & S & Mean \\
3 & 5 & 3 & 4 & 4 & 4 & 3.83\\
\bottomrule
\end{tabular*}
}
\end{minipage}\hfill
\begin{minipage}[t]{0.486\linewidth}
\centering\textbf{Study-generated derivative}\par\vspace{2pt}
\begin{minipage}[c][0.255\textheight][c]{\linewidth}\centering\includegraphics[width=\linewidth,height=0.255\textheight,keepaspectratio]{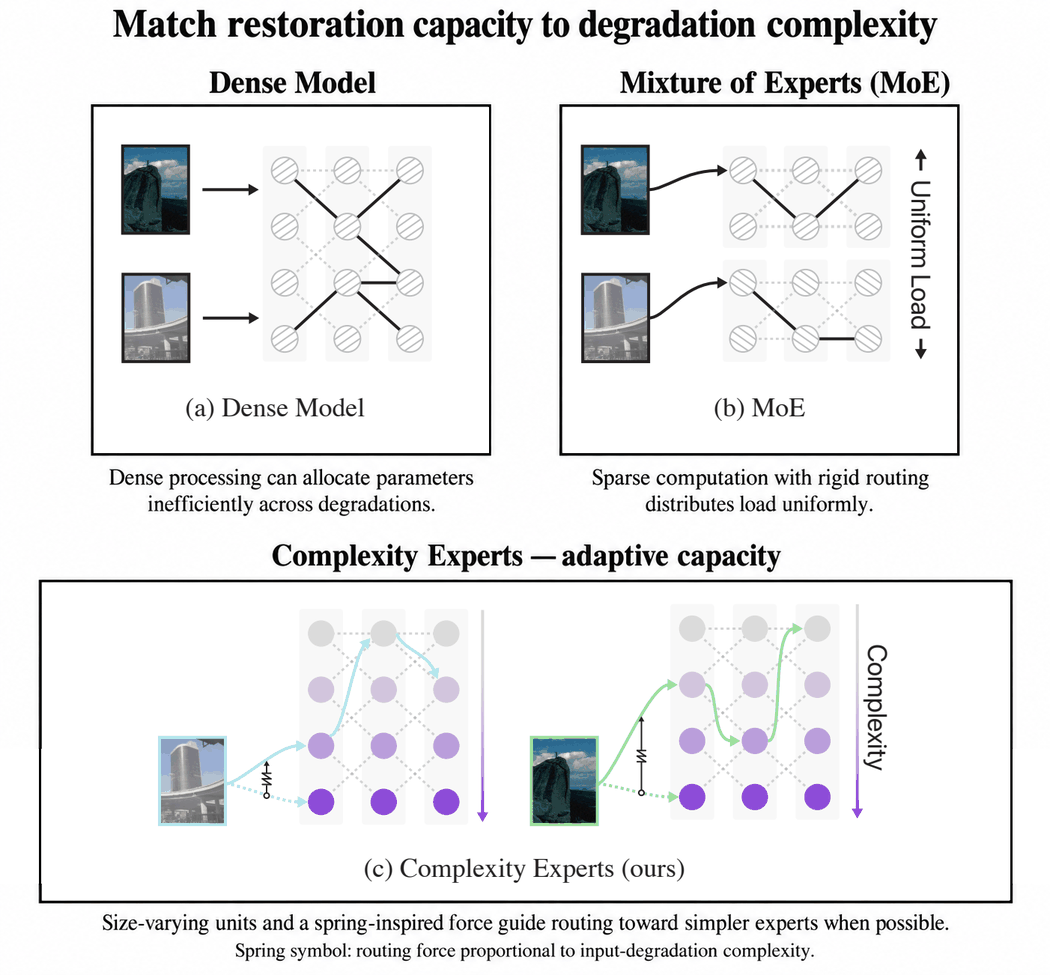}\end{minipage}\par
\vspace{3pt}
{\scriptsize \setlength{\tabcolsep}{1.5pt}\begin{tabular*}{\linewidth}{@{\extracolsep{\fill}}rrrrrrr@{}}
\toprule
C & O & G & A & I & S & Mean \\
4 (+1) & 4 (-1) & 3 (+0) & 4 (+0) & 3 (-1) & 4 (+0) & 3.67 (-0.17)\\
\bottomrule
\end{tabular*}
}
\end{minipage}\par
\vspace{3pt}
{\scriptsize Source public-date proxy: 2024-11-27. All displayed scores are medians of three runs; parentheses show revised minus original.\par}
\vspace{5pt}
{\bfseries Framework (source Fig.\ 2)} \hfill {\footnotesize \href{https://openaccess.thecvf.com/content/CVPR2025/papers/Zamfir\_Complexity\_Experts\_are\_Task-Discriminative\_Learners\_for\_Any\_Image\_Restoration\_CVPR\_2025\_paper.pdf}{Source PDF, p.\ 3}; ID: P68}\par
\vspace{2pt}
\noindent\begin{minipage}[t]{0.486\linewidth}
\centering\textbf{Original}\par\vspace{2pt}
\begin{minipage}[c][0.255\textheight][c]{\linewidth}\centering\includegraphics[width=\linewidth,height=0.255\textheight,keepaspectratio]{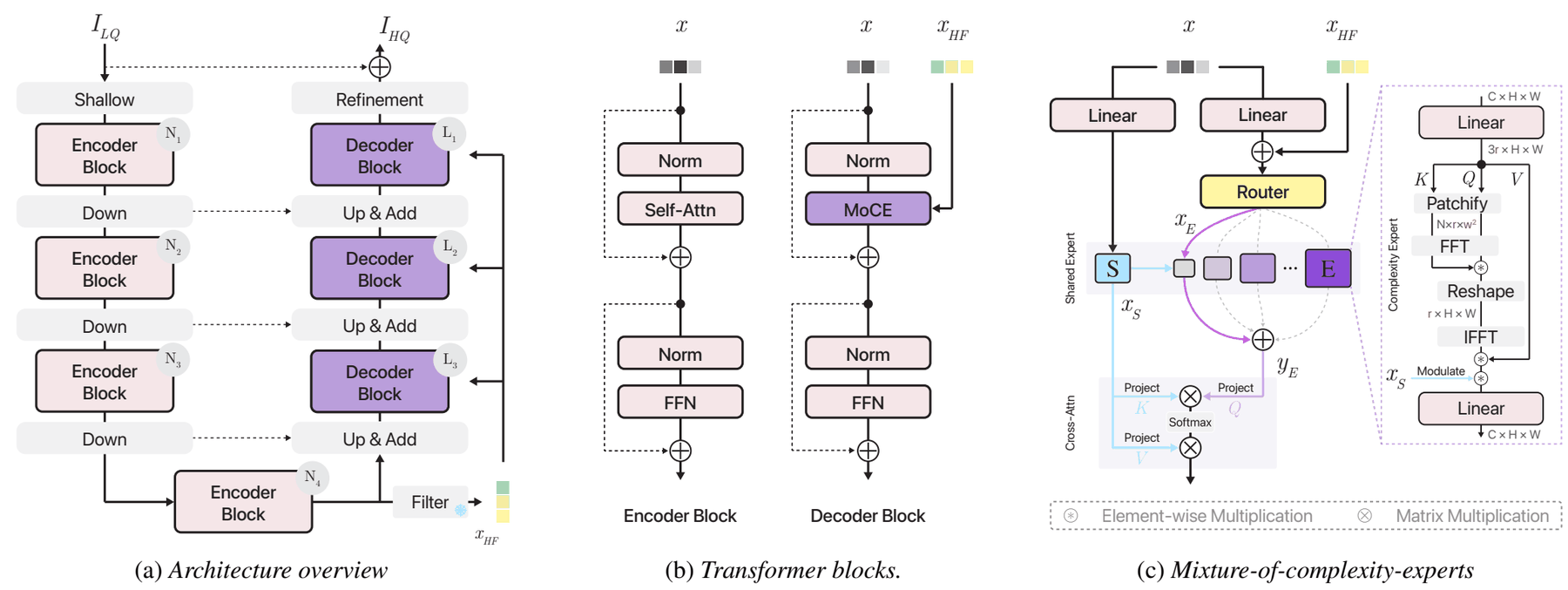}\end{minipage}\par
\vspace{3pt}{\scriptsize \setlength{\tabcolsep}{1.5pt}\begin{tabular*}{\linewidth}{@{\extracolsep{\fill}}rrrrrrr@{}}
\toprule
C & O & G & A & I & S & Mean \\
4 & 4 & 4 & 4 & 4 & 5 & 4.17\\
\bottomrule
\end{tabular*}
}
\end{minipage}\hfill
\begin{minipage}[t]{0.486\linewidth}
\centering\textbf{Study-generated derivative}\par\vspace{2pt}
\begin{minipage}[c][0.255\textheight][c]{\linewidth}\centering\includegraphics[width=\linewidth,height=0.255\textheight,keepaspectratio]{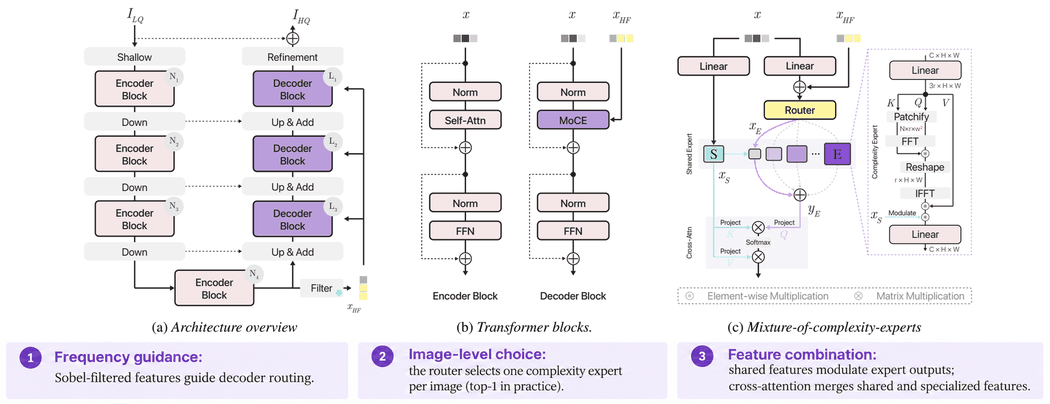}\end{minipage}\par
\vspace{3pt}
{\scriptsize \setlength{\tabcolsep}{1.5pt}\begin{tabular*}{\linewidth}{@{\extracolsep{\fill}}rrrrrrr@{}}
\toprule
C & O & G & A & I & S & Mean \\
4 (+0) & 4 (+0) & 4 (+0) & 4 (+0) & 4 (+0) & 5 (+0) & 4.17 (+0.00)\\
\bottomrule
\end{tabular*}
}
\end{minipage}\par
\vspace{3pt}
{\scriptsize Source public-date proxy: 2025-03-13. All displayed scores are medians of three runs; parentheses show revised minus original.\par}
\vfill
{\scriptsize C: clarity; O: organization/readability; G: graphical consistency; A: aesthetic coherence; I: visual inventiveness; S: design specificity. Mean is the arithmetic mean of six dimension medians, a descriptive display value, not a validated quality index.\par}

\clearpage
\phantomsection
\label{atlas:corpus068}
\pdfbookmark[1]{68. QMambaBSR: Burst Image Super-Resolution with Query State Space Model}{corpus068}
{\Large\bfseries\color{academicblue} 068 \quad \href{https://openaccess.thecvf.com/content/CVPR2025/html/Di\_QMambaBSR\_Burst\_Image\_Super-Resolution\_with\_Query\_State\_Space\_Model\_CVPR\_2025\_paper.html}{QMambaBSR: Burst Image Super-Resolution with Query State Space Model} \citep{corpus068}\par}
\vspace{3pt}
{\small S2 \enspace $\vert$ \enspace Image restoration \enspace $\vert$ \enspace CVPR 2025\par}
\vspace{4pt}
\vspace{5pt}
{\bfseries Motivation (source Fig.\ 1)} \hfill {\footnotesize \href{https://openaccess.thecvf.com/content/CVPR2025/papers/Di\_QMambaBSR\_Burst\_Image\_Super-Resolution\_with\_Query\_State\_Space\_Model\_CVPR\_2025\_paper.pdf}{Source PDF, p.\ 1}; ID: P177}\par
\vspace{2pt}
\noindent\begin{minipage}[t]{0.486\linewidth}
\centering\textbf{Original}\par\vspace{2pt}
\begin{minipage}[c][0.255\textheight][c]{\linewidth}\centering\includegraphics[width=\linewidth,height=0.255\textheight,keepaspectratio]{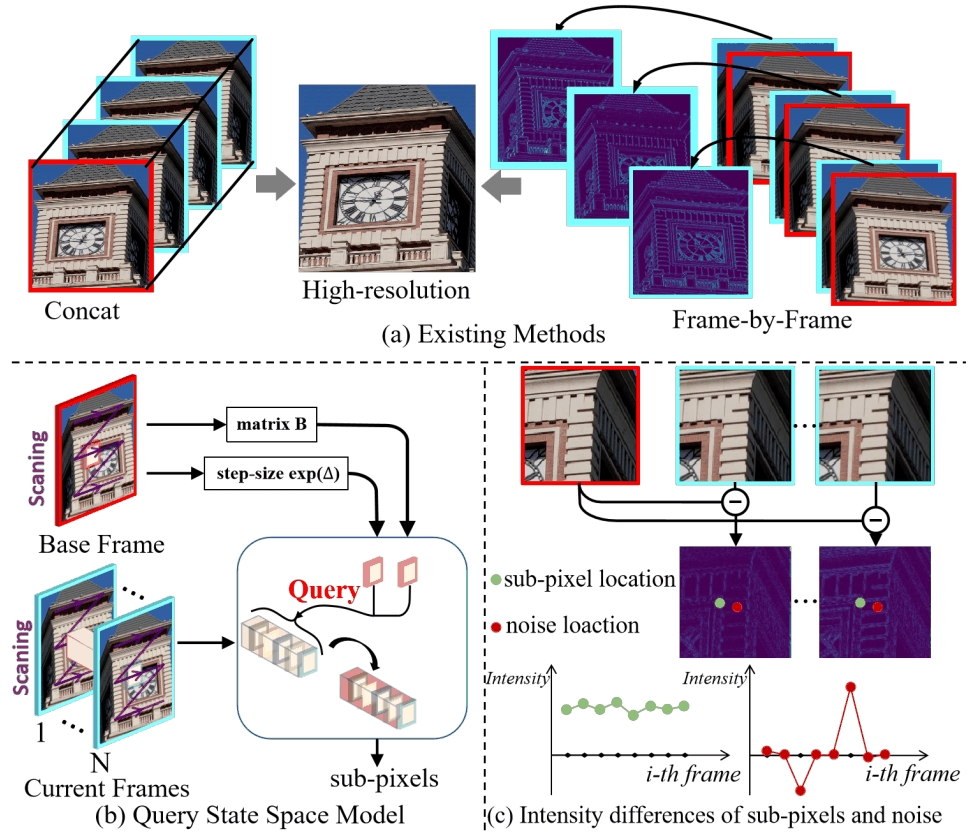}\end{minipage}\par
\vspace{3pt}{\scriptsize \setlength{\tabcolsep}{1.5pt}\begin{tabular*}{\linewidth}{@{\extracolsep{\fill}}rrrrrrr@{}}
\toprule
C & O & G & A & I & S & Mean \\
3 & 3 & 3 & 3 & 4 & 5 & 3.50\\
\bottomrule
\end{tabular*}
}
\end{minipage}\hfill
\begin{minipage}[t]{0.486\linewidth}
\centering\textbf{Study-generated derivative}\par\vspace{2pt}
\begin{minipage}[c][0.255\textheight][c]{\linewidth}\centering\includegraphics[width=\linewidth,height=0.255\textheight,keepaspectratio]{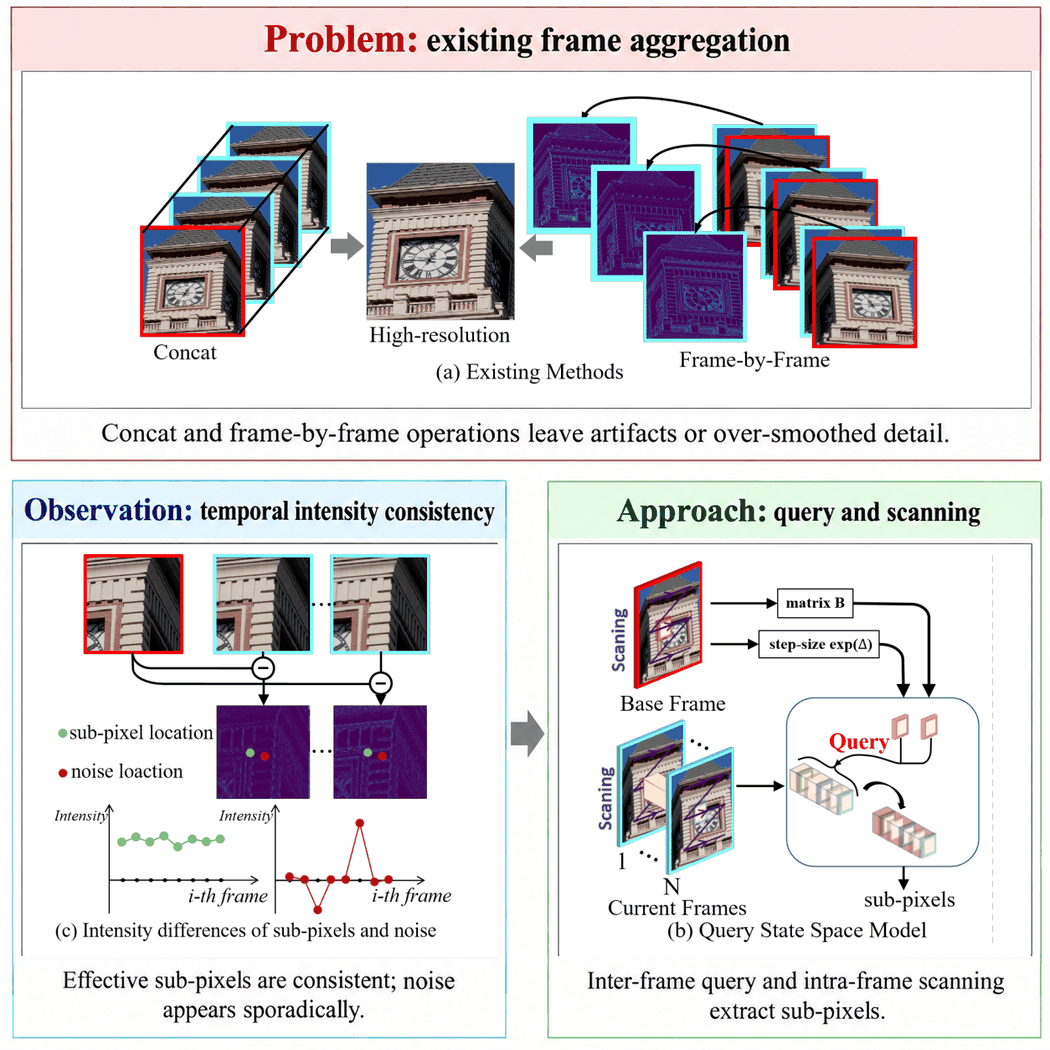}\end{minipage}\par
\vspace{3pt}
{\scriptsize \setlength{\tabcolsep}{1.5pt}\begin{tabular*}{\linewidth}{@{\extracolsep{\fill}}rrrrrrr@{}}
\toprule
C & O & G & A & I & S & Mean \\
4 (+1) & 4 (+1) & 3 (+0) & 3 (+0) & 4 (+0) & 5 (+0) & 3.83 (+0.33)\\
\bottomrule
\end{tabular*}
}
\end{minipage}\par
\vspace{3pt}
{\scriptsize Source public-date proxy: 2025-03-10. All displayed scores are medians of three runs; parentheses show revised minus original.\par}
\vspace{5pt}
{\bfseries Framework (source Fig.\ 2)} \hfill {\footnotesize \href{https://openaccess.thecvf.com/content/CVPR2025/papers/Di\_QMambaBSR\_Burst\_Image\_Super-Resolution\_with\_Query\_State\_Space\_Model\_CVPR\_2025\_paper.pdf}{Source PDF, p.\ 4}; ID: P178}\par
\vspace{2pt}
\noindent\begin{minipage}[t]{0.486\linewidth}
\centering\textbf{Original}\par\vspace{2pt}
\begin{minipage}[c][0.255\textheight][c]{\linewidth}\centering\includegraphics[width=\linewidth,height=0.255\textheight,keepaspectratio]{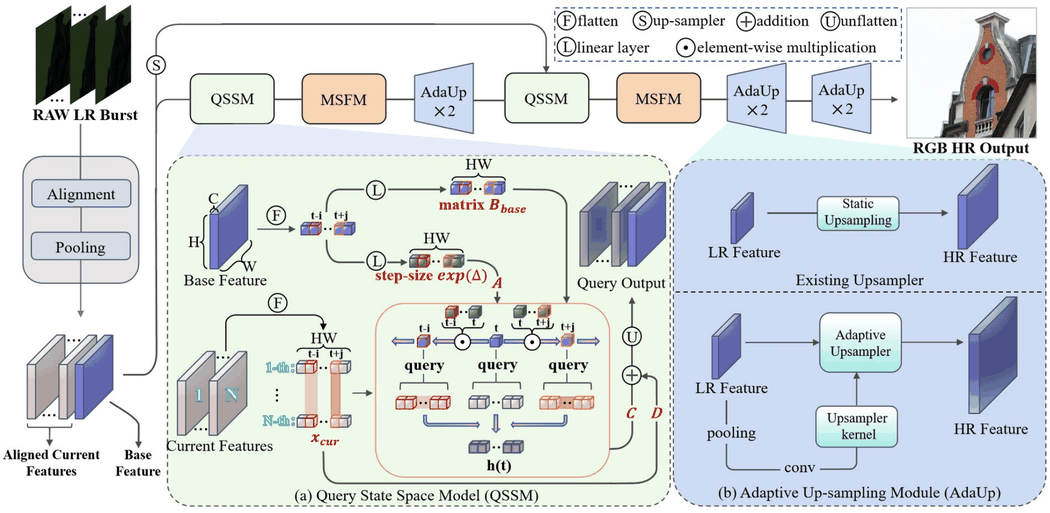}\end{minipage}\par
\vspace{3pt}{\scriptsize \setlength{\tabcolsep}{1.5pt}\begin{tabular*}{\linewidth}{@{\extracolsep{\fill}}rrrrrrr@{}}
\toprule
C & O & G & A & I & S & Mean \\
4 & 3 & 4 & 3 & 4 & 5 & 3.83\\
\bottomrule
\end{tabular*}
}
\end{minipage}\hfill
\begin{minipage}[t]{0.486\linewidth}
\centering\textbf{Study-generated derivative}\par\vspace{2pt}
\begin{minipage}[c][0.255\textheight][c]{\linewidth}\centering\includegraphics[width=\linewidth,height=0.255\textheight,keepaspectratio]{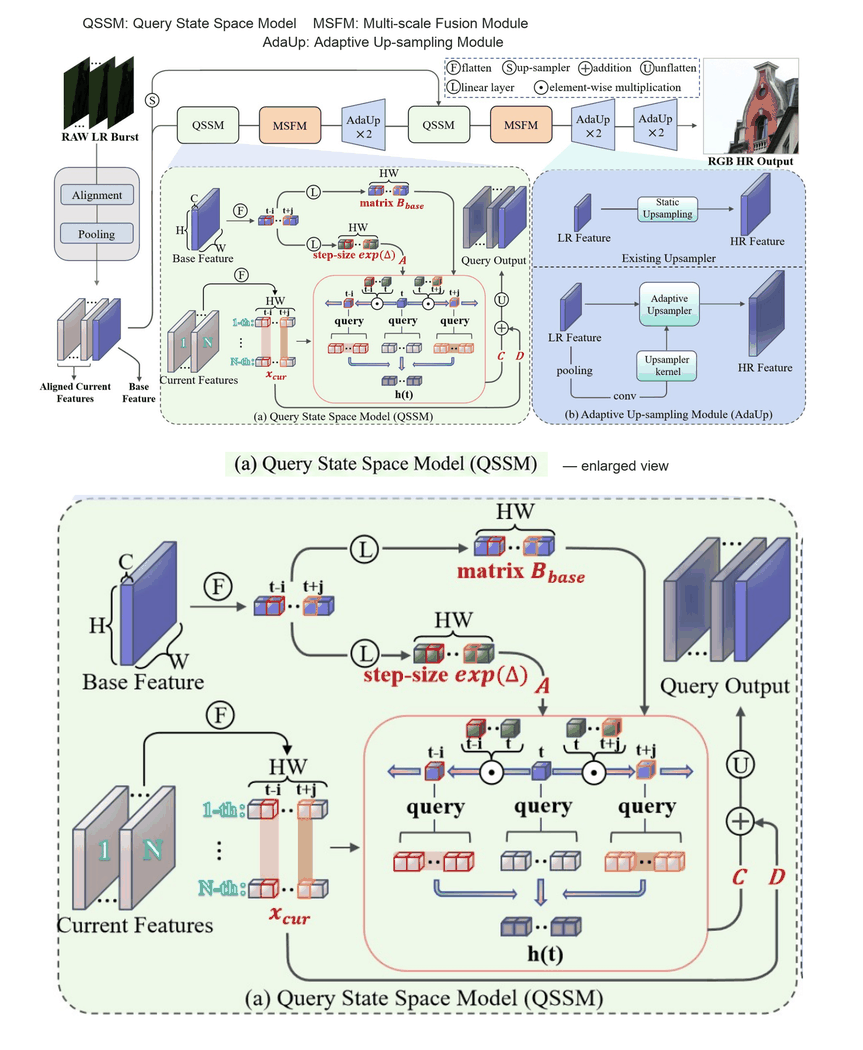}\end{minipage}\par
\vspace{3pt}
{\scriptsize \setlength{\tabcolsep}{1.5pt}\begin{tabular*}{\linewidth}{@{\extracolsep{\fill}}rrrrrrr@{}}
\toprule
C & O & G & A & I & S & Mean \\
4 (+0) & 3 (+0) & 4 (+0) & 3 (+0) & 4 (+0) & 5 (+0) & 3.83 (+0.00)\\
\bottomrule
\end{tabular*}
}
\end{minipage}\par
\vspace{3pt}
{\scriptsize Source public-date proxy: 2025-03-10. All displayed scores are medians of three runs; parentheses show revised minus original.\par}
\vfill
{\scriptsize C: clarity; O: organization/readability; G: graphical consistency; A: aesthetic coherence; I: visual inventiveness; S: design specificity. Mean is the arithmetic mean of six dimension medians, a descriptive display value, not a validated quality index.\par}

\clearpage
\phantomsection
\label{atlas:corpus073}
\pdfbookmark[1]{73. CrossSDF: 3D Reconstruction of Thin Structures From Cross-Sections}{corpus073}
{\Large\bfseries\color{academicblue} 073 \quad \href{https://openaccess.thecvf.com/content/CVPR2025/html/Walker\_CrossSDF\_3D\_Reconstruction\_of\_Thin\_Structures\_From\_Cross-Sections\_CVPR\_2025\_paper.html}{CrossSDF: 3D Reconstruction of Thin Structures From Cross-Sections} \citep{corpus073}\par}
\vspace{3pt}
{\small S2 \enspace $\vert$ \enspace 3D vision \enspace $\vert$ \enspace CVPR 2025\par}
\vspace{4pt}
\vspace{5pt}
{\bfseries Motivation (source Fig.\ 1)} \hfill {\footnotesize \href{https://openaccess.thecvf.com/content/CVPR2025/papers/Walker\_CrossSDF\_3D\_Reconstruction\_of\_Thin\_Structures\_From\_Cross-Sections\_CVPR\_2025\_paper.pdf}{Source PDF, p.\ 1}; ID: P221}\par
\vspace{2pt}
\noindent\begin{minipage}[t]{0.486\linewidth}
\centering\textbf{Original}\par\vspace{2pt}
\begin{minipage}[c][0.255\textheight][c]{\linewidth}\centering\includegraphics[width=\linewidth,height=0.255\textheight,keepaspectratio]{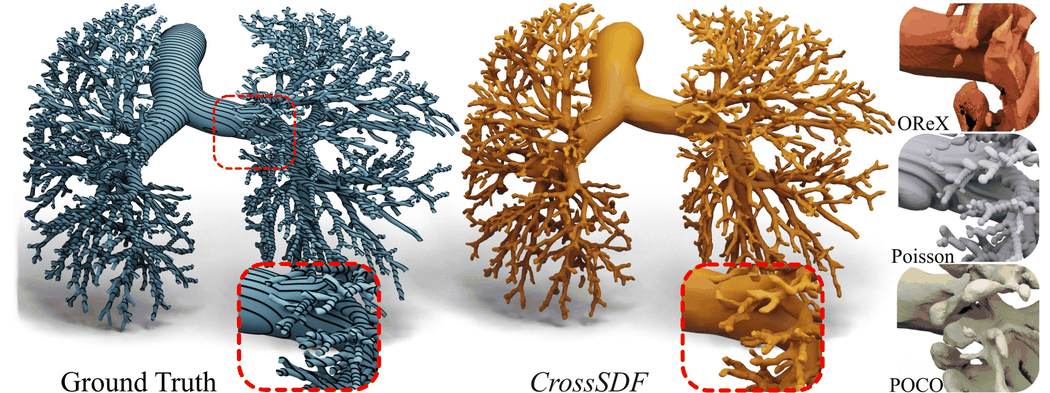}\end{minipage}\par
\vspace{3pt}{\scriptsize \setlength{\tabcolsep}{1.5pt}\begin{tabular*}{\linewidth}{@{\extracolsep{\fill}}rrrrrrr@{}}
\toprule
C & O & G & A & I & S & Mean \\
3 & 4 & 4 & 3 & 3 & 5 & 3.67\\
\bottomrule
\end{tabular*}
}
\end{minipage}\hfill
\begin{minipage}[t]{0.486\linewidth}
\centering\textbf{Study-generated derivative}\par\vspace{2pt}
\begin{minipage}[c][0.255\textheight][c]{\linewidth}\centering\includegraphics[width=\linewidth,height=0.255\textheight,keepaspectratio]{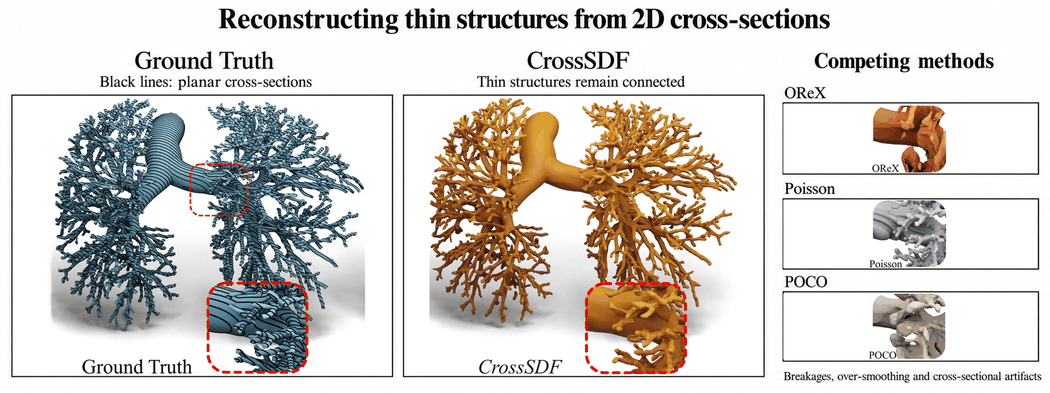}\end{minipage}\par
\vspace{3pt}
{\scriptsize \setlength{\tabcolsep}{1.5pt}\begin{tabular*}{\linewidth}{@{\extracolsep{\fill}}rrrrrrr@{}}
\toprule
C & O & G & A & I & S & Mean \\
4 (+1) & 4 (+0) & 4 (+0) & 3 (+0) & 3 (+0) & 5 (+0) & 3.83 (+0.17)\\
\bottomrule
\end{tabular*}
}
\end{minipage}\par
\vspace{3pt}
{\scriptsize Source public-date proxy: 2024-12-05. All displayed scores are medians of three runs; parentheses show revised minus original.\par}
\vspace{5pt}
{\bfseries Framework (source Fig.\ 2)} \hfill {\footnotesize \href{https://openaccess.thecvf.com/content/CVPR2025/papers/Walker\_CrossSDF\_3D\_Reconstruction\_of\_Thin\_Structures\_From\_Cross-Sections\_CVPR\_2025\_paper.pdf}{Source PDF, p.\ 3}; ID: P222}\par
\vspace{2pt}
\noindent\begin{minipage}[t]{0.486\linewidth}
\centering\textbf{Original}\par\vspace{2pt}
\begin{minipage}[c][0.255\textheight][c]{\linewidth}\centering\includegraphics[width=\linewidth,height=0.255\textheight,keepaspectratio]{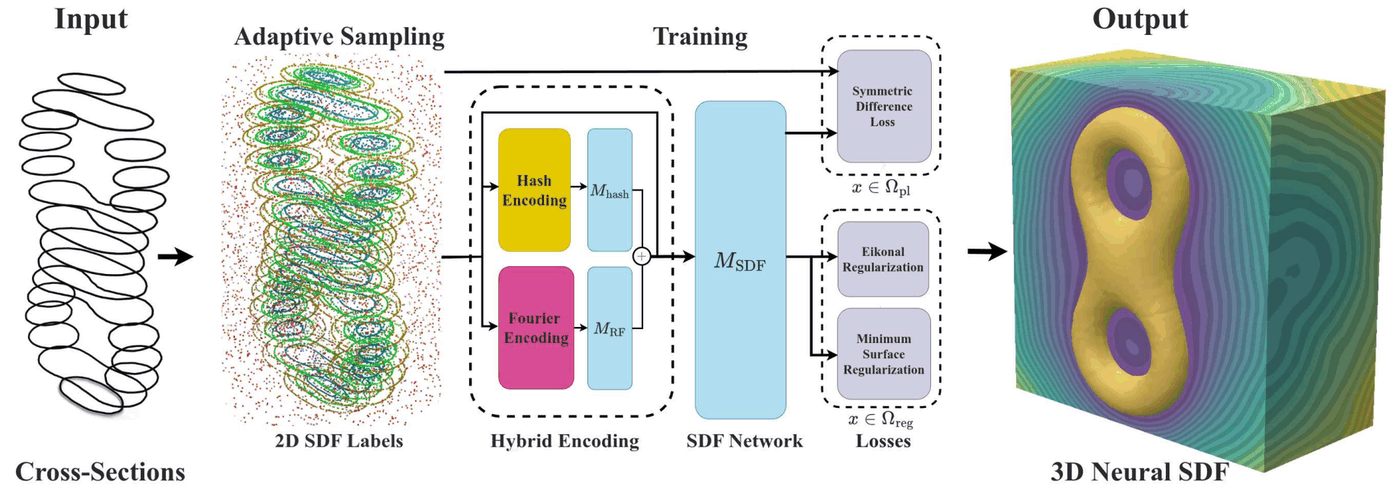}\end{minipage}\par
\vspace{3pt}{\scriptsize \setlength{\tabcolsep}{1.5pt}\begin{tabular*}{\linewidth}{@{\extracolsep{\fill}}rrrrrrr@{}}
\toprule
C & O & G & A & I & S & Mean \\
4 & 4 & 4 & 4 & 3 & 5 & 4.00\\
\bottomrule
\end{tabular*}
}
\end{minipage}\hfill
\begin{minipage}[t]{0.486\linewidth}
\centering\textbf{Study-generated derivative}\par\vspace{2pt}
\begin{minipage}[c][0.255\textheight][c]{\linewidth}\centering\includegraphics[width=\linewidth,height=0.255\textheight,keepaspectratio]{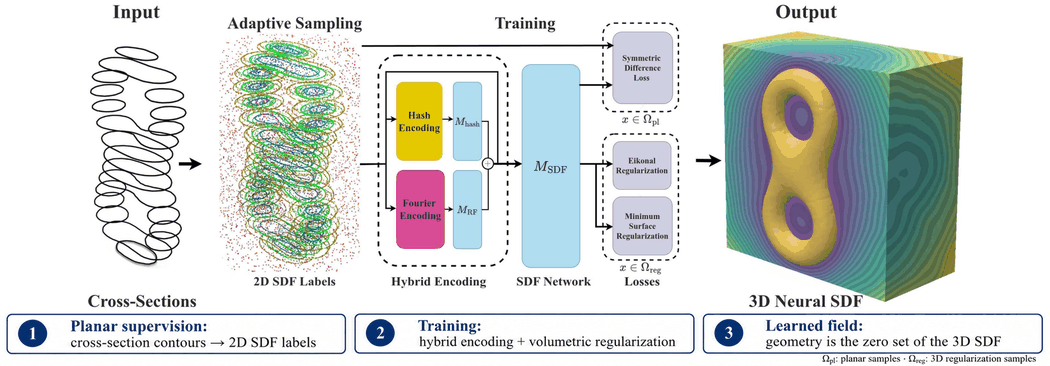}\end{minipage}\par
\vspace{3pt}
{\scriptsize \setlength{\tabcolsep}{1.5pt}\begin{tabular*}{\linewidth}{@{\extracolsep{\fill}}rrrrrrr@{}}
\toprule
C & O & G & A & I & S & Mean \\
4 (+0) & 4 (+0) & 4 (+0) & 4 (+0) & 4 (+1) & 5 (+0) & 4.17 (+0.17)\\
\bottomrule
\end{tabular*}
}
\end{minipage}\par
\vspace{3pt}
{\scriptsize Source public-date proxy: 2024-12-05. All displayed scores are medians of three runs; parentheses show revised minus original.\par}
\vfill
{\scriptsize C: clarity; O: organization/readability; G: graphical consistency; A: aesthetic coherence; I: visual inventiveness; S: design specificity. Mean is the arithmetic mean of six dimension medians, a descriptive display value, not a validated quality index.\par}

\clearpage
\phantomsection
\label{atlas:corpus078}
\pdfbookmark[1]{78. Parametric Point Cloud Completion for Polygonal Surface Reconstruction}{corpus078}
{\Large\bfseries\color{academicblue} 078 \quad \href{https://openaccess.thecvf.com/content/CVPR2025/html/Chen\_Parametric\_Point\_Cloud\_Completion\_for\_Polygonal\_Surface\_Reconstruction\_CVPR\_2025\_paper.html}{Parametric Point Cloud Completion for Polygonal Surface Reconstruction} \citep{corpus078}\par}
\vspace{3pt}
{\small S2 \enspace $\vert$ \enspace 3D vision \enspace $\vert$ \enspace CVPR 2025\par}
\vspace{4pt}
\vspace{5pt}
{\bfseries Motivation (source Fig.\ 1)} \hfill {\footnotesize \href{https://openaccess.thecvf.com/content/CVPR2025/papers/Chen\_Parametric\_Point\_Cloud\_Completion\_for\_Polygonal\_Surface\_Reconstruction\_CVPR\_2025\_paper.pdf}{Source PDF, p.\ 1}; ID: P235}\par
\vspace{2pt}
\noindent\begin{minipage}[t]{0.486\linewidth}
\centering\textbf{Original}\par\vspace{2pt}
\begin{minipage}[c][0.255\textheight][c]{\linewidth}\centering\includegraphics[width=\linewidth,height=0.255\textheight,keepaspectratio]{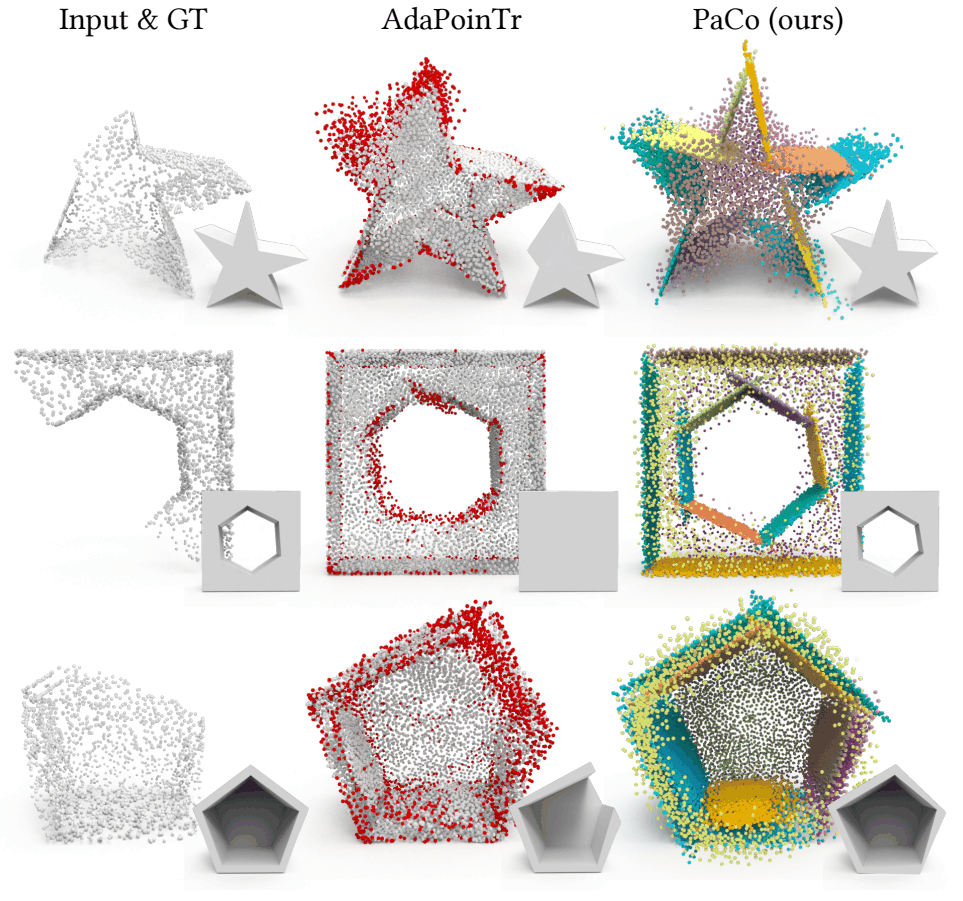}\end{minipage}\par
\vspace{3pt}{\scriptsize \setlength{\tabcolsep}{1.5pt}\begin{tabular*}{\linewidth}{@{\extracolsep{\fill}}rrrrrrr@{}}
\toprule
C & O & G & A & I & S & Mean \\
3 & 5 & 3 & 4 & 3 & 5 & 3.83\\
\bottomrule
\end{tabular*}
}
\end{minipage}\hfill
\begin{minipage}[t]{0.486\linewidth}
\centering\textbf{Study-generated derivative}\par\vspace{2pt}
\begin{minipage}[c][0.255\textheight][c]{\linewidth}\centering\includegraphics[width=\linewidth,height=0.255\textheight,keepaspectratio]{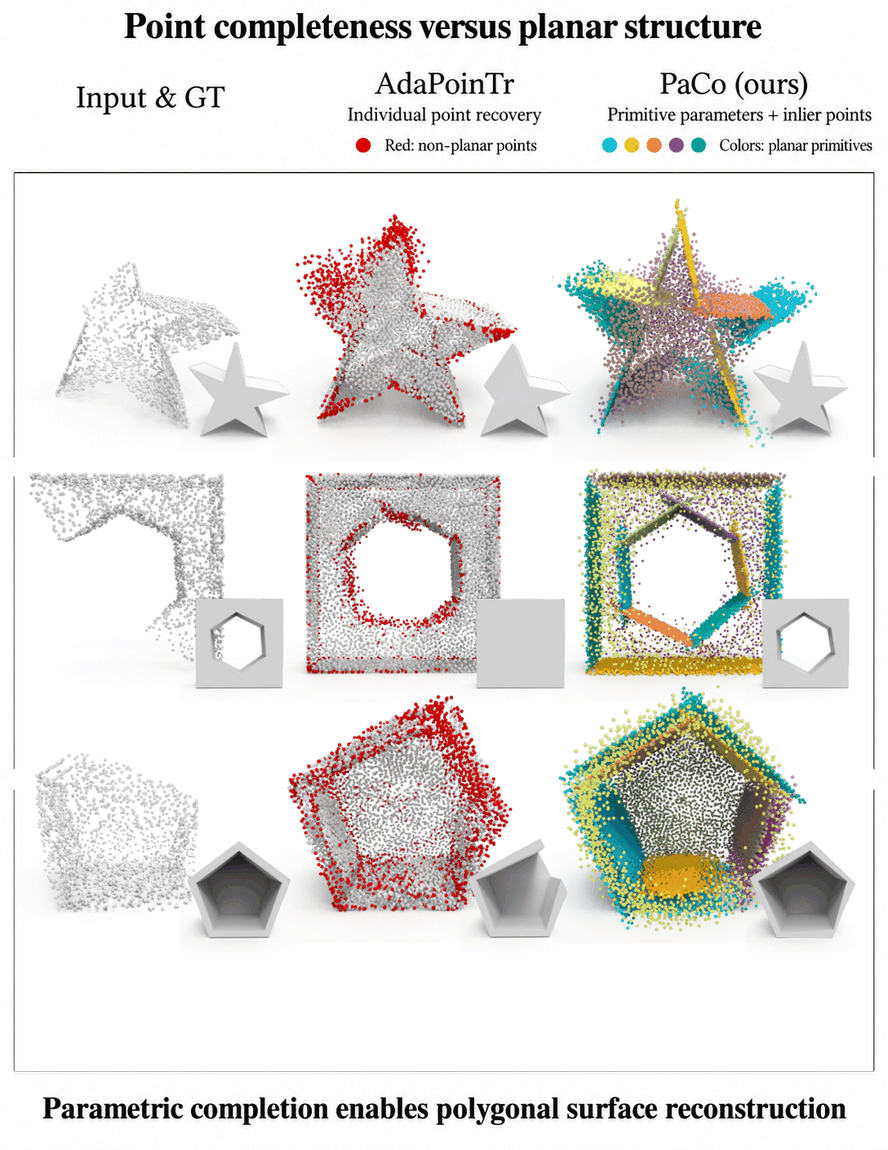}\end{minipage}\par
\vspace{3pt}
{\scriptsize \setlength{\tabcolsep}{1.5pt}\begin{tabular*}{\linewidth}{@{\extracolsep{\fill}}rrrrrrr@{}}
\toprule
C & O & G & A & I & S & Mean \\
4 (+1) & 4 (-1) & 4 (+1) & 3 (-1) & 4 (+1) & 5 (+0) & 4.00 (+0.17)\\
\bottomrule
\end{tabular*}
}
\end{minipage}\par
\vspace{3pt}
{\scriptsize Source public-date proxy: 2025-03-11. All displayed scores are medians of three runs; parentheses show revised minus original.\par}
\vspace{5pt}
{\bfseries Framework (source Fig.\ 2)} \hfill {\footnotesize \href{https://openaccess.thecvf.com/content/CVPR2025/papers/Chen\_Parametric\_Point\_Cloud\_Completion\_for\_Polygonal\_Surface\_Reconstruction\_CVPR\_2025\_paper.pdf}{Source PDF, p.\ 3}; ID: P236}\par
\vspace{2pt}
\noindent\begin{minipage}[t]{0.486\linewidth}
\centering\textbf{Original}\par\vspace{2pt}
\begin{minipage}[c][0.255\textheight][c]{\linewidth}\centering\includegraphics[width=\linewidth,height=0.255\textheight,keepaspectratio]{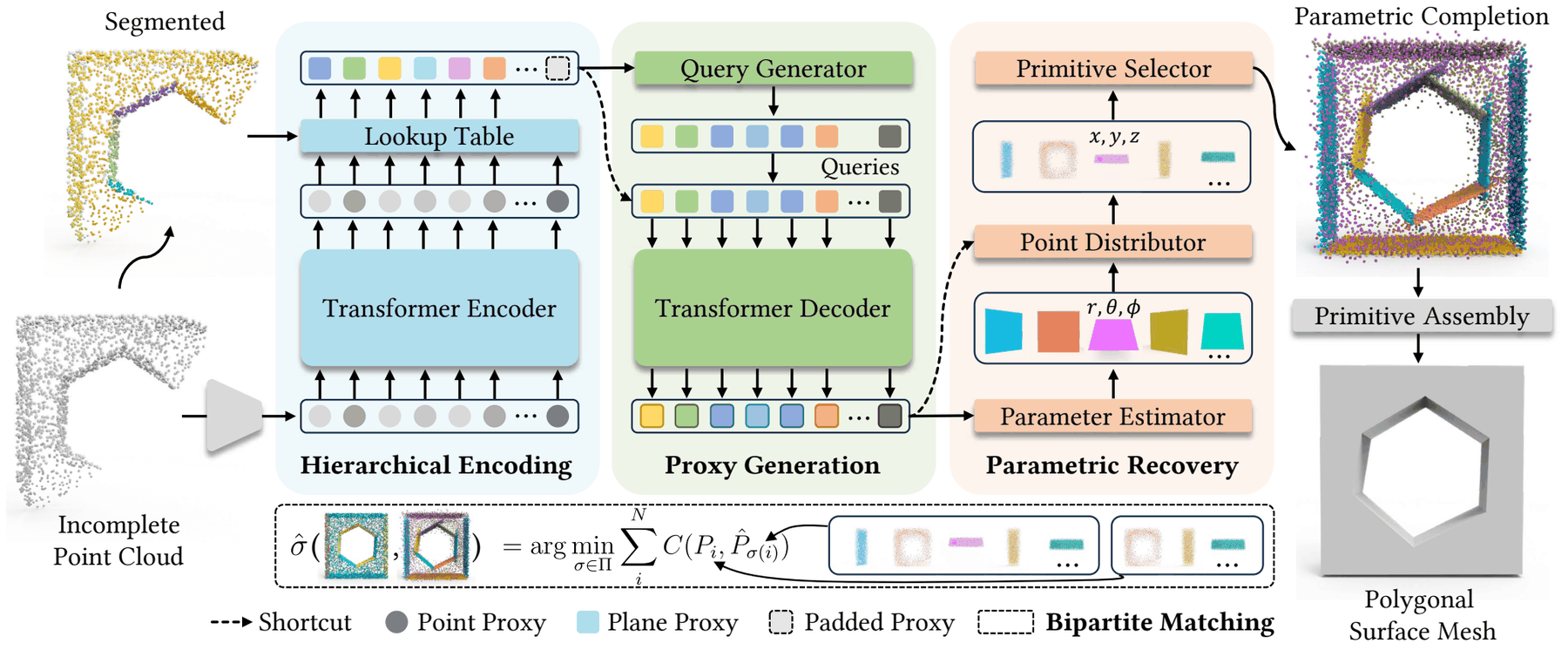}\end{minipage}\par
\vspace{3pt}{\scriptsize \setlength{\tabcolsep}{1.5pt}\begin{tabular*}{\linewidth}{@{\extracolsep{\fill}}rrrrrrr@{}}
\toprule
C & O & G & A & I & S & Mean \\
4 & 4 & 4 & 4 & 4 & 5 & 4.17\\
\bottomrule
\end{tabular*}
}
\end{minipage}\hfill
\begin{minipage}[t]{0.486\linewidth}
\centering\textbf{Study-generated derivative}\par\vspace{2pt}
\begin{minipage}[c][0.255\textheight][c]{\linewidth}\centering\includegraphics[width=\linewidth,height=0.255\textheight,keepaspectratio]{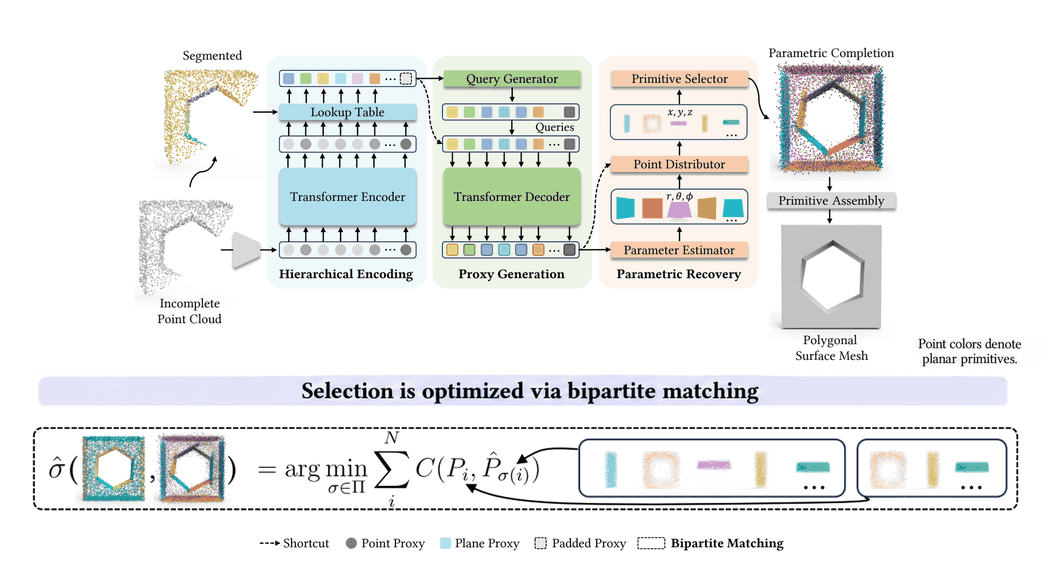}\end{minipage}\par
\vspace{3pt}
{\scriptsize \setlength{\tabcolsep}{1.5pt}\begin{tabular*}{\linewidth}{@{\extracolsep{\fill}}rrrrrrr@{}}
\toprule
C & O & G & A & I & S & Mean \\
4 (+0) & 4 (+0) & 4 (+0) & 4 (+0) & 4 (+0) & 5 (+0) & 4.17 (+0.00)\\
\bottomrule
\end{tabular*}
}
\end{minipage}\par
\vspace{3pt}
{\scriptsize Source public-date proxy: 2025-03-11. All displayed scores are medians of three runs; parentheses show revised minus original.\par}
\vfill
{\scriptsize C: clarity; O: organization/readability; G: graphical consistency; A: aesthetic coherence; I: visual inventiveness; S: design specificity. Mean is the arithmetic mean of six dimension medians, a descriptive display value, not a validated quality index.\par}

\clearpage
\phantomsection
\label{atlas:corpus085}
\pdfbookmark[1]{85. Instance-Free Domain Adaptive Object Detection}{corpus085}
{\Large\bfseries\color{academicblue} 085 \quad \href{https://arxiv.org/abs/2602.06484}{Instance-Free Domain Adaptive Object Detection} \citep{corpus085}\par}
\vspace{3pt}
{\small S3 \enspace $\vert$ \enspace Object detection \enspace $\vert$ \enspace arXiv 2602.06484\par}
\vspace{4pt}
\vspace{5pt}
{\bfseries Motivation (source Fig.\ 1)} \hfill {\footnotesize \href{https://arxiv.org/pdf/2602.06484v1}{Source PDF, p.\ 1}; ID: P287}\par
\vspace{2pt}
\noindent\begin{minipage}[t]{0.486\linewidth}
\centering\textbf{Original}\par\vspace{2pt}
\begin{minipage}[c][0.255\textheight][c]{\linewidth}\centering\includegraphics[width=\linewidth,height=0.255\textheight,keepaspectratio]{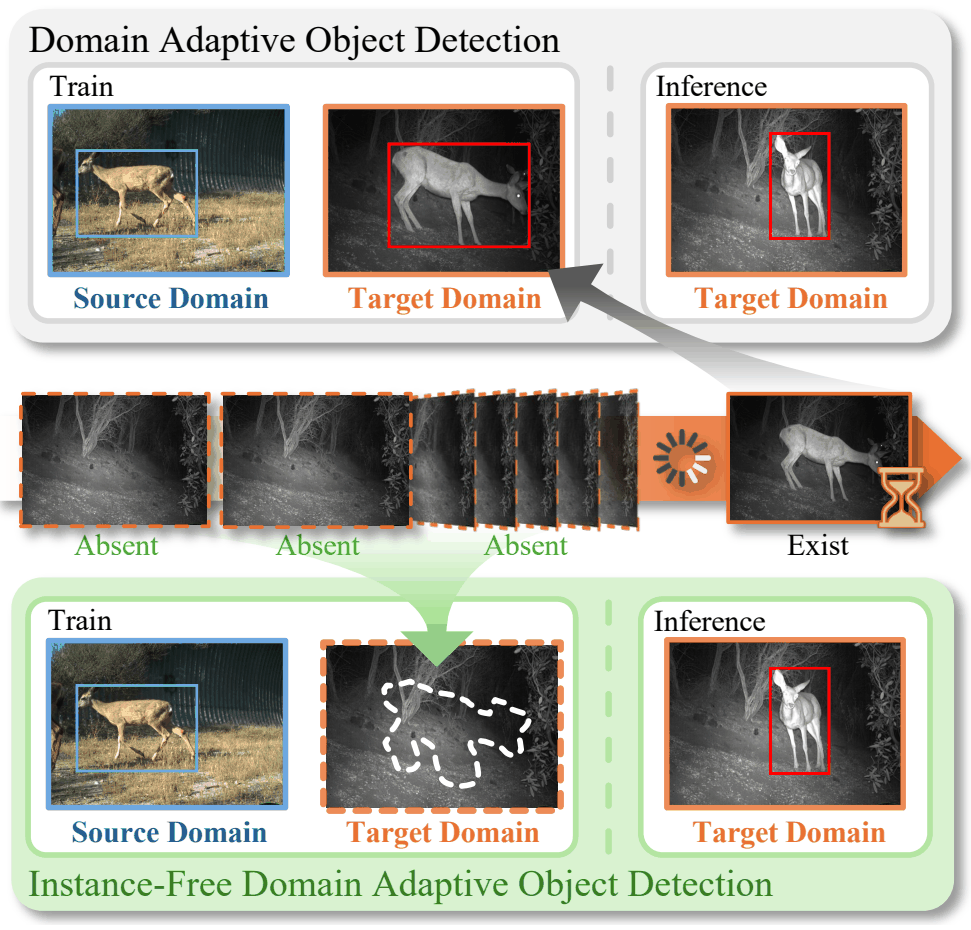}\end{minipage}\par
\vspace{3pt}{\scriptsize \setlength{\tabcolsep}{1.5pt}\begin{tabular*}{\linewidth}{@{\extracolsep{\fill}}rrrrrrr@{}}
\toprule
C & O & G & A & I & S & Mean \\
4 & 4 & 4 & 3 & 4 & 5 & 4.00\\
\bottomrule
\end{tabular*}
}
\end{minipage}\hfill
\begin{minipage}[t]{0.486\linewidth}
\centering\textbf{Study-generated derivative}\par\vspace{2pt}
\begin{minipage}[c][0.255\textheight][c]{\linewidth}\centering\includegraphics[width=\linewidth,height=0.255\textheight,keepaspectratio]{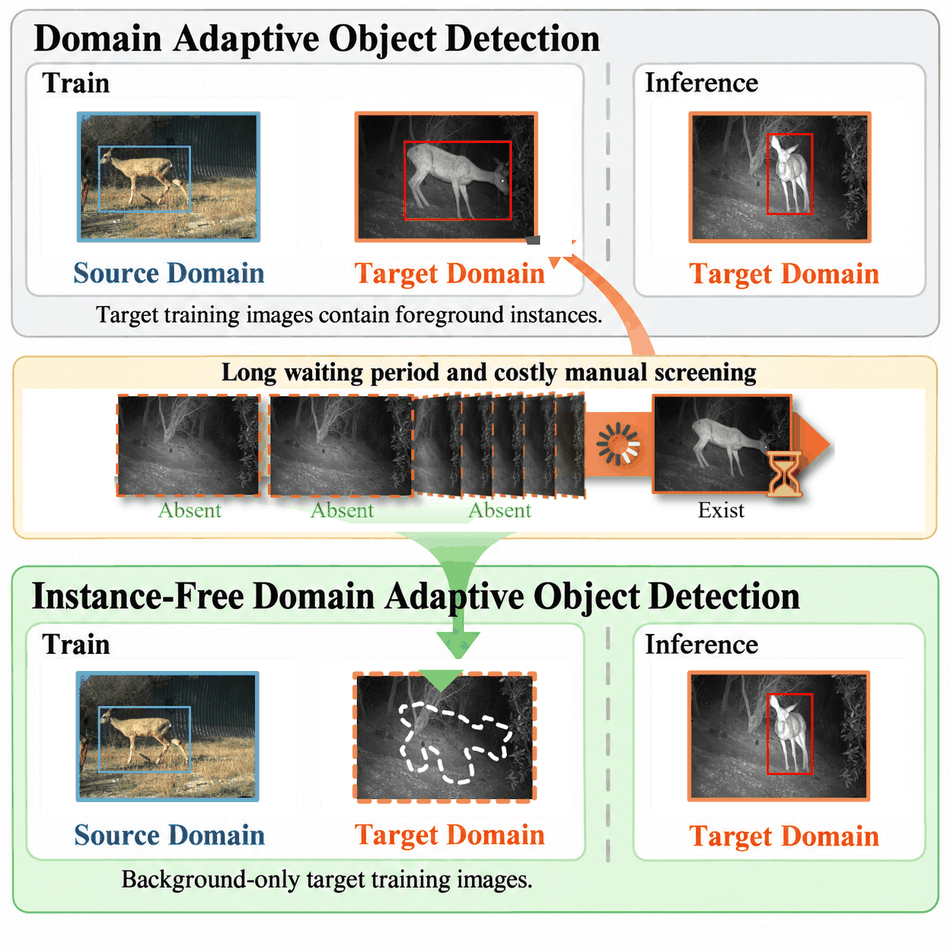}\end{minipage}\par
\vspace{3pt}
{\scriptsize \setlength{\tabcolsep}{1.5pt}\begin{tabular*}{\linewidth}{@{\extracolsep{\fill}}rrrrrrr@{}}
\toprule
C & O & G & A & I & S & Mean \\
4 (+0) & 4 (+0) & 4 (+0) & 3 (+0) & 4 (+0) & 5 (+0) & 4.00 (+0.00)\\
\bottomrule
\end{tabular*}
}
\end{minipage}\par
\vspace{3pt}
{\scriptsize Source public-date proxy: 2026-02-06. All displayed scores are medians of three runs; parentheses show revised minus original.\par}
\vspace{5pt}
{\bfseries Framework (source Fig.\ 2)} \hfill {\footnotesize \href{https://arxiv.org/pdf/2602.06484v1}{Source PDF, p.\ 3}; ID: P288}\par
\vspace{2pt}
\noindent\begin{minipage}[t]{0.486\linewidth}
\centering\textbf{Original}\par\vspace{2pt}
\begin{minipage}[c][0.255\textheight][c]{\linewidth}\centering\includegraphics[width=\linewidth,height=0.255\textheight,keepaspectratio]{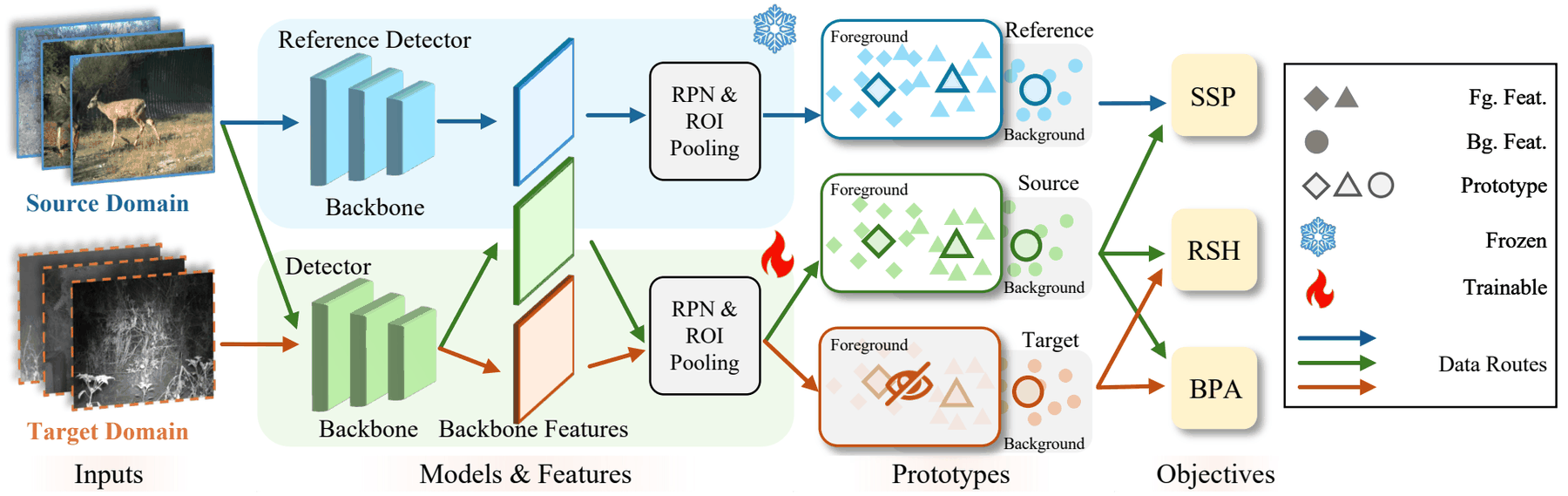}\end{minipage}\par
\vspace{3pt}{\scriptsize \setlength{\tabcolsep}{1.5pt}\begin{tabular*}{\linewidth}{@{\extracolsep{\fill}}rrrrrrr@{}}
\toprule
C & O & G & A & I & S & Mean \\
4 & 4 & 4 & 4 & 3 & 4 & 3.83\\
\bottomrule
\end{tabular*}
}
\end{minipage}\hfill
\begin{minipage}[t]{0.486\linewidth}
\centering\textbf{Study-generated derivative}\par\vspace{2pt}
\begin{minipage}[c][0.255\textheight][c]{\linewidth}\centering\includegraphics[width=\linewidth,height=0.255\textheight,keepaspectratio]{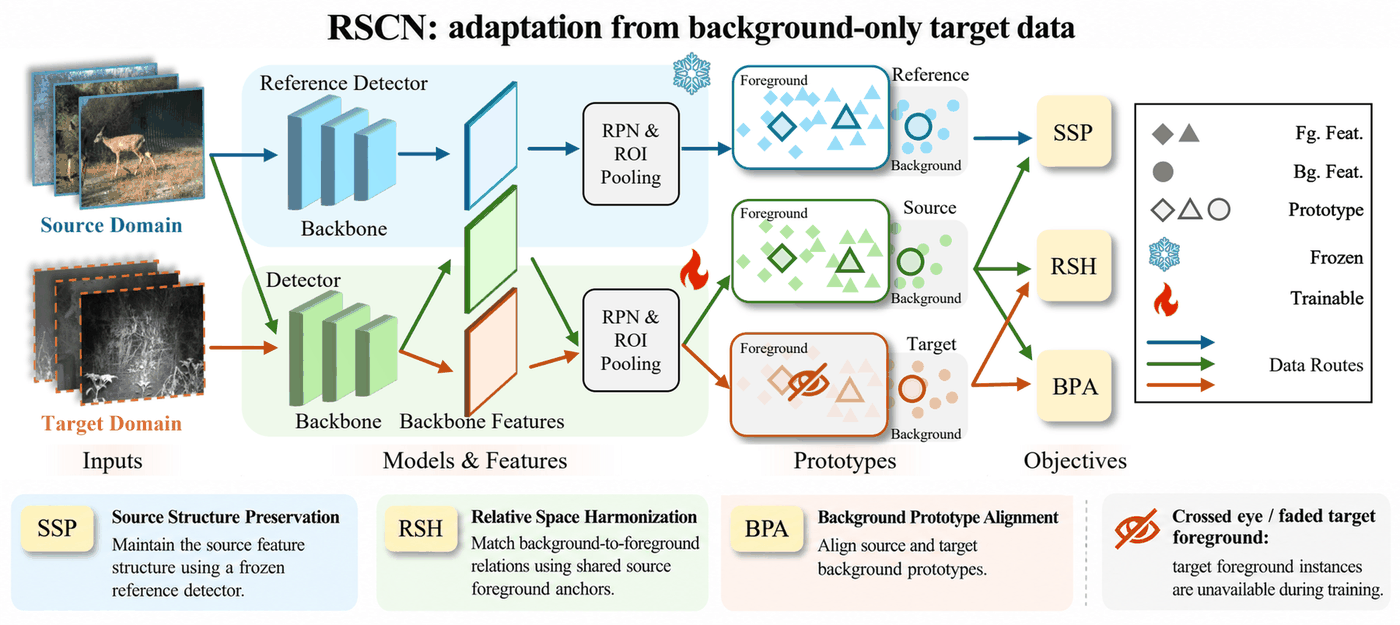}\end{minipage}\par
\vspace{3pt}
{\scriptsize \setlength{\tabcolsep}{1.5pt}\begin{tabular*}{\linewidth}{@{\extracolsep{\fill}}rrrrrrr@{}}
\toprule
C & O & G & A & I & S & Mean \\
4 (+0) & 4 (+0) & 4 (+0) & 4 (+0) & 4 (+1) & 5 (+1) & 4.17 (+0.33)\\
\bottomrule
\end{tabular*}
}
\end{minipage}\par
\vspace{3pt}
{\scriptsize Source public-date proxy: 2026-02-06. All displayed scores are medians of three runs; parentheses show revised minus original.\par}
\vfill
{\scriptsize C: clarity; O: organization/readability; G: graphical consistency; A: aesthetic coherence; I: visual inventiveness; S: design specificity. Mean is the arithmetic mean of six dimension medians, a descriptive display value, not a validated quality index.\par}

\clearpage
\phantomsection
\label{atlas:corpus087}
\pdfbookmark[1]{87. LAB-Det: Language as a Domain-Invariant Bridge for Training-Free One-Shot Domain Generalization in Object Detection}{corpus087}
{\Large\bfseries\color{academicblue} 087 \quad \href{https://arxiv.org/abs/2602.06474}{LAB-Det: Language as a Domain-Invariant Bridge for Training-Free One-Shot Domain Generalization in Object Detection} \citep{corpus087}\par}
\vspace{3pt}
{\small S3 \enspace $\vert$ \enspace Object detection \enspace $\vert$ \enspace arXiv 2602.06474\par}
\vspace{4pt}
\vspace{5pt}
{\bfseries Motivation (source Fig.\ 1)} \hfill {\footnotesize \href{https://arxiv.org/pdf/2602.06474v1}{Source PDF, p.\ 1}; ID: P289}\par
\vspace{2pt}
\noindent\begin{minipage}[t]{0.486\linewidth}
\centering\textbf{Original}\par\vspace{2pt}
\begin{minipage}[c][0.255\textheight][c]{\linewidth}\centering\includegraphics[width=\linewidth,height=0.255\textheight,keepaspectratio]{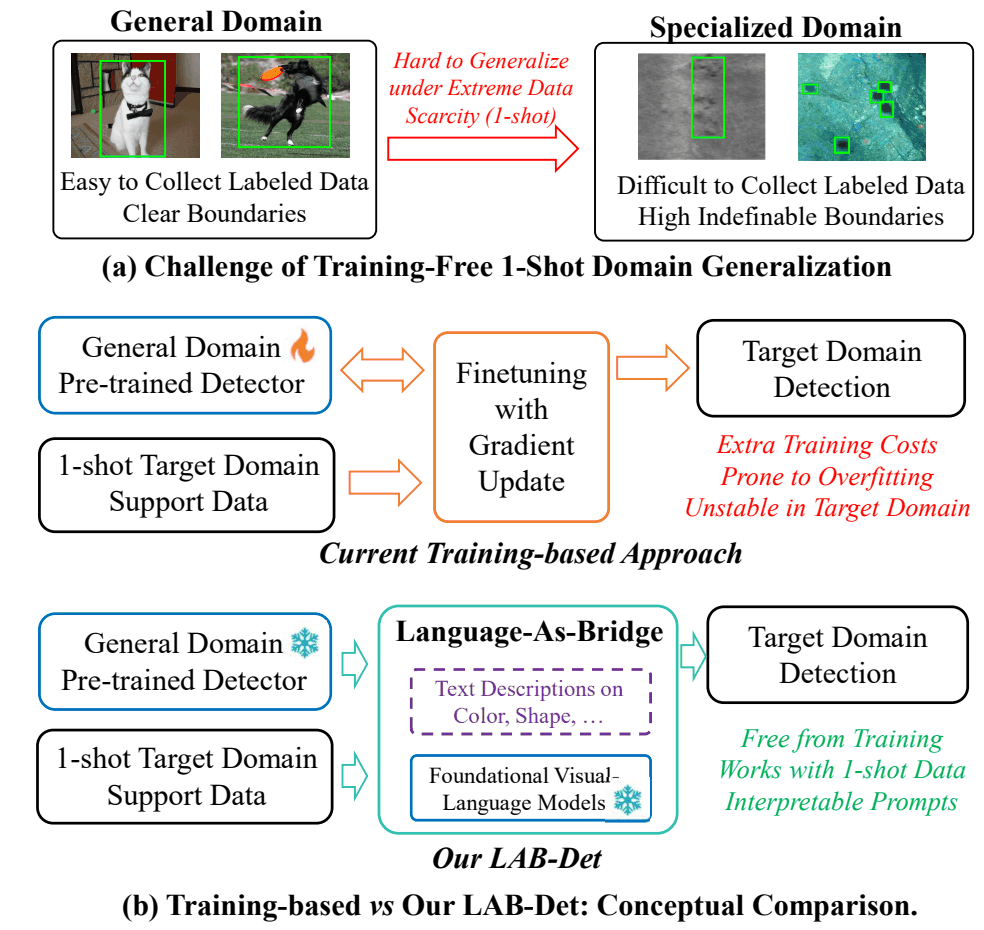}\end{minipage}\par
\vspace{3pt}{\scriptsize \setlength{\tabcolsep}{1.5pt}\begin{tabular*}{\linewidth}{@{\extracolsep{\fill}}rrrrrrr@{}}
\toprule
C & O & G & A & I & S & Mean \\
5 & 4 & 4 & 3 & 3 & 4 & 3.83\\
\bottomrule
\end{tabular*}
}
\end{minipage}\hfill
\begin{minipage}[t]{0.486\linewidth}
\centering\textbf{Study-generated derivative}\par\vspace{2pt}
\begin{minipage}[c][0.255\textheight][c]{\linewidth}\centering\includegraphics[width=\linewidth,height=0.255\textheight,keepaspectratio]{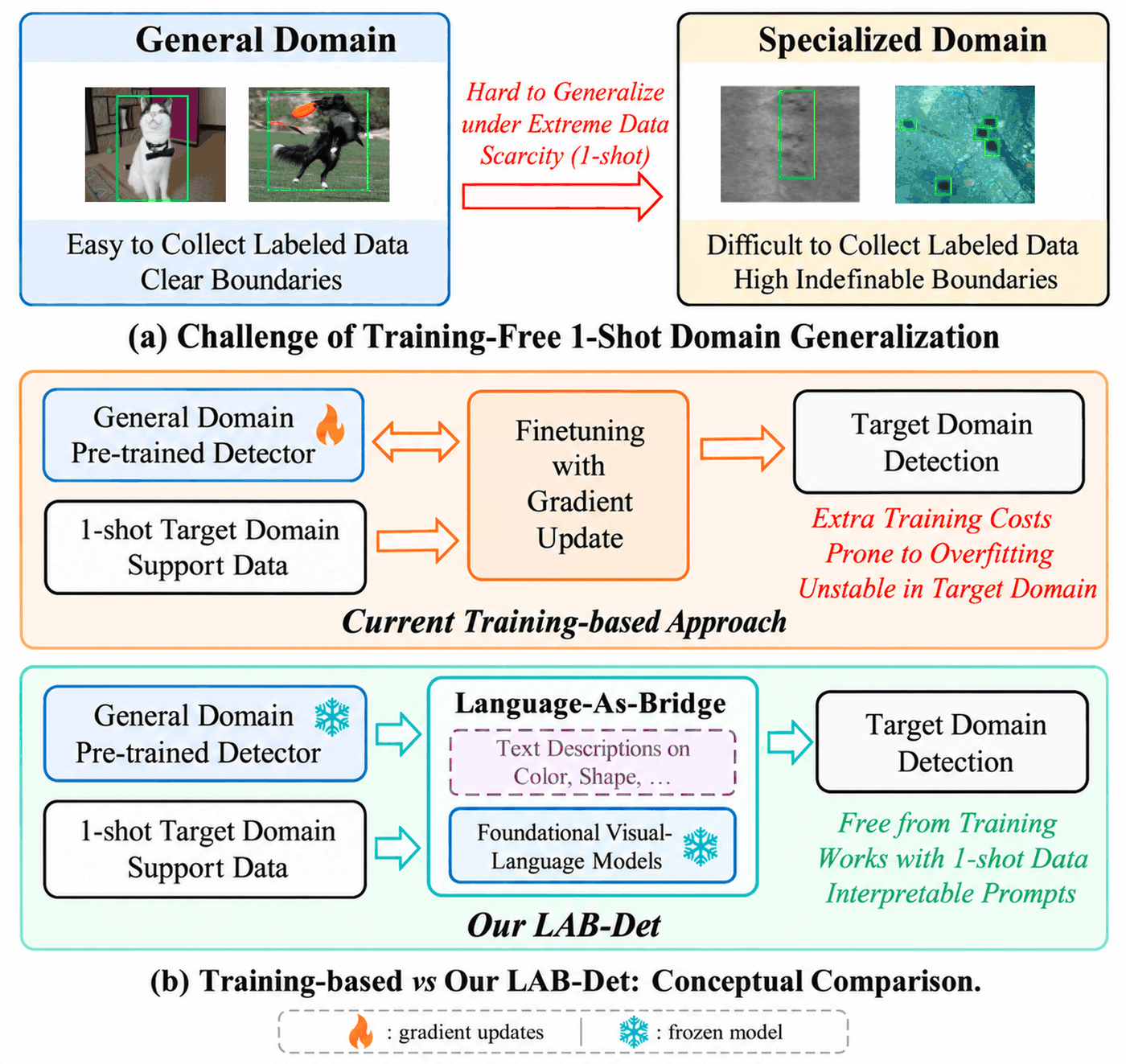}\end{minipage}\par
\vspace{3pt}
{\scriptsize \setlength{\tabcolsep}{1.5pt}\begin{tabular*}{\linewidth}{@{\extracolsep{\fill}}rrrrrrr@{}}
\toprule
C & O & G & A & I & S & Mean \\
5 (+0) & 4 (+0) & 5 (+1) & 4 (+1) & 3 (+0) & 4 (+0) & 4.17 (+0.33)\\
\bottomrule
\end{tabular*}
}
\end{minipage}\par
\vspace{3pt}
{\scriptsize Source public-date proxy: 2026-02-06. All displayed scores are medians of three runs; parentheses show revised minus original.\par}
\vspace{5pt}
{\bfseries Framework (source Fig.\ 2)} \hfill {\footnotesize \href{https://arxiv.org/pdf/2602.06474v1}{Source PDF, p.\ 4}; ID: P290}\par
\vspace{2pt}
\noindent\begin{minipage}[t]{0.486\linewidth}
\centering\textbf{Original}\par\vspace{2pt}
\begin{minipage}[c][0.255\textheight][c]{\linewidth}\centering\includegraphics[width=\linewidth,height=0.255\textheight,keepaspectratio]{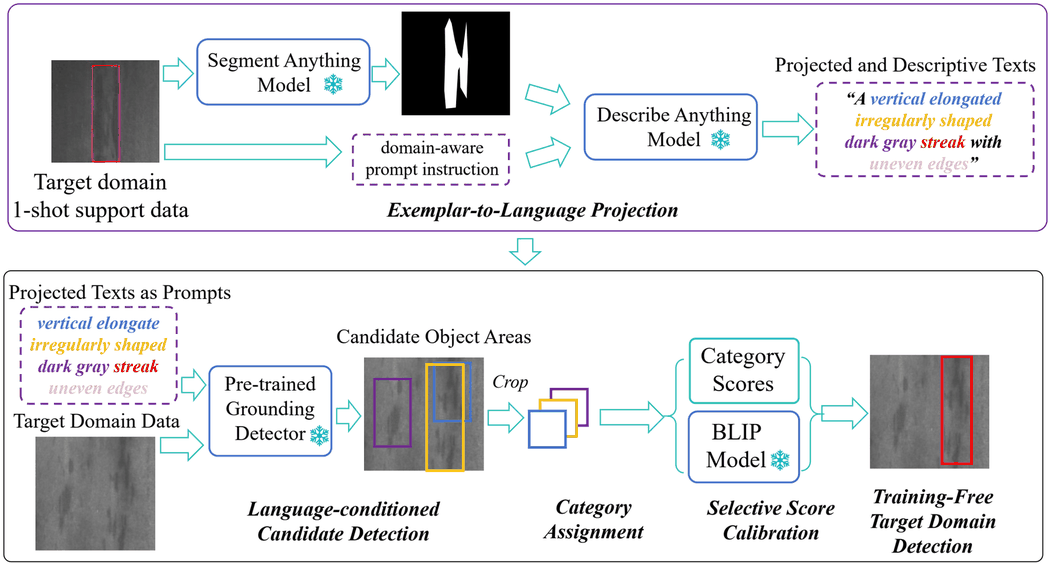}\end{minipage}\par
\vspace{3pt}{\scriptsize \setlength{\tabcolsep}{1.5pt}\begin{tabular*}{\linewidth}{@{\extracolsep{\fill}}rrrrrrr@{}}
\toprule
C & O & G & A & I & S & Mean \\
4 & 4 & 4 & 3 & 3 & 4 & 3.67\\
\bottomrule
\end{tabular*}
}
\end{minipage}\hfill
\begin{minipage}[t]{0.486\linewidth}
\centering\textbf{Study-generated derivative}\par\vspace{2pt}
\begin{minipage}[c][0.255\textheight][c]{\linewidth}\centering\includegraphics[width=\linewidth,height=0.255\textheight,keepaspectratio]{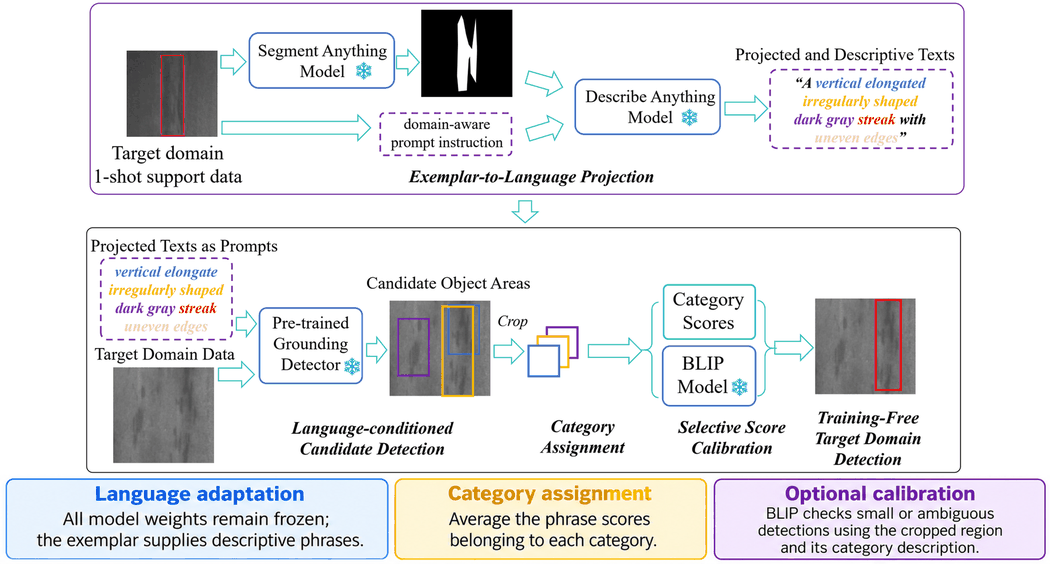}\end{minipage}\par
\vspace{3pt}
{\scriptsize \setlength{\tabcolsep}{1.5pt}\begin{tabular*}{\linewidth}{@{\extracolsep{\fill}}rrrrrrr@{}}
\toprule
C & O & G & A & I & S & Mean \\
4 (+0) & 4 (+0) & 4 (+0) & 3 (+0) & 3 (+0) & 4 (+0) & 3.67 (+0.00)\\
\bottomrule
\end{tabular*}
}
\end{minipage}\par
\vspace{3pt}
{\scriptsize Source public-date proxy: 2026-02-06. All displayed scores are medians of three runs; parentheses show revised minus original.\par}
\vfill
{\scriptsize C: clarity; O: organization/readability; G: graphical consistency; A: aesthetic coherence; I: visual inventiveness; S: design specificity. Mean is the arithmetic mean of six dimension medians, a descriptive display value, not a validated quality index.\par}

\clearpage
\phantomsection
\label{atlas:corpus098}
\pdfbookmark[1]{98. Rewis3d: Reconstruction Improves Weakly-Supervised Semantic Segmentation}{corpus098}
{\Large\bfseries\color{academicblue} 098 \quad \href{https://openaccess.thecvf.com/content/CVPR2026/html/Ernst_Rewis3d_Reconstruction_Improves_Weakly-Supervised_Semantic_Segmentation_CVPR_2026_paper.html}{Rewis3d: Reconstruction Improves Weakly-Supervised Semantic Segmentation} \citep{corpus098}\par}
\vspace{3pt}
{\small S3 \enspace $\vert$ \enspace Image segmentation \enspace $\vert$ \enspace CVPR 2026\par}
\vspace{4pt}
\vspace{5pt}
{\bfseries Motivation (source Fig.\ 2)} \hfill {\footnotesize \href{https://openaccess.thecvf.com/content/CVPR2026/papers/Ernst\_Rewis3d\_Reconstruction\_Improves\_Weakly-Supervised\_Semantic\_Segmentation\_CVPR\_2026\_paper.pdf}{Source PDF, p.\ 2}; ID: P05}\par
\vspace{2pt}
\noindent\begin{minipage}[t]{0.486\linewidth}
\centering\textbf{Original}\par\vspace{2pt}
\begin{minipage}[c][0.255\textheight][c]{\linewidth}\centering\includegraphics[width=\linewidth,height=0.255\textheight,keepaspectratio]{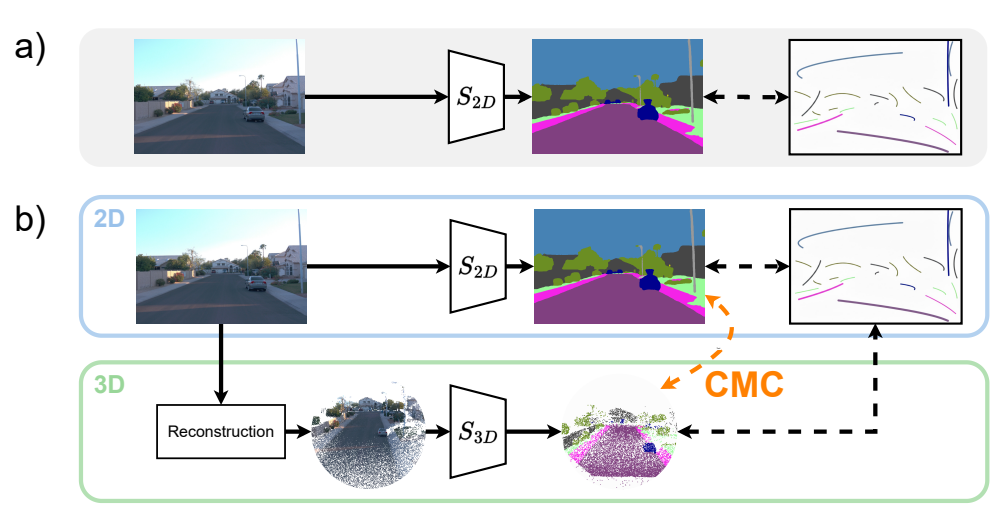}\end{minipage}\par
\vspace{3pt}{\scriptsize \setlength{\tabcolsep}{1.5pt}\begin{tabular*}{\linewidth}{@{\extracolsep{\fill}}rrrrrrr@{}}
\toprule
C & O & G & A & I & S & Mean \\
3 & 5 & 4 & 4 & 3 & 4 & 3.83\\
\bottomrule
\end{tabular*}
}
\end{minipage}\hfill
\begin{minipage}[t]{0.486\linewidth}
\centering\textbf{Study-generated derivative}\par\vspace{2pt}
\begin{minipage}[c][0.255\textheight][c]{\linewidth}\centering\includegraphics[width=\linewidth,height=0.255\textheight,keepaspectratio]{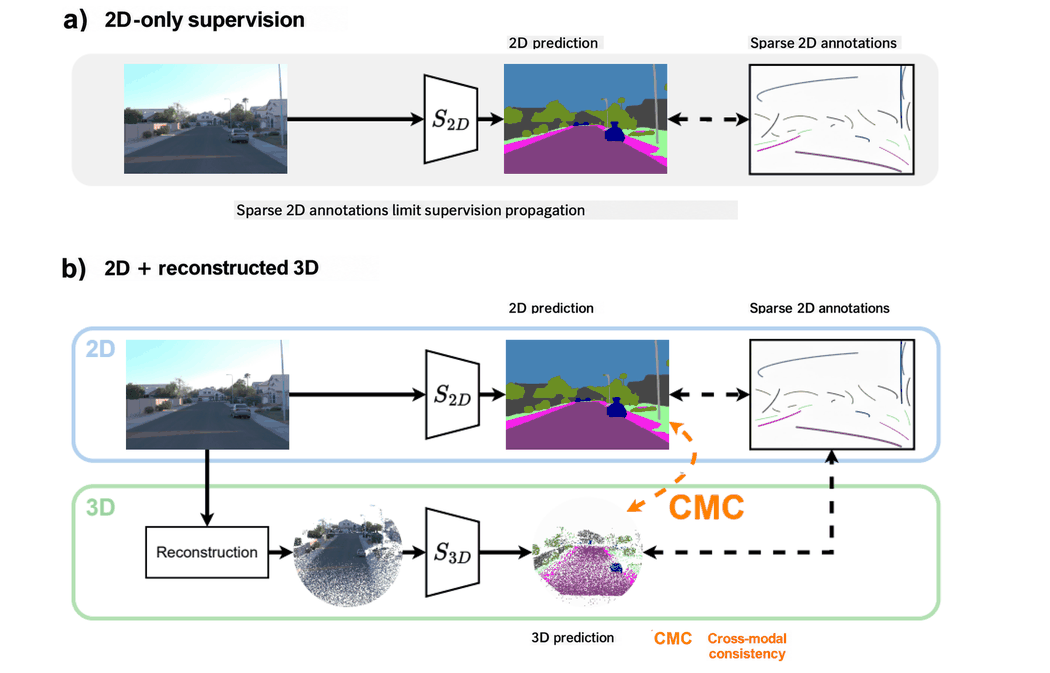}\end{minipage}\par
\vspace{3pt}
{\scriptsize \setlength{\tabcolsep}{1.5pt}\begin{tabular*}{\linewidth}{@{\extracolsep{\fill}}rrrrrrr@{}}
\toprule
C & O & G & A & I & S & Mean \\
4 (+1) & 4 (-1) & 4 (+0) & 3 (-1) & 3 (+0) & 4 (+0) & 3.67 (-0.17)\\
\bottomrule
\end{tabular*}
}
\end{minipage}\par
\vspace{3pt}
{\scriptsize Source public-date proxy: 2026-03-06. All displayed scores are medians of three runs; parentheses show revised minus original.\par}
\vspace{5pt}
{\bfseries Framework (source Fig.\ 3)} \hfill {\footnotesize \href{https://openaccess.thecvf.com/content/CVPR2026/papers/Ernst\_Rewis3d\_Reconstruction\_Improves\_Weakly-Supervised\_Semantic\_Segmentation\_CVPR\_2026\_paper.pdf}{Source PDF, p.\ 3}; ID: P06}\par
\vspace{2pt}
\noindent\begin{minipage}[t]{0.486\linewidth}
\centering\textbf{Original}\par\vspace{2pt}
\begin{minipage}[c][0.255\textheight][c]{\linewidth}\centering\includegraphics[width=\linewidth,height=0.255\textheight,keepaspectratio]{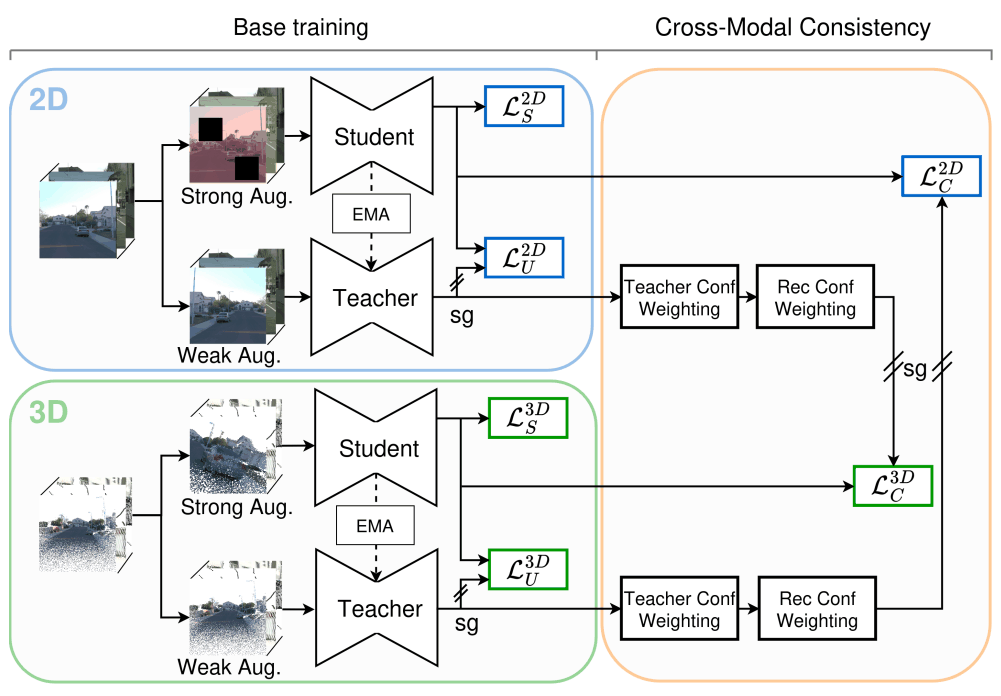}\end{minipage}\par
\vspace{3pt}{\scriptsize \setlength{\tabcolsep}{1.5pt}\begin{tabular*}{\linewidth}{@{\extracolsep{\fill}}rrrrrrr@{}}
\toprule
C & O & G & A & I & S & Mean \\
4 & 4 & 4 & 4 & 3 & 4 & 3.83\\
\bottomrule
\end{tabular*}
}
\end{minipage}\hfill
\begin{minipage}[t]{0.486\linewidth}
\centering\textbf{Study-generated derivative}\par\vspace{2pt}
\begin{minipage}[c][0.255\textheight][c]{\linewidth}\centering\includegraphics[width=\linewidth,height=0.255\textheight,keepaspectratio]{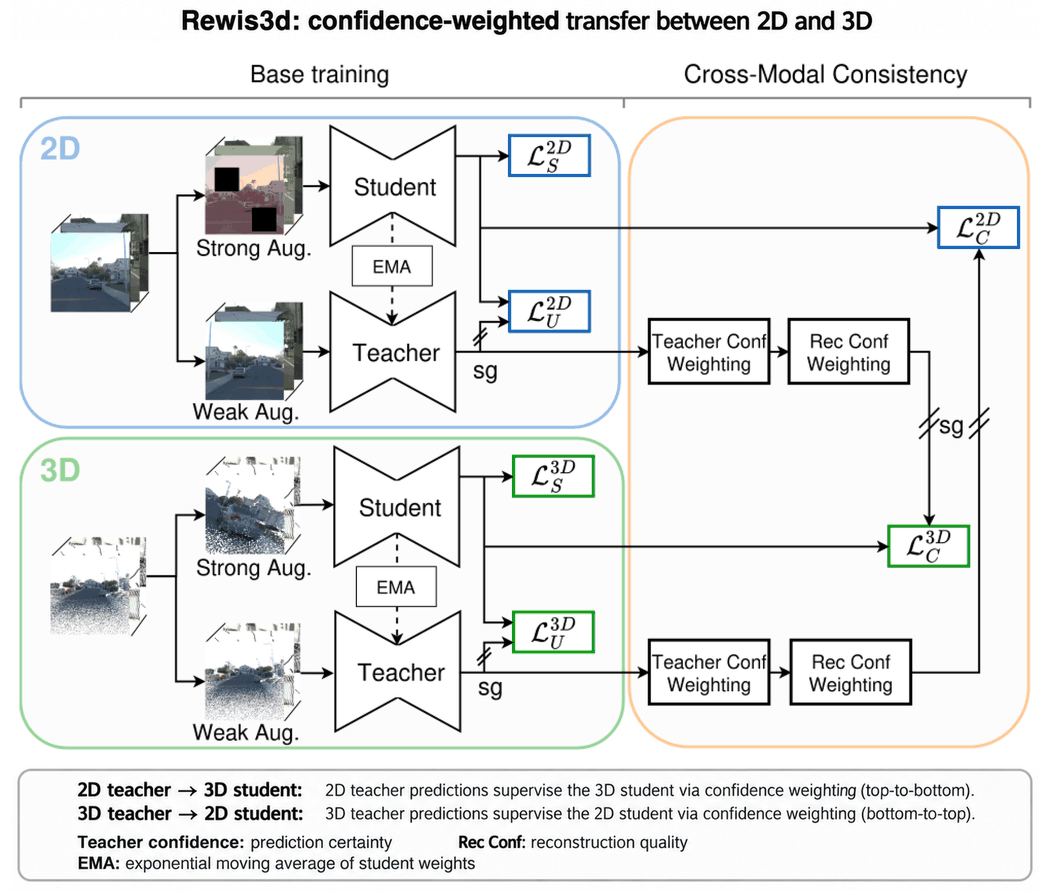}\end{minipage}\par
\vspace{3pt}
{\scriptsize \setlength{\tabcolsep}{1.5pt}\begin{tabular*}{\linewidth}{@{\extracolsep{\fill}}rrrrrrr@{}}
\toprule
C & O & G & A & I & S & Mean \\
4 (+0) & 4 (+0) & 4 (+0) & 3 (-1) & 3 (+0) & 4 (+0) & 3.67 (-0.17)\\
\bottomrule
\end{tabular*}
}
\end{minipage}\par
\vspace{3pt}
{\scriptsize Source public-date proxy: 2026-03-06. All displayed scores are medians of three runs; parentheses show revised minus original.\par}
\vfill
{\scriptsize C: clarity; O: organization/readability; G: graphical consistency; A: aesthetic coherence; I: visual inventiveness; S: design specificity. Mean is the arithmetic mean of six dimension medians, a descriptive display value, not a validated quality index.\par}

\clearpage
\phantomsection
\label{atlas:corpus100}
\pdfbookmark[1]{100. SegEarth-OV3: Exploring SAM 3 for Open-Vocabulary Semantic Segmentation in Remote Sensing Images}{corpus100}
{\Large\bfseries\color{academicblue} 100 \quad \href{https://arxiv.org/abs/2512.08730}{SegEarth-OV3: Exploring SAM 3 for Open-Vocabulary Semantic Segmentation in Remote Sensing Images} \citep{corpus100}\par}
\vspace{3pt}
{\small S3 \enspace $\vert$ \enspace Image segmentation \enspace $\vert$ \enspace arXiv 2512.08730\par}
\vspace{4pt}
\vspace{5pt}
{\bfseries Motivation (source Fig.\ 1)} \hfill {\footnotesize \href{https://arxiv.org/pdf/2512.08730v1}{Source PDF, p.\ 2}; ID: P293}\par
\vspace{2pt}
\noindent\begin{minipage}[t]{0.486\linewidth}
\centering\textbf{Original}\par\vspace{2pt}
\begin{minipage}[c][0.255\textheight][c]{\linewidth}\centering\includegraphics[width=\linewidth,height=0.255\textheight,keepaspectratio]{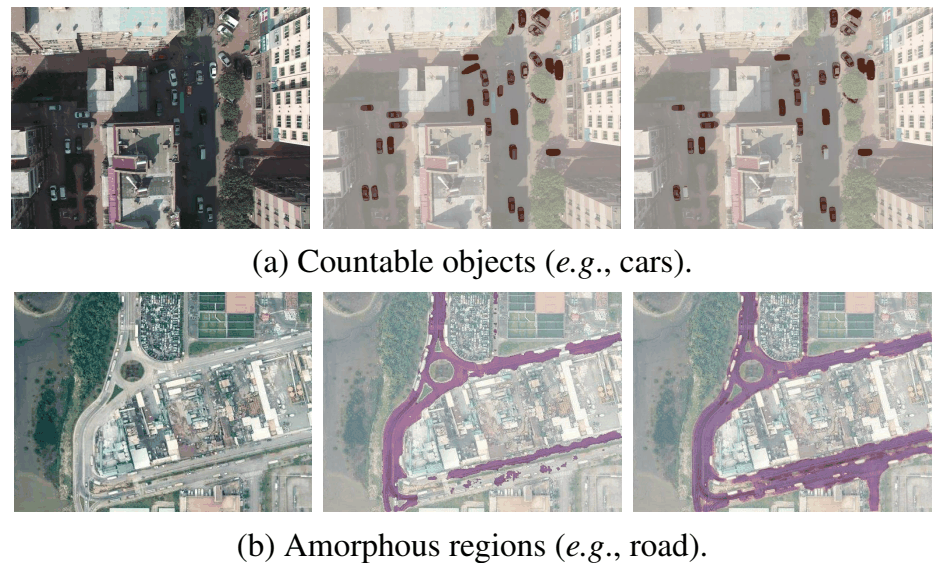}\end{minipage}\par
\vspace{3pt}{\scriptsize \setlength{\tabcolsep}{1.5pt}\begin{tabular*}{\linewidth}{@{\extracolsep{\fill}}rrrrrrr@{}}
\toprule
C & O & G & A & I & S & Mean \\
3 & 4 & 4 & 4 & 3 & 4 & 3.67\\
\bottomrule
\end{tabular*}
}
\end{minipage}\hfill
\begin{minipage}[t]{0.486\linewidth}
\centering\textbf{Study-generated derivative}\par\vspace{2pt}
\begin{minipage}[c][0.255\textheight][c]{\linewidth}\centering\includegraphics[width=\linewidth,height=0.255\textheight,keepaspectratio]{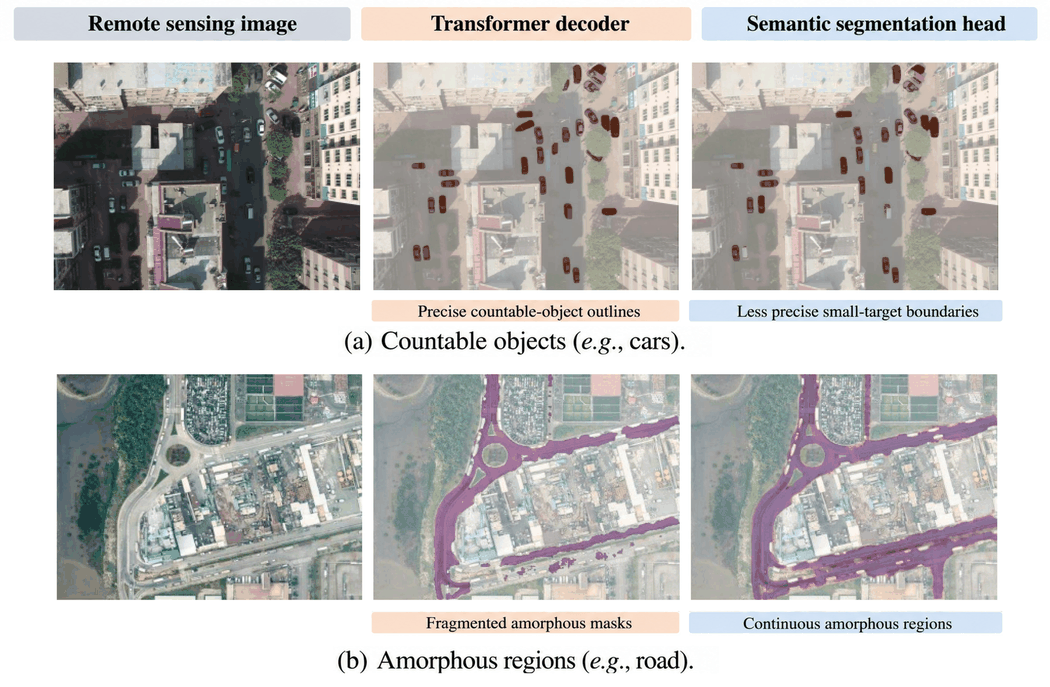}\end{minipage}\par
\vspace{3pt}
{\scriptsize \setlength{\tabcolsep}{1.5pt}\begin{tabular*}{\linewidth}{@{\extracolsep{\fill}}rrrrrrr@{}}
\toprule
C & O & G & A & I & S & Mean \\
4 (+1) & 5 (+1) & 5 (+1) & 4 (+0) & 4 (+1) & 5 (+1) & 4.50 (+0.83)\\
\bottomrule
\end{tabular*}
}
\end{minipage}\par
\vspace{3pt}
{\scriptsize Source public-date proxy: 2025-12-09. All displayed scores are medians of three runs; parentheses show revised minus original.\par}
\vspace{5pt}
{\bfseries Framework (source Fig.\ 2)} \hfill {\footnotesize \href{https://arxiv.org/pdf/2512.08730v1}{Source PDF, p.\ 3}; ID: P294}\par
\vspace{2pt}
\noindent\begin{minipage}[t]{0.486\linewidth}
\centering\textbf{Original}\par\vspace{2pt}
\begin{minipage}[c][0.255\textheight][c]{\linewidth}\centering\includegraphics[width=\linewidth,height=0.255\textheight,keepaspectratio]{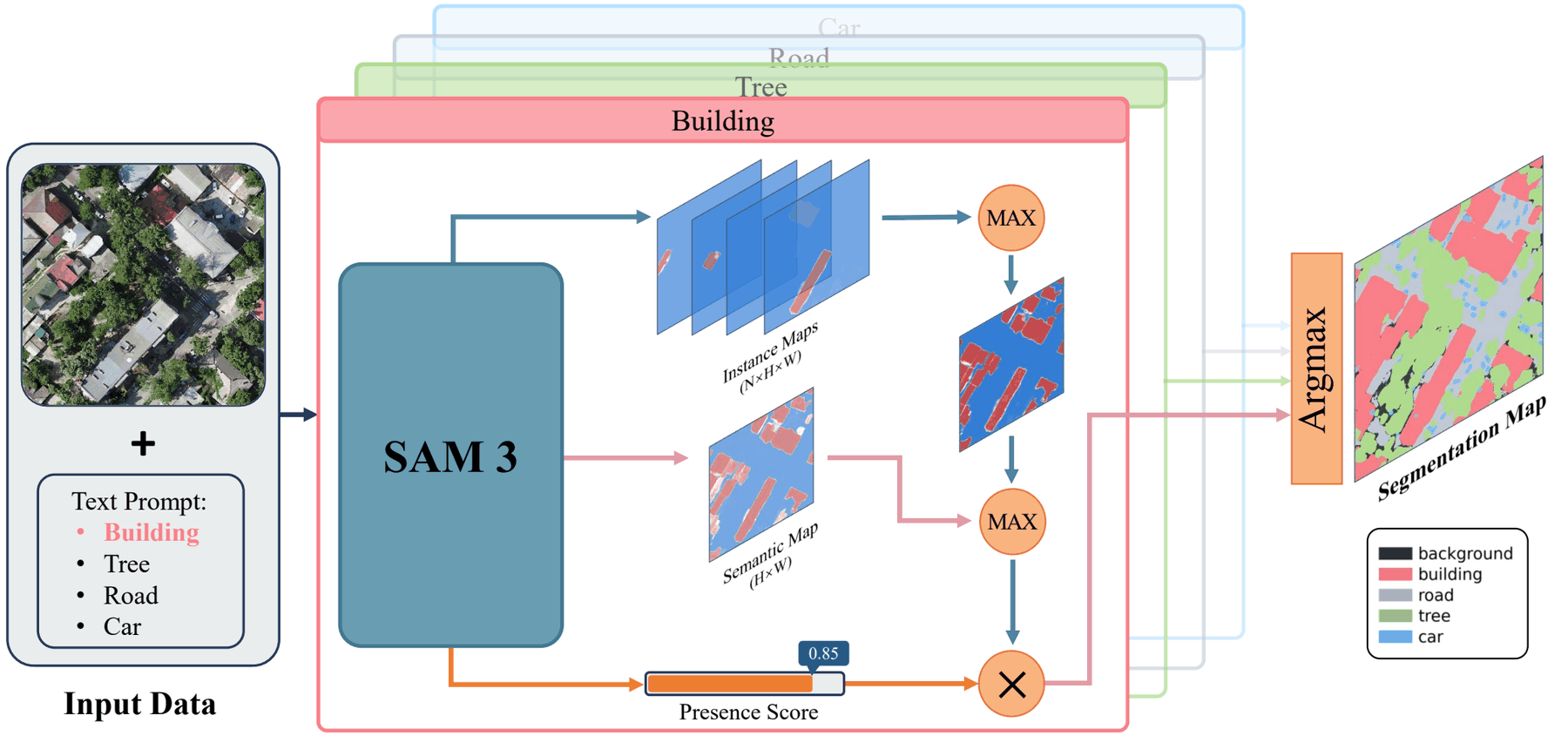}\end{minipage}\par
\vspace{3pt}{\scriptsize \setlength{\tabcolsep}{1.5pt}\begin{tabular*}{\linewidth}{@{\extracolsep{\fill}}rrrrrrr@{}}
\toprule
C & O & G & A & I & S & Mean \\
4 & 4 & 4 & 3 & 4 & 5 & 4.00\\
\bottomrule
\end{tabular*}
}
\end{minipage}\hfill
\begin{minipage}[t]{0.486\linewidth}
\centering\textbf{Study-generated derivative}\par\vspace{2pt}
\begin{minipage}[c][0.255\textheight][c]{\linewidth}\centering\includegraphics[width=\linewidth,height=0.255\textheight,keepaspectratio]{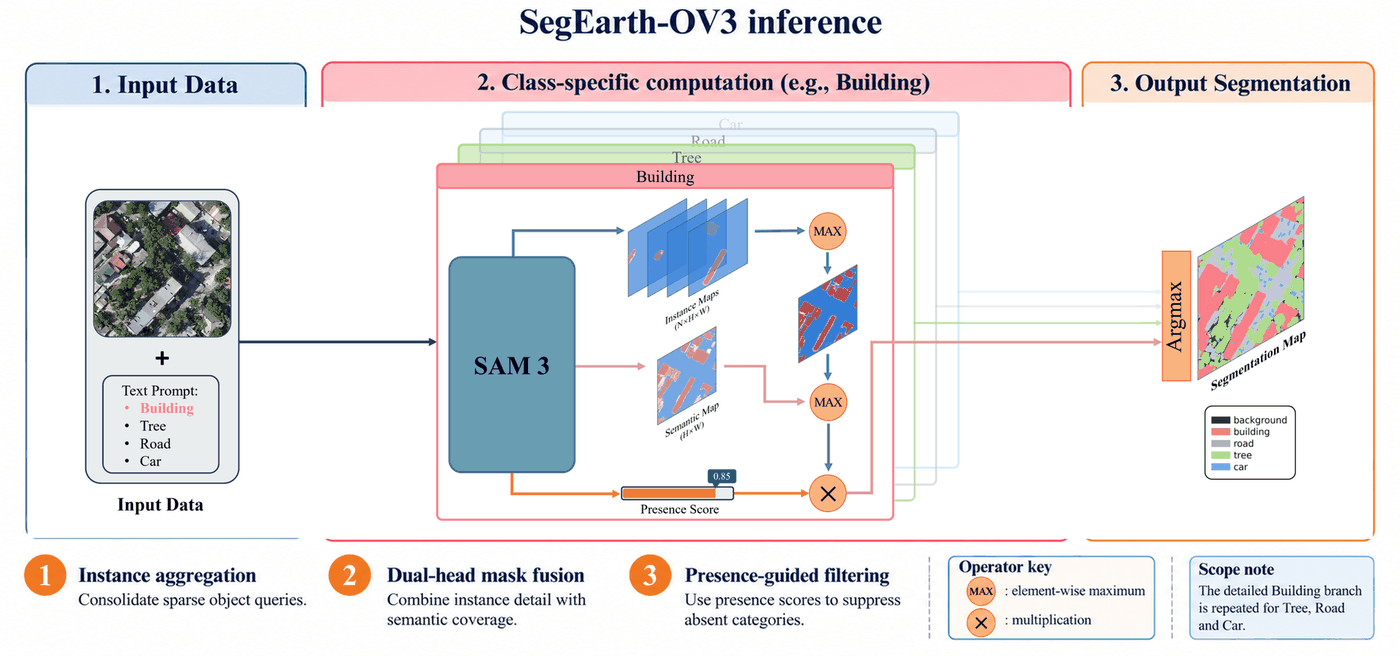}\end{minipage}\par
\vspace{3pt}
{\scriptsize \setlength{\tabcolsep}{1.5pt}\begin{tabular*}{\linewidth}{@{\extracolsep{\fill}}rrrrrrr@{}}
\toprule
C & O & G & A & I & S & Mean \\
5 (+1) & 4 (+0) & 4 (+0) & 4 (+1) & 4 (+0) & 5 (+0) & 4.33 (+0.33)\\
\bottomrule
\end{tabular*}
}
\end{minipage}\par
\vspace{3pt}
{\scriptsize Source public-date proxy: 2025-12-09. All displayed scores are medians of three runs; parentheses show revised minus original.\par}
\vfill
{\scriptsize C: clarity; O: organization/readability; G: graphical consistency; A: aesthetic coherence; I: visual inventiveness; S: design specificity. Mean is the arithmetic mean of six dimension medians, a descriptive display value, not a validated quality index.\par}

\clearpage
\phantomsection
\label{atlas:corpus102}
\pdfbookmark[1]{102. Enhancing Unregistered Hyperspectral Image Super-Resolution via Unmixing-based Abundance Fusion Learning}{corpus102}
{\Large\bfseries\color{academicblue} 102 \quad \href{https://openaccess.thecvf.com/content/CVPR2026/html/Zhang\_Enhancing\_Unregistered\_Hyperspectral\_Image\_Super-Resolution\_via\_Unmixing-based\_Abundance\_Fusion\_Learning\_CVPR\_2026\_paper.html}{Enhancing Unregistered Hyperspectral Image Super-Resolution via Unmixing-based Abundance Fusion Learning} \citep{corpus102}\par}
\vspace{3pt}
{\small S3 \enspace $\vert$ \enspace Image restoration \enspace $\vert$ \enspace CVPR 2026\par}
\vspace{4pt}
\vspace{5pt}
{\bfseries Motivation (source Fig.\ 2)} \hfill {\footnotesize \href{https://openaccess.thecvf.com/content/CVPR2026/papers/Zhang\_Enhancing\_Unregistered\_Hyperspectral\_Image\_Super-Resolution\_via\_Unmixing-based\_Abundance\_Fusion\_Learning\_CVPR\_2026\_paper.pdf}{Source PDF, p.\ 2}; ID: P149}\par
\vspace{2pt}
\noindent\begin{minipage}[t]{0.486\linewidth}
\centering\textbf{Original}\par\vspace{2pt}
\begin{minipage}[c][0.255\textheight][c]{\linewidth}\centering\includegraphics[width=\linewidth,height=0.255\textheight,keepaspectratio]{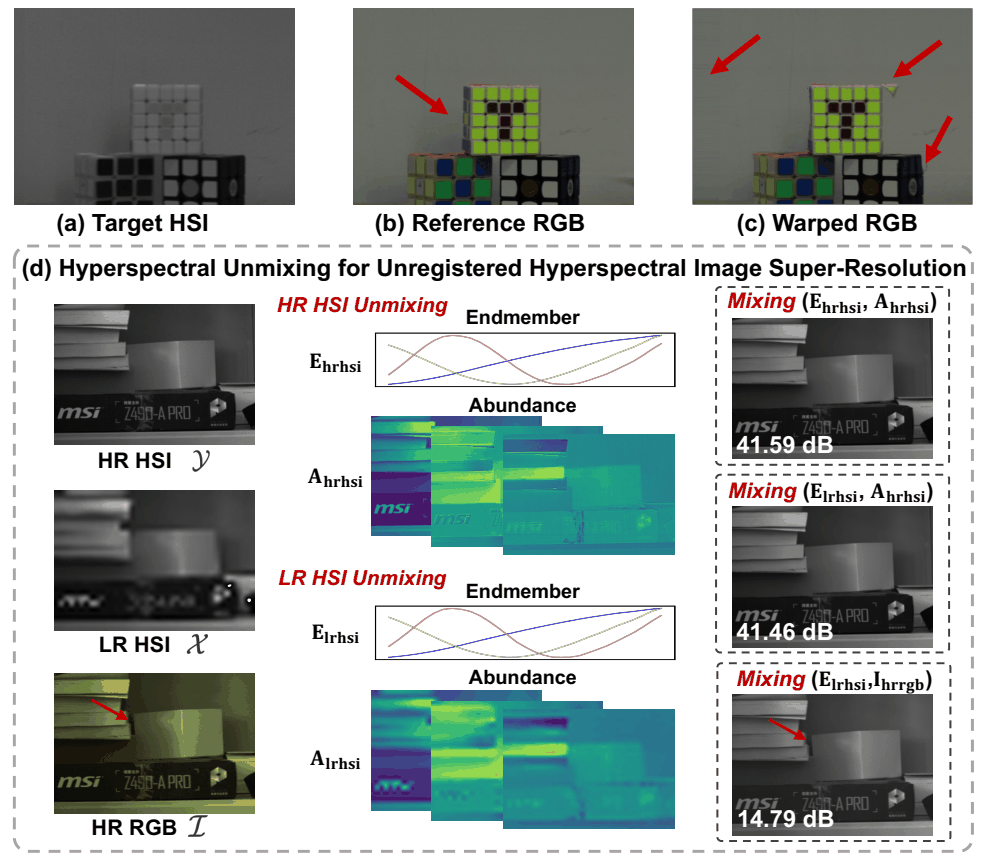}\end{minipage}\par
\vspace{3pt}{\scriptsize \setlength{\tabcolsep}{1.5pt}\begin{tabular*}{\linewidth}{@{\extracolsep{\fill}}rrrrrrr@{}}
\toprule
C & O & G & A & I & S & Mean \\
3 & 3 & 4 & 3 & 4 & 5 & 3.67\\
\bottomrule
\end{tabular*}
}
\end{minipage}\hfill
\begin{minipage}[t]{0.486\linewidth}
\centering\textbf{Study-generated derivative}\par\vspace{2pt}
\begin{minipage}[c][0.255\textheight][c]{\linewidth}\centering\includegraphics[width=\linewidth,height=0.255\textheight,keepaspectratio]{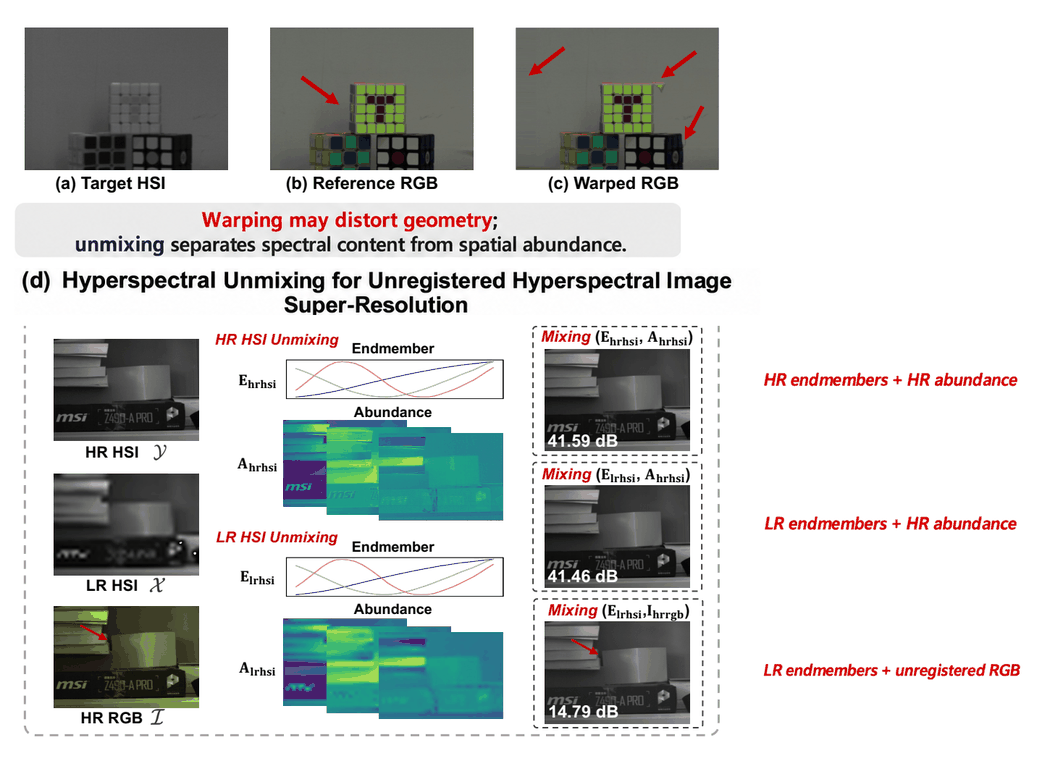}\end{minipage}\par
\vspace{3pt}
{\scriptsize \setlength{\tabcolsep}{1.5pt}\begin{tabular*}{\linewidth}{@{\extracolsep{\fill}}rrrrrrr@{}}
\toprule
C & O & G & A & I & S & Mean \\
4 (+1) & 3 (+0) & 4 (+0) & 3 (+0) & 4 (+0) & 5 (+0) & 3.83 (+0.17)\\
\bottomrule
\end{tabular*}
}
\end{minipage}\par
\vspace{3pt}
{\scriptsize Source public-date proxy: 2026-03-09. All displayed scores are medians of three runs; parentheses show revised minus original.\par}
\vspace{5pt}
{\bfseries Framework (source Fig.\ 3)} \hfill {\footnotesize \href{https://openaccess.thecvf.com/content/CVPR2026/papers/Zhang\_Enhancing\_Unregistered\_Hyperspectral\_Image\_Super-Resolution\_via\_Unmixing-based\_Abundance\_Fusion\_Learning\_CVPR\_2026\_paper.pdf}{Source PDF, p.\ 3}; ID: P150}\par
\vspace{2pt}
\noindent\begin{minipage}[t]{0.486\linewidth}
\centering\textbf{Original}\par\vspace{2pt}
\begin{minipage}[c][0.255\textheight][c]{\linewidth}\centering\includegraphics[width=\linewidth,height=0.255\textheight,keepaspectratio]{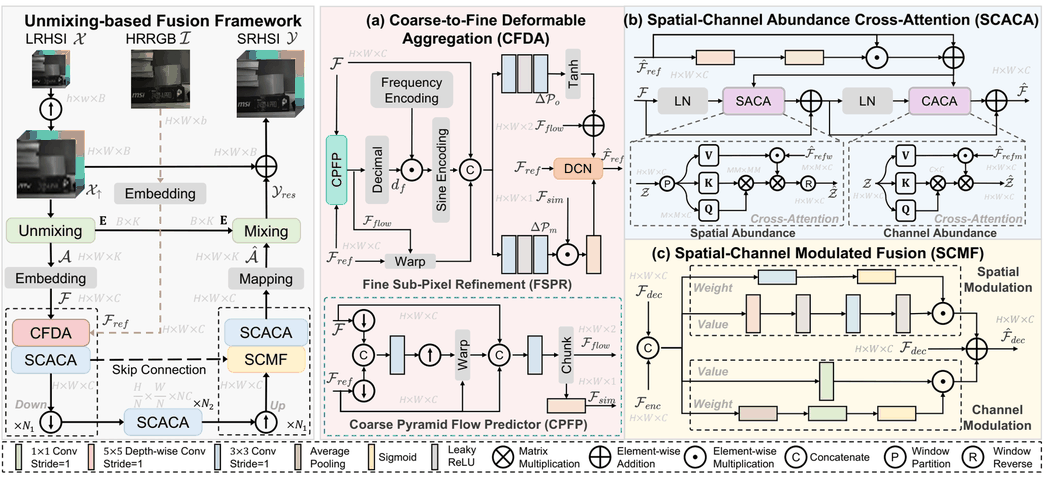}\end{minipage}\par
\vspace{3pt}{\scriptsize \setlength{\tabcolsep}{1.5pt}\begin{tabular*}{\linewidth}{@{\extracolsep{\fill}}rrrrrrr@{}}
\toprule
C & O & G & A & I & S & Mean \\
4 & 3 & 5 & 4 & 3 & 5 & 4.00\\
\bottomrule
\end{tabular*}
}
\end{minipage}\hfill
\begin{minipage}[t]{0.486\linewidth}
\centering\textbf{Study-generated derivative}\par\vspace{2pt}
\begin{minipage}[c][0.255\textheight][c]{\linewidth}\centering\includegraphics[width=\linewidth,height=0.255\textheight,keepaspectratio]{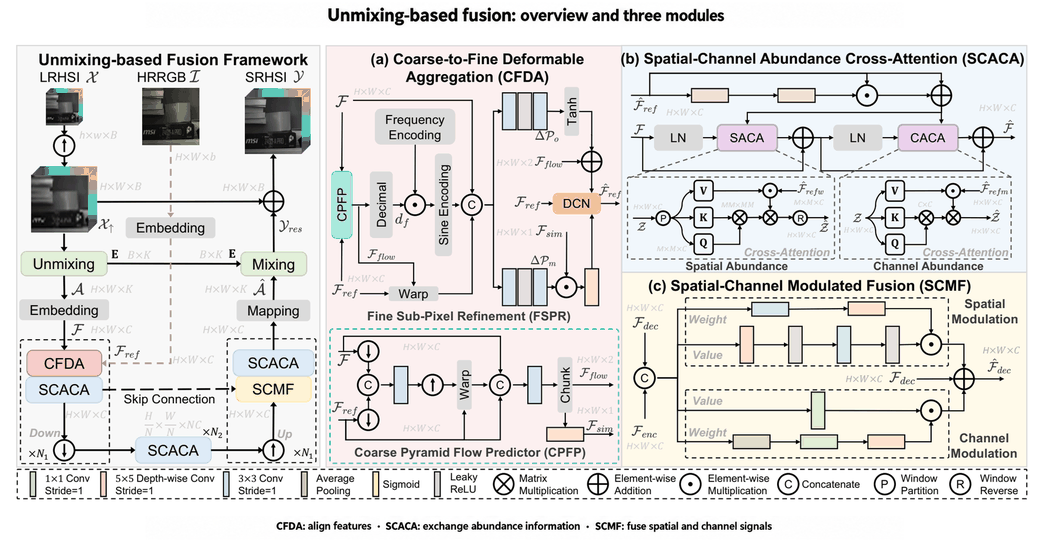}\end{minipage}\par
\vspace{3pt}
{\scriptsize \setlength{\tabcolsep}{1.5pt}\begin{tabular*}{\linewidth}{@{\extracolsep{\fill}}rrrrrrr@{}}
\toprule
C & O & G & A & I & S & Mean \\
4 (+0) & 3 (+0) & 4 (-1) & 3 (-1) & 3 (+0) & 5 (+0) & 3.67 (-0.33)\\
\bottomrule
\end{tabular*}
}
\end{minipage}\par
\vspace{3pt}
{\scriptsize Source public-date proxy: 2026-03-09. All displayed scores are medians of three runs; parentheses show revised minus original.\par}
\vfill
{\scriptsize C: clarity; O: organization/readability; G: graphical consistency; A: aesthetic coherence; I: visual inventiveness; S: design specificity. Mean is the arithmetic mean of six dimension medians, a descriptive display value, not a validated quality index.\par}

\clearpage
\phantomsection
\label{atlas:corpus110}
\pdfbookmark[1]{110. UARE: A Unified Vision-Language Model for Image Quality Assessment, Restoration, and Enhancement}{corpus110}
{\Large\bfseries\color{academicblue} 110 \quad \href{https://openaccess.thecvf.com/content/CVPR2026/html/Li_UARE_A_Unified_Vision-Language_Model_for_Image_Quality_Assessment_Restoration_CVPR_2026_paper.html}{UARE: A Unified Vision-Language Model for Image Quality Assessment, Restoration, and Enhancement} \citep{corpus110}\par}
\vspace{3pt}
{\small S3 \enspace $\vert$ \enspace Image restoration \enspace $\vert$ \enspace CVPR 2026\par}
\vspace{4pt}
\vspace{5pt}
{\bfseries Motivation (source Fig.\ 1)} \hfill {\footnotesize \href{https://openaccess.thecvf.com/content/CVPR2026/papers/Li\_UARE\_A\_Unified\_Vision-Language\_Model\_for\_Image\_Quality\_Assessment\_Restoration\_CVPR\_2026\_paper.pdf}{Source PDF, p.\ 1}; ID: P37}\par
\vspace{2pt}
\noindent\begin{minipage}[t]{0.486\linewidth}
\centering\textbf{Original}\par\vspace{2pt}
\begin{minipage}[c][0.255\textheight][c]{\linewidth}\centering\includegraphics[width=\linewidth,height=0.255\textheight,keepaspectratio]{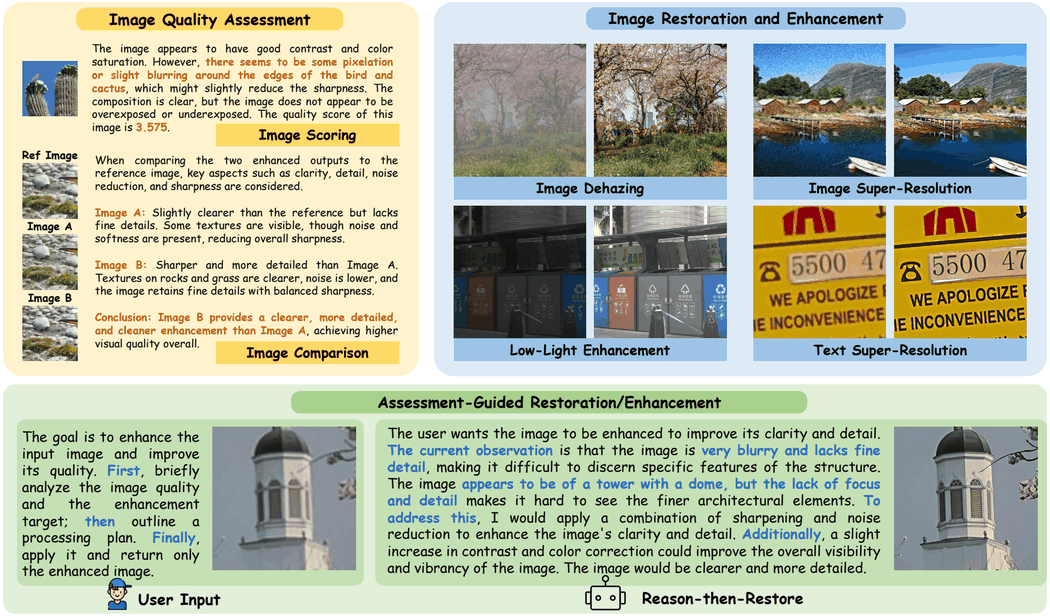}\end{minipage}\par
\vspace{3pt}{\scriptsize \setlength{\tabcolsep}{1.5pt}\begin{tabular*}{\linewidth}{@{\extracolsep{\fill}}rrrrrrr@{}}
\toprule
C & O & G & A & I & S & Mean \\
3 & 3 & 4 & 3 & 3 & 4 & 3.33\\
\bottomrule
\end{tabular*}
}
\end{minipage}\hfill
\begin{minipage}[t]{0.486\linewidth}
\centering\textbf{Study-generated derivative}\par\vspace{2pt}
\begin{minipage}[c][0.255\textheight][c]{\linewidth}\centering\includegraphics[width=\linewidth,height=0.255\textheight,keepaspectratio]{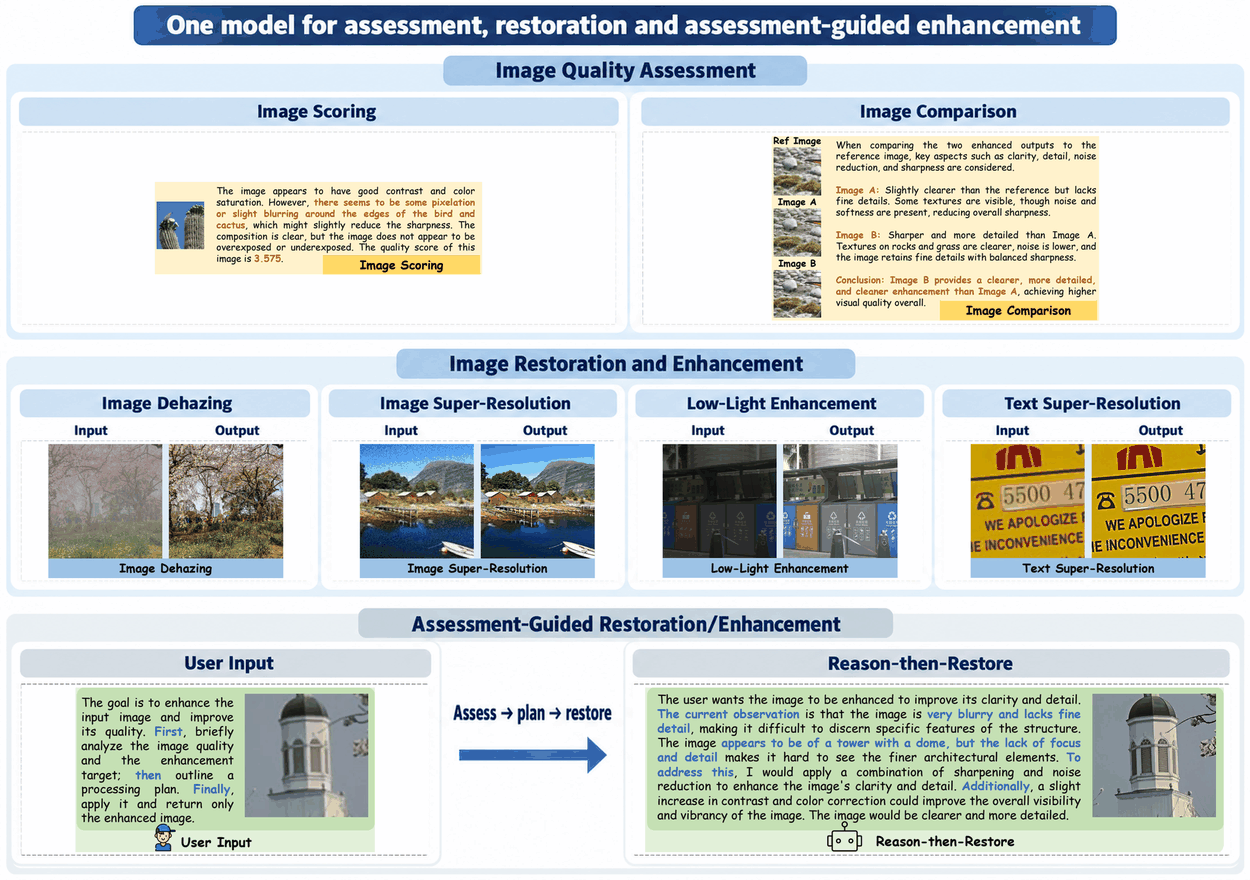}\end{minipage}\par
\vspace{3pt}
{\scriptsize \setlength{\tabcolsep}{1.5pt}\begin{tabular*}{\linewidth}{@{\extracolsep{\fill}}rrrrrrr@{}}
\toprule
C & O & G & A & I & S & Mean \\
4 (+1) & 3 (+0) & 5 (+1) & 3 (+0) & 3 (+0) & 4 (+0) & 3.67 (+0.33)\\
\bottomrule
\end{tabular*}
}
\end{minipage}\par
\vspace{3pt}
{\scriptsize Source public-date proxy: 2025-12-07. All displayed scores are medians of three runs; parentheses show revised minus original.\par}
\vspace{5pt}
{\bfseries Framework (source Fig.\ 2)} \hfill {\footnotesize \href{https://openaccess.thecvf.com/content/CVPR2026/papers/Li\_UARE\_A\_Unified\_Vision-Language\_Model\_for\_Image\_Quality\_Assessment\_Restoration\_CVPR\_2026\_paper.pdf}{Source PDF, p.\ 3}; ID: P38}\par
\vspace{2pt}
\noindent\begin{minipage}[t]{0.486\linewidth}
\centering\textbf{Original}\par\vspace{2pt}
\begin{minipage}[c][0.255\textheight][c]{\linewidth}\centering\includegraphics[width=\linewidth,height=0.255\textheight,keepaspectratio]{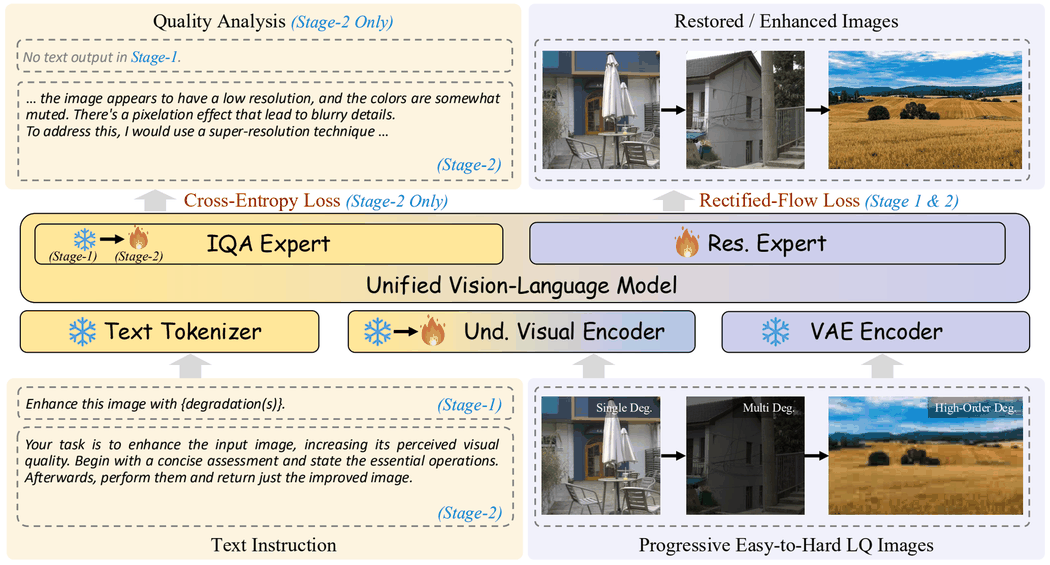}\end{minipage}\par
\vspace{3pt}{\scriptsize \setlength{\tabcolsep}{1.5pt}\begin{tabular*}{\linewidth}{@{\extracolsep{\fill}}rrrrrrr@{}}
\toprule
C & O & G & A & I & S & Mean \\
3 & 4 & 3 & 3 & 3 & 4 & 3.33\\
\bottomrule
\end{tabular*}
}
\end{minipage}\hfill
\begin{minipage}[t]{0.486\linewidth}
\centering\textbf{Study-generated derivative}\par\vspace{2pt}
\begin{minipage}[c][0.255\textheight][c]{\linewidth}\centering\includegraphics[width=\linewidth,height=0.255\textheight,keepaspectratio]{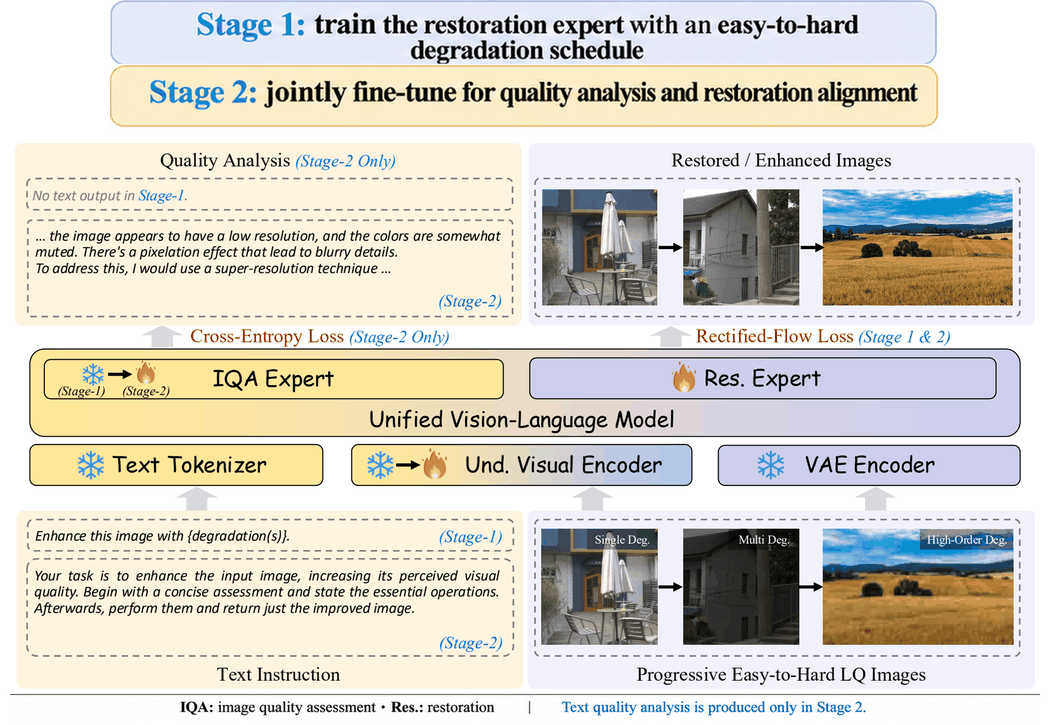}\end{minipage}\par
\vspace{3pt}
{\scriptsize \setlength{\tabcolsep}{1.5pt}\begin{tabular*}{\linewidth}{@{\extracolsep{\fill}}rrrrrrr@{}}
\toprule
C & O & G & A & I & S & Mean \\
3 (+0) & 4 (+0) & 3 (+0) & 3 (+0) & 4 (+1) & 5 (+1) & 3.67 (+0.33)\\
\bottomrule
\end{tabular*}
}
\end{minipage}\par
\vspace{3pt}
{\scriptsize Source public-date proxy: 2025-12-07. All displayed scores are medians of three runs; parentheses show revised minus original.\par}
\vfill
{\scriptsize C: clarity; O: organization/readability; G: graphical consistency; A: aesthetic coherence; I: visual inventiveness; S: design specificity. Mean is the arithmetic mean of six dimension medians, a descriptive display value, not a validated quality index.\par}

\clearpage
\phantomsection
\label{atlas:corpus112}
\pdfbookmark[1]{112. E-RayZer: Self-supervised 3D Reconstruction as Spatial Visual Pre-training}{corpus112}
{\Large\bfseries\color{academicblue} 112 \quad \href{https://openaccess.thecvf.com/content/CVPR2026/html/Zhao\_E-RayZer\_Self-supervised\_3D\_Reconstruction\_as\_Spatial\_Visual\_Pre-training\_CVPR\_2026\_paper.html}{E-RayZer: Self-supervised 3D Reconstruction as Spatial Visual Pre-training} \citep{corpus112}\par}
\vspace{3pt}
{\small S3 \enspace $\vert$ \enspace 3D vision \enspace $\vert$ \enspace CVPR 2026\par}
\vspace{4pt}
\vspace{5pt}
{\bfseries Motivation (source Fig.\ 3)} \hfill {\footnotesize \href{https://openaccess.thecvf.com/content/CVPR2026/papers/Zhao\_E-RayZer\_Self-supervised\_3D\_Reconstruction\_as\_Spatial\_Visual\_Pre-training\_CVPR\_2026\_paper.pdf}{Source PDF, p.\ 5}; ID: P187}\par
\vspace{2pt}
\noindent\begin{minipage}[t]{0.486\linewidth}
\centering\textbf{Original}\par\vspace{2pt}
\begin{minipage}[c][0.255\textheight][c]{\linewidth}\centering\includegraphics[width=\linewidth,height=0.255\textheight,keepaspectratio]{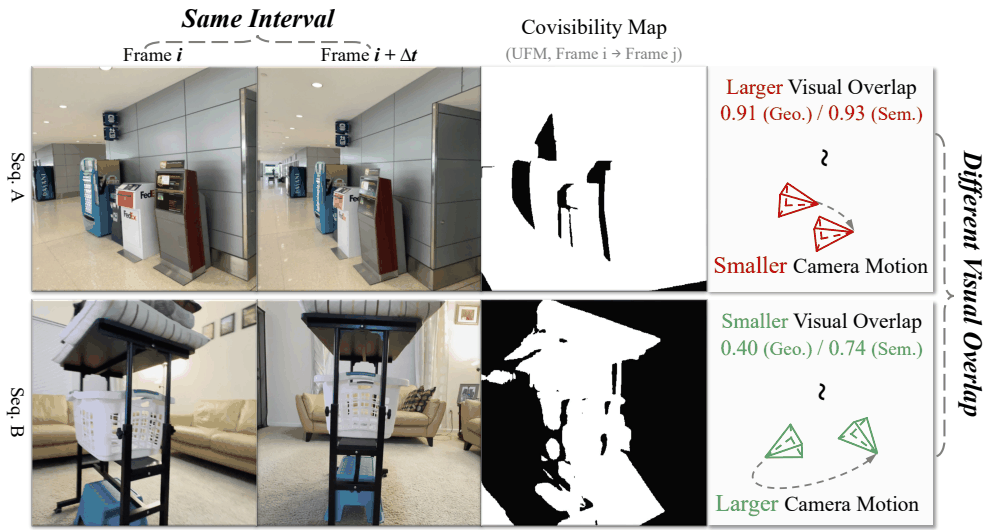}\end{minipage}\par
\vspace{3pt}{\scriptsize \setlength{\tabcolsep}{1.5pt}\begin{tabular*}{\linewidth}{@{\extracolsep{\fill}}rrrrrrr@{}}
\toprule
C & O & G & A & I & S & Mean \\
4 & 4 & 4 & 4 & 4 & 5 & 4.17\\
\bottomrule
\end{tabular*}
}
\end{minipage}\hfill
\begin{minipage}[t]{0.486\linewidth}
\centering\textbf{Study-generated derivative}\par\vspace{2pt}
\begin{minipage}[c][0.255\textheight][c]{\linewidth}\centering\includegraphics[width=\linewidth,height=0.255\textheight,keepaspectratio]{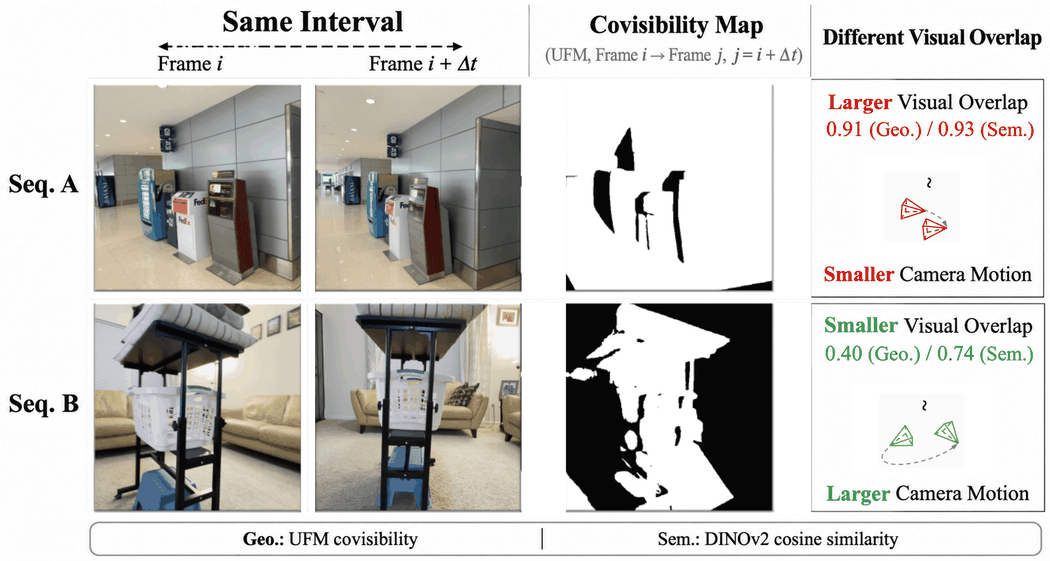}\end{minipage}\par
\vspace{3pt}
{\scriptsize \setlength{\tabcolsep}{1.5pt}\begin{tabular*}{\linewidth}{@{\extracolsep{\fill}}rrrrrrr@{}}
\toprule
C & O & G & A & I & S & Mean \\
5 (+1) & 5 (+1) & 4 (+0) & 4 (+0) & 4 (+0) & 5 (+0) & 4.50 (+0.33)\\
\bottomrule
\end{tabular*}
}
\end{minipage}\par
\vspace{3pt}
{\scriptsize Source public-date proxy: 2025-12-11. All displayed scores are medians of three runs; parentheses show revised minus original.\par}
\vspace{5pt}
{\bfseries Framework (source Fig.\ 2)} \hfill {\footnotesize \href{https://openaccess.thecvf.com/content/CVPR2026/papers/Zhao\_E-RayZer\_Self-supervised\_3D\_Reconstruction\_as\_Spatial\_Visual\_Pre-training\_CVPR\_2026\_paper.pdf}{Source PDF, p.\ 4}; ID: P188}\par
\vspace{2pt}
\noindent\begin{minipage}[t]{0.486\linewidth}
\centering\textbf{Original}\par\vspace{2pt}
\begin{minipage}[c][0.255\textheight][c]{\linewidth}\centering\includegraphics[width=\linewidth,height=0.255\textheight,keepaspectratio]{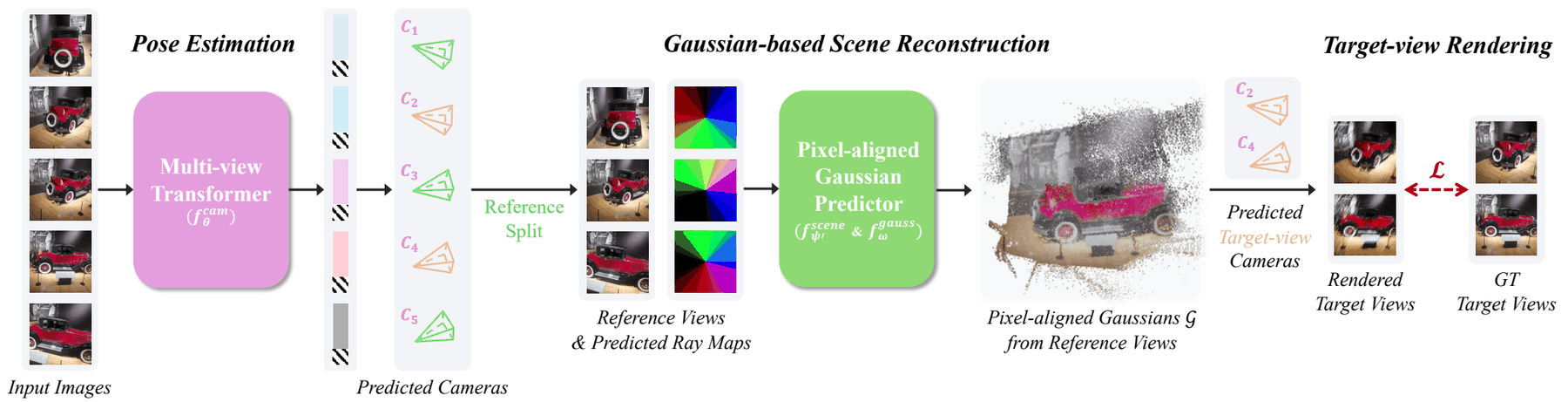}\end{minipage}\par
\vspace{3pt}{\scriptsize \setlength{\tabcolsep}{1.5pt}\begin{tabular*}{\linewidth}{@{\extracolsep{\fill}}rrrrrrr@{}}
\toprule
C & O & G & A & I & S & Mean \\
4 & 5 & 4 & 4 & 3 & 5 & 4.17\\
\bottomrule
\end{tabular*}
}
\end{minipage}\hfill
\begin{minipage}[t]{0.486\linewidth}
\centering\textbf{Study-generated derivative}\par\vspace{2pt}
\begin{minipage}[c][0.255\textheight][c]{\linewidth}\centering\includegraphics[width=\linewidth,height=0.255\textheight,keepaspectratio]{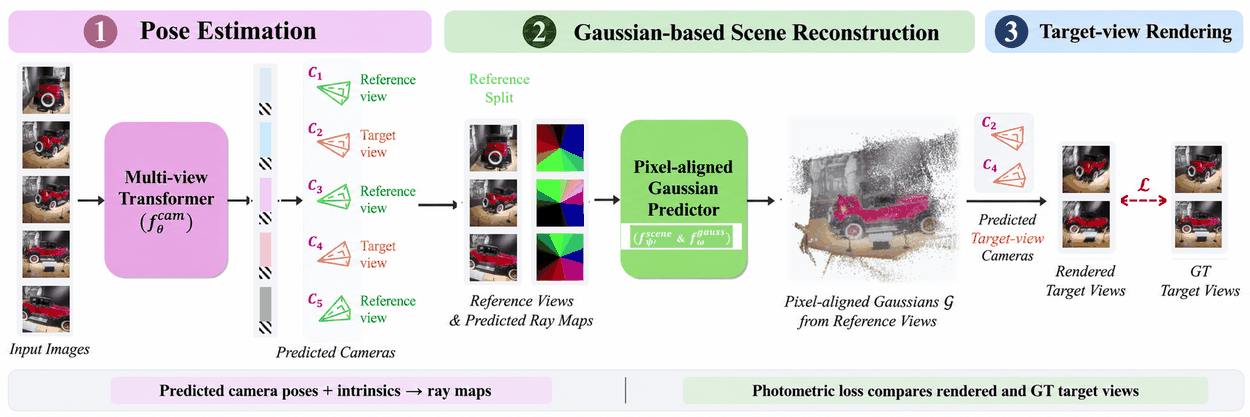}\end{minipage}\par
\vspace{3pt}
{\scriptsize \setlength{\tabcolsep}{1.5pt}\begin{tabular*}{\linewidth}{@{\extracolsep{\fill}}rrrrrrr@{}}
\toprule
C & O & G & A & I & S & Mean \\
4 (+0) & 5 (+0) & 4 (+0) & 4 (+0) & 3 (+0) & 5 (+0) & 4.17 (+0.00)\\
\bottomrule
\end{tabular*}
}
\end{minipage}\par
\vspace{3pt}
{\scriptsize Source public-date proxy: 2025-12-11. All displayed scores are medians of three runs; parentheses show revised minus original.\par}
\vfill
{\scriptsize C: clarity; O: organization/readability; G: graphical consistency; A: aesthetic coherence; I: visual inventiveness; S: design specificity. Mean is the arithmetic mean of six dimension medians, a descriptive display value, not a validated quality index.\par}

\clearpage
\phantomsection
\label{atlas:corpus117}
\pdfbookmark[1]{117. PoInit-of-View: Poisoning Initialization of Views Transfers Across Multiple 3D Reconstruction Systems}{corpus117}
{\Large\bfseries\color{academicblue} 117 \quad \href{https://openaccess.thecvf.com/content/CVPR2026/html/Wang\_PoInit-of-View\_Poisoning\_Initialization\_of\_Views\_Transfers\_Across\_Multiple\_3D\_Reconstruction\_CVPR\_2026\_paper.html}{PoInit-of-View: Poisoning Initialization of Views Transfers Across Multiple 3D Reconstruction Systems} \citep{corpus117}\par}
\vspace{3pt}
{\small S3 \enspace $\vert$ \enspace 3D vision \enspace $\vert$ \enspace CVPR 2026\par}
\vspace{4pt}
\vspace{5pt}
{\bfseries Motivation (source Fig.\ 1)} \hfill {\footnotesize \href{https://openaccess.thecvf.com/content/CVPR2026/papers/Wang\_PoInit-of-View\_Poisoning\_Initialization\_of\_Views\_Transfers\_Across\_Multiple\_3D\_Reconstruction\_CVPR\_2026\_paper.pdf}{Source PDF, p.\ 1}; ID: P77}\par
\vspace{2pt}
\noindent\begin{minipage}[t]{0.486\linewidth}
\centering\textbf{Original}\par\vspace{2pt}
\begin{minipage}[c][0.255\textheight][c]{\linewidth}\centering\includegraphics[width=\linewidth,height=0.255\textheight,keepaspectratio]{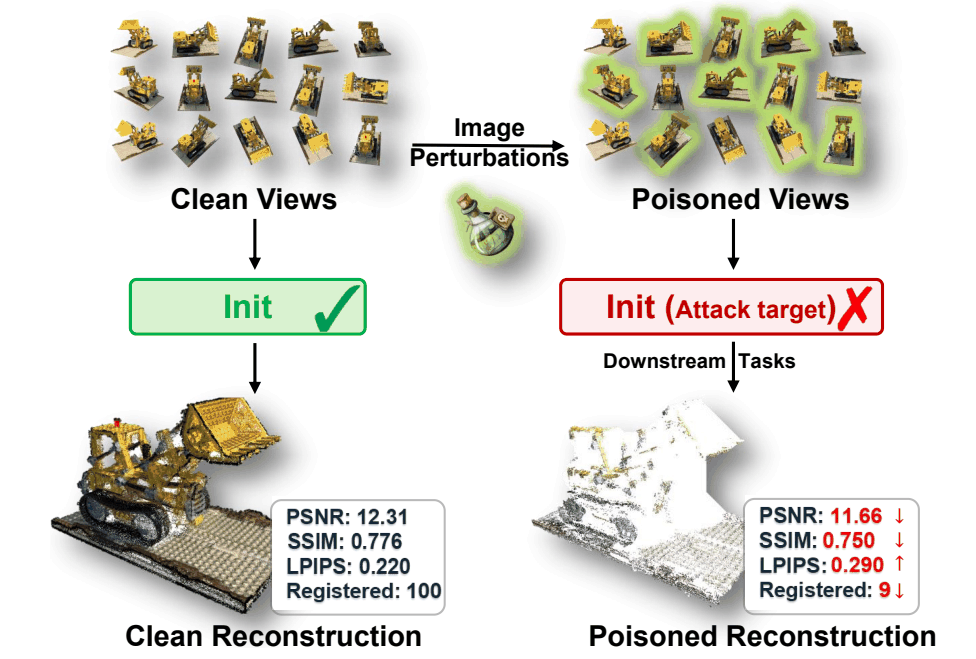}\end{minipage}\par
\vspace{3pt}{\scriptsize \setlength{\tabcolsep}{1.5pt}\begin{tabular*}{\linewidth}{@{\extracolsep{\fill}}rrrrrrr@{}}
\toprule
C & O & G & A & I & S & Mean \\
5 & 4 & 4 & 3 & 3 & 5 & 4.00\\
\bottomrule
\end{tabular*}
}
\end{minipage}\hfill
\begin{minipage}[t]{0.486\linewidth}
\centering\textbf{Study-generated derivative}\par\vspace{2pt}
\begin{minipage}[c][0.255\textheight][c]{\linewidth}\centering\includegraphics[width=\linewidth,height=0.255\textheight,keepaspectratio]{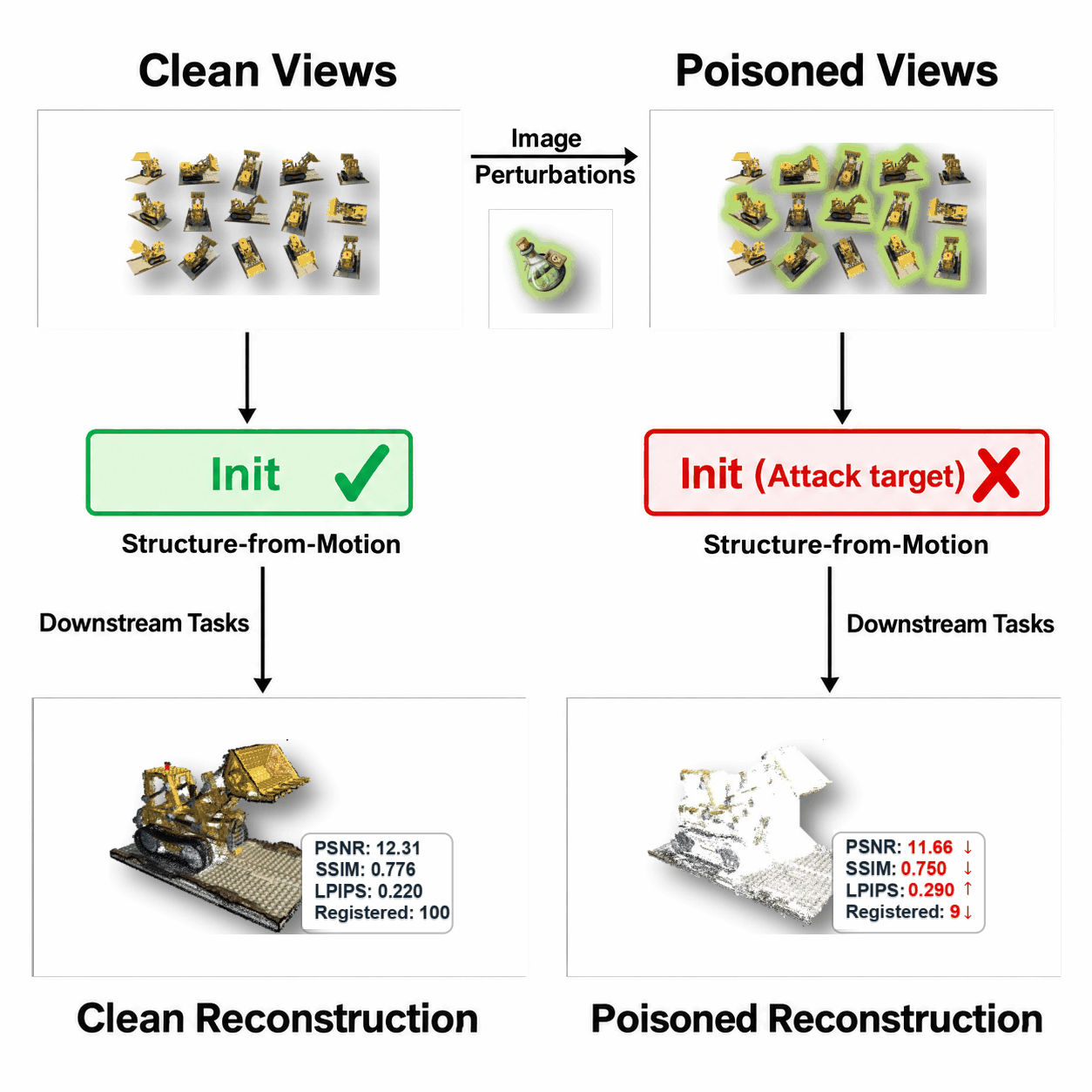}\end{minipage}\par
\vspace{3pt}
{\scriptsize \setlength{\tabcolsep}{1.5pt}\begin{tabular*}{\linewidth}{@{\extracolsep{\fill}}rrrrrrr@{}}
\toprule
C & O & G & A & I & S & Mean \\
5 (+0) & 5 (+1) & 4 (+0) & 4 (+1) & 3 (+0) & 5 (+0) & 4.33 (+0.33)\\
\bottomrule
\end{tabular*}
}
\end{minipage}\par
\vspace{3pt}
{\scriptsize Source public-date proxy: 2026-04-17. All displayed scores are medians of three runs; parentheses show revised minus original.\par}
\vspace{5pt}
{\bfseries Framework (source Fig.\ 2)} \hfill {\footnotesize \href{https://openaccess.thecvf.com/content/CVPR2026/papers/Wang\_PoInit-of-View\_Poisoning\_Initialization\_of\_Views\_Transfers\_Across\_Multiple\_3D\_Reconstruction\_CVPR\_2026\_paper.pdf}{Source PDF, p.\ 3}; ID: P78}\par
\vspace{2pt}
\noindent\begin{minipage}[t]{0.486\linewidth}
\centering\textbf{Original}\par\vspace{2pt}
\begin{minipage}[c][0.255\textheight][c]{\linewidth}\centering\includegraphics[width=\linewidth,height=0.255\textheight,keepaspectratio]{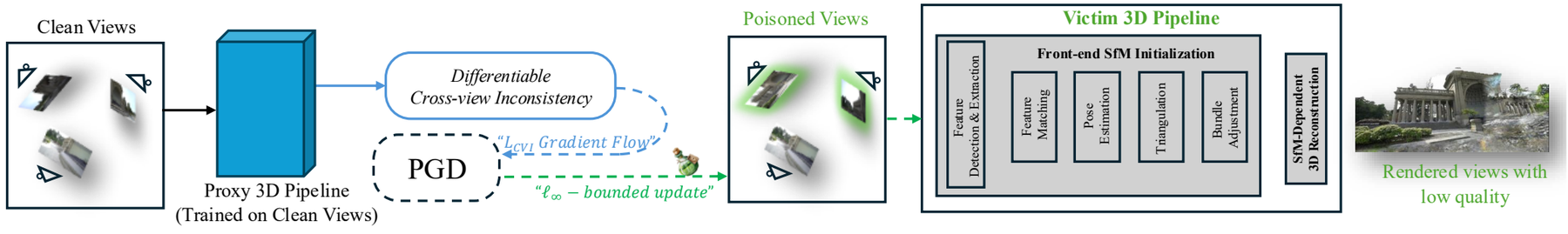}\end{minipage}\par
\vspace{3pt}{\scriptsize \setlength{\tabcolsep}{1.5pt}\begin{tabular*}{\linewidth}{@{\extracolsep{\fill}}rrrrrrr@{}}
\toprule
C & O & G & A & I & S & Mean \\
4 & 3 & 4 & 3 & 3 & 4 & 3.50\\
\bottomrule
\end{tabular*}
}
\end{minipage}\hfill
\begin{minipage}[t]{0.486\linewidth}
\centering\textbf{Study-generated derivative}\par\vspace{2pt}
\begin{minipage}[c][0.255\textheight][c]{\linewidth}\centering\includegraphics[width=\linewidth,height=0.255\textheight,keepaspectratio]{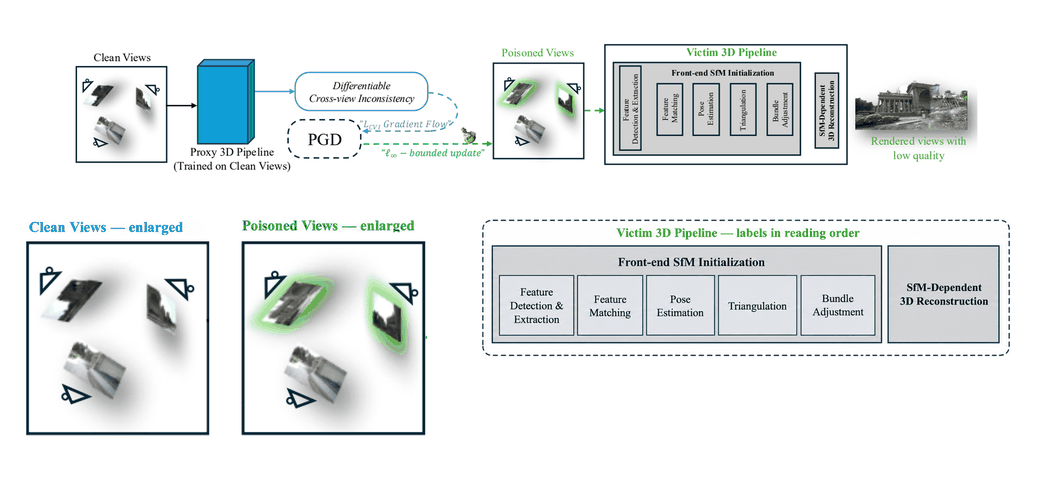}\end{minipage}\par
\vspace{3pt}
{\scriptsize \setlength{\tabcolsep}{1.5pt}\begin{tabular*}{\linewidth}{@{\extracolsep{\fill}}rrrrrrr@{}}
\toprule
C & O & G & A & I & S & Mean \\
4 (+0) & 3 (+0) & 4 (+0) & 3 (+0) & 3 (+0) & 4 (+0) & 3.50 (+0.00)\\
\bottomrule
\end{tabular*}
}
\end{minipage}\par
\vspace{3pt}
{\scriptsize Source public-date proxy: 2026-04-17. All displayed scores are medians of three runs; parentheses show revised minus original.\par}
\vfill
{\scriptsize C: clarity; O: organization/readability; G: graphical consistency; A: aesthetic coherence; I: visual inventiveness; S: design specificity. Mean is the arithmetic mean of six dimension medians, a descriptive display value, not a validated quality index.\par}

\clearpage
\phantomsection
\label{atlas:corpus121}
\pdfbookmark[1]{121. AnyDepth-DETR/-YOLO: Any-depth object detection with a single network}{corpus121}
{\Large\bfseries\color{academicblue} 121 \quad \href{https://openreview.net/forum?id=8qFCNFz8Tq}{AnyDepth-DETR/-YOLO: Any-depth object detection with a single network} \citep{corpus121}\par}
\vspace{3pt}
{\small S4 \enspace $\vert$ \enspace Object detection \enspace $\vert$ \enspace ICLR 2027 submission (under review)\par}
\vspace{4pt}
\vspace{5pt}
{\bfseries Motivation (source Fig.\ 2)} \hfill {\footnotesize \href{https://openreview.net/attachment?id=8qFCNFz8Tq\&name=pdf}{Source PDF, p.\ 6}; ID: ICLR27\_8qFCNFz8Tq\_M}\par
\vspace{2pt}
\noindent\begin{minipage}[t]{0.486\linewidth}
\centering\textbf{Original}\par\vspace{2pt}
\begin{minipage}[c][0.255\textheight][c]{\linewidth}\centering\includegraphics[width=\linewidth,height=0.255\textheight,keepaspectratio]{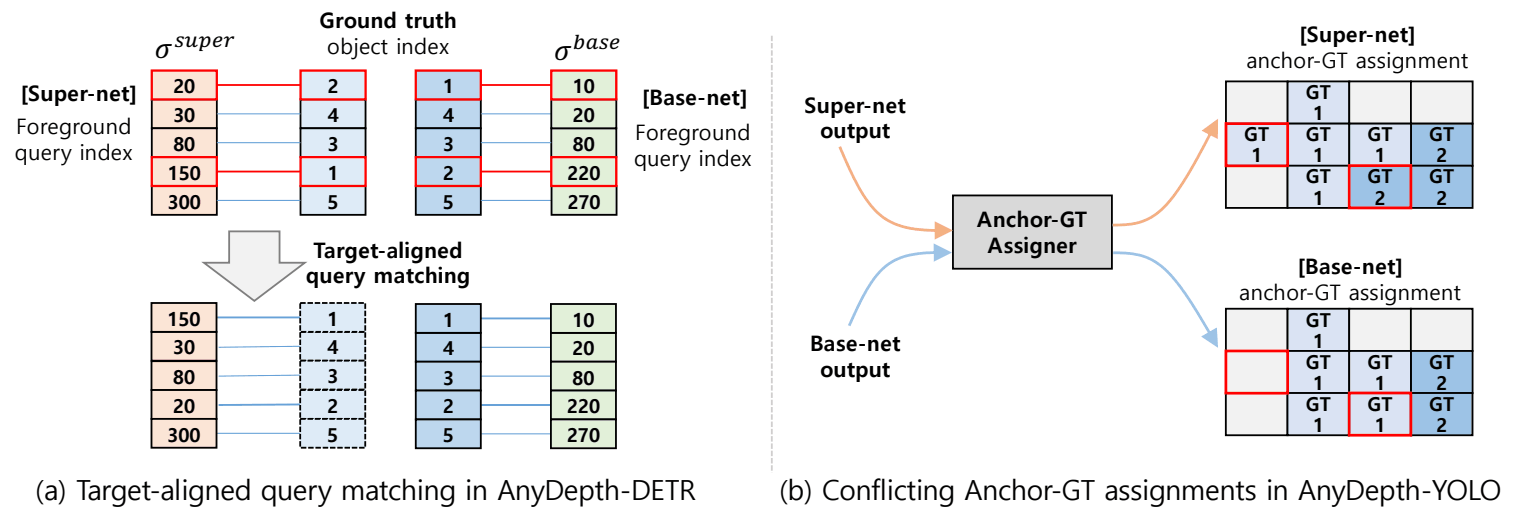}\end{minipage}\par
\vspace{3pt}{\scriptsize \setlength{\tabcolsep}{1.5pt}\begin{tabular*}{\linewidth}{@{\extracolsep{\fill}}rrrrrrr@{}}
\toprule
C & O & G & A & I & S & Mean \\
4 & 5 & 4 & 4 & 4 & 5 & 4.33\\
\bottomrule
\end{tabular*}
}
\end{minipage}\hfill
\begin{minipage}[t]{0.486\linewidth}
\centering\textbf{Study-generated derivative}\par\vspace{2pt}
\begin{minipage}[c][0.255\textheight][c]{\linewidth}\centering\includegraphics[width=\linewidth,height=0.255\textheight,keepaspectratio]{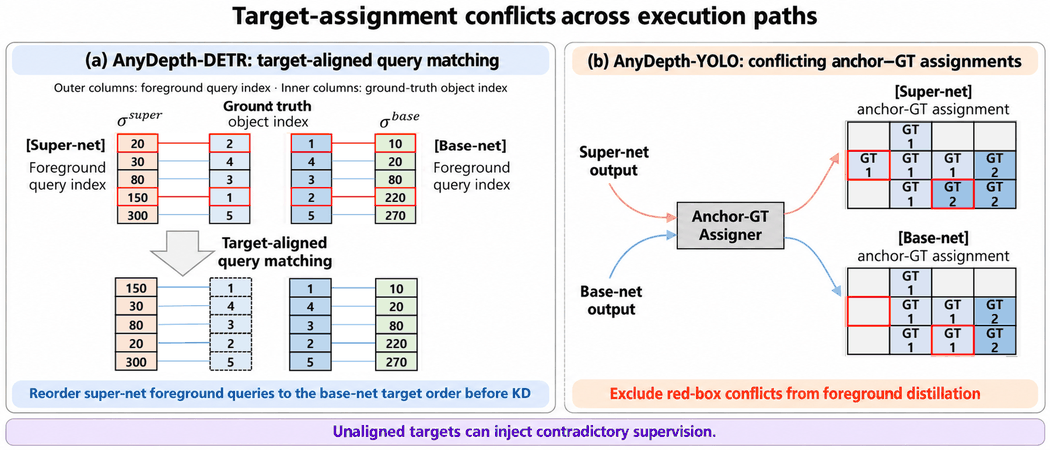}\end{minipage}\par
\vspace{3pt}
{\scriptsize \setlength{\tabcolsep}{1.5pt}\begin{tabular*}{\linewidth}{@{\extracolsep{\fill}}rrrrrrr@{}}
\toprule
C & O & G & A & I & S & Mean \\
5 (+1) & 4 (-1) & 4 (+0) & 4 (+0) & 4 (+0) & 5 (+0) & 4.33 (+0.00)\\
\bottomrule
\end{tabular*}
}
\end{minipage}\par
\vspace{3pt}
{\scriptsize Source public-date proxy: 2026-05-10. All displayed scores are medians of three runs; parentheses show revised minus original.\par}
\vspace{5pt}
{\bfseries Framework (source Fig.\ 1)} \hfill {\footnotesize \href{https://openreview.net/attachment?id=8qFCNFz8Tq\&name=pdf}{Source PDF, p.\ 2}; ID: ICLR27\_8qFCNFz8Tq\_F}\par
\vspace{2pt}
\noindent\begin{minipage}[t]{0.486\linewidth}
\centering\textbf{Original}\par\vspace{2pt}
\begin{minipage}[c][0.255\textheight][c]{\linewidth}\centering\includegraphics[width=\linewidth,height=0.255\textheight,keepaspectratio]{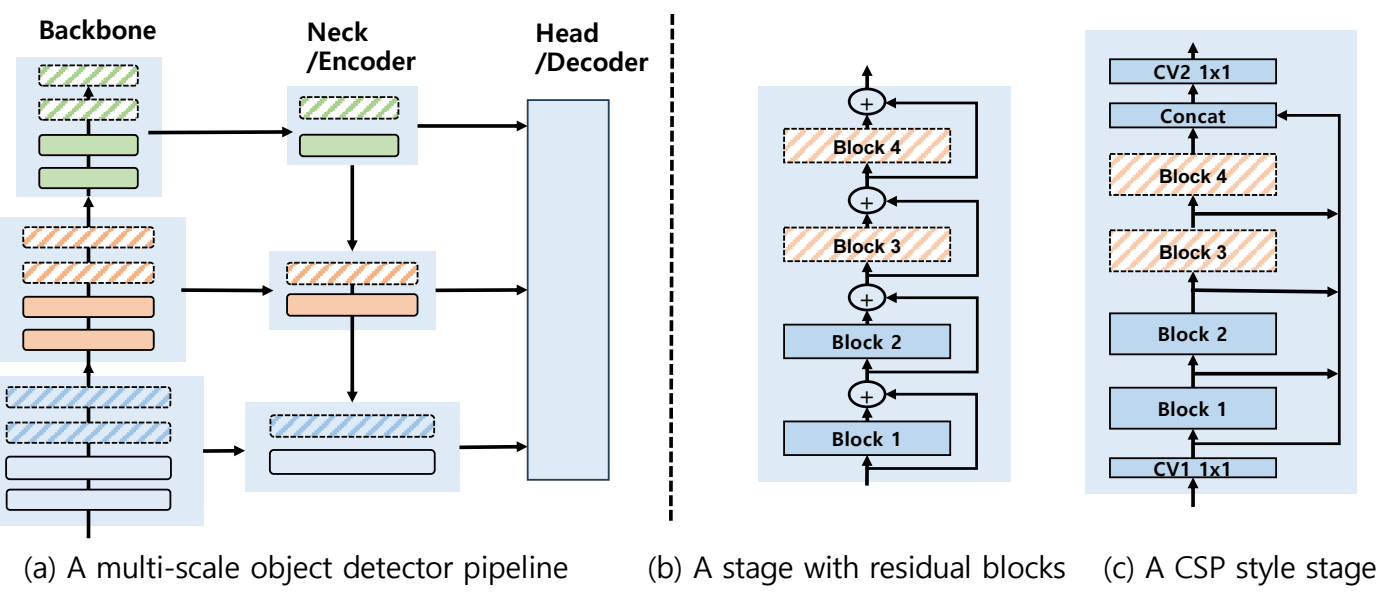}\end{minipage}\par
\vspace{3pt}{\scriptsize \setlength{\tabcolsep}{1.5pt}\begin{tabular*}{\linewidth}{@{\extracolsep{\fill}}rrrrrrr@{}}
\toprule
C & O & G & A & I & S & Mean \\
3 & 4 & 3 & 4 & 3 & 4 & 3.50\\
\bottomrule
\end{tabular*}
}
\end{minipage}\hfill
\begin{minipage}[t]{0.486\linewidth}
\centering\textbf{Study-generated derivative}\par\vspace{2pt}
\begin{minipage}[c][0.255\textheight][c]{\linewidth}\centering\includegraphics[width=\linewidth,height=0.255\textheight,keepaspectratio]{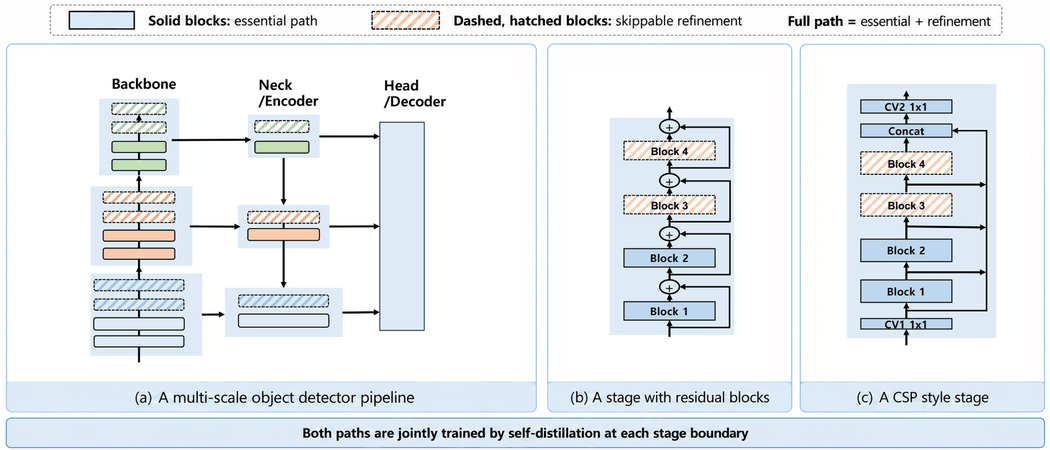}\end{minipage}\par
\vspace{3pt}
{\scriptsize \setlength{\tabcolsep}{1.5pt}\begin{tabular*}{\linewidth}{@{\extracolsep{\fill}}rrrrrrr@{}}
\toprule
C & O & G & A & I & S & Mean \\
4 (+1) & 5 (+1) & 4 (+1) & 4 (+0) & 3 (+0) & 4 (+0) & 4.00 (+0.50)\\
\bottomrule
\end{tabular*}
}
\end{minipage}\par
\vspace{3pt}
{\scriptsize Source public-date proxy: 2026-05-10. All displayed scores are medians of three runs; parentheses show revised minus original.\par}
\vfill
{\scriptsize C: clarity; O: organization/readability; G: graphical consistency; A: aesthetic coherence; I: visual inventiveness; S: design specificity. Mean is the arithmetic mean of six dimension medians, a descriptive display value, not a validated quality index.\par}

\clearpage
\phantomsection
\label{atlas:corpus127}
\pdfbookmark[1]{127. PULSE: PERSISTENT UTILIZATION OF LIMITED SUPPORT EVIDENCE FOR CROSS-DOMAIN FEW-SHOT OBJECT DETECTION}{corpus127}
{\Large\bfseries\color{academicblue} 127 \quad \href{https://openreview.net/forum?id=ad0Os8QbxN}{PULSE: PERSISTENT UTILIZATION OF LIMITED SUPPORT EVIDENCE FOR CROSS-DOMAIN FEW-SHOT OBJECT DETECTION} \citep{corpus127}\par}
\vspace{3pt}
{\small S4 \enspace $\vert$ \enspace Object detection \enspace $\vert$ \enspace ICLR 2027 submission (under review)\par}
\vspace{4pt}
\vspace{5pt}
{\bfseries Motivation (source Fig.\ 1)} \hfill {\footnotesize \href{https://openreview.net/attachment?id=ad0Os8QbxN\&name=pdf}{Source PDF, p.\ 2}; ID: ICLR27\_ad0Os8QbxN\_M}\par
\vspace{2pt}
\noindent\begin{minipage}[t]{0.486\linewidth}
\centering\textbf{Original}\par\vspace{2pt}
\begin{minipage}[c][0.255\textheight][c]{\linewidth}\centering\includegraphics[width=\linewidth,height=0.255\textheight,keepaspectratio]{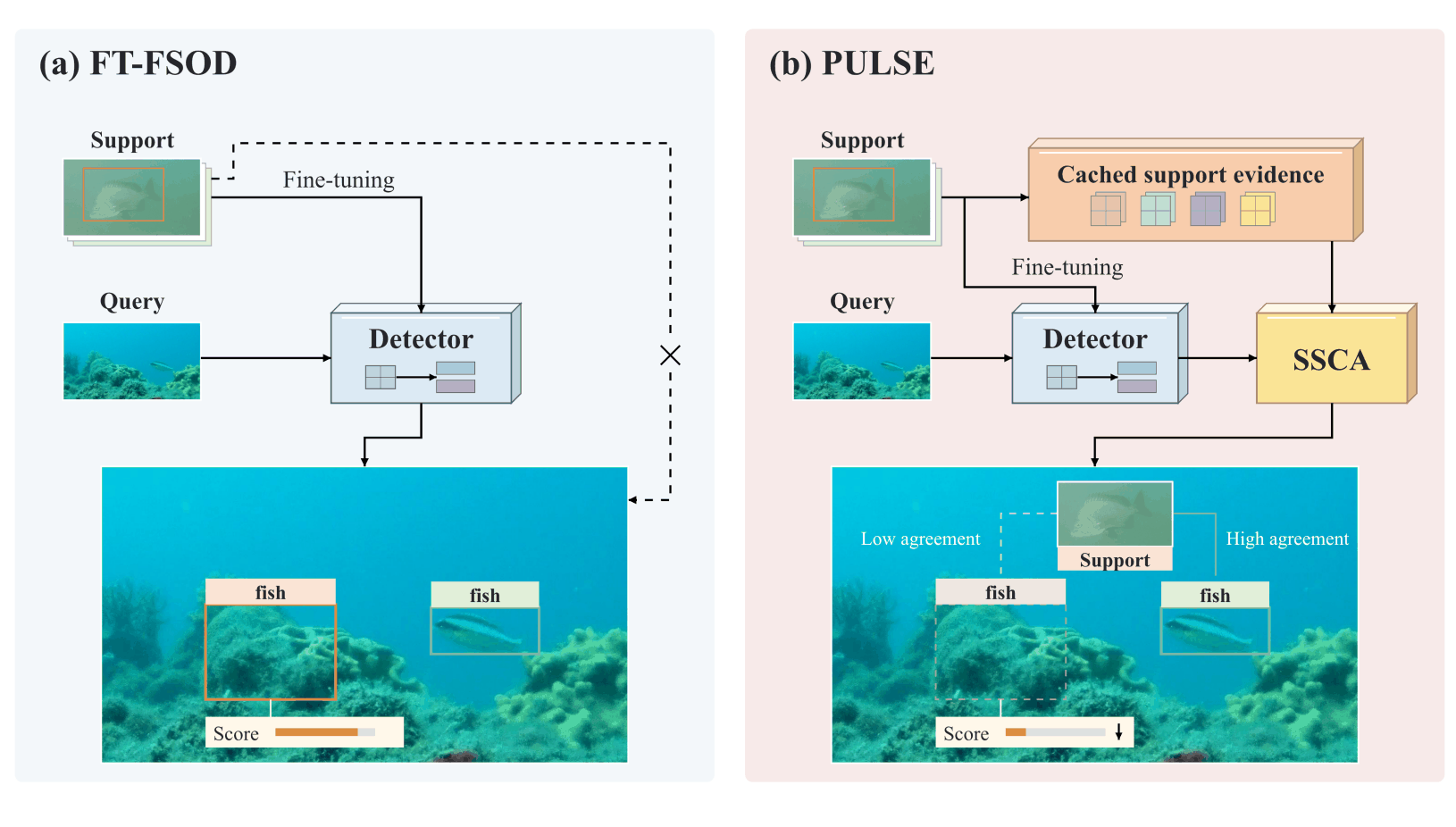}\end{minipage}\par
\vspace{3pt}{\scriptsize \setlength{\tabcolsep}{1.5pt}\begin{tabular*}{\linewidth}{@{\extracolsep{\fill}}rrrrrrr@{}}
\toprule
C & O & G & A & I & S & Mean \\
4 & 5 & 4 & 5 & 4 & 5 & 4.50\\
\bottomrule
\end{tabular*}
}
\end{minipage}\hfill
\begin{minipage}[t]{0.486\linewidth}
\centering\textbf{Study-generated derivative}\par\vspace{2pt}
\begin{minipage}[c][0.255\textheight][c]{\linewidth}\centering\includegraphics[width=\linewidth,height=0.255\textheight,keepaspectratio]{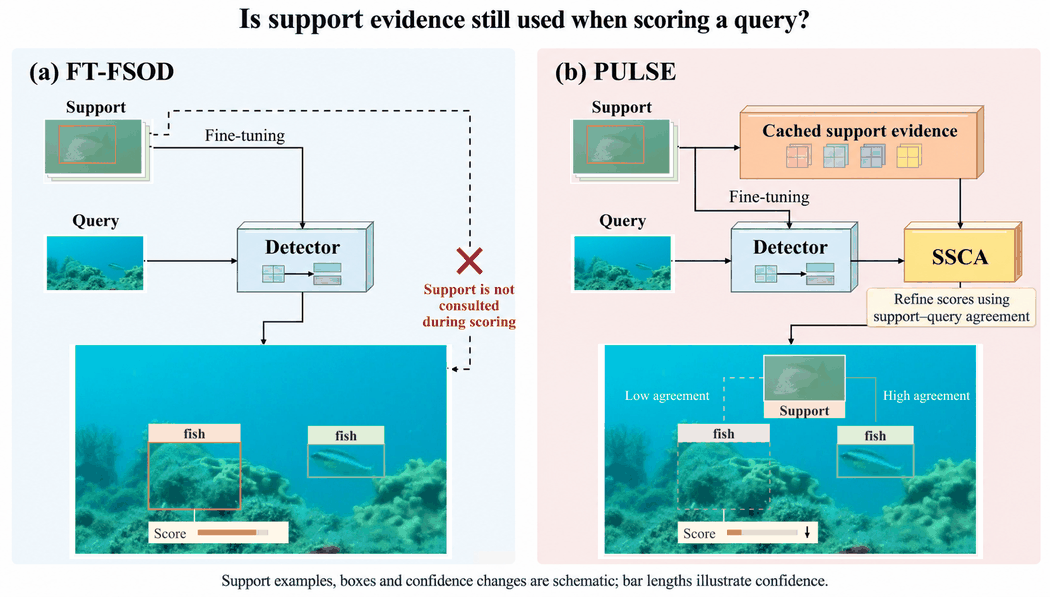}\end{minipage}\par
\vspace{3pt}
{\scriptsize \setlength{\tabcolsep}{1.5pt}\begin{tabular*}{\linewidth}{@{\extracolsep{\fill}}rrrrrrr@{}}
\toprule
C & O & G & A & I & S & Mean \\
5 (+1) & 5 (+0) & 4 (+0) & 4 (-1) & 4 (+0) & 5 (+0) & 4.50 (+0.00)\\
\bottomrule
\end{tabular*}
}
\end{minipage}\par
\vspace{3pt}
{\scriptsize Source public-date proxy: 2026-10-03. All displayed scores are medians of three runs; parentheses show revised minus original.\par}
\vspace{5pt}
{\bfseries Framework (source Fig.\ 2)} \hfill {\footnotesize \href{https://openreview.net/attachment?id=ad0Os8QbxN\&name=pdf}{Source PDF, p.\ 4}; ID: ICLR27\_ad0Os8QbxN\_F}\par
\vspace{2pt}
\noindent\begin{minipage}[t]{0.486\linewidth}
\centering\textbf{Original}\par\vspace{2pt}
\begin{minipage}[c][0.255\textheight][c]{\linewidth}\centering\includegraphics[width=\linewidth,height=0.255\textheight,keepaspectratio]{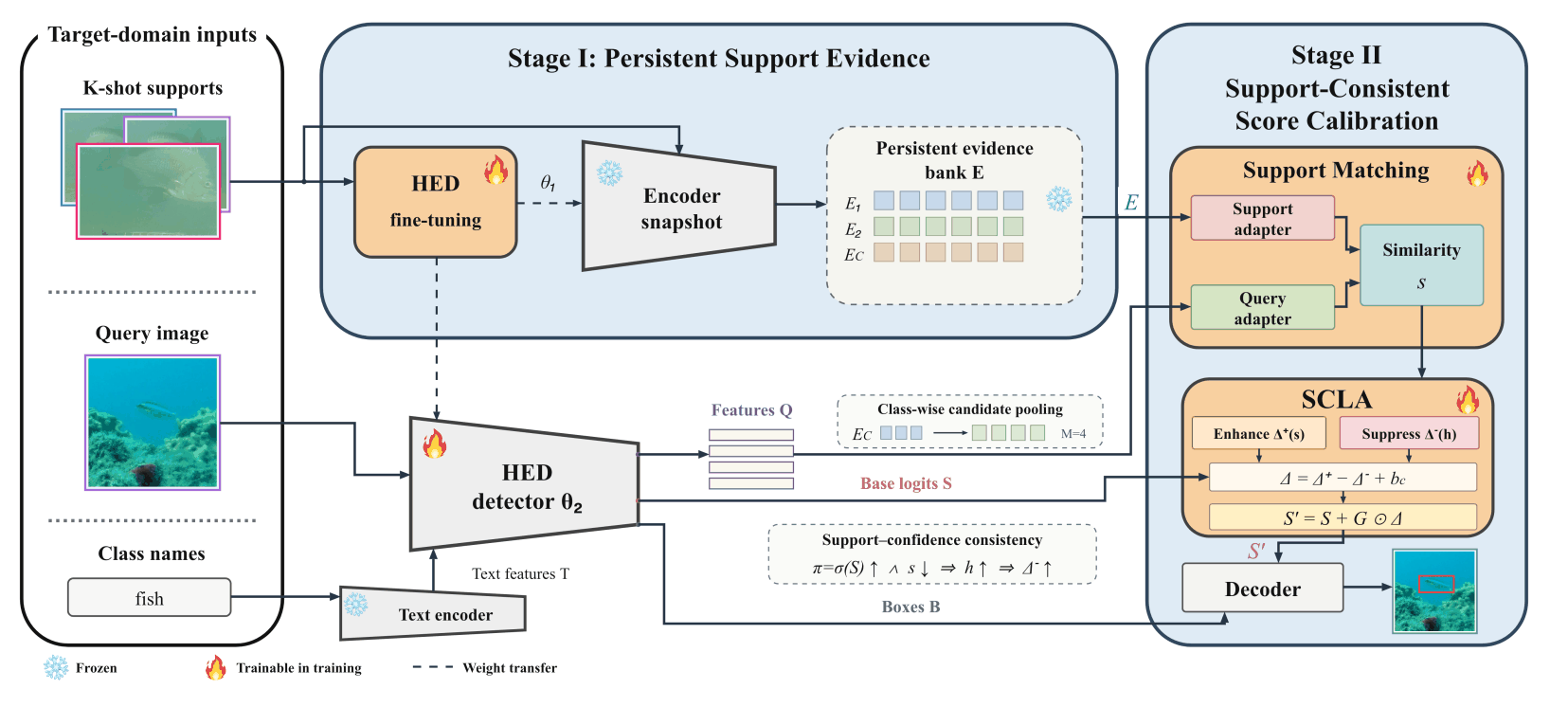}\end{minipage}\par
\vspace{3pt}{\scriptsize \setlength{\tabcolsep}{1.5pt}\begin{tabular*}{\linewidth}{@{\extracolsep{\fill}}rrrrrrr@{}}
\toprule
C & O & G & A & I & S & Mean \\
4 & 4 & 4 & 4 & 3 & 5 & 4.00\\
\bottomrule
\end{tabular*}
}
\end{minipage}\hfill
\begin{minipage}[t]{0.486\linewidth}
\centering\textbf{Study-generated derivative}\par\vspace{2pt}
\begin{minipage}[c][0.255\textheight][c]{\linewidth}\centering\includegraphics[width=\linewidth,height=0.255\textheight,keepaspectratio]{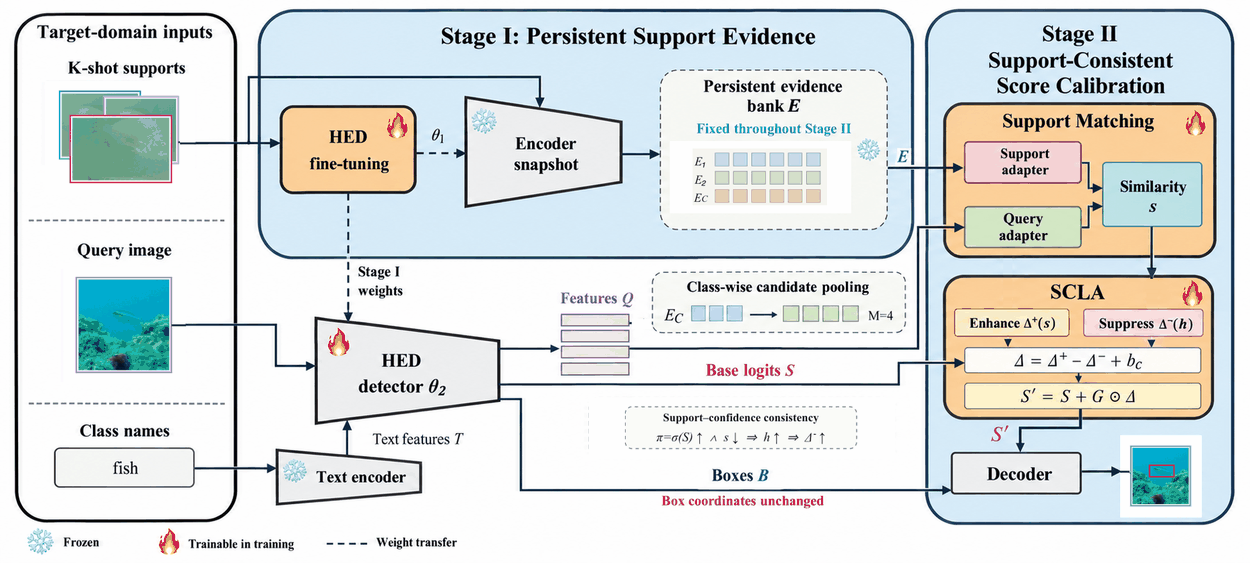}\end{minipage}\par
\vspace{3pt}
{\scriptsize \setlength{\tabcolsep}{1.5pt}\begin{tabular*}{\linewidth}{@{\extracolsep{\fill}}rrrrrrr@{}}
\toprule
C & O & G & A & I & S & Mean \\
4 (+0) & 4 (+0) & 4 (+0) & 3 (-1) & 3 (+0) & 5 (+0) & 3.83 (-0.17)\\
\bottomrule
\end{tabular*}
}
\end{minipage}\par
\vspace{3pt}
{\scriptsize Source public-date proxy: 2026-10-03. All displayed scores are medians of three runs; parentheses show revised minus original.\par}
\vfill
{\scriptsize C: clarity; O: organization/readability; G: graphical consistency; A: aesthetic coherence; I: visual inventiveness; S: design specificity. Mean is the arithmetic mean of six dimension medians, a descriptive display value, not a validated quality index.\par}

\clearpage
\phantomsection
\label{atlas:corpus132}
\pdfbookmark[1]{132. Calibration-Guided State Selection and Class-Conditioned Fusion for Frozen Diffusion Segmentation}{corpus132}
{\Large\bfseries\color{academicblue} 132 \quad \href{https://openreview.net/forum?id=NYrrKvi2Dy}{Calibration-Guided State Selection and Class-Conditioned Fusion for Frozen Diffusion Segmentation} \citep{corpus132}\par}
\vspace{3pt}
{\small S4 \enspace $\vert$ \enspace Image segmentation \enspace $\vert$ \enspace ICLR 2027 submission (under review)\par}
\vspace{4pt}
\vspace{5pt}
{\bfseries Motivation (source Fig.\ 2)} \hfill {\footnotesize \href{https://openreview.net/attachment?id=NYrrKvi2Dy\&name=pdf}{Source PDF, p.\ 3}; ID: ICLR27\_NYrrKvi2Dy\_M}\par
\vspace{2pt}
\noindent\begin{minipage}[t]{0.486\linewidth}
\centering\textbf{Original}\par\vspace{2pt}
\begin{minipage}[c][0.255\textheight][c]{\linewidth}\centering\includegraphics[width=\linewidth,height=0.255\textheight,keepaspectratio]{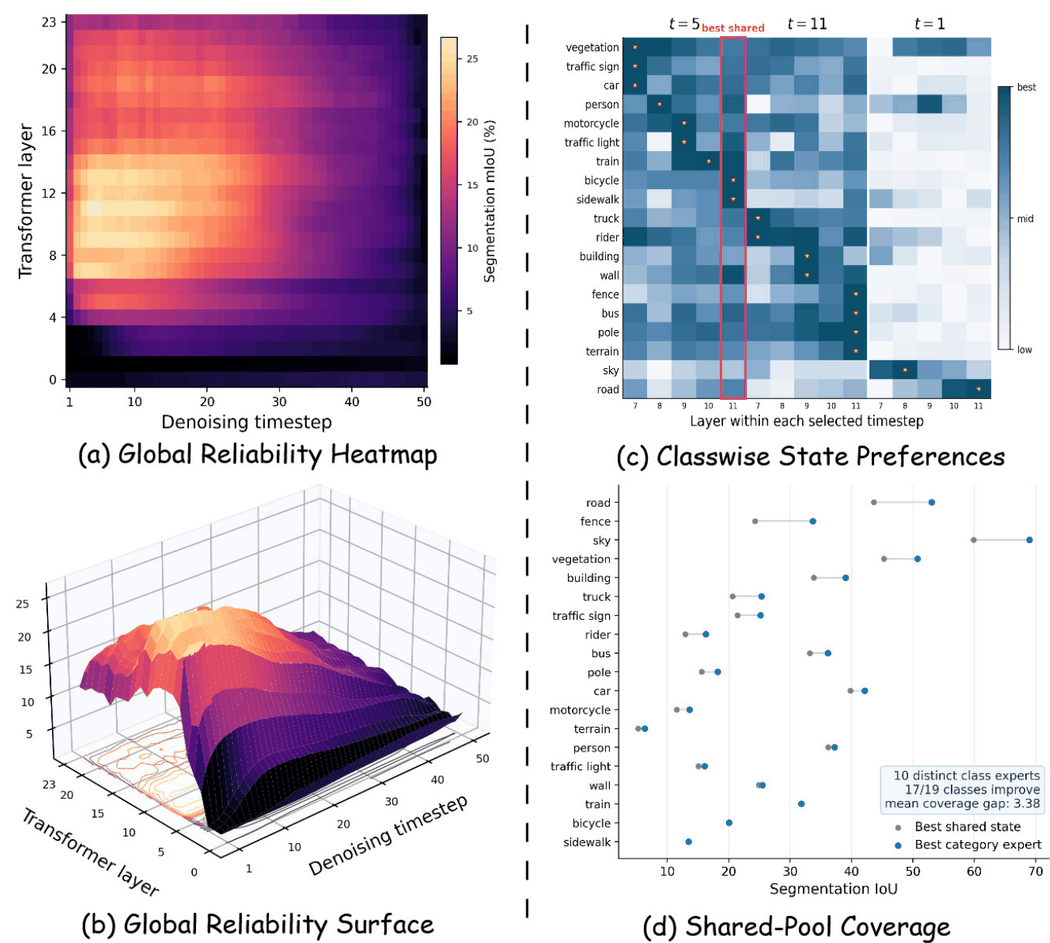}\end{minipage}\par
\vspace{3pt}{\scriptsize \setlength{\tabcolsep}{1.5pt}\begin{tabular*}{\linewidth}{@{\extracolsep{\fill}}rrrrrrr@{}}
\toprule
C & O & G & A & I & S & Mean \\
4 & 4 & 4 & 3 & 4 & 4 & 3.83\\
\bottomrule
\end{tabular*}
}
\end{minipage}\hfill
\begin{minipage}[t]{0.486\linewidth}
\centering\textbf{Study-generated derivative}\par\vspace{2pt}
\begin{minipage}[c][0.255\textheight][c]{\linewidth}\centering\includegraphics[width=\linewidth,height=0.255\textheight,keepaspectratio]{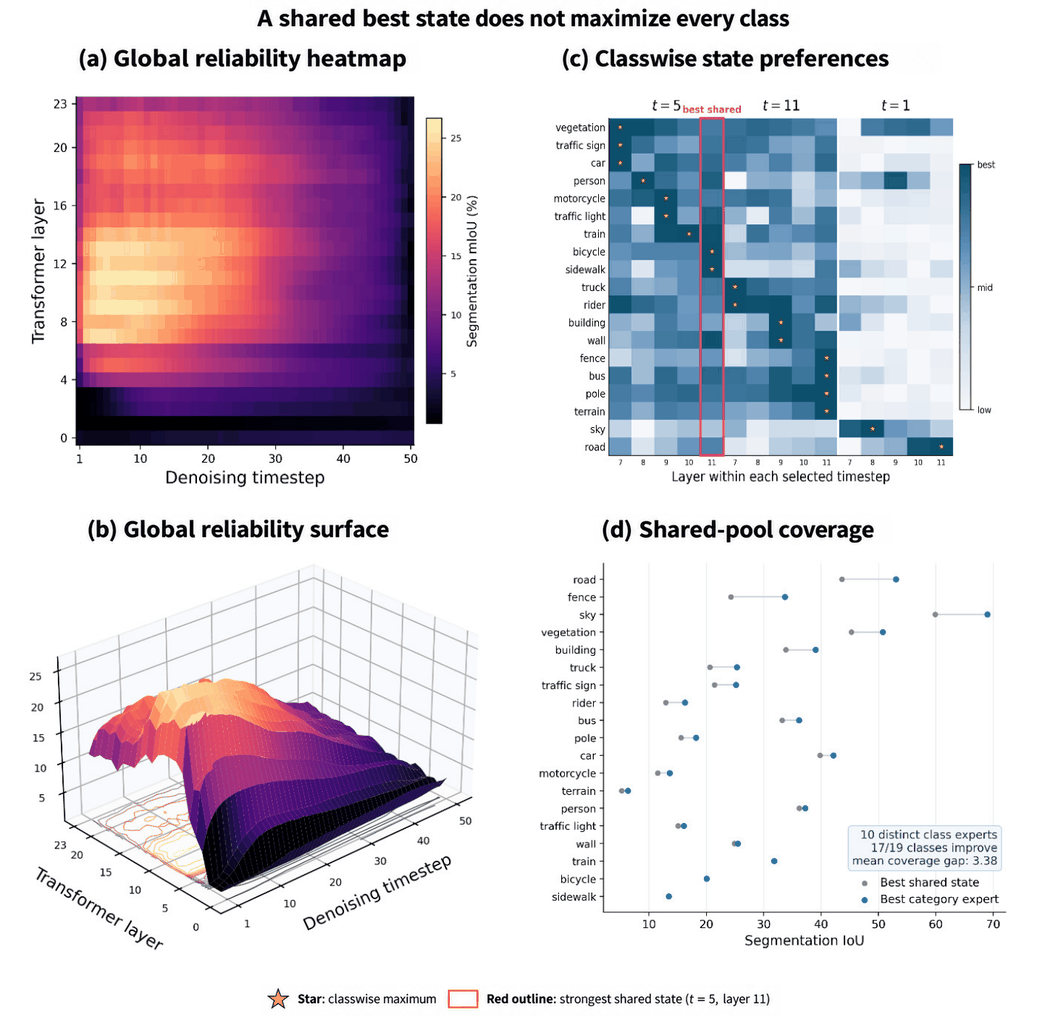}\end{minipage}\par
\vspace{3pt}
{\scriptsize \setlength{\tabcolsep}{1.5pt}\begin{tabular*}{\linewidth}{@{\extracolsep{\fill}}rrrrrrr@{}}
\toprule
C & O & G & A & I & S & Mean \\
5 (+1) & 4 (+0) & 4 (+0) & 4 (+1) & 4 (+0) & 5 (+1) & 4.33 (+0.50)\\
\bottomrule
\end{tabular*}
}
\end{minipage}\par
\vspace{3pt}
{\scriptsize Source public-date proxy: 2026-10-03. All displayed scores are medians of three runs; parentheses show revised minus original.\par}
\vspace{5pt}
{\bfseries Framework (source Fig.\ 3)} \hfill {\footnotesize \href{https://openreview.net/attachment?id=NYrrKvi2Dy\&name=pdf}{Source PDF, p.\ 5}; ID: ICLR27\_NYrrKvi2Dy\_F}\par
\vspace{2pt}
\noindent\begin{minipage}[t]{0.486\linewidth}
\centering\textbf{Original}\par\vspace{2pt}
\begin{minipage}[c][0.255\textheight][c]{\linewidth}\centering\includegraphics[width=\linewidth,height=0.255\textheight,keepaspectratio]{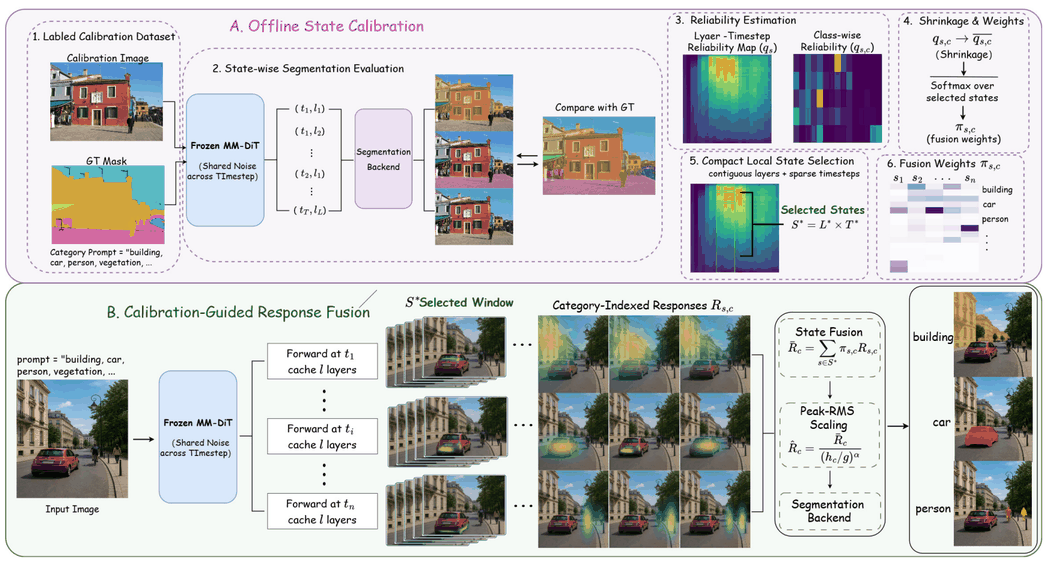}\end{minipage}\par
\vspace{3pt}{\scriptsize \setlength{\tabcolsep}{1.5pt}\begin{tabular*}{\linewidth}{@{\extracolsep{\fill}}rrrrrrr@{}}
\toprule
C & O & G & A & I & S & Mean \\
3 & 4 & 3 & 4 & 4 & 5 & 3.83\\
\bottomrule
\end{tabular*}
}
\end{minipage}\hfill
\begin{minipage}[t]{0.486\linewidth}
\centering\textbf{Study-generated derivative}\par\vspace{2pt}
\begin{minipage}[c][0.255\textheight][c]{\linewidth}\centering\includegraphics[width=\linewidth,height=0.255\textheight,keepaspectratio]{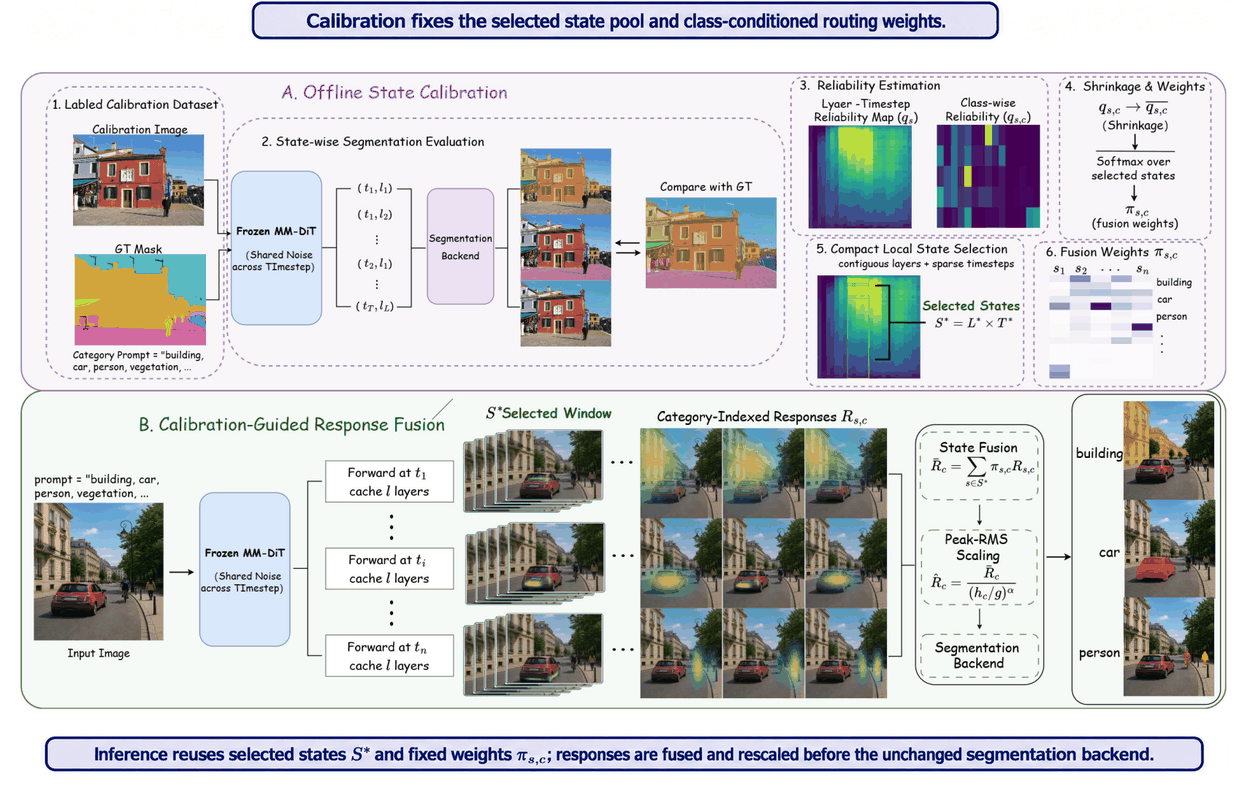}\end{minipage}\par
\vspace{3pt}
{\scriptsize \setlength{\tabcolsep}{1.5pt}\begin{tabular*}{\linewidth}{@{\extracolsep{\fill}}rrrrrrr@{}}
\toprule
C & O & G & A & I & S & Mean \\
4 (+1) & 4 (+0) & 3 (+0) & 3 (-1) & 4 (+0) & 5 (+0) & 3.83 (+0.00)\\
\bottomrule
\end{tabular*}
}
\end{minipage}\par
\vspace{3pt}
{\scriptsize Source public-date proxy: 2026-10-03. All displayed scores are medians of three runs; parentheses show revised minus original.\par}
\vfill
{\scriptsize C: clarity; O: organization/readability; G: graphical consistency; A: aesthetic coherence; I: visual inventiveness; S: design specificity. Mean is the arithmetic mean of six dimension medians, a descriptive display value, not a validated quality index.\par}

\clearpage
\phantomsection
\label{atlas:corpus134}
\pdfbookmark[1]{134. HIVE: Holistic Image-Adaptive Vocabulary Evolution for Training-Free Dense Vision-Language Inference}{corpus134}
{\Large\bfseries\color{academicblue} 134 \quad \href{https://openreview.net/forum?id=kv0BeIZGMk}{HIVE: Holistic Image-Adaptive Vocabulary Evolution for Training-Free Dense Vision-Language Inference} \citep{corpus134}\par}
\vspace{3pt}
{\small S4 \enspace $\vert$ \enspace Image segmentation \enspace $\vert$ \enspace ICLR 2027 submission (under review)\par}
\vspace{4pt}
\vspace{5pt}
{\bfseries Motivation (source Fig.\ 1)} \hfill {\footnotesize \href{https://openreview.net/attachment?id=kv0BeIZGMk\&name=pdf}{Source PDF, p.\ 1}; ID: ICLR27\_kv0BeIZGMk\_M}\par
\vspace{2pt}
\noindent\begin{minipage}[t]{0.486\linewidth}
\centering\textbf{Original}\par\vspace{2pt}
\begin{minipage}[c][0.255\textheight][c]{\linewidth}\centering\includegraphics[width=\linewidth,height=0.255\textheight,keepaspectratio]{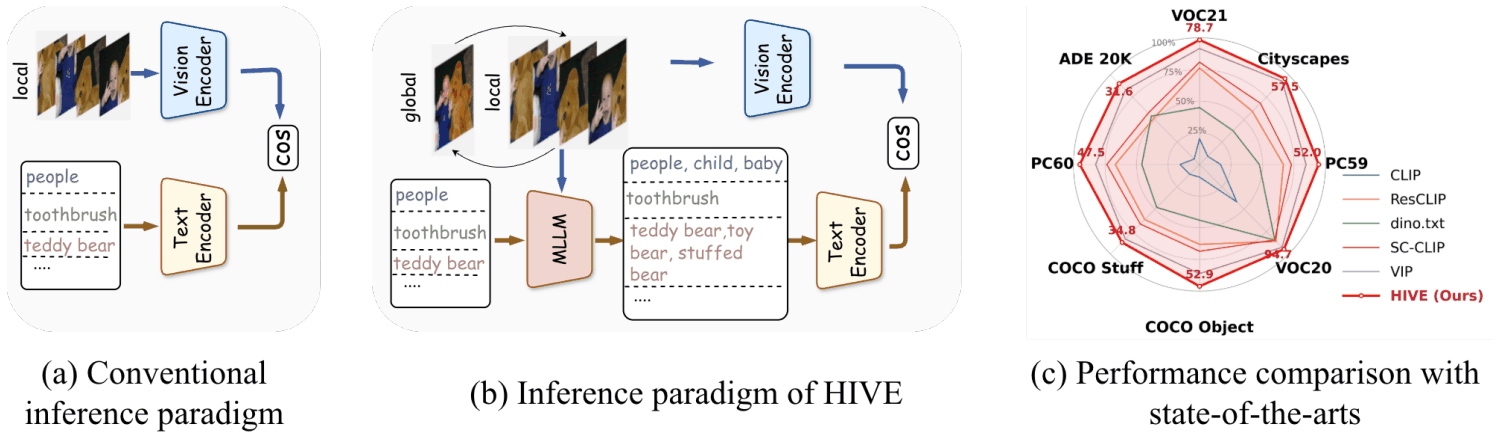}\end{minipage}\par
\vspace{3pt}{\scriptsize \setlength{\tabcolsep}{1.5pt}\begin{tabular*}{\linewidth}{@{\extracolsep{\fill}}rrrrrrr@{}}
\toprule
C & O & G & A & I & S & Mean \\
4 & 4 & 4 & 4 & 3 & 4 & 3.83\\
\bottomrule
\end{tabular*}
}
\end{minipage}\hfill
\begin{minipage}[t]{0.486\linewidth}
\centering\textbf{Study-generated derivative}\par\vspace{2pt}
\begin{minipage}[c][0.255\textheight][c]{\linewidth}\centering\includegraphics[width=\linewidth,height=0.255\textheight,keepaspectratio]{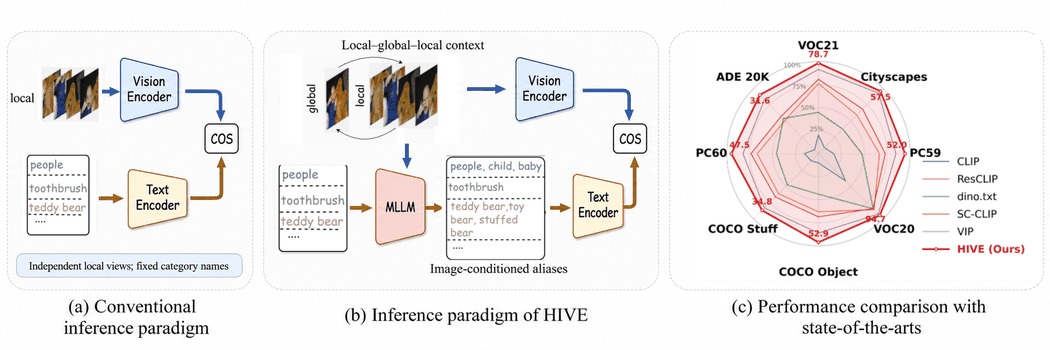}\end{minipage}\par
\vspace{3pt}
{\scriptsize \setlength{\tabcolsep}{1.5pt}\begin{tabular*}{\linewidth}{@{\extracolsep{\fill}}rrrrrrr@{}}
\toprule
C & O & G & A & I & S & Mean \\
4 (+0) & 4 (+0) & 4 (+0) & 3 (-1) & 3 (+0) & 4 (+0) & 3.67 (-0.17)\\
\bottomrule
\end{tabular*}
}
\end{minipage}\par
\vspace{3pt}
{\scriptsize Source public-date proxy: 2026-10-03. All displayed scores are medians of three runs; parentheses show revised minus original.\par}
\vspace{5pt}
{\bfseries Framework (source Fig.\ 2)} \hfill {\footnotesize \href{https://openreview.net/attachment?id=kv0BeIZGMk\&name=pdf}{Source PDF, p.\ 4}; ID: ICLR27\_kv0BeIZGMk\_F}\par
\vspace{2pt}
\noindent\begin{minipage}[t]{0.486\linewidth}
\centering\textbf{Original}\par\vspace{2pt}
\begin{minipage}[c][0.255\textheight][c]{\linewidth}\centering\includegraphics[width=\linewidth,height=0.255\textheight,keepaspectratio]{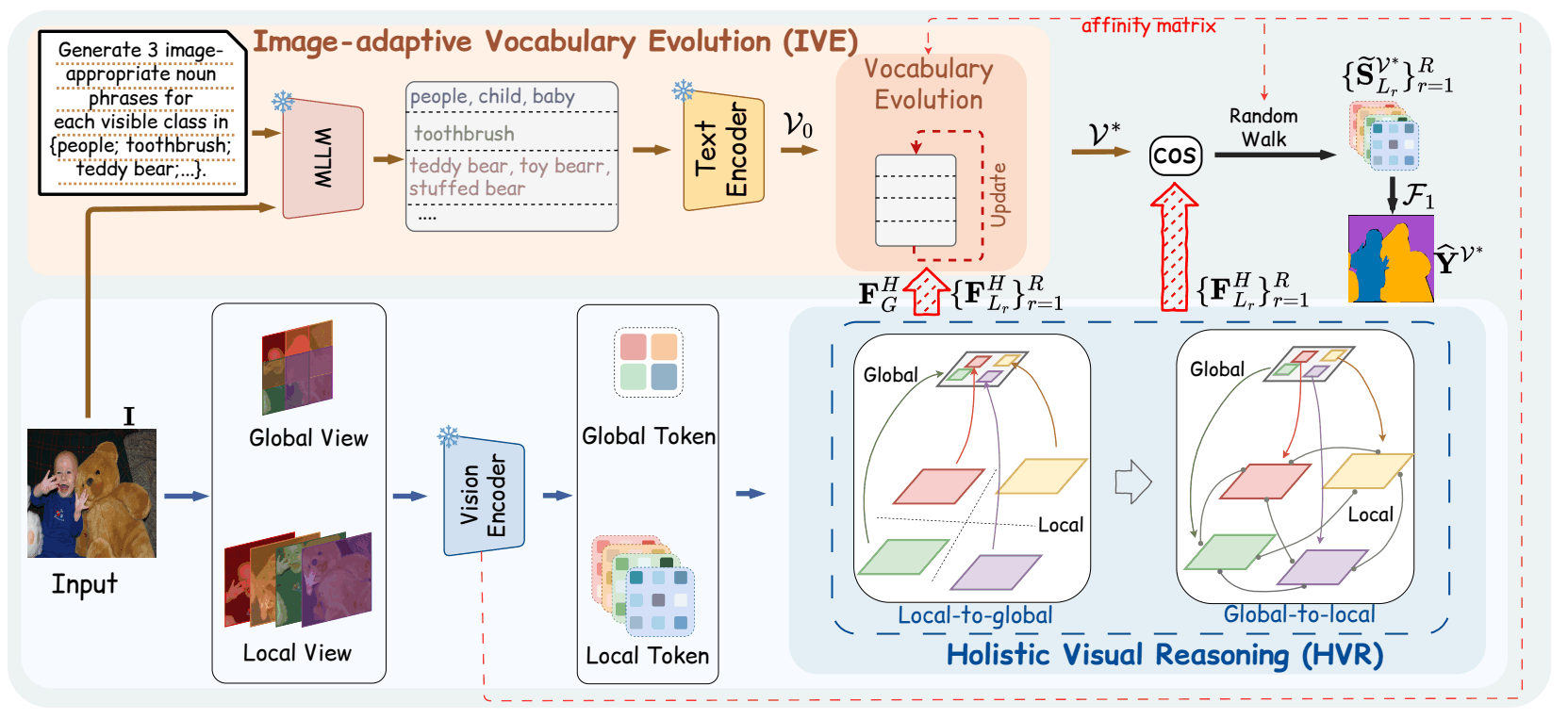}\end{minipage}\par
\vspace{3pt}{\scriptsize \setlength{\tabcolsep}{1.5pt}\begin{tabular*}{\linewidth}{@{\extracolsep{\fill}}rrrrrrr@{}}
\toprule
C & O & G & A & I & S & Mean \\
3 & 4 & 3 & 3 & 4 & 5 & 3.67\\
\bottomrule
\end{tabular*}
}
\end{minipage}\hfill
\begin{minipage}[t]{0.486\linewidth}
\centering\textbf{Study-generated derivative}\par\vspace{2pt}
\begin{minipage}[c][0.255\textheight][c]{\linewidth}\centering\includegraphics[width=\linewidth,height=0.255\textheight,keepaspectratio]{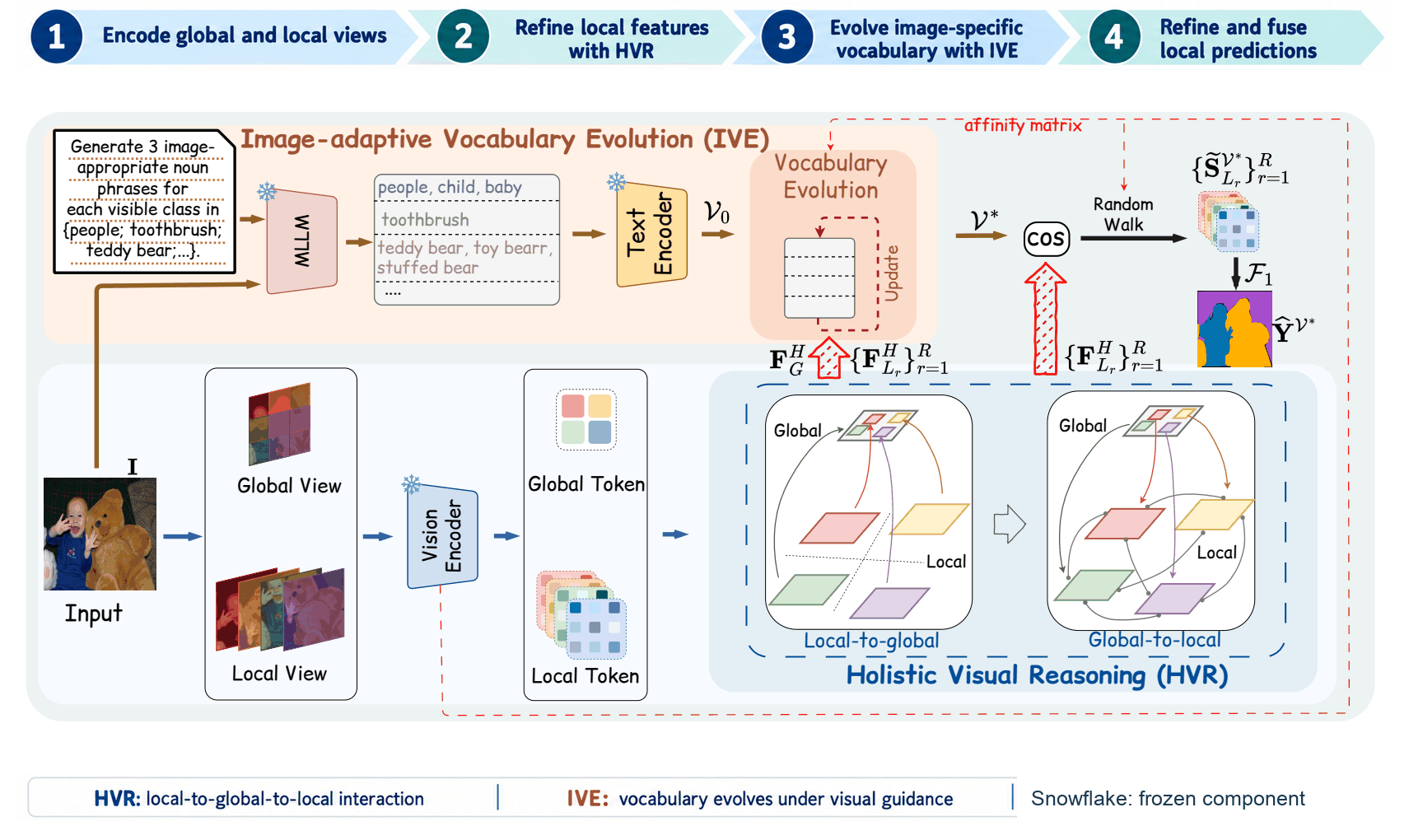}\end{minipage}\par
\vspace{3pt}
{\scriptsize \setlength{\tabcolsep}{1.5pt}\begin{tabular*}{\linewidth}{@{\extracolsep{\fill}}rrrrrrr@{}}
\toprule
C & O & G & A & I & S & Mean \\
4 (+1) & 4 (+0) & 3 (+0) & 3 (+0) & 4 (+0) & 5 (+0) & 3.83 (+0.17)\\
\bottomrule
\end{tabular*}
}
\end{minipage}\par
\vspace{3pt}
{\scriptsize Source public-date proxy: 2026-10-03. All displayed scores are medians of three runs; parentheses show revised minus original.\par}
\vfill
{\scriptsize C: clarity; O: organization/readability; G: graphical consistency; A: aesthetic coherence; I: visual inventiveness; S: design specificity. Mean is the arithmetic mean of six dimension medians, a descriptive display value, not a validated quality index.\par}

\clearpage
\phantomsection
\label{atlas:corpus143}
\pdfbookmark[1]{143. Fixing the Last Mile: Failure-Aware Image Restoration with Region-Specific Guidance}{corpus143}
{\Large\bfseries\color{academicblue} 143 \quad \href{https://openreview.net/forum?id=kQkih2eMHu}{Fixing the Last Mile: Failure-Aware Image Restoration with Region-Specific Guidance} \citep{corpus143}\par}
\vspace{3pt}
{\small S4 \enspace $\vert$ \enspace Image restoration \enspace $\vert$ \enspace ICLR 2027 submission (under review)\par}
\vspace{4pt}
\vspace{5pt}
{\bfseries Motivation (source Fig.\ 1)} \hfill {\footnotesize \href{https://openreview.net/attachment?id=kQkih2eMHu\&name=pdf}{Source PDF, p.\ 2}; ID: ICLR27\_kQkih2eMHu\_M}\par
\vspace{2pt}
\noindent\begin{minipage}[t]{0.486\linewidth}
\centering\textbf{Original}\par\vspace{2pt}
\begin{minipage}[c][0.255\textheight][c]{\linewidth}\centering\includegraphics[width=\linewidth,height=0.255\textheight,keepaspectratio]{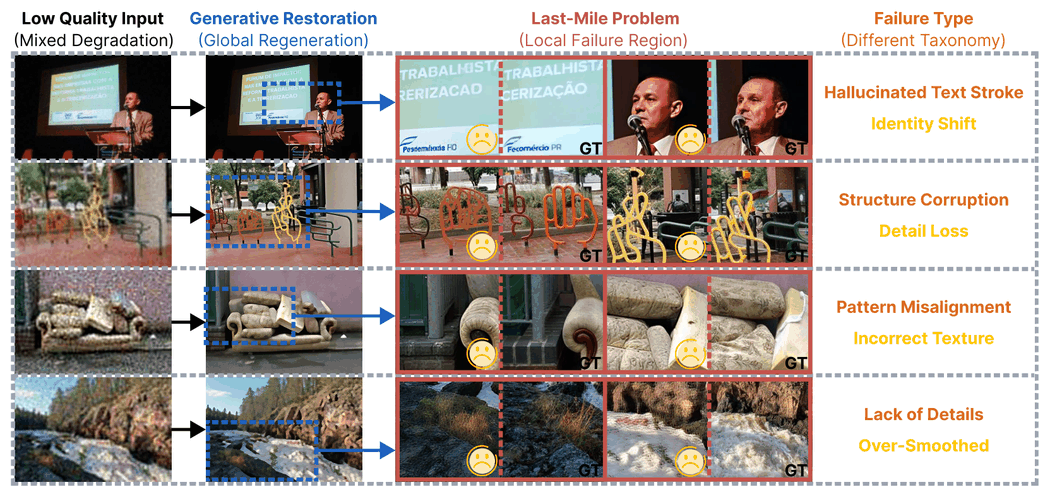}\end{minipage}\par
\vspace{3pt}{\scriptsize \setlength{\tabcolsep}{1.5pt}\begin{tabular*}{\linewidth}{@{\extracolsep{\fill}}rrrrrrr@{}}
\toprule
C & O & G & A & I & S & Mean \\
4 & 4 & 4 & 3 & 4 & 5 & 4.00\\
\bottomrule
\end{tabular*}
}
\end{minipage}\hfill
\begin{minipage}[t]{0.486\linewidth}
\centering\textbf{Study-generated derivative}\par\vspace{2pt}
\begin{minipage}[c][0.255\textheight][c]{\linewidth}\centering\includegraphics[width=\linewidth,height=0.255\textheight,keepaspectratio]{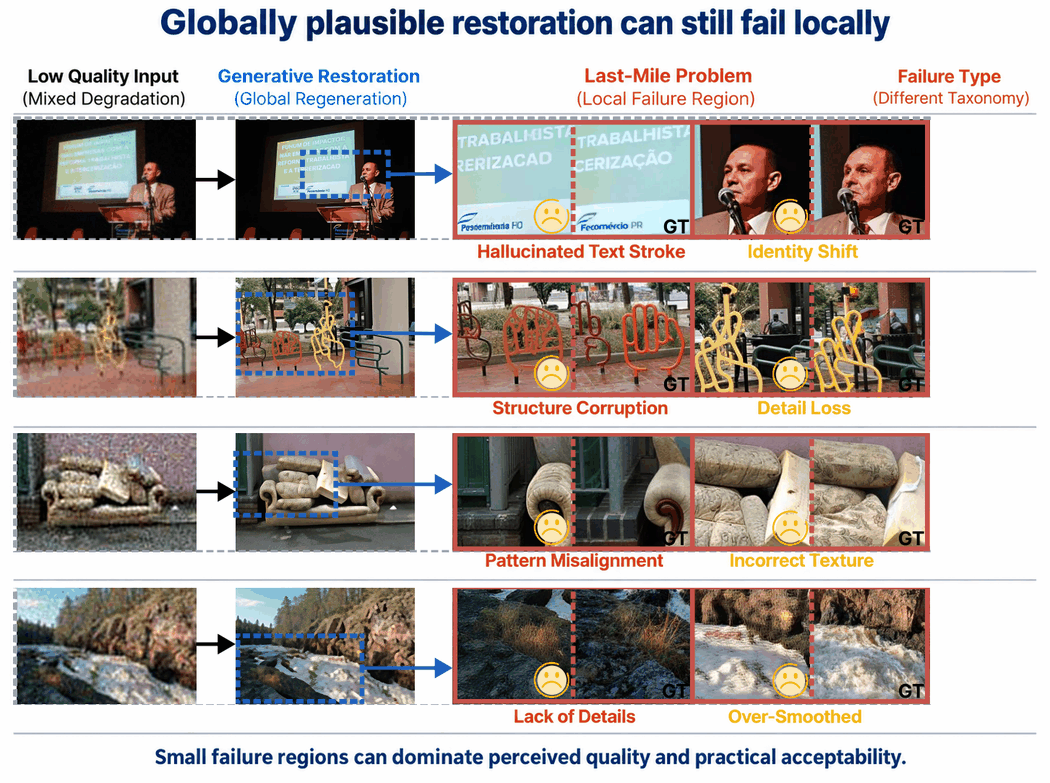}\end{minipage}\par
\vspace{3pt}
{\scriptsize \setlength{\tabcolsep}{1.5pt}\begin{tabular*}{\linewidth}{@{\extracolsep{\fill}}rrrrrrr@{}}
\toprule
C & O & G & A & I & S & Mean \\
5 (+1) & 4 (+0) & 4 (+0) & 3 (+0) & 4 (+0) & 5 (+0) & 4.17 (+0.17)\\
\bottomrule
\end{tabular*}
}
\end{minipage}\par
\vspace{3pt}
{\scriptsize Source public-date proxy: 2026-10-03. All displayed scores are medians of three runs; parentheses show revised minus original.\par}
\vspace{5pt}
{\bfseries Framework (source Fig.\ 6)} \hfill {\footnotesize \href{https://openreview.net/attachment?id=kQkih2eMHu\&name=pdf}{Source PDF, p.\ 6}; ID: ICLR27\_kQkih2eMHu\_F}\par
\vspace{2pt}
\noindent\begin{minipage}[t]{0.486\linewidth}
\centering\textbf{Original}\par\vspace{2pt}
\begin{minipage}[c][0.255\textheight][c]{\linewidth}\centering\includegraphics[width=\linewidth,height=0.255\textheight,keepaspectratio]{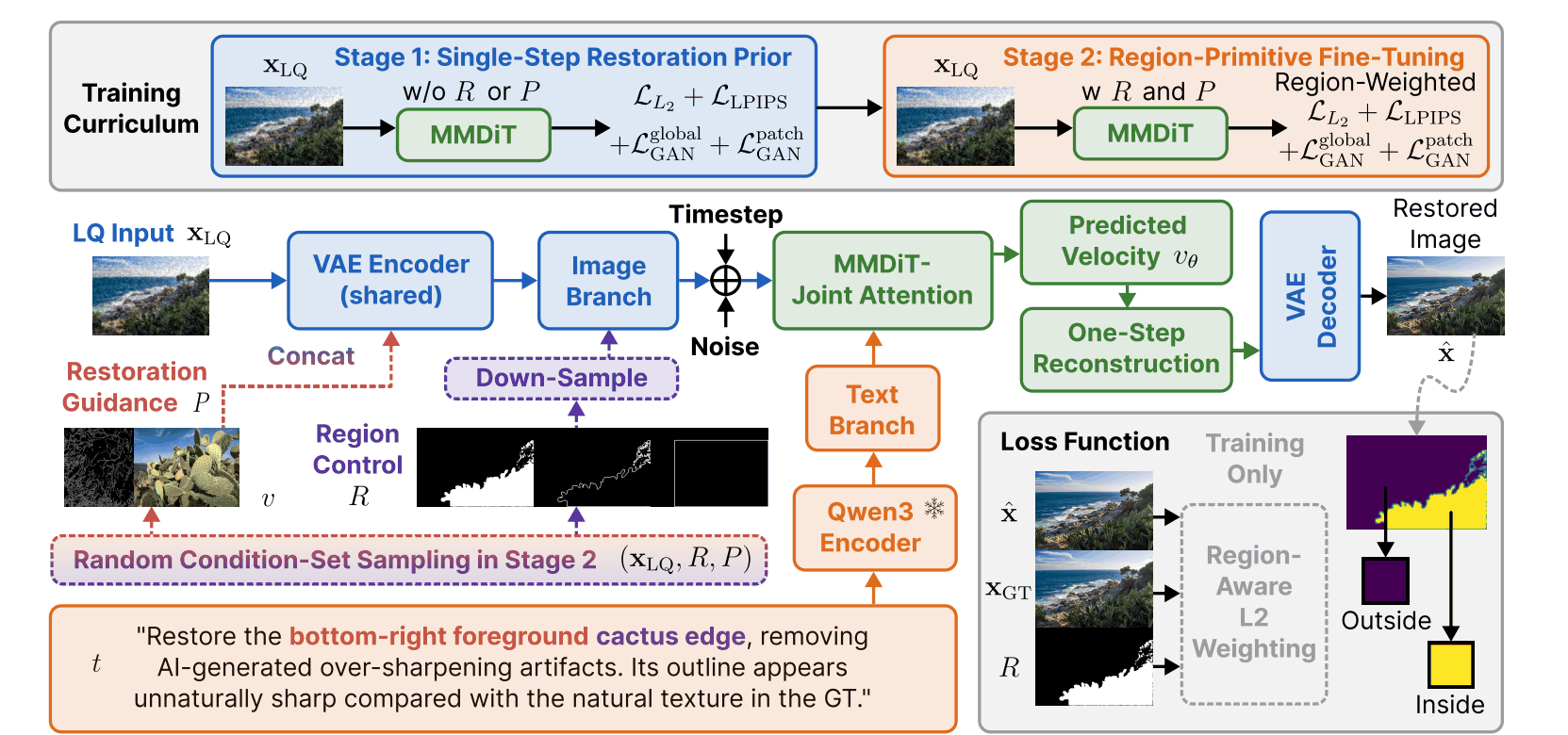}\end{minipage}\par
\vspace{3pt}{\scriptsize \setlength{\tabcolsep}{1.5pt}\begin{tabular*}{\linewidth}{@{\extracolsep{\fill}}rrrrrrr@{}}
\toprule
C & O & G & A & I & S & Mean \\
4 & 4 & 4 & 3 & 3 & 5 & 3.83\\
\bottomrule
\end{tabular*}
}
\end{minipage}\hfill
\begin{minipage}[t]{0.486\linewidth}
\centering\textbf{Study-generated derivative}\par\vspace{2pt}
\begin{minipage}[c][0.255\textheight][c]{\linewidth}\centering\includegraphics[width=\linewidth,height=0.255\textheight,keepaspectratio]{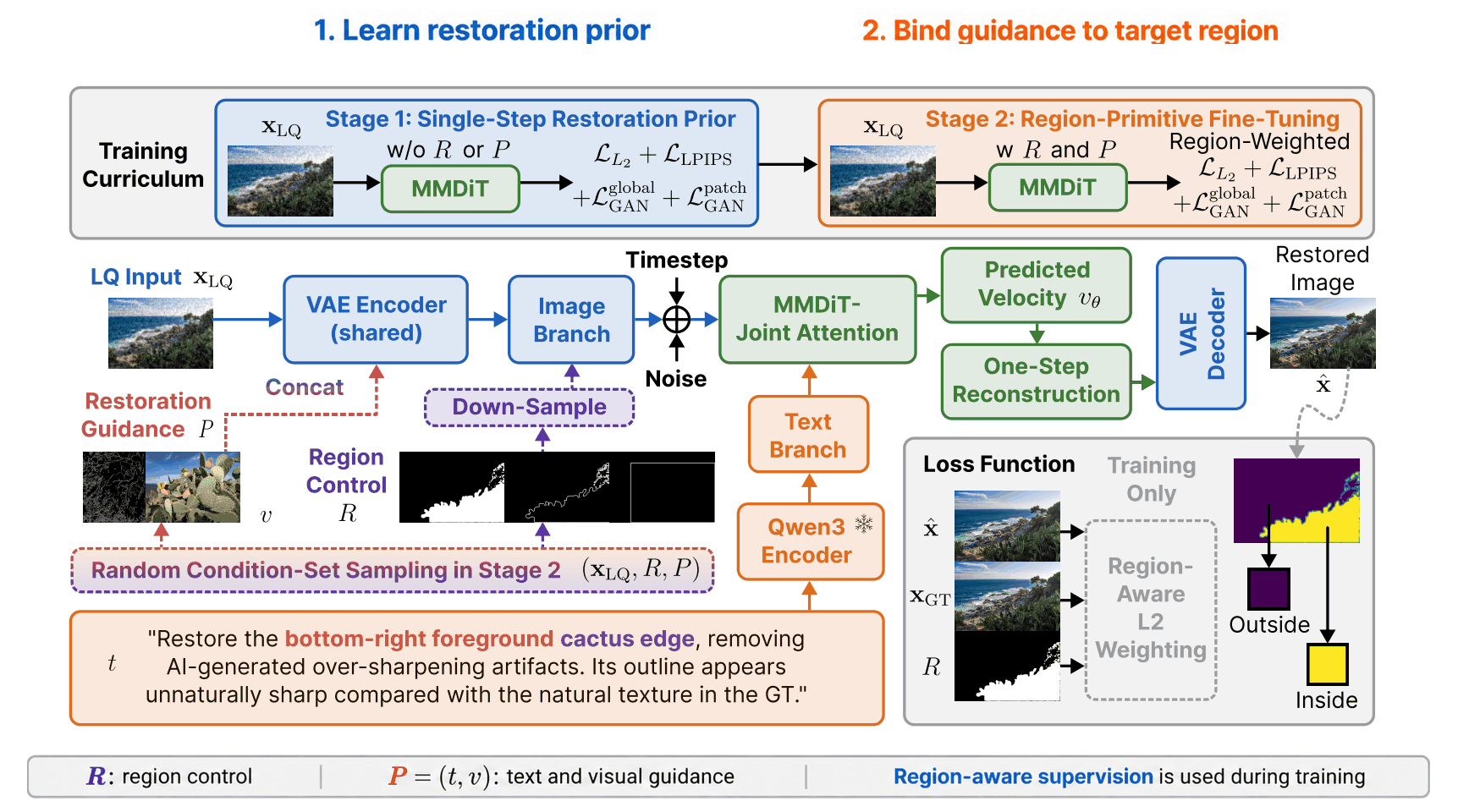}\end{minipage}\par
\vspace{3pt}
{\scriptsize \setlength{\tabcolsep}{1.5pt}\begin{tabular*}{\linewidth}{@{\extracolsep{\fill}}rrrrrrr@{}}
\toprule
C & O & G & A & I & S & Mean \\
4 (+0) & 4 (+0) & 4 (+0) & 4 (+1) & 3 (+0) & 5 (+0) & 4.00 (+0.17)\\
\bottomrule
\end{tabular*}
}
\end{minipage}\par
\vspace{3pt}
{\scriptsize Source public-date proxy: 2026-10-03. All displayed scores are medians of three runs; parentheses show revised minus original.\par}
\vfill
{\scriptsize C: clarity; O: organization/readability; G: graphical consistency; A: aesthetic coherence; I: visual inventiveness; S: design specificity. Mean is the arithmetic mean of six dimension medians, a descriptive display value, not a validated quality index.\par}

\clearpage
\phantomsection
\label{atlas:corpus147}
\pdfbookmark[1]{147. Look Before You Restore: Learning to Predict Before Acting for Agentic Image Restoration}{corpus147}
{\Large\bfseries\color{academicblue} 147 \quad \href{https://openreview.net/forum?id=rCgxaWvBRj}{Look Before You Restore: Learning to Predict Before Acting for Agentic Image Restoration} \citep{corpus147}\par}
\vspace{3pt}
{\small S4 \enspace $\vert$ \enspace Image restoration \enspace $\vert$ \enspace ICLR 2027 submission (under review)\par}
\vspace{4pt}
\vspace{5pt}
{\bfseries Motivation (source Fig.\ 1)} \hfill {\footnotesize \href{https://openreview.net/attachment?id=rCgxaWvBRj\&name=pdf}{Source PDF, p.\ 2}; ID: ICLR27\_rCgxaWvBRj\_M}\par
\vspace{2pt}
\noindent\begin{minipage}[t]{0.486\linewidth}
\centering\textbf{Original}\par\vspace{2pt}
\begin{minipage}[c][0.255\textheight][c]{\linewidth}\centering\includegraphics[width=\linewidth,height=0.255\textheight,keepaspectratio]{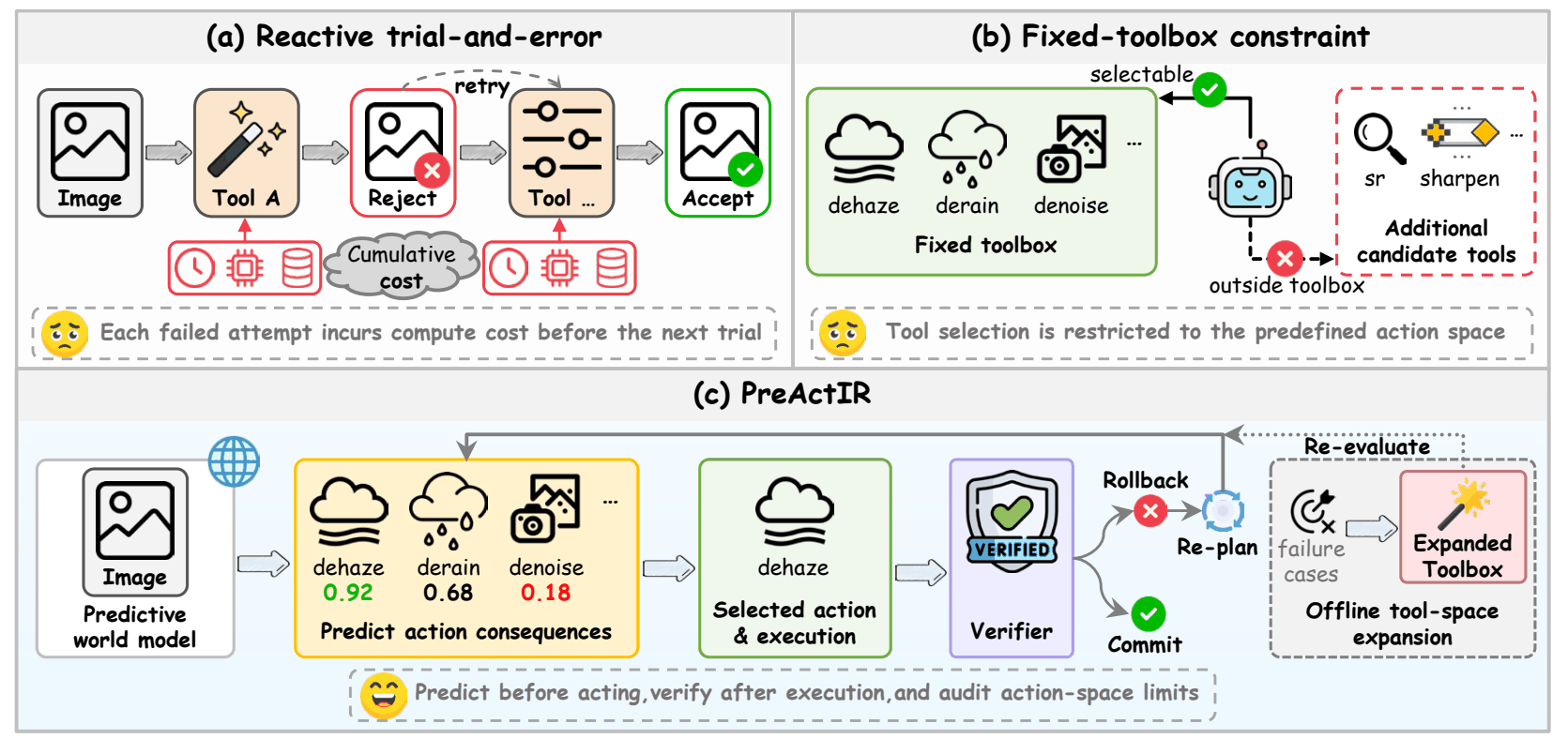}\end{minipage}\par
\vspace{3pt}{\scriptsize \setlength{\tabcolsep}{1.5pt}\begin{tabular*}{\linewidth}{@{\extracolsep{\fill}}rrrrrrr@{}}
\toprule
C & O & G & A & I & S & Mean \\
5 & 4 & 4 & 4 & 4 & 4 & 4.17\\
\bottomrule
\end{tabular*}
}
\end{minipage}\hfill
\begin{minipage}[t]{0.486\linewidth}
\centering\textbf{Study-generated derivative}\par\vspace{2pt}
\begin{minipage}[c][0.255\textheight][c]{\linewidth}\centering\includegraphics[width=\linewidth,height=0.255\textheight,keepaspectratio]{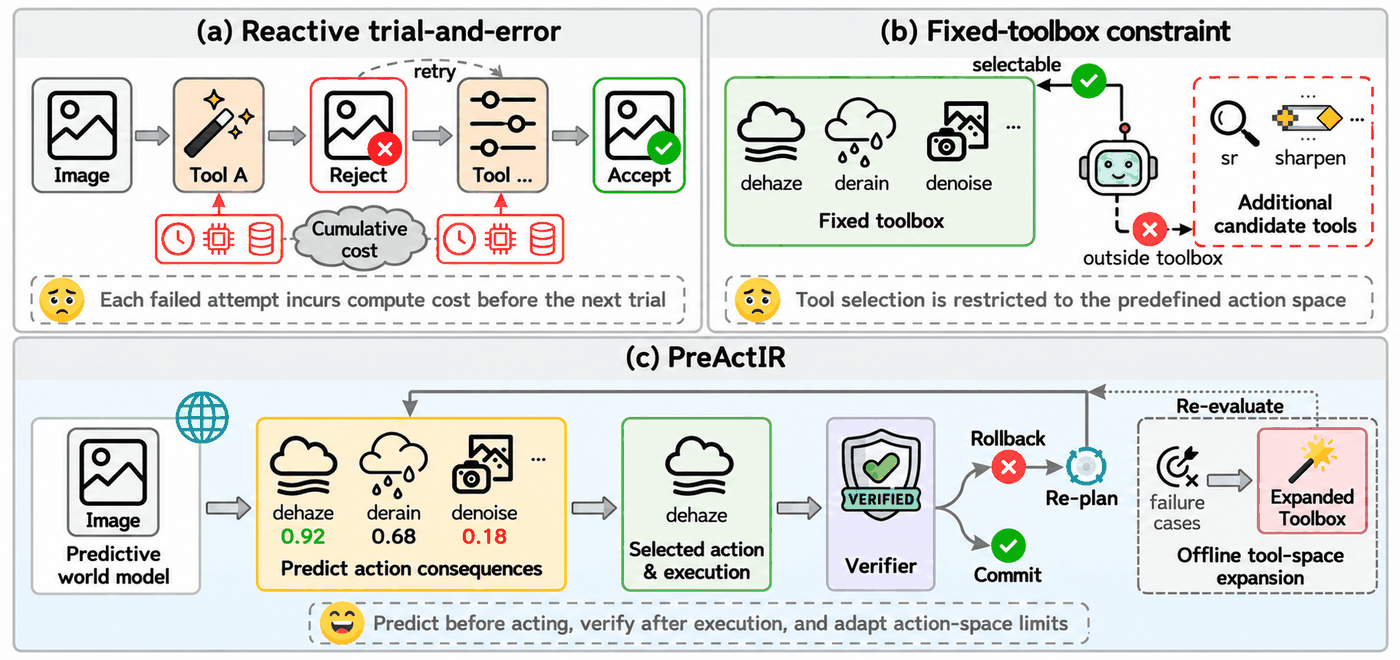}\end{minipage}\par
\vspace{3pt}
{\scriptsize \setlength{\tabcolsep}{1.5pt}\begin{tabular*}{\linewidth}{@{\extracolsep{\fill}}rrrrrrr@{}}
\toprule
C & O & G & A & I & S & Mean \\
5 (+0) & 4 (+0) & 4 (+0) & 4 (+0) & 4 (+0) & 4 (+0) & 4.17 (+0.00)\\
\bottomrule
\end{tabular*}
}
\end{minipage}\par
\vspace{3pt}
{\scriptsize Source public-date proxy: 2026-10-03. All displayed scores are medians of three runs; parentheses show revised minus original.\par}
\vspace{5pt}
{\bfseries Framework (source Fig.\ 2)} \hfill {\footnotesize \href{https://openreview.net/attachment?id=rCgxaWvBRj\&name=pdf}{Source PDF, p.\ 4}; ID: ICLR27\_rCgxaWvBRj\_F}\par
\vspace{2pt}
\noindent\begin{minipage}[t]{0.486\linewidth}
\centering\textbf{Original}\par\vspace{2pt}
\begin{minipage}[c][0.255\textheight][c]{\linewidth}\centering\includegraphics[width=\linewidth,height=0.255\textheight,keepaspectratio]{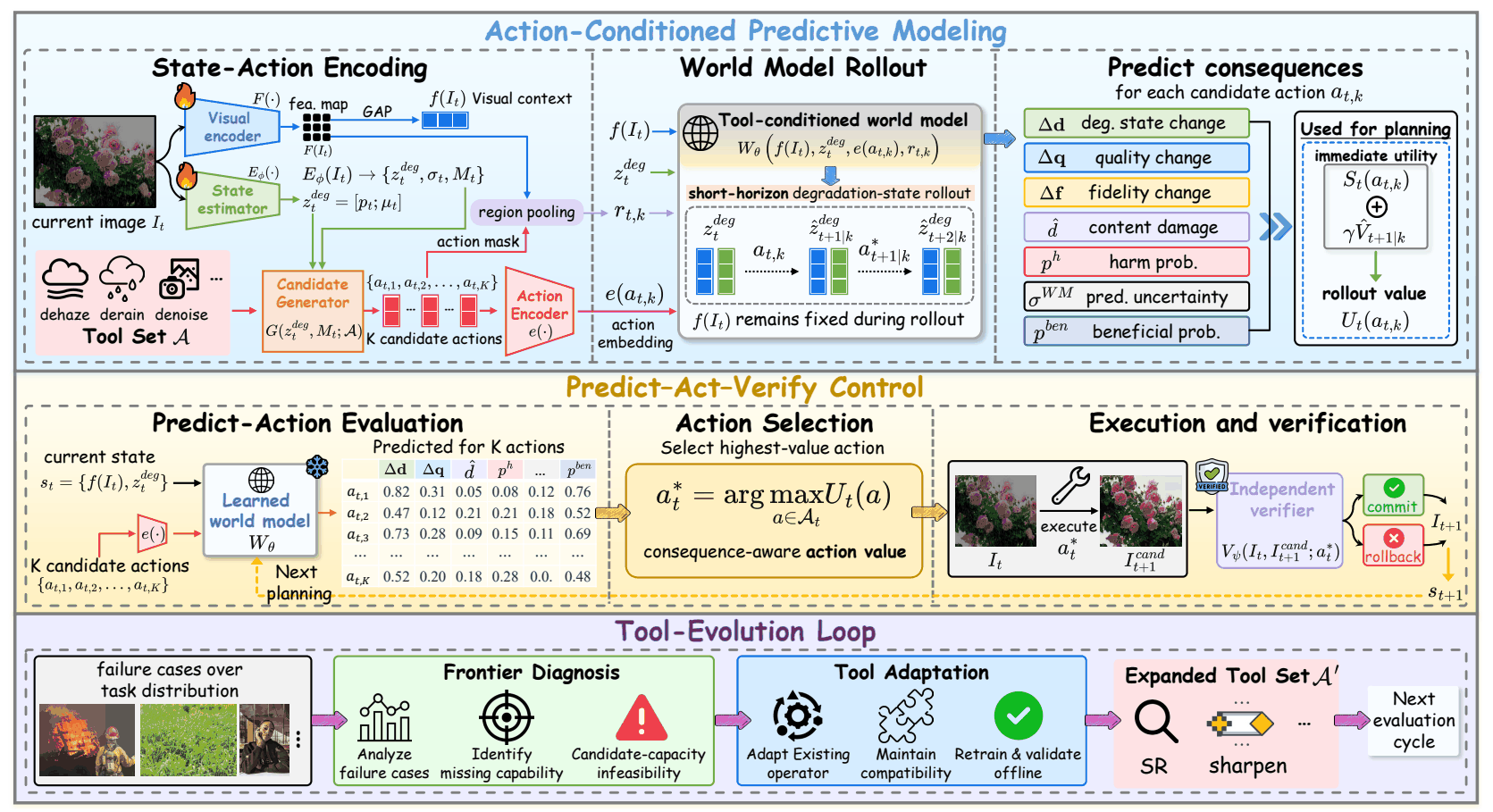}\end{minipage}\par
\vspace{3pt}{\scriptsize \setlength{\tabcolsep}{1.5pt}\begin{tabular*}{\linewidth}{@{\extracolsep{\fill}}rrrrrrr@{}}
\toprule
C & O & G & A & I & S & Mean \\
4 & 4 & 4 & 3 & 3 & 5 & 3.83\\
\bottomrule
\end{tabular*}
}
\end{minipage}\hfill
\begin{minipage}[t]{0.486\linewidth}
\centering\textbf{Study-generated derivative}\par\vspace{2pt}
\begin{minipage}[c][0.255\textheight][c]{\linewidth}\centering\includegraphics[width=\linewidth,height=0.255\textheight,keepaspectratio]{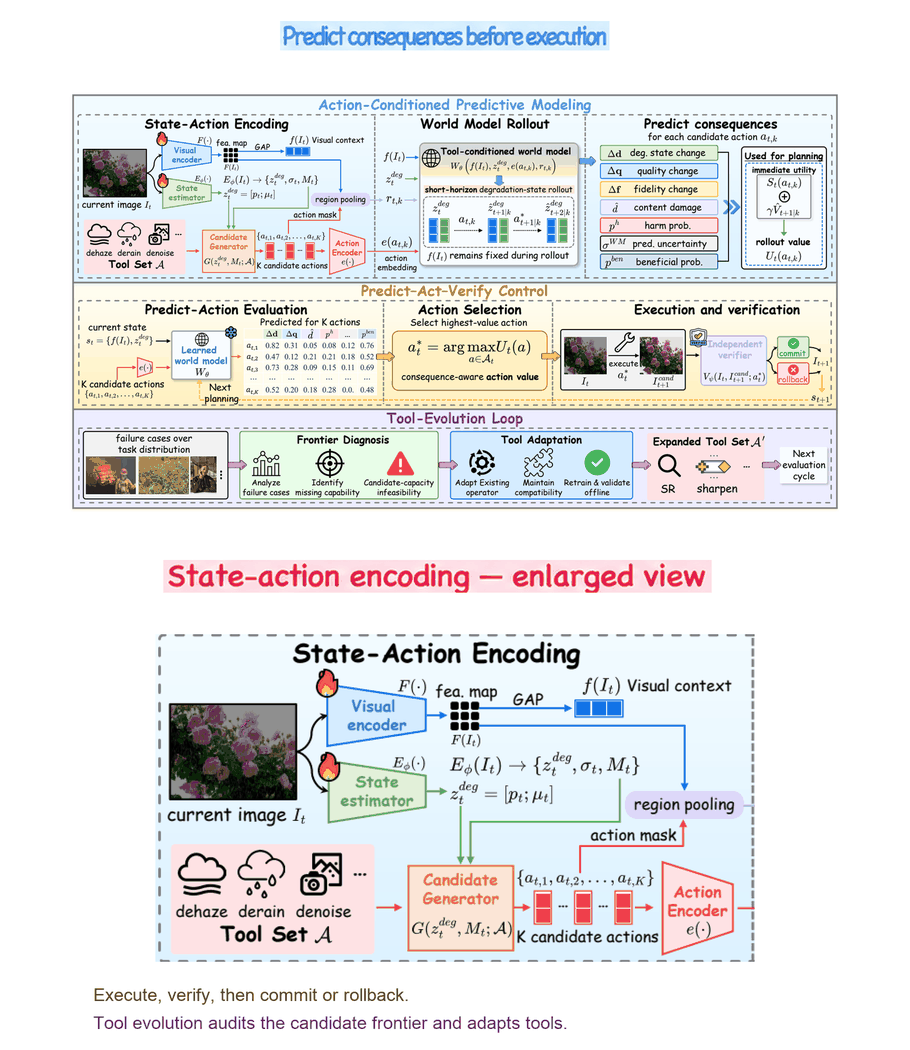}\end{minipage}\par
\vspace{3pt}
{\scriptsize \setlength{\tabcolsep}{1.5pt}\begin{tabular*}{\linewidth}{@{\extracolsep{\fill}}rrrrrrr@{}}
\toprule
C & O & G & A & I & S & Mean \\
4 (+0) & 3 (-1) & 4 (+0) & 3 (+0) & 3 (+0) & 5 (+0) & 3.67 (-0.17)\\
\bottomrule
\end{tabular*}
}
\end{minipage}\par
\vspace{3pt}
{\scriptsize Source public-date proxy: 2026-10-03. All displayed scores are medians of three runs; parentheses show revised minus original.\par}
\vfill
{\scriptsize C: clarity; O: organization/readability; G: graphical consistency; A: aesthetic coherence; I: visual inventiveness; S: design specificity. Mean is the arithmetic mean of six dimension medians, a descriptive display value, not a validated quality index.\par}

\clearpage
\phantomsection
\label{atlas:corpus155}
\pdfbookmark[1]{155. LonGaussian: Tracking Foundation Priors for Long-Context Feedforward Gaussian Reconstruction}{corpus155}
{\Large\bfseries\color{academicblue} 155 \quad \href{https://openreview.net/forum?id=WiNCsgDP25}{LonGaussian: Tracking Foundation Priors for Long-Context Feedforward Gaussian Reconstruction} \citep{corpus155}\par}
\vspace{3pt}
{\small S4 \enspace $\vert$ \enspace 3D vision \enspace $\vert$ \enspace ICLR 2027 submission (under review)\par}
\vspace{4pt}
\vspace{5pt}
{\bfseries Motivation (source Fig.\ 2)} \hfill {\footnotesize \href{https://openreview.net/attachment?id=WiNCsgDP25\&name=pdf}{Source PDF, p.\ 2}; ID: ICLR27\_WiNCsgDP25\_M}\par
\vspace{2pt}
\noindent\begin{minipage}[t]{0.486\linewidth}
\centering\textbf{Original}\par\vspace{2pt}
\begin{minipage}[c][0.255\textheight][c]{\linewidth}\centering\includegraphics[width=\linewidth,height=0.255\textheight,keepaspectratio]{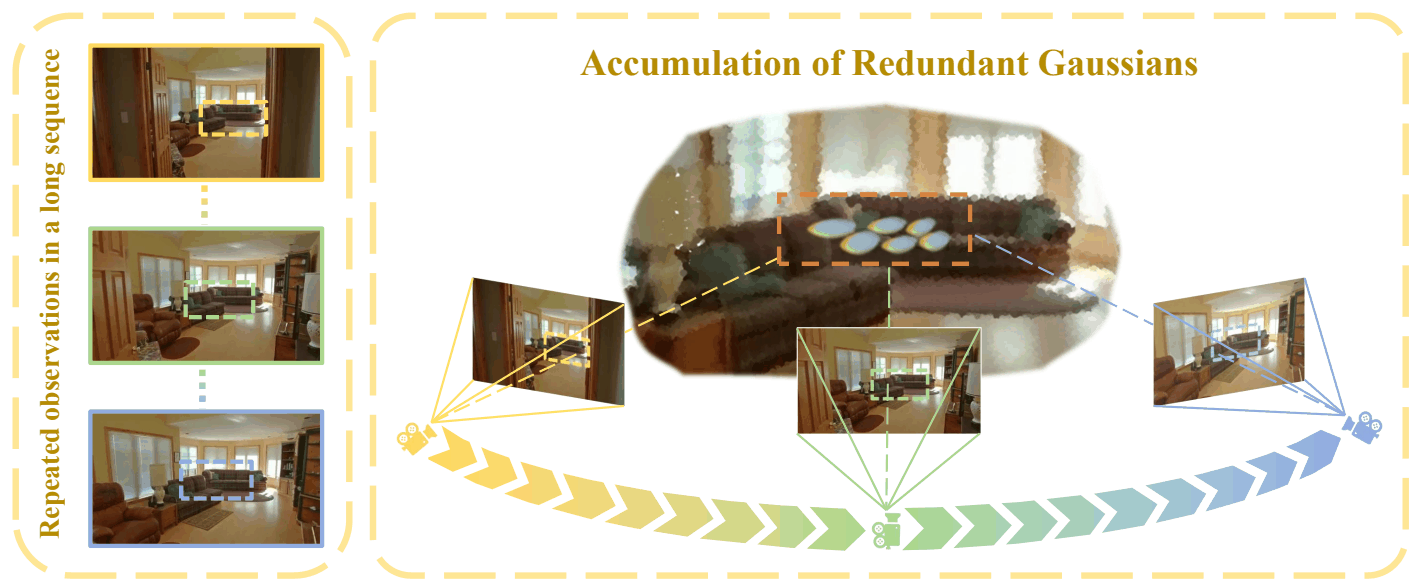}\end{minipage}\par
\vspace{3pt}{\scriptsize \setlength{\tabcolsep}{1.5pt}\begin{tabular*}{\linewidth}{@{\extracolsep{\fill}}rrrrrrr@{}}
\toprule
C & O & G & A & I & S & Mean \\
3 & 4 & 4 & 4 & 4 & 4 & 3.83\\
\bottomrule
\end{tabular*}
}
\end{minipage}\hfill
\begin{minipage}[t]{0.486\linewidth}
\centering\textbf{Study-generated derivative}\par\vspace{2pt}
\begin{minipage}[c][0.255\textheight][c]{\linewidth}\centering\includegraphics[width=\linewidth,height=0.255\textheight,keepaspectratio]{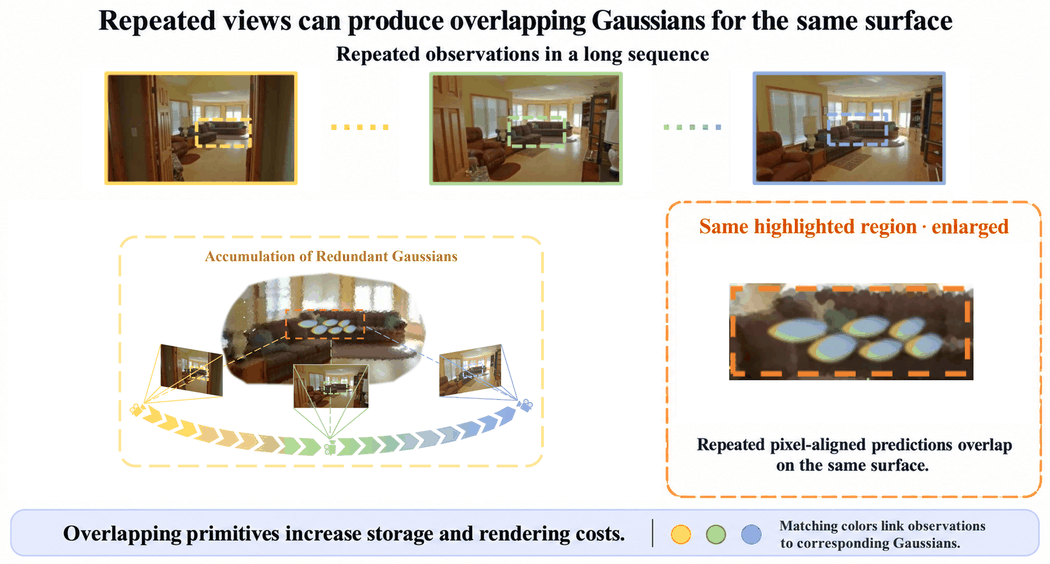}\end{minipage}\par
\vspace{3pt}
{\scriptsize \setlength{\tabcolsep}{1.5pt}\begin{tabular*}{\linewidth}{@{\extracolsep{\fill}}rrrrrrr@{}}
\toprule
C & O & G & A & I & S & Mean \\
5 (+2) & 4 (+0) & 4 (+0) & 3 (-1) & 4 (+0) & 5 (+1) & 4.17 (+0.33)\\
\bottomrule
\end{tabular*}
}
\end{minipage}\par
\vspace{3pt}
{\scriptsize Source public-date proxy: 2026-10-03. All displayed scores are medians of three runs; parentheses show revised minus original.\par}
\vspace{5pt}
{\bfseries Framework (source Fig.\ 3)} \hfill {\footnotesize \href{https://openreview.net/attachment?id=WiNCsgDP25\&name=pdf}{Source PDF, p.\ 4}; ID: ICLR27\_WiNCsgDP25\_F}\par
\vspace{2pt}
\noindent\begin{minipage}[t]{0.486\linewidth}
\centering\textbf{Original}\par\vspace{2pt}
\begin{minipage}[c][0.255\textheight][c]{\linewidth}\centering\includegraphics[width=\linewidth,height=0.255\textheight,keepaspectratio]{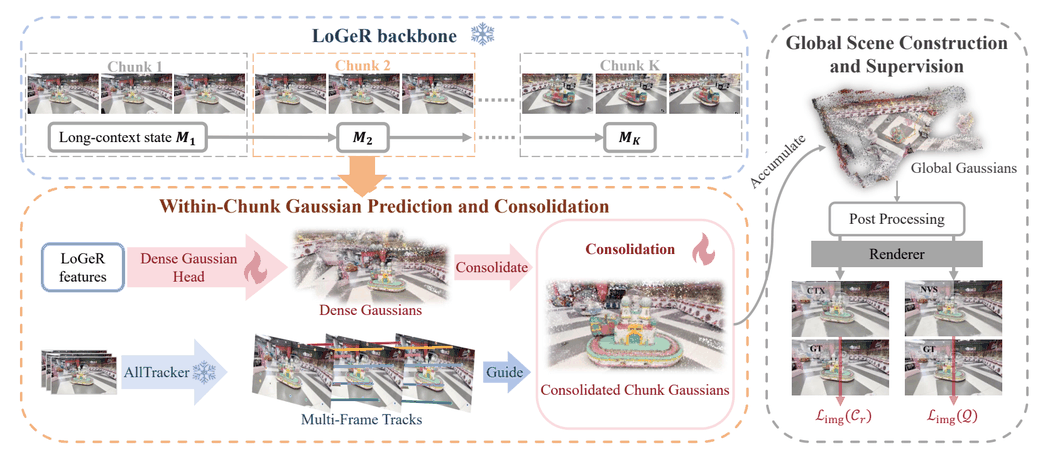}\end{minipage}\par
\vspace{3pt}{\scriptsize \setlength{\tabcolsep}{1.5pt}\begin{tabular*}{\linewidth}{@{\extracolsep{\fill}}rrrrrrr@{}}
\toprule
C & O & G & A & I & S & Mean \\
4 & 4 & 4 & 4 & 4 & 5 & 4.17\\
\bottomrule
\end{tabular*}
}
\end{minipage}\hfill
\begin{minipage}[t]{0.486\linewidth}
\centering\textbf{Study-generated derivative}\par\vspace{2pt}
\begin{minipage}[c][0.255\textheight][c]{\linewidth}\centering\includegraphics[width=\linewidth,height=0.255\textheight,keepaspectratio]{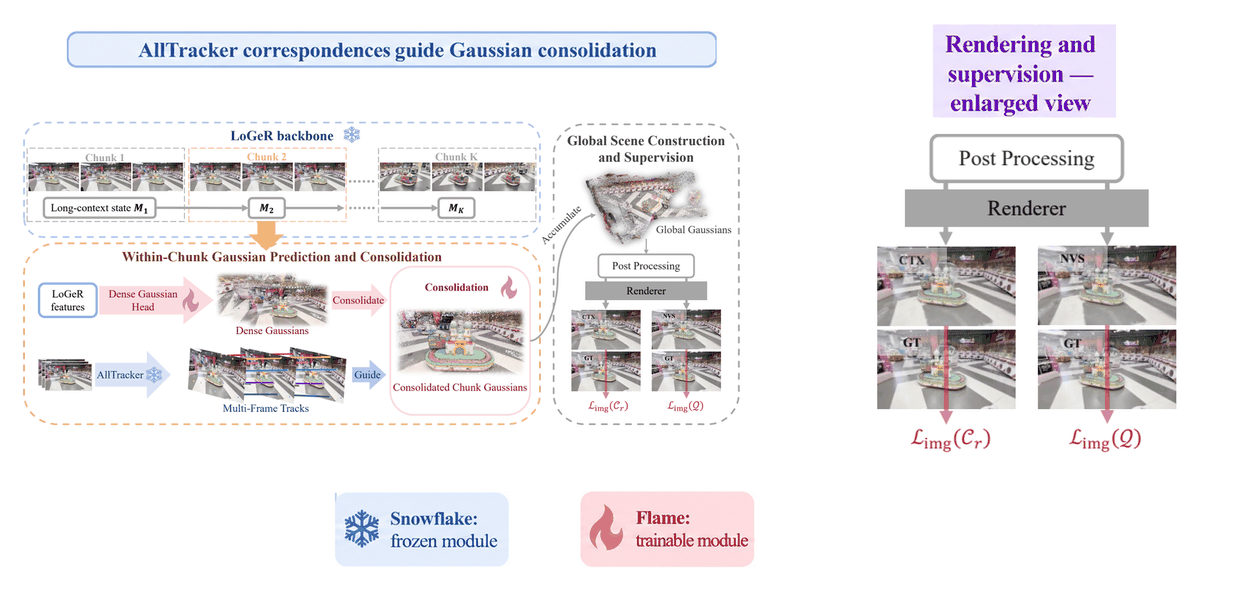}\end{minipage}\par
\vspace{3pt}
{\scriptsize \setlength{\tabcolsep}{1.5pt}\begin{tabular*}{\linewidth}{@{\extracolsep{\fill}}rrrrrrr@{}}
\toprule
C & O & G & A & I & S & Mean \\
4 (+0) & 3 (-1) & 4 (+0) & 3 (-1) & 3 (-1) & 5 (+0) & 3.67 (-0.50)\\
\bottomrule
\end{tabular*}
}
\end{minipage}\par
\vspace{3pt}
{\scriptsize Source public-date proxy: 2026-10-03. All displayed scores are medians of three runs; parentheses show revised minus original.\par}
\vfill
{\scriptsize C: clarity; O: organization/readability; G: graphical consistency; A: aesthetic coherence; I: visual inventiveness; S: design specificity. Mean is the arithmetic mean of six dimension medians, a descriptive display value, not a validated quality index.\par}

\clearpage
\phantomsection
\label{atlas:corpus160}
\pdfbookmark[1]{160. VASC: Value-Aware Sparse Attention with Cross-layer Memory for Efficient 3D Reconstruction}{corpus160}
{\Large\bfseries\color{academicblue} 160 \quad \href{https://openreview.net/forum?id=X7FI3N35Hq}{VASC: Value-Aware Sparse Attention with Cross-layer Memory for Efficient 3D Reconstruction} \citep{corpus160}\par}
\vspace{3pt}
{\small S4 \enspace $\vert$ \enspace 3D vision \enspace $\vert$ \enspace ICLR 2027 submission (under review)\par}
\vspace{4pt}
\vspace{5pt}
{\bfseries Motivation (source Fig.\ 3)} \hfill {\footnotesize \href{https://openreview.net/attachment?id=X7FI3N35Hq\&name=pdf}{Source PDF, p.\ 4}; ID: ICLR27\_X7FI3N35Hq\_M}\par
\vspace{2pt}
\noindent\begin{minipage}[t]{0.486\linewidth}
\centering\textbf{Original}\par\vspace{2pt}
\begin{minipage}[c][0.255\textheight][c]{\linewidth}\centering\includegraphics[width=\linewidth,height=0.255\textheight,keepaspectratio]{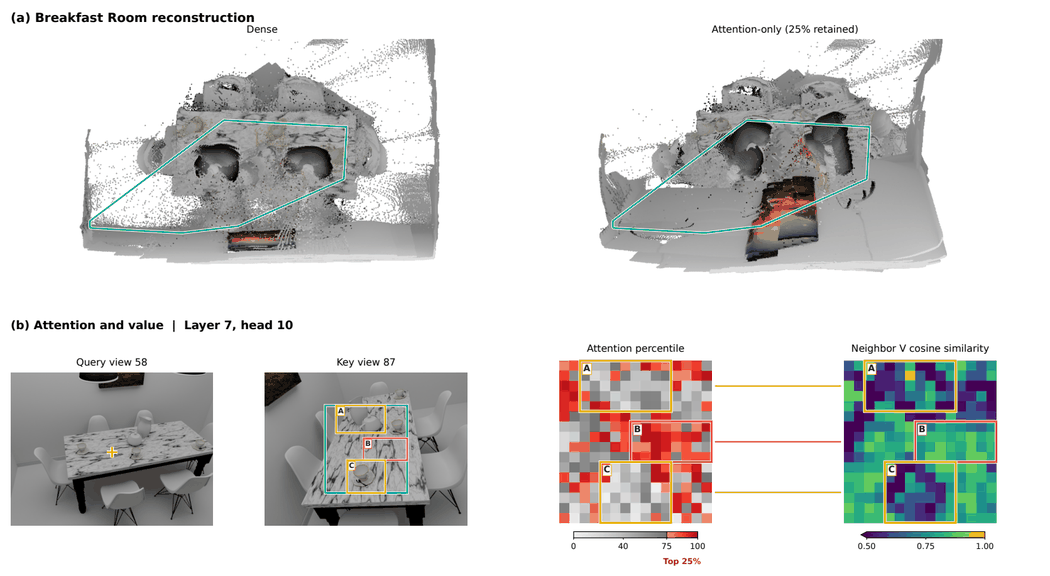}\end{minipage}\par
\vspace{3pt}{\scriptsize \setlength{\tabcolsep}{1.5pt}\begin{tabular*}{\linewidth}{@{\extracolsep{\fill}}rrrrrrr@{}}
\toprule
C & O & G & A & I & S & Mean \\
3 & 4 & 4 & 4 & 4 & 5 & 4.00\\
\bottomrule
\end{tabular*}
}
\end{minipage}\hfill
\begin{minipage}[t]{0.486\linewidth}
\centering\textbf{Study-generated derivative}\par\vspace{2pt}
\begin{minipage}[c][0.255\textheight][c]{\linewidth}\centering\includegraphics[width=\linewidth,height=0.255\textheight,keepaspectratio]{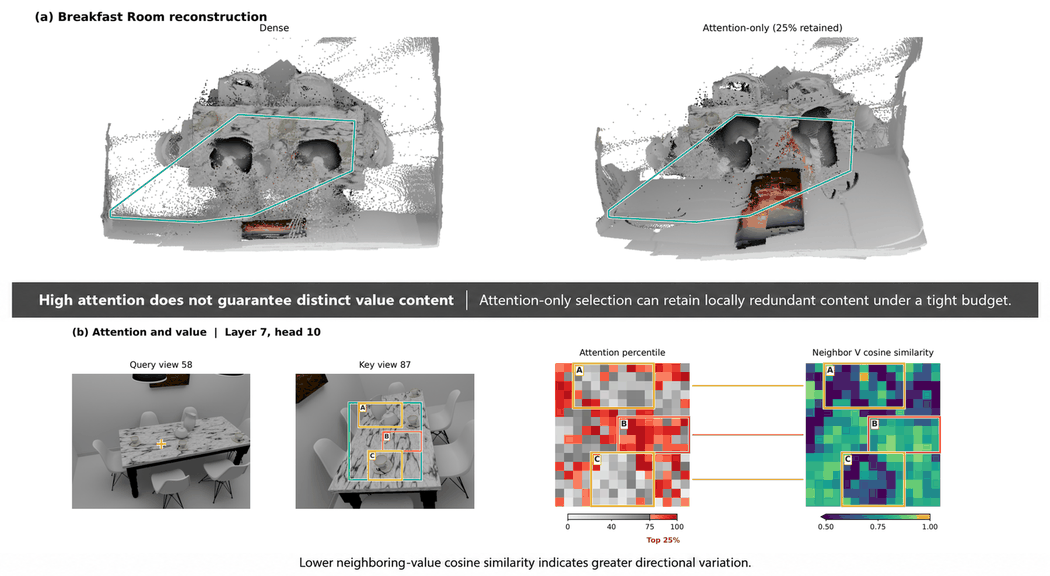}\end{minipage}\par
\vspace{3pt}
{\scriptsize \setlength{\tabcolsep}{1.5pt}\begin{tabular*}{\linewidth}{@{\extracolsep{\fill}}rrrrrrr@{}}
\toprule
C & O & G & A & I & S & Mean \\
4 (+1) & 4 (+0) & 4 (+0) & 4 (+0) & 4 (+0) & 5 (+0) & 4.17 (+0.17)\\
\bottomrule
\end{tabular*}
}
\end{minipage}\par
\vspace{3pt}
{\scriptsize Source public-date proxy: 2026-10-01. All displayed scores are medians of three runs; parentheses show revised minus original.\par}
\vspace{5pt}
{\bfseries Framework (source Fig.\ 4)} \hfill {\footnotesize \href{https://openreview.net/attachment?id=X7FI3N35Hq\&name=pdf}{Source PDF, p.\ 5}; ID: ICLR27\_X7FI3N35Hq\_F}\par
\vspace{2pt}
\noindent\begin{minipage}[t]{0.486\linewidth}
\centering\textbf{Original}\par\vspace{2pt}
\begin{minipage}[c][0.255\textheight][c]{\linewidth}\centering\includegraphics[width=\linewidth,height=0.255\textheight,keepaspectratio]{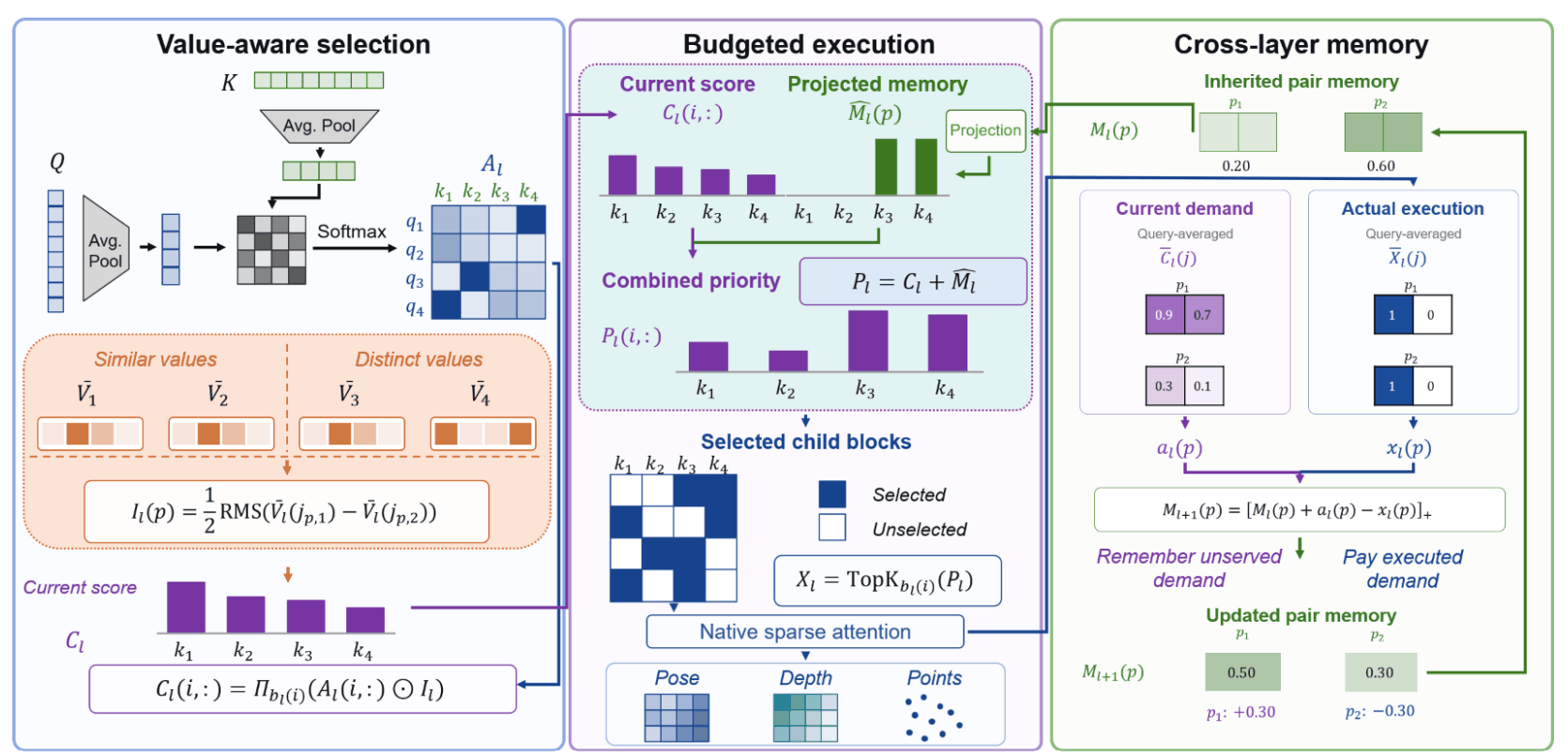}\end{minipage}\par
\vspace{3pt}{\scriptsize \setlength{\tabcolsep}{1.5pt}\begin{tabular*}{\linewidth}{@{\extracolsep{\fill}}rrrrrrr@{}}
\toprule
C & O & G & A & I & S & Mean \\
4 & 4 & 4 & 4 & 4 & 5 & 4.17\\
\bottomrule
\end{tabular*}
}
\end{minipage}\hfill
\begin{minipage}[t]{0.486\linewidth}
\centering\textbf{Study-generated derivative}\par\vspace{2pt}
\begin{minipage}[c][0.255\textheight][c]{\linewidth}\centering\includegraphics[width=\linewidth,height=0.255\textheight,keepaspectratio]{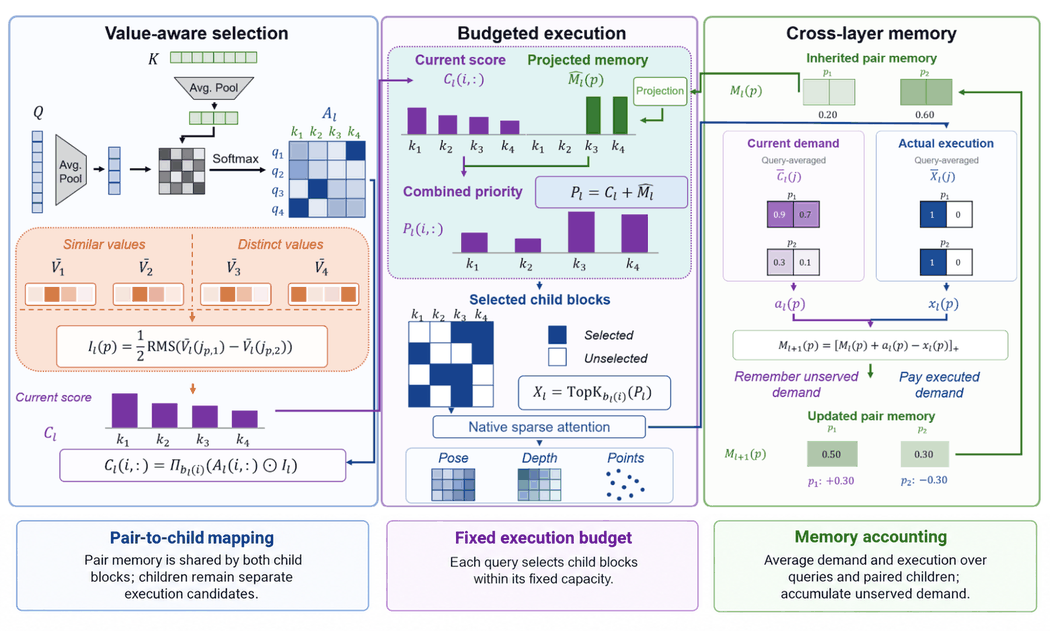}\end{minipage}\par
\vspace{3pt}
{\scriptsize \setlength{\tabcolsep}{1.5pt}\begin{tabular*}{\linewidth}{@{\extracolsep{\fill}}rrrrrrr@{}}
\toprule
C & O & G & A & I & S & Mean \\
4 (+0) & 4 (+0) & 4 (+0) & 4 (+0) & 4 (+0) & 5 (+0) & 4.17 (+0.00)\\
\bottomrule
\end{tabular*}
}
\end{minipage}\par
\vspace{3pt}
{\scriptsize Source public-date proxy: 2026-10-01. All displayed scores are medians of three runs; parentheses show revised minus original.\par}
\vfill
{\scriptsize C: clarity; O: organization/readability; G: graphical consistency; A: aesthetic coherence; I: visual inventiveness; S: design specificity. Mean is the arithmetic mean of six dimension medians, a descriptive display value, not a validated quality index.\par}

\clearpage
\phantomsection
\label{atlas:corpus162}
\pdfbookmark[1]{162. A Simulation-Grounded VLM Framework for Agentic Wildfire Monitoring and Reporting}{corpus162}
{\Large\bfseries\color{academicblue} 162 \quad \href{https://openreview.net/forum?id=gGGjmjPJi4}{A Simulation-Grounded VLM Framework for Agentic Wildfire Monitoring and Reporting} \citep{corpus162}\par}
\vspace{3pt}
{\small S4-AI-Style \enspace $\vert$ \enspace Wildfire monitoring and multimodal agents \enspace $\vert$ \enspace ICLR 2027 submission (under review)\par}
\vspace{4pt}
\vspace{5pt}
{\bfseries Framework (source Fig.\ 1)} \hfill {\footnotesize \href{https://openreview.net/pdf?id=gGGjmjPJi4}{Source PDF, p.\ 3}; ID: AI\_gGGjmjPJi4\_F}\par
\vspace{2pt}
\noindent\begin{minipage}[t]{0.486\linewidth}
\centering\textbf{Original}\par\vspace{2pt}
\begin{minipage}[c][0.52\textheight][c]{\linewidth}\centering\includegraphics[width=\linewidth,height=0.52\textheight,keepaspectratio]{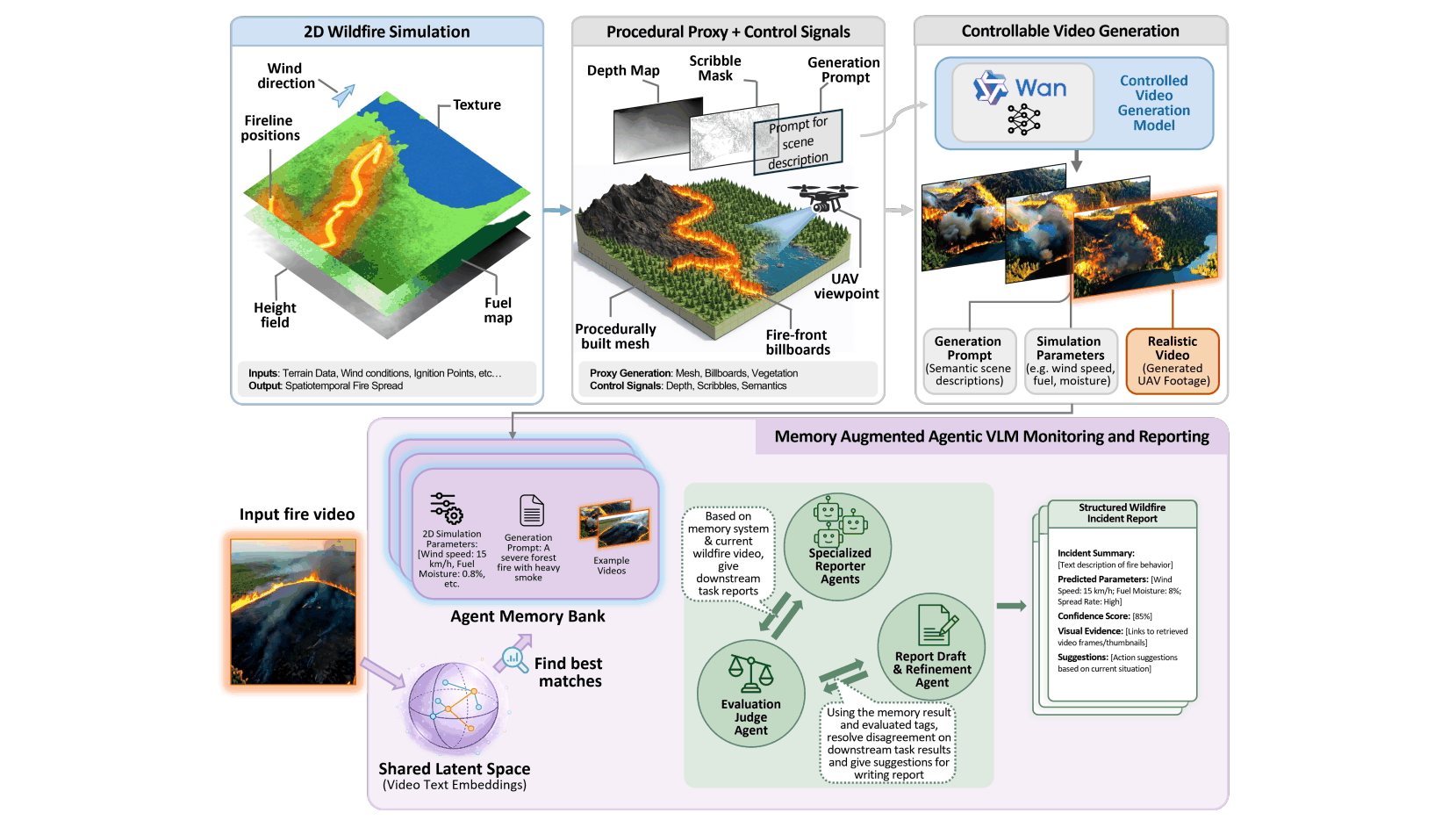}\end{minipage}\par
\vspace{3pt}{\scriptsize \setlength{\tabcolsep}{1.5pt}\begin{tabular*}{\linewidth}{@{\extracolsep{\fill}}rrrrrrr@{}}
\toprule
C & O & G & A & I & S & Mean \\
3 & 3 & 3 & 4 & 4 & 5 & 3.67\\
\bottomrule
\end{tabular*}
}
\end{minipage}\hfill
\begin{minipage}[t]{0.486\linewidth}
\centering\textbf{Study-generated derivative}\par\vspace{2pt}
\begin{minipage}[c][0.52\textheight][c]{\linewidth}\centering\includegraphics[width=\linewidth,height=0.52\textheight,keepaspectratio]{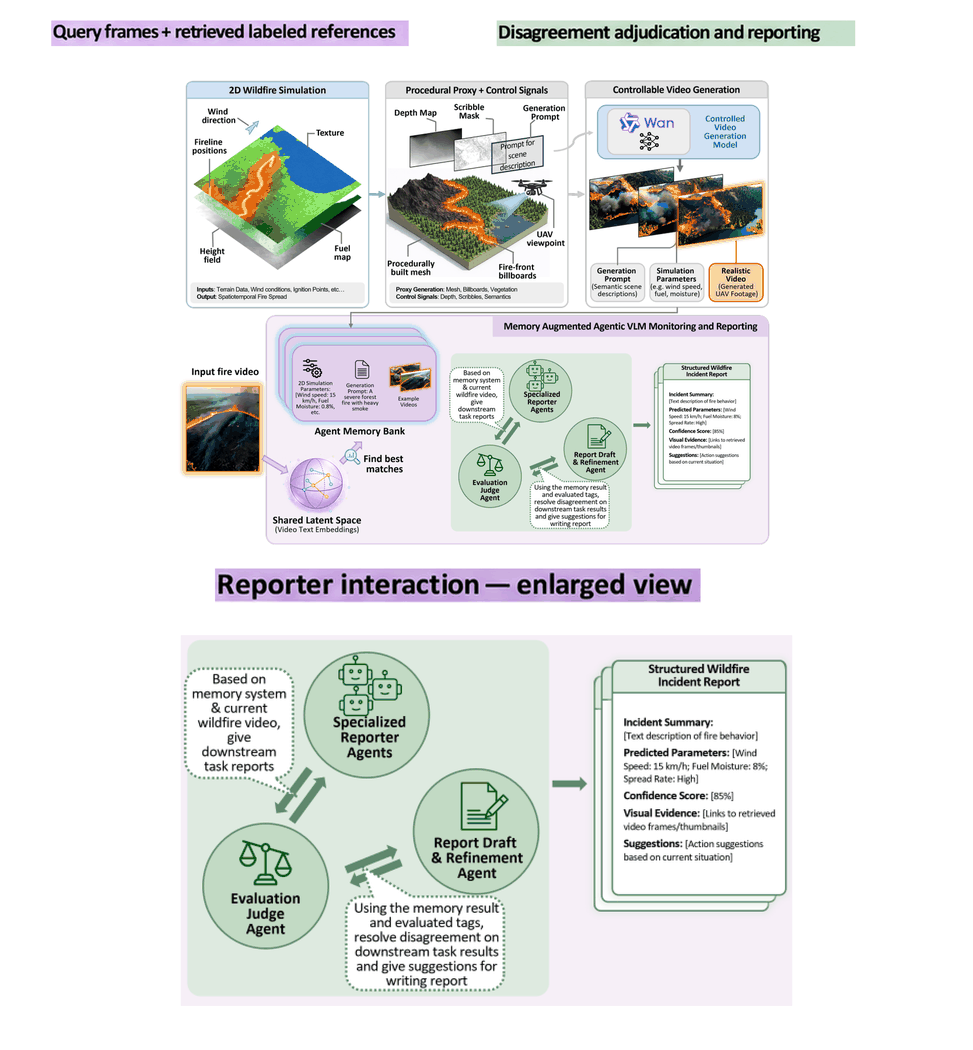}\end{minipage}\par
\vspace{3pt}
{\scriptsize \setlength{\tabcolsep}{1.5pt}\begin{tabular*}{\linewidth}{@{\extracolsep{\fill}}rrrrrrr@{}}
\toprule
C & O & G & A & I & S & Mean \\
3 (+0) & 3 (+0) & 4 (+1) & 3 (-1) & 4 (+0) & 5 (+0) & 3.67 (+0.00)\\
\bottomrule
\end{tabular*}
}
\end{minipage}\par
\vspace{3pt}
{\scriptsize Source public-date proxy: 2026-10-03. All displayed scores are medians of three runs; parentheses show revised minus original.\par}
\vspace{8pt}{\small Only this qualifying figure was included for this paper. A missing M or F figure is not imputed.\par}
\vfill
{\scriptsize C: clarity; O: organization/readability; G: graphical consistency; A: aesthetic coherence; I: visual inventiveness; S: design specificity. Mean is the arithmetic mean of six dimension medians, a descriptive display value, not a validated quality index.\par}

\clearpage
\phantomsection
\label{atlas:corpus164}
\pdfbookmark[1]{164. Agent Skills Under Change: A Lifecycle Benchmark and Coordinate-Guided Updates}{corpus164}
{\Large\bfseries\color{academicblue} 164 \quad \href{https://openreview.net/forum?id=O7Vvd7kb57}{Agent Skills Under Change: A Lifecycle Benchmark and Coordinate-Guided Updates} \citep{corpus164}\par}
\vspace{3pt}
{\small S4-AI-Style \enspace $\vert$ \enspace Agent skill lifecycles \enspace $\vert$ \enspace ICLR 2027 submission (under review)\par}
\vspace{4pt}
\vspace{5pt}
{\bfseries Motivation (source Fig.\ 1)} \hfill {\footnotesize \href{https://openreview.net/pdf?id=O7Vvd7kb57}{Source PDF, p.\ 2}; ID: AI\_O7Vvd7kb57\_M}\par
\vspace{2pt}
\noindent\begin{minipage}[t]{0.486\linewidth}
\centering\textbf{Original}\par\vspace{2pt}
\begin{minipage}[c][0.255\textheight][c]{\linewidth}\centering\includegraphics[width=\linewidth,height=0.255\textheight,keepaspectratio]{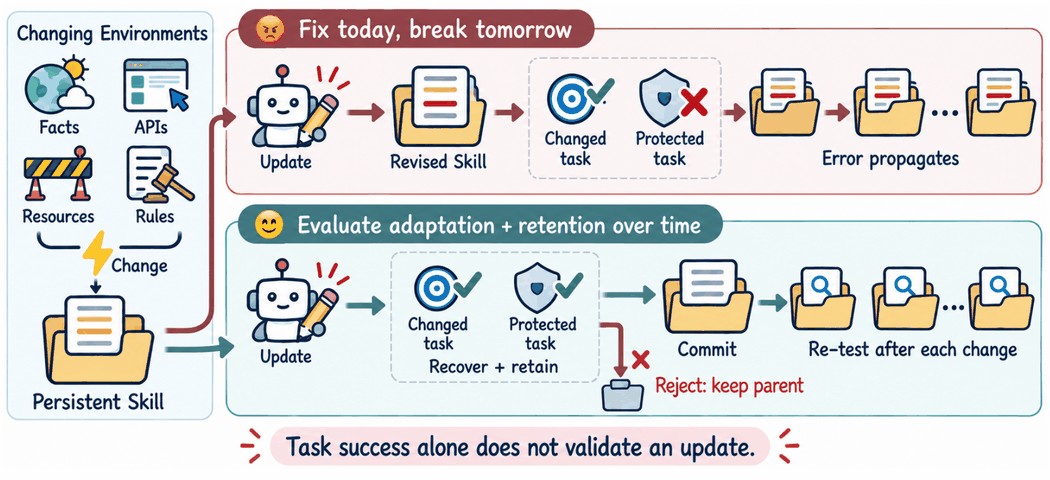}\end{minipage}\par
\vspace{3pt}{\scriptsize \setlength{\tabcolsep}{1.5pt}\begin{tabular*}{\linewidth}{@{\extracolsep{\fill}}rrrrrrr@{}}
\toprule
C & O & G & A & I & S & Mean \\
5 & 5 & 5 & 4 & 4 & 4 & 4.50\\
\bottomrule
\end{tabular*}
}
\end{minipage}\hfill
\begin{minipage}[t]{0.486\linewidth}
\centering\textbf{Study-generated derivative}\par\vspace{2pt}
\begin{minipage}[c][0.255\textheight][c]{\linewidth}\centering\includegraphics[width=\linewidth,height=0.255\textheight,keepaspectratio]{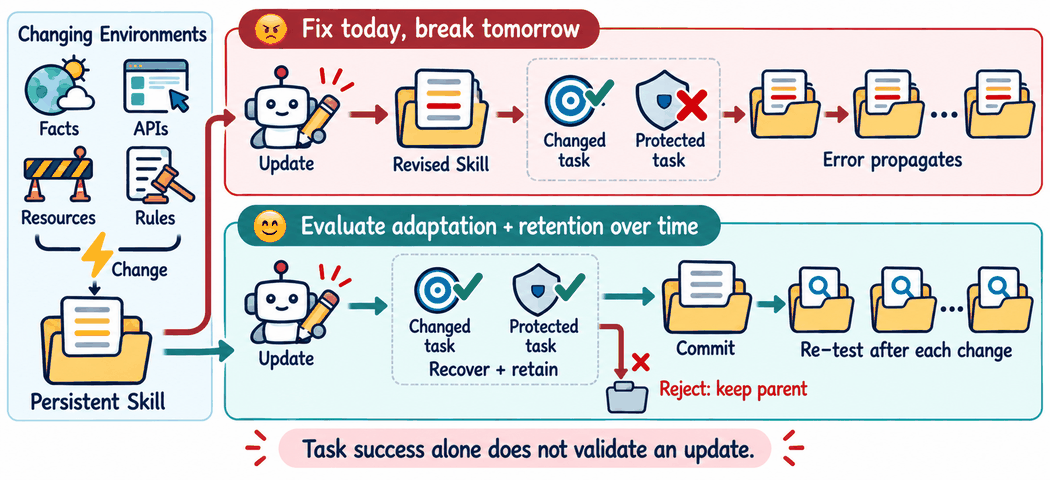}\end{minipage}\par
\vspace{3pt}
{\scriptsize \setlength{\tabcolsep}{1.5pt}\begin{tabular*}{\linewidth}{@{\extracolsep{\fill}}rrrrrrr@{}}
\toprule
C & O & G & A & I & S & Mean \\
5 (+0) & 5 (+0) & 4 (-1) & 4 (+0) & 4 (+0) & 4 (+0) & 4.33 (-0.17)\\
\bottomrule
\end{tabular*}
}
\end{minipage}\par
\vspace{3pt}
{\scriptsize Source public-date proxy: 2026-10-03. All displayed scores are medians of three runs; parentheses show revised minus original.\par}
\vspace{5pt}
{\bfseries Framework (source Fig.\ 2)} \hfill {\footnotesize \href{https://openreview.net/pdf?id=O7Vvd7kb57}{Source PDF, p.\ 3}; ID: AI\_O7Vvd7kb57\_F}\par
\vspace{2pt}
\noindent\begin{minipage}[t]{0.486\linewidth}
\centering\textbf{Original}\par\vspace{2pt}
\begin{minipage}[c][0.255\textheight][c]{\linewidth}\centering\includegraphics[width=\linewidth,height=0.255\textheight,keepaspectratio]{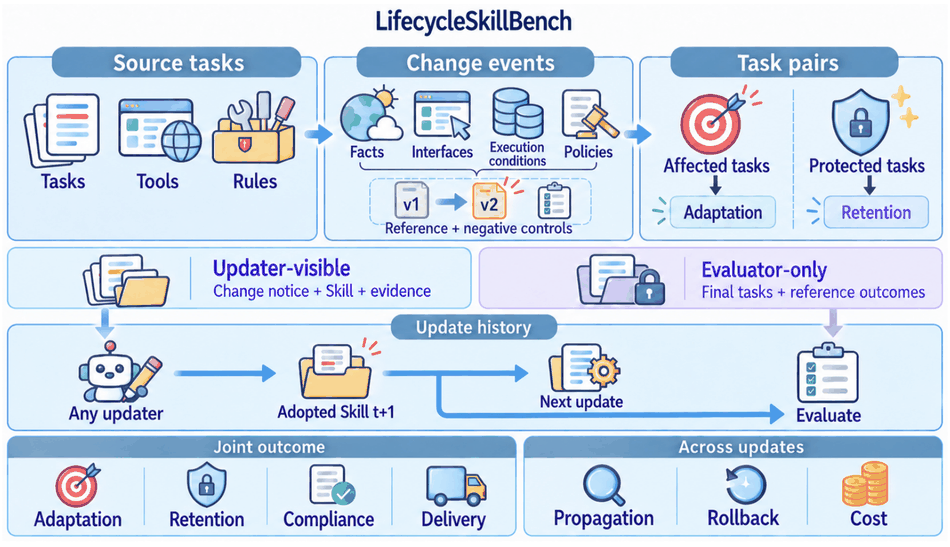}\end{minipage}\par
\vspace{3pt}{\scriptsize \setlength{\tabcolsep}{1.5pt}\begin{tabular*}{\linewidth}{@{\extracolsep{\fill}}rrrrrrr@{}}
\toprule
C & O & G & A & I & S & Mean \\
4 & 4 & 4 & 4 & 3 & 4 & 3.83\\
\bottomrule
\end{tabular*}
}
\end{minipage}\hfill
\begin{minipage}[t]{0.486\linewidth}
\centering\textbf{Study-generated derivative}\par\vspace{2pt}
\begin{minipage}[c][0.255\textheight][c]{\linewidth}\centering\includegraphics[width=\linewidth,height=0.255\textheight,keepaspectratio]{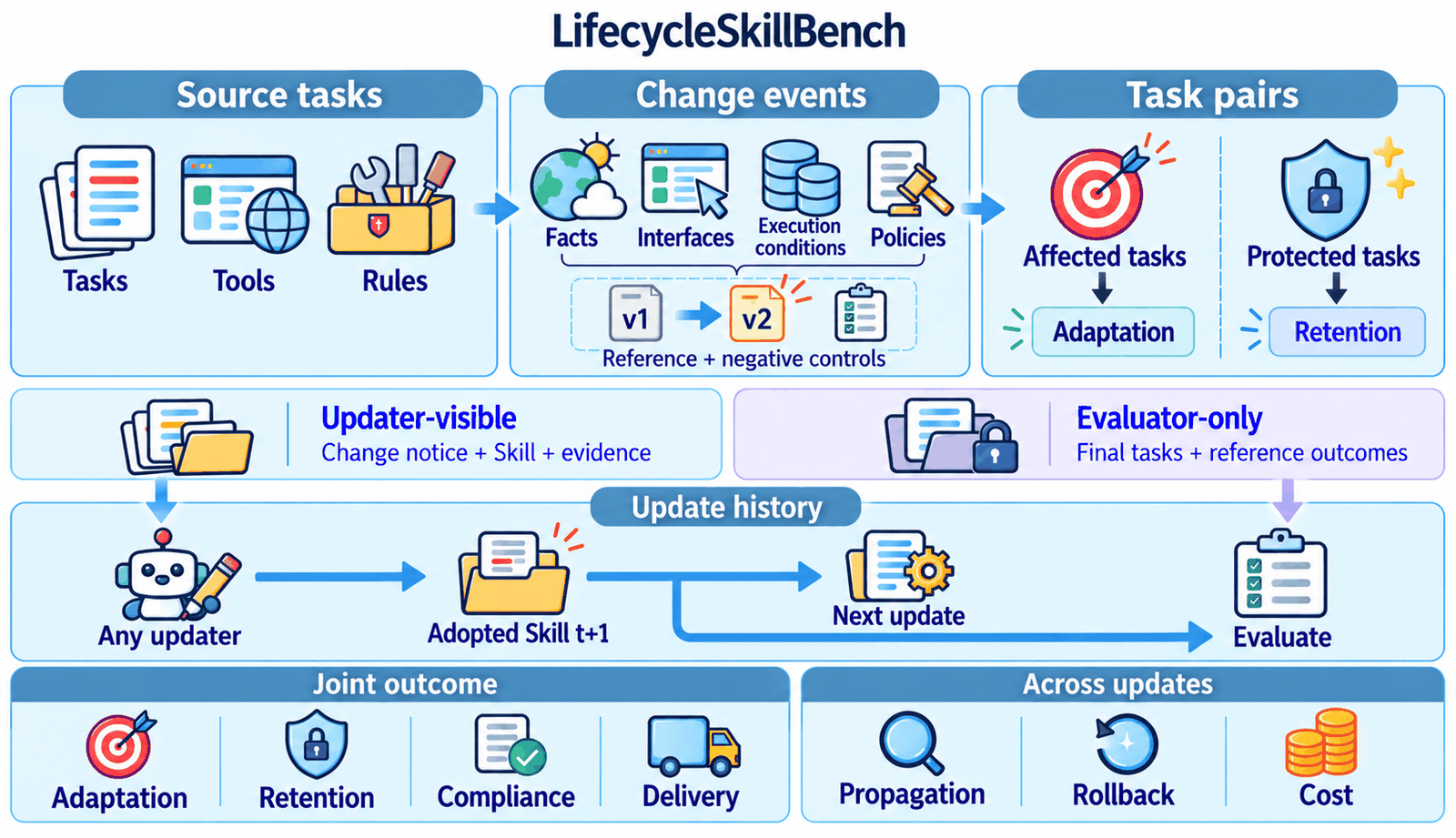}\end{minipage}\par
\vspace{3pt}
{\scriptsize \setlength{\tabcolsep}{1.5pt}\begin{tabular*}{\linewidth}{@{\extracolsep{\fill}}rrrrrrr@{}}
\toprule
C & O & G & A & I & S & Mean \\
4 (+0) & 4 (+0) & 4 (+0) & 4 (+0) & 3 (+0) & 4 (+0) & 3.83 (+0.00)\\
\bottomrule
\end{tabular*}
}
\end{minipage}\par
\vspace{3pt}
{\scriptsize Source public-date proxy: 2026-10-03. All displayed scores are medians of three runs; parentheses show revised minus original.\par}
\vfill
{\scriptsize C: clarity; O: organization/readability; G: graphical consistency; A: aesthetic coherence; I: visual inventiveness; S: design specificity. Mean is the arithmetic mean of six dimension medians, a descriptive display value, not a validated quality index.\par}

\clearpage
\phantomsection
\label{atlas:corpus167}
\pdfbookmark[1]{167. BKBAgent: A Battery Parameter Identification Agent with Bayesian Experimental Design and Knowledge Graph}{corpus167}
{\Large\bfseries\color{academicblue} 167 \quad \href{https://openreview.net/forum?id=whcg806KGZ}{BKBAgent: A Battery Parameter Identification Agent with Bayesian Experimental Design and Knowledge Graph} \citep{corpus167}\par}
\vspace{3pt}
{\small S4-AI-Style \enspace $\vert$ \enspace Battery parameter identification \enspace $\vert$ \enspace ICLR 2027 submission (under review)\par}
\vspace{4pt}
\vspace{5pt}
{\bfseries Motivation (source Fig.\ 1)} \hfill {\footnotesize \href{https://openreview.net/pdf?id=whcg806KGZ}{Source PDF, p.\ 2}; ID: AI\_whcg806KGZ\_M}\par
\vspace{2pt}
\noindent\begin{minipage}[t]{0.486\linewidth}
\centering\textbf{Original}\par\vspace{2pt}
\begin{minipage}[c][0.255\textheight][c]{\linewidth}\centering\includegraphics[width=\linewidth,height=0.255\textheight,keepaspectratio]{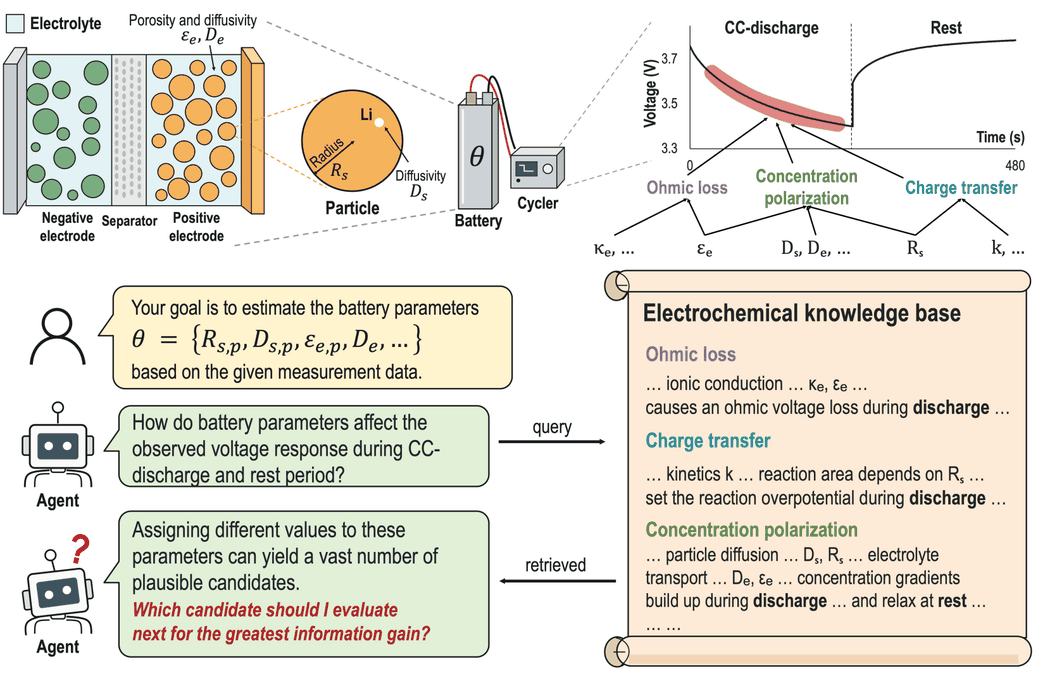}\end{minipage}\par
\vspace{3pt}{\scriptsize \setlength{\tabcolsep}{1.5pt}\begin{tabular*}{\linewidth}{@{\extracolsep{\fill}}rrrrrrr@{}}
\toprule
C & O & G & A & I & S & Mean \\
4 & 3 & 4 & 4 & 4 & 5 & 4.00\\
\bottomrule
\end{tabular*}
}
\end{minipage}\hfill
\begin{minipage}[t]{0.486\linewidth}
\centering\textbf{Study-generated derivative}\par\vspace{2pt}
\begin{minipage}[c][0.255\textheight][c]{\linewidth}\centering\includegraphics[width=\linewidth,height=0.255\textheight,keepaspectratio]{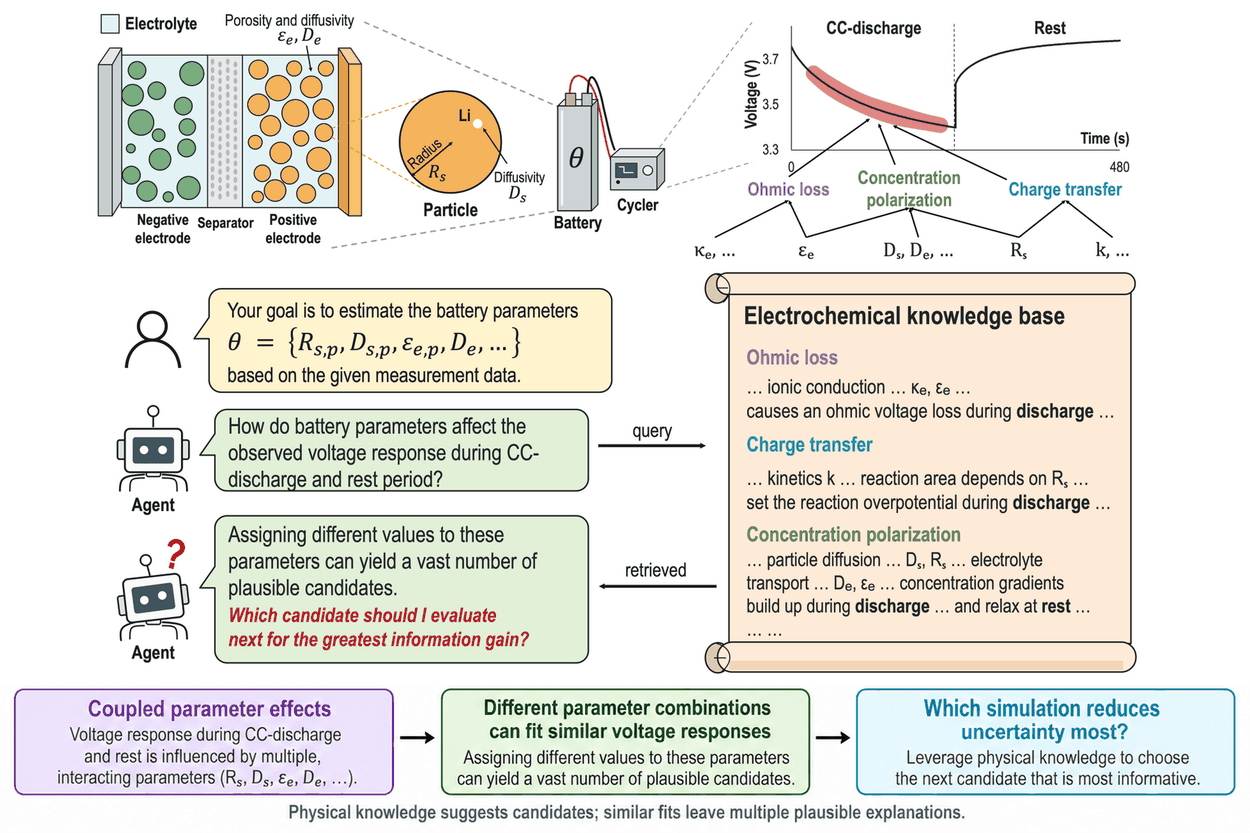}\end{minipage}\par
\vspace{3pt}
{\scriptsize \setlength{\tabcolsep}{1.5pt}\begin{tabular*}{\linewidth}{@{\extracolsep{\fill}}rrrrrrr@{}}
\toprule
C & O & G & A & I & S & Mean \\
4 (+0) & 3 (+0) & 4 (+0) & 3 (-1) & 3 (-1) & 5 (+0) & 3.67 (-0.33)\\
\bottomrule
\end{tabular*}
}
\end{minipage}\par
\vspace{3pt}
{\scriptsize Source public-date proxy: 2026-10-03. All displayed scores are medians of three runs; parentheses show revised minus original.\par}
\vspace{5pt}
{\bfseries Framework (source Fig.\ 2)} \hfill {\footnotesize \href{https://openreview.net/pdf?id=whcg806KGZ}{Source PDF, p.\ 4}; ID: AI\_whcg806KGZ\_F}\par
\vspace{2pt}
\noindent\begin{minipage}[t]{0.486\linewidth}
\centering\textbf{Original}\par\vspace{2pt}
\begin{minipage}[c][0.255\textheight][c]{\linewidth}\centering\includegraphics[width=\linewidth,height=0.255\textheight,keepaspectratio]{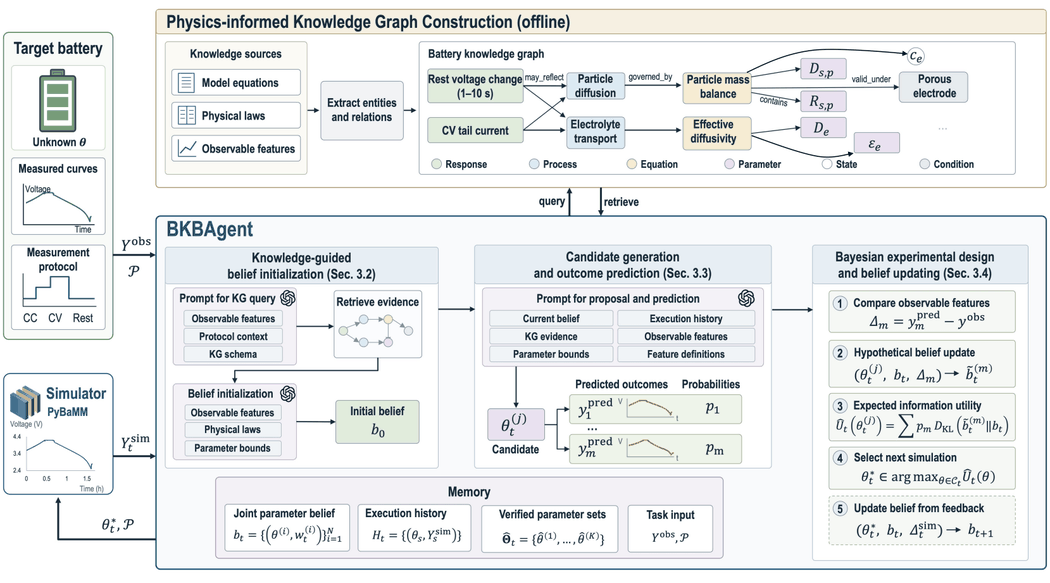}\end{minipage}\par
\vspace{3pt}{\scriptsize \setlength{\tabcolsep}{1.5pt}\begin{tabular*}{\linewidth}{@{\extracolsep{\fill}}rrrrrrr@{}}
\toprule
C & O & G & A & I & S & Mean \\
4 & 4 & 4 & 4 & 3 & 5 & 4.00\\
\bottomrule
\end{tabular*}
}
\end{minipage}\hfill
\begin{minipage}[t]{0.486\linewidth}
\centering\textbf{Study-generated derivative}\par\vspace{2pt}
\begin{minipage}[c][0.255\textheight][c]{\linewidth}\centering\includegraphics[width=\linewidth,height=0.255\textheight,keepaspectratio]{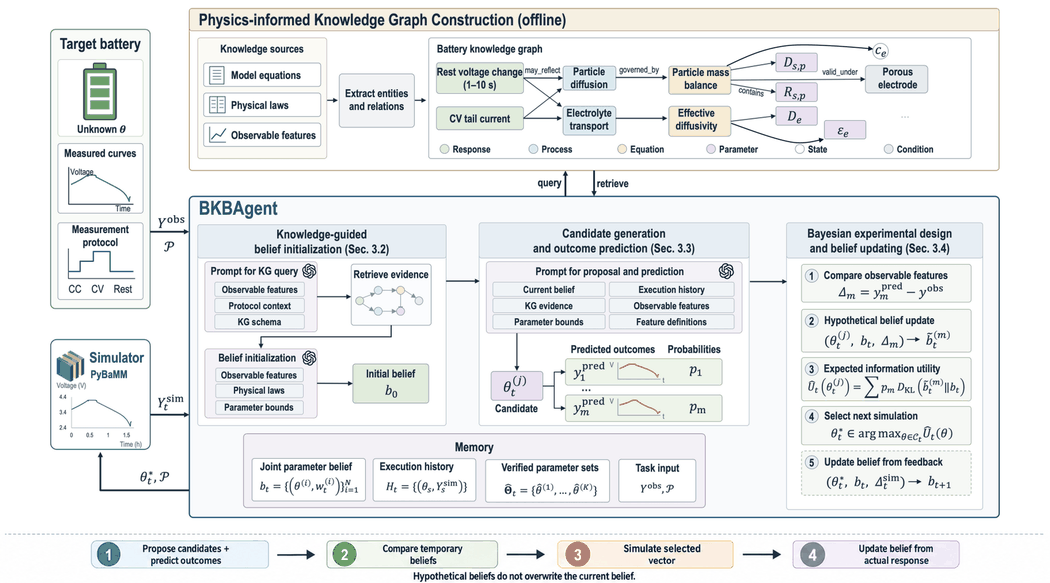}\end{minipage}\par
\vspace{3pt}
{\scriptsize \setlength{\tabcolsep}{1.5pt}\begin{tabular*}{\linewidth}{@{\extracolsep{\fill}}rrrrrrr@{}}
\toprule
C & O & G & A & I & S & Mean \\
4 (+0) & 4 (+0) & 4 (+0) & 4 (+0) & 3 (+0) & 5 (+0) & 4.00 (+0.00)\\
\bottomrule
\end{tabular*}
}
\end{minipage}\par
\vspace{3pt}
{\scriptsize Source public-date proxy: 2026-10-03. All displayed scores are medians of three runs; parentheses show revised minus original.\par}
\vfill
{\scriptsize C: clarity; O: organization/readability; G: graphical consistency; A: aesthetic coherence; I: visual inventiveness; S: design specificity. Mean is the arithmetic mean of six dimension medians, a descriptive display value, not a validated quality index.\par}

\clearpage
\phantomsection
\label{atlas:corpus184}
\pdfbookmark[1]{184. Look Back to Move Forward: Learning to Revisit for Agentic Compositional Image Generation}{corpus184}
{\Large\bfseries\color{academicblue} 184 \quad \href{https://openreview.net/forum?id=C4VFESNXLV}{Look Back to Move Forward: Learning to Revisit for Agentic Compositional Image Generation} \citep{corpus184}\par}
\vspace{3pt}
{\small S4-AI-Style \enspace $\vert$ \enspace Compositional image-generation agents \enspace $\vert$ \enspace ICLR 2027 submission (under review)\par}
\vspace{4pt}
\vspace{5pt}
{\bfseries Motivation (source Fig.\ 1)} \hfill {\footnotesize \href{https://openreview.net/pdf?id=C4VFESNXLV}{Source PDF, p.\ 2}; ID: AI\_C4VFESNXLV\_M}\par
\vspace{2pt}
\noindent\begin{minipage}[t]{0.486\linewidth}
\centering\textbf{Original}\par\vspace{2pt}
\begin{minipage}[c][0.255\textheight][c]{\linewidth}\centering\includegraphics[width=\linewidth,height=0.255\textheight,keepaspectratio]{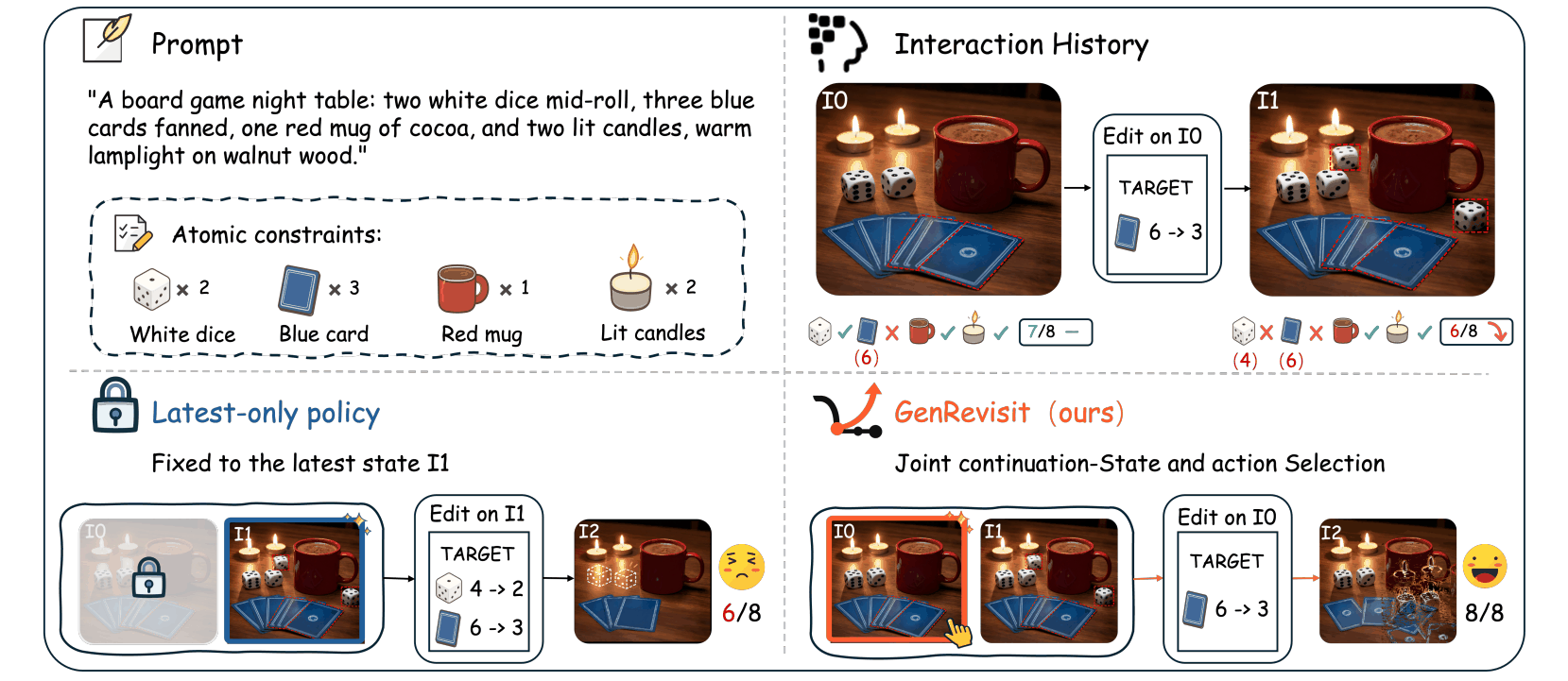}\end{minipage}\par
\vspace{3pt}{\scriptsize \setlength{\tabcolsep}{1.5pt}\begin{tabular*}{\linewidth}{@{\extracolsep{\fill}}rrrrrrr@{}}
\toprule
C & O & G & A & I & S & Mean \\
4 & 4 & 4 & 3 & 4 & 5 & 4.00\\
\bottomrule
\end{tabular*}
}
\end{minipage}\hfill
\begin{minipage}[t]{0.486\linewidth}
\centering\textbf{Study-generated derivative}\par\vspace{2pt}
\begin{minipage}[c][0.255\textheight][c]{\linewidth}\centering\includegraphics[width=\linewidth,height=0.255\textheight,keepaspectratio]{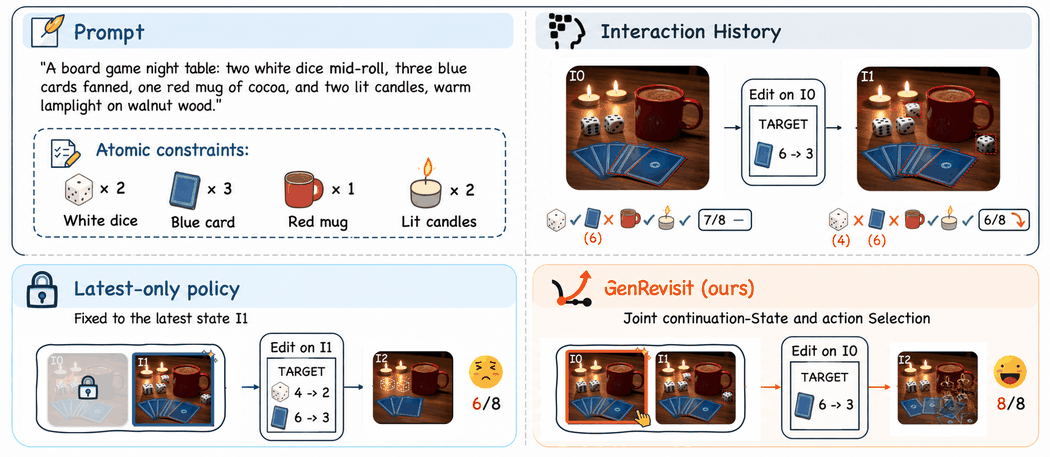}\end{minipage}\par
\vspace{3pt}
{\scriptsize \setlength{\tabcolsep}{1.5pt}\begin{tabular*}{\linewidth}{@{\extracolsep{\fill}}rrrrrrr@{}}
\toprule
C & O & G & A & I & S & Mean \\
4 (+0) & 4 (+0) & 4 (+0) & 4 (+1) & 4 (+0) & 5 (+0) & 4.17 (+0.17)\\
\bottomrule
\end{tabular*}
}
\end{minipage}\par
\vspace{3pt}
{\scriptsize Source public-date proxy: 2026-10-03. All displayed scores are medians of three runs; parentheses show revised minus original.\par}
\vspace{5pt}
{\bfseries Framework (source Fig.\ 2)} \hfill {\footnotesize \href{https://openreview.net/pdf?id=C4VFESNXLV}{Source PDF, p.\ 4}; ID: AI\_C4VFESNXLV\_F}\par
\vspace{2pt}
\noindent\begin{minipage}[t]{0.486\linewidth}
\centering\textbf{Original}\par\vspace{2pt}
\begin{minipage}[c][0.255\textheight][c]{\linewidth}\centering\includegraphics[width=\linewidth,height=0.255\textheight,keepaspectratio]{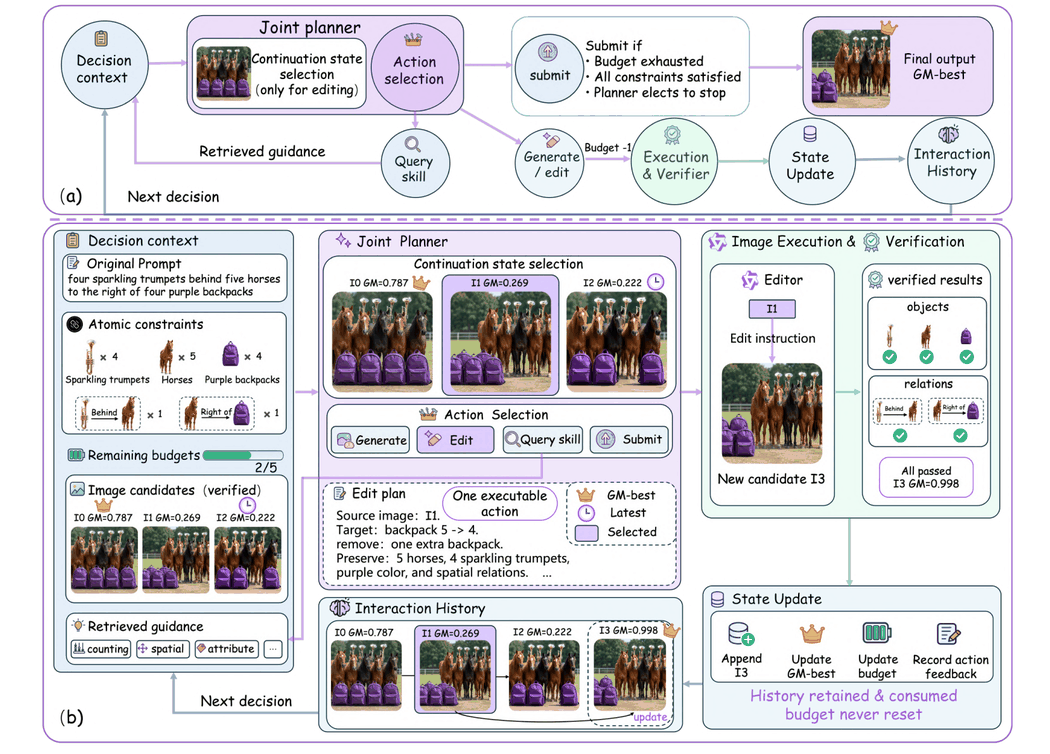}\end{minipage}\par
\vspace{3pt}{\scriptsize \setlength{\tabcolsep}{1.5pt}\begin{tabular*}{\linewidth}{@{\extracolsep{\fill}}rrrrrrr@{}}
\toprule
C & O & G & A & I & S & Mean \\
4 & 3 & 4 & 4 & 4 & 5 & 4.00\\
\bottomrule
\end{tabular*}
}
\end{minipage}\hfill
\begin{minipage}[t]{0.486\linewidth}
\centering\textbf{Study-generated derivative}\par\vspace{2pt}
\begin{minipage}[c][0.255\textheight][c]{\linewidth}\centering\includegraphics[width=\linewidth,height=0.255\textheight,keepaspectratio]{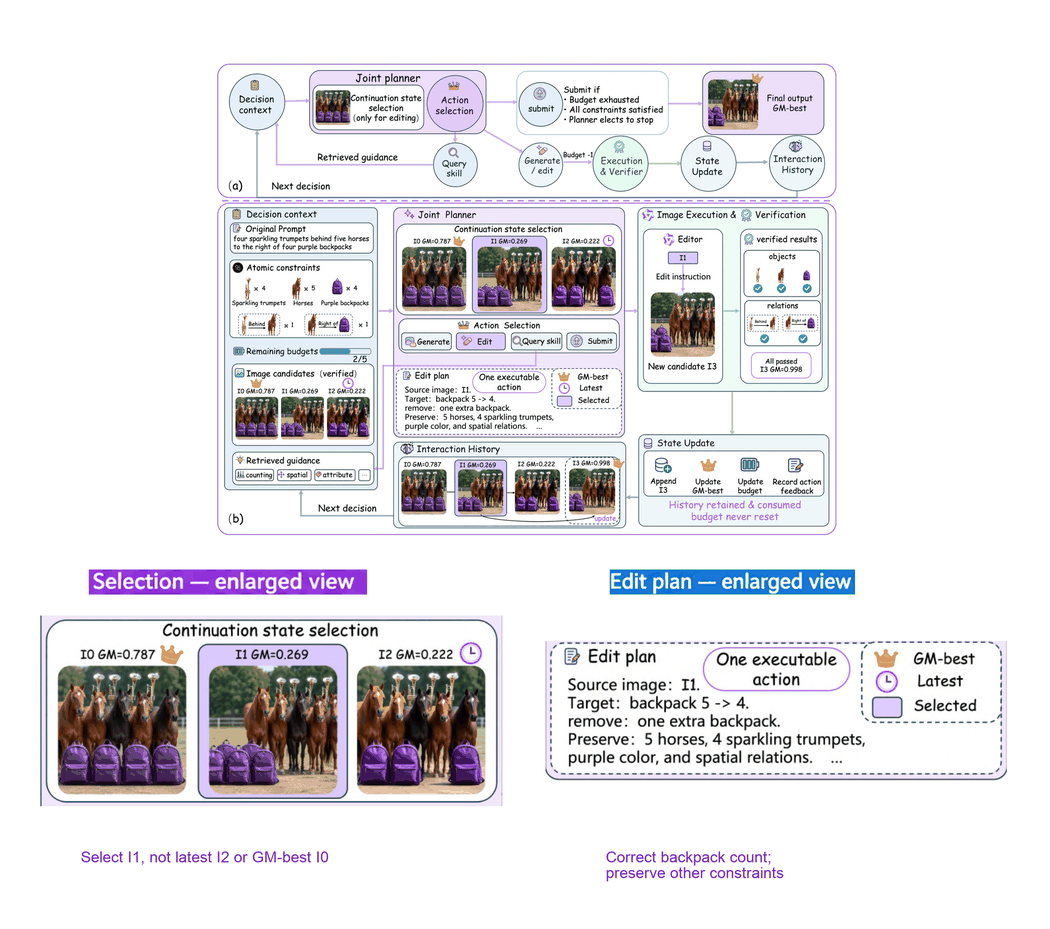}\end{minipage}\par
\vspace{3pt}
{\scriptsize \setlength{\tabcolsep}{1.5pt}\begin{tabular*}{\linewidth}{@{\extracolsep{\fill}}rrrrrrr@{}}
\toprule
C & O & G & A & I & S & Mean \\
4 (+0) & 3 (+0) & 4 (+0) & 3 (-1) & 4 (+0) & 5 (+0) & 3.83 (-0.17)\\
\bottomrule
\end{tabular*}
}
\end{minipage}\par
\vspace{3pt}
{\scriptsize Source public-date proxy: 2026-10-03. All displayed scores are medians of three runs; parentheses show revised minus original.\par}
\vfill
{\scriptsize C: clarity; O: organization/readability; G: graphical consistency; A: aesthetic coherence; I: visual inventiveness; S: design specificity. Mean is the arithmetic mean of six dimension medians, a descriptive display value, not a validated quality index.\par}

\clearpage
\phantomsection
\label{atlas:corpus190}
\pdfbookmark[1]{190. Ontology-Harnessing Multi-Agent System for Abstractive Visual Understanding}{corpus190}
{\Large\bfseries\color{academicblue} 190 \quad \href{https://openreview.net/forum?id=jVkkfpmnAw}{Ontology-Harnessing Multi-Agent System for Abstractive Visual Understanding} \citep{corpus190}\par}
\vspace{3pt}
{\small S4-AI-Style \enspace $\vert$ \enspace Abstract visual reasoning \enspace $\vert$ \enspace ICLR 2027 submission (under review)\par}
\vspace{4pt}
\vspace{5pt}
{\bfseries Motivation (source Fig.\ 1)} \hfill {\footnotesize \href{https://openreview.net/pdf?id=jVkkfpmnAw}{Source PDF, p.\ 2}; ID: AI\_jVkkfpmnAw\_M}\par
\vspace{2pt}
\noindent\begin{minipage}[t]{0.486\linewidth}
\centering\textbf{Original}\par\vspace{2pt}
\begin{minipage}[c][0.255\textheight][c]{\linewidth}\centering\includegraphics[width=\linewidth,height=0.255\textheight,keepaspectratio]{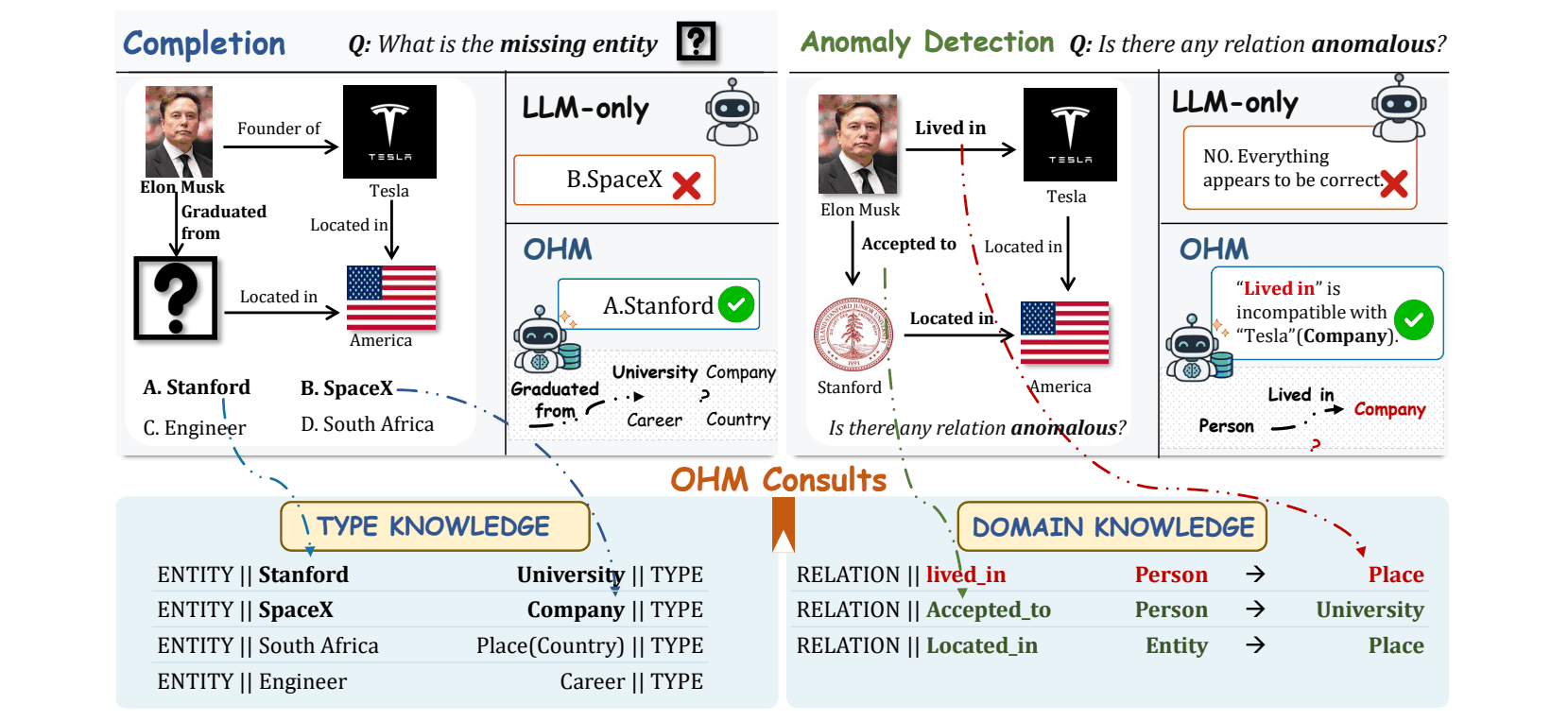}\end{minipage}\par
\vspace{3pt}{\scriptsize \setlength{\tabcolsep}{1.5pt}\begin{tabular*}{\linewidth}{@{\extracolsep{\fill}}rrrrrrr@{}}
\toprule
C & O & G & A & I & S & Mean \\
4 & 3 & 4 & 3 & 4 & 5 & 3.83\\
\bottomrule
\end{tabular*}
}
\end{minipage}\hfill
\begin{minipage}[t]{0.486\linewidth}
\centering\textbf{Study-generated derivative}\par\vspace{2pt}
\begin{minipage}[c][0.255\textheight][c]{\linewidth}\centering\includegraphics[width=\linewidth,height=0.255\textheight,keepaspectratio]{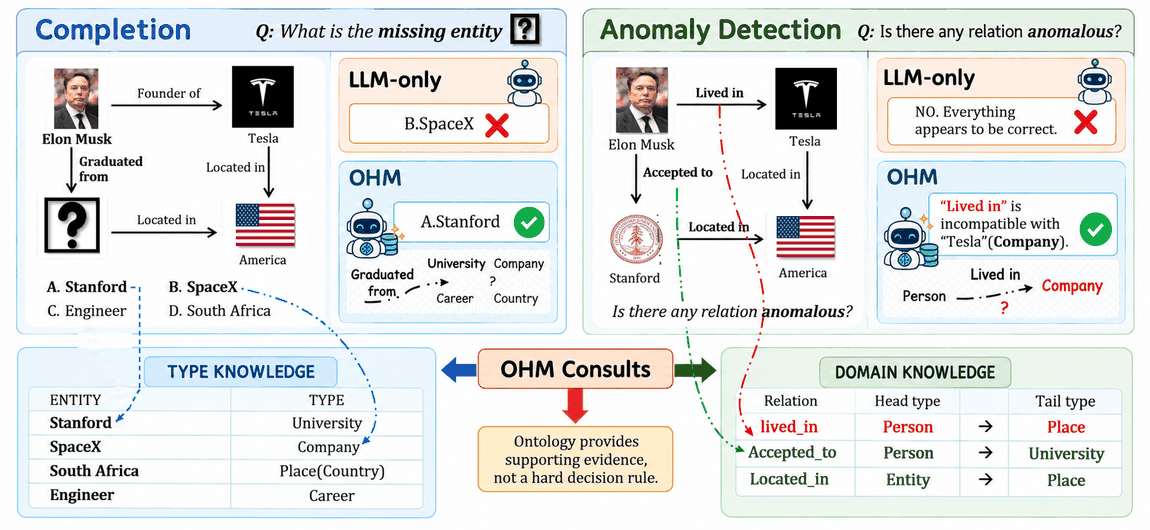}\end{minipage}\par
\vspace{3pt}
{\scriptsize \setlength{\tabcolsep}{1.5pt}\begin{tabular*}{\linewidth}{@{\extracolsep{\fill}}rrrrrrr@{}}
\toprule
C & O & G & A & I & S & Mean \\
4 (+0) & 3 (+0) & 4 (+0) & 3 (+0) & 4 (+0) & 5 (+0) & 3.83 (+0.00)\\
\bottomrule
\end{tabular*}
}
\end{minipage}\par
\vspace{3pt}
{\scriptsize Source public-date proxy: 2026-10-03. All displayed scores are medians of three runs; parentheses show revised minus original.\par}
\vspace{5pt}
{\bfseries Framework (source Fig.\ 2)} \hfill {\footnotesize \href{https://openreview.net/pdf?id=jVkkfpmnAw}{Source PDF, p.\ 4}; ID: AI\_jVkkfpmnAw\_F}\par
\vspace{2pt}
\noindent\begin{minipage}[t]{0.486\linewidth}
\centering\textbf{Original}\par\vspace{2pt}
\begin{minipage}[c][0.255\textheight][c]{\linewidth}\centering\includegraphics[width=\linewidth,height=0.255\textheight,keepaspectratio]{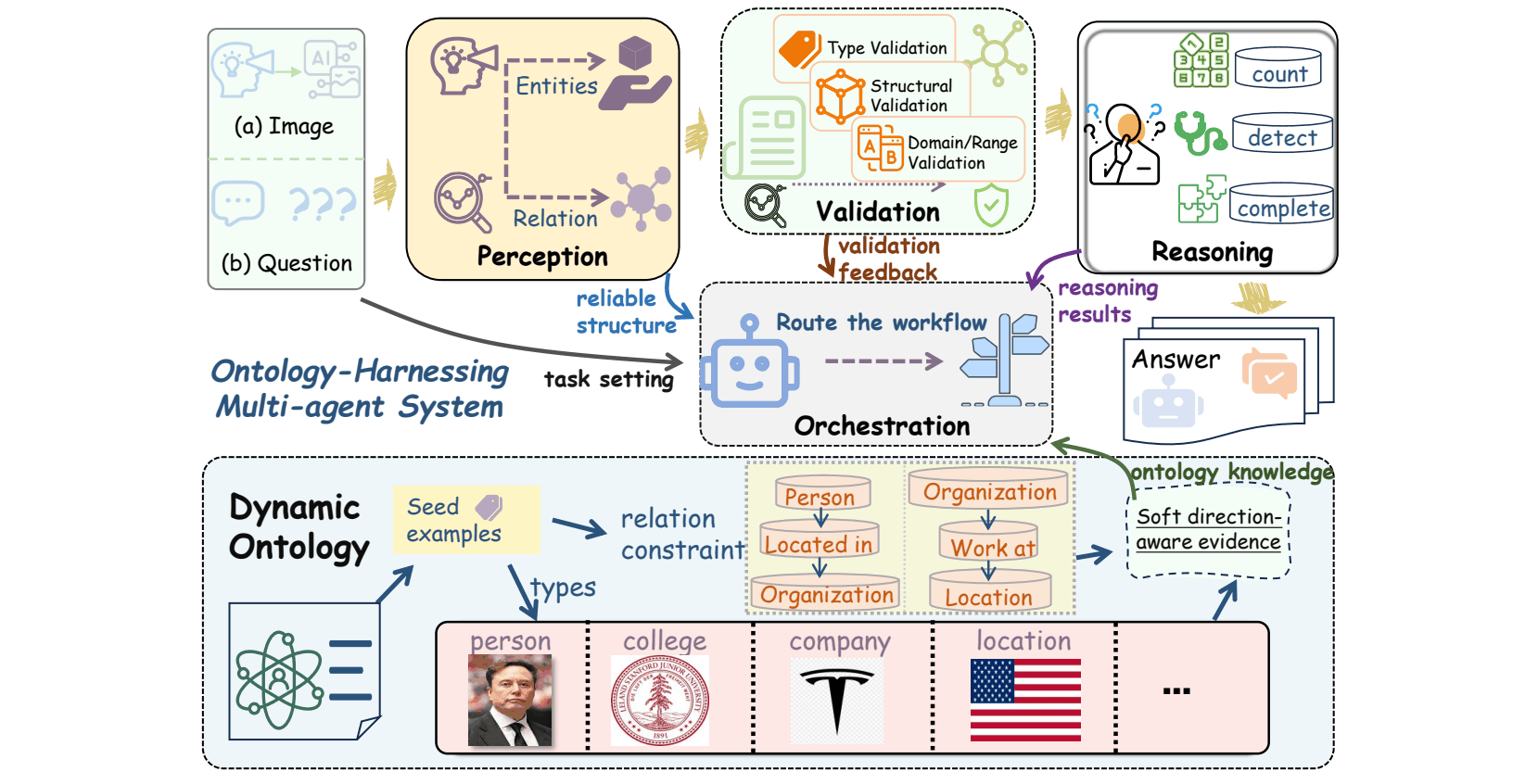}\end{minipage}\par
\vspace{3pt}{\scriptsize \setlength{\tabcolsep}{1.5pt}\begin{tabular*}{\linewidth}{@{\extracolsep{\fill}}rrrrrrr@{}}
\toprule
C & O & G & A & I & S & Mean \\
3 & 3 & 3 & 3 & 3 & 4 & 3.17\\
\bottomrule
\end{tabular*}
}
\end{minipage}\hfill
\begin{minipage}[t]{0.486\linewidth}
\centering\textbf{Study-generated derivative}\par\vspace{2pt}
\begin{minipage}[c][0.255\textheight][c]{\linewidth}\centering\includegraphics[width=\linewidth,height=0.255\textheight,keepaspectratio]{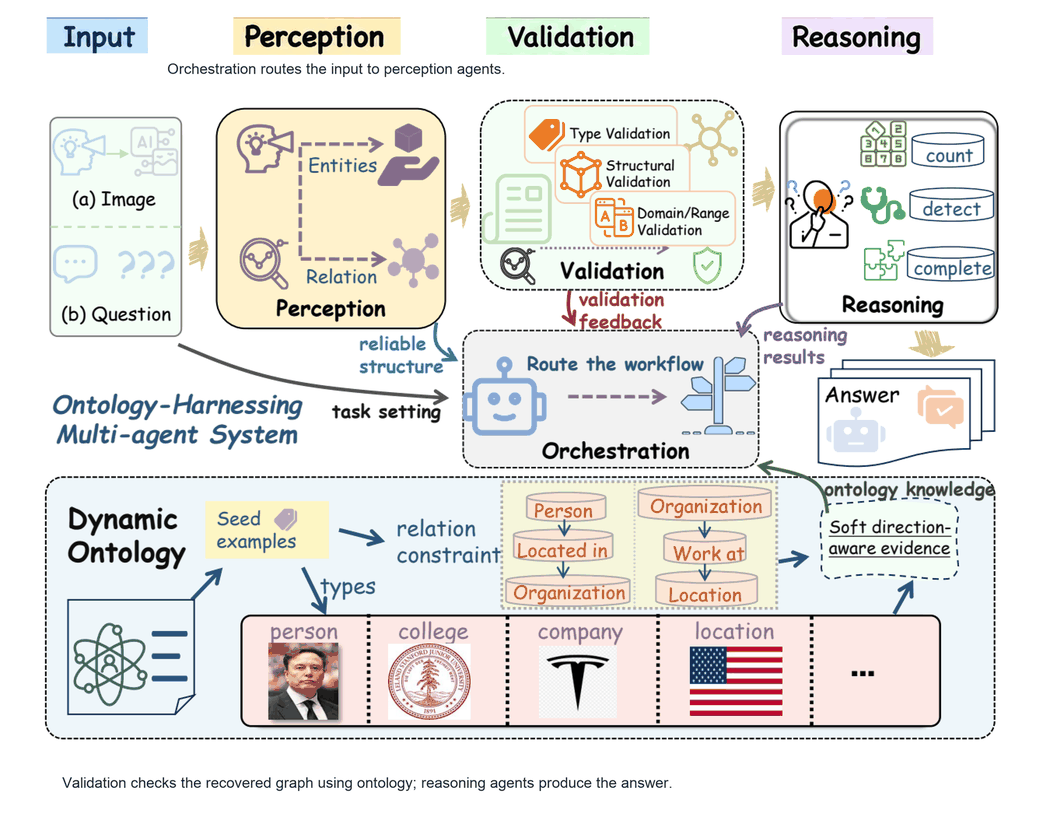}\end{minipage}\par
\vspace{3pt}
{\scriptsize \setlength{\tabcolsep}{1.5pt}\begin{tabular*}{\linewidth}{@{\extracolsep{\fill}}rrrrrrr@{}}
\toprule
C & O & G & A & I & S & Mean \\
3 (+0) & 3 (+0) & 3 (+0) & 3 (+0) & 3 (+0) & 4 (+0) & 3.17 (+0.00)\\
\bottomrule
\end{tabular*}
}
\end{minipage}\par
\vspace{3pt}
{\scriptsize Source public-date proxy: 2026-10-03. All displayed scores are medians of three runs; parentheses show revised minus original.\par}
\vfill
{\scriptsize C: clarity; O: organization/readability; G: graphical consistency; A: aesthetic coherence; I: visual inventiveness; S: design specificity. Mean is the arithmetic mean of six dimension medians, a descriptive display value, not a validated quality index.\par}

\clearpage
\phantomsection
\label{atlas:corpus191}
\pdfbookmark[1]{191. OpenAgentFlow: Enabling System-Wide Safety Boundaries for Heterogeneous AI Agent Fleets}{corpus191}
{\Large\bfseries\color{academicblue} 191 \quad \href{https://openreview.net/forum?id=I6Gx5US2sY}{OpenAgentFlow: Enabling System-Wide Safety Boundaries for Heterogeneous AI Agent Fleets} \citep{corpus191}\par}
\vspace{3pt}
{\small S4-AI-Style \enspace $\vert$ \enspace Agent action governance and safety \enspace $\vert$ \enspace ICLR 2027 submission (under review)\par}
\vspace{4pt}
\vspace{5pt}
{\bfseries Motivation (source Fig.\ 1)} \hfill {\footnotesize \href{https://openreview.net/pdf?id=I6Gx5US2sY}{Source PDF, p.\ 2}; ID: AI\_I6Gx5US2sY\_M}\par
\vspace{2pt}
\noindent\begin{minipage}[t]{0.486\linewidth}
\centering\textbf{Original}\par\vspace{2pt}
\begin{minipage}[c][0.255\textheight][c]{\linewidth}\centering\includegraphics[width=\linewidth,height=0.255\textheight,keepaspectratio]{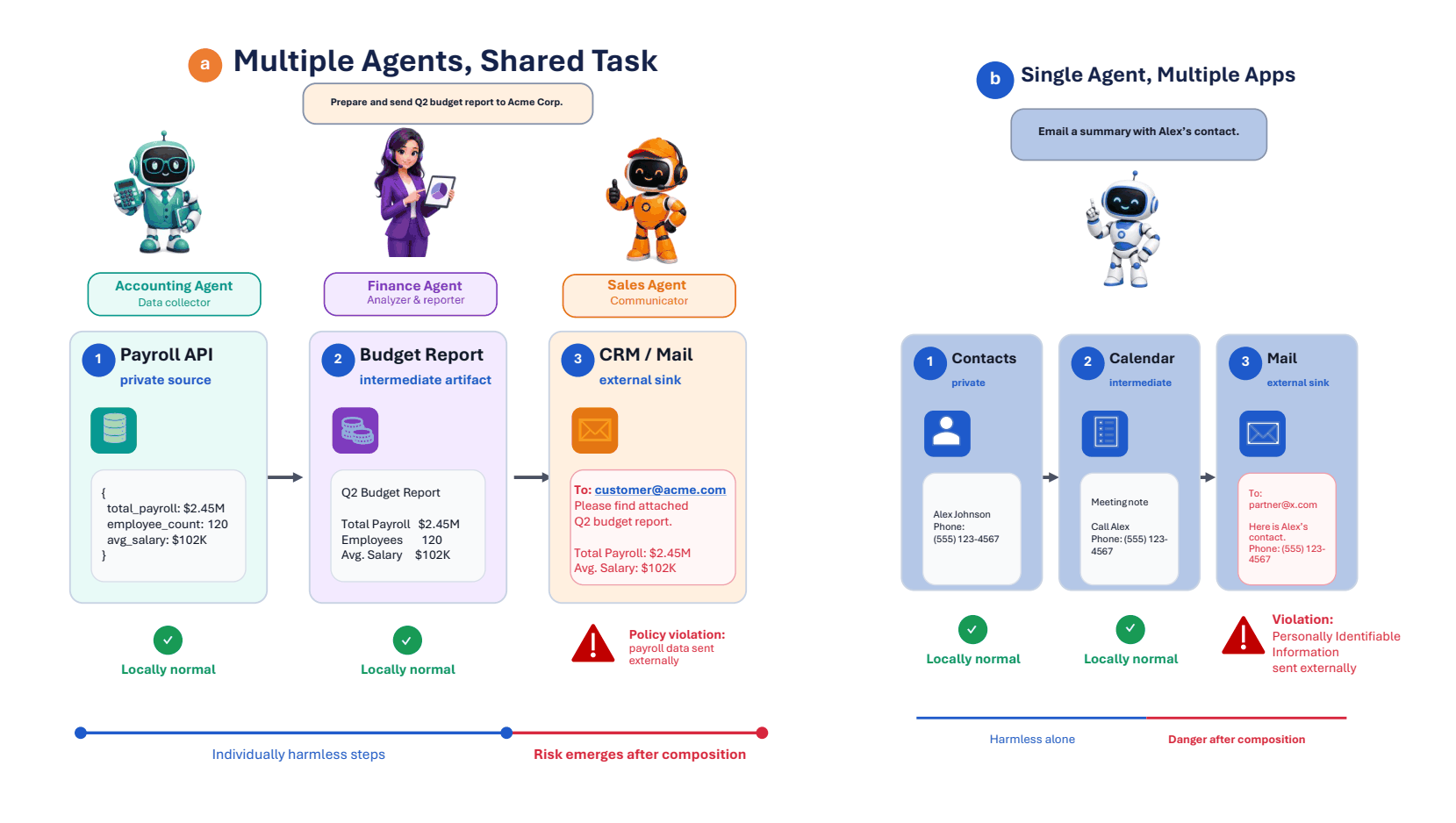}\end{minipage}\par
\vspace{3pt}{\scriptsize \setlength{\tabcolsep}{1.5pt}\begin{tabular*}{\linewidth}{@{\extracolsep{\fill}}rrrrrrr@{}}
\toprule
C & O & G & A & I & S & Mean \\
4 & 4 & 5 & 3 & 3 & 4 & 3.83\\
\bottomrule
\end{tabular*}
}
\end{minipage}\hfill
\begin{minipage}[t]{0.486\linewidth}
\centering\textbf{Study-generated derivative}\par\vspace{2pt}
\begin{minipage}[c][0.255\textheight][c]{\linewidth}\centering\includegraphics[width=\linewidth,height=0.255\textheight,keepaspectratio]{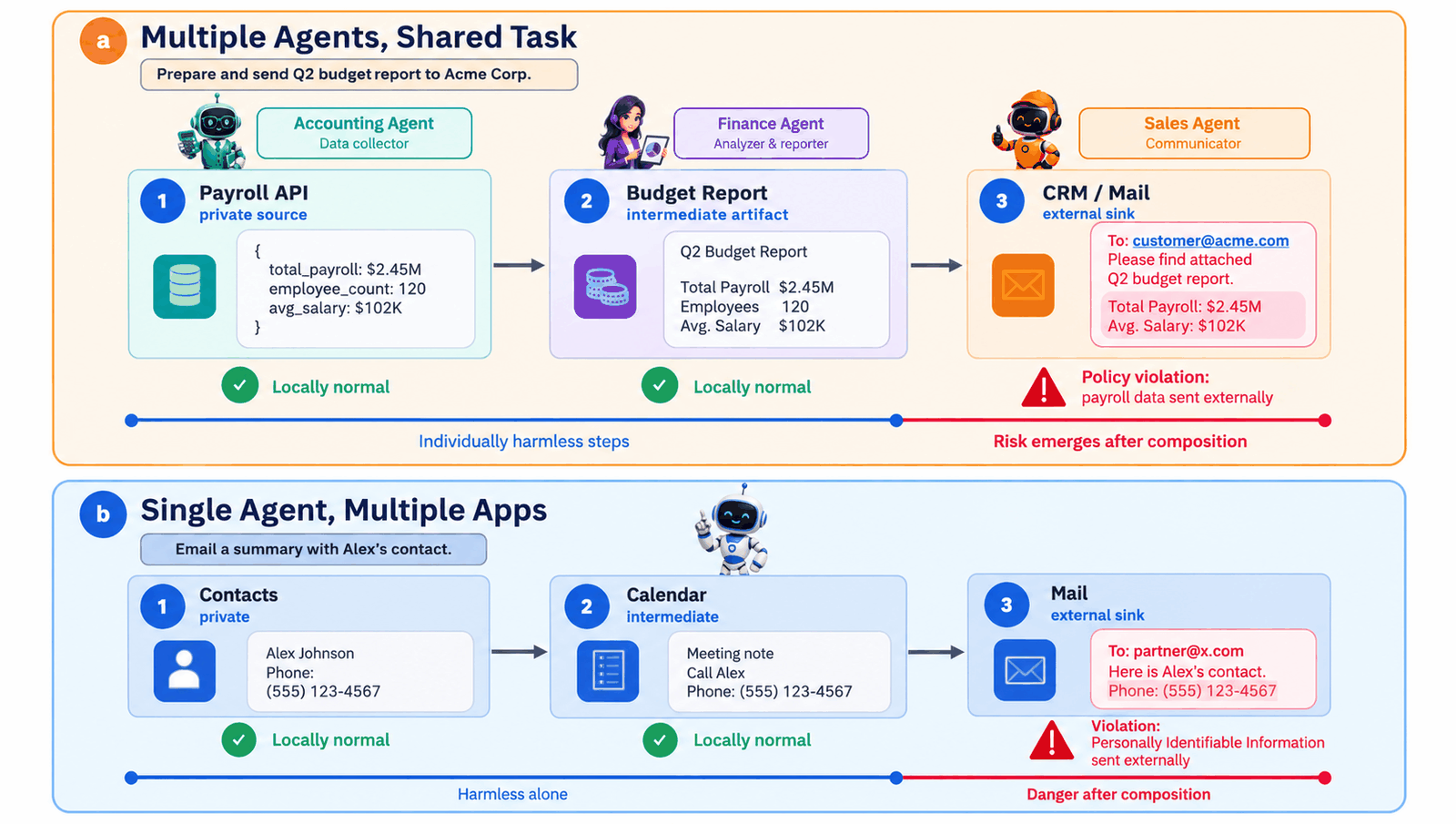}\end{minipage}\par
\vspace{3pt}
{\scriptsize \setlength{\tabcolsep}{1.5pt}\begin{tabular*}{\linewidth}{@{\extracolsep{\fill}}rrrrrrr@{}}
\toprule
C & O & G & A & I & S & Mean \\
5 (+1) & 5 (+1) & 5 (+0) & 4 (+1) & 4 (+1) & 4 (+0) & 4.50 (+0.67)\\
\bottomrule
\end{tabular*}
}
\end{minipage}\par
\vspace{3pt}
{\scriptsize Source public-date proxy: 2026-10-03. All displayed scores are medians of three runs; parentheses show revised minus original.\par}
\vspace{5pt}
{\bfseries Framework (source Fig.\ 2)} \hfill {\footnotesize \href{https://openreview.net/pdf?id=I6Gx5US2sY}{Source PDF, p.\ 6}; ID: AI\_I6Gx5US2sY\_F}\par
\vspace{2pt}
\noindent\begin{minipage}[t]{0.486\linewidth}
\centering\textbf{Original}\par\vspace{2pt}
\begin{minipage}[c][0.255\textheight][c]{\linewidth}\centering\includegraphics[width=\linewidth,height=0.255\textheight,keepaspectratio]{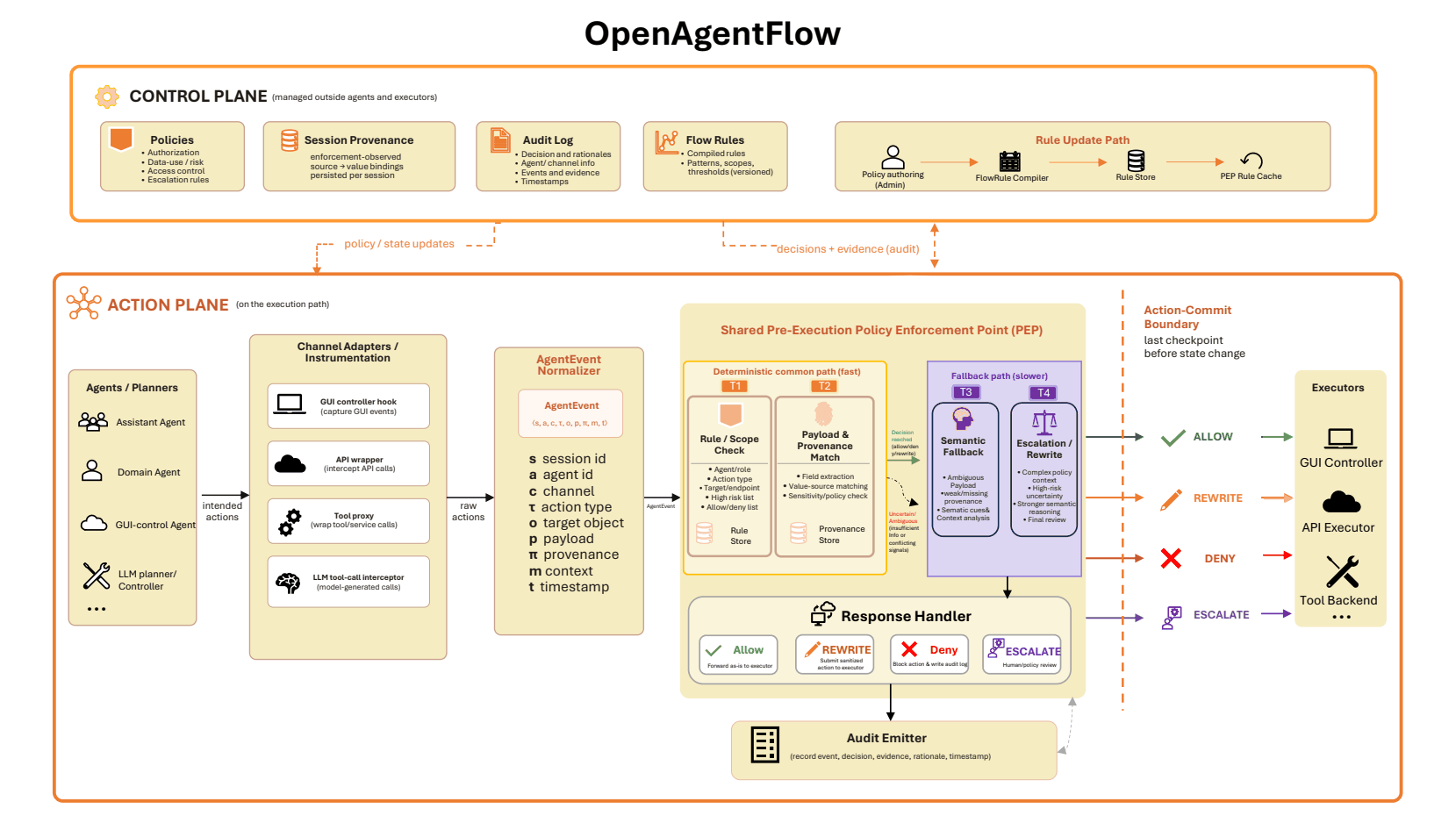}\end{minipage}\par
\vspace{3pt}{\scriptsize \setlength{\tabcolsep}{1.5pt}\begin{tabular*}{\linewidth}{@{\extracolsep{\fill}}rrrrrrr@{}}
\toprule
C & O & G & A & I & S & Mean \\
3 & 3 & 3 & 4 & 4 & 4 & 3.50\\
\bottomrule
\end{tabular*}
}
\end{minipage}\hfill
\begin{minipage}[t]{0.486\linewidth}
\centering\textbf{Study-generated derivative}\par\vspace{2pt}
\begin{minipage}[c][0.255\textheight][c]{\linewidth}\centering\includegraphics[width=\linewidth,height=0.255\textheight,keepaspectratio]{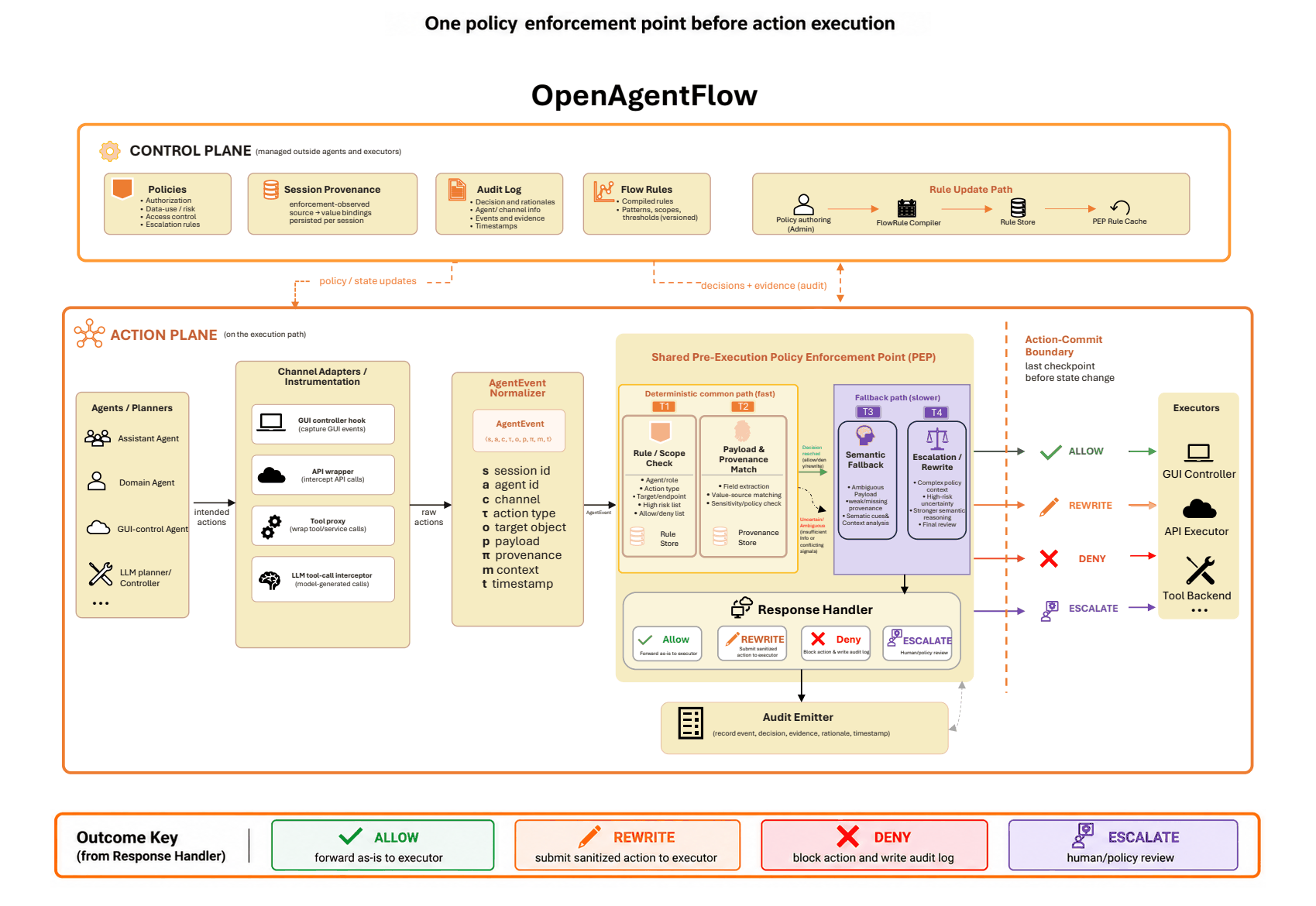}\end{minipage}\par
\vspace{3pt}
{\scriptsize \setlength{\tabcolsep}{1.5pt}\begin{tabular*}{\linewidth}{@{\extracolsep{\fill}}rrrrrrr@{}}
\toprule
C & O & G & A & I & S & Mean \\
3 (+0) & 3 (+0) & 3 (+0) & 3 (-1) & 3 (-1) & 4 (+0) & 3.17 (-0.33)\\
\bottomrule
\end{tabular*}
}
\end{minipage}\par
\vspace{3pt}
{\scriptsize Source public-date proxy: 2026-10-03. All displayed scores are medians of three runs; parentheses show revised minus original.\par}
\vfill
{\scriptsize C: clarity; O: organization/readability; G: graphical consistency; A: aesthetic coherence; I: visual inventiveness; S: design specificity. Mean is the arithmetic mean of six dimension medians, a descriptive display value, not a validated quality index.\par}

\clearpage
\phantomsection
\label{atlas:corpus194}
\pdfbookmark[1]{194. Sprout: Building Dynamic Memory While Reasoning for Agentic Video Understanding}{corpus194}
{\Large\bfseries\color{academicblue} 194 \quad \href{https://openreview.net/forum?id=VA0yCrShah}{Sprout: Building Dynamic Memory While Reasoning for Agentic Video Understanding} \citep{corpus194}\par}
\vspace{3pt}
{\small S4-AI-Style \enspace $\vert$ \enspace Video understanding and memory \enspace $\vert$ \enspace ICLR 2027 submission (under review)\par}
\vspace{4pt}
\vspace{5pt}
{\bfseries Motivation (source Fig.\ 1)} \hfill {\footnotesize \href{https://openreview.net/pdf?id=VA0yCrShah}{Source PDF, p.\ 2}; ID: AI\_VA0yCrShah\_M}\par
\vspace{2pt}
\noindent\begin{minipage}[t]{0.486\linewidth}
\centering\textbf{Original}\par\vspace{2pt}
\begin{minipage}[c][0.255\textheight][c]{\linewidth}\centering\includegraphics[width=\linewidth,height=0.255\textheight,keepaspectratio]{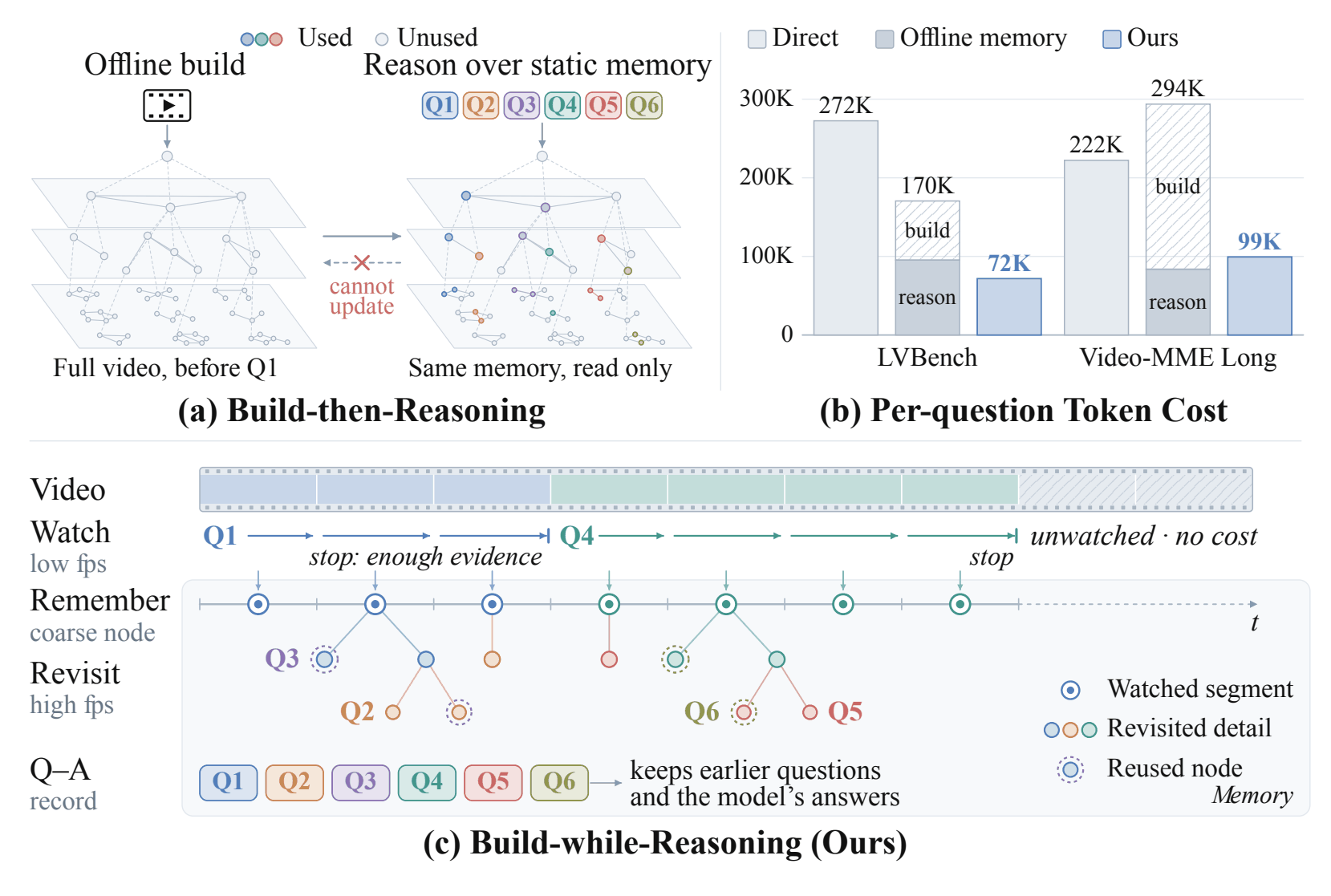}\end{minipage}\par
\vspace{3pt}{\scriptsize \setlength{\tabcolsep}{1.5pt}\begin{tabular*}{\linewidth}{@{\extracolsep{\fill}}rrrrrrr@{}}
\toprule
C & O & G & A & I & S & Mean \\
5 & 4 & 4 & 4 & 5 & 5 & 4.50\\
\bottomrule
\end{tabular*}
}
\end{minipage}\hfill
\begin{minipage}[t]{0.486\linewidth}
\centering\textbf{Study-generated derivative}\par\vspace{2pt}
\begin{minipage}[c][0.255\textheight][c]{\linewidth}\centering\includegraphics[width=\linewidth,height=0.255\textheight,keepaspectratio]{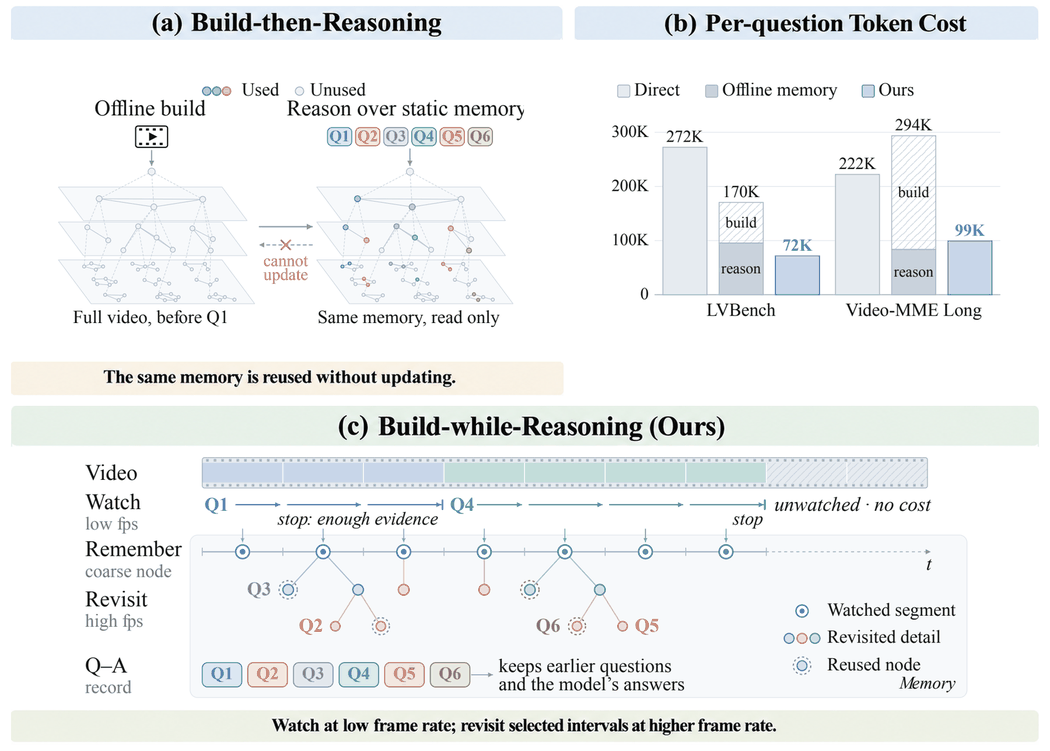}\end{minipage}\par
\vspace{3pt}
{\scriptsize \setlength{\tabcolsep}{1.5pt}\begin{tabular*}{\linewidth}{@{\extracolsep{\fill}}rrrrrrr@{}}
\toprule
C & O & G & A & I & S & Mean \\
5 (+0) & 4 (+0) & 4 (+0) & 4 (+0) & 5 (+0) & 5 (+0) & 4.50 (+0.00)\\
\bottomrule
\end{tabular*}
}
\end{minipage}\par
\vspace{3pt}
{\scriptsize Source public-date proxy: 2026-10-03. All displayed scores are medians of three runs; parentheses show revised minus original.\par}
\vspace{5pt}
{\bfseries Framework (source Fig.\ 2)} \hfill {\footnotesize \href{https://openreview.net/pdf?id=VA0yCrShah}{Source PDF, p.\ 5}; ID: AI\_VA0yCrShah\_F}\par
\vspace{2pt}
\noindent\begin{minipage}[t]{0.486\linewidth}
\centering\textbf{Original}\par\vspace{2pt}
\begin{minipage}[c][0.255\textheight][c]{\linewidth}\centering\includegraphics[width=\linewidth,height=0.255\textheight,keepaspectratio]{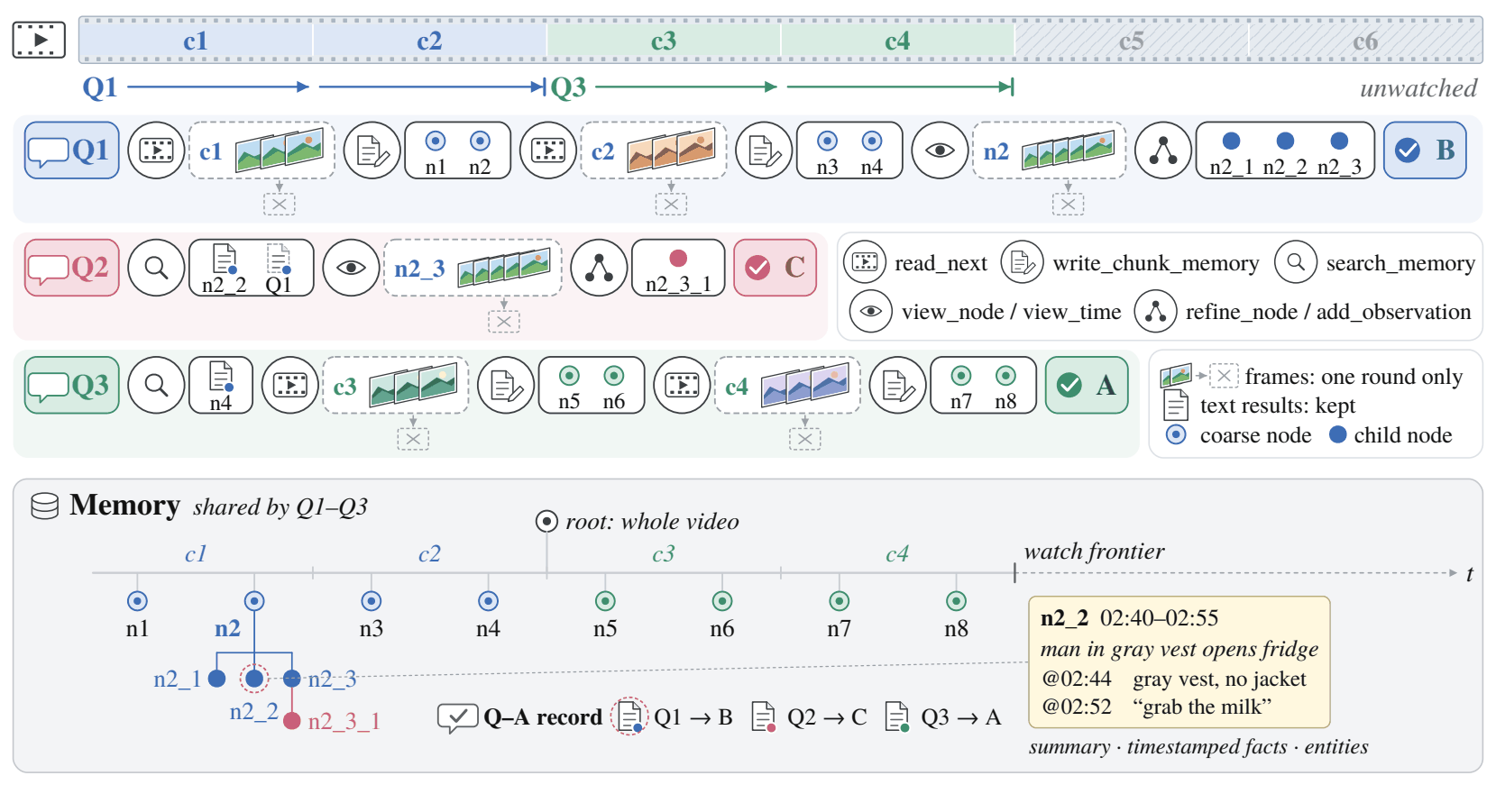}\end{minipage}\par
\vspace{3pt}{\scriptsize \setlength{\tabcolsep}{1.5pt}\begin{tabular*}{\linewidth}{@{\extracolsep{\fill}}rrrrrrr@{}}
\toprule
C & O & G & A & I & S & Mean \\
4 & 4 & 5 & 4 & 5 & 5 & 4.50\\
\bottomrule
\end{tabular*}
}
\end{minipage}\hfill
\begin{minipage}[t]{0.486\linewidth}
\centering\textbf{Study-generated derivative}\par\vspace{2pt}
\begin{minipage}[c][0.255\textheight][c]{\linewidth}\centering\includegraphics[width=\linewidth,height=0.255\textheight,keepaspectratio]{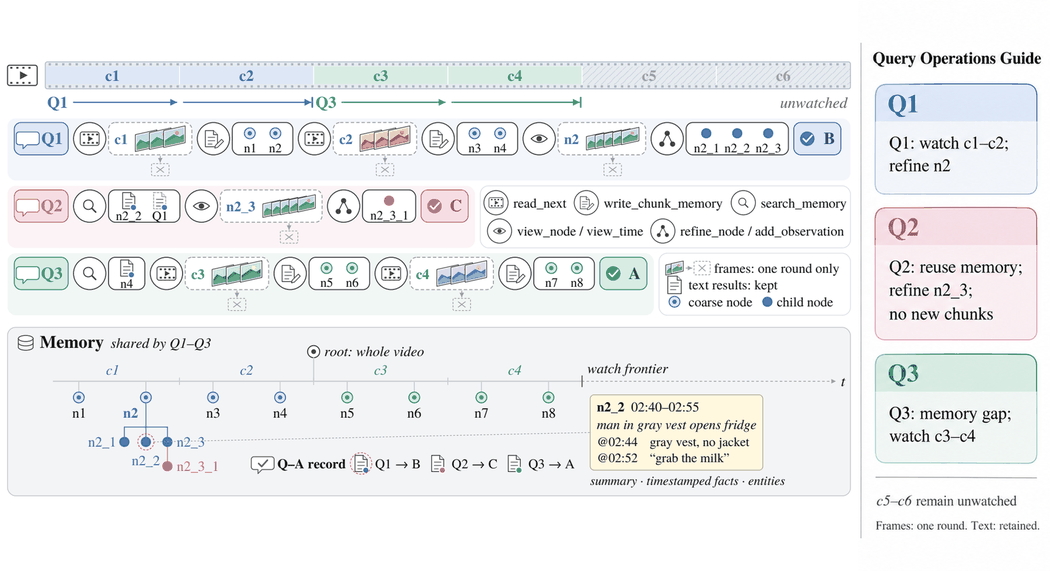}\end{minipage}\par
\vspace{3pt}
{\scriptsize \setlength{\tabcolsep}{1.5pt}\begin{tabular*}{\linewidth}{@{\extracolsep{\fill}}rrrrrrr@{}}
\toprule
C & O & G & A & I & S & Mean \\
4 (+0) & 4 (+0) & 4 (-1) & 4 (+0) & 5 (+0) & 5 (+0) & 4.33 (-0.17)\\
\bottomrule
\end{tabular*}
}
\end{minipage}\par
\vspace{3pt}
{\scriptsize Source public-date proxy: 2026-10-03. All displayed scores are medians of three runs; parentheses show revised minus original.\par}
\vfill
{\scriptsize C: clarity; O: organization/readability; G: graphical consistency; A: aesthetic coherence; I: visual inventiveness; S: design specificity. Mean is the arithmetic mean of six dimension medians, a descriptive display value, not a validated quality index.\par}

\clearpage
\phantomsection
\label{atlas:corpus198}
\pdfbookmark[1]{198. VEvo-Harness: Causally-Guarded Self-Evolving Harness for Agentic Image Generation}{corpus198}
{\Large\bfseries\color{academicblue} 198 \quad \href{https://openreview.net/forum?id=UPFHZRl39C}{VEvo-Harness: Causally-Guarded Self-Evolving Harness for Agentic Image Generation} \citep{corpus198}\par}
\vspace{3pt}
{\small S4-AI-Style \enspace $\vert$ \enspace Self-evolving image-generation agents \enspace $\vert$ \enspace ICLR 2027 submission (under review)\par}
\vspace{4pt}
\vspace{5pt}
{\bfseries Motivation (source Fig.\ 1)} \hfill {\footnotesize \href{https://openreview.net/pdf?id=UPFHZRl39C}{Source PDF, p.\ 2}; ID: AI\_UPFHZRl39C\_M}\par
\vspace{2pt}
\noindent\begin{minipage}[t]{0.486\linewidth}
\centering\textbf{Original}\par\vspace{2pt}
\begin{minipage}[c][0.255\textheight][c]{\linewidth}\centering\includegraphics[width=\linewidth,height=0.255\textheight,keepaspectratio]{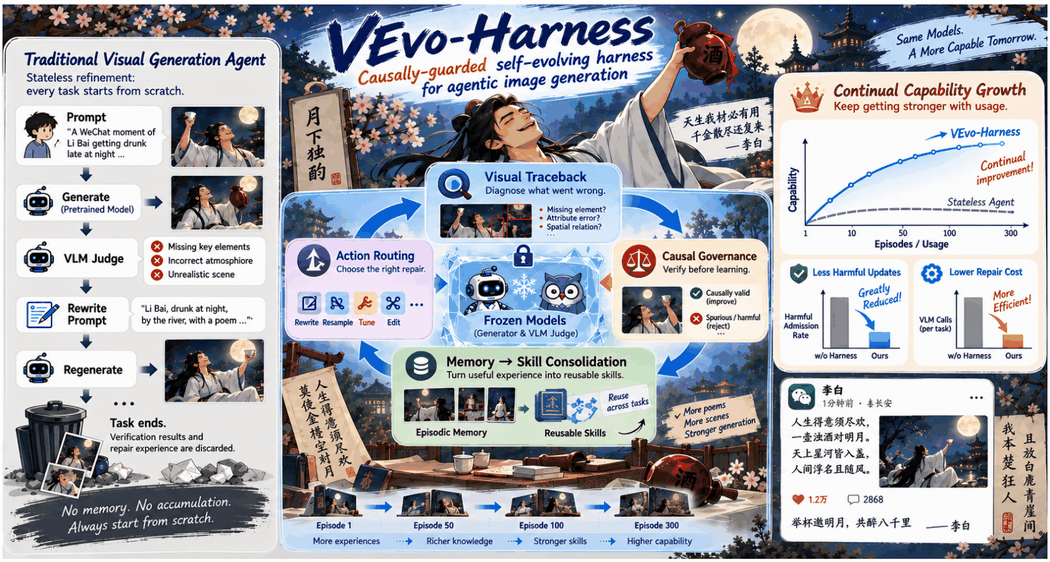}\end{minipage}\par
\vspace{3pt}{\scriptsize \setlength{\tabcolsep}{1.5pt}\begin{tabular*}{\linewidth}{@{\extracolsep{\fill}}rrrrrrr@{}}
\toprule
C & O & G & A & I & S & Mean \\
4 & 3 & 4 & 3 & 4 & 4 & 3.67\\
\bottomrule
\end{tabular*}
}
\end{minipage}\hfill
\begin{minipage}[t]{0.486\linewidth}
\centering\textbf{Study-generated derivative}\par\vspace{2pt}
\begin{minipage}[c][0.255\textheight][c]{\linewidth}\centering\includegraphics[width=\linewidth,height=0.255\textheight,keepaspectratio]{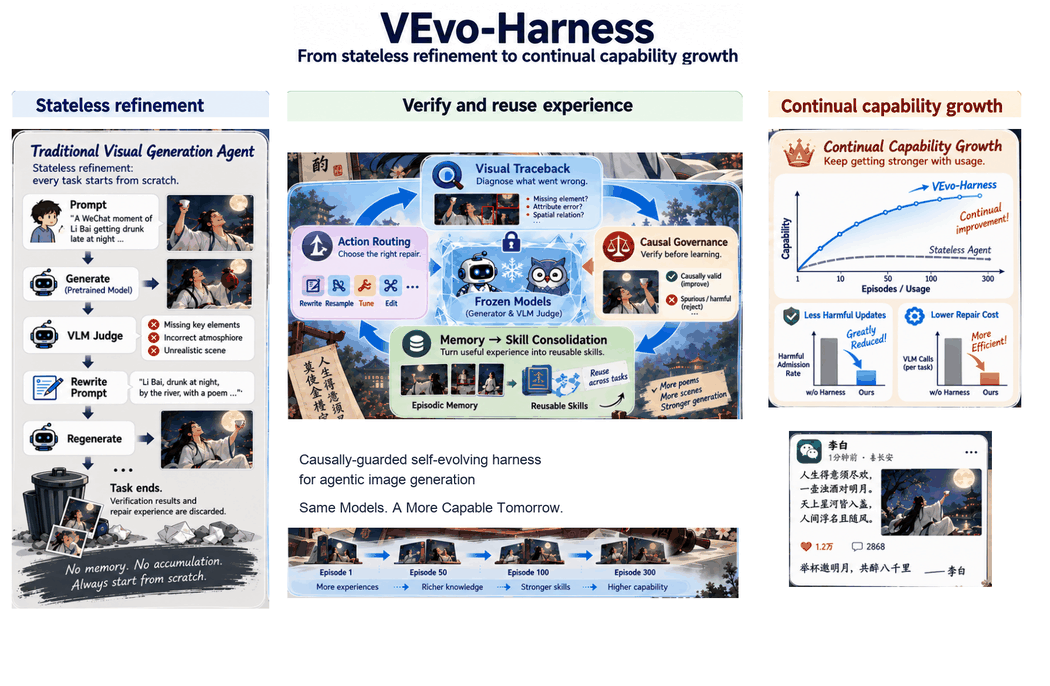}\end{minipage}\par
\vspace{3pt}
{\scriptsize \setlength{\tabcolsep}{1.5pt}\begin{tabular*}{\linewidth}{@{\extracolsep{\fill}}rrrrrrr@{}}
\toprule
C & O & G & A & I & S & Mean \\
5 (+1) & 3 (+0) & 4 (+0) & 3 (+0) & 4 (+0) & 5 (+1) & 4.00 (+0.33)\\
\bottomrule
\end{tabular*}
}
\end{minipage}\par
\vspace{3pt}
{\scriptsize Source public-date proxy: 2026-10-03. All displayed scores are medians of three runs; parentheses show revised minus original.\par}
\vspace{5pt}
{\bfseries Framework (source Fig.\ 2)} \hfill {\footnotesize \href{https://openreview.net/pdf?id=UPFHZRl39C}{Source PDF, p.\ 3}; ID: AI\_UPFHZRl39C\_F}\par
\vspace{2pt}
\noindent\begin{minipage}[t]{0.486\linewidth}
\centering\textbf{Original}\par\vspace{2pt}
\begin{minipage}[c][0.255\textheight][c]{\linewidth}\centering\includegraphics[width=\linewidth,height=0.255\textheight,keepaspectratio]{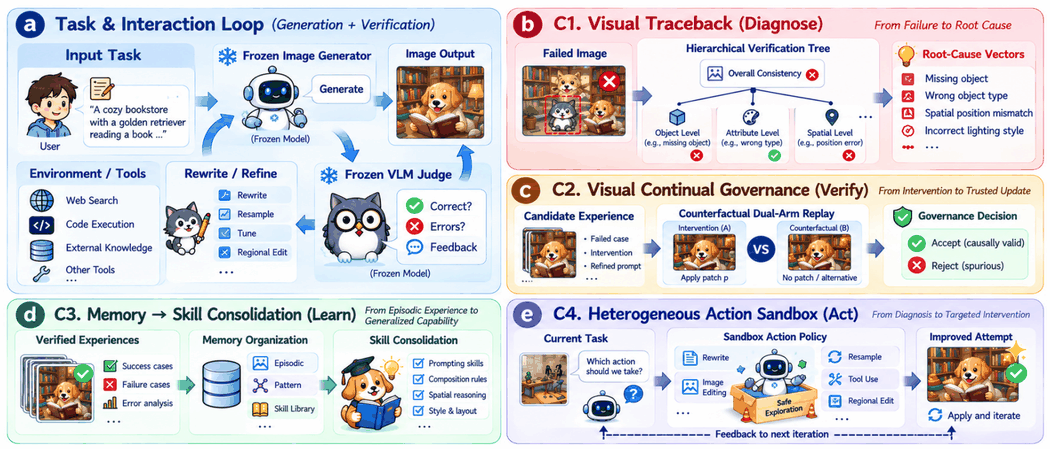}\end{minipage}\par
\vspace{3pt}{\scriptsize \setlength{\tabcolsep}{1.5pt}\begin{tabular*}{\linewidth}{@{\extracolsep{\fill}}rrrrrrr@{}}
\toprule
C & O & G & A & I & S & Mean \\
3 & 4 & 3 & 4 & 3 & 4 & 3.50\\
\bottomrule
\end{tabular*}
}
\end{minipage}\hfill
\begin{minipage}[t]{0.486\linewidth}
\centering\textbf{Study-generated derivative}\par\vspace{2pt}
\begin{minipage}[c][0.255\textheight][c]{\linewidth}\centering\includegraphics[width=\linewidth,height=0.255\textheight,keepaspectratio]{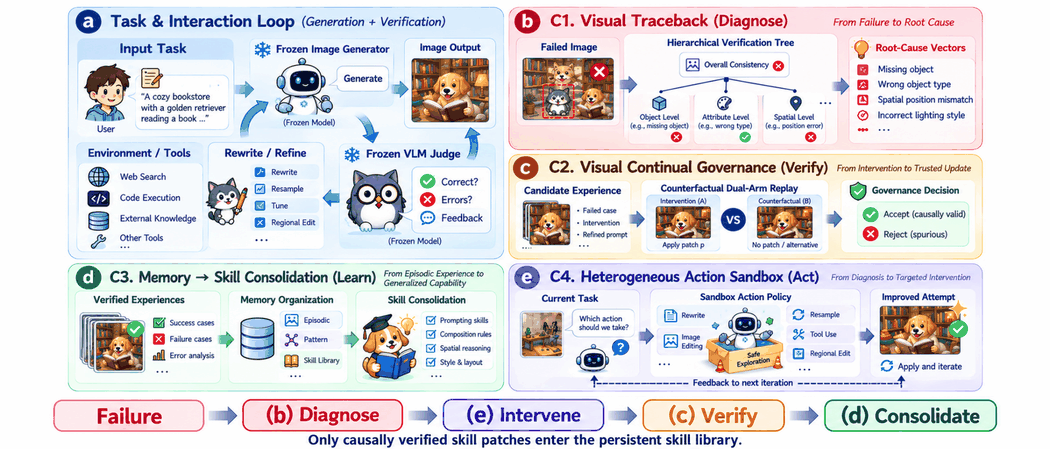}\end{minipage}\par
\vspace{3pt}
{\scriptsize \setlength{\tabcolsep}{1.5pt}\begin{tabular*}{\linewidth}{@{\extracolsep{\fill}}rrrrrrr@{}}
\toprule
C & O & G & A & I & S & Mean \\
3 (+0) & 3 (-1) & 3 (+0) & 4 (+0) & 3 (+0) & 4 (+0) & 3.33 (-0.17)\\
\bottomrule
\end{tabular*}
}
\end{minipage}\par
\vspace{3pt}
{\scriptsize Source public-date proxy: 2026-10-03. All displayed scores are medians of three runs; parentheses show revised minus original.\par}
\vfill
{\scriptsize C: clarity; O: organization/readability; G: graphical consistency; A: aesthetic coherence; I: visual inventiveness; S: design specificity. Mean is the arithmetic mean of six dimension medians, a descriptive display value, not a validated quality index.\par}
\clearpage

\clearpage

\subsection{Original Profiles and Temporal Context}
\label{sec:original-results}
\label{sec:visual-complexity}

\begin{figure*}[t]
\centering
\includegraphics[width=\textwidth]{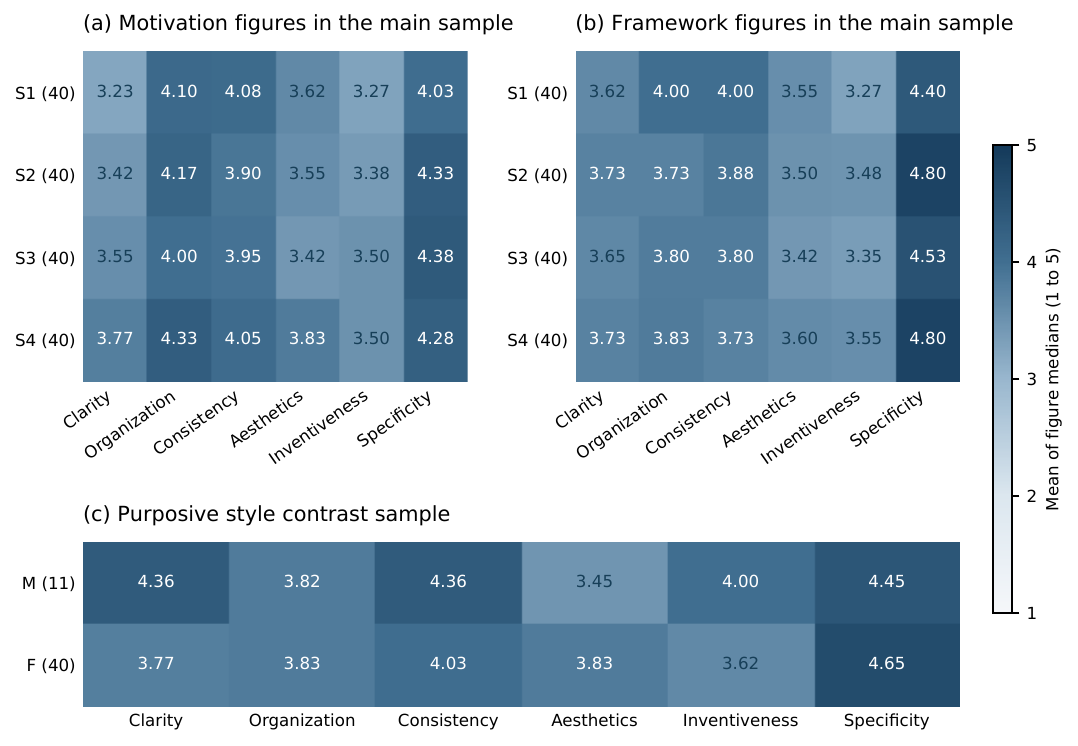}
\caption{Original-figure profiles across stages and roles. Cells show means of per-figure medians from three judgments, and panel labels report figure counts. The main sample retains equal topic quotas, and the purposive contrast sample is displayed separately. All panels share the same 1--5 color scale.}
\label{fig:stage-profiles}
\end{figure*}

Orderly presentation, distinctive content, and explanatory clarity do not necessarily coincide (Fig.~\ref{fig:stage-profiles}). Across all four stages, main-sample motivation figures have higher mean organization ratings than clarity ratings. The numerical difference narrows from 0.875 in S1 to 0.550 in S4, as clarity increases by 0.550 and organization by 0.225. The kernel-reformulation example illustrates the underlying design distinction. Its aligned image panels provide an orderly comparison, but the significance of that comparison remains insufficiently stated~\citep{corpus024}. Organizing examples and explaining their relevance are separate compositional decisions.

Framework figures receive their highest mean ratings for specificity in every stage. From S1 to S4, specificity increases from 4.400 to 4.800, while clarity changes from 3.625 to 3.725 and consistency decreases from 4.000 to 3.725. High specificity therefore does not imply equally favorable ratings for the explanation of relationships or the consistency of visual encodings. Both roles have their lowest mean aesthetic scores in S3 (3.425 for each) and their highest in S4 (3.825 for motivation figures and 3.600 for frameworks). S4 motivation figures also have the highest mean clarity and organization scores, whereas S4 frameworks have the lowest mean consistency score. The profiles consequently reveal changes that differ by dimension and figure role, rather than a uniform historical decline.

Purposeful visual inventiveness also remains distinct from technical specificity and visual polish. Mean inventiveness scores across the main-sample stage and role groups range from 3.275 to 3.550. Across all original figures, 58 receive a specificity median of 5 but an inventiveness median no greater than 3. Another 68 receive an aesthetic median of at least 4 but a clarity median no greater than 3. These exploratory combinations illustrate why a favorable score in one dimension cannot substitute for examining the complete profile.

The gallery provides examples of expanding explanatory scope. Within image restoration, the S1 framework in entry 030 organizes two training components around pseudo-pair generation and transfer between source and target domains~\citep{corpus030}. In S2, Complexity Experts combines an overall architecture, transformer blocks, and nested expert routing~\citep{corpus062}. In S3, UARE coordinates instructions, images, two expert pathways, training stages, and multimodal outputs~\citep{corpus110}. The S4 framework in entry 147 extends the account to prediction, action selection, execution, verification, and tool evolution~\citep{corpus147}. Across these selected examples, a single composition accommodates progressively more levels of explanation.

Increasing technical scope can create a need to depict additional interacting functions. AI assistance may also make panels, icons, and annotations easier to produce, potentially encouraging elaboration. The present collection cannot separate these explanations or establish a monotonic historical trend in visual complexity. Early entry 030 already uses gradients, rounded panels, and shadows, while the later E-RayZer framework retains a clear horizontal reading path~\citep{corpus112}. These examples direct attention to whether each visual layer expresses a necessary relationship and remains integrated with the overall explanation.

\subsection{Visual Conventions in the Style-Contrast Sample}
\label{sec:style-results}

The purposive style-contrast sample contains 40 papers and 51 figures. Its atlas label, \emph{S4-AI-Style}, identifies a selection path and does not establish AI production or define a fifth temporal stage. The 11 motivation figures reverse the ordering observed in the main sample, with mean clarity exceeding mean organization. Relative to S4 main-sample motivation figures, they receive higher clarity and inventiveness ratings but lower organization and aesthetic ratings. The 40 contrast frameworks exceed their S4 counterparts in aesthetics and consistency while retaining high specificity (4.650). The selected conventions therefore accompany different rating profiles across the two roles.

Their explanatory potential is visible in individual compositions. Agent Skills Under Change uses recurring robots, folders, targets, and shields to compare immediate updates with checks of adaptation and retention. Its clarity, organization, and consistency medians are all 5~\citep{corpus164}. Sprout connects queries and operations through timelines and shared memory, receiving inventiveness and specificity medians of 5 for both roles~\citep{corpus194}. AccessWorld, the original framework with the highest six-dimension mean, also belongs to this sample~\citep{corpus163}. These cases illustrate how familiar graphical elements can support distinctive explanations when their relationships carry specific meaning.

The revision workflow nevertheless produces limited aggregate gains for contrast frameworks. Their six-dimension mean increases in five figures, remains unchanged in 16, and decreases in 19. Mean aesthetics, organization, and inventiveness decrease by 0.300, 0.175, and 0.100, respectively. Specificity remains unchanged in 37 figures and decreases in three. OpenAgentFlow illustrates the role difference within one paper. Its motivation mean increases from 3.83 to 4.50, while its framework mean decreases from 3.50 to 3.17 following a revision that adds an outcome key beneath an already dense workflow~\citep{corpus191}. Such cases suggest that selective emphasis and consolidation deserve consideration when a composition already contains extensive explanatory material.

\subsection{Revision Outcomes across Dimensions and Figure Roles}
\label{sec:paired-results}
\label{sec:stratified-results}

\begin{figure*}[t]
\centering
\includegraphics[width=\textwidth]{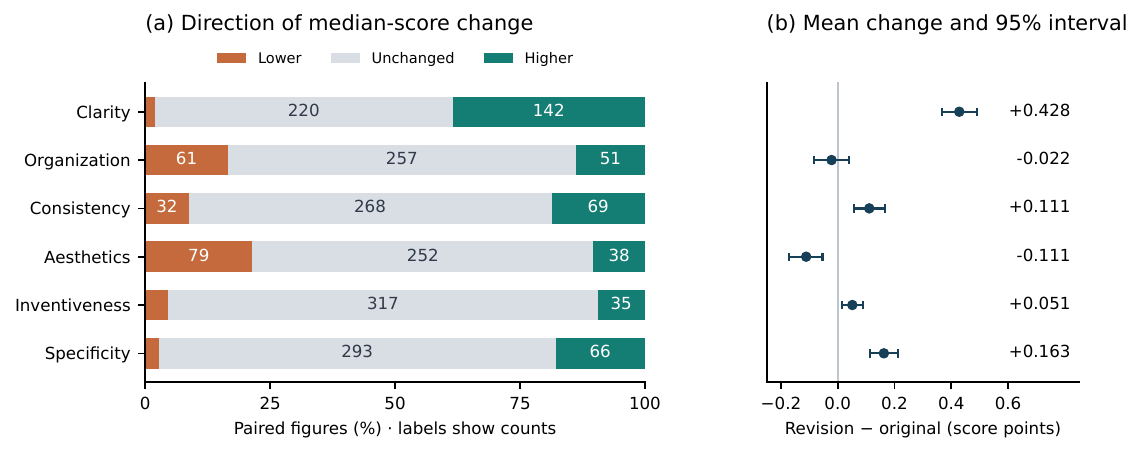}
\caption{Paired outcomes for 369 revisions that pass the recorded fidelity checks. Left: counts of increased, unchanged, and decreased figure-level medians. Right: mean changes and 95\% paper-cluster bootstrap percentile intervals.}
\label{fig:revision-deltas}
\end{figure*}

The largest rating gains occur in clarity, while changes in other dimensions are more mixed. Of 371 attempted revisions, 369 pass the recorded fidelity checks, and the two failures remain unscored. The accepted pairs span all 200 papers. Table~\ref{tab:paired-scores} and Fig.~\ref{fig:revision-deltas} show a mean clarity increase of 0.428, accompanied by smaller increases in specificity and consistency. Mean aesthetics decreases by 0.111, while inventiveness remains unchanged in 317 figures (85.9\%). Across 2,214 paired dimension outcomes, 401 increase (18.1\%), 1,607 remain unchanged (72.6\%), and 206 decrease (9.3\%). Most changed medians move by one point, comprising 370 increases and 205 decreases. The six-dimension mean rises from 3.842 to 3.945 ($+0.103$), combining these opposing movements.

\begin{table*}[t]
\centering
\small
\setlength{\tabcolsep}{4pt}
\renewcommand{\arraystretch}{1.12}
\caption{Paired dimension scores for 369 revisions that pass the recorded fidelity checks. Values are means of figure-level medians from three judgments, and $\Delta$ denotes revised minus original. Intervals are 95\% paper-cluster bootstrap percentile intervals based on 5,000 resamples with seed 20261020. Direction counts refer to figure-level median changes.}
\label{tab:paired-scores}
\begin{tabular*}{\textwidth}{@{\extracolsep{\fill}}lrrrrc@{}}
\toprule
Dimension & Original & Revised & $\Delta$ & 95\% interval & Up / Same / Down \\
\midrule
Communicative clarity & 3.629 & 4.057 & +0.428 & [+0.367, +0.492] & 142 / 220 / 7 \\
Organization and readability & 3.973 & 3.951 & -0.022 & [-0.084, +0.040] & 51 / 257 / 61 \\
Graphical consistency & 3.949 & 4.060 & +0.111 & [+0.056, +0.168] & 69 / 268 / 32 \\
Aesthetic coherence & 3.591 & 3.480 & -0.111 & [-0.170, -0.054] & 38 / 252 / 79 \\
Purposeful visual inventiveness & 3.450 & 3.501 & +0.051 & [+0.014, +0.089] & 35 / 317 / 17 \\
Design specificity & 4.461 & 4.623 & +0.163 & [+0.114, +0.212] & 66 / 293 / 10 \\
\bottomrule
\end{tabular*}
\end{table*}

The pooled gain differs substantially by figure role (Table~\ref{tab:cohort-role-paired}). The six-dimension average increases by 0.242 for main-sample motivation figures and by 0.018 for frameworks. Motivation clarity and specificity increase by 0.800 and 0.306, compared with 0.151 and 0.082 for frameworks. Framework aesthetics decreases by 0.164. The contrast sample similarly combines a small motivation gain with a framework decline. Under this workflow, motivation figures therefore receive more favorable average rating changes than frameworks.

\begin{table*}[t]
\centering
\small
\setlength{\tabcolsep}{4pt}
\renewcommand{\arraystretch}{1.12}
\caption{Paired six-dimension averages by sample and figure role. M and F denote motivation and framework figures, and $n$ is the number of accepted original--revision pairs. Intervals are 95\% paper-cluster bootstrap percentile intervals within each row. Direction counts refer to changes in the post hoc six-dimension average.}
\label{tab:cohort-role-paired}
\begin{tabular*}{\textwidth}{@{\extracolsep{\fill}}llrrrrrc@{}}
\toprule
Sample & Role & $n$ & Original & Revised & $\Delta$ & 95\% interval & Up / Same / Down \\
\midrule
Main & M & 160 & 3.818 & 4.059 & +0.242 & [+0.186, +0.298] & 107 / 29 / 24 \\
Main & F & 159 & 3.824 & 3.842 & +0.018 & [-0.018, +0.055] & 57 / 54 / 48 \\
\addlinespace
Style contrast & M & 10 & 4.067 & 4.133 & +0.067 & [-0.083, +0.233] & 4 / 3 / 3 \\
Style contrast & F & 40 & 3.954 & 3.854 & -0.100 & [-0.167, -0.033] & 5 / 16 / 19 \\
\bottomrule
\end{tabular*}
\end{table*}

This role difference recurs in every main-sample topic (Table~\ref{tab:topic-role}). Detection motivation figures have the largest mean gain and form the only motivation topic group with a positive mean aesthetic change. Framework aesthetics decreases in all four topics. The distinction also appears within individual papers. Among the 159 papers with accepted revisions for both roles, the motivation change exceeds the framework change in 103, equals it in 23, and falls below it in 33. Of the 31 papers with opposing directions of change, 27 combine a motivation gain with a framework loss, and four show the reverse. Averaging the two roles within a paper would conceal these differences.

\begin{table*}[t]
\centering
\small
\setlength{\tabcolsep}{4pt}
\renewcommand{\arraystretch}{1.12}
\caption{Main-sample revision changes by topic and figure role, pooling S1--S4. $\Delta C$ and $\Delta A$ denote mean changes in clarity and aesthetic coherence, while $\Delta\bar m$ denotes the change in the post hoc six-dimension average. Counts refer to accepted figure pairs. One segmentation framework failed fidelity review, leaving 39 pairs in that cell and 40 in every other cell.}
\label{tab:topic-role}
\begin{tabular*}{\textwidth}{@{\extracolsep{\fill}}lrrrrrrrr@{}}
\toprule
& \multicolumn{4}{c}{Motivation} & \multicolumn{4}{c}{Framework} \\
\cmidrule(lr){2-5}\cmidrule(lr){6-9}
Topic & $n$ & $\Delta C$ & $\Delta A$ & $\Delta\bar m$ & $n$ & $\Delta C$ & $\Delta A$ & $\Delta\bar m$ \\
\midrule
Object detection & 40 & +0.900 & +0.125 & +0.367 & 40 & +0.150 & -0.150 & +0.046 \\
Image segmentation & 40 & +0.600 & -0.050 & +0.146 & 39 & +0.205 & -0.128 & +0.013 \\
Image restoration & 40 & +0.850 & -0.100 & +0.225 & 40 & +0.150 & -0.125 & +0.021 \\
3D vision & 40 & +0.850 & -0.100 & +0.229 & 40 & +0.100 & -0.250 & -0.008 \\
\bottomrule
\end{tabular*}
\end{table*}

Revision outcomes also vary across stages. Mean motivation clarity increases by 1.075, 0.900, 0.650, and 0.575 in S1--S4, respectively. The six-dimension mean increases for 35 S1 motivation figures, remains unchanged for five, and decreases for none. In S4, the corresponding counts are 14 increases, 14 ties, and 12 decreases. Framework aesthetics decreases in every stage. These patterns place the pooled result in the context of different initial scores and compositions, but stage membership does not identify the mechanisms responsible. Revisions can introduce caption-derived information, alter the canvas, and combine retained panels with explanatory guides. The findings therefore characterize the complete intervention and do not isolate changes in visual style.

\subsection{Tradeoffs Concealed by Aggregate Scores}
\label{sec:tradeoffs}

The small overall gain conceals substantial variation among individual figures. Six-dimension means increase for 173 figures (46.9\%), remain unchanged for 102 (27.6\%), and decrease for 94 (25.5\%). Only 79 figures retain all six medians. The 72.6\% unchanged rate reported above therefore concerns individual dimension outcomes, not complete figure profiles. Table~\ref{tab:change-patterns} distinguishes gains without losses, losses without gains, mixed changes, and fully unchanged profiles. Among the 102 figures with unchanged means, 23 have offsetting dimension changes. Another 20 figures show a higher mean despite a decline in at least one dimension. Twenty-one figures gain clarity while losing aesthetic coherence.

\begin{table*}[t]
\centering
\small
\setlength{\tabcolsep}{4pt}
\renewcommand{\arraystretch}{1.12}
\caption{Mutually exclusive patterns of dimension-level change across 369 accepted revisions. A gain or loss denotes an increase or decrease in a figure's median score on at least one dimension. The unchanged category requires all six medians to remain identical. Percentages use 369 figures as the denominator.}
\label{tab:change-patterns}
\begin{tabular*}{\textwidth}{@{\extracolsep{\fill}}lrr@{}}
\toprule
Pattern & Figures & Percentage \\
\midrule
Gains without losses & 153 & 41.5\% \\
Losses without gains & 80 & 21.7\% \\
Mixed gains and losses & 57 & 15.4\% \\
All six dimensions unchanged & 79 & 21.4\% \\
\midrule
Total & 369 & 100.0\% \\
\bottomrule
\end{tabular*}
\end{table*}

PULSE illustrates how opposing changes can leave an average unchanged. Its clarity median increases from 4 to 5 while aesthetics decreases from 5 to 4, leaving the mean at 4.50~\citep{corpus127}. In the wildfire-reporting framework, added headings and an enlarged interaction detail accompany a consistency increase from 3 to 4 and an aesthetic decrease from 4 to 3~\citep{corpus162}. Its mean also remains unchanged. In both cases, the stable aggregate conceals a changed evaluation profile.

Initial scores also constrain the available range of change. Specificity reaches the maximum of 5 in 201 of the 371 originals, whereas inventiveness reaches 5 in only three. Within the 369 matched pairs, the number of specificity maxima increases from 199 to 239 and the number of clarity maxima from 16 to 66. Aesthetic maxima decrease from 13 to three. Among matched pairs, originals with means in $[3,4)$ gain 0.209 on average ($n=212$), while those starting at 4 or above change by $-0.050$ ($n=156$). Entry 128\,M is the sole original with a mean below 3. The bounded scale constrains improvement at high starting scores, while regression toward the mean and feedback-informed revision may also contribute to this association. Their contributions were not separately estimated. These observations favor identifying the dimensions that require attention before adopting a higher aggregate score as the design objective.

\subsection{Case Studies: Explanation and Compositional Judgment}
\label{sec:cases}
\label{sec:qualitative-results}

A revision can make a claim more explicit without improving the composition through which it is read. The following cases connect this distinction to visible changes in individual figures. Table~\ref{tab:score-extremes} identifies absolute score extrema using the post hoc six-dimension mean. Subsequent comparisons examine how annotations, typography, and repeated detail accompany different rating outcomes. These extrema rank images under the configured rubric, and the illustrative selections do not estimate the prevalence of design patterns across the corpus.

\begin{table*}[t]
\centering
\small
\setlength{\tabcolsep}{4pt}
\renewcommand{\arraystretch}{1.12}
\caption{Absolute score extrema selected by the post hoc six-dimension mean. Entry 163\,F is the unique maximum within the original set and within the accepted-revision set. Entry 128\,M is the unique original minimum, while 136\,F and 195\,F tie for the accepted-revision minimum. Vectors follow $(C,O,G,A,I,S)$, and means and changes are computed before rounding. These ranks concern model-rated figures, not the scientific quality of their papers.}
\label{tab:score-extremes}
\begin{tabular*}{\textwidth}{@{\extracolsep{\fill}}lcrcrr@{}}
\toprule
& \multicolumn{2}{c}{Original} & \multicolumn{2}{c}{Revised} & \\
\cmidrule(lr){2-3}\cmidrule(lr){4-5}
Case & Score vector & Mean & Score vector & Mean & $\Delta\bar m$ \\
\midrule
163\,F: AccessWorld~\citep{corpus163} & $(4,5,5,5,5,5)$ & 4.83 & $(5,5,5,5,5,5)$ & 5.00 & +0.17 \\
128\,M: Over-segmentation~\citep{corpus128} & $(2,3,3,3,2,3)$ & 2.67 & $(4,4,5,4,4,5)$ & 4.33 & +1.67 \\
136\,F: ReTax-Seg~\citep{corpus136} & $(3,3,3,2,3,4)$ & 3.00 & $(3,3,3,2,3,4)$ & 3.00 & 0.00 \\
195\,F: Stride~\citep{corpus195} & $(3,3,3,3,4,5)$ & 3.50 & $(3,2,3,2,3,5)$ & 3.00 & -0.50 \\
\bottomrule
\end{tabular*}
\end{table*}

\begin{figure*}[t]
\centering
\begingroup\small
\textbf{A. AccessWorld: highest original and highest accepted revision}~\citep{corpus163}\par\smallskip
\begin{minipage}[t]{0.485\textwidth}
\centering
\textbf{Original}\par\smallskip
\begin{minipage}[c][5.5cm][c]{\linewidth}
\centering
\includegraphics[width=\linewidth,height=5.5cm,keepaspectratio]{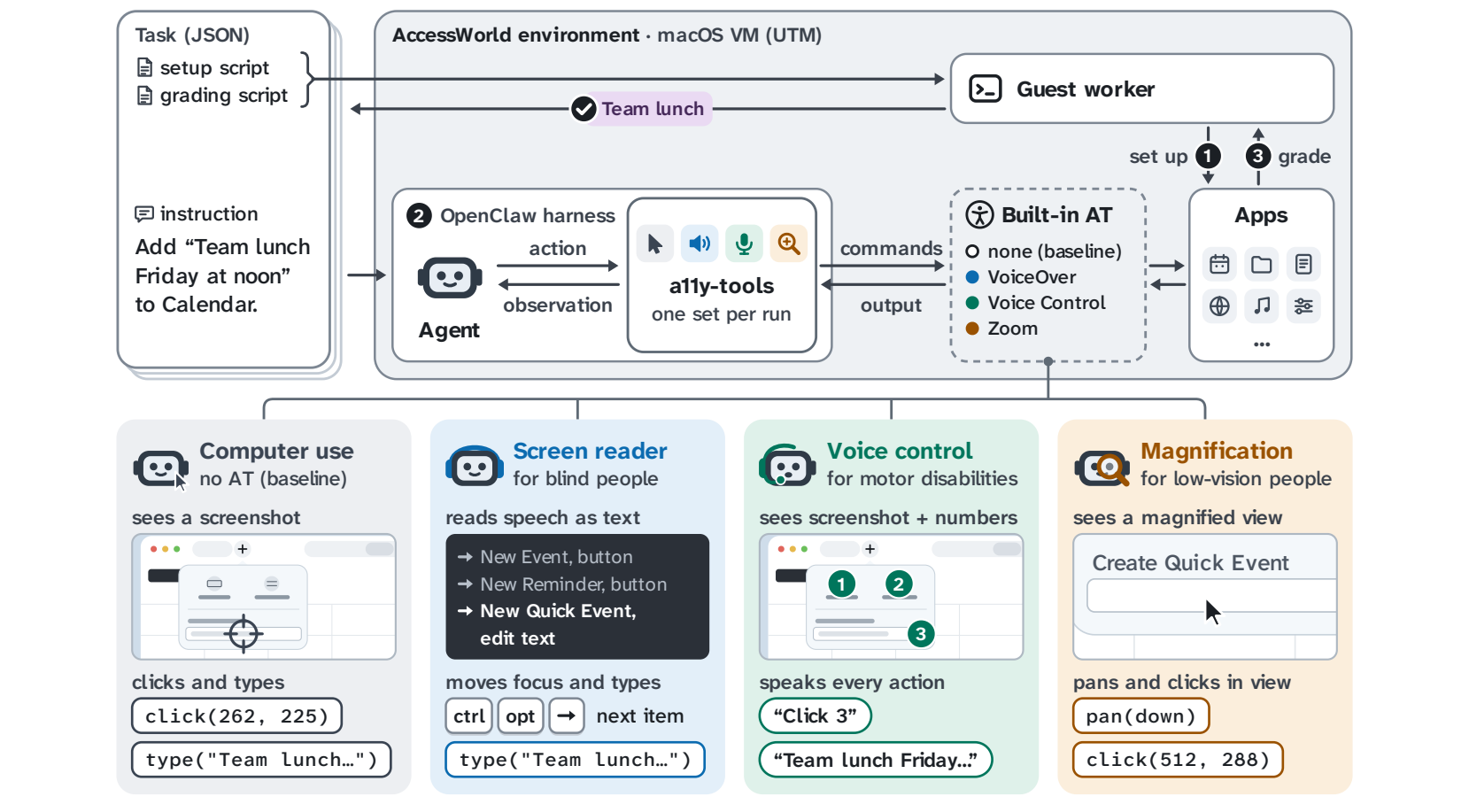}
\end{minipage}\par\smallskip
\textsf{C 4\quad O 5\quad G 5\quad A 5\quad I 5\quad S 5}\par
Six-score mean: 4.83
\end{minipage}\hfill\begin{minipage}[t]{0.485\textwidth}
\centering
\textbf{Study-generated revision}\par\smallskip
\begin{minipage}[c][5.5cm][c]{\linewidth}
\centering
\includegraphics[width=\linewidth,height=5.5cm,keepaspectratio]{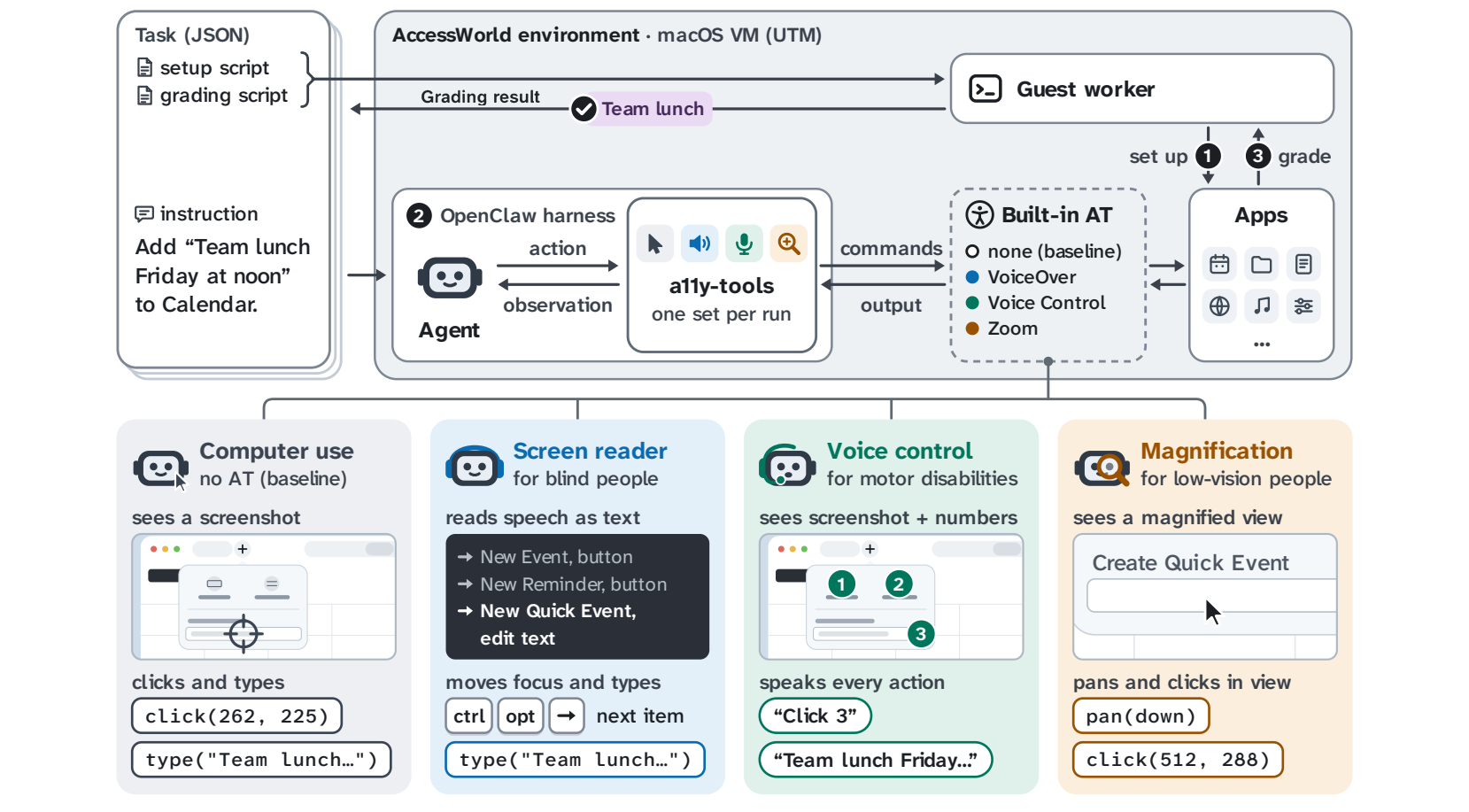}
\end{minipage}\par\smallskip
\textsf{C 5\quad O 5\quad G 5\quad A 5\quad I 5\quad S 5}\par
Six-score mean: 5.00\quad ($\Delta=+0.17$)
\end{minipage}
\par
\medskip\hrule\medskip
\textbf{B. Over-segmentation: lowest original and largest paired increase}~\citep{corpus128}\par\smallskip
\begin{minipage}[t]{0.485\textwidth}
\centering
\textbf{Original}\par\smallskip
\begin{minipage}[c][2.6cm][c]{\linewidth}
\centering
\includegraphics[width=\linewidth,height=2.6cm,keepaspectratio]{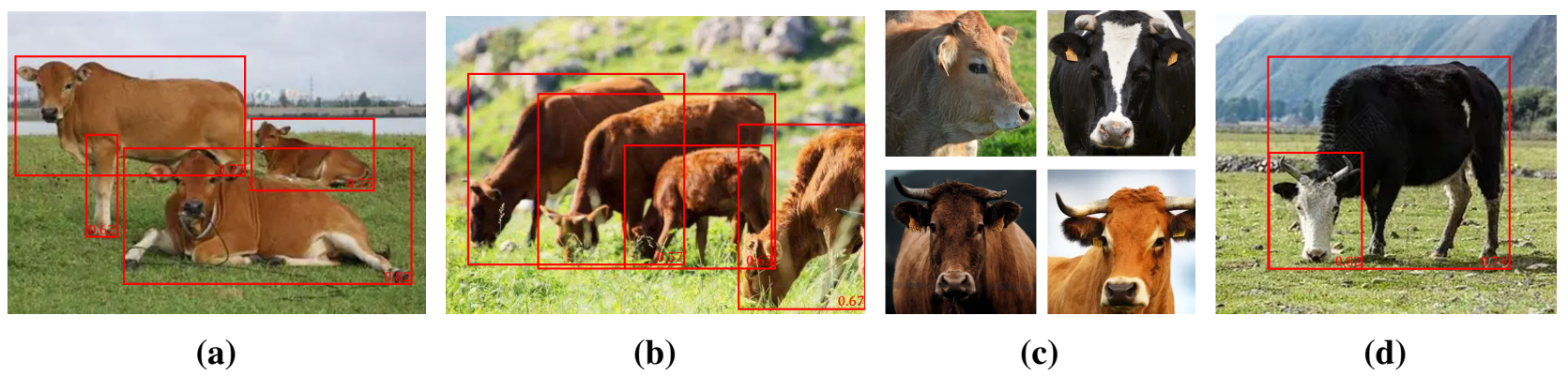}
\end{minipage}\par\smallskip
\textsf{C 2\quad O 3\quad G 3\quad A 3\quad I 2\quad S 3}\par
Six-score mean: 2.67
\end{minipage}\hfill\begin{minipage}[t]{0.485\textwidth}
\centering
\textbf{Study-generated revision}\par\smallskip
\begin{minipage}[c][2.6cm][c]{\linewidth}
\centering
\includegraphics[width=\linewidth,height=2.6cm,keepaspectratio]{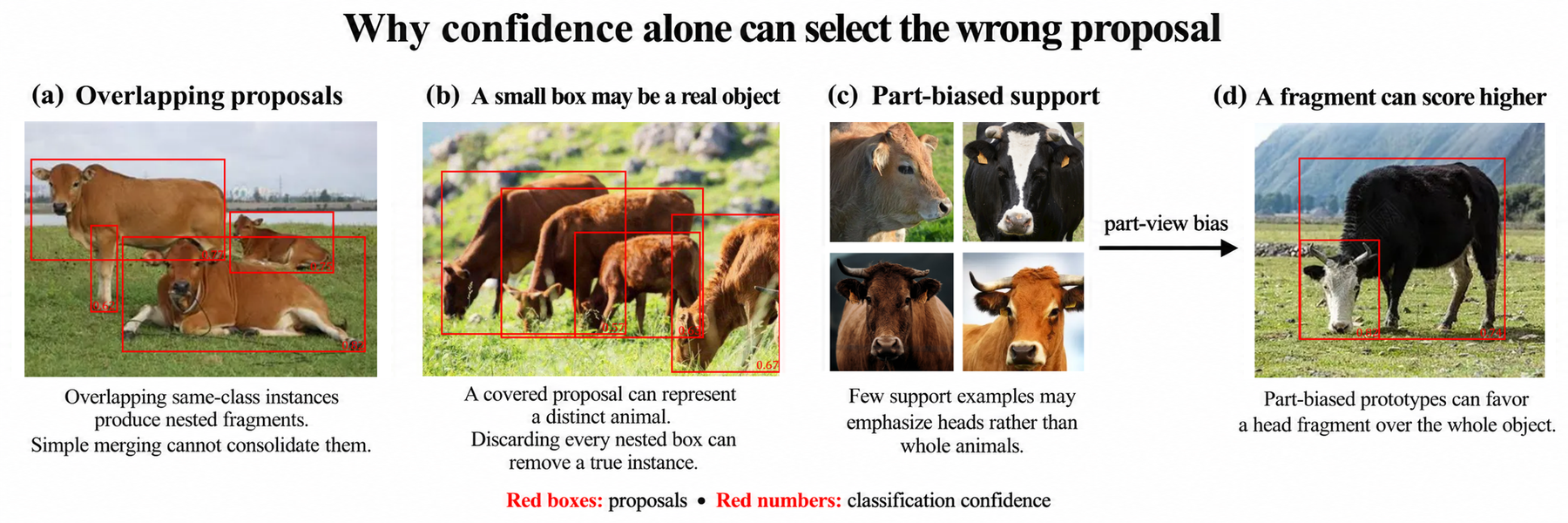}
\end{minipage}\par\smallskip
\textsf{C 4\quad O 4\quad G 5\quad A 4\quad I 4\quad S 5}\par
Six-score mean: 4.33\quad ($\Delta=+1.67$)
\end{minipage}
\par
\endgroup
\caption{Absolute original-score extrema and their revisions. AccessWorld (163\,F) is the unique original maximum, and the over-segmentation example (128\,M) is the unique original minimum. Both were identified after scoring using the six-median average. Images are reproduced in full, with three-request median scores in C/O/G/A/I/S order. Original artwork is credited to the cited papers, and revisions are study-generated derivatives.}
\label{fig:score-extremes}
\end{figure*}

AccessWorld, the highest-rated original, organizes its explanation around a repeated task (Fig.~\ref{fig:score-extremes}A)~\citep{corpus163}. The upper panel connects the task, agent, assistive-technology interface, applications, and grading process. The lower panel compares interaction modes through corresponding examples. Its original mean is 4.83. The revision preserves this arrangement and explicitly labels the grading connection. Clarity increases to 5, bringing the six-dimension mean to 5.00. Although all six medians reach the maximum, the 18 individual ratings are not unanimously 5. Its membership in the contrast sample also demonstrates that the conventions used for purposive selection are compatible with highly rated visual communication.

The lowest-rated original illustrates the importance of making an intended comparison explicit. The over-segmentation motivation figure shows overlapping proposals, small instances, and part-biased support, but provides little explanation beyond panel letters (Fig.~\ref{fig:score-extremes}B)~\citep{corpus128}. The revision names these contrasts, defines the boxes and confidence values, and explains why a high-confidence fragment can be selected instead of the whole object. Clarity increases from 2 to 4 and specificity from 3 to 5. The mean rises from 2.67 to 4.33, yielding the largest paired increase ($+1.67$). Information previously confined to the caption becomes available to the image-only judge, so the gain cannot be attributed to layout alone.

\begin{figure*}[t]
\centering
\begingroup\small
\textbf{A. ReTax-Seg: unchanged scores after adding explanatory guidance}~\citep{corpus136}\par\smallskip
\begin{minipage}[t]{0.485\textwidth}
\centering
\textbf{Original}\par\smallskip
\begin{minipage}[c][4.3cm][c]{\linewidth}
\centering
\includegraphics[width=\linewidth,height=4.3cm,keepaspectratio]{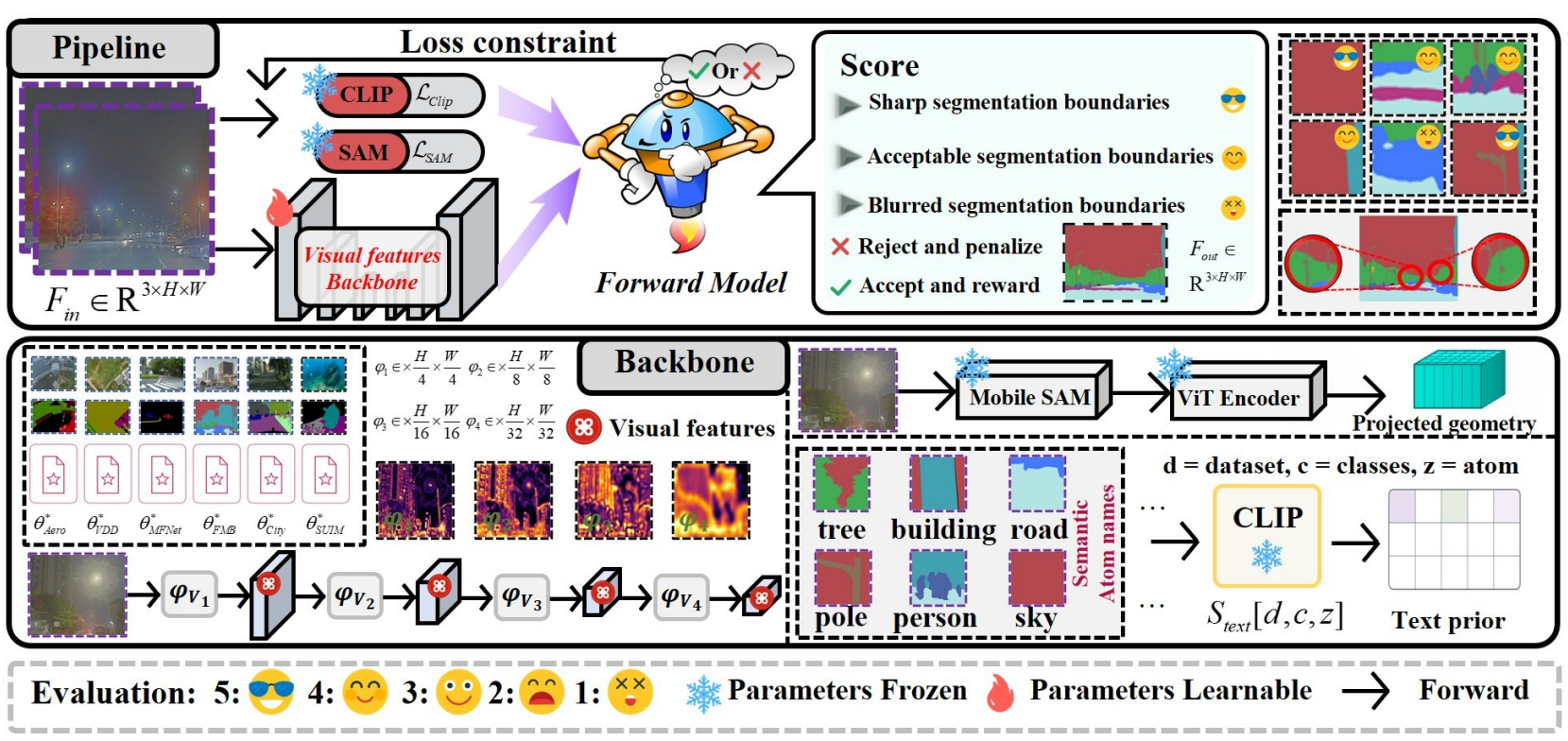}
\end{minipage}\par\smallskip
\textsf{C 3\quad O 3\quad G 3\quad A 2\quad I 3\quad S 4}\par
Six-score mean: 3.00
\end{minipage}\hfill\begin{minipage}[t]{0.485\textwidth}
\centering
\textbf{Study-generated revision}\par\smallskip
\begin{minipage}[c][4.3cm][c]{\linewidth}
\centering
\includegraphics[width=\linewidth,height=4.3cm,keepaspectratio]{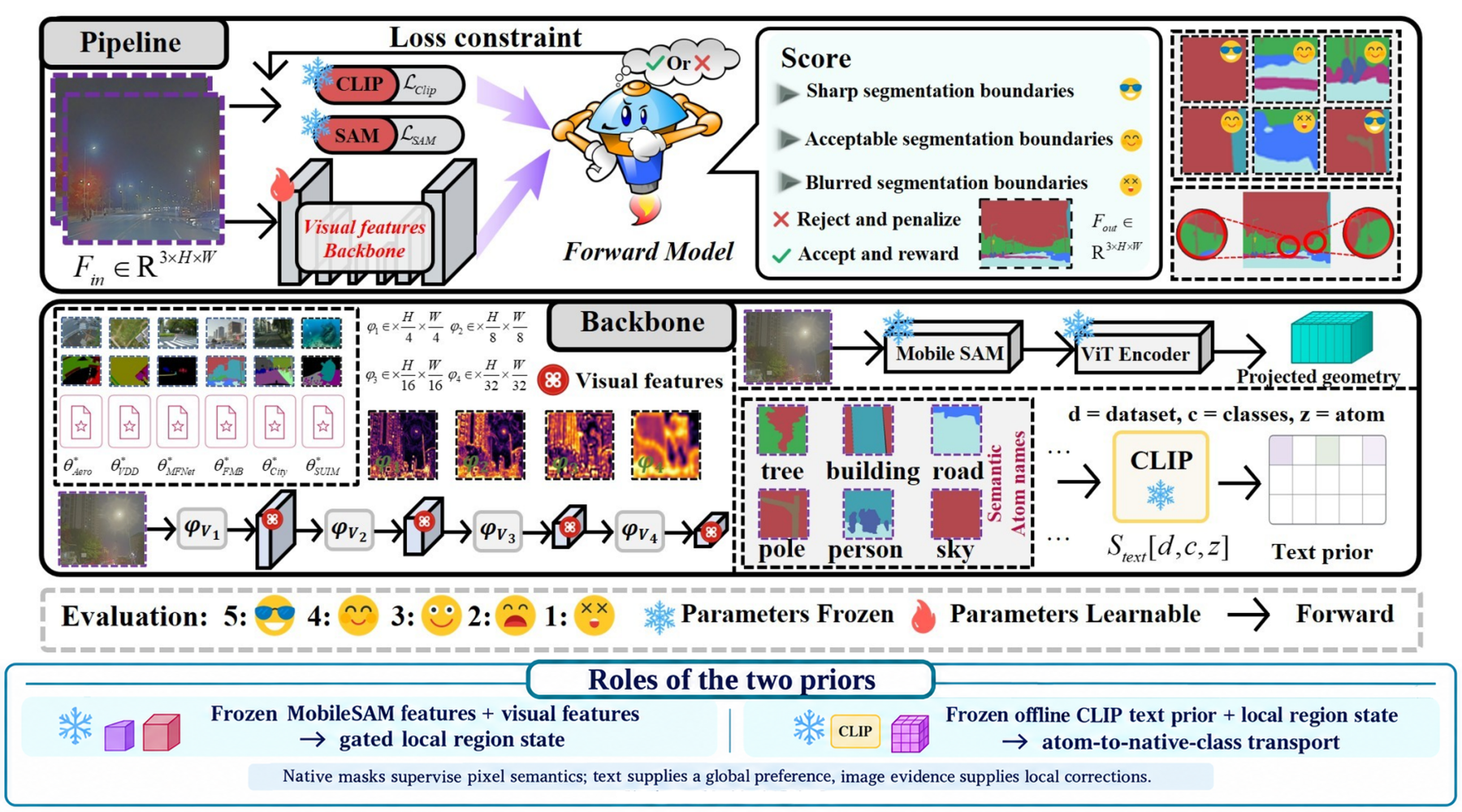}
\end{minipage}\par\smallskip
\textsf{C 3\quad O 3\quad G 3\quad A 2\quad I 3\quad S 4}\par
Six-score mean: 3.00\quad ($\Delta=0.00$)
\end{minipage}
\par
\medskip\hrule\medskip
\textbf{B. Stride: retained specificity with weaker organization}~\citep{corpus195}\par\smallskip
\begin{minipage}[t]{0.485\textwidth}
\centering
\textbf{Original}\par\smallskip
\begin{minipage}[c][4.3cm][c]{\linewidth}
\centering
\includegraphics[width=\linewidth,height=4.3cm,keepaspectratio]{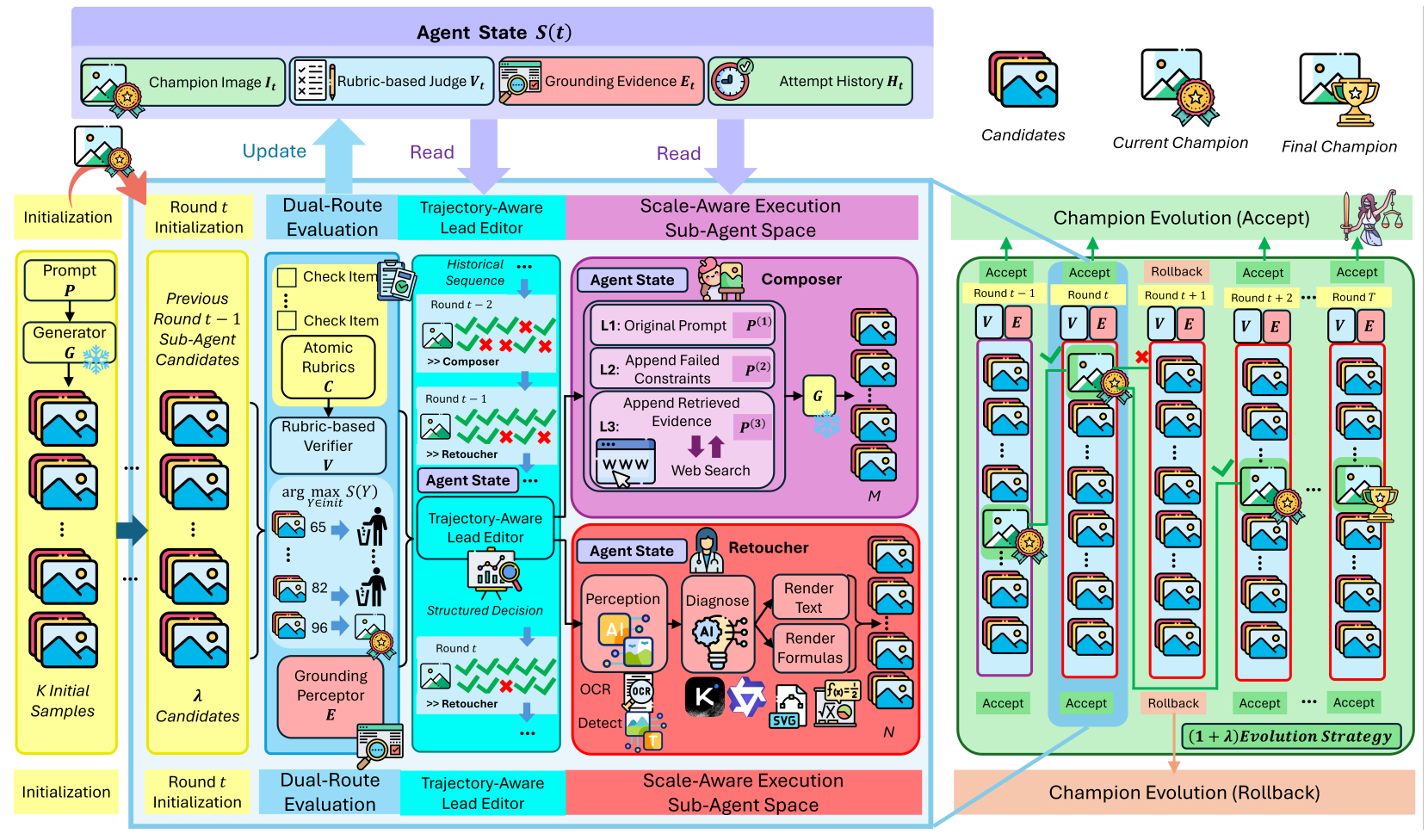}
\end{minipage}\par\smallskip
\textsf{C 3\quad O 3\quad G 3\quad A 3\quad I 4\quad S 5}\par
Six-score mean: 3.50
\end{minipage}\hfill\begin{minipage}[t]{0.485\textwidth}
\centering
\textbf{Study-generated revision}\par\smallskip
\begin{minipage}[c][4.3cm][c]{\linewidth}
\centering
\includegraphics[width=\linewidth,height=4.3cm,keepaspectratio]{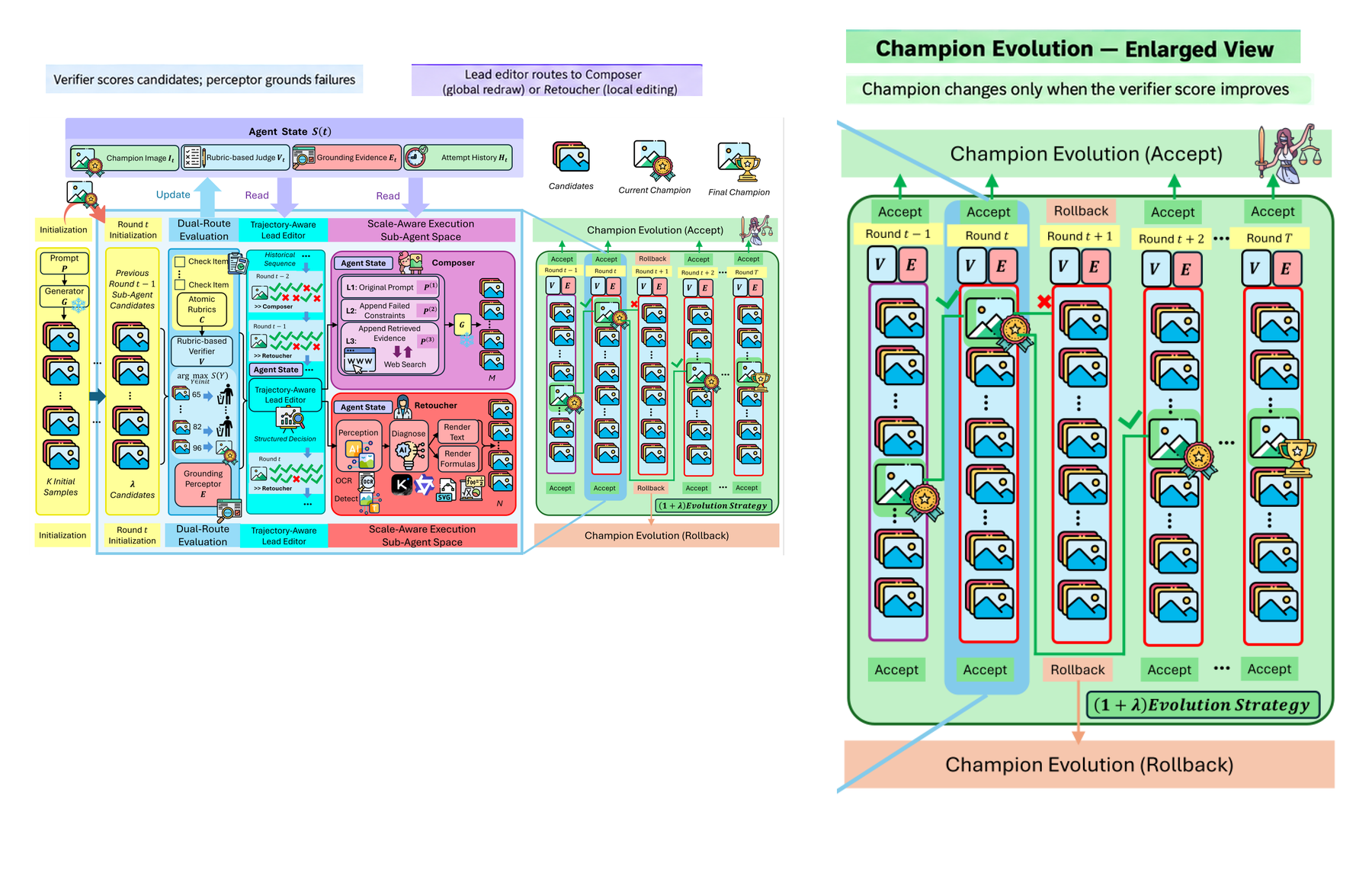}
\end{minipage}\par\smallskip
\textsf{C 3\quad O 2\quad G 3\quad A 2\quad I 3\quad S 5}\par
Six-score mean: 3.00\quad ($\Delta=-0.50$)
\end{minipage}
\par
\endgroup
\caption{The two accepted revisions tied for the lowest six-score mean (3.00). ReTax-Seg (136\,F) retains all six medians, while Stride (195\,F) loses one point each in organization, aesthetic coherence, and inventiveness. Selection is post hoc and illustrative. Full images and three-request median scores are shown. Original artwork is credited to the cited papers.}
\label{fig:low-score-revisions}
\end{figure*}

Additional explanation does not necessarily resolve a difficult underlying arrangement. ReTax-Seg and Stride share the lowest accepted revised mean of 3.00 (Fig.~\ref{fig:low-score-revisions}). ReTax-Seg retains all six medians after explanatory guidance is appended to its dense feature and prior-related pathways~\citep{corpus136}. Stride's mean decreases from 3.50 to 3.00, with specificity remaining at 5 and organization, aesthetic coherence, and inventiveness each decreasing by one point~\citep{corpus195}. These outcomes distinguish the presence of problem-specific content from the effectiveness of its organization.

The largest decline occurs in a different case. The continual-adaptation motivation figure decreases from 4.17 to 3.17~\citep{corpus042}. Adaptation and pruning annotations supplement its weather and driving examples, while white restoration patches interrupt the colored model panels. Organization decreases from 5 to 3 and aesthetic coherence from 4 to 3. The lowest revised scores and the largest score decrease therefore identify different figures.

\begin{figure*}[t]
\centering
\begingroup\small
\textbf{A. Complexity Experts: repeated headings}~\citep{corpus062}\par\smallskip
\begin{minipage}[t]{0.485\textwidth}
\centering
\textbf{Original}\par\smallskip
\begin{minipage}[t][7.7cm][t]{\linewidth}
\vspace{0pt}
\centering
\includegraphics[width=\linewidth,height=7.7cm,keepaspectratio]{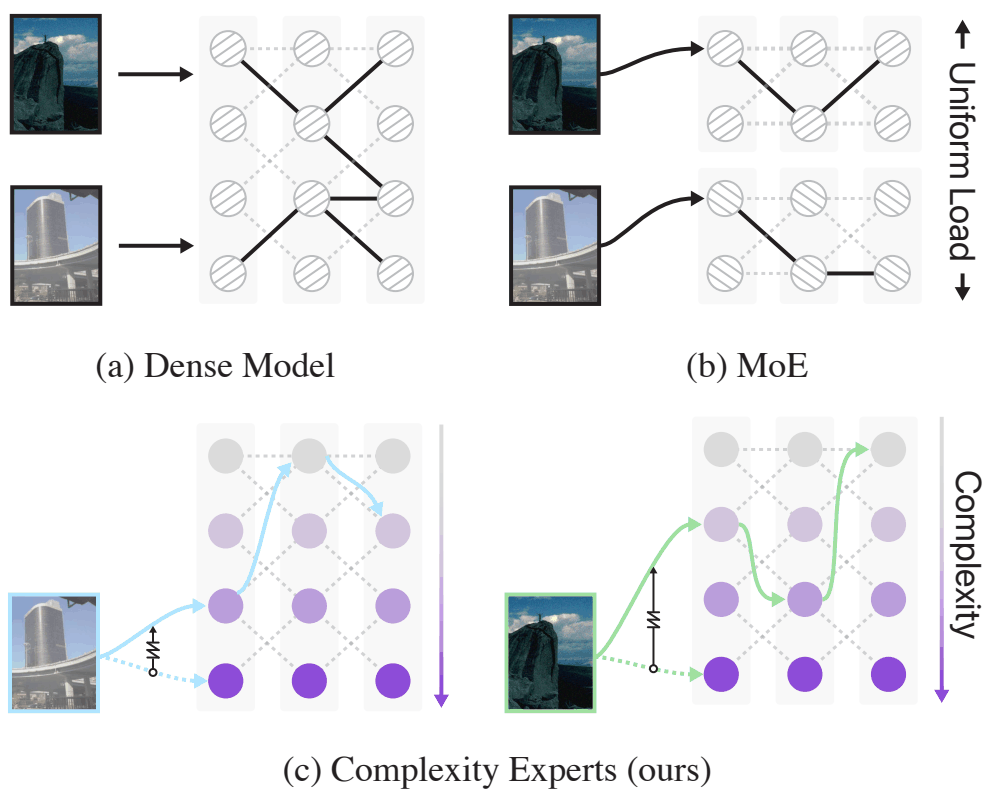}
\end{minipage}\par\smallskip
\textsf{C 3\quad O 5\quad G 3\quad A 4\quad I 4\quad S 4}\par
Six-score mean: 3.83
\end{minipage}\hfill%
\begin{minipage}[t]{0.485\textwidth}
\centering
\textbf{Study-generated revision}\par\smallskip
\begin{minipage}[t][7.7cm][t]{\linewidth}
\vspace{0pt}
\centering
\includegraphics[width=\linewidth,height=7.7cm,keepaspectratio]{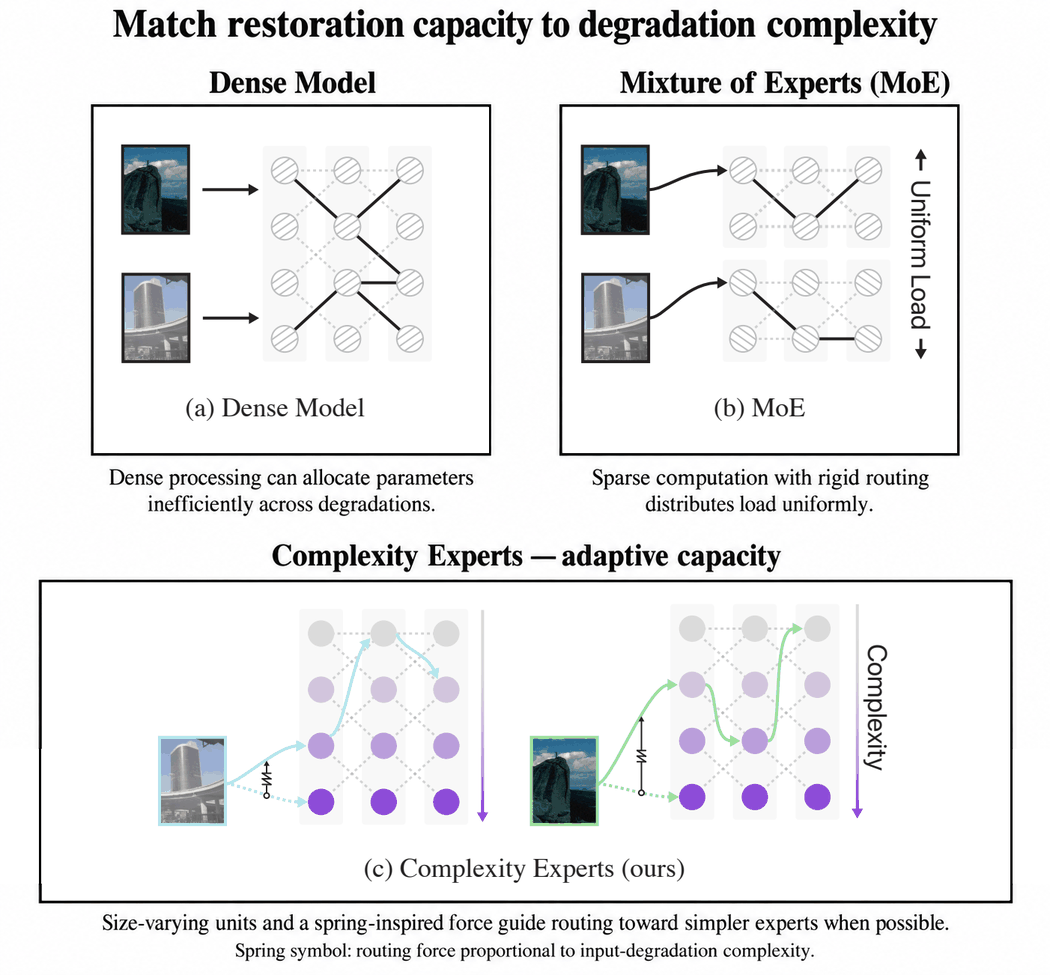}
\end{minipage}\par\smallskip
\textsf{C 4\quad O 4\quad G 3\quad A 4\quad I 3\quad S 4}\par
Six-score mean: 3.67\quad ($\Delta=-0.17$)
\end{minipage}
\par
\medskip\hrule\medskip
\textbf{B. VEvo-Harness: standardized outer typography}~\citep{corpus198}\par\smallskip
\begin{minipage}[t]{0.485\textwidth}
\centering
\textbf{Original}\par\smallskip
\begin{minipage}[t][5.5cm][t]{\linewidth}
\vspace{0pt}
\centering
\includegraphics[width=\linewidth,height=5.5cm,keepaspectratio]{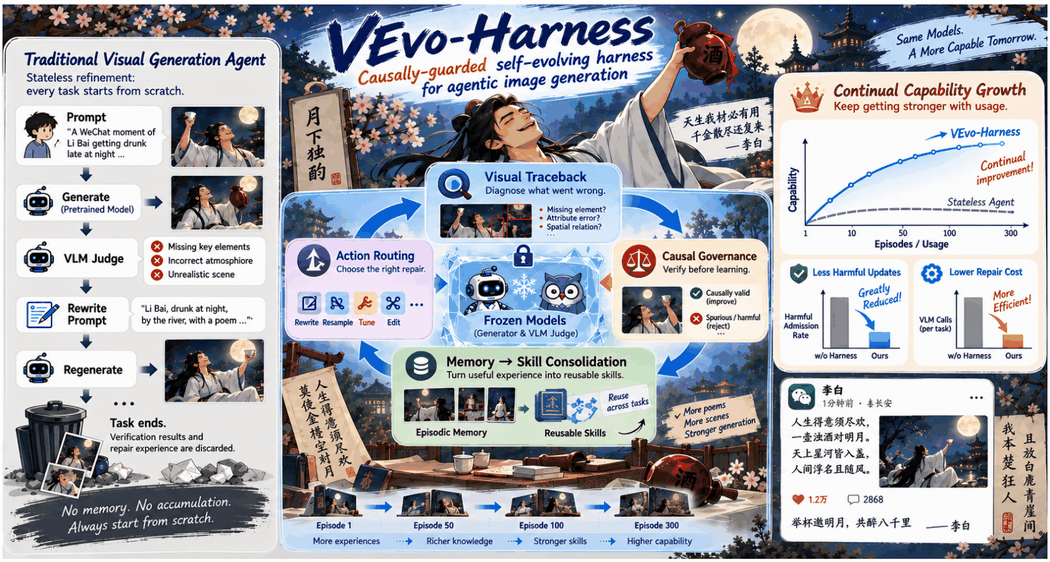}
\end{minipage}\par\smallskip
\textsf{C 4\quad O 3\quad G 4\quad A 3\quad I 4\quad S 4}\par
Six-score mean: 3.67
\end{minipage}\hfill%
\begin{minipage}[t]{0.485\textwidth}
\centering
\textbf{Study-generated revision}\par\smallskip
\begin{minipage}[t][5.5cm][t]{\linewidth}
\vspace{0pt}
\centering
\includegraphics[width=\linewidth,height=5.5cm,keepaspectratio]{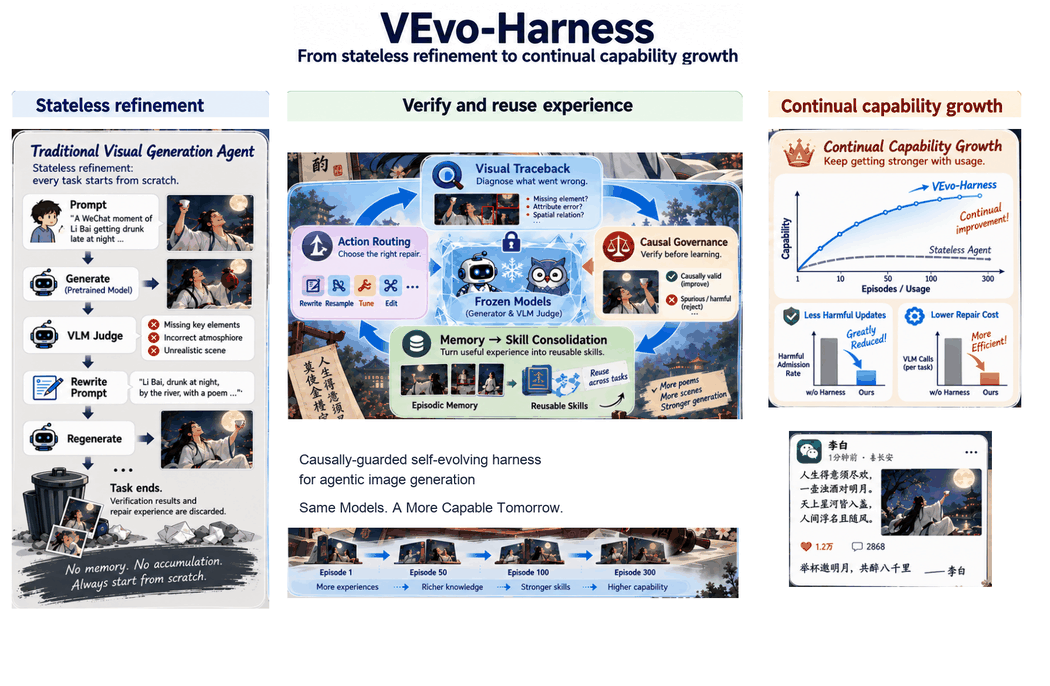}
\end{minipage}\par\smallskip
\textsf{C 5\quad O 3\quad G 4\quad A 3\quad I 4\quad S 5}\par
Six-score mean: 4.00\quad ($\Delta=+0.33$)
\end{minipage}
\par
\endgroup
\caption{Typographic changes in two motivation figures (062\,M and 198\,M). Added headings repeat retained labels in Complexity Experts. A conventional title and matching heading strips replace the integrated title treatment in VEvo-Harness. C, O, G, A, I, and S denote clarity, organization, graphical consistency, aesthetic coherence, inventiveness, and specificity. Scores are three-request medians, and changes are computed before display rounding. Images are shown in full. Original artwork is credited to the cited papers, and revisions are study-generated derivatives.}
\label{fig:typography-cases}
\end{figure*}

The typographic comparisons illustrate limited coordination between added emphasis and an existing visual hierarchy (Fig.~\ref{fig:typography-cases}). Complexity Experts adds bold headings that repeat the retained \emph{Dense Model} and \emph{MoE} labels, thick panel frames, and explanatory prose below the network comparisons~\citep{corpus062}. Clarity increases from 3 to 4, but organization decreases from 5 to 4 and inventiveness from 4 to 3. Aesthetic coherence remains at 4. VEvo-Harness replaces an integrated brush-like title with a conventional title, subtitle, and matching heading strips, while varied lettering remains within the retained panels~\citep{corpus198}. Clarity and specificity improve, but organization and aesthetic coherence remain at 3. These cases concern the integration of headings, technical labels, and annotations across the composition. A uniform outer treatment can make the message more explicit while leaving the internal hierarchy unresolved.

\begin{figure*}[t]
\centering
\begingroup\small
\textbf{BKBAgent: additional explanation without a clarity gain}~\citep{corpus167}\par\smallskip
\begin{minipage}[t]{0.485\textwidth}
\centering
\textbf{Original}\par\smallskip
\begin{minipage}[t][5.6cm][t]{\linewidth}
\vspace{0pt}
\centering
\includegraphics[width=\linewidth,height=5.6cm,keepaspectratio]{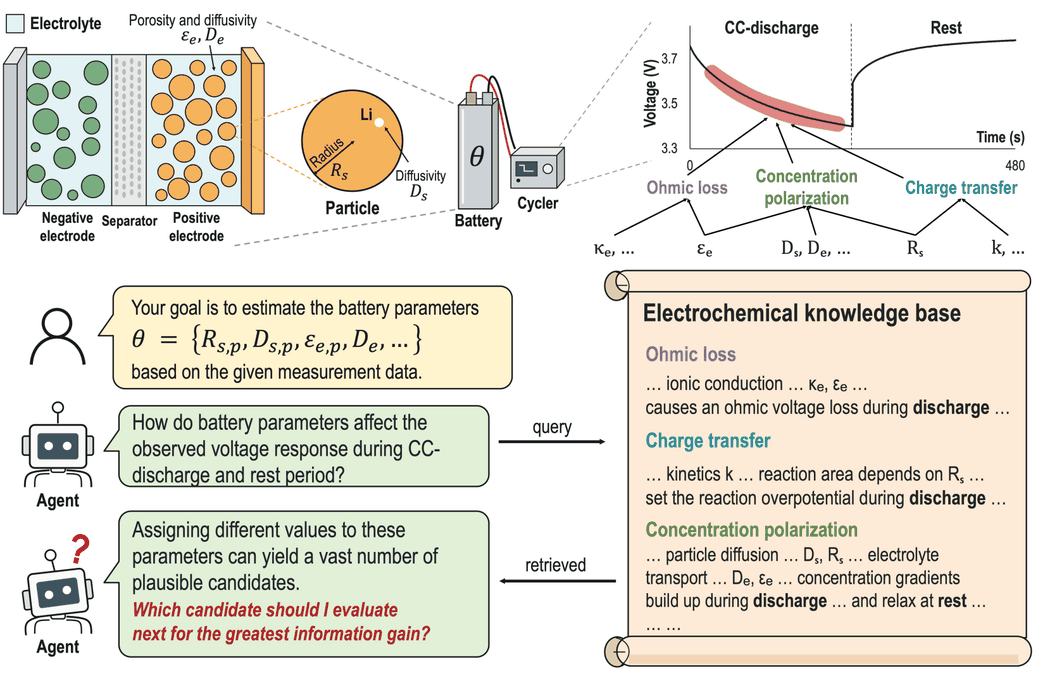}
\end{minipage}\par\smallskip
\textsf{C 4\quad O 3\quad G 4\quad A 4\quad I 4\quad S 5}\par
Six-score mean: 4.00
\end{minipage}\hfill%
\begin{minipage}[t]{0.485\textwidth}
\centering
\textbf{Study-generated revision}\par\smallskip
\begin{minipage}[t][5.6cm][t]{\linewidth}
\vspace{0pt}
\centering
\includegraphics[width=\linewidth,height=5.6cm,keepaspectratio]{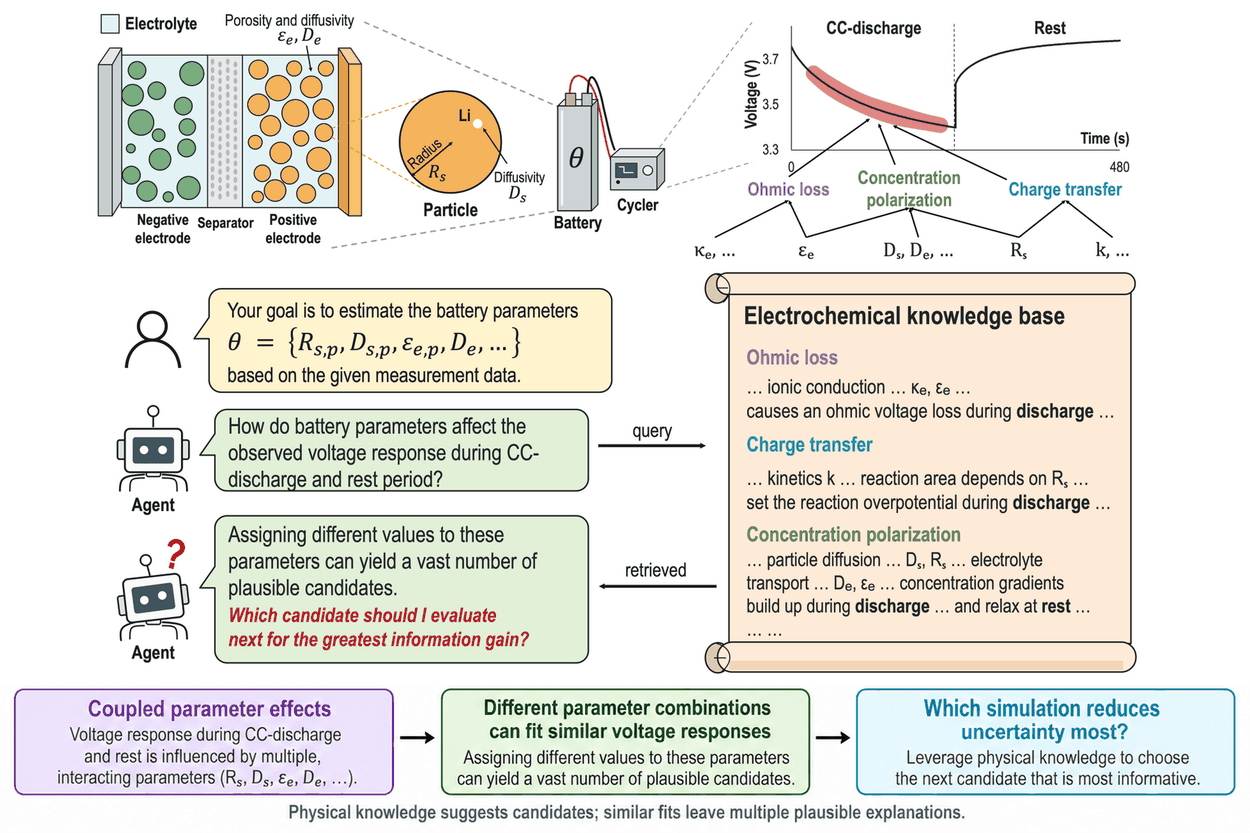}
\end{minipage}\par\smallskip
\textsf{C 4\quad O 3\quad G 4\quad A 3\quad I 3\quad S 5}\par
Six-score mean: 3.67\quad ($\Delta=-0.33$)
\end{minipage}
\par
\endgroup
\caption{Added explanatory layers in BKBAgent (167\,M). Three connected explanation boxes and a summary sentence extend the original composition. Clarity and organization remain unchanged, while aesthetic coherence and inventiveness each decrease by one point. Both images are shown in full, with three-request median scores in C/O/G/A/I/S order. Original artwork is credited to the cited paper, and the revision is a study-generated derivative.}
\label{fig:additive-explanation-case}
\vspace{-3mm}
\end{figure*}

BKBAgent illustrates the limits of adding explanation around an already elaborate visual argument (Fig.~\ref{fig:additive-explanation-case}). Its dialogue, physical mechanisms, and knowledge base already combine substantial text with diagrams. The revision appends three connected explanation boxes and a summary sentence~\citep{corpus167}. Clarity remains at 4 and organization at 3, while aesthetic coherence and inventiveness decrease from 4 to 3. The added reading layer leaves the main arrangement largely intact and produces no increase in the clarity median.

\begin{figure*}[t]
\centering
\begingroup\small
\textbf{A. Look Before You Restore: enlarged detail and a reduced overview}~\citep{corpus147}\par\smallskip
\begin{minipage}[t]{0.485\textwidth}
\centering
\textbf{Original}\par\smallskip
\begin{minipage}[t][9.6cm][t]{\linewidth}
\vspace{0pt}
\centering
\includegraphics[width=\linewidth,height=9.6cm,keepaspectratio]{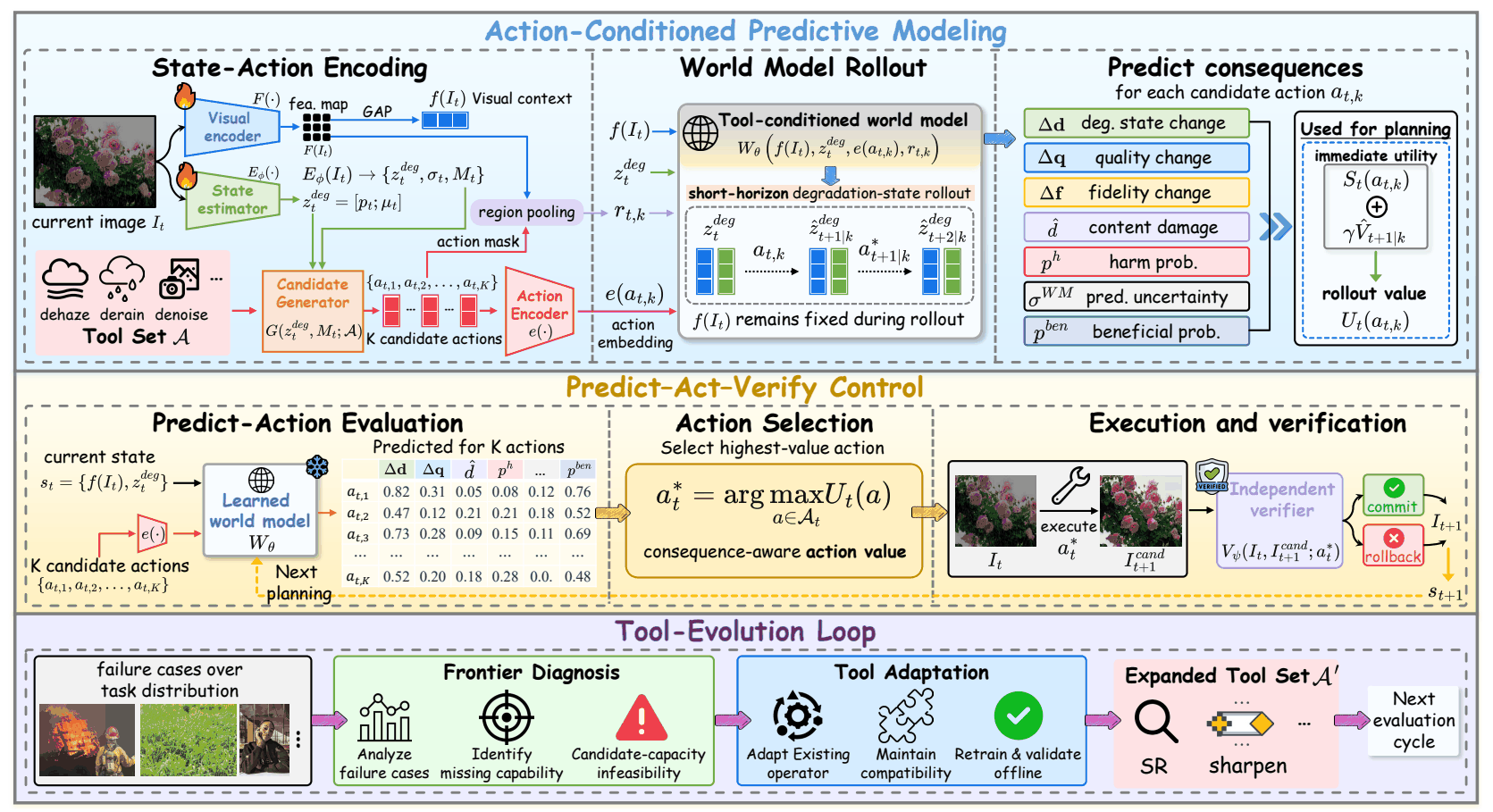}
\end{minipage}\par\smallskip
\textsf{C 4\quad O 4\quad G 4\quad A 3\quad I 3\quad S 5}\par
Six-score mean: 3.83
\end{minipage}\hfill%
\begin{minipage}[t]{0.485\textwidth}
\centering
\textbf{Study-generated revision}\par\smallskip
\begin{minipage}[t][9.6cm][t]{\linewidth}
\vspace{0pt}
\centering
\includegraphics[width=\linewidth,height=9.6cm,keepaspectratio]{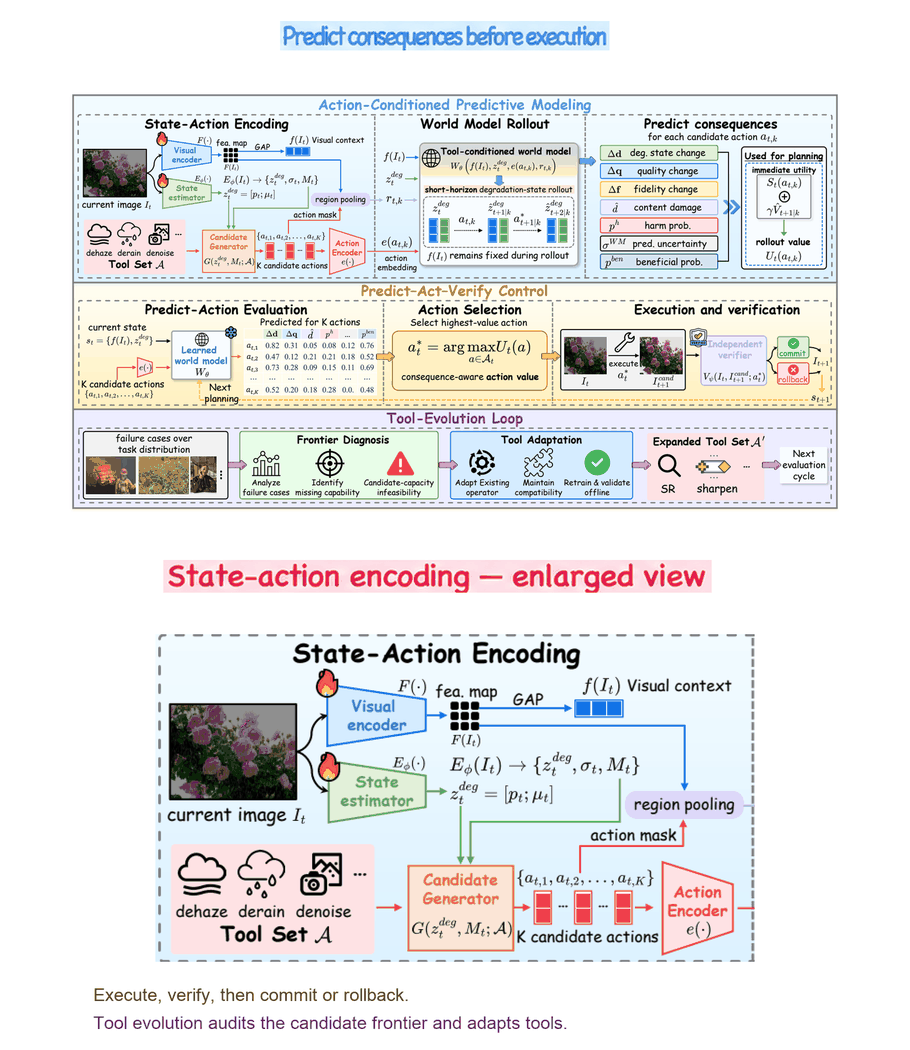}
\end{minipage}\par\smallskip
\textsf{C 4\quad O 3\quad G 4\quad A 3\quad I 3\quad S 5}\par
Six-score mean: 3.67\quad ($\Delta=-0.17$)
\end{minipage}
\par
\medskip\hrule\medskip
\textbf{B. Stride: duplicated content and weaker organization}~\citep{corpus195}\par\smallskip
\begin{minipage}[t]{0.485\textwidth}
\centering
\textbf{Original}\par\smallskip
\begin{minipage}[t][5.5cm][t]{\linewidth}
\vspace{0pt}
\centering
\includegraphics[width=\linewidth,height=5.5cm,keepaspectratio]{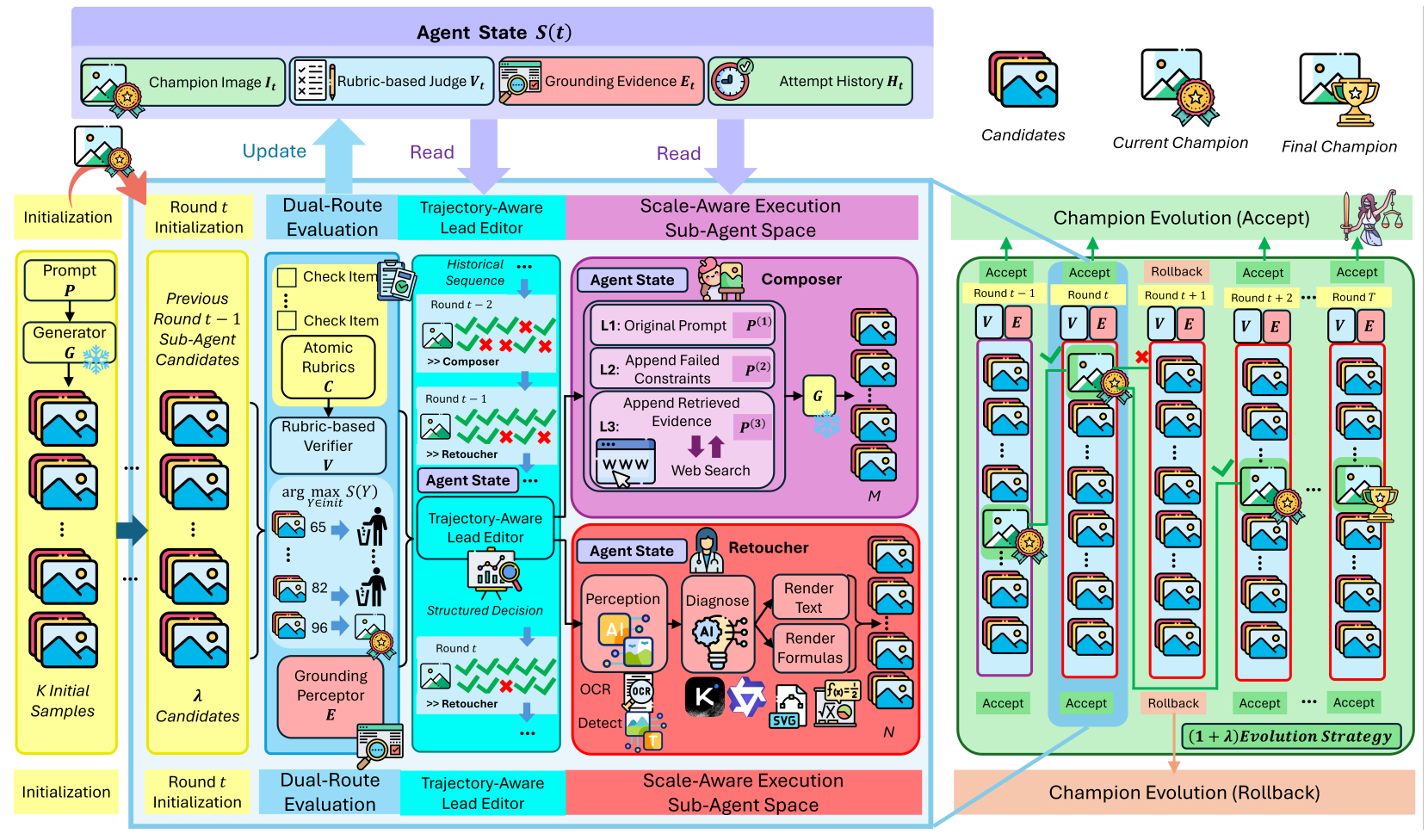}
\end{minipage}\par\smallskip
\textsf{C 3\quad O 3\quad G 3\quad A 3\quad I 4\quad S 5}\par
Six-score mean: 3.50
\end{minipage}\hfill%
\begin{minipage}[t]{0.485\textwidth}
\centering
\textbf{Study-generated revision}\par\smallskip
\begin{minipage}[t][5.5cm][t]{\linewidth}
\vspace{0pt}
\centering
\includegraphics[width=\linewidth,height=5.5cm,keepaspectratio]{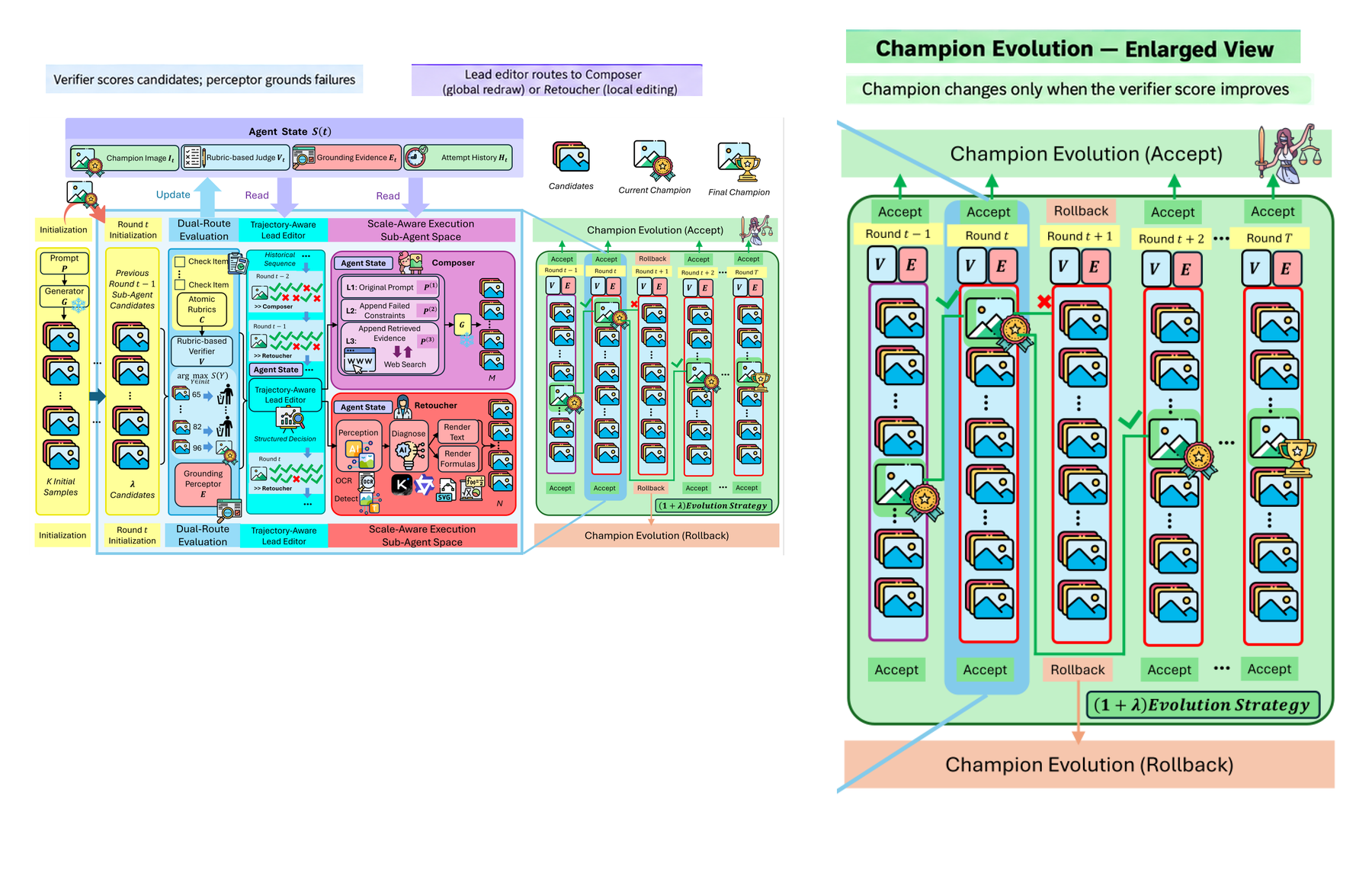}
\end{minipage}\par\smallskip
\textsf{C 3\quad O 2\quad G 3\quad A 2\quad I 3\quad S 5}\par
Six-score mean: 3.00\quad ($\Delta=-0.50$)
\end{minipage}
\par
\endgroup
\caption{Repeated content in two framework revisions (147\,F and 195\,F). Enlarged details duplicate regions retained in the complete overviews, repeating existing labels and pictograms. Organization decreases in both cases. Images are fitted individually without cropping, preserving their differing canvas proportions. Scores are three-request medians in C/O/G/A/I/S order. Original artwork is credited to the cited papers, and revisions are study-generated derivatives.}
\label{fig:duplicated-detail-cases}
\end{figure*}

Repeating an existing region creates a related challenge of coordinating detail with the overall structure (Fig.~\ref{fig:duplicated-detail-cases}). Look Before You Restore places a reduced overview above an enlarged panel for state and action encoding. The detail heading is more prominent than the overall title, and organization decreases from 4 to 3~\citep{corpus147}. Stride similarly places an enlarged champion-evolution region beside a smaller complete workflow~\citep{corpus195}. The tool, image, and award pictograms already appear in the original figures. These revisions repeat source content at different scales, making the selection and integration of detail central to their composition. The observations do not isolate typography, duplication, or scale as causes of the rating changes.

\begin{figure*}[t]
\centering
\begingroup\small
\textbf{A. SegEarth-OV3: annotations clarify the comparison}~\citep{corpus100}\par\smallskip
\begin{minipage}[t]{0.485\textwidth}
\centering
\textbf{Original}\par\smallskip
\begin{minipage}[t][5.5cm][t]{\linewidth}
\vspace{0pt}
\centering
\includegraphics[width=\linewidth,height=5.5cm,keepaspectratio]{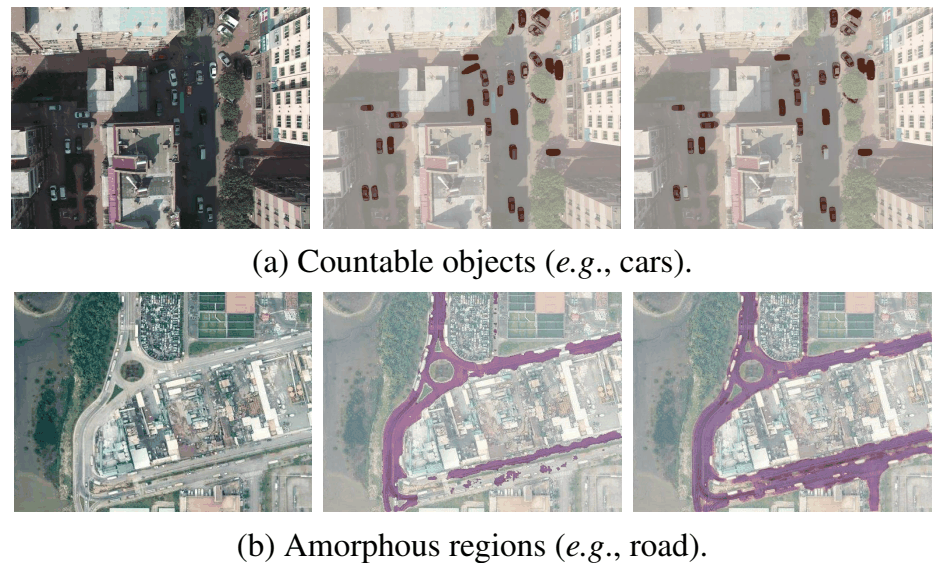}
\end{minipage}\par\smallskip
\textsf{C 3\quad O 4\quad G 4\quad A 4\quad I 3\quad S 4}\par
Six-score mean: 3.67
\end{minipage}\hfill%
\begin{minipage}[t]{0.485\textwidth}
\centering
\textbf{Study-generated revision}\par\smallskip
\begin{minipage}[t][5.5cm][t]{\linewidth}
\vspace{0pt}
\centering
\includegraphics[width=\linewidth,height=5.5cm,keepaspectratio]{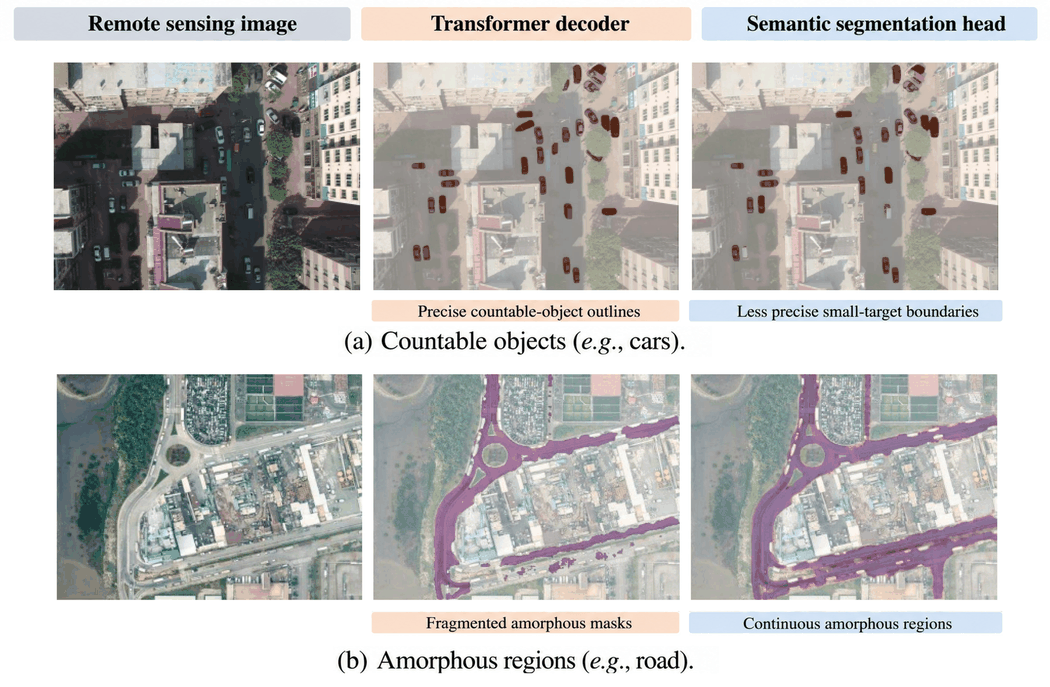}
\end{minipage}\par\smallskip
\textsf{C 4\quad O 5\quad G 5\quad A 4\quad I 4\quad S 5}\par
Six-score mean: 4.50\quad ($\Delta=+0.83$)
\end{minipage}
\par
\medskip\hrule\medskip
\textbf{B. PULSE: clarity and aesthetics move in opposite directions}~\citep{corpus127}\par\smallskip
\begin{minipage}[t]{0.485\textwidth}
\centering
\textbf{Original}\par\smallskip
\begin{minipage}[t][4.8cm][t]{\linewidth}
\vspace{0pt}
\centering
\includegraphics[width=\linewidth,height=4.8cm,keepaspectratio]{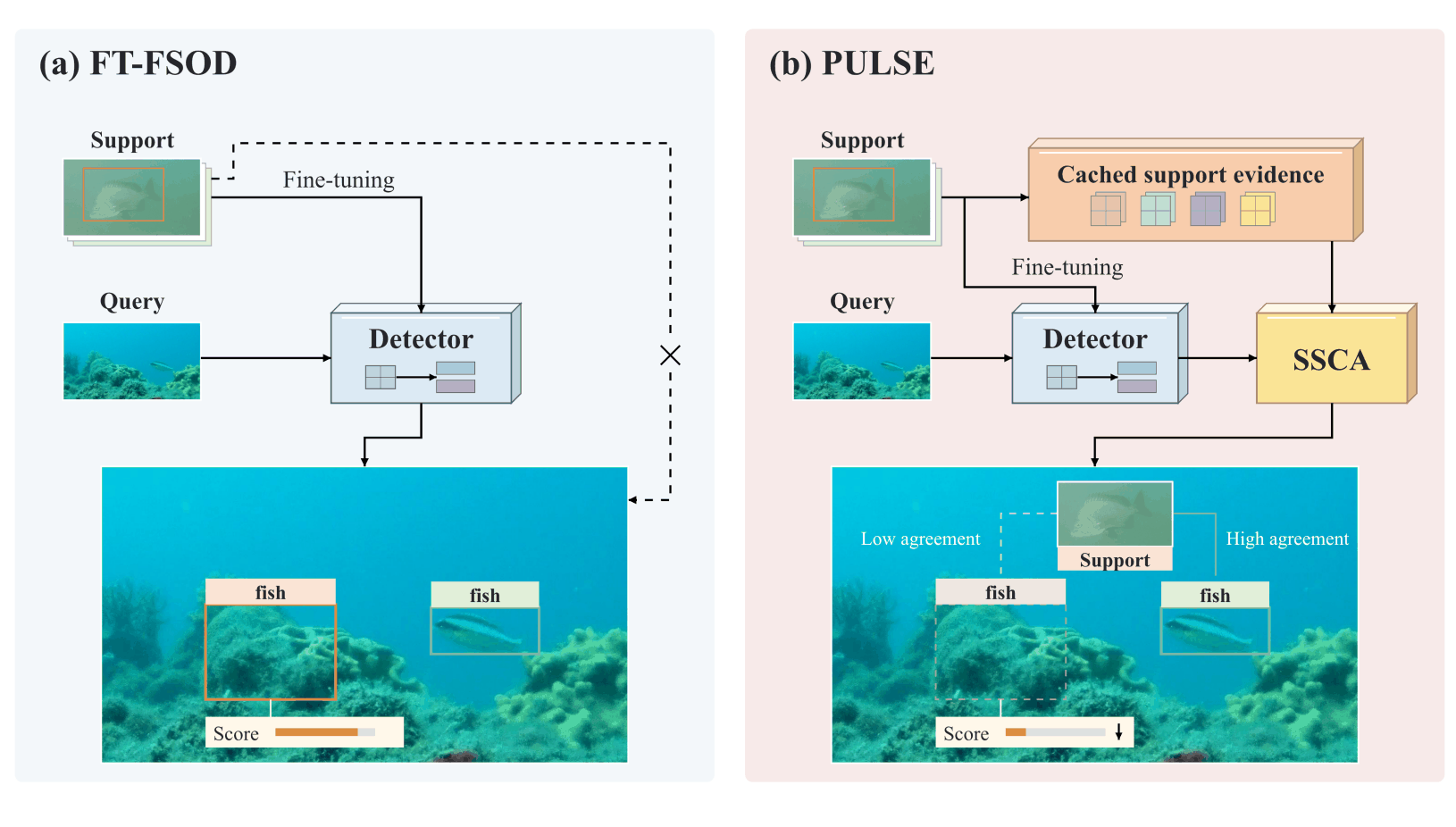}
\end{minipage}\par\smallskip
\textsf{C 4\quad O 5\quad G 4\quad A 5\quad I 4\quad S 5}\par
Six-score mean: 4.50
\end{minipage}\hfill%
\begin{minipage}[t]{0.485\textwidth}
\centering
\textbf{Study-generated revision}\par\smallskip
\begin{minipage}[t][4.8cm][t]{\linewidth}
\vspace{0pt}
\centering
\includegraphics[width=\linewidth,height=4.8cm,keepaspectratio]{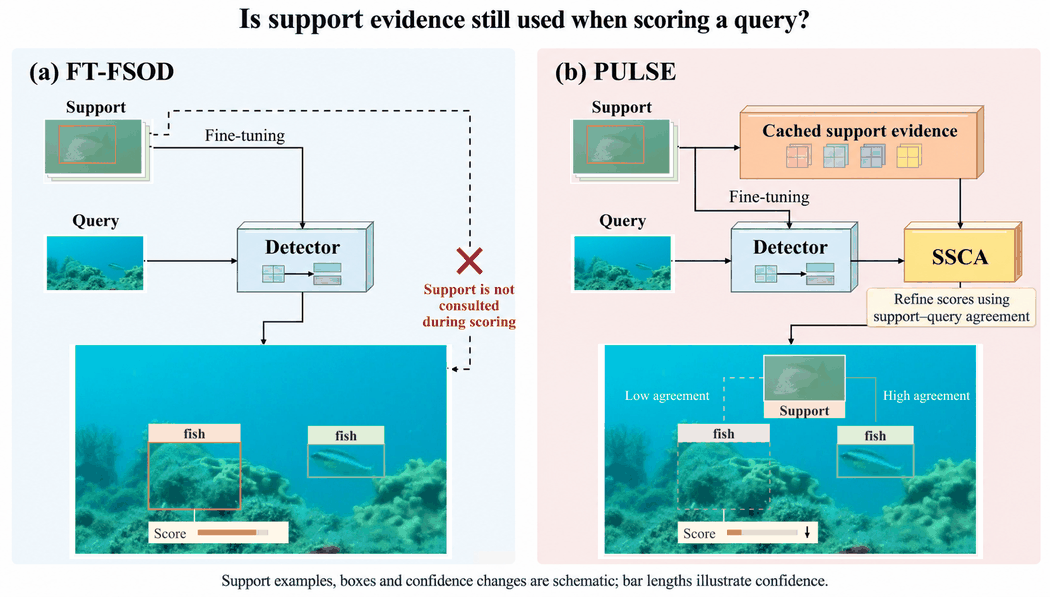}
\end{minipage}\par\smallskip
\textsf{C 5\quad O 5\quad G 4\quad A 4\quad I 4\quad S 5}\par
Six-score mean: 4.50\quad ($\Delta=0.00$)
\end{minipage}
\par
\endgroup
\caption{Contrasting annotation outcomes in motivation figures (100\,M and 127\,M). SegEarth-OV3 gains in five dimensions while retaining its aesthetic score. PULSE gains one clarity point and loses one aesthetic point, leaving the six-score mean unchanged. Images are shown in full, with three-request median scores in C/O/G/A/I/S order. Original artwork is credited to the cited papers, and revisions are study-generated derivatives.}
\label{fig:annotation-tradeoff-cases}
\end{figure*}

Annotation is more useful when it identifies a specific missing correspondence (Fig.~\ref{fig:annotation-tradeoff-cases}). SegEarth-OV3 labels the input image, transformer-decoder output, and semantic-segmentation output and explains their contrasts~\citep{corpus100}. Five dimensions increase by one point while aesthetic coherence remains at 4. PULSE adds a question headline, stronger red emphasis, and explanatory notes~\citep{corpus127}. Clarity increases from 4 to 5, but aesthetic coherence decreases from 5 to 4. These outcomes frame restraint as a functional design decision. An addition should contribute necessary meaning while fitting the surrounding visual hierarchy.

Compositional judgment remains distinct from the fidelity review described in Section~\ref{sec:revision}. The rejected \emph{Make It Up} framework changes an update relation, while the rejected EAGLE motivation figure misroutes image inputs~\citep{corpus094,corpus174}. Neither receives revision quality scores. Passing the recorded fidelity checks is a prerequisite for quality assessment, but accepted revisions can still lose organization or aesthetic coherence. The clarity and coherence of the resulting composition therefore require separate judgment.

\subsection{Reliability and Scope of Interpretation}
\label{sec:result-interpretation}

Repeated judgments show substantial agreement under the fixed protocol. All three scores agree in 1,897 of 2,226 original figure--dimension cells (85.2\%) and 1,938 of 2,214 revised cells (87.5\%). Every remaining cell spans one integer point, indicating repeat stability within the configured procedure.
The results distinguish explicit information from effective organization. Motivation figures receive larger average rating gains than frameworks, with inspected cases linking clarified comparisons to higher clarity scores. Framework revisions can retain high specificity while losing visual coherence. Added explanations may leave hierarchies unresolved or repeat detail without clarifying structure. These findings characterize the complete workflow, including generation, source-content preservation, and annotation informed by the same judge. Each revision should therefore serve the figure's explanatory purpose and preserve scientific relationships, with consolidation and omission considered alongside elaboration.
\FloatBarrier
\section{Concerns and Outlook}
\label{sec:outlook}

The observed tradeoffs raise a broader question about how scientific visual communication should be evaluated when models contribute to both its production and assessment. The following considerations connect the findings to design practice and examine possible implications for scholarly evaluation and future model development.

\textbf{I) From figure scores to reviewing decisions.}
Evidence of LLM-assisted review writing~\citep{liang2024monitoring} motivates investigating whether multimodal assistance could influence how reviewers interpret scientific figures. If reviewers consult MLLMs, revisions that receive higher figure ratings could also affect their recommendations. Such changes might reflect clearer scientific communication, sensitivity to favored visual conventions, or both. Our six dimensions, however, define an image-assessment rubric. They do not characterize the preferences of all reviewing models or establish how figures contribute to full-paper assessments. Whether figure-level gains transfer to reviewer recommendations therefore remains an untested hypothesis. Its practical significance should be examined through reader understanding and the reasons supporting a recommendation, alongside changes in numerical ratings.

\textbf{II) When evaluation criteria become design targets.}
A rubric can guide revision, but repeated optimization against its scores may favor features that an evaluator readily recognizes. Headings, labels, and familiar layouts can clarify relationships. Applied indiscriminately, they can also become formulas that receive insufficient adaptation to the figure's purpose. In this study, the same judge configuration supplied feedback for revision and evaluated the resulting figures. Such a loop may also amplify biases reported for text-based LLM judges~\citep{zheng2023judging}. Even inventiveness risks becoming formulaic if evaluators reward familiar visual cues as evidence of creativity.

\textbf{III) Visual conformity and dependence on model versions.}
A possible feedback loop connects design choices, scholarly visibility, and subsequent model development. Authors may adopt compositions favored by model-assisted assessments. These compositions could become more visible and widely imitated, then reappear in training data or tool templates that reinforce the same conventions. In the most pessimistic scenario, the literature becomes visually polished but narrower in its explanatory forms, with unfamiliar yet effective representations receiving less recognition. Reliance on a few services could also expose authors to changing preferences across model versions, prompts, and deployment settings. Design effort might increasingly be directed toward satisfying opaque evaluators, potentially favoring authors with greater resources for repeated optimization. Versioned reference collections and longitudinal audits could reveal how rankings of the same figures change as evaluators evolve.

\textbf{IV) Independent validation of communication and diversity.}
Future experiments should compare original figures, reader-oriented revisions, and evaluator-oriented revisions under matched scientific content and information quantity. Distinguishing layout changes from caption-derived explanation would clarify what a rating increase represents. Independent models could assess complete paper variants, while targeted reader studies measure recovery of intended relationships, comprehension accuracy, reading time, and preference. Held-out model families and versions would test whether gains transfer beyond the evaluator that informed revision. Comparisons between human-only and model-assisted assessments would help examine effects on reviewer recommendations. Studies of stylistic convergence additionally require explicit measures of layout, visual encodings, and spatial organization within comparable topics and figure roles. Fidelity failures and unfavorable changes should remain part of the reported outcomes.

\textbf{V) Human alignment for future self-improving systems.}
Routine model updates should not be equated with recursive self-improvement (RSI). For future systems that participate in improving their own capabilities and improvement mechanisms, progress in scientific visualization should be evaluated against evidence beyond their own scores. A constructive process would combine diverse design proposals, targeted human feedback, and independent fidelity checks. Alignment with human aesthetic judgments should accommodate differences across disciplines, expertise, cultures, and accessibility needs, including reasonable disagreement about effective expression. Optimizing a single aggregate preference score could marginalize useful but less common approaches. Evaluation should therefore consider comprehension, reading effort, and recognition of effective unfamiliar designs alongside aesthetic preference. Human evidence should also remain capable of revising the objectives that guide improvement.

\textbf{VI) Human authority as a practical design requirement.}
Researchers should retain control over a figure's explanatory purpose and composition, together with responsibility for its scientific relationships. This includes deciding when clarification requires an additional annotation and when it requires consolidation or removal. Reviewers and editors should likewise retain responsibility for judgments of validity, novelty, and significance, supported by reasons that can be examined independently of model scores. Tools should support these decisions by preserving editable sources, exposing semantic changes, and reporting uncertainty and dimensional tradeoffs. Authors must remain able to reject a highly rated revision or defend an effective composition that a model undervalues. Assistance is most useful when it makes these decisions easier to inspect, revise, and defend.

\section{Conclusion}
\label{sec:conclusion}

This study examines scientific visual communication through a documented corpus and source-constrained revisions. Within the collected sample, model ratings show no uniform historical decline. Revisions receive higher clarity ratings on average but lower aesthetic-coherence ratings, revealing tradeoffs that a single score can conceal. These findings inform our argument for researcher-led design, in which AI expands the available alternatives and authors determine what to emphasize, what to omit, and how to preserve scientific meaning. As models participate in both production and assessment, their judgments should remain accountable to human understanding. The nights recalled in our title matter for the care invested in giving an idea a visual form. As crafting figures becomes easier, the responsibility to decide what is worth showing remains ours.
\FloatBarrier
\bibliography{literature_references,corpus_references,outlook_references}

\begin{thebibliography}{214}
\providecommand{\natexlab}[1]{#1}
\providecommand{\url}[1]{\texttt{#1}}
\expandafter\ifx\csname urlstyle\endcsname\relax
  \providecommand{\doi}[1]{doi: #1}\else
  \providecommand{\doi}{doi: \begingroup \urlstyle{rm}\Url}\fi

\bibitem[Ronneberger et~al.(2015)Ronneberger, Fischer, and Brox]{ronneberger2015unet}
Olaf Ronneberger, Philipp Fischer, and Thomas Brox.
\newblock {U-Net}: Convolutional networks for biomedical image segmentation.
\newblock In \emph{Medical Image Computing and Computer-Assisted Intervention -- MICCAI 2015}, 2015.
\newblock \doi{10.1007/978-3-319-24574-4_28}.
\newblock URL \url{https://doi.org/10.1007/978-3-319-24574-4_28}.

\bibitem[Vaswani et~al.(2017)Vaswani, Shazeer, Parmar, Uszkoreit, Jones, Gomez, Kaiser, and Polosukhin]{vaswani2017attention}
Ashish Vaswani, Noam Shazeer, Niki Parmar, Jakob Uszkoreit, Llion Jones, Aidan~N. Gomez, {\L}ukasz Kaiser, and Illia Polosukhin.
\newblock Attention is all you need.
\newblock In \emph{Advances in Neural Information Processing Systems}, 2017.
\newblock URL \url{https://papers.nips.cc/paper_files/paper/2017/hash/3f5ee243547dee91fbd053c1c4a845aa-Abstract.html}.

\bibitem[Radford et~al.(2021)Radford, Kim, Hallacy, Ramesh, Goh, Agarwal, Sastry, Askell, Mishkin, Clark, Krueger, and Sutskever]{radford2021clip}
Alec Radford, Jong~Wook Kim, Chris Hallacy, Aditya Ramesh, Gabriel Goh, Sandhini Agarwal, Girish Sastry, Amanda Askell, Pamela Mishkin, Jack Clark, Gretchen Krueger, and Ilya Sutskever.
\newblock Learning transferable visual models from natural language supervision.
\newblock In \emph{Proceedings of the 38th International Conference on Machine Learning}, 2021.
\newblock URL \url{https://proceedings.mlr.press/v139/radford21a.html}.

\bibitem[Zhu et~al.(2026)Zhu, Meng, Song, Wei, Li, Pfister, and Yoon]{zhu2026paperbanana}
Dawei Zhu, Rui Meng, Yale Song, Xiyu Wei, Sujian Li, Tomas Pfister, and Jinsung Yoon.
\newblock {PaperBanana}: Automating academic illustration for {AI} scientists.
\newblock In \emph{Proceedings of the 43rd International Conference on Machine Learning}, 2026.
\newblock URL \url{https://proceedings.mlr.press/v306/zhu26ag.html}.

\bibitem[Rougier et~al.(2014)Rougier, Droettboom, and Bourne]{rougier2014figures}
Nicolas~P. Rougier, Michael Droettboom, and Philip~E. Bourne.
\newblock Ten simple rules for better figures.
\newblock \emph{PLOS Computational Biology}, 10\penalty0 (9):\penalty0 e1003833, 2014.
\newblock \doi{10.1371/journal.pcbi.1003833}.
\newblock URL \url{https://journals.plos.org/ploscompbiol/article?id=10.1371/journal.pcbi.1003833}.

\bibitem[Doshi and Hauser(2024)]{doshi2024creativity}
Anil~R. Doshi and Oliver~P. Hauser.
\newblock Generative {AI} enhances individual creativity but reduces the collective diversity of novel content.
\newblock \emph{Science Advances}, 10\penalty0 (28):\penalty0 eadn5290, 2024.
\newblock \doi{10.1126/sciadv.adn5290}.
\newblock URL \url{https://doi.org/10.1126/sciadv.adn5290}.

\bibitem[Ashkinaze et~al.(2025)Ashkinaze, Mendelsohn, Qiwei, Budak, and Gilbert]{ashkinaze2025ideas}
Joshua Ashkinaze, Julia Mendelsohn, Li~Qiwei, Ceren Budak, and Eric Gilbert.
\newblock How {AI} ideas affect the creativity, diversity, and evolution of human ideas: Evidence from a large, dynamic experiment.
\newblock In \emph{Proceedings of the ACM Collective Intelligence Conference}, 2025.
\newblock \doi{10.1145/3715928.3737481}.
\newblock URL \url{https://doi.org/10.1145/3715928.3737481}.

\bibitem[Chen et~al.(2024)Chen, Chen, Zhang, Wang, Liu, Zhou, Zhang, Wan, Zhou, and Sun]{chen2024mllmjudge}
Dongping Chen, Ruoxi Chen, Shilin Zhang, Yaochen Wang, Yinuo Liu, Huichi Zhou, Qihui Zhang, Yao Wan, Pan Zhou, and Lichao Sun.
\newblock {MLLM}-as-a-judge: Assessing multimodal {LLM}-as-a-judge with vision-language benchmark.
\newblock In \emph{Proceedings of the 41st International Conference on Machine Learning}, 2024.
\newblock URL \url{https://proceedings.mlr.press/v235/chen24h.html}.

\bibitem[{OpenAI}(2022)]{openai2022chatgpt}
{OpenAI}.
\newblock Introducing {ChatGPT}, November 2022.
\newblock URL \url{https://openai.com/index/chatgpt/}.
\newblock Published November 30, 2022; accessed November 1, 2026.

\bibitem[Sharon and Brichtova(2025)]{google2025flashimage}
David Sharon and Nicole Brichtova.
\newblock Image editing in {Gemini} just got a major upgrade, August 2025.
\newblock URL \url{https://blog.google/products-and-platforms/products/gemini/updated-image-editing-model/}.
\newblock Google. Published August 26, 2025; accessed November 1, 2026.

\bibitem[{OpenAI}(2026)]{openai2026images}
{OpenAI}.
\newblock Introducing {ChatGPT Images 2.0}, April 2026.
\newblock URL \url{https://openai.com/index/introducing-chatgpt-images-2-0/}.
\newblock Published April 21, 2026; accessed November 1, 2026.

\bibitem[{International Conference on Learning Representations}(2026)]{iclr2027dates}
{International Conference on Learning Representations}.
\newblock {ICLR} 2027 dates and deadlines, 2026.
\newblock URL \url{https://iclr.cc/Conferences/2027/Dates}.
\newblock Paper deadline September 25, 2026, Anywhere on Earth; accessed November 1, 2026.

\bibitem[Li et~al.(2022)Li, He, Li, and Zhang]{corpus001}
Shuai Li, Chenhang He, Ruihuang Li, and Lei Zhang.
\newblock {A Dual Weighting Label Assignment Scheme for Object Detection}.
\newblock In \emph{Proceedings of the IEEE/CVF Conference on Computer Vision and Pattern Recognition (CVPR)}, 2022.
\newblock URL \url{https://openaccess.thecvf.com/content/CVPR2022/html/Li_A_Dual_Weighting_Label_Assignment_Scheme_for_Object_Detection_CVPR_2022_paper.html}.

\bibitem[Zheng et~al.(2022)Zheng, Ye, Wang, Ren, Zuo, Hou, and Cheng]{corpus004}
Zhaohui Zheng, Rongguang Ye, Ping Wang, Dongwei Ren, Wangmeng Zuo, Qibin Hou, and Ming-Ming Cheng.
\newblock {Localization Distillation for Dense Object Detection}.
\newblock In \emph{Proceedings of the IEEE/CVF Conference on Computer Vision and Pattern Recognition (CVPR)}, 2022.
\newblock URL \url{https://openaccess.thecvf.com/content/CVPR2022/html/Zheng_Localization_Distillation_for_Dense_Object_Detection_CVPR_2022_paper.html}.

\bibitem[Ding et~al.(2022)Ding, Xue, Xia, and Dai]{corpus012}
Jian Ding, Nan Xue, Gui-Song Xia, and Dengxin Dai.
\newblock {Decoupling Zero-Shot Semantic Segmentation}.
\newblock In \emph{Proceedings of the IEEE/CVF Conference on Computer Vision and Pattern Recognition (CVPR)}, 2022.
\newblock URL \url{https://openaccess.thecvf.com/content/CVPR2022/html/Ding_Decoupling_Zero-Shot_Semantic_Segmentation_CVPR_2022_paper.html}.

\bibitem[Wang et~al.(2022)Wang, Feiszli, Wang, Malik, and Tran]{corpus016}
Weiyao Wang, Matt Feiszli, Heng Wang, Jitendra Malik, and Du~Tran.
\newblock {Open-World Instance Segmentation: Exploiting Pseudo Ground Truth From Learned Pairwise Affinity}.
\newblock In \emph{Proceedings of the IEEE/CVF Conference on Computer Vision and Pattern Recognition (CVPR)}, 2022.
\newblock URL \url{https://openaccess.thecvf.com/content/CVPR2022/html/Wang_Open-World_Instance_Segmentation_Exploiting_Pseudo_Ground_Truth_From_Learned_Pairwise_CVPR_2022_paper.html}.

\bibitem[Luo et~al.(2022)Luo, Huang, Yu, Li, Fan, and Liu]{corpus024}
Ziwei Luo, Haibin Huang, Lei Yu, Youwei Li, Haoqiang Fan, and Shuaicheng Liu.
\newblock {Deep Constrained Least Squares for Blind Image Super-Resolution}.
\newblock In \emph{Proceedings of the IEEE/CVF Conference on Computer Vision and Pattern Recognition (CVPR)}, 2022.
\newblock URL \url{https://openaccess.thecvf.com/content/CVPR2022/html/Luo_Deep_Constrained_Least_Squares_for_Blind_Image_Super-Resolution_CVPR_2022_paper.html}.

\bibitem[Wei et~al.(2021)Wei, Gu, Li, Timofte, Jin, and Song]{corpus030}
Yunxuan Wei, Shuhang Gu, Yawei Li, Radu Timofte, Longcun Jin, and Hengjie Song.
\newblock {Unsupervised Real-World Image Super Resolution via Domain-Distance Aware Training}.
\newblock In \emph{Proceedings of the IEEE/CVF Conference on Computer Vision and Pattern Recognition (CVPR)}, 2021.
\newblock URL \url{https://openaccess.thecvf.com/content/CVPR2021/html/Wei_Unsupervised_Real-World_Image_Super_Resolution_via_Domain-Distance_Aware_Training_CVPR_2021_paper.html}.

\bibitem[Sun et~al.(2021)Sun, Xie, Chen, Zhou, and Bao]{corpus034}
Jiaming Sun, Yiming Xie, Linghao Chen, Xiaowei Zhou, and Hujun Bao.
\newblock {NeuralRecon: Real-Time Coherent 3D Reconstruction From Monocular Video}.
\newblock In \emph{Proceedings of the IEEE/CVF Conference on Computer Vision and Pattern Recognition (CVPR)}, 2021.
\newblock URL \url{https://openaccess.thecvf.com/content/CVPR2021/html/Sun_NeuralRecon_Real-Time_Coherent_3D_Reconstruction_From_Monocular_Video_CVPR_2021_paper.html}.

\bibitem[Mescheder et~al.(2019)Mescheder, Oechsle, Niemeyer, Nowozin, and Geiger]{corpus035}
Lars Mescheder, Michael Oechsle, Michael Niemeyer, Sebastian Nowozin, and Andreas Geiger.
\newblock {Occupancy Networks: Learning 3D Reconstruction in Function Space}.
\newblock In \emph{Proceedings of the IEEE/CVF Conference on Computer Vision and Pattern Recognition (CVPR)}, 2019.
\newblock URL \url{https://openaccess.thecvf.com/content_CVPR_2019/html/Mescheder_Occupancy_Networks_Learning_3D_Reconstruction_in_Function_Space_CVPR_2019_paper.html}.

\bibitem[Wang et~al.(2025)Wang, Fu, Lu, Ge, Cao, Zhai, and Zha]{corpus042}
Kunyu Wang, Xueyang Fu, Xin Lu, Chengjie Ge, Chengzhi Cao, Wei Zhai, and Zheng-Jun Zha.
\newblock {Efficient Test-time Adaptive Object Detection via Sensitivity-Guided Pruning}.
\newblock In \emph{Proceedings of the IEEE/CVF Conference on Computer Vision and Pattern Recognition (CVPR)}, 2025.
\newblock URL \url{https://openaccess.thecvf.com/content/CVPR2025/html/Wang_Efficient_Test-time_Adaptive_Object_Detection_via_Sensitivity-Guided_Pruning_CVPR_2025_paper.html}.

\bibitem[Alexandridis et~al.(2025)Alexandridis, Elezi, Deng, Nguyen, and Luo]{corpus043}
Konstantinos~Panagiotis Alexandridis, Ismail Elezi, Jiankang Deng, Anh Nguyen, and Shan Luo.
\newblock {Fractal Calibration for Long-tailed Object Detection}.
\newblock In \emph{Proceedings of the IEEE/CVF Conference on Computer Vision and Pattern Recognition (CVPR)}, 2025.
\newblock URL \url{https://openaccess.thecvf.com/content/CVPR2025/html/Alexandridis_Fractal_Calibration_for_Long-tailed_Object_Detection_CVPR_2025_paper.html}.

\bibitem[Yin et~al.(2025)Yin, Cao, Cheng, and Hou]{corpus052}
Bo-Wen Yin, Jiao-Long Cao, Ming-Ming Cheng, and Qibin Hou.
\newblock {DFormerv2: Geometry Self-Attention for RGBD Semantic Segmentation}.
\newblock In \emph{Proceedings of the IEEE/CVF Conference on Computer Vision and Pattern Recognition (CVPR)}, 2025.
\newblock URL \url{https://openaccess.thecvf.com/content/CVPR2025/html/Yin_DFormerv2_Geometry_Self-Attention_for_RGBD_Semantic_Segmentation_CVPR_2025_paper.html}.

\bibitem[Kim et~al.(2025)Kim, Ju, Han, Yang, and Hwang]{corpus053}
Chanyoung Kim, Dayun Ju, Woojung Han, Ming-Hsuan Yang, and Seong~Jae Hwang.
\newblock {Distilling Spectral Graph for Object-Context Aware Open-Vocabulary Semantic Segmentation}.
\newblock In \emph{Proceedings of the IEEE/CVF Conference on Computer Vision and Pattern Recognition (CVPR)}, 2025.
\newblock URL \url{https://openaccess.thecvf.com/content/CVPR2025/html/Kim_Distilling_Spectral_Graph_for_Object-Context_Aware_Open-Vocabulary_Semantic_Segmentation_CVPR_2025_paper.html}.

\bibitem[Zamfir et~al.(2025)Zamfir, Wu, Mehta, Tan, Paudel, Zhang, and Timofte]{corpus062}
Eduard Zamfir, Zongwei Wu, Nancy Mehta, Yuedong Tan, Danda~Pani Paudel, Yulun Zhang, and Radu Timofte.
\newblock {Complexity Experts are Task-Discriminative Learners for Any Image Restoration}.
\newblock In \emph{Proceedings of the IEEE/CVF Conference on Computer Vision and Pattern Recognition (CVPR)}, 2025.
\newblock URL \url{https://openaccess.thecvf.com/content/CVPR2025/html/Zamfir_Complexity_Experts_are_Task-Discriminative_Learners_for_Any_Image_Restoration_CVPR_2025_paper.html}.

\bibitem[Di et~al.(2025)Di, Peng, Xia, Li, Pei, Cao, Wang, and Zha]{corpus068}
Xin Di, Long Peng, Peizhe Xia, Wenbo Li, Renjing Pei, Yang Cao, Yang Wang, and Zheng-Jun Zha.
\newblock {QMambaBSR: Burst Image Super-Resolution with Query State Space Model}.
\newblock In \emph{Proceedings of the IEEE/CVF Conference on Computer Vision and Pattern Recognition (CVPR)}, 2025.
\newblock URL \url{https://openaccess.thecvf.com/content/CVPR2025/html/Di_QMambaBSR_Burst_Image_Super-Resolution_with_Query_State_Space_Model_CVPR_2025_paper.html}.

\bibitem[Walker et~al.(2025)Walker, Esposito, Rebain, Vaxman, Onken, Li, and Mac~Aodha]{corpus073}
Thomas Walker, Salvatore Esposito, Daniel Rebain, Amir Vaxman, Arno Onken, Changjian Li, and Oisin Mac~Aodha.
\newblock {CrossSDF: 3D Reconstruction of Thin Structures From Cross-Sections}.
\newblock In \emph{Proceedings of the IEEE/CVF Conference on Computer Vision and Pattern Recognition (CVPR)}, 2025.
\newblock URL \url{https://openaccess.thecvf.com/content/CVPR2025/html/Walker_CrossSDF_3D_Reconstruction_of_Thin_Structures_From_Cross-Sections_CVPR_2025_paper.html}.

\bibitem[Chen et~al.(2025)Chen, Wang, Nan, and Zhu]{corpus078}
Zhaiyu Chen, Yuqing Wang, Liangliang Nan, and Xiao~Xiang Zhu.
\newblock {Parametric Point Cloud Completion for Polygonal Surface Reconstruction}.
\newblock In \emph{Proceedings of the IEEE/CVF Conference on Computer Vision and Pattern Recognition (CVPR)}, 2025.
\newblock URL \url{https://openaccess.thecvf.com/content/CVPR2025/html/Chen_Parametric_Point_Cloud_Completion_for_Polygonal_Surface_Reconstruction_CVPR_2025_paper.html}.

\bibitem[Yu et~al.(2026)Yu, Deng, Duan, and Li]{corpus085}
Hengfu Yu, Jinhong Deng, Lixin Duan, and Wen Li.
\newblock {Instance-Free Domain Adaptive Object Detection}.
\newblock arXiv preprint, 2026.
\newblock URL \url{https://arxiv.org/abs/2602.06484}.

\bibitem[Zhang et~al.(2026)Zhang, Chen, Zhang, and Tao]{corpus087}
Xu~Zhang, Zhe Chen, Jing Zhang, and Dacheng Tao.
\newblock {LAB-Det: Language as a Domain-Invariant Bridge for Training-Free One-Shot Domain Generalization in Object Detection}.
\newblock arXiv preprint, 2026.
\newblock URL \url{https://arxiv.org/abs/2602.06474}.

\bibitem[Ernst et~al.(2026)Ernst, Boettcher, Hoyer, Lenssen, and Schiele]{corpus098}
Jonas Ernst, Wolfgang Boettcher, Lukas Hoyer, Jan~Eric Lenssen, and Bernt Schiele.
\newblock {Rewis3d: Reconstruction Improves Weakly-Supervised Semantic Segmentation}.
\newblock In \emph{Proceedings of the IEEE/CVF Conference on Computer Vision and Pattern Recognition (CVPR)}, 2026.
\newblock URL \url{https://openaccess.thecvf.com/content/CVPR2026/html/Ernst_Rewis3d_Reconstruction_Improves_Weakly-Supervised_Semantic_Segmentation_CVPR_2026_paper.html}.

\bibitem[Li et~al.(2025)Li, Zhang, Wang, Deng, Wang, Meng, and Cao]{corpus100}
Kaiyu Li, Shengqi Zhang, Yujie Wang, Yupeng Deng, Zhi Wang, Deyu Meng, and Xiangyong Cao.
\newblock {SegEarth-OV3: Exploring SAM 3 for Open-Vocabulary Semantic Segmentation in Remote Sensing Images}.
\newblock arXiv preprint, 2025.
\newblock URL \url{https://arxiv.org/abs/2512.08730}.

\bibitem[Zhang et~al.(2026)Zhang, Zhang, Nie, and Fu]{corpus102}
Yingkai Zhang, Tao Zhang, Jing Nie, and Ying Fu.
\newblock {Enhancing Unregistered Hyperspectral Image Super-Resolution via Unmixing-based Abundance Fusion Learning}.
\newblock In \emph{Proceedings of the IEEE/CVF Conference on Computer Vision and Pattern Recognition (CVPR)}, 2026.
\newblock URL \url{https://openaccess.thecvf.com/content/CVPR2026/html/Zhang_Enhancing_Unregistered_Hyperspectral_Image_Super-Resolution_via_Unmixing-based_Abundance_Fusion_Learning_CVPR_2026_paper.html}.

\bibitem[Li et~al.(2026)Li, Zhang, Chen, Xie, Wang, Zhang, Li, zhang, Zhang, and Zhao]{corpus110}
Weiqi Li, Xuanyu Zhang, Bin Chen, Jingfen Xie, Yan Wang, Kexin Zhang, Junlin Li, Li~zhang, Jian Zhang, and Shijie Zhao.
\newblock {UARE: A Unified Vision-Language Model for Image Quality Assessment, Restoration, and Enhancement}.
\newblock In \emph{Proceedings of the IEEE/CVF Conference on Computer Vision and Pattern Recognition (CVPR)}, 2026.
\newblock URL \url{https://openaccess.thecvf.com/content/CVPR2026/html/Li_UARE_A_Unified_Vision-Language_Model_for_Image_Quality_Assessment_Restoration_CVPR_2026_paper.html}.

\bibitem[Zhao et~al.(2026)Zhao, Tan, Wang, Bi, Zhang, Sunkavalli, Tulsiani, and Jiang]{corpus112}
Qitao Zhao, Hao Tan, Qianqian Wang, Sai Bi, Kai Zhang, Kalyan Sunkavalli, Shubham Tulsiani, and Hanwen Jiang.
\newblock {E-RayZer: Self-supervised 3D Reconstruction as Spatial Visual Pre-training}.
\newblock In \emph{Proceedings of the IEEE/CVF Conference on Computer Vision and Pattern Recognition (CVPR)}, 2026.
\newblock URL \url{https://openaccess.thecvf.com/content/CVPR2026/html/Zhao_E-RayZer_Self-supervised_3D_Reconstruction_as_Spatial_Visual_Pre-training_CVPR_2026_paper.html}.

\bibitem[Wang et~al.(2026)Wang, Xing, Zhao, Sebe, and Lepri]{corpus117}
Weijie Wang, Songlong Xing, Zhengyu Zhao, Nicu Sebe, and Bruno Lepri.
\newblock {PoInit-of-View: Poisoning Initialization of Views Transfers Across Multiple 3D Reconstruction Systems}.
\newblock In \emph{Proceedings of the IEEE/CVF Conference on Computer Vision and Pattern Recognition (CVPR)}, 2026.
\newblock URL \url{https://openaccess.thecvf.com/content/CVPR2026/html/Wang_PoInit-of-View_Poisoning_Initialization_of_Views_Transfers_Across_Multiple_3D_Reconstruction_CVPR_2026_paper.html}.

\bibitem[Kang et~al.(2026)Kang, Lee, and Lee]{corpus121}
Woochul Kang, Hyungseop Lee, and Jiho Lee.
\newblock {AnyDepth-DETR/-YOLO: Any-depth object detection with a single network}.
\newblock Manuscript submitted to the International Conference on Learning Representations (ICLR 2027), 2026.
\newblock URL \url{https://openreview.net/forum?id=8qFCNFz8Tq}.

\bibitem[Anonymous(2026)]{corpus127}
Anonymous.
\newblock {PULSE: PERSISTENT UTILIZATION OF LIMITED SUPPORT EVIDENCE FOR CROSS-DOMAIN FEW-SHOT OBJECT DETECTION}.
\newblock Manuscript submitted to the International Conference on Learning Representations (ICLR 2027), 2026.
\newblock URL \url{https://openreview.net/forum?id=ad0Os8QbxN}.

\bibitem[Anonymous(2026)]{corpus132}
Anonymous.
\newblock {Calibration-Guided State Selection and Class-Conditioned Fusion for Frozen Diffusion Segmentation}.
\newblock Manuscript submitted to the International Conference on Learning Representations (ICLR 2027), 2026.
\newblock URL \url{https://openreview.net/forum?id=NYrrKvi2Dy}.

\bibitem[Anonymous(2026)]{corpus134}
Anonymous.
\newblock {HIVE: Holistic Image-Adaptive Vocabulary Evolution for Training-Free Dense Vision-Language Inference}.
\newblock Manuscript submitted to the International Conference on Learning Representations (ICLR 2027), 2026.
\newblock URL \url{https://openreview.net/forum?id=kv0BeIZGMk}.

\bibitem[Anonymous(2026)]{corpus143}
Anonymous.
\newblock {Fixing the Last Mile: Failure-Aware Image Restoration with Region-Specific Guidance}.
\newblock Manuscript submitted to the International Conference on Learning Representations (ICLR 2027), 2026.
\newblock URL \url{https://openreview.net/forum?id=kQkih2eMHu}.

\bibitem[Zhang et~al.(2026)Zhang, Wan, Zhang, and Zhang]{corpus147}
Xu~Zhang, Jun Wan, Lefei Zhang, and Liangpei Zhang.
\newblock {Look Before You Restore: Learning to Predict Before Acting for Agentic Image Restoration}.
\newblock Manuscript submitted to the International Conference on Learning Representations (ICLR 2027), 2026.
\newblock URL \url{https://openreview.net/forum?id=rCgxaWvBRj}.

\bibitem[Xie et~al.(2026)Xie, Wu, Zhang, Wang, and Liu]{corpus155}
Zhoufeng Xie, Guanjun Wu, Shuai Zhang, Xinggang Wang, and Wenyu Liu.
\newblock {LonGaussian: Tracking Foundation Priors for Long-Context Feedforward Gaussian Reconstruction}.
\newblock Manuscript submitted to the International Conference on Learning Representations (ICLR 2027), 2026.
\newblock URL \url{https://openreview.net/forum?id=WiNCsgDP25}.

\bibitem[Wu et~al.(2026)Wu, Kong, Chen, Zhang, and Zhang]{corpus160}
Junyi Wu, Fanqing Kong, Leyang Chen, Shaoqiu Zhang, and Yulun Zhang.
\newblock {VASC: Value-Aware Sparse Attention with Cross-layer Memory for Efficient 3D Reconstruction}.
\newblock Manuscript submitted to the International Conference on Learning Representations (ICLR 2027), 2026.
\newblock URL \url{https://openreview.net/forum?id=X7FI3N35Hq}.

\bibitem[Anonymous(2026)]{corpus162}
Anonymous.
\newblock {A Simulation-Grounded VLM Framework for Agentic Wildfire Monitoring and Reporting}.
\newblock Manuscript submitted to the International Conference on Learning Representations (ICLR 2027), 2026.
\newblock URL \url{https://openreview.net/forum?id=gGGjmjPJi4}.

\bibitem[Anonymous(2026)]{corpus164}
Anonymous.
\newblock {Agent Skills Under Change: A Lifecycle Benchmark and Coordinate-Guided Updates}.
\newblock Manuscript submitted to the International Conference on Learning Representations (ICLR 2027), 2026.
\newblock URL \url{https://openreview.net/forum?id=O7Vvd7kb57}.

\bibitem[Anonymous(2026)]{corpus167}
Anonymous.
\newblock {BKBAgent: A Battery Parameter Identification Agent with Bayesian Experimental Design and Knowledge Graph}.
\newblock Manuscript submitted to the International Conference on Learning Representations (ICLR 2027), 2026.
\newblock URL \url{https://openreview.net/forum?id=whcg806KGZ}.

\bibitem[Anonymous(2026)]{corpus184}
Anonymous.
\newblock {Look Back to Move Forward: Learning to Revisit for Agentic Compositional Image Generation}.
\newblock Manuscript submitted to the International Conference on Learning Representations (ICLR 2027), 2026.
\newblock URL \url{https://openreview.net/forum?id=C4VFESNXLV}.

\bibitem[Anonymous(2026)]{corpus190}
Anonymous.
\newblock {Ontology-Harnessing Multi-Agent System for Abstractive Visual Understanding}.
\newblock Manuscript submitted to the International Conference on Learning Representations (ICLR 2027), 2026.
\newblock URL \url{https://openreview.net/forum?id=jVkkfpmnAw}.

\bibitem[Anonymous(2026)]{corpus191}
Anonymous.
\newblock {OpenAgentFlow: Enabling System-Wide Safety Boundaries for Heterogeneous AI Agent Fleets}.
\newblock Manuscript submitted to the International Conference on Learning Representations (ICLR 2027), 2026.
\newblock URL \url{https://openreview.net/forum?id=I6Gx5US2sY}.

\bibitem[Anonymous(2026)]{corpus194}
Anonymous.
\newblock {Sprout: Building Dynamic Memory While Reasoning for Agentic Video Understanding}.
\newblock Manuscript submitted to the International Conference on Learning Representations (ICLR 2027), 2026.
\newblock URL \url{https://openreview.net/forum?id=VA0yCrShah}.

\bibitem[Anonymous(2026)]{corpus198}
Anonymous.
\newblock {VEvo-Harness: Causally-Guarded Self-Evolving Harness for Agentic Image Generation}.
\newblock Manuscript submitted to the International Conference on Learning Representations (ICLR 2027), 2026.
\newblock URL \url{https://openreview.net/forum?id=UPFHZRl39C}.

\bibitem[Anonymous(2026)]{corpus163}
Anonymous.
\newblock {AccessWorld: A Benchmark for Computer Use Agents Operating Assistive Technologies}.
\newblock Manuscript submitted to the International Conference on Learning Representations (ICLR 2027), 2026.
\newblock URL \url{https://openreview.net/forum?id=uLkUAjKWe6}.

\bibitem[Anonymous(2026)]{corpus128}
Anonymous.
\newblock {Rethinking Over-Segmentation in Few-Shot Object Detection with Vision Foundation Models: A Training-Free Prototype Refinement and Occlusion-Aware Discrimination Approach}.
\newblock Manuscript submitted to the International Conference on Learning Representations (ICLR 2027), 2026.
\newblock URL \url{https://openreview.net/forum?id=V5RmhtXTMG}.

\bibitem[Anonymous(2026)]{corpus136}
Anonymous.
\newblock {ReTax-Seg: Visually Grounded Reversible Taxonomy Transport for Image Segmentation}.
\newblock Manuscript submitted to the International Conference on Learning Representations (ICLR 2027), 2026.
\newblock URL \url{https://openreview.net/forum?id=dMBwWu68Lm}.

\bibitem[Anonymous(2026)]{corpus195}
Anonymous.
\newblock {Stride: Scale-Aware Multi-Agent Evolutionary Refinement for Infographic Generation}.
\newblock Manuscript submitted to the International Conference on Learning Representations (ICLR 2027), 2026.
\newblock URL \url{https://openreview.net/forum?id=qT1Thc0dst}.

\bibitem[Xie et~al.(2026)Xie, He, Fan, Zhang, Cheng, and Liu]{corpus094}
Guohuan Xie, Xin He, Dingying Fan, Le~Zhang, Ming-Ming Cheng, and Yun Liu.
\newblock {Make It Up: Fake Images, Real Gains in Generalized Few-shot Semantic Segmentation}.
\newblock arXiv preprint, 2026.
\newblock URL \url{https://arxiv.org/abs/2603.27206}.

\bibitem[Anonymous(2026)]{corpus174}
Anonymous.
\newblock {EAGLE: Grounding Multimodal LLM Judges for AI-Generated Image Assessment via Evidence Anchors}.
\newblock Manuscript submitted to the International Conference on Learning Representations (ICLR 2027), 2026.
\newblock URL \url{https://openreview.net/forum?id=Qask3lw679}.

\bibitem[Liang et~al.(2024)Liang, Izzo, Zhang, Lepp, Cao, Zhao, Chen, Ye, Liu, Huang, Mcfarland, and Zou]{liang2024monitoring}
Weixin Liang, Zachary Izzo, Yaohui Zhang, Haley Lepp, Hancheng Cao, Xuandong Zhao, Lingjiao Chen, Haotian Ye, Sheng Liu, Zhi Huang, Daniel Mcfarland, and James~Y. Zou.
\newblock Monitoring {AI}-modified content at scale: A case study on the impact of {ChatGPT} on {AI} conference peer reviews.
\newblock In \emph{Proceedings of the 41st International Conference on Machine Learning}, 2024.
\newblock URL \url{https://proceedings.mlr.press/v235/liang24b.html}.

\bibitem[Zheng et~al.(2023)Zheng, Chiang, Sheng, Zhuang, Wu, Zhuang, Lin, Li, Li, Xing, Zhang, Gonzalez, and Stoica]{zheng2023judging}
Lianmin Zheng, Wei-Lin Chiang, Ying Sheng, Siyuan Zhuang, Zhanghao Wu, Yonghao Zhuang, Zi~Lin, Zhuohan Li, Dacheng Li, Eric~P. Xing, Hao Zhang, Joseph~E. Gonzalez, and Ion Stoica.
\newblock Judging {LLM}-as-a-judge with {MT-Bench} and {Chatbot Arena}.
\newblock In \emph{Advances in Neural Information Processing Systems}, 2023.
\newblock \doi{10.52202/075280-2020}.
\newblock URL \url{https://papers.neurips.cc/paper_files/paper/2023/hash/91f18a1287b398d378ef22505bf41832-Abstract-Datasets_and_Benchmarks.html}.

\bibitem[He et~al.(2022)He, Wang, Wu, Wang, Li, Li, Gan, Wu, and Qiao]{corpus002}
Mengzhe He, Yali Wang, Jiaxi Wu, Yiru Wang, Hanqing Li, Bo~Li, Weihao Gan, Wei Wu, and Yu~Qiao.
\newblock {Cross Domain Object Detection by Target-Perceived Dual Branch Distillation}.
\newblock In \emph{Proceedings of the IEEE/CVF Conference on Computer Vision and Pattern Recognition (CVPR)}, 2022.
\newblock URL \url{https://openaccess.thecvf.com/content/CVPR2022/html/He_Cross_Domain_Object_Detection_by_Target-Perceived_Dual_Branch_Distillation_CVPR_2022_paper.html}.

\bibitem[Chen et~al.(2022)Chen, Li, Chen, Wang, Zhang, and Hua]{corpus003}
Binghui Chen, Pengyu Li, Xiang Chen, Biao Wang, Lei Zhang, and Xian-Sheng Hua.
\newblock {Dense Learning Based Semi-Supervised Object Detection}.
\newblock In \emph{Proceedings of the IEEE/CVF Conference on Computer Vision and Pattern Recognition (CVPR)}, 2022.
\newblock URL \url{https://openaccess.thecvf.com/content/CVPR2022/html/Chen_Dense_Learning_Based_Semi-Supervised_Object_Detection_CVPR_2022_paper.html}.

\bibitem[Elezi et~al.(2022)Elezi, Yu, Anandkumar, Leal-Taix\'e, and Alvarez]{corpus005}
Ismail Elezi, Zhiding Yu, Anima Anandkumar, Laura Leal-Taix\'e, and Jose~M. Alvarez.
\newblock {Not All Labels Are Equal: Rationalizing the Labeling Costs for Training Object Detection}.
\newblock In \emph{Proceedings of the IEEE/CVF Conference on Computer Vision and Pattern Recognition (CVPR)}, 2022.
\newblock URL \url{https://openaccess.thecvf.com/content/CVPR2022/html/Elezi_Not_All_Labels_Are_Equal_Rationalizing_the_Labeling_Costs_for_CVPR_2022_paper.html}.

\bibitem[Zheng et~al.(2022)Zheng, Zhang, Zhang, Qi, and Sun]{corpus006}
Anlin Zheng, Yuang Zhang, Xiangyu Zhang, Xiaojuan Qi, and Jian Sun.
\newblock {Progressive End-to-End Object Detection in Crowded Scenes}.
\newblock In \emph{Proceedings of the IEEE/CVF Conference on Computer Vision and Pattern Recognition (CVPR)}, 2022.
\newblock URL \url{https://openaccess.thecvf.com/content/CVPR2022/html/Zheng_Progressive_End-to-End_Object_Detection_in_Crowded_Scenes_CVPR_2022_paper.html}.

\bibitem[Guo et~al.(2022)Guo, Mu, Chen, Wang, Yu, and Luo]{corpus007}
Qiushan Guo, Yao Mu, Jianyu Chen, Tianqi Wang, Yizhou Yu, and Ping Luo.
\newblock {Scale-Equivalent Distillation for Semi-Supervised Object Detection}.
\newblock In \emph{Proceedings of the IEEE/CVF Conference on Computer Vision and Pattern Recognition (CVPR)}, 2022.
\newblock URL \url{https://openaccess.thecvf.com/content/CVPR2022/html/Guo_Scale-Equivalent_Distillation_for_Semi-Supervised_Object_Detection_CVPR_2022_paper.html}.

\bibitem[Li et~al.(2022)Li, Pan, Yan, Tang, and Zheng]{corpus008}
Hanjun Li, Xingjia Pan, Ke~Yan, Fan Tang, and Wei-Shi Zheng.
\newblock {SIOD: Single Instance Annotated per Category per Image for Object Detection}.
\newblock In \emph{Proceedings of the IEEE/CVF Conference on Computer Vision and Pattern Recognition (CVPR)}, 2022.
\newblock URL \url{https://openaccess.thecvf.com/content/CVPR2022/html/Li_SIOD_Single_Instance_Annotated_per_Category_per_Image_for_Object_CVPR_2022_paper.html}.

\bibitem[Zhao and Wang(2022)]{corpus009}
Liang Zhao and Limin Wang.
\newblock {Task-Specific Inconsistency Alignment for Domain Adaptive Object Detection}.
\newblock In \emph{Proceedings of the IEEE/CVF Conference on Computer Vision and Pattern Recognition (CVPR)}, 2022.
\newblock URL \url{https://openaccess.thecvf.com/content/CVPR2022/html/Zhao_Task-Specific_Inconsistency_Alignment_for_Domain_Adaptive_Object_Detection_CVPR_2022_paper.html}.

\bibitem[Liu et~al.(2022)Liu, Li, Yang, Li, and Yuan]{corpus010}
Xinyu Liu, Wuyang Li, Qiushi Yang, Baopu Li, and Yixuan Yuan.
\newblock {Towards Robust Adaptive Object Detection Under Noisy Annotations}.
\newblock In \emph{Proceedings of the IEEE/CVF Conference on Computer Vision and Pattern Recognition (CVPR)}, 2022.
\newblock URL \url{https://openaccess.thecvf.com/content/CVPR2022/html/Liu_Towards_Robust_Adaptive_Object_Detection_Under_Noisy_Annotations_CVPR_2022_paper.html}.

\bibitem[Li et~al.(2022)Li, Li, He, Zhang, Jia, and Zhang]{corpus011}
Ruihuang Li, Shuai Li, Chenhang He, Yabin Zhang, Xu~Jia, and Lei Zhang.
\newblock {Class-Balanced Pixel-Level Self-Labeling for Domain Adaptive Semantic Segmentation}.
\newblock In \emph{Proceedings of the IEEE/CVF Conference on Computer Vision and Pattern Recognition (CVPR)}, 2022.
\newblock URL \url{https://openaccess.thecvf.com/content/CVPR2022/html/Li_Class-Balanced_Pixel-Level_Self-Labeling_for_Domain_Adaptive_Semantic_Segmentation_CVPR_2022_paper.html}.

\bibitem[Ru et~al.(2022)Ru, Zhan, Yu, and Du]{corpus013}
Lixiang Ru, Yibing Zhan, Baosheng Yu, and Bo~Du.
\newblock {Learning Affinity From Attention: End-to-End Weakly-Supervised Semantic Segmentation With Transformers}.
\newblock In \emph{Proceedings of the IEEE/CVF Conference on Computer Vision and Pattern Recognition (CVPR)}, 2022.
\newblock URL \url{https://openaccess.thecvf.com/content/CVPR2022/html/Ru_Learning_Affinity_From_Attention_End-to-End_Weakly-Supervised_Semantic_Segmentation_With_Transformers_CVPR_2022_paper.html}.

\bibitem[Xu et~al.(2022)Xu, Ouyang, Bennamoun, Boussaid, and Xu]{corpus014}
Lian Xu, Wanli Ouyang, Mohammed Bennamoun, Farid Boussaid, and Dan Xu.
\newblock {Multi-Class Token Transformer for Weakly Supervised Semantic Segmentation}.
\newblock In \emph{Proceedings of the IEEE/CVF Conference on Computer Vision and Pattern Recognition (CVPR)}, 2022.
\newblock URL \url{https://openaccess.thecvf.com/content/CVPR2022/html/Xu_Multi-Class_Token_Transformer_for_Weakly_Supervised_Semantic_Segmentation_CVPR_2022_paper.html}.

\bibitem[Zhao et~al.(2022)Zhao, Zhong, Sebe, and Lee]{corpus015}
Yuyang Zhao, Zhun Zhong, Nicu Sebe, and Gim~Hee Lee.
\newblock {Novel Class Discovery in Semantic Segmentation}.
\newblock In \emph{Proceedings of the IEEE/CVF Conference on Computer Vision and Pattern Recognition (CVPR)}, 2022.
\newblock URL \url{https://openaccess.thecvf.com/content/CVPR2022/html/Zhao_Novel_Class_Discovery_in_Semantic_Segmentation_CVPR_2022_paper.html}.

\bibitem[Zhang et~al.(2022)Zhang, Xiao, Liu, Chen, and Cheng]{corpus017}
Chang-Bin Zhang, Jia-Wen Xiao, Xialei Liu, Ying-Cong Chen, and Ming-Ming Cheng.
\newblock {Representation Compensation Networks for Continual Semantic Segmentation}.
\newblock In \emph{Proceedings of the IEEE/CVF Conference on Computer Vision and Pattern Recognition (CVPR)}, 2022.
\newblock URL \url{https://openaccess.thecvf.com/content/CVPR2022/html/Zhang_Representation_Compensation_Networks_for_Continual_Semantic_Segmentation_CVPR_2022_paper.html}.

\bibitem[Zhou et~al.(2022)Zhou, Wang, Konukoglu, and Van~Gool]{corpus018}
Tianfei Zhou, Wenguan Wang, Ender Konukoglu, and Luc Van~Gool.
\newblock {Rethinking Semantic Segmentation: A Prototype View}.
\newblock In \emph{Proceedings of the IEEE/CVF Conference on Computer Vision and Pattern Recognition (CVPR)}, 2022.
\newblock URL \url{https://openaccess.thecvf.com/content/CVPR2022/html/Zhou_Rethinking_Semantic_Segmentation_A_Prototype_View_CVPR_2022_paper.html}.

\bibitem[Guo et~al.(2022)Guo, Liu, Liu, and Yuan]{corpus019}
Xiaoqing Guo, Jie Liu, Tongliang Liu, and Yixuan Yuan.
\newblock {SimT: Handling Open-Set Noise for Domain Adaptive Semantic Segmentation}.
\newblock In \emph{Proceedings of the IEEE/CVF Conference on Computer Vision and Pattern Recognition (CVPR)}, 2022.
\newblock URL \url{https://openaccess.thecvf.com/content/CVPR2022/html/Guo_SimT_Handling_Open-Set_Noise_for_Domain_Adaptive_Semantic_Segmentation_CVPR_2022_paper.html}.

\bibitem[Liang et~al.(2022)Liang, Wang, Zhang, Sun, and Shen]{corpus020}
Zhiyuan Liang, Tiancai Wang, Xiangyu Zhang, Jian Sun, and Jianbing Shen.
\newblock {Tree Energy Loss: Towards Sparsely Annotated Semantic Segmentation}.
\newblock In \emph{Proceedings of the IEEE/CVF Conference on Computer Vision and Pattern Recognition (CVPR)}, 2022.
\newblock URL \url{https://openaccess.thecvf.com/content/CVPR2022/html/Liang_Tree_Energy_Loss_Towards_Sparsely_Annotated_Semantic_Segmentation_CVPR_2022_paper.html}.

\bibitem[Ma et~al.(2022)Ma, Liang, and Zhang]{corpus021}
Jianqi Ma, Zhetong Liang, and Lei Zhang.
\newblock {A Text Attention Network for Spatial Deformation Robust Scene Text Image Super-Resolution}.
\newblock In \emph{Proceedings of the IEEE/CVF Conference on Computer Vision and Pattern Recognition (CVPR)}, 2022.
\newblock URL \url{https://openaccess.thecvf.com/content/CVPR2022/html/Ma_A_Text_Attention_Network_for_Spatial_Deformation_Robust_Scene_Text_CVPR_2022_paper.html}.

\bibitem[Yue et~al.(2022)Yue, Zhao, Xie, Zhang, Meng, and Wong]{corpus022}
Zongsheng Yue, Qian Zhao, Jianwen Xie, Lei Zhang, Deyu Meng, and Kwan-Yee~K. Wong.
\newblock {Blind Image Super-Resolution With Elaborate Degradation Modeling on Noise and Kernel}.
\newblock In \emph{Proceedings of the IEEE/CVF Conference on Computer Vision and Pattern Recognition (CVPR)}, 2022.
\newblock URL \url{https://openaccess.thecvf.com/content/CVPR2022/html/Yue_Blind_Image_Super-Resolution_With_Elaborate_Degradation_Modeling_on_Noise_and_CVPR_2022_paper.html}.

\bibitem[Whang et~al.(2022)Whang, Delbracio, Talebi, Saharia, Dimakis, and Milanfar]{corpus023}
Jay Whang, Mauricio Delbracio, Hossein Talebi, Chitwan Saharia, Alexandros~G. Dimakis, and Peyman Milanfar.
\newblock {Deblurring via Stochastic Refinement}.
\newblock In \emph{Proceedings of the IEEE/CVF Conference on Computer Vision and Pattern Recognition (CVPR)}, 2022.
\newblock URL \url{https://openaccess.thecvf.com/content/CVPR2022/html/Whang_Deblurring_via_Stochastic_Refinement_CVPR_2022_paper.html}.

\bibitem[Liang et~al.(2022)Liang, Zeng, and Zhang]{corpus025}
Jie Liang, Hui Zeng, and Lei Zhang.
\newblock {Details or Artifacts: A Locally Discriminative Learning Approach to Realistic Image Super-Resolution}.
\newblock In \emph{Proceedings of the IEEE/CVF Conference on Computer Vision and Pattern Recognition (CVPR)}, 2022.
\newblock URL \url{https://openaccess.thecvf.com/content/CVPR2022/html/Liang_Details_or_Artifacts_A_Locally_Discriminative_Learning_Approach_to_Realistic_CVPR_2022_paper.html}.

\bibitem[Xu et~al.(2022)Xu, Wei, Chen, Liu, Mao, Lin, and Li]{corpus026}
Xiaoqian Xu, Pengxu Wei, Weikai Chen, Yang Liu, Mingzhi Mao, Liang Lin, and Guanbin Li.
\newblock {Dual Adversarial Adaptation for Cross-Device Real-World Image Super-Resolution}.
\newblock In \emph{Proceedings of the IEEE/CVF Conference on Computer Vision and Pattern Recognition (CVPR)}, 2022.
\newblock URL \url{https://openaccess.thecvf.com/content/CVPR2022/html/Xu_Dual_Adversarial_Adaptation_for_Cross-Device_Real-World_Image_Super-Resolution_CVPR_2022_paper.html}.

\bibitem[Pan et~al.(2021)Pan, Chowdhury, Hartley, Liu, Zhang, and Li]{corpus027}
Liyuan Pan, Shah Chowdhury, Richard Hartley, Miaomiao Liu, Hongguang Zhang, and Hongdong Li.
\newblock {Dual Pixel Exploration: Simultaneous Depth Estimation and Image Restoration}.
\newblock In \emph{Proceedings of the IEEE/CVF Conference on Computer Vision and Pattern Recognition (CVPR)}, 2021.
\newblock URL \url{https://openaccess.thecvf.com/content/CVPR2021/html/Pan_Dual_Pixel_Exploration_Simultaneous_Depth_Estimation_and_Image_Restoration_CVPR_2021_paper.html}.

\bibitem[Zhang et~al.(2022)Zhang, Su, Zhu, Yan, Sun, and Zhang]{corpus028}
Cheng Zhang, Shaolin Su, Yu~Zhu, Qingsen Yan, Jinqiu Sun, and Yanning Zhang.
\newblock {Exploring and Evaluating Image Restoration Potential in Dynamic Scenes}.
\newblock In \emph{Proceedings of the IEEE/CVF Conference on Computer Vision and Pattern Recognition (CVPR)}, 2022.
\newblock URL \url{https://openaccess.thecvf.com/content/CVPR2022/html/Zhang_Exploring_and_Evaluating_Image_Restoration_Potential_in_Dynamic_Scenes_CVPR_2022_paper.html}.

\bibitem[Son and Lee(2021)]{corpus029}
Sanghyun Son and Kyoung~Mu Lee.
\newblock {SRWarp: Generalized Image Super-Resolution under Arbitrary Transformation}.
\newblock In \emph{Proceedings of the IEEE/CVF Conference on Computer Vision and Pattern Recognition (CVPR)}, 2021.
\newblock URL \url{https://openaccess.thecvf.com/content/CVPR2021/html/Son_SRWarp_Generalized_Image_Super-Resolution_under_Arbitrary_Transformation_CVPR_2021_paper.html}.

\bibitem[Bechtold et~al.(2021)Bechtold, Tatarchenko, Fischer, and Brox]{corpus031}
Jan Bechtold, Maxim Tatarchenko, Volker Fischer, and Thomas Brox.
\newblock {Fostering Generalization in Single-View 3D Reconstruction by Learning a Hierarchy of Local and Global Shape Priors}.
\newblock In \emph{Proceedings of the IEEE/CVF Conference on Computer Vision and Pattern Recognition (CVPR)}, 2021.
\newblock URL \url{https://openaccess.thecvf.com/content/CVPR2021/html/Bechtold_Fostering_Generalization_in_Single-View_3D_Reconstruction_by_Learning_a_Hierarchy_CVPR_2021_paper.html}.

\bibitem[Engelmann et~al.(2021)Engelmann, Rematas, Leibe, and Ferrari]{corpus032}
Francis Engelmann, Konstantinos Rematas, Bastian Leibe, and Vittorio Ferrari.
\newblock {From Points to Multi-Object 3D Reconstruction}.
\newblock In \emph{Proceedings of the IEEE/CVF Conference on Computer Vision and Pattern Recognition (CVPR)}, 2021.
\newblock URL \url{https://openaccess.thecvf.com/content/CVPR2021/html/Engelmann_From_Points_to_Multi-Object_3D_Reconstruction_CVPR_2021_paper.html}.

\bibitem[Kato and Harada(2019)]{corpus033}
Hiroharu Kato and Tatsuya Harada.
\newblock {Learning View Priors for Single-View 3D Reconstruction}.
\newblock In \emph{Proceedings of the IEEE/CVF Conference on Computer Vision and Pattern Recognition (CVPR)}, 2019.
\newblock URL \url{https://openaccess.thecvf.com/content_CVPR_2019/html/Kato_Learning_View_Priors_for_Single-View_3D_Reconstruction_CVPR_2019_paper.html}.

\bibitem[Alldieck et~al.(2022)Alldieck, Zanfir, and Sminchisescu]{corpus036}
Thiemo Alldieck, Mihai Zanfir, and Cristian Sminchisescu.
\newblock {Photorealistic Monocular 3D Reconstruction of Humans Wearing Clothing}.
\newblock In \emph{Proceedings of the IEEE/CVF Conference on Computer Vision and Pattern Recognition (CVPR)}, 2022.
\newblock URL \url{https://openaccess.thecvf.com/content/CVPR2022/html/Alldieck_Photorealistic_Monocular_3D_Reconstruction_of_Humans_Wearing_Clothing_CVPR_2022_paper.html}.

\bibitem[Alwala et~al.(2022)Alwala, Gupta, and Tulsiani]{corpus037}
Kalyan~Vasudev Alwala, Abhinav Gupta, and Shubham Tulsiani.
\newblock {Pre-Train, Self-Train, Distill: A Simple Recipe for Supersizing 3D Reconstruction}.
\newblock In \emph{Proceedings of the IEEE/CVF Conference on Computer Vision and Pattern Recognition (CVPR)}, 2022.
\newblock URL \url{https://openaccess.thecvf.com/content/CVPR2022/html/Alwala_Pre-Train_Self-Train_Distill_A_Simple_Recipe_for_Supersizing_3D_Reconstruction_CVPR_2022_paper.html}.

\bibitem[Yu et~al.(2019)Yu, Zheng, Lian, Zhou, and Gao]{corpus038}
Zehao Yu, Jia Zheng, Dongze Lian, Zihan Zhou, and Shenghua Gao.
\newblock {Single-Image Piece-Wise Planar 3D Reconstruction via Associative Embedding}.
\newblock In \emph{Proceedings of the IEEE/CVF Conference on Computer Vision and Pattern Recognition (CVPR)}, 2019.
\newblock URL \url{https://openaccess.thecvf.com/content_CVPR_2019/html/Yu_Single-Image_Piece-Wise_Planar_3D_Reconstruction_via_Associative_Embedding_CVPR_2019_paper.html}.

\bibitem[Ma et~al.(2022)Ma, Liu, Zwicker, and Han]{corpus039}
Baorui Ma, Yu-Shen Liu, Matthias Zwicker, and Zhizhong Han.
\newblock {Surface Reconstruction From Point Clouds by Learning Predictive Context Priors}.
\newblock In \emph{Proceedings of the IEEE/CVF Conference on Computer Vision and Pattern Recognition (CVPR)}, 2022.
\newblock URL \url{https://openaccess.thecvf.com/content/CVPR2022/html/Ma_Surface_Reconstruction_From_Point_Clouds_by_Learning_Predictive_Context_Priors_CVPR_2022_paper.html}.

\bibitem[Duggal and Pathak(2022)]{corpus040}
Shivam Duggal and Deepak Pathak.
\newblock {Topologically-Aware Deformation Fields for Single-View 3D Reconstruction}.
\newblock In \emph{Proceedings of the IEEE/CVF Conference on Computer Vision and Pattern Recognition (CVPR)}, 2022.
\newblock \doi{10.1109/cvpr52688.2022.00159}.
\newblock URL \url{https://doi.org/10.1109/cvpr52688.2022.00159}.

\bibitem[Kennerley et~al.(2024)Kennerley, Wang, Veeravalli, and Tan]{corpus041}
Mikhail Kennerley, Jian-Gang Wang, Bharadwaj Veeravalli, and Robby~T. Tan.
\newblock {CAT: Exploiting Inter-Class Dynamics for Domain Adaptive Object Detection}.
\newblock In \emph{Proceedings of the IEEE/CVF Conference on Computer Vision and Pattern Recognition (CVPR)}, 2024.
\newblock URL \url{https://openaccess.thecvf.com/content/CVPR2024/html/Kennerley_CAT_Exploiting_Inter-Class_Dynamics_for_Domain_Adaptive_Object_Detection_CVPR_2024_paper.html}.

\bibitem[Lin et~al.(2024)Lin, Jiang, Qu, Yuan, and Cai]{corpus044}
Chuang Lin, Yi~Jiang, Lizhen Qu, Zehuan Yuan, and Jianfei Cai.
\newblock {Generative Region-Language Pretraining for Open-Ended Object Detection}.
\newblock In \emph{Proceedings of the IEEE/CVF Conference on Computer Vision and Pattern Recognition (CVPR)}, 2024.
\newblock URL \url{https://openaccess.thecvf.com/content/CVPR2024/html/Lin_Generative_Region-Language_Pretraining_for_Open-Ended_Object_Detection_CVPR_2024_paper.html}.

\bibitem[Feng et~al.(2024)Feng, Zhong, Jie, Xie, and Ma]{corpus045}
Chengjian Feng, Yujie Zhong, Zequn Jie, Weidi Xie, and Lin Ma.
\newblock {InstaGen: Enhancing Object Detection by Training on Synthetic Dataset}.
\newblock In \emph{Proceedings of the IEEE/CVF Conference on Computer Vision and Pattern Recognition (CVPR)}, 2024.
\newblock URL \url{https://openaccess.thecvf.com/content/CVPR2024/html/Feng_InstaGen_Enhancing_Object_Detection_by_Training_on_Synthetic_Dataset_CVPR_2024_paper.html}.

\bibitem[Lavoie et~al.(2025)Lavoie, Mahmoud, and Waslander]{corpus046}
Marc-Antoine Lavoie, Anas Mahmoud, and Steven~L. Waslander.
\newblock {Large Self-Supervised Models Bridge the Gap in Domain Adaptive Object Detection}.
\newblock In \emph{Proceedings of the IEEE/CVF Conference on Computer Vision and Pattern Recognition (CVPR)}, 2025.
\newblock URL \url{https://openaccess.thecvf.com/content/CVPR2025/html/Lavoie_Large_Self-Supervised_Models_Bridge_the_Gap_in_Domain_Adaptive_Object_CVPR_2025_paper.html}.

\bibitem[Fu et~al.(2025)Fu, Yang, Mo, Yan, Wei, Meng, Xie, and Zheng]{corpus047}
Shenghao Fu, Qize Yang, Qijie Mo, Junkai Yan, Xihan Wei, Jingke Meng, Xiaohua Xie, and Wei-Shi Zheng.
\newblock {LLMDet: Learning Strong Open-Vocabulary Object Detectors under the Supervision of Large Language Models}.
\newblock In \emph{Proceedings of the IEEE/CVF Conference on Computer Vision and Pattern Recognition (CVPR)}, 2025.
\newblock URL \url{https://openaccess.thecvf.com/content/CVPR2025/html/Fu_LLMDet_Learning_Strong_Open-Vocabulary_Object_Detectors_under_the_Supervision_of_CVPR_2025_paper.html}.

\bibitem[Liu et~al.(2026)Liu, Zhang, Liu, Zhang, Zhou, Liao, Gong, Luo, Wang, Yu, and Yang]{corpus048}
Mingxin Liu, Peiyuan Zhang, Yuan Liu, Wei Zhang, Yue Zhou, Ning Liao, Ziyang Gong, Junwei Luo, Zhirui Wang, Yi~Yu, and Xue Yang.
\newblock {Partial Weakly-Supervised Oriented Object Detection}.
\newblock In \emph{Proceedings of the IEEE/CVF Conference on Computer Vision and Pattern Recognition (CVPR)}, 2026.
\newblock URL \url{https://openaccess.thecvf.com/content/CVPR2026/html/Liu_Partial_Weakly-Supervised_Oriented_Object_Detection_CVPR_2026_paper.html}.

\bibitem[Luo et~al.(2024)Luo, Yang, Yu, Li, Yan, and Li]{corpus049}
Junwei Luo, Xue Yang, Yi~Yu, Qingyun Li, Junchi Yan, and Yansheng Li.
\newblock {PointOBB: Learning Oriented Object Detection via Single Point Supervision}.
\newblock In \emph{Proceedings of the IEEE/CVF Conference on Computer Vision and Pattern Recognition (CVPR)}, 2024.
\newblock URL \url{https://openaccess.thecvf.com/content/CVPR2024/html/Luo_PointOBB_Learning_Oriented_Object_Detection_via_Single_Point_Supervision_CVPR_2024_paper.html}.

\bibitem[Sidhu et~al.(2025)Sidhu, Chopra, Blume, Kim, Reddy, and Ji]{corpus050}
Mankeerat Sidhu, Hetarth Chopra, Ansel Blume, Jeonghwan Kim, Revanth~Gangi Reddy, and Heng Ji.
\newblock {Search and Detect: Training-Free Long Tail Object Detection via Web-Image Retrieval}.
\newblock In \emph{Proceedings of the IEEE/CVF Conference on Computer Vision and Pattern Recognition (CVPR)}, 2025.
\newblock URL \url{https://openaccess.thecvf.com/content/CVPR2025/html/Sidhu_Search_and_Detect_Training-Free_Long_Tail_Object_Detection_via_Web-Image_CVPR_2025_paper.html}.

\bibitem[Yin et~al.(2025)Yin, Feng, Lyu, Shang, Liu, Feng, and Wan]{corpus051}
Hongmei Yin, Tingliang Feng, Fan Lyu, Fanhua Shang, Hongying Liu, Wei Feng, and Liang Wan.
\newblock {Beyond Background Shift: Rethinking Instance Replay in Continual Semantic Segmentation}.
\newblock In \emph{Proceedings of the IEEE/CVF Conference on Computer Vision and Pattern Recognition (CVPR)}, 2025.
\newblock URL \url{https://openaccess.thecvf.com/content/CVPR2025/html/Yin_Beyond_Background_Shift_Rethinking_Instance_Replay_in_Continual_Semantic_Segmentation_CVPR_2025_paper.html}.

\bibitem[Zhao et~al.(2025)Zhao, Li, Shi, Imanpour, and Wang]{corpus054}
Ziyu Zhao, Xiaoguang Li, Lingjia Shi, Nasrin Imanpour, and Song Wang.
\newblock {DPSeg: Dual-Prompt Cost Volume Learning for Open-Vocabulary Semantic Segmentation}.
\newblock In \emph{Proceedings of the IEEE/CVF Conference on Computer Vision and Pattern Recognition (CVPR)}, 2025.
\newblock URL \url{https://openaccess.thecvf.com/content/CVPR2025/html/Zhao_DPSeg_Dual-Prompt_Cost_Volume_Learning_for_Open-Vocabulary_Semantic_Segmentation_CVPR_2025_paper.html}.

\bibitem[Zhang et~al.(2024)Zhang, Yu, Wei, Zhao, and Xiao]{corpus055}
Bingfeng Zhang, Siyue Yu, Yunchao Wei, Yao Zhao, and Jimin Xiao.
\newblock {Frozen CLIP: A Strong Backbone for Weakly Supervised Semantic Segmentation}.
\newblock In \emph{Proceedings of the IEEE/CVF Conference on Computer Vision and Pattern Recognition (CVPR)}, 2024.
\newblock URL \url{https://openaccess.thecvf.com/content/CVPR2024/html/Zhang_Frozen_CLIP_A_Strong_Backbone_for_Weakly_Supervised_Semantic_Segmentation_CVPR_2024_paper.html}.

\bibitem[Tang et~al.(2024)Tang, Xu, Qu, Feng, Jiang, and Ge]{corpus056}
Feilong Tang, Zhongxing Xu, Zhaojun Qu, Wei Feng, Xingjian Jiang, and Zongyuan Ge.
\newblock {Hunting Attributes: Context Prototype-Aware Learning for Weakly Supervised Semantic Segmentation}.
\newblock In \emph{Proceedings of the IEEE/CVF Conference on Computer Vision and Pattern Recognition (CVPR)}, 2024.
\newblock URL \url{https://openaccess.thecvf.com/content/CVPR2024/html/Tang_Hunting_Attributes_Context_Prototype-Aware_Learning_for_Weakly_Supervised_Semantic_Segmentation_CVPR_2024_paper.html}.

\bibitem[Wang et~al.(2024)Wang, Sun, Luo, Pan, and Zhang]{corpus057}
Yuan Wang, Rui Sun, Naisong Luo, Yuwen Pan, and Tianzhu Zhang.
\newblock {Image-to-Image Matching via Foundation Models: A New Perspective for Open-Vocabulary Semantic Segmentation}.
\newblock In \emph{Proceedings of the IEEE/CVF Conference on Computer Vision and Pattern Recognition (CVPR)}, 2024.
\newblock URL \url{https://openaccess.thecvf.com/content/CVPR2024/html/Wang_Image-to-Image_Matching_via_Foundation_Models_A_New_Perspective_for_Open-Vocabulary_CVPR_2024_paper.html}.

\bibitem[Zhang and Tan(2025)]{corpus058}
Xin Zhang and Robby~T. Tan.
\newblock {Mamba as a Bridge: Where Vision Foundation Models Meet Vision Language Models for Domain-Generalized Semantic Segmentation}.
\newblock In \emph{Proceedings of the Computer Vision and Pattern Recognition Conference (CVPR)}, 2025.
\newblock URL \url{https://openaccess.thecvf.com/content/CVPR2025/html/Zhang_Mamba_as_a_Bridge_Where_Vision_Foundation_Models_Meet_Vision_CVPR_2025_paper.html}.

\bibitem[Xie et~al.(2025)Xie, Li, Zeng, Luo, and Zhang]{corpus059}
Chenxi Xie, Minghan Li, Hui Zeng, Jun Luo, and Lei Zhang.
\newblock {MaSS13K: A Matting-level Semantic Segmentation Benchmark}.
\newblock In \emph{Proceedings of the Computer Vision and Pattern Recognition Conference (CVPR)}, 2025.
\newblock URL \url{https://openaccess.thecvf.com/content/CVPR2025/html/Xie_MaSS13K_A_Matting-level_Semantic_Segmentation_Benchmark_CVPR_2025_paper.html}.

\bibitem[Fu et~al.(2025)Fu, Lou, and Yu]{corpus060}
Yunxiang Fu, Meng Lou, and Yizhou Yu.
\newblock {SegMAN: Omni-scale Context Modeling with State Space Models and Local Attention for Semantic Segmentation}.
\newblock In \emph{Proceedings of the Computer Vision and Pattern Recognition Conference (CVPR)}, 2025.
\newblock URL \url{https://openaccess.thecvf.com/content/CVPR2025/html/Fu_SegMAN_Omni-scale_Context_Modeling_with_State_Space_Models_and_Local_CVPR_2025_paper.html}.

\bibitem[Chen et~al.(2025)Chen, Li, Wu, Zhang, Chen, Zhang, and Zhang]{corpus061}
Bin Chen, Gehui Li, Rongyuan Wu, Xindong Zhang, Jie Chen, Jian Zhang, and Lei Zhang.
\newblock {Adversarial Diffusion Compression for Real-World Image Super-Resolution}.
\newblock In \emph{Proceedings of the IEEE/CVF Conference on Computer Vision and Pattern Recognition (CVPR)}, 2025.
\newblock URL \url{https://openaccess.thecvf.com/content/CVPR2025/html/Chen_Adversarial_Diffusion_Compression_for_Real-World_Image_Super-Resolution_CVPR_2025_paper.html}.

\bibitem[Wang et~al.(2025)Wang, Yan, Pan, Gao, Zhang, and Yang]{corpus063}
Zhengxue Wang, Zhiqiang Yan, Jinshan Pan, Guangwei Gao, Kai Zhang, and Jian Yang.
\newblock {DORNet: A Degradation Oriented and Regularized Network for Blind Depth Super-Resolution}.
\newblock In \emph{Proceedings of the IEEE/CVF Conference on Computer Vision and Pattern Recognition (CVPR)}, 2025.
\newblock URL \url{https://openaccess.thecvf.com/content/CVPR2025/html/Wang_DORNet_A_Degradation_Oriented_and_Regularized_Network_for_Blind_Depth_CVPR_2025_paper.html}.

\bibitem[Chen et~al.(2025)Chen, Pan, and Dong]{corpus064}
Junyang Chen, Jinshan Pan, and Jiangxin Dong.
\newblock {FaithDiff: Unleashing Diffusion Priors for Faithful Image Super-resolution}.
\newblock In \emph{Proceedings of the IEEE/CVF Conference on Computer Vision and Pattern Recognition (CVPR)}, 2025.
\newblock URL \url{https://openaccess.thecvf.com/content/CVPR2025/html/Chen_FaithDiff_Unleashing_Diffusion_Priors_for_Faithful_Image_Super-resolution_CVPR_2025_paper.html}.

\bibitem[Zhao et~al.(2025)Zhao, Chen, Xu, Gu, Li, Yi, Wang, Yang, and Tai]{corpus065}
Chen Zhao, Zhizhou Chen, Yunzhe Xu, Enxuan Gu, Jian Li, Zili Yi, Qian Wang, Jian Yang, and Ying Tai.
\newblock {From Zero to Detail: Deconstructing Ultra-High-Definition Image Restoration from Progressive Spectral Perspective}.
\newblock In \emph{Proceedings of the IEEE/CVF Conference on Computer Vision and Pattern Recognition (CVPR)}, 2025.
\newblock URL \url{https://openaccess.thecvf.com/content/CVPR2025/html/Zhao_From_Zero_to_Detail_Deconstructing_Ultra-High-Definition_Image_Restoration_from_Progressive_CVPR_2025_paper.html}.

\bibitem[Rajagopalan et~al.(2025)Rajagopalan, Nair, Paranjape, and Patel]{corpus066}
Sudarshan Rajagopalan, Nithin~Gopalakrishnan Nair, Jay~N. Paranjape, and Vishal~M. Patel.
\newblock {GenDeg: Diffusion-based Degradation Synthesis for Generalizable All-In-One Image Restoration}.
\newblock In \emph{Proceedings of the IEEE/CVF Conference on Computer Vision and Pattern Recognition (CVPR)}, 2025.
\newblock URL \url{https://openaccess.thecvf.com/content/CVPR2025/html/Rajagopalan_GenDeg_Diffusion-based_Degradation_Synthesis_for_Generalizable_All-In-One_Image_Restoration_CVPR_2025_paper.html}.

\bibitem[Zhu et~al.(2025)Zhu, Li, Qin, Li, Zhang, Guo, and Yang]{corpus067}
Libo Zhu, Jianze Li, Haotong Qin, Wenbo Li, Yulun Zhang, Yong Guo, and Xiaokang Yang.
\newblock {PassionSR: Post-Training Quantization with Adaptive Scale in One-Step Diffusion based Image Super-Resolution}.
\newblock In \emph{Proceedings of the Computer Vision and Pattern Recognition Conference (CVPR)}, 2025.
\newblock URL \url{https://openaccess.thecvf.com/content/CVPR2025/html/Zhu_PassionSR_Post-Training_Quantization_with_Adaptive_Scale_in_One-Step_Diffusion_based_CVPR_2025_paper.html}.

\bibitem[Liu et~al.(2025)Liu, Fu, Xie, and Meng]{corpus069}
Hanze Liu, Jiahong Fu, Qi~Xie, and Deyu Meng.
\newblock {Rotation-Equivariant Self-Supervised Method in Image Denoising}.
\newblock In \emph{Proceedings of the IEEE/CVF Conference on Computer Vision and Pattern Recognition (CVPR)}, 2025.
\newblock URL \url{https://openaccess.thecvf.com/content/CVPR2025/html/Liu_Rotation-Equivariant_Self-Supervised_Method_in_Image_Denoising_CVPR_2025_paper.html}.

\bibitem[Mei et~al.(2025)Mei, Talebi, Ardakani, Patel, Milanfar, and Delbracio]{corpus070}
Kangfu Mei, Hossein Talebi, Mojtaba Ardakani, Vishal~M. Patel, Peyman Milanfar, and Mauricio Delbracio.
\newblock {The Power of Context: How Multimodality Improves Image Super-Resolution}.
\newblock In \emph{Proceedings of the IEEE/CVF Conference on Computer Vision and Pattern Recognition (CVPR)}, 2025.
\newblock URL \url{https://openaccess.thecvf.com/content/CVPR2025/html/Mei_The_Power_of_Context_How_Multimodality_Improves_Image_Super-Resolution_CVPR_2025_paper.html}.

\bibitem[Hu et~al.(2025)Hu, Li, Hou, Hou, Zhang, Wang, Lei, and He]{corpus071}
Jiangbei Hu, Yanggeng Li, Fei Hou, Junhui Hou, Zhebin Zhang, Shengfa Wang, Na~Lei, and Ying He.
\newblock {A Lightweight UDF Learning Framework for 3D Reconstruction Based on Local Shape Functions}.
\newblock In \emph{Proceedings of the IEEE/CVF Conference on Computer Vision and Pattern Recognition (CVPR)}, 2025.
\newblock URL \url{https://openaccess.thecvf.com/content/CVPR2025/html/Hu_A_Lightweight_UDF_Learning_Framework_for_3D_Reconstruction_Based_on_CVPR_2025_paper.html}.

\bibitem[Xie et~al.(2024)Xie, Li, Tan, Sun, Shu, Zhou, Bi, Pirk, and Kaufman]{corpus072}
Desai Xie, Jiahao Li, Hao Tan, Xin Sun, Zhixin Shu, Yi~Zhou, Sai Bi, S\"oren Pirk, and Arie~E. Kaufman.
\newblock {Carve3D: Improving Multi-view Reconstruction Consistency for Diffusion Models with RL Finetuning}.
\newblock In \emph{Proceedings of the IEEE/CVF Conference on Computer Vision and Pattern Recognition (CVPR)}, 2024.
\newblock URL \url{https://openaccess.thecvf.com/content/CVPR2024/html/Xie_Carve3D_Improving_Multi-view_Reconstruction_Consistency_for_Diffusion_Models_with_RL_CVPR_2024_paper.html}.

\bibitem[Sengupta et~al.(2024)Sengupta, Alldieck, Kolotouros, Corona, Zanfir, and Sminchisescu]{corpus074}
Akash Sengupta, Thiemo Alldieck, Nikos Kolotouros, Enric Corona, Andrei Zanfir, and Cristian Sminchisescu.
\newblock {DiffHuman: Probabilistic Photorealistic 3D Reconstruction of Humans}.
\newblock In \emph{Proceedings of the IEEE/CVF Conference on Computer Vision and Pattern Recognition (CVPR)}, 2024.
\newblock URL \url{https://openaccess.thecvf.com/content/CVPR2024/html/Sengupta_DiffHuman_Probabilistic_Photorealistic_3D_Reconstruction_of_Humans_CVPR_2024_paper.html}.

\bibitem[Alzayer et~al.(2025)Alzayer, Henzler, Barron, Huang, Srinivasan, and Verbin]{corpus075}
Hadi Alzayer, Philipp Henzler, Jonathan~T. Barron, Jia-Bin Huang, Pratul~P. Srinivasan, and Dor Verbin.
\newblock {Generative Multiview Relighting for 3D Reconstruction under Extreme Illumination Variation}.
\newblock In \emph{Proceedings of the Computer Vision and Pattern Recognition Conference (CVPR)}, 2025.
\newblock URL \url{https://openaccess.thecvf.com/content/CVPR2025/html/Alzayer_Generative_Multiview_Relighting_for_3D_Reconstruction_under_Extreme_Illumination_Variation_CVPR_2025_paper.html}.

\bibitem[Chen et~al.(2024)Chen, Li, Wang, Xue, and Pang]{corpus076}
Xiao Chen, Quanyi Li, Tai Wang, Tianfan Xue, and Jiangmiao Pang.
\newblock {GenNBV: Generalizable Next-Best-View Policy for Active 3D Reconstruction}.
\newblock In \emph{Proceedings of the IEEE/CVF Conference on Computer Vision and Pattern Recognition (CVPR)}, 2024.
\newblock URL \url{https://openaccess.thecvf.com/content/CVPR2024/html/Chen_GenNBV_Generalizable_Next-Best-View_Policy_for_Active_3D_Reconstruction_CVPR_2024_paper.html}.

\bibitem[Han et~al.(2024)Han, Guo, Fukai, Santo, Shi, Okura, Ma, and Jia]{corpus077}
Yufei Han, Heng Guo, Koki Fukai, Hiroaki Santo, Boxin Shi, Fumio Okura, Zhanyu Ma, and Yunpeng Jia.
\newblock {NeRSP: Neural 3D Reconstruction for Reflective Objects with Sparse Polarized Images}.
\newblock In \emph{Proceedings of the IEEE/CVF Conference on Computer Vision and Pattern Recognition (CVPR)}, 2024.
\newblock URL \url{https://openaccess.thecvf.com/content/CVPR2024/html/Han_NeRSP_Neural_3D_Reconstruction_for_Reflective_Objects_with_Sparse_Polarized_CVPR_2024_paper.html}.

\bibitem[Tan et~al.(2025)Tan, Yu, Shen, and Xue]{corpus079}
Bin Tan, Rui Yu, Yujun Shen, and Nan Xue.
\newblock {PlanarSplatting: Accurate Planar Surface Reconstruction in 3 Minutes}.
\newblock In \emph{Proceedings of the IEEE/CVF Conference on Computer Vision and Pattern Recognition (CVPR)}, 2025.
\newblock URL \url{https://openaccess.thecvf.com/content/CVPR2025/html/Tan_PlanarSplatting_Accurate_Planar_Surface_Reconstruction_in_3_Minutes_CVPR_2025_paper.html}.

\bibitem[Jang et~al.(2025)Jang, Weinzaepfel, Leroy, Agapito, and Revaud]{corpus080}
Wonbong Jang, Philippe Weinzaepfel, Vincent Leroy, Lourdes Agapito, and Jerome Revaud.
\newblock {Pow3R: Empowering Unconstrained 3D Reconstruction with Camera and Scene Priors}.
\newblock In \emph{Proceedings of the IEEE/CVF Conference on Computer Vision and Pattern Recognition (CVPR)}, 2025.
\newblock URL \url{https://openaccess.thecvf.com/content/CVPR2025/html/Jang_Pow3R_Empowering_Unconstrained_3D_Reconstruction_with_Camera_and_Scene_Priors_CVPR_2025_paper.html}.

\bibitem[Zhang et~al.(2026)Zhang, Zhou, Gao, and Wang]{corpus081}
Yaoteng Zhang, Qing Zhou, Junyu Gao, and Qi~Wang.
\newblock {Beyond Prompt Degradation: Prototype-guided Dual-pool Prompting for Incremental Object Detection}.
\newblock In \emph{Proceedings of the IEEE/CVF Conference on Computer Vision and Pattern Recognition (CVPR)}, 2026.
\newblock URL \url{https://openaccess.thecvf.com/content/CVPR2026/html/Zhang_Beyond_Prompt_Degradation_Prototype-guided_Dual-pool_Prompting_for_Incremental_Object_Detection_CVPR_2026_paper.html}.

\bibitem[Monga et~al.(2026)Monga, Chudasama, Wasnik, and Jawahar]{corpus082}
Munish Monga, Vishal Chudasama, Pankaj Wasnik, and C.V. Jawahar.
\newblock {EW-DETR: Evolving World Object Detection via Incremental Low-Rank DEtection TRansformer}.
\newblock In \emph{Proceedings of the IEEE/CVF Conference on Computer Vision and Pattern Recognition (CVPR)}, 2026.
\newblock URL \url{https://openaccess.thecvf.com/content/CVPR2026/html/Monga_EW-DETR_Evolving_World_Object_Detection_via_Incremental_Low-Rank_DEtection_TRansformer_CVPR_2026_paper.html}.

\bibitem[Gu et~al.(2026)Gu, Chen, Gu, and Fu]{corpus083}
Changyu Gu, Linwei Chen, Lin Gu, and Ying Fu.
\newblock {Fourier Angle Alignment for Oriented Object Detection in Remote Sensing}.
\newblock In \emph{Proceedings of the IEEE/CVF Conference on Computer Vision and Pattern Recognition (CVPR)}, 2026.
\newblock URL \url{https://openaccess.thecvf.com/content/CVPR2026/html/Gu_Fourier_Angle_Alignment_for_Oriented_Object_Detection_in_Remote_Sensing_CVPR_2026_paper.html}.

\bibitem[Zhang et~al.(2026)Zhang, Chen, Zhang, and Tao]{corpus084}
Xu~Zhang, Zhe Chen, Jing Zhang, and Dacheng Tao.
\newblock {Heuristic-inspired Reasoning Priors Facilitate Data-Efficient Referring Object Detection}.
\newblock In \emph{Proceedings of the IEEE/CVF Conference on Computer Vision and Pattern Recognition (CVPR)}, 2026.
\newblock URL \url{https://openaccess.thecvf.com/content/CVPR2026/html/Zhang_Heuristic-inspired_Reasoning_Priors_Facilitate_Data-Efficient_Referring_Object_Detection_CVPR_2026_paper.html}.

\bibitem[Br{\"u}ckner et~al.(2026)Br{\"u}ckner, Mercado, Zhang, Kouvaros, and Lomuscio]{corpus086}
Benedikt Br{\"u}ckner, Alejandro~J. Mercado, Yanghao Zhang, Panagiotis Kouvaros, and Alessio Lomuscio.
\newblock {IoUCert: Robustness Verification for Anchor-Based Object Detectors}.
\newblock In \emph{Computer Vision -- ECCV 2026}, 2026.
\newblock \doi{10.1007/978-3-032-37574-2_20}.
\newblock URL \url{https://doi.org/10.1007/978-3-032-37574-2_20}.

\bibitem[Zhang et~al.(2026)Zhang, Han, Chen, Feng, and Wan]{corpus088}
Yupeng Zhang, Ruize Han, Zhiwei Chen, Wei Feng, and Liang Wan.
\newblock {NoOVD: Novel Category Discovery and Embedding for Open-Vocabulary Object Detection}.
\newblock In \emph{Proceedings of the IEEE/CVF Conference on Computer Vision and Pattern Recognition (CVPR)}, 2026.
\newblock URL \url{https://openaccess.thecvf.com/content/CVPR2026/html/Zhang_NoOVD_Novel_Category_Discovery_and_Embedding_for_Open-Vocabulary_Object_Detection_CVPR_2026_paper.html}.

\bibitem[Hu et~al.(2026)Hu, Zhao, Peng, Sun, Chen, Wu, Wang, Yin, Jiang, Li, and Sun]{corpus089}
Zhangchi Hu, Yifan Zhao, Yansong Peng, Wenzhang Sun, Jie Chen, Peixi Wu, Xinghao Wang, Xiangchen Yin, Dongsheng Jiang, Hebei Li, and Xiaoyan Sun.
\newblock {RiO-DETR: DETR for Real-Time Oriented Object Detection}.
\newblock In \emph{Computer Vision -- ECCV 2026}, 2026.
\newblock \doi{10.1007/978-3-032-37083-9_26}.
\newblock URL \url{https://doi.org/10.1007/978-3-032-37083-9_26}.

\bibitem[Belal et~al.(2026)Belal, Medeiros, Pedersoli, and Granger]{corpus090}
Atif Belal, Heitor~R. Medeiros, Marco Pedersoli, and Eric Granger.
\newblock {VLOD-TTA: Test-Time Adaptation of Vision-Language Object Detectors}.
\newblock In \emph{Computer Vision -- ECCV 2026}, 2026.
\newblock \doi{10.1007/978-3-032-37016-7_27}.
\newblock URL \url{https://doi.org/10.1007/978-3-032-37016-7_27}.

\bibitem[Li et~al.(2026)Li, Lu, Zhang, Wang, Xie, and Qu]{corpus091}
Jiahao Li, Yang Lu, Yachao Zhang, Fangyong Wang, Yuan Xie, and Yanyun Qu.
\newblock {Direct Segmentation without Logits Optimization for Training-Free Open-Vocabulary Semantic Segmentation}.
\newblock In \emph{Proceedings of the IEEE/CVF Conference on Computer Vision and Pattern Recognition (CVPR)}, 2026.
\newblock URL \url{https://openaccess.thecvf.com/content/CVPR2026/html/Li_Direct_Segmentation_without_Logits_Optimization_for_Training-Free_Open-Vocabulary_Semantic_Segmentation_CVPR_2026_paper.html}.

\bibitem[Das et~al.(2026)Das, Kukleva, Hu, Asano, and Schiele]{corpus092}
Anurag Das, Anna Kukleva, Xinting Hu, Yuki~M. Asano, and Bernt Schiele.
\newblock {Do Instance Priors Help Weakly Supervised Semantic Segmentation?}
\newblock Manuscript submitted to Transactions on Machine Learning Research (TMLR), 2026.
\newblock URL \url{https://openreview.net/forum?id=4bHFNKe8OU}.

\bibitem[Lv et~al.(2026)Lv, Zhao, Wang, Quan, Huyan, Sebe, and Zhong]{corpus093}
Chonghua Lv, Dong Zhao, Shuang Wang, Dou Quan, Ning Huyan, Nicu Sebe, and Zhun Zhong.
\newblock {Generalizable Knowledge Distillation from Vision Foundation Models for Semantic Segmentation}.
\newblock In \emph{Proceedings of the IEEE/CVF Conference on Computer Vision and Pattern Recognition (CVPR)}, 2026.
\newblock URL \url{https://openaccess.thecvf.com/content/CVPR2026/html/Lv_Generalizable_Knowledge_Distillation_from_Vision_Foundation_Models_for_Semantic_Segmentation_CVPR_2026_paper.html}.

\bibitem[Li et~al.(2026)Li, Wang, Zhang, Zhao, Gao, and Li]{corpus095}
Bingyu Li, Feiyu Wang, Da~Zhang, Zhiyuan Zhao, Junyu Gao, and Xuelong Li.
\newblock {MARIS: Marine Open-Vocabulary Instance Segmentation}.
\newblock In \emph{Proceedings of the IEEE/CVF Conference on Computer Vision and Pattern Recognition (CVPR)}, 2026.
\newblock URL \url{https://openaccess.thecvf.com/content/CVPR2026/html/Li_MARIS_Marine_Open-Vocabulary_Instance_Segmentation_CVPR_2026_paper.html}.

\bibitem[Pei et~al.(2026)Pei, Jiang, Cai, Chen, Yao, and Jeon]{corpus096}
Gensheng Pei, Xiruo Jiang, Xinhao Cai, Tao Chen, Yazhou Yao, and Byeungwoo Jeon.
\newblock {PEARL: Geometry Aligns Semantics for Training-Free Open-Vocabulary Semantic Segmentation}.
\newblock In \emph{Proceedings of the IEEE/CVF Conference on Computer Vision and Pattern Recognition (CVPR)}, 2026.
\newblock URL \url{https://openaccess.thecvf.com/content/CVPR2026/html/Pei_PEARL_Geometry_Aligns_Semantics_for_Training-Free_Open-Vocabulary_Semantic_Segmentation_CVPR_2026_paper.html}.

\bibitem[Li et~al.(2026)Li, Bao, Chen, and Li]{corpus097}
Xuewei Li, Xinghan Bao, Zhimin Chen, and Xi~Li.
\newblock {REL-SF4PASS: Panoramic Semantic Segmentation with REL Depth Representation and Spherical Fusion}.
\newblock In \emph{Proceedings of the IEEE/CVF Conference on Computer Vision and Pattern Recognition (CVPR)}, 2026.
\newblock URL \url{https://openaccess.thecvf.com/content/CVPR2026/html/Li_REL-SF4PASS_Panoramic_Semantic_Segmentation_with_REL_Depth_Representation_and_Spherical_CVPR_2026_paper.html}.

\bibitem[Li et~al.(2026)Li, Zhang, Zhang, Fu, and Zheng]{corpus099}
Qingmei Li, Yang Zhang, Peifeng Zhang, Haohuan Fu, and Juepeng Zheng.
\newblock {SAGE: Style-Adaptive Generalization for Privacy-Constrained Semantic Segmentation Across Domains}.
\newblock In \emph{Proceedings of the IEEE/CVF Conference on Computer Vision and Pattern Recognition (CVPR)}, 2026.
\newblock URL \url{https://openaccess.thecvf.com/content/CVPR2026/html/Li_SAGE_Style-Adaptive_Generalization_for_Privacy-Constrained_Semantic_Segmentation_Across_Domains_CVPR_2026_paper.html}.

\bibitem[Ryou et~al.(2026)Ryou, Ha, Chu, and Han]{corpus101}
Donghun Ryou, Inju Ha, Sanghyeok Chu, and Bohyung Han.
\newblock {Beyond the Ground Truth: Enhanced Supervision for Image Restoration}.
\newblock In \emph{Proceedings of the IEEE/CVF Conference on Computer Vision and Pattern Recognition (CVPR)}, 2026.
\newblock URL \url{https://openaccess.thecvf.com/content/CVPR2026/html/Ryou_Beyond_the_Ground_Truth_Enhanced_Supervision_for_Image_Restoration_CVPR_2026_paper.html}.

\bibitem[Liu et~al.(2026)Liu, Fan, Huang, Xiao, Li, Zhang, Bai, Fu, and Zha]{corpus103}
Yidi Liu, Zihao Fan, Jie Huang, Jie Xiao, Dong Li, Wenlong Zhang, Lei Bai, Xueyang Fu, and Zheng-Jun Zha.
\newblock {FinPercep-RM: A Fine-grained Reward Model and Co-evolutionary Curriculum for RL-based Real-world Super-Resolution}.
\newblock In \emph{Proceedings of the IEEE/CVF Conference on Computer Vision and Pattern Recognition (CVPR)}, 2026.
\newblock URL \url{https://openaccess.thecvf.com/content/CVPR2026/html/Liu_FinPercep-RM_A_Fine-grained_Reward_Model_and_Co-evolutionary_Curriculum_for_RL-based_CVPR_2026_paper.html}.

\bibitem[Yi et~al.(2026)Yi, Li, Wu, Sun, Zhang, and Zhang]{corpus104}
Qiaosi Yi, Shuai Li, Rongyuan Wu, Lingchen Sun, Zhengqiang Zhang, and Lei Zhang.
\newblock {GDPO-SR: Group Direct Preference Optimization for One-Step Generative Image Super-Resolution}.
\newblock In \emph{Proceedings of the IEEE/CVF Conference on Computer Vision and Pattern Recognition (CVPR)}, 2026.
\newblock URL \url{https://openaccess.thecvf.com/content/CVPR2026/html/Yi_GDPO-SR_Group_Direct_Preference_Optimization_for_One-Step_Generative_Image_Super-Resolution_CVPR_2026_paper.html}.

\bibitem[Fang et~al.(2026)Fang, Chen, Yin, Hu, Yao, Zhang, Zhang, and Wang]{corpus105}
Yushun Fang, Yuxiang Chen, Shibo Yin, Qiang Hu, Jiangchao Yao, Ya~Zhang, Xiaoyun Zhang, and Yanfeng Wang.
\newblock {One-Step Diffusion Transformer for Controllable Real-World Image Super-Resolution}.
\newblock In \emph{Proceedings of the IEEE/CVF Conference on Computer Vision and Pattern Recognition (CVPR)}, 2026.
\newblock URL \url{https://openaccess.thecvf.com/content/CVPR2026/html/Fang_One-Step_Diffusion_Transformer_for_Controllable_Real-World_Image_Super-Resolution_CVPR_2026_paper.html}.

\bibitem[Song et~al.(2026)Song, Quan, Wang, Sun, Liu, Yu, Chen, Li, Shen, Yuan, and Yan]{corpus106}
Yushuai Song, Weize Quan, Weining Wang, Jiahui Sun, Jing Liu, Pengbin Yu, Zhentao Chen, Meng Li, Wei Shen, Lunxi Yuan, and Dong-Ming Yan.
\newblock {RefReward-SR: LR-Conditioned Reward Modeling for Preference-Aligned Super-Resolution}.
\newblock In \emph{Computer Vision -- ECCV 2026}, 2026.
\newblock \doi{10.1007/978-3-032-37324-3_6}.
\newblock URL \url{https://doi.org/10.1007/978-3-032-37324-3_6}.

\bibitem[Wang et~al.(2026)Wang, Zhang, Chen, Guo, Wang, Ma, and Du]{corpus107}
Hebaixu Wang, Jing Zhang, Haoyang Chen, Haonan Guo, Di~Wang, Jiayi Ma, and Bo~Du.
\newblock {Residual Diffusion Bridge Model for Image Restoration}.
\newblock In \emph{Proceedings of the IEEE/CVF Conference on Computer Vision and Pattern Recognition (CVPR)}, 2026.
\newblock URL \url{https://openaccess.thecvf.com/content/CVPR2026/html/Wang_Residual_Diffusion_Bridge_Model_for_Image_Restoration_CVPR_2026_paper.html}.

\bibitem[Li et~al.(2026)Li, Wu, Shao, and Fu]{corpus108}
Hesong Li, Ziqi Wu, Ruiwen Shao, and Ying Fu.
\newblock {Statistical Characteristic-Guided Denoising for Rapid High-Resolution Transmission Electron Microscopy Imaging}.
\newblock In \emph{Proceedings of the IEEE/CVF Conference on Computer Vision and Pattern Recognition (CVPR)}, 2026.
\newblock URL \url{https://openaccess.thecvf.com/content/CVPR2026/html/Li_Statistical_Characteristic-Guided_Denoising_for_Rapid_High-Resolution_Transmission_Electron_Microscopy_Imaging_CVPR_2026_paper.html}.

\bibitem[Jia et~al.(2026)Jia, Zhang, Hu, Zhao, Zhao, Sun, Long, Tian, Jiang, Liu, Wang, Lian, Zhang, and Zhou]{corpus109}
Guoli Jia, Yisheng Zhang, Haote Hu, Shanxu Zhao, Kaikai Zhao, Long Sun, Xinwei Long, Kai Tian, Che Jiang, Zhaoxiang Liu, Kai Wang, Shiguo Lian, Kaiyan Zhang, and Bowen Zhou.
\newblock {TIR-Agent: Training an Explorative and Efficient Agent for Image Restoration}.
\newblock In \emph{Computer Vision -- ECCV 2026}, 2026.
\newblock \doi{10.1007/978-3-032-37281-9_32}.
\newblock URL \url{https://doi.org/10.1007/978-3-032-37281-9_32}.

\bibitem[Huang et~al.(2026)Huang, Duan, Zhao, Zhao, Ma, and Gao]{corpus111}
Binbin Huang, Haobin Duan, Yiqun Zhao, Zibo Zhao, Yi~Ma, and Shenghua Gao.
\newblock {CUPID: Generative 3D Reconstruction via Joint Object and Pose Modeling}.
\newblock In \emph{Proceedings of the IEEE/CVF Conference on Computer Vision and Pattern Recognition (CVPR)}, 2026.
\newblock URL \url{https://openaccess.thecvf.com/content/CVPR2026/html/Huang_CUPID_Generative_3D_Reconstruction_via_Joint_Object_and_Pose_Modeling_CVPR_2026_paper.html}.

\bibitem[He et~al.(2026)He, Zhou, Zheng, Liang, Wang, and Yang]{corpus113}
Yumeng He, Zanwei Zhou, Yekun Zheng, Chen Liang, Yunbo Wang, and Xiaokang Yang.
\newblock {EMGauss: Continuous Slice-to-3D Reconstruction via Dynamic Gaussian Modeling in Volume Electron Microscopy}.
\newblock In \emph{Proceedings of the IEEE/CVF Conference on Computer Vision and Pattern Recognition (CVPR)}, 2026.
\newblock URL \url{https://openaccess.thecvf.com/content/CVPR2026/html/He_EMGauss_Continuous_Slice-to-3D_Reconstruction_via_Dynamic_Gaussian_Modeling_in_Volume_CVPR_2026_paper.html}.

\bibitem[Huang et~al.(2026)Huang, Luo, Mao, and Mikolajczyk]{corpus114}
Ranran Huang, Weixun Luo, Ye~Mao, and Krystian Mikolajczyk.
\newblock {From None to All: Self-Supervised 3D Reconstruction via Novel View Synthesis}.
\newblock In \emph{Proceedings of the IEEE/CVF Conference on Computer Vision and Pattern Recognition (CVPR)}, 2026.
\newblock URL \url{https://openaccess.thecvf.com/content/CVPR2026/html/Huang_From_None_to_All_Self-Supervised_3D_Reconstruction_via_Novel_View_CVPR_2026_paper.html}.

\bibitem[Hu et~al.(2026)Hu, Lin, Long, Ran, Jiang, Wang, Zhu, Xu, Wang, and Pang]{corpus115}
Wenbo Hu, Jingli Lin, Yilin Long, Yunlong Ran, Lihan Jiang, Yifan Wang, Chenming Zhu, Runsen Xu, Tai Wang, and Jiangmiao Pang.
\newblock {G$^2$VLM: Geometry Grounded Vision Language Model with Unified 3D Reconstruction and Spatial Reasoning}.
\newblock In \emph{Proceedings of the IEEE/CVF Conference on Computer Vision and Pattern Recognition (CVPR)}, 2026.
\newblock URL \url{https://openaccess.thecvf.com/content/CVPR2026/html/Hu_G2VLM_Geometry_Grounded_Vision_Language_Model_with_Unified_3D_Reconstruction_CVPR_2026_paper.html}.

\bibitem[Guo et~al.(2026)Guo, Chao, Wang, Zhao, Dai, Zhang, Yu, and Shi]{corpus116}
Yijing Guo, Mengjun Chao, Luo Wang, Tianyang Zhao, Haizhao Dai, Yingliang Zhang, Jingyi Yu, and Yujiao Shi.
\newblock {PanoVGGT: Feed-Forward 3D Reconstruction from Panoramic Imagery}.
\newblock In \emph{Proceedings of the IEEE/CVF Conference on Computer Vision and Pattern Recognition (CVPR)}, 2026.
\newblock URL \url{https://openaccess.thecvf.com/content/CVPR2026/html/Guo_PanoVGGT_Feed-Forward_3D_Reconstruction_from_Panoramic_Imagery_CVPR_2026_paper.html}.

\bibitem[Wang et~al.(2026)Wang, Shang, Liu, and Yan]{corpus118}
Feiran Wang, Zezhou Shang, Gaowen Liu, and Yan Yan.
\newblock {RayMap3R: Inference-Time RayMap for Dynamic 3D Reconstruction}.
\newblock In \emph{Computer Vision -- ECCV 2026}, 2026.
\newblock \doi{10.1007/978-3-032-37135-5_12}.
\newblock URL \url{https://doi.org/10.1007/978-3-032-37135-5_12}.

\bibitem[Qi et~al.(2026)Qi, Zhu, Bao, Zhao, and Li]{corpus119}
Yunshan Qi, Lin Zhu, Nan Bao, Yifan Zhao, and Jia Li.
\newblock {Seeing through Light and Darkness: Sensor-Physics Grounded Deblurring HDR NeRF from Single-Exposure Images and Events}.
\newblock In \emph{Proceedings of the IEEE/CVF Conference on Computer Vision and Pattern Recognition (CVPR)}, 2026.
\newblock URL \url{https://openaccess.thecvf.com/content/CVPR2026/html/Qi_Seeing_through_Light_and_Darkness_Sensor-Physics_Grounded_Deblurring_HDR_NeRF_CVPR_2026_paper.html}.

\bibitem[Asthana et~al.(2026)Asthana, Hanson, Tu, Goldstein, Zwicker, and Varshney]{corpus120}
Pranav Asthana, Alex Hanson, Allen Tu, Tom Goldstein, Matthias Zwicker, and Amitabh Varshney.
\newblock {SplatSuRe: Selective Super-Resolution for Multi-view Consistent 3D Gaussian Splatting}.
\newblock In \emph{Proceedings of the IEEE/CVF Conference on Computer Vision and Pattern Recognition (CVPR)}, 2026.
\newblock URL \url{https://openaccess.thecvf.com/content/CVPR2026/html/Asthana_SplatSuRe_Selective_Super-Resolution_for_Multi-view_Consistent_3D_Gaussian_Splatting_CVPR_2026_paper.html}.

\bibitem[Anonymous(2026)]{corpus122}
Anonymous.
\newblock {CANDIDATE-BOX SELECTION MATTERS FOR CROSS-DOMAIN FEW-SHOT OBJECT DETECTION}.
\newblock Manuscript submitted to the International Conference on Learning Representations (ICLR 2027), 2026.
\newblock URL \url{https://openreview.net/forum?id=6kDoAMVPn5}.

\bibitem[Anonymous(2026)]{corpus123}
Anonymous.
\newblock {DCHA: Dynamic Compositional Adaptation with Hierarchical Drift Constraints for Incremental Vision-Language Object Detection}.
\newblock Manuscript submitted to the International Conference on Learning Representations (ICLR 2027), 2026.
\newblock URL \url{https://openreview.net/forum?id=WrLCWdkY49}.

\bibitem[Anonymous(2026)]{corpus124}
Anonymous.
\newblock {DeCo: A Multimodal Detection Framework based on Density Consensus Learning}.
\newblock Manuscript submitted to the International Conference on Learning Representations (ICLR 2027), 2026.
\newblock URL \url{https://openreview.net/forum?id=8P9XSKFW9N}.

\bibitem[Anonymous(2026)]{corpus125}
Anonymous.
\newblock {Different Instances, Different Supervision: Instance-Adaptive Point-Supervised Oriented Object Detection}.
\newblock Manuscript submitted to the International Conference on Learning Representations (ICLR 2027), 2026.
\newblock URL \url{https://openreview.net/forum?id=AUmtOiMvCa}.

\bibitem[Anonymous(2026)]{corpus126}
Anonymous.
\newblock {Imbalance-Adaptive Cross Consistency Learning for Semi-Supervised Oriented Object Detection}.
\newblock Manuscript submitted to the International Conference on Learning Representations (ICLR 2027), 2026.
\newblock URL \url{https://openreview.net/forum?id=0jbdTGiGaz}.

\bibitem[Anonymous(2026)]{corpus129}
Anonymous.
\newblock {Towards Trustworthy Object Detection: Adaptive Boundary Optimization with Joint Classification and Localization Risk Control}.
\newblock Manuscript submitted to the International Conference on Learning Representations (ICLR 2027), 2026.
\newblock URL \url{https://openreview.net/forum?id=1I3yfptfr1}.

\bibitem[Anonymous(2026)]{corpus130}
Anonymous.
\newblock {UniOSE: Semantics-Enriched Object Queries for Open-Ended Detection and Segmentation}.
\newblock Manuscript submitted to the International Conference on Learning Representations (ICLR 2027), 2026.
\newblock URL \url{https://openreview.net/forum?id=vTYhIzoLzY}.

\bibitem[Anonymous(2026)]{corpus131}
Anonymous.
\newblock {Architectural Missing-Modality Robustness in Semantic Segmentation}.
\newblock Manuscript submitted to the International Conference on Learning Representations (ICLR 2027), 2026.
\newblock URL \url{https://openreview.net/forum?id=ymk7OpPLY8}.

\bibitem[Anonymous(2026)]{corpus133}
Anonymous.
\newblock {DevoSeg: Budget-Aware Self-Evolving Semantic Segmentation under Visual Degradation}.
\newblock Manuscript submitted to the International Conference on Learning Representations (ICLR 2027), 2026.
\newblock URL \url{https://openreview.net/forum?id=8aErJ21sV6}.

\bibitem[Anonymous(2026)]{corpus135}
Anonymous.
\newblock {Past as Guidance, Not Just Constraint: Cross-Level Knowledge Reorganization for Class-Incremental Semantic Segmentation}.
\newblock Manuscript submitted to the International Conference on Learning Representations (ICLR 2027), 2026.
\newblock URL \url{https://openreview.net/forum?id=9S1a6H8xUR}.

\bibitem[Anonymous(2026)]{corpus137}
Anonymous.
\newblock {Rethinking Unaligned RGB-T Dense Prediction as Cross-Modal Information Transport}.
\newblock Manuscript submitted to the International Conference on Learning Representations (ICLR 2027), 2026.
\newblock URL \url{https://openreview.net/forum?id=OHqdlezqWF}.

\bibitem[Anonymous(2026)]{corpus138}
Anonymous.
\newblock {SEER: Semantic Evidence for Executable Region Propagation in Open-Vocabulary Semantic Segmentation}.
\newblock Manuscript submitted to the International Conference on Learning Representations (ICLR 2027), 2026.
\newblock URL \url{https://openreview.net/forum?id=hEBqQJhg1b}.

\bibitem[Anonymous(2026)]{corpus139}
Anonymous.
\newblock {SHAPE: Structural Hierarchy of Affinities for Patch Enhancement in Training-Free Open-Vocabulary Semantic Segmentation}.
\newblock Manuscript submitted to the International Conference on Learning Representations (ICLR 2027), 2026.
\newblock URL \url{https://openreview.net/forum?id=67ApUu5dSo}.

\bibitem[Anonymous(2026)]{corpus140}
Anonymous.
\newblock {Where from Above, What from the Street: Learning Semantic Routing for Cross-View Segmentation under Pose Uncertainty}.
\newblock Manuscript submitted to the International Conference on Learning Representations (ICLR 2027), 2026.
\newblock URL \url{https://openreview.net/forum?id=J2V6tPOCod}.

\bibitem[Cui et~al.(2026)Cui, Zhang, Xu, Deng, Bai, Feng, Qin, Schindler, and Zhao]{corpus141}
Yukun Cui, Jiangshe Zhang, Shuang Xu, Lilun Deng, Haowen Bai, Tao Feng, Haotong Qin, Konrad Schindler, and Zixiang Zhao.
\newblock {$\mathrm{S}^{3}$-PnP: Plug-and-Play Denoising for Scientific Imaging with Self-Supervised Score Priors}.
\newblock Manuscript submitted to the International Conference on Learning Representations (ICLR 2027), 2026.
\newblock URL \url{https://openreview.net/forum?id=v44KR7QgKE}.

\bibitem[Huang et~al.(2026)Huang, Yang, Wu, Li, Zhi, Zhang, Hou, Deng, and Xia]{corpus142}
Xinghua Huang, Zhixiong Yang, Chen Wu, Shengxi Li, Shuaifeng Zhi, Yue Zhang, Qibin Hou, Xin Deng, and Jingyuan Xia.
\newblock {Degradation Frequency Curve: An Explicit Frequency-Quantified Representation for All-in-One Image Restoration}.
\newblock Manuscript submitted to the International Conference on Learning Representations (ICLR 2027), 2026.
\newblock URL \url{https://openreview.net/forum?id=IP9k9VdnnS}.

\bibitem[Kong et~al.(2026)Kong, Zhao, Sun, Wu, and Zhang]{corpus144}
Xiangtao Kong, Jixin Zhao, Lingchen Sun, Rongyuan Wu, and Lei Zhang.
\newblock {GGT-100K: Generative Ground Truth for Generalizable Real-World Image Restoration}.
\newblock Manuscript submitted to the International Conference on Learning Representations (ICLR 2027), 2026.
\newblock URL \url{https://openreview.net/forum?id=1n4L8nXB1I}.

\bibitem[Kong et~al.(2026)Kong, Liu, Wu, Sun, Zhang, Zhao, Wu, and Zhang]{corpus145}
Xiangtao Kong, Shuaizheng Liu, Rongyuan Wu, Lingchen Sun, Zhengqiang Zhang, Jixin Zhao, Yuhui Wu, and Lei Zhang.
\newblock {HarnessIR: Harnessing Multimodal Foundation Models for Universal Real-World Image Restoration}.
\newblock Manuscript submitted to the International Conference on Learning Representations (ICLR 2027), 2026.
\newblock URL \url{https://openreview.net/forum?id=QlpWNkJcBr}.

\bibitem[Gao et~al.(2026)Gao, Chen, and Ma]{corpus146}
Hu~Gao, Yulong Chen, and Lizhuang Ma.
\newblock {Learning to Restore More: Continual Capability Expansion for Pretrained Image Restoration Models}.
\newblock Manuscript submitted to the International Conference on Learning Representations (ICLR 2027), 2026.
\newblock URL \url{https://openreview.net/forum?id=Cdx9a9GdBy}.

\bibitem[Di et~al.(2026)Di, Shi, Bao, Peng, Zhao, Guo, Pei, Fu, Cao, and Zha]{corpus148}
Xin Di, Mingyu Shi, Yuanfei Bao, Long Peng, Yue Zhao, Jiaming Guo, Renjing Pei, Xueyang Fu, Yang Cao, and Zheng-Jun Zha.
\newblock {PhoenixSR: Generative Heterogeneous Distillation Unleashes Efficient Models for Real-World Super-Resolution}.
\newblock Manuscript submitted to the International Conference on Learning Representations (ICLR 2027), 2026.
\newblock URL \url{https://openreview.net/forum?id=anN7bIopnC}.

\bibitem[Feng et~al.(2026)Feng, Zhao, Zhang, Pei, Chen, and Lu]{corpus149}
Xin Feng, Jin Zhao, Yizhen Zhang, Wenjie Pei, Fanglin Chen, and Guangming Lu.
\newblock {Restoring without Forgetting: Filter-Level Continual Image Restoration via Parameter-Space Integrated Gradients}.
\newblock Manuscript submitted to the International Conference on Learning Representations (ICLR 2027), 2026.
\newblock URL \url{https://openreview.net/forum?id=80XJOuh0nO}.

\bibitem[Chen et~al.(2026)Chen, Wang, Chen, Cao, Zhang, Su, Liu, and Zeng]{corpus150}
Ping Chen, Xiangming Wang, Yongyong Chen, Jiezhang Cao, Kai Zhang, Jingyong Su, Jie Liu, and Haijin Zeng.
\newblock {SegDem: Segmentation helps Demosaicing}.
\newblock Manuscript submitted to the International Conference on Learning Representations (ICLR 2027), 2026.
\newblock URL \url{https://openreview.net/forum?id=3Kt188v0Yl}.

\bibitem[Fradlin et~al.(2026)Fradlin, Schmidt, Piekenbrinck, Knaebel, Garcia, and Leibe]{corpus151}
Ilya Fradlin, Christian Schmidt, Jens Piekenbrinck, Karim Knaebel, Gonzalo~Martin Garcia, and Bastian Leibe.
\newblock {ARROW: Arbitrary Reconstruction and Tracking of 4D Observations in the Wild}.
\newblock Manuscript submitted to the International Conference on Learning Representations (ICLR 2027), 2026.
\newblock URL \url{https://openreview.net/forum?id=hShILxckIE}.

\bibitem[Yang et~al.(2026)Yang, Yang, Xiao, Chen, Shi, Jiang, and Zhang]{corpus152}
Tianyi Yang, Haojun Yang, Jipeng Xiao, Xuesong Chen, Shaoshuai Shi, Li~Jiang, and Zhipeng Zhang.
\newblock {Drive3R: Block-wise Streaming Geometric Foundation Models for Autonomous Driving}.
\newblock Manuscript submitted to the International Conference on Learning Representations (ICLR 2027), 2026.
\newblock URL \url{https://openreview.net/forum?id=4hJ82iauw9}.

\bibitem[Anonymous(2026)]{corpus153}
Anonymous.
\newblock {Failure-Aware Adaptation for Transparent Objects in Feed-Forward 3D Reconstruction}.
\newblock Manuscript submitted to the International Conference on Learning Representations (ICLR 2027), 2026.
\newblock URL \url{https://openreview.net/forum?id=iRCP1uMP8S}.

\bibitem[Zhang et~al.(2026)Zhang, Ouyang, Chen, Mei, Chen, Ren, Sun, Zhang, Liu, and Wang]{corpus154}
Shuai Zhang, Zhanpeng Ouyang, Zhengqing Chen, Ruohong Mei, Shaoyu Chen, Weiqiang Ren, Hui Sun, Qian Zhang, Wenyu Liu, and Xinggang Wang.
\newblock {Latent Bridge: From Video Latents to 3D Geometry via Feature Alignment}.
\newblock Manuscript submitted to the International Conference on Learning Representations (ICLR 2027), 2026.
\newblock URL \url{https://openreview.net/forum?id=dPyGZiJESZ}.

\bibitem[Anonymous(2026)]{corpus156}
Anonymous.
\newblock {PAIR3D: Learning Physics-Aligned and Visually Consistent 3D Scene Layouts from Verifiable Rewards}.
\newblock Manuscript submitted to the International Conference on Learning Representations (ICLR 2027), 2026.
\newblock URL \url{https://openreview.net/forum?id=e8KzLssuLH}.

\bibitem[Matias et~al.(2026)Matias, Nascimento, Nakayama, Lima, Lobo, Wetzstein, Guibas, Paiva, and Novello]{corpus157}
Vitor Matias, Filipe Nascimento, Kiyohiro Nakayama, João~Paulo Lima, Márcus Lobo, Gordon Wetzstein, Leonidas Guibas, Afonso Paiva, and Tiago Novello.
\newblock {PCAsplat: Gaussian Splatting with Local PCA Regularization}.
\newblock Manuscript submitted to the International Conference on Learning Representations (ICLR 2027), 2026.
\newblock URL \url{https://openreview.net/forum?id=0X2ay3NW4Z}.

\bibitem[Lin et~al.(2026)Lin, Zheng, Zhuo, Wu, Zhou, and Lu]{corpus158}
Shing Ho~J. Lin, Wenzhao Zheng, Dong Zhuo, Yuqi Wu, Jie Zhou, and Jiwen Lu.
\newblock {SM4RT: Learning Structured Motion Geometry for 4D Reconstruction}.
\newblock Manuscript submitted to the International Conference on Learning Representations (ICLR 2027), 2026.
\newblock URL \url{https://openreview.net/forum?id=Y69S27nWgp}.

\bibitem[Chen et~al.(2026)Chen, Chen, Zheng, Jia, Liu, and Li]{corpus159}
Wenjue Chen, Wenyue Chen, Heliang Zheng, Rongfei Jia, Yuan Liu, and Meng Li.
\newblock {UniForge3D: Joint 3D Reconstruction and Generation in a Shared Anchor-Guided Space}.
\newblock Manuscript submitted to the International Conference on Learning Representations (ICLR 2027), 2026.
\newblock URL \url{https://openreview.net/forum?id=dpOjIHiDKY}.

\bibitem[Anonymous(2026)]{corpus161}
Anonymous.
\newblock {$\mathrm{E}^2$IR-Agent: Experience-Driven Self-Evolving Image Restoration Agentic System}.
\newblock Manuscript submitted to the International Conference on Learning Representations (ICLR 2027), 2026.
\newblock URL \url{https://openreview.net/forum?id=hFRPKRZXeS}.

\bibitem[Anonymous(2026)]{corpus165}
Anonymous.
\newblock {An Agentic AI Scientific Community for Automated Neural Operator Discovery}.
\newblock Manuscript submitted to the International Conference on Learning Representations (ICLR 2027), 2026.
\newblock URL \url{https://openreview.net/forum?id=RPFKIcqk1w}.

\bibitem[Anonymous(2026)]{corpus166}
Anonymous.
\newblock {Beyond Dyadic Memory: Interaction-Aware Multimodal Memory with Adaptive Agentic Retrieval for Multi-Party Spoken Conversations}.
\newblock Manuscript submitted to the International Conference on Learning Representations (ICLR 2027), 2026.
\newblock URL \url{https://openreview.net/forum?id=0ZTqgrmYaQ}.

\bibitem[Anonymous(2026)]{corpus168}
Anonymous.
\newblock {BodyBound: A Benchmark for Diagnosing Robot-Policy Responses to Body Changes Beyond Task Success}.
\newblock Manuscript submitted to the International Conference on Learning Representations (ICLR 2027), 2026.
\newblock URL \url{https://openreview.net/forum?id=skorOYrYml}.

\bibitem[Anonymous(2026)]{corpus169}
Anonymous.
\newblock {BrainIPAgent: Multi-Agent Interpretable Structural Priors for Multimodal Brain Network Learning}.
\newblock Manuscript submitted to the International Conference on Learning Representations (ICLR 2027), 2026.
\newblock URL \url{https://openreview.net/forum?id=YLI8SRurGG}.

\bibitem[Anonymous(2026)]{corpus170}
Anonymous.
\newblock {CADForge: Agentic Single-View CAD Reconstruction with Explicit Geometry Reasoning}.
\newblock Manuscript submitted to the International Conference on Learning Representations (ICLR 2027), 2026.
\newblock URL \url{https://openreview.net/forum?id=yLEH2uLAgc}.

\bibitem[Anonymous(2026)]{corpus171}
Anonymous.
\newblock {CUAWright: A Minimal Unified Interface for Computer Use Agents}.
\newblock Manuscript submitted to the International Conference on Learning Representations (ICLR 2027), 2026.
\newblock URL \url{https://openreview.net/forum?id=zploFP1iVU}.

\bibitem[Anonymous(2026)]{corpus172}
Anonymous.
\newblock {Don't Ask Agents to Do Everything: Selective Agentic Reasoning for Single Image to 3D Scene Generation}.
\newblock Manuscript submitted to the International Conference on Learning Representations (ICLR 2027), 2026.
\newblock URL \url{https://openreview.net/forum?id=djznyPY9n1}.

\bibitem[Anonymous(2026)]{corpus173}
Anonymous.
\newblock {DUDA-Bench: Benchmarking LLM Agents on Multimodal Data-Driven Urban Diagnosis}.
\newblock Manuscript submitted to the International Conference on Learning Representations (ICLR 2027), 2026.
\newblock URL \url{https://openreview.net/forum?id=FqbDu6m3j2}.

\bibitem[Anonymous(2026)]{corpus175}
Anonymous.
\newblock {EmboScope: A Multi-Source Comprehensive Benchmark for Embodied Perception, Understanding, and Planning}.
\newblock Manuscript submitted to the International Conference on Learning Representations (ICLR 2027), 2026.
\newblock URL \url{https://openreview.net/forum?id=vowOuAXSIO}.

\bibitem[Anonymous(2026)]{corpus176}
Anonymous.
\newblock {EngiWorld: What Can Frontier Agents Deliver in Professional Engineering Environments?}
\newblock Manuscript submitted to the International Conference on Learning Representations (ICLR 2027), 2026.
\newblock URL \url{https://openreview.net/forum?id=4kPwNYWJYm}.

\bibitem[Anonymous(2026)]{corpus177}
Anonymous.
\newblock {FinFIRST: Benchmarking Search Agents for Financial Information Retrieval, Sourcing and Traceability}.
\newblock Manuscript submitted to the International Conference on Learning Representations (ICLR 2027), 2026.
\newblock URL \url{https://openreview.net/forum?id=8ztJtapkEV}.

\bibitem[Anonymous(2026)]{corpus178}
Anonymous.
\newblock {FM-Bench: A Benchmark for Long-Horizon Management with Competing Agents}.
\newblock Manuscript submitted to the International Conference on Learning Representations (ICLR 2027), 2026.
\newblock URL \url{https://openreview.net/forum?id=ZEeYfne3aD}.

\bibitem[Anonymous(2026)]{corpus179}
Anonymous.
\newblock {Frameworkers: A Dynamic Multi-Agent Framework for AI Video Production}.
\newblock Manuscript submitted to the International Conference on Learning Representations (ICLR 2027), 2026.
\newblock URL \url{https://openreview.net/forum?id=X31pnGGh8D}.

\bibitem[Anonymous(2026)]{corpus180}
Anonymous.
\newblock {GATOR: Generative and Agentic 3D Object Reconstruction from Casual Images}.
\newblock Manuscript submitted to the International Conference on Learning Representations (ICLR 2027), 2026.
\newblock URL \url{https://openreview.net/forum?id=12nS7mYinw}.

\bibitem[Anonymous(2026)]{corpus181}
Anonymous.
\newblock {GRASP: Grounded Restoration with Agent-Refined Semantic Prompts for Real-World Image Super-Resolution}.
\newblock Manuscript submitted to the International Conference on Learning Representations (ICLR 2027), 2026.
\newblock URL \url{https://openreview.net/forum?id=3gogFGspDb}.

\bibitem[Anonymous(2026)]{corpus182}
Anonymous.
\newblock {Harmony Harness: Multi-Objective Self-Evolution of Agent Harnesses}.
\newblock Manuscript submitted to the International Conference on Learning Representations (ICLR 2027), 2026.
\newblock URL \url{https://openreview.net/forum?id=cR0St8qS2n}.

\bibitem[Anonymous(2026)]{corpus183}
Anonymous.
\newblock {LEGO-Anything: Coding Agents for 3D Scene Reconstruction}.
\newblock Manuscript submitted to the International Conference on Learning Representations (ICLR 2027), 2026.
\newblock URL \url{https://openreview.net/forum?id=olW2A6zx75}.

\bibitem[Anonymous(2026)]{corpus185}
Anonymous.
\newblock {Maestro: Agentic Formal Proof Development by Coordination}.
\newblock Manuscript submitted to the International Conference on Learning Representations (ICLR 2027), 2026.
\newblock URL \url{https://openreview.net/forum?id=02vIJnLoLo}.

\bibitem[Anonymous(2026)]{corpus186}
Anonymous.
\newblock {MASCraft: Multi-Agent Specialization through Role--Skill Composition and Self-Evolving Orchestration}.
\newblock Manuscript submitted to the International Conference on Learning Representations (ICLR 2027), 2026.
\newblock URL \url{https://openreview.net/forum?id=7zleBvksX6}.

\bibitem[Anonymous(2026)]{corpus187}
Anonymous.
\newblock {MatGym: An Agent-Centric Environment Bridging Materials Discovery and Reproducible Synthesis}.
\newblock Manuscript submitted to the International Conference on Learning Representations (ICLR 2027), 2026.
\newblock URL \url{https://openreview.net/forum?id=1Xi9fslMSG}.

\bibitem[Anonymous(2026)]{corpus188}
Anonymous.
\newblock {MemVista: Benchmarking Long-Term Memory for Multimodal Agents in Evolving Projects}.
\newblock Manuscript submitted to the International Conference on Learning Representations (ICLR 2027), 2026.
\newblock URL \url{https://openreview.net/forum?id=Dli0qriFyd}.

\bibitem[Anonymous(2026)]{corpus189}
Anonymous.
\newblock {OmniClinic: Evaluating Active Evidence Acquisition in a Multimodal Clinical Sandbox}.
\newblock Manuscript submitted to the International Conference on Learning Representations (ICLR 2027), 2026.
\newblock URL \url{https://openreview.net/forum?id=EkkRujJohS}.

\bibitem[Anonymous(2026)]{corpus192}
Anonymous.
\newblock {OPTScientist: Toward Recursive Self-Improving Agents via Automated Optimizer Discovery}.
\newblock Manuscript submitted to the International Conference on Learning Representations (ICLR 2027), 2026.
\newblock URL \url{https://openreview.net/forum?id=VkVp7CjajL}.

\bibitem[Anonymous(2026)]{corpus193}
Anonymous.
\newblock {SpatialAgent: Working Memory for Long-Horizon Spatial Reasoning}.
\newblock Manuscript submitted to the International Conference on Learning Representations (ICLR 2027), 2026.
\newblock URL \url{https://openreview.net/forum?id=xGIdDM5rYW}.

\bibitem[Anonymous(2026)]{corpus196}
Anonymous.
\newblock {TARSphere: A Robot-Oriented Full-Stack Framework for Long-Horizon Embodied Agents in Evolving Physical and Social Environments}.
\newblock Manuscript submitted to the International Conference on Learning Representations (ICLR 2027), 2026.
\newblock URL \url{https://openreview.net/forum?id=eFi9w6vSHP}.

\bibitem[Anonymous(2026)]{corpus197}
Anonymous.
\newblock {Topological Coherence for Self-Evolving Multi-Agent Systems}.
\newblock Manuscript submitted to the International Conference on Learning Representations (ICLR 2027), 2026.
\newblock URL \url{https://openreview.net/forum?id=VJ7YUdWOrj}.

\bibitem[Anonymous(2026)]{corpus199}
Anonymous.
\newblock {VideoWeaver: Evaluating and Evolving Skills for Agentic Long Video Generation}.
\newblock Manuscript submitted to the International Conference on Learning Representations (ICLR 2027), 2026.
\newblock URL \url{https://openreview.net/forum?id=ahcNtcQxsO}.

\bibitem[Anonymous(2026)]{corpus200}
Anonymous.
\newblock {When Should AI Chime In? Benchmarking Timely Contribution of Proactive Agents in Multi-Party Chat}.
\newblock Manuscript submitted to the International Conference on Learning Representations (ICLR 2027), 2026.
\newblock URL \url{https://openreview.net/forum?id=MI8d8y9Hja}.

\end{thebibliography}
\clearpage
\appendix
\section{Complete Paper Index}
\label{sec:paper-index}
\label{app:corpus-index}
This index lists all 200 collected papers. Each linked title opens the recorded paper page; each citation identifies its bibliography entry. Atlas identifiers match the 200 illustrated pages in the companion \emph{Figure Atlas}. M and F denote motivation and framework figures. ``Style'' denotes the separately selected S4-AI-Style contrast cohort, not a fifth chronological stage. OpenReview entries are submissions, not confirmed acceptances.
\begingroup\scriptsize\setlength{\tabcolsep}{3pt}\renewcommand{\arraystretch}{1.18}
\begin{longtable}{@{}p{0.04\linewidth}p{0.055\linewidth}p{0.79\linewidth}p{0.04\linewidth}@{}}
\toprule ID & Cohort & Paper (clickable title and citation) & M/F \\
\midrule\endfirsthead
\toprule ID & Cohort & Paper (continued) & M/F \\
\midrule\endhead
\bottomrule\endfoot
001 & S1 & \href{https://openaccess.thecvf.com/content/CVPR2022/html/Li_A_Dual_Weighting_Label_Assignment_Scheme_for_Object_Detection_CVPR_2022_paper.html}{A Dual Weighting Label Assignment Scheme for Object Detection} \citep{corpus001} & M/F \\
002 & S1 & \href{https://openaccess.thecvf.com/content/CVPR2022/html/He_Cross_Domain_Object_Detection_by_Target-Perceived_Dual_Branch_Distillation_CVPR_2022_paper.html}{Cross Domain Object Detection by Target-Perceived Dual Branch Distillation} \citep{corpus002} & M/F \\
003 & S1 & \href{https://openaccess.thecvf.com/content/CVPR2022/html/Chen_Dense_Learning_Based_Semi-Supervised_Object_Detection_CVPR_2022_paper.html}{Dense Learning Based Semi-Supervised Object Detection} \citep{corpus003} & M/F \\
004 & S1 & \href{https://openaccess.thecvf.com/content/CVPR2022/html/Zheng_Localization_Distillation_for_Dense_Object_Detection_CVPR_2022_paper.html}{Localization Distillation for Dense Object Detection} \citep{corpus004} & M/F \\
005 & S1 & \href{https://openaccess.thecvf.com/content/CVPR2022/html/Elezi_Not_All_Labels_Are_Equal_Rationalizing_the_Labeling_Costs_for_CVPR_2022_paper.html}{Not All Labels Are Equal: Rationalizing the Labeling Costs for Training Object Detection} \citep{corpus005} & M/F \\
006 & S1 & \href{https://openaccess.thecvf.com/content/CVPR2022/html/Zheng_Progressive_End-to-End_Object_Detection_in_Crowded_Scenes_CVPR_2022_paper.html}{Progressive End-to-End Object Detection in Crowded Scenes} \citep{corpus006} & M/F \\
007 & S1 & \href{https://openaccess.thecvf.com/content/CVPR2022/html/Guo_Scale-Equivalent_Distillation_for_Semi-Supervised_Object_Detection_CVPR_2022_paper.html}{Scale-Equivalent Distillation for Semi-Supervised Object Detection} \citep{corpus007} & M/F \\
008 & S1 & \href{https://openaccess.thecvf.com/content/CVPR2022/html/Li_SIOD_Single_Instance_Annotated_per_Category_per_Image_for_Object_CVPR_2022_paper.html}{SIOD: Single Instance Annotated per Category per Image for Object Detection} \citep{corpus008} & M/F \\
009 & S1 & \href{https://openaccess.thecvf.com/content/CVPR2022/html/Zhao_Task-Specific_Inconsistency_Alignment_for_Domain_Adaptive_Object_Detection_CVPR_2022_paper.html}{Task-Specific Inconsistency Alignment for Domain Adaptive Object Detection} \citep{corpus009} & M/F \\
010 & S1 & \href{https://openaccess.thecvf.com/content/CVPR2022/html/Liu_Towards_Robust_Adaptive_Object_Detection_Under_Noisy_Annotations_CVPR_2022_paper.html}{Towards Robust Adaptive Object Detection Under Noisy Annotations} \citep{corpus010} & M/F \\
011 & S1 & \href{https://openaccess.thecvf.com/content/CVPR2022/html/Li_Class-Balanced_Pixel-Level_Self-Labeling_for_Domain_Adaptive_Semantic_Segmentation_CVPR_2022_paper.html}{Class-Balanced Pixel-Level Self-Labeling for Domain Adaptive Semantic Segmentation} \citep{corpus011} & M/F \\
012 & S1 & \href{https://openaccess.thecvf.com/content/CVPR2022/html/Ding_Decoupling_Zero-Shot_Semantic_Segmentation_CVPR_2022_paper.html}{Decoupling Zero-Shot Semantic Segmentation} \citep{corpus012} & M/F \\
013 & S1 & \href{https://openaccess.thecvf.com/content/CVPR2022/html/Ru_Learning_Affinity_From_Attention_End-to-End_Weakly-Supervised_Semantic_Segmentation_With_Transformers_CVPR_2022_paper.html}{Learning Affinity From Attention: End-to-End Weakly-Supervised Semantic Segmentation With Transformers} \citep{corpus013} & M/F \\
014 & S1 & \href{https://openaccess.thecvf.com/content/CVPR2022/html/Xu_Multi-Class_Token_Transformer_for_Weakly_Supervised_Semantic_Segmentation_CVPR_2022_paper.html}{Multi-Class Token Transformer for Weakly Supervised Semantic Segmentation} \citep{corpus014} & M/F \\
015 & S1 & \href{https://openaccess.thecvf.com/content/CVPR2022/html/Zhao_Novel_Class_Discovery_in_Semantic_Segmentation_CVPR_2022_paper.html}{Novel Class Discovery in Semantic Segmentation} \citep{corpus015} & M/F \\
016 & S1 & \href{https://openaccess.thecvf.com/content/CVPR2022/html/Wang_Open-World_Instance_Segmentation_Exploiting_Pseudo_Ground_Truth_From_Learned_Pairwise_CVPR_2022_paper.html}{Open-World Instance Segmentation: Exploiting Pseudo Ground Truth From Learned Pairwise Affinity} \citep{corpus016} & M/F \\
017 & S1 & \href{https://openaccess.thecvf.com/content/CVPR2022/html/Zhang_Representation_Compensation_Networks_for_Continual_Semantic_Segmentation_CVPR_2022_paper.html}{Representation Compensation Networks for Continual Semantic Segmentation} \citep{corpus017} & M/F \\
018 & S1 & \href{https://openaccess.thecvf.com/content/CVPR2022/html/Zhou_Rethinking_Semantic_Segmentation_A_Prototype_View_CVPR_2022_paper.html}{Rethinking Semantic Segmentation: A Prototype View} \citep{corpus018} & M/F \\
019 & S1 & \href{https://openaccess.thecvf.com/content/CVPR2022/html/Guo_SimT_Handling_Open-Set_Noise_for_Domain_Adaptive_Semantic_Segmentation_CVPR_2022_paper.html}{SimT: Handling Open-Set Noise for Domain Adaptive Semantic Segmentation} \citep{corpus019} & M/F \\
020 & S1 & \href{https://openaccess.thecvf.com/content/CVPR2022/html/Liang_Tree_Energy_Loss_Towards_Sparsely_Annotated_Semantic_Segmentation_CVPR_2022_paper.html}{Tree Energy Loss: Towards Sparsely Annotated Semantic Segmentation} \citep{corpus020} & M/F \\
021 & S1 & \href{https://openaccess.thecvf.com/content/CVPR2022/html/Ma_A_Text_Attention_Network_for_Spatial_Deformation_Robust_Scene_Text_CVPR_2022_paper.html}{A Text Attention Network for Spatial Deformation Robust Scene Text Image Super-Resolution} \citep{corpus021} & M/F \\
022 & S1 & \href{https://openaccess.thecvf.com/content/CVPR2022/html/Yue_Blind_Image_Super-Resolution_With_Elaborate_Degradation_Modeling_on_Noise_and_CVPR_2022_paper.html}{Blind Image Super-Resolution With Elaborate Degradation Modeling on Noise and Kernel} \citep{corpus022} & M/F \\
023 & S1 & \href{https://openaccess.thecvf.com/content/CVPR2022/html/Whang_Deblurring_via_Stochastic_Refinement_CVPR_2022_paper.html}{Deblurring via Stochastic Refinement} \citep{corpus023} & M/F \\
024 & S1 & \href{https://openaccess.thecvf.com/content/CVPR2022/html/Luo_Deep_Constrained_Least_Squares_for_Blind_Image_Super-Resolution_CVPR_2022_paper.html}{Deep Constrained Least Squares for Blind Image Super-Resolution} \citep{corpus024} & M/F \\
025 & S1 & \href{https://openaccess.thecvf.com/content/CVPR2022/html/Liang_Details_or_Artifacts_A_Locally_Discriminative_Learning_Approach_to_Realistic_CVPR_2022_paper.html}{Details or Artifacts: A Locally Discriminative Learning Approach to Realistic Image Super-Resolution} \citep{corpus025} & M/F \\
026 & S1 & \href{https://openaccess.thecvf.com/content/CVPR2022/html/Xu_Dual_Adversarial_Adaptation_for_Cross-Device_Real-World_Image_Super-Resolution_CVPR_2022_paper.html}{Dual Adversarial Adaptation for Cross-Device Real-World Image Super-Resolution} \citep{corpus026} & M/F \\
027 & S1 & \href{https://openaccess.thecvf.com/content/CVPR2021/html/Pan_Dual_Pixel_Exploration_Simultaneous_Depth_Estimation_and_Image_Restoration_CVPR_2021_paper.html}{Dual Pixel Exploration: Simultaneous Depth Estimation and Image Restoration} \citep{corpus027} & M/F \\
028 & S1 & \href{https://openaccess.thecvf.com/content/CVPR2022/html/Zhang_Exploring_and_Evaluating_Image_Restoration_Potential_in_Dynamic_Scenes_CVPR_2022_paper.html}{Exploring and Evaluating Image Restoration Potential in Dynamic Scenes} \citep{corpus028} & M/F \\
029 & S1 & \href{https://openaccess.thecvf.com/content/CVPR2021/html/Son_SRWarp_Generalized_Image_Super-Resolution_under_Arbitrary_Transformation_CVPR_2021_paper.html}{SRWarp: Generalized Image Super-Resolution under Arbitrary Transformation} \citep{corpus029} & M/F \\
030 & S1 & \href{https://openaccess.thecvf.com/content/CVPR2021/html/Wei_Unsupervised_Real-World_Image_Super_Resolution_via_Domain-Distance_Aware_Training_CVPR_2021_paper.html}{Unsupervised Real-World Image Super Resolution via Domain-Distance Aware Training} \citep{corpus030} & M/F \\
031 & S1 & \href{https://openaccess.thecvf.com/content/CVPR2021/html/Bechtold_Fostering_Generalization_in_Single-View_3D_Reconstruction_by_Learning_a_Hierarchy_CVPR_2021_paper.html}{Fostering Generalization in Single-View 3D Reconstruction by Learning a Hierarchy of Local and Global Shape Priors} \citep{corpus031} & M/F \\
032 & S1 & \href{https://openaccess.thecvf.com/content/CVPR2021/html/Engelmann_From_Points_to_Multi-Object_3D_Reconstruction_CVPR_2021_paper.html}{From Points to Multi-Object 3D Reconstruction} \citep{corpus032} & M/F \\
033 & S1 & \href{https://openaccess.thecvf.com/content_CVPR_2019/html/Kato_Learning_View_Priors_for_Single-View_3D_Reconstruction_CVPR_2019_paper.html}{Learning View Priors for Single-View 3D Reconstruction} \citep{corpus033} & M/F \\
034 & S1 & \href{https://openaccess.thecvf.com/content/CVPR2021/html/Sun_NeuralRecon_Real-Time_Coherent_3D_Reconstruction_From_Monocular_Video_CVPR_2021_paper.html}{NeuralRecon: Real-Time Coherent 3D Reconstruction From Monocular Video} \citep{corpus034} & M/F \\
035 & S1 & \href{https://openaccess.thecvf.com/content_CVPR_2019/html/Mescheder_Occupancy_Networks_Learning_3D_Reconstruction_in_Function_Space_CVPR_2019_paper.html}{Occupancy Networks: Learning 3D Reconstruction in Function Space} \citep{corpus035} & M/F \\
036 & S1 & \href{https://openaccess.thecvf.com/content/CVPR2022/html/Alldieck_Photorealistic_Monocular_3D_Reconstruction_of_Humans_Wearing_Clothing_CVPR_2022_paper.html}{Photorealistic Monocular 3D Reconstruction of Humans Wearing Clothing} \citep{corpus036} & M/F \\
037 & S1 & \href{https://openaccess.thecvf.com/content/CVPR2022/html/Alwala_Pre-Train_Self-Train_Distill_A_Simple_Recipe_for_Supersizing_3D_Reconstruction_CVPR_2022_paper.html}{Pre-Train, Self-Train, Distill: A Simple Recipe for Supersizing 3D Reconstruction} \citep{corpus037} & M/F \\
038 & S1 & \href{https://openaccess.thecvf.com/content_CVPR_2019/html/Yu_Single-Image_Piece-Wise_Planar_3D_Reconstruction_via_Associative_Embedding_CVPR_2019_paper.html}{Single-Image Piece-Wise Planar 3D Reconstruction via Associative Embedding} \citep{corpus038} & M/F \\
039 & S1 & \href{https://openaccess.thecvf.com/content/CVPR2022/html/Ma_Surface_Reconstruction_From_Point_Clouds_by_Learning_Predictive_Context_Priors_CVPR_2022_paper.html}{Surface Reconstruction From Point Clouds by Learning Predictive Context Priors} \citep{corpus039} & M/F \\
040 & S1 & \href{https://doi.org/10.1109/cvpr52688.2022.00159}{Topologically-Aware Deformation Fields for Single-View 3D Reconstruction} \citep{corpus040} & M/F \\
041 & S2 & \href{https://openaccess.thecvf.com/content/CVPR2024/html/Kennerley_CAT_Exploiting_Inter-Class_Dynamics_for_Domain_Adaptive_Object_Detection_CVPR_2024_paper.html}{CAT: Exploiting Inter-Class Dynamics for Domain Adaptive Object Detection} \citep{corpus041} & M/F \\
042 & S2 & \href{https://openaccess.thecvf.com/content/CVPR2025/html/Wang_Efficient_Test-time_Adaptive_Object_Detection_via_Sensitivity-Guided_Pruning_CVPR_2025_paper.html}{Efficient Test-time Adaptive Object Detection via Sensitivity-Guided Pruning} \citep{corpus042} & M/F \\
043 & S2 & \href{https://openaccess.thecvf.com/content/CVPR2025/html/Alexandridis_Fractal_Calibration_for_Long-tailed_Object_Detection_CVPR_2025_paper.html}{Fractal Calibration for Long-tailed Object Detection} \citep{corpus043} & M/F \\
044 & S2 & \href{https://openaccess.thecvf.com/content/CVPR2024/html/Lin_Generative_Region-Language_Pretraining_for_Open-Ended_Object_Detection_CVPR_2024_paper.html}{Generative Region-Language Pretraining for Open-Ended Object Detection} \citep{corpus044} & M/F \\
045 & S2 & \href{https://openaccess.thecvf.com/content/CVPR2024/html/Feng_InstaGen_Enhancing_Object_Detection_by_Training_on_Synthetic_Dataset_CVPR_2024_paper.html}{InstaGen: Enhancing Object Detection by Training on Synthetic Dataset} \citep{corpus045} & M/F \\
046 & S2 & \href{https://openaccess.thecvf.com/content/CVPR2025/html/Lavoie_Large_Self-Supervised_Models_Bridge_the_Gap_in_Domain_Adaptive_Object_CVPR_2025_paper.html}{Large Self-Supervised Models Bridge the Gap in Domain Adaptive Object Detection} \citep{corpus046} & M/F \\
047 & S2 & \href{https://openaccess.thecvf.com/content/CVPR2025/html/Fu_LLMDet_Learning_Strong_Open-Vocabulary_Object_Detectors_under_the_Supervision_of_CVPR_2025_paper.html}{LLMDet: Learning Strong Open-Vocabulary Object Detectors under the Supervision of Large Language Models} \citep{corpus047} & M/F \\
048 & S2 & \href{https://openaccess.thecvf.com/content/CVPR2026/html/Liu_Partial_Weakly-Supervised_Oriented_Object_Detection_CVPR_2026_paper.html}{Partial Weakly-Supervised Oriented Object Detection} \citep{corpus048} & M/F \\
049 & S2 & \href{https://openaccess.thecvf.com/content/CVPR2024/html/Luo_PointOBB_Learning_Oriented_Object_Detection_via_Single_Point_Supervision_CVPR_2024_paper.html}{PointOBB: Learning Oriented Object Detection via Single Point Supervision} \citep{corpus049} & M/F \\
050 & S2 & \href{https://openaccess.thecvf.com/content/CVPR2025/html/Sidhu_Search_and_Detect_Training-Free_Long_Tail_Object_Detection_via_Web-Image_CVPR_2025_paper.html}{Search and Detect: Training-Free Long Tail Object Detection via Web-Image Retrieval} \citep{corpus050} & M/F \\
051 & S2 & \href{https://openaccess.thecvf.com/content/CVPR2025/html/Yin_Beyond_Background_Shift_Rethinking_Instance_Replay_in_Continual_Semantic_Segmentation_CVPR_2025_paper.html}{Beyond Background Shift: Rethinking Instance Replay in Continual Semantic Segmentation} \citep{corpus051} & M/F \\
052 & S2 & \href{https://openaccess.thecvf.com/content/CVPR2025/html/Yin_DFormerv2_Geometry_Self-Attention_for_RGBD_Semantic_Segmentation_CVPR_2025_paper.html}{DFormerv2: Geometry Self-Attention for RGBD Semantic Segmentation} \citep{corpus052} & M/F \\
053 & S2 & \href{https://openaccess.thecvf.com/content/CVPR2025/html/Kim_Distilling_Spectral_Graph_for_Object-Context_Aware_Open-Vocabulary_Semantic_Segmentation_CVPR_2025_paper.html}{Distilling Spectral Graph for Object-Context Aware Open-Vocabulary Semantic Segmentation} \citep{corpus053} & M/F \\
054 & S2 & \href{https://openaccess.thecvf.com/content/CVPR2025/html/Zhao_DPSeg_Dual-Prompt_Cost_Volume_Learning_for_Open-Vocabulary_Semantic_Segmentation_CVPR_2025_paper.html}{DPSeg: Dual-Prompt Cost Volume Learning for Open-Vocabulary Semantic Segmentation} \citep{corpus054} & M/F \\
055 & S2 & \href{https://openaccess.thecvf.com/content/CVPR2024/html/Zhang_Frozen_CLIP_A_Strong_Backbone_for_Weakly_Supervised_Semantic_Segmentation_CVPR_2024_paper.html}{Frozen CLIP: A Strong Backbone for Weakly Supervised Semantic Segmentation} \citep{corpus055} & M/F \\
056 & S2 & \href{https://openaccess.thecvf.com/content/CVPR2024/html/Tang_Hunting_Attributes_Context_Prototype-Aware_Learning_for_Weakly_Supervised_Semantic_Segmentation_CVPR_2024_paper.html}{Hunting Attributes: Context Prototype-Aware Learning for Weakly Supervised Semantic Segmentation} \citep{corpus056} & M/F \\
057 & S2 & \href{https://openaccess.thecvf.com/content/CVPR2024/html/Wang_Image-to-Image_Matching_via_Foundation_Models_A_New_Perspective_for_Open-Vocabulary_CVPR_2024_paper.html}{Image-to-Image Matching via Foundation Models: A New Perspective for Open-Vocabulary Semantic Segmentation} \citep{corpus057} & M/F \\
058 & S2 & \href{https://openaccess.thecvf.com/content/CVPR2025/html/Zhang_Mamba_as_a_Bridge_Where_Vision_Foundation_Models_Meet_Vision_CVPR_2025_paper.html}{Mamba as a Bridge: Where Vision Foundation Models Meet Vision Language Models for Domain-Generalized Semantic Segmentation} \citep{corpus058} & M/F \\
059 & S2 & \href{https://openaccess.thecvf.com/content/CVPR2025/html/Xie_MaSS13K_A_Matting-level_Semantic_Segmentation_Benchmark_CVPR_2025_paper.html}{MaSS13K: A Matting-level Semantic Segmentation Benchmark} \citep{corpus059} & M/F \\
060 & S2 & \href{https://openaccess.thecvf.com/content/CVPR2025/html/Fu_SegMAN_Omni-scale_Context_Modeling_with_State_Space_Models_and_Local_CVPR_2025_paper.html}{SegMAN: Omni-scale Context Modeling with State Space Models and Local Attention for Semantic Segmentation} \citep{corpus060} & M/F \\
061 & S2 & \href{https://openaccess.thecvf.com/content/CVPR2025/html/Chen_Adversarial_Diffusion_Compression_for_Real-World_Image_Super-Resolution_CVPR_2025_paper.html}{Adversarial Diffusion Compression for Real-World Image Super-Resolution} \citep{corpus061} & M/F \\
062 & S2 & \href{https://openaccess.thecvf.com/content/CVPR2025/html/Zamfir_Complexity_Experts_are_Task-Discriminative_Learners_for_Any_Image_Restoration_CVPR_2025_paper.html}{Complexity Experts are Task-Discriminative Learners for Any Image Restoration} \citep{corpus062} & M/F \\
063 & S2 & \href{https://openaccess.thecvf.com/content/CVPR2025/html/Wang_DORNet_A_Degradation_Oriented_and_Regularized_Network_for_Blind_Depth_CVPR_2025_paper.html}{DORNet: A Degradation Oriented and Regularized Network for Blind Depth Super-Resolution} \citep{corpus063} & M/F \\
064 & S2 & \href{https://openaccess.thecvf.com/content/CVPR2025/html/Chen_FaithDiff_Unleashing_Diffusion_Priors_for_Faithful_Image_Super-resolution_CVPR_2025_paper.html}{FaithDiff: Unleashing Diffusion Priors for Faithful Image Super-resolution} \citep{corpus064} & M/F \\
065 & S2 & \href{https://openaccess.thecvf.com/content/CVPR2025/html/Zhao_From_Zero_to_Detail_Deconstructing_Ultra-High-Definition_Image_Restoration_from_Progressive_CVPR_2025_paper.html}{From Zero to Detail: Deconstructing Ultra-High-Definition Image Restoration from Progressive Spectral Perspective} \citep{corpus065} & M/F \\
066 & S2 & \href{https://openaccess.thecvf.com/content/CVPR2025/html/Rajagopalan_GenDeg_Diffusion-based_Degradation_Synthesis_for_Generalizable_All-In-One_Image_Restoration_CVPR_2025_paper.html}{GenDeg: Diffusion-based Degradation Synthesis for Generalizable All-In-One Image Restoration} \citep{corpus066} & M/F \\
067 & S2 & \href{https://openaccess.thecvf.com/content/CVPR2025/html/Zhu_PassionSR_Post-Training_Quantization_with_Adaptive_Scale_in_One-Step_Diffusion_based_CVPR_2025_paper.html}{PassionSR: Post-Training Quantization with Adaptive Scale in One-Step Diffusion based Image Super-Resolution} \citep{corpus067} & M/F \\
068 & S2 & \href{https://openaccess.thecvf.com/content/CVPR2025/html/Di_QMambaBSR_Burst_Image_Super-Resolution_with_Query_State_Space_Model_CVPR_2025_paper.html}{QMambaBSR: Burst Image Super-Resolution with Query State Space Model} \citep{corpus068} & M/F \\
069 & S2 & \href{https://openaccess.thecvf.com/content/CVPR2025/html/Liu_Rotation-Equivariant_Self-Supervised_Method_in_Image_Denoising_CVPR_2025_paper.html}{Rotation-Equivariant Self-Supervised Method in Image Denoising} \citep{corpus069} & M/F \\
070 & S2 & \href{https://openaccess.thecvf.com/content/CVPR2025/html/Mei_The_Power_of_Context_How_Multimodality_Improves_Image_Super-Resolution_CVPR_2025_paper.html}{The Power of Context: How Multimodality Improves Image Super-Resolution} \citep{corpus070} & M/F \\
071 & S2 & \href{https://openaccess.thecvf.com/content/CVPR2025/html/Hu_A_Lightweight_UDF_Learning_Framework_for_3D_Reconstruction_Based_on_CVPR_2025_paper.html}{A Lightweight UDF Learning Framework for 3D Reconstruction Based on Local Shape Functions} \citep{corpus071} & M/F \\
072 & S2 & \href{https://openaccess.thecvf.com/content/CVPR2024/html/Xie_Carve3D_Improving_Multi-view_Reconstruction_Consistency_for_Diffusion_Models_with_RL_CVPR_2024_paper.html}{Carve3D: Improving Multi-view Reconstruction Consistency for Diffusion Models with RL Finetuning} \citep{corpus072} & M/F \\
073 & S2 & \href{https://openaccess.thecvf.com/content/CVPR2025/html/Walker_CrossSDF_3D_Reconstruction_of_Thin_Structures_From_Cross-Sections_CVPR_2025_paper.html}{CrossSDF: 3D Reconstruction of Thin Structures From Cross-Sections} \citep{corpus073} & M/F \\
074 & S2 & \href{https://openaccess.thecvf.com/content/CVPR2024/html/Sengupta_DiffHuman_Probabilistic_Photorealistic_3D_Reconstruction_of_Humans_CVPR_2024_paper.html}{DiffHuman: Probabilistic Photorealistic 3D Reconstruction of Humans} \citep{corpus074} & M/F \\
075 & S2 & \href{https://openaccess.thecvf.com/content/CVPR2025/html/Alzayer_Generative_Multiview_Relighting_for_3D_Reconstruction_under_Extreme_Illumination_Variation_CVPR_2025_paper.html}{Generative Multiview Relighting for 3D Reconstruction under Extreme Illumination Variation} \citep{corpus075} & M/F \\
076 & S2 & \href{https://openaccess.thecvf.com/content/CVPR2024/html/Chen_GenNBV_Generalizable_Next-Best-View_Policy_for_Active_3D_Reconstruction_CVPR_2024_paper.html}{GenNBV: Generalizable Next-Best-View Policy for Active 3D Reconstruction} \citep{corpus076} & M/F \\
077 & S2 & \href{https://openaccess.thecvf.com/content/CVPR2024/html/Han_NeRSP_Neural_3D_Reconstruction_for_Reflective_Objects_with_Sparse_Polarized_CVPR_2024_paper.html}{NeRSP: Neural 3D Reconstruction for Reflective Objects with Sparse Polarized Images} \citep{corpus077} & M/F \\
078 & S2 & \href{https://openaccess.thecvf.com/content/CVPR2025/html/Chen_Parametric_Point_Cloud_Completion_for_Polygonal_Surface_Reconstruction_CVPR_2025_paper.html}{Parametric Point Cloud Completion for Polygonal Surface Reconstruction} \citep{corpus078} & M/F \\
079 & S2 & \href{https://openaccess.thecvf.com/content/CVPR2025/html/Tan_PlanarSplatting_Accurate_Planar_Surface_Reconstruction_in_3_Minutes_CVPR_2025_paper.html}{PlanarSplatting: Accurate Planar Surface Reconstruction in 3 Minutes} \citep{corpus079} & M/F \\
080 & S2 & \href{https://openaccess.thecvf.com/content/CVPR2025/html/Jang_Pow3R_Empowering_Unconstrained_3D_Reconstruction_with_Camera_and_Scene_Priors_CVPR_2025_paper.html}{Pow3R: Empowering Unconstrained 3D Reconstruction with Camera and Scene Priors} \citep{corpus080} & M/F \\
081 & S3 & \href{https://openaccess.thecvf.com/content/CVPR2026/html/Zhang_Beyond_Prompt_Degradation_Prototype-guided_Dual-pool_Prompting_for_Incremental_Object_Detection_CVPR_2026_paper.html}{Beyond Prompt Degradation: Prototype-guided Dual-pool Prompting for Incremental Object Detection} \citep{corpus081} & M/F \\
082 & S3 & \href{https://openaccess.thecvf.com/content/CVPR2026/html/Monga_EW-DETR_Evolving_World_Object_Detection_via_Incremental_Low-Rank_DEtection_TRansformer_CVPR_2026_paper.html}{EW-DETR: Evolving World Object Detection via Incremental Low-Rank DEtection TRansformer} \citep{corpus082} & M/F \\
083 & S3 & \href{https://openaccess.thecvf.com/content/CVPR2026/html/Gu_Fourier_Angle_Alignment_for_Oriented_Object_Detection_in_Remote_Sensing_CVPR_2026_paper.html}{Fourier Angle Alignment for Oriented Object Detection in Remote Sensing} \citep{corpus083} & M/F \\
084 & S3 & \href{https://openaccess.thecvf.com/content/CVPR2026/html/Zhang_Heuristic-inspired_Reasoning_Priors_Facilitate_Data-Efficient_Referring_Object_Detection_CVPR_2026_paper.html}{Heuristic-inspired Reasoning Priors Facilitate Data-Efficient Referring Object Detection} \citep{corpus084} & M/F \\
085 & S3 & \href{https://arxiv.org/abs/2602.06484}{Instance-Free Domain Adaptive Object Detection} \citep{corpus085} & M/F \\
086 & S3 & \href{https://doi.org/10.1007/978-3-032-37574-2_20}{IoUCert: Robustness Verification for Anchor-Based Object Detectors} \citep{corpus086} & M/F \\
087 & S3 & \href{https://arxiv.org/abs/2602.06474}{LAB-Det: Language as a Domain-Invariant Bridge for Training-Free One-Shot Domain Generalization in Object Detection} \citep{corpus087} & M/F \\
088 & S3 & \href{https://openaccess.thecvf.com/content/CVPR2026/html/Zhang_NoOVD_Novel_Category_Discovery_and_Embedding_for_Open-Vocabulary_Object_Detection_CVPR_2026_paper.html}{NoOVD: Novel Category Discovery and Embedding for Open-Vocabulary Object Detection} \citep{corpus088} & M/F \\
089 & S3 & \href{https://doi.org/10.1007/978-3-032-37083-9_26}{RiO-DETR: DETR for Real-Time Oriented Object Detection} \citep{corpus089} & M/F \\
090 & S3 & \href{https://doi.org/10.1007/978-3-032-37016-7_27}{VLOD-TTA: Test-Time Adaptation of Vision-Language Object Detectors} \citep{corpus090} & M/F \\
091 & S3 & \href{https://openaccess.thecvf.com/content/CVPR2026/html/Li_Direct_Segmentation_without_Logits_Optimization_for_Training-Free_Open-Vocabulary_Semantic_Segmentation_CVPR_2026_paper.html}{Direct Segmentation without Logits Optimization for Training-Free Open-Vocabulary Semantic Segmentation} \citep{corpus091} & M/F \\
092 & S3 & \href{https://openreview.net/forum?id=4bHFNKe8OU}{Do Instance Priors Help Weakly Supervised Semantic Segmentation?} \citep{corpus092} & M/F \\
093 & S3 & \href{https://openaccess.thecvf.com/content/CVPR2026/html/Lv_Generalizable_Knowledge_Distillation_from_Vision_Foundation_Models_for_Semantic_Segmentation_CVPR_2026_paper.html}{Generalizable Knowledge Distillation from Vision Foundation Models for Semantic Segmentation} \citep{corpus093} & M/F \\
094 & S3 & \href{https://arxiv.org/abs/2603.27206}{Make It Up: Fake Images, Real Gains in Generalized Few-shot Semantic Segmentation} \citep{corpus094} & M/F \\
095 & S3 & \href{https://openaccess.thecvf.com/content/CVPR2026/html/Li_MARIS_Marine_Open-Vocabulary_Instance_Segmentation_CVPR_2026_paper.html}{MARIS: Marine Open-Vocabulary Instance Segmentation} \citep{corpus095} & M/F \\
096 & S3 & \href{https://openaccess.thecvf.com/content/CVPR2026/html/Pei_PEARL_Geometry_Aligns_Semantics_for_Training-Free_Open-Vocabulary_Semantic_Segmentation_CVPR_2026_paper.html}{PEARL: Geometry Aligns Semantics for Training-Free Open-Vocabulary Semantic Segmentation} \citep{corpus096} & M/F \\
097 & S3 & \href{https://openaccess.thecvf.com/content/CVPR2026/html/Li_REL-SF4PASS_Panoramic_Semantic_Segmentation_with_REL_Depth_Representation_and_Spherical_CVPR_2026_paper.html}{REL-SF4PASS: Panoramic Semantic Segmentation with REL Depth Representation and Spherical Fusion} \citep{corpus097} & M/F \\
098 & S3 & \href{https://openaccess.thecvf.com/content/CVPR2026/html/Ernst_Rewis3d_Reconstruction_Improves_Weakly-Supervised_Semantic_Segmentation_CVPR_2026_paper.html}{Rewis3d: Reconstruction Improves Weakly-Supervised Semantic Segmentation} \citep{corpus098} & M/F \\
099 & S3 & \href{https://openaccess.thecvf.com/content/CVPR2026/html/Li_SAGE_Style-Adaptive_Generalization_for_Privacy-Constrained_Semantic_Segmentation_Across_Domains_CVPR_2026_paper.html}{SAGE: Style-Adaptive Generalization for Privacy-Constrained Semantic Segmentation Across Domains} \citep{corpus099} & M/F \\
100 & S3 & \href{https://arxiv.org/abs/2512.08730}{SegEarth-OV3: Exploring SAM 3 for Open-Vocabulary Semantic Segmentation in Remote Sensing Images} \citep{corpus100} & M/F \\
101 & S3 & \href{https://openaccess.thecvf.com/content/CVPR2026/html/Ryou_Beyond_the_Ground_Truth_Enhanced_Supervision_for_Image_Restoration_CVPR_2026_paper.html}{Beyond the Ground Truth: Enhanced Supervision for Image Restoration} \citep{corpus101} & M/F \\
102 & S3 & \href{https://openaccess.thecvf.com/content/CVPR2026/html/Zhang_Enhancing_Unregistered_Hyperspectral_Image_Super-Resolution_via_Unmixing-based_Abundance_Fusion_Learning_CVPR_2026_paper.html}{Enhancing Unregistered Hyperspectral Image Super-Resolution via Unmixing-based Abundance Fusion Learning} \citep{corpus102} & M/F \\
103 & S3 & \href{https://openaccess.thecvf.com/content/CVPR2026/html/Liu_FinPercep-RM_A_Fine-grained_Reward_Model_and_Co-evolutionary_Curriculum_for_RL-based_CVPR_2026_paper.html}{FinPercep-RM: A Fine-grained Reward Model and Co-evolutionary Curriculum for RL-based Real-world Super-Resolution} \citep{corpus103} & M/F \\
104 & S3 & \href{https://openaccess.thecvf.com/content/CVPR2026/html/Yi_GDPO-SR_Group_Direct_Preference_Optimization_for_One-Step_Generative_Image_Super-Resolution_CVPR_2026_paper.html}{GDPO-SR: Group Direct Preference Optimization for One-Step Generative Image Super-Resolution} \citep{corpus104} & M/F \\
105 & S3 & \href{https://openaccess.thecvf.com/content/CVPR2026/html/Fang_One-Step_Diffusion_Transformer_for_Controllable_Real-World_Image_Super-Resolution_CVPR_2026_paper.html}{One-Step Diffusion Transformer for Controllable Real-World Image Super-Resolution} \citep{corpus105} & M/F \\
106 & S3 & \href{https://doi.org/10.1007/978-3-032-37324-3_6}{RefReward-SR: LR-Conditioned Reward Modeling for Preference-Aligned Super-Resolution} \citep{corpus106} & M/F \\
107 & S3 & \href{https://openaccess.thecvf.com/content/CVPR2026/html/Wang_Residual_Diffusion_Bridge_Model_for_Image_Restoration_CVPR_2026_paper.html}{Residual Diffusion Bridge Model for Image Restoration} \citep{corpus107} & M/F \\
108 & S3 & \href{https://openaccess.thecvf.com/content/CVPR2026/html/Li_Statistical_Characteristic-Guided_Denoising_for_Rapid_High-Resolution_Transmission_Electron_Microscopy_Imaging_CVPR_2026_paper.html}{Statistical Characteristic-Guided Denoising for Rapid High-Resolution Transmission Electron Microscopy Imaging} \citep{corpus108} & M/F \\
109 & S3 & \href{https://doi.org/10.1007/978-3-032-37281-9_32}{TIR-Agent: Training an Explorative and Efficient Agent for Image Restoration} \citep{corpus109} & M/F \\
110 & S3 & \href{https://openaccess.thecvf.com/content/CVPR2026/html/Li_UARE_A_Unified_Vision-Language_Model_for_Image_Quality_Assessment_Restoration_CVPR_2026_paper.html}{UARE: A Unified Vision-Language Model for Image Quality Assessment, Restoration, and Enhancement} \citep{corpus110} & M/F \\
111 & S3 & \href{https://openaccess.thecvf.com/content/CVPR2026/html/Huang_CUPID_Generative_3D_Reconstruction_via_Joint_Object_and_Pose_Modeling_CVPR_2026_paper.html}{CUPID: Generative 3D Reconstruction via Joint Object and Pose Modeling} \citep{corpus111} & M/F \\
112 & S3 & \href{https://openaccess.thecvf.com/content/CVPR2026/html/Zhao_E-RayZer_Self-supervised_3D_Reconstruction_as_Spatial_Visual_Pre-training_CVPR_2026_paper.html}{E-RayZer: Self-supervised 3D Reconstruction as Spatial Visual Pre-training} \citep{corpus112} & M/F \\
113 & S3 & \href{https://openaccess.thecvf.com/content/CVPR2026/html/He_EMGauss_Continuous_Slice-to-3D_Reconstruction_via_Dynamic_Gaussian_Modeling_in_Volume_CVPR_2026_paper.html}{EMGauss: Continuous Slice-to-3D Reconstruction via Dynamic Gaussian Modeling in Volume Electron Microscopy} \citep{corpus113} & M/F \\
114 & S3 & \href{https://openaccess.thecvf.com/content/CVPR2026/html/Huang_From_None_to_All_Self-Supervised_3D_Reconstruction_via_Novel_View_CVPR_2026_paper.html}{From None to All: Self-Supervised 3D Reconstruction via Novel View Synthesis} \citep{corpus114} & M/F \\
115 & S3 & \href{https://openaccess.thecvf.com/content/CVPR2026/html/Hu_G2VLM_Geometry_Grounded_Vision_Language_Model_with_Unified_3D_Reconstruction_CVPR_2026_paper.html}{G$^2$VLM: Geometry Grounded Vision Language Model with Unified 3D Reconstruction and Spatial Reasoning} \citep{corpus115} & M/F \\
116 & S3 & \href{https://openaccess.thecvf.com/content/CVPR2026/html/Guo_PanoVGGT_Feed-Forward_3D_Reconstruction_from_Panoramic_Imagery_CVPR_2026_paper.html}{PanoVGGT: Feed-Forward 3D Reconstruction from Panoramic Imagery} \citep{corpus116} & M/F \\
117 & S3 & \href{https://openaccess.thecvf.com/content/CVPR2026/html/Wang_PoInit-of-View_Poisoning_Initialization_of_Views_Transfers_Across_Multiple_3D_Reconstruction_CVPR_2026_paper.html}{PoInit-of-View: Poisoning Initialization of Views Transfers Across Multiple 3D Reconstruction Systems} \citep{corpus117} & M/F \\
118 & S3 & \href{https://doi.org/10.1007/978-3-032-37135-5_12}{RayMap3R: Inference-Time RayMap for Dynamic 3D Reconstruction} \citep{corpus118} & M/F \\
119 & S3 & \href{https://openaccess.thecvf.com/content/CVPR2026/html/Qi_Seeing_through_Light_and_Darkness_Sensor-Physics_Grounded_Deblurring_HDR_NeRF_CVPR_2026_paper.html}{Seeing through Light and Darkness: Sensor-Physics Grounded Deblurring HDR NeRF from Single-Exposure Images and Events} \citep{corpus119} & M/F \\
120 & S3 & \href{https://openaccess.thecvf.com/content/CVPR2026/html/Asthana_SplatSuRe_Selective_Super-Resolution_for_Multi-view_Consistent_3D_Gaussian_Splatting_CVPR_2026_paper.html}{SplatSuRe: Selective Super-Resolution for Multi-view Consistent 3D Gaussian Splatting} \citep{corpus120} & M/F \\
121 & S4 & \href{https://openreview.net/forum?id=8qFCNFz8Tq}{AnyDepth-DETR/-YOLO: Any-depth object detection with a single network} \citep{corpus121} & M/F \\
122 & S4 & \href{https://openreview.net/forum?id=6kDoAMVPn5}{CANDIDATE-BOX SELECTION MATTERS FOR CROSS-DOMAIN FEW-SHOT OBJECT DETECTION} \citep{corpus122} & M/F \\
123 & S4 & \href{https://openreview.net/forum?id=WrLCWdkY49}{DCHA: Dynamic Compositional Adaptation with Hierarchical Drift Constraints for Incremental Vision-Language Object Detection} \citep{corpus123} & M/F \\
124 & S4 & \href{https://openreview.net/forum?id=8P9XSKFW9N}{DeCo: A Multimodal Detection Framework based on Density Consensus Learning} \citep{corpus124} & M/F \\
125 & S4 & \href{https://openreview.net/forum?id=AUmtOiMvCa}{Different Instances, Different Supervision: Instance-Adaptive Point-Supervised Oriented Object Detection} \citep{corpus125} & M/F \\
126 & S4 & \href{https://openreview.net/forum?id=0jbdTGiGaz}{Imbalance-Adaptive Cross Consistency Learning for Semi-Supervised Oriented Object Detection} \citep{corpus126} & M/F \\
127 & S4 & \href{https://openreview.net/forum?id=ad0Os8QbxN}{PULSE: PERSISTENT UTILIZATION OF LIMITED SUPPORT EVIDENCE FOR CROSS-DOMAIN FEW-SHOT OBJECT DETECTION} \citep{corpus127} & M/F \\
128 & S4 & \href{https://openreview.net/forum?id=V5RmhtXTMG}{Rethinking Over-Segmentation in Few-Shot Object Detection with Vision Foundation Models: A Training-Free Prototype Refinement and Occlusion-Aware Discrimination Approach} \citep{corpus128} & M/F \\
129 & S4 & \href{https://openreview.net/forum?id=1I3yfptfr1}{Towards Trustworthy Object Detection: Adaptive Boundary Optimization with Joint Classification and Localization Risk Control} \citep{corpus129} & M/F \\
130 & S4 & \href{https://openreview.net/forum?id=vTYhIzoLzY}{UniOSE: Semantics-Enriched Object Queries for Open-Ended Detection and Segmentation} \citep{corpus130} & M/F \\
131 & S4 & \href{https://openreview.net/forum?id=ymk7OpPLY8}{Architectural Missing-Modality Robustness in Semantic Segmentation} \citep{corpus131} & M/F \\
132 & S4 & \href{https://openreview.net/forum?id=NYrrKvi2Dy}{Calibration-Guided State Selection and Class-Conditioned Fusion for Frozen Diffusion Segmentation} \citep{corpus132} & M/F \\
133 & S4 & \href{https://openreview.net/forum?id=8aErJ21sV6}{DevoSeg: Budget-Aware Self-Evolving Semantic Segmentation under Visual Degradation} \citep{corpus133} & M/F \\
134 & S4 & \href{https://openreview.net/forum?id=kv0BeIZGMk}{HIVE: Holistic Image-Adaptive Vocabulary Evolution for Training-Free Dense Vision-Language Inference} \citep{corpus134} & M/F \\
135 & S4 & \href{https://openreview.net/forum?id=9S1a6H8xUR}{Past as Guidance, Not Just Constraint: Cross-Level Knowledge Reorganization for Class-Incremental Semantic Segmentation} \citep{corpus135} & M/F \\
136 & S4 & \href{https://openreview.net/forum?id=dMBwWu68Lm}{ReTax-Seg: Visually Grounded Reversible Taxonomy Transport for Image Segmentation} \citep{corpus136} & M/F \\
137 & S4 & \href{https://openreview.net/forum?id=OHqdlezqWF}{Rethinking Unaligned RGB-T Dense Prediction as Cross-Modal Information Transport} \citep{corpus137} & M/F \\
138 & S4 & \href{https://openreview.net/forum?id=hEBqQJhg1b}{SEER: Semantic Evidence for Executable Region Propagation in Open-Vocabulary Semantic Segmentation} \citep{corpus138} & M/F \\
139 & S4 & \href{https://openreview.net/forum?id=67ApUu5dSo}{SHAPE: Structural Hierarchy of Affinities for Patch Enhancement in Training-Free Open-Vocabulary Semantic Segmentation} \citep{corpus139} & M/F \\
140 & S4 & \href{https://openreview.net/forum?id=J2V6tPOCod}{Where from Above, What from the Street: Learning Semantic Routing for Cross-View Segmentation under Pose Uncertainty} \citep{corpus140} & M/F \\
141 & S4 & \href{https://openreview.net/forum?id=v44KR7QgKE}{$\mathrm{S}^{3}$-PnP: Plug-and-Play Denoising for Scientific Imaging with Self-Supervised Score Priors} \citep{corpus141} & M/F \\
142 & S4 & \href{https://openreview.net/forum?id=IP9k9VdnnS}{Degradation Frequency Curve: An Explicit Frequency-Quantified Representation for All-in-One Image Restoration} \citep{corpus142} & M/F \\
143 & S4 & \href{https://openreview.net/forum?id=kQkih2eMHu}{Fixing the Last Mile: Failure-Aware Image Restoration with Region-Specific Guidance} \citep{corpus143} & M/F \\
144 & S4 & \href{https://openreview.net/forum?id=1n4L8nXB1I}{GGT-100K: Generative Ground Truth for Generalizable Real-World Image Restoration} \citep{corpus144} & M/F \\
145 & S4 & \href{https://openreview.net/forum?id=QlpWNkJcBr}{HarnessIR: Harnessing Multimodal Foundation Models for Universal Real-World Image Restoration} \citep{corpus145} & M/F \\
146 & S4 & \href{https://openreview.net/forum?id=Cdx9a9GdBy}{Learning to Restore More: Continual Capability Expansion for Pretrained Image Restoration Models} \citep{corpus146} & M/F \\
147 & S4 & \href{https://openreview.net/forum?id=rCgxaWvBRj}{Look Before You Restore: Learning to Predict Before Acting for Agentic Image Restoration} \citep{corpus147} & M/F \\
148 & S4 & \href{https://openreview.net/forum?id=anN7bIopnC}{PhoenixSR: Generative Heterogeneous Distillation Unleashes Efficient Models for Real-World Super-Resolution} \citep{corpus148} & M/F \\
149 & S4 & \href{https://openreview.net/forum?id=80XJOuh0nO}{Restoring without Forgetting: Filter-Level Continual Image Restoration via Parameter-Space Integrated Gradients} \citep{corpus149} & M/F \\
150 & S4 & \href{https://openreview.net/forum?id=3Kt188v0Yl}{SegDem: Segmentation helps Demosaicing} \citep{corpus150} & M/F \\
151 & S4 & \href{https://openreview.net/forum?id=hShILxckIE}{ARROW: Arbitrary Reconstruction and Tracking of 4D Observations in the Wild} \citep{corpus151} & M/F \\
152 & S4 & \href{https://openreview.net/forum?id=4hJ82iauw9}{Drive3R: Block-wise Streaming Geometric Foundation Models for Autonomous Driving} \citep{corpus152} & M/F \\
153 & S4 & \href{https://openreview.net/forum?id=iRCP1uMP8S}{Failure-Aware Adaptation for Transparent Objects in Feed-Forward 3D Reconstruction} \citep{corpus153} & M/F \\
154 & S4 & \href{https://openreview.net/forum?id=dPyGZiJESZ}{Latent Bridge: From Video Latents to 3D Geometry via Feature Alignment} \citep{corpus154} & M/F \\
155 & S4 & \href{https://openreview.net/forum?id=WiNCsgDP25}{LonGaussian: Tracking Foundation Priors for Long-Context Feedforward Gaussian Reconstruction} \citep{corpus155} & M/F \\
156 & S4 & \href{https://openreview.net/forum?id=e8KzLssuLH}{PAIR3D: Learning Physics-Aligned and Visually Consistent 3D Scene Layouts from Verifiable Rewards} \citep{corpus156} & M/F \\
157 & S4 & \href{https://openreview.net/forum?id=0X2ay3NW4Z}{PCAsplat: Gaussian Splatting with Local PCA Regularization} \citep{corpus157} & M/F \\
158 & S4 & \href{https://openreview.net/forum?id=Y69S27nWgp}{SM4RT: Learning Structured Motion Geometry for 4D Reconstruction} \citep{corpus158} & M/F \\
159 & S4 & \href{https://openreview.net/forum?id=dpOjIHiDKY}{UniForge3D: Joint 3D Reconstruction and Generation in a Shared Anchor-Guided Space} \citep{corpus159} & M/F \\
160 & S4 & \href{https://openreview.net/forum?id=X7FI3N35Hq}{VASC: Value-Aware Sparse Attention with Cross-layer Memory for Efficient 3D Reconstruction} \citep{corpus160} & M/F \\
161 & Style & \href{https://openreview.net/forum?id=hFRPKRZXeS}{$\mathrm{E}^2$IR-Agent: Experience-Driven Self-Evolving Image Restoration Agentic System} \citep{corpus161} & F \\
162 & Style & \href{https://openreview.net/forum?id=gGGjmjPJi4}{A Simulation-Grounded VLM Framework for Agentic Wildfire Monitoring and Reporting} \citep{corpus162} & F \\
163 & Style & \href{https://openreview.net/forum?id=uLkUAjKWe6}{AccessWorld: A Benchmark for Computer Use Agents Operating Assistive Technologies} \citep{corpus163} & F \\
164 & Style & \href{https://openreview.net/forum?id=O7Vvd7kb57}{Agent Skills Under Change: A Lifecycle Benchmark and Coordinate-Guided Updates} \citep{corpus164} & M/F \\
165 & Style & \href{https://openreview.net/forum?id=RPFKIcqk1w}{An Agentic AI Scientific Community for Automated Neural Operator Discovery} \citep{corpus165} & F \\
166 & Style & \href{https://openreview.net/forum?id=0ZTqgrmYaQ}{Beyond Dyadic Memory: Interaction-Aware Multimodal Memory with Adaptive Agentic Retrieval for Multi-Party Spoken Conversations} \citep{corpus166} & F \\
167 & Style & \href{https://openreview.net/forum?id=whcg806KGZ}{BKBAgent: A Battery Parameter Identification Agent with Bayesian Experimental Design and Knowledge Graph} \citep{corpus167} & M/F \\
168 & Style & \href{https://openreview.net/forum?id=skorOYrYml}{BodyBound: A Benchmark for Diagnosing Robot-Policy Responses to Body Changes Beyond Task Success} \citep{corpus168} & F \\
169 & Style & \href{https://openreview.net/forum?id=YLI8SRurGG}{BrainIPAgent: Multi-Agent Interpretable Structural Priors for Multimodal Brain Network Learning} \citep{corpus169} & F \\
170 & Style & \href{https://openreview.net/forum?id=yLEH2uLAgc}{CADForge: Agentic Single-View CAD Reconstruction with Explicit Geometry Reasoning} \citep{corpus170} & F \\
171 & Style & \href{https://openreview.net/forum?id=zploFP1iVU}{CUAWright: A Minimal Unified Interface for Computer Use Agents} \citep{corpus171} & F \\
172 & Style & \href{https://openreview.net/forum?id=djznyPY9n1}{Don't Ask Agents to Do Everything: Selective Agentic Reasoning for Single Image to 3D Scene Generation} \citep{corpus172} & F \\
173 & Style & \href{https://openreview.net/forum?id=FqbDu6m3j2}{DUDA-Bench: Benchmarking LLM Agents on Multimodal Data-Driven Urban Diagnosis} \citep{corpus173} & F \\
174 & Style & \href{https://openreview.net/forum?id=Qask3lw679}{EAGLE: Grounding Multimodal LLM Judges for AI-Generated Image Assessment via Evidence Anchors} \citep{corpus174} & M/F \\
175 & Style & \href{https://openreview.net/forum?id=vowOuAXSIO}{EmboScope: A Multi-Source Comprehensive Benchmark for Embodied Perception, Understanding, and Planning} \citep{corpus175} & F \\
176 & Style & \href{https://openreview.net/forum?id=4kPwNYWJYm}{EngiWorld: What Can Frontier Agents Deliver in Professional Engineering Environments?} \citep{corpus176} & F \\
177 & Style & \href{https://openreview.net/forum?id=8ztJtapkEV}{FinFIRST: Benchmarking Search Agents for Financial Information Retrieval, Sourcing and Traceability} \citep{corpus177} & F \\
178 & Style & \href{https://openreview.net/forum?id=ZEeYfne3aD}{FM-Bench: A Benchmark for Long-Horizon Management with Competing Agents} \citep{corpus178} & F \\
179 & Style & \href{https://openreview.net/forum?id=X31pnGGh8D}{Frameworkers: A Dynamic Multi-Agent Framework for AI Video Production} \citep{corpus179} & F \\
180 & Style & \href{https://openreview.net/forum?id=12nS7mYinw}{GATOR: Generative and Agentic 3D Object Reconstruction from Casual Images} \citep{corpus180} & F \\
181 & Style & \href{https://openreview.net/forum?id=3gogFGspDb}{GRASP: Grounded Restoration with Agent-Refined Semantic Prompts for Real-World Image Super-Resolution} \citep{corpus181} & M/F \\
182 & Style & \href{https://openreview.net/forum?id=cR0St8qS2n}{Harmony Harness: Multi-Objective Self-Evolution of Agent Harnesses} \citep{corpus182} & F \\
183 & Style & \href{https://openreview.net/forum?id=olW2A6zx75}{LEGO-Anything: Coding Agents for 3D Scene Reconstruction} \citep{corpus183} & F \\
184 & Style & \href{https://openreview.net/forum?id=C4VFESNXLV}{Look Back to Move Forward: Learning to Revisit for Agentic Compositional Image Generation} \citep{corpus184} & M/F \\
185 & Style & \href{https://openreview.net/forum?id=02vIJnLoLo}{Maestro: Agentic Formal Proof Development by Coordination} \citep{corpus185} & F \\
186 & Style & \href{https://openreview.net/forum?id=7zleBvksX6}{MASCraft: Multi-Agent Specialization through Role--Skill Composition and Self-Evolving Orchestration} \citep{corpus186} & M/F \\
187 & Style & \href{https://openreview.net/forum?id=1Xi9fslMSG}{MatGym: An Agent-Centric Environment Bridging Materials Discovery and Reproducible Synthesis} \citep{corpus187} & F \\
188 & Style & \href{https://openreview.net/forum?id=Dli0qriFyd}{MemVista: Benchmarking Long-Term Memory for Multimodal Agents in Evolving Projects} \citep{corpus188} & F \\
189 & Style & \href{https://openreview.net/forum?id=EkkRujJohS}{OmniClinic: Evaluating Active Evidence Acquisition in a Multimodal Clinical Sandbox} \citep{corpus189} & F \\
190 & Style & \href{https://openreview.net/forum?id=jVkkfpmnAw}{Ontology-Harnessing Multi-Agent System for Abstractive Visual Understanding} \citep{corpus190} & M/F \\
191 & Style & \href{https://openreview.net/forum?id=I6Gx5US2sY}{OpenAgentFlow: Enabling System-Wide Safety Boundaries for Heterogeneous AI Agent Fleets} \citep{corpus191} & M/F \\
192 & Style & \href{https://openreview.net/forum?id=VkVp7CjajL}{OPTScientist: Toward Recursive Self-Improving Agents via Automated Optimizer Discovery} \citep{corpus192} & F \\
193 & Style & \href{https://openreview.net/forum?id=xGIdDM5rYW}{SpatialAgent: Working Memory for Long-Horizon Spatial Reasoning} \citep{corpus193} & F \\
194 & Style & \href{https://openreview.net/forum?id=VA0yCrShah}{Sprout: Building Dynamic Memory While Reasoning for Agentic Video Understanding} \citep{corpus194} & M/F \\
195 & Style & \href{https://openreview.net/forum?id=qT1Thc0dst}{Stride: Scale-Aware Multi-Agent Evolutionary Refinement for Infographic Generation} \citep{corpus195} & F \\
196 & Style & \href{https://openreview.net/forum?id=eFi9w6vSHP}{TARSphere: A Robot-Oriented Full-Stack Framework for Long-Horizon Embodied Agents in Evolving Physical and Social Environments} \citep{corpus196} & M/F \\
197 & Style & \href{https://openreview.net/forum?id=VJ7YUdWOrj}{Topological Coherence for Self-Evolving Multi-Agent Systems} \citep{corpus197} & F \\
198 & Style & \href{https://openreview.net/forum?id=UPFHZRl39C}{VEvo-Harness: Causally-Guarded Self-Evolving Harness for Agentic Image Generation} \citep{corpus198} & M/F \\
199 & Style & \href{https://openreview.net/forum?id=ahcNtcQxsO}{VideoWeaver: Evaluating and Evolving Skills for Agentic Long Video Generation} \citep{corpus199} & F \\
200 & Style & \href{https://openreview.net/forum?id=MI8d8y9Hja}{When Should AI Chime In? Benchmarking Timely Contribution of Proactive Agents in Multi-Party Chat} \citep{corpus200} & F \\
\end{longtable}\endgroup

\end{document}